%% file: arxiv.tex
\documentclass[twoside]{article}

\usepackage[accepted]{aistats2025}
\usepackage[round]{natbib}
\renewcommand{\bibname}{References}
\renewcommand{\bibsection}{\subsubsection*{\bibname}}

\usepackage{lmodern}  %
\usepackage[utf8]{inputenc} %
\usepackage[T1]{fontenc}    %
\usepackage[colorlinks = True, linkcolor = blue, citecolor = violet!70!white]{hyperref}       %
\usepackage{url}            %
\usepackage{booktabs}       %
\usepackage{amsfonts}       %
\usepackage{nicefrac}       %
\usepackage{microtype}      %
\usepackage{xcolor}         %
\usepackage{algorithm}      %
\usepackage{algpseudocode}  %

\usepackage{etoc}
\usepackage{bookmark}

\usepackage{mathtools}
\usepackage{subcaption}
\usepackage{amssymb}

\hypersetup{
    colorlinks=true,    
    linktocpage=true,    
    pdfborder={0 0 0},   
}
\usepackage[capitalise,nameinlink]{cleveref}

\usepackage{bm}
\usepackage{amsmath}
\usepackage{amsthm}
\usepackage{booktabs}
\usepackage{multirow}
\usepackage{enumitem}
\usepackage{xr}
\usepackage{xspace}
\usepackage{wrapfig}

\usepackage{graphicx}
\usepackage{subcaption}
\usepackage{tikz}
\usetikzlibrary{fit}
\usepackage{pgfplots}
\pgfplotsset{compat=newest}
\usetikzlibrary{shapes.geometric} 
\usepgfplotslibrary{groupplots}
\usetikzlibrary{decorations.pathmorphing} %
\usetikzlibrary{positioning}
\usepackage{tcolorbox}

\algrenewcommand\algorithmicrequire{\textbf{Input:}}
\algrenewcommand\algorithmicensure{\textbf{Output:}}
\newcommand{\e}{\mathbf{e}}
\newcommand{\E}{\mathbf{E}}
\renewcommand{\r}{\mathbf{r}}

\newcommand{\vtheta}{{\bm{\theta}}}
\newcommand{\vxi}{\bm{\xi}}
\newcommand{\vpi}{\bm{\pi}}

\renewcommand{\ldots}{\mathinner{.\mkern1.5mu.\mkern1.5mu.}}

\newcommand{\x}{\mathbf{x}}
\newcommand{\xs}{\mathbf{x}^\star}
\newcommand{\y}{\mathbf{y}}

\newcommand{\z}{\mathbf{z}}
\newcommand{\X}{\mathbf{X}}

\newcommand{\data}{\mathcal{D}}
\newcommand{\dataplus}{\mathfrak{D}} %

\newcommand{\xt}{\mathbf{x}^\star}
\newcommand{\yt}{{y}^\star}
\newcommand{\zt}{{z}^\star}
\newcommand{\vxit}{\bm{\xi}^\star}

\newcommand{\XX}{\mathcal{X}}
\newcommand{\YY}{\mathcal{Y}}
\newcommand{\xopt}{\x_\text{opt}}
\newcommand{\yopt}{y_\text{opt}}

\newcommand{\ie}{\textit{i.e.}\@\xspace}
\newcommand{\eg}{\textit{e.g.}\@\xspace}
\newlength\figureheight
\newlength\figurewidth

\newcommand{\colorline}[1]{\protect\tikz[baseline=-0.5ex]\protect\draw[very thick,#1] (0,0) -- (5mm,0);}
\newcommand{\colorlinedashed}[1]{\protect\tikz[baseline=-0.5ex]\protect\draw[very thick,#1, dashed] (0,0) -- (5mm,0);}
\newcommand{\colorlinedotted}[1]{\protect\tikz[baseline=-0.5ex]\protect\draw[very thick,#1, dotted] (0,0) -- (5mm,0);}
\newcommand{\colorlinedotdash}[1]{\protect\tikz[baseline=-0.5ex]\protect\draw[very thick,#1, dash dot] (0,0) -- (5mm,0);}
\newcommand{\colordot}[1]{\protect\tikz[baseline=-0.5ex]\protect\draw[fill=#1, very thick] (0,0) circle (0.5mm);}
\newcommand{\colorlinedashedlight}[1]{\protect\tikz[baseline=-0.5ex]\protect\draw[ thick,#1, dashed] (0,0) -- (5mm,0);}

\newcommand{\legendACETS}{\colorline{blue} ACE-TS}
\newcommand{\legendACEPTS}{\colorlinedotdash{blue} ACEP-TS}
\newcommand{\legendpiBOTS}{\colorlinedotdash{color1} $\pi$BO-TS}
\newcommand{\legendACEMES}{\colorlinedashed{blue!54.5098039215686!black} ACE-MES}
\newcommand{\legendRandom}{\colorlinedotted{color0} Random}
\newcommand{\legendGPMES}{\colorlinedashed{color2} GP-MES}
\newcommand{\legendGPTS}{\colorline{color1} GP-TS}
\newcommand{\legendTNPDTS}{\colorline{color3} AR-TNPD-TS}

\newcommand{\legendTrueFunction}{\colorlinedashedlight{white!50.1960784313725!black} True function}
\newcommand{\legendObservations}{\colordot{black} Observations}
\newcommand{\legendpyopt}{\colorline{color1} $p(\yopt | \data_N)$}
\newcommand{\legendpxopt}
{\colorline{blue} $p(x_\text{opt} | \data_N)$}
\newcommand{\legendpygivenx}{\colorlinedotted{color4} $p(y| x, \data_N)$}
\newcommand{\legendyoptcond}{\colorlinedashed{orange} $\yopt$}

\definecolor{color0}{rgb}{0.968627450980392,0.505882352941176,0.749019607843137} %
\definecolor{color1}{rgb}{1,0.549019607843137,0} %
\definecolor{color2}{rgb}{1,0.647058823529412,0} %
\definecolor{color3}{rgb}{0.564705882352941,0.933333333333333,0.564705882352941} %
\definecolor{color4}{rgb}{0.501960784313725,0,0.501960784313725} %
\definecolor{cbAblue}{HTML}{1f78b4}
\definecolor{cbAgreen}{HTML}{33a02c}
\definecolor{cbBpurple}{HTML}{7b3294}
\definecolor{cbBgreen}{HTML}{008837}
\definecolor{latent2}{HTML}{1b9e77}

\colorlet{parametrized}{cbBpurple}

\colorlet{relational}{cbBgreen}

\newcommand{\ignore}[1]{}

\begin{document}
\addtocontents{toc}{\protect\setcounter{tocdepth}{0}}

\runningtitle{Amortized Probabilistic Conditioning for Optimization, Simulation and Inference}

\runningauthor{Chang$^{*}$, Loka$^{*}$, Huang$^{*}$, Remes, Kaski, Acerbi}

\twocolumn[
\aistatstitle{Amortized Probabilistic Conditioning \\ for Optimization, Simulation and Inference}

\aistatsauthor{Paul E. Chang$^{*1}$  Nasrulloh Loka$^{*1}$ \  Daolang Huang$^{*2}$  Ulpu Remes$^3$  \\ \textbf{Samuel Kaski}$^{2,4}$  \textbf{Luigi Acerbi}$^1$}

\aistatsaddress{$^1$Department of Computer Science, University of Helsinki, Helsinki, Finland 
\\  $^2$Department of Computer Science, Aalto University, Espoo, Finland \\ 
$^3$Department of Mathematics and Statistics, University of Helsinki, Helsinki, Finland 
\\
$^4$Department of Computer Science, University of Manchester, Manchester, United Kingdom
} ]

\begin{abstract}
Amortized meta-learning methods based on pre-training have propelled fields like natural language processing and vision. Transformer-based neural processes and their variants are leading models for probabilistic meta-learning with a tractable objective. Often trained on synthetic data, these models implicitly capture essential latent information in the data-generation process. However, existing methods do not allow users to flexibly \emph{inject} (condition on) and \emph{extract} (predict) this probabilistic latent information at runtime, which is key to many tasks.
We introduce the Amortized Conditioning Engine (ACE), a new transformer-based meta-learning model that explicitly represents latent variables of interest.
ACE affords conditioning on both observed data and interpretable latent variables, the inclusion of priors at runtime, and outputs predictive distributions for discrete and continuous data and latents. 
We show ACE's practical utility  across diverse tasks such as image completion and classification, Bayesian optimization, and simulation-based inference, demonstrating how a general conditioning framework can replace task-specific solutions.\looseness=-1
\end{abstract}

\section{INTRODUCTION}

\begin{figure}[h!]
    \centering
\begin{tikzpicture}

        \node[anchor=south west] (image1) at (-4.3, -0.8) {\includegraphics[width=1.5cm]{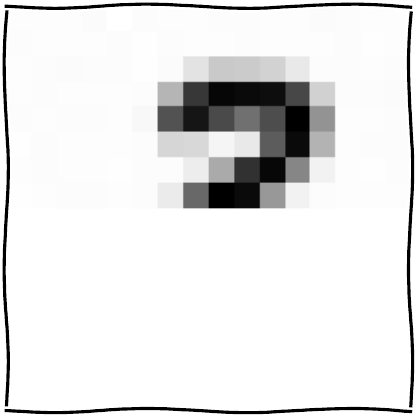}};
        \node[anchor=south west] (image3) at (-0.1, -1.6) {\includegraphics[width=1.5cm]{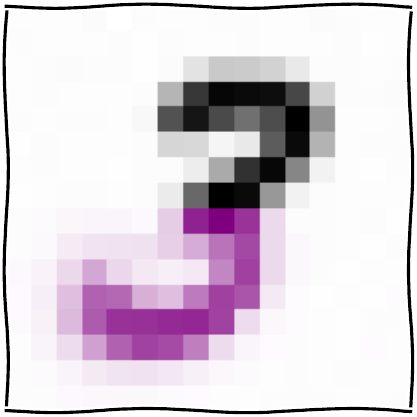}};

        \node[anchor=south west, right=0cm of image1, yshift=0.80cm] (xkcd1) {\includegraphics[width=1.5cm]{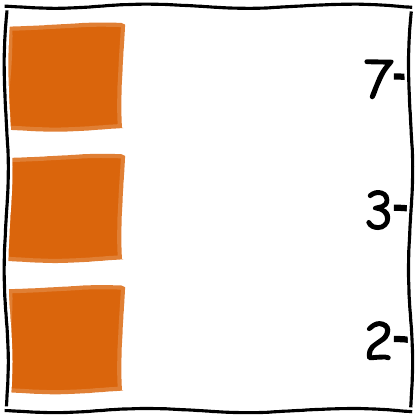}};
        \node[anchor=south west, right=0cm of image1, yshift=-0.80cm] (xkcd2)  {\includegraphics[width=1.5cm]{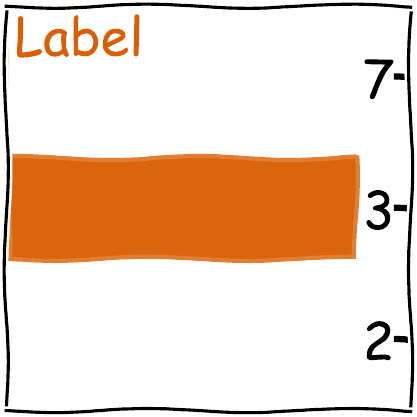}};

        \node[anchor=south west, right=0.70cm of xkcd1] (image2) {\includegraphics[width=1.5cm]{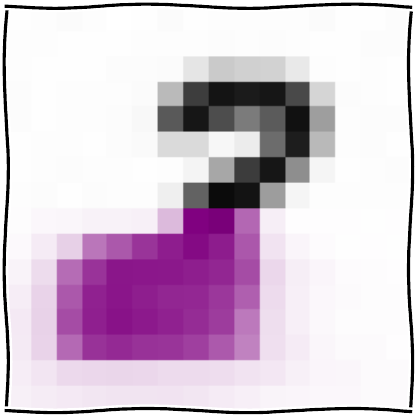}};

        \node[anchor=south west, right=0cm of image2] (xkcd3) {\includegraphics[width=1.5cm]{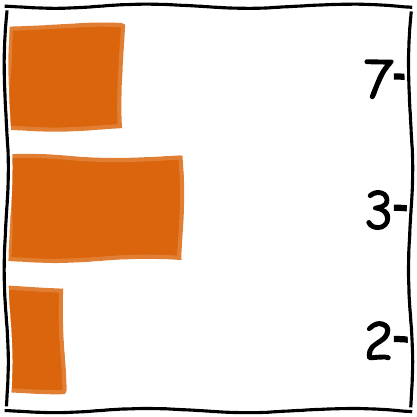}};
        \node[anchor=south west,right=0cm of image3] (xkcd4) {\includegraphics[width=1.5cm]{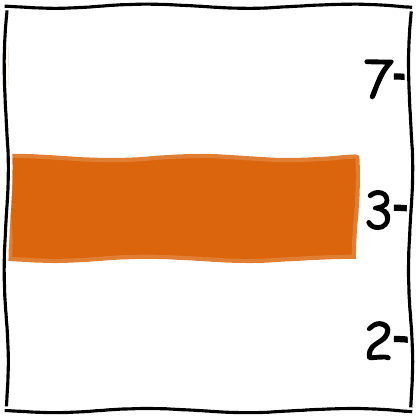}};

        \node[anchor=south west] (image4) at (-4.3, -3.5) {\includegraphics[width=1.5cm]{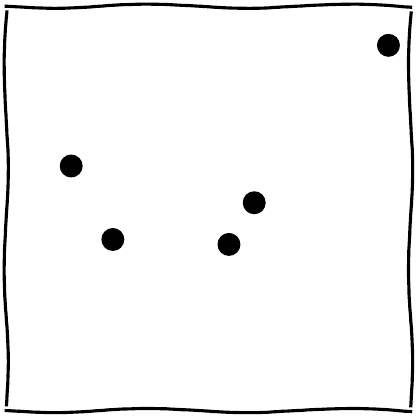}};
        \node[anchor=south west, right=0cm of image4] (image5)  {\includegraphics[width=1.5cm]{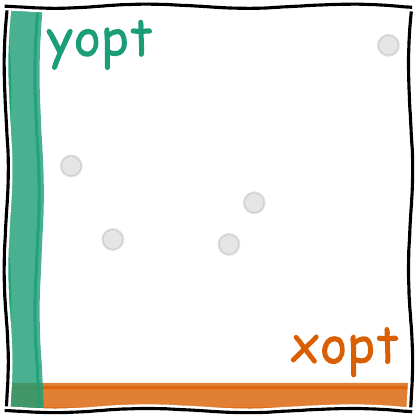}};

        \node[anchor=south west] (image6) at (-0.1, -3.5) {\includegraphics[width=1.5cm]{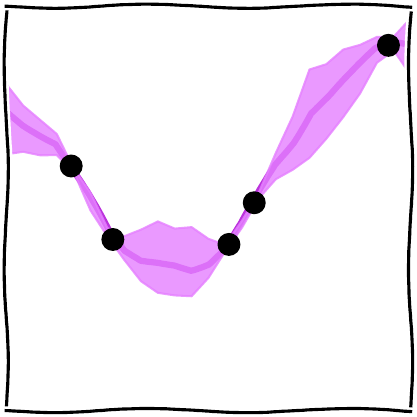}};
        \node[anchor=south west, right=0cm of image6] (image7) {\includegraphics[width=1.5cm]{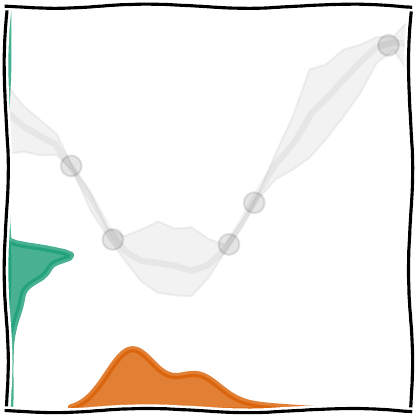}};

        \node[anchor=south west] (sbi_1) at (-4.3, -5.4) {\includegraphics[width=1.5cm]{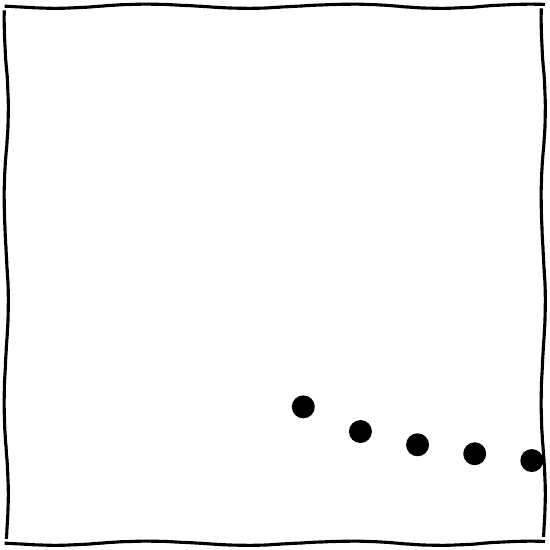}};
        \node[anchor=south west, right=0cm of sbi_1] (sbi_2) {\includegraphics[width=1.5cm]{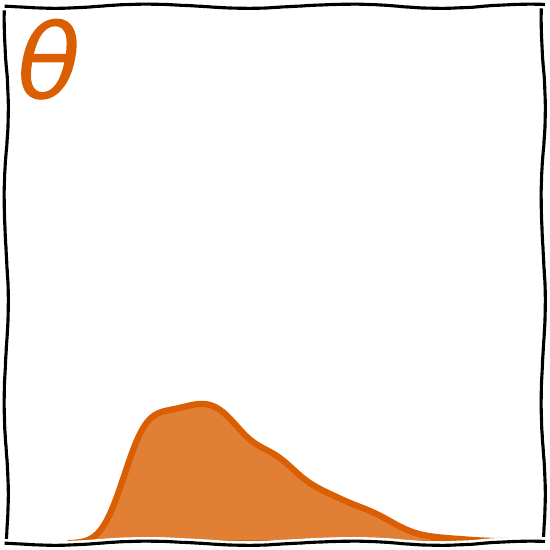}};

        \node[anchor=south west] (sbi_3) at (-0.1, -5.4) {\includegraphics[width=1.5cm]{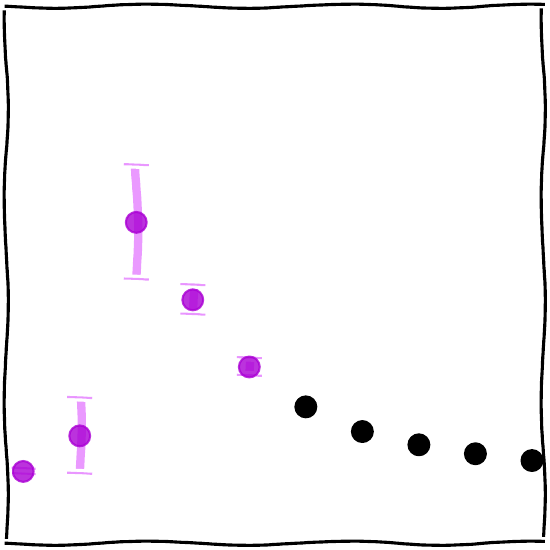}};
        \node[anchor=south west, right=0cm of sbi_3] (sbi_4) {\includegraphics[width=1.5cm]{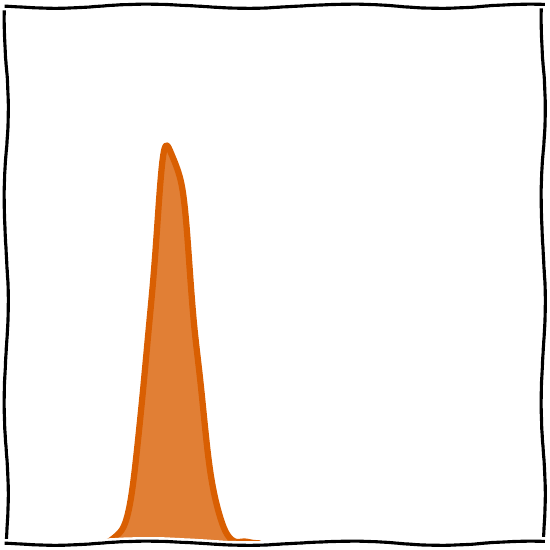}};

        \draw[->, line width=0.7mm, black] (xkcd1.east) -- (image2.west);
        \draw[->, line width=0.7mm, black] (xkcd2.east) -- (image3.west);

        \draw[->, line width=0.7mm, black] (image5.east) -- (image6.west);
        \draw[->, line width=0.7mm, black] (sbi_2.east) -- (sbi_3.west);

        \draw[dashed, line width=0.7mm, gray] (-4.3, -1.7) -- (3.0, -1.7);  
        \draw[dashed, line width=0.7mm, gray] (-4.3, -3.6) -- (3.0, -3.6);  %

        \node[anchor=south west] at (-5.0, -0.2) {\textbf{(a)}};
        \node[anchor=south west] at (-5.0, -2.9) {\textbf{(b)}};
        \node[anchor=south west] at (-5.0, -4.8) {\textbf{(c)}};

        \node[anchor=south east] (prob_cond)  at (-0.1, 2.4) {\textbf{Probabilistic Conditioning}};
          
        \node[anchor=south east] (data)  at (-3.0, 1.9) {\textbf{\small Data}};
        
         \node[anchor=south east,, right=0.65cm of data] (latent)   {\textbf{\small Latent}};

        \node[anchor=south west, right=0.80cm of prob_cond] {\textbf{Predictions}};
        
        \node[anchor=south east] (data2)  at (1.25, 1.9) {\textbf{\small Data}};
        \node[anchor=south east,, right=0.65cm of data2] (latent2)   {\textbf{\small Latent}};
        
    \end{tikzpicture}

    \caption{\textbf{Probabilistic conditioning and prediction.} Many tasks reduce to \emph{probabilistic conditioning} on data and key latent variables (left) and then \emph{predicting} data and latents (right). (\textcolor{blue}{a}) Image completion and classification (data: pixels; latents: classes). \emph{Top}: Class prediction. \emph{Bottom}: Conditional generation. (\textcolor{blue}{b}) Bayesian optimization (data: function values; latents: optimum location $x_\text{opt}$ and value $y_\text{opt}$). We predict both the function values and $x_\text{opt}$, $y_\text{opt}$ given function observations and a prior over $x_\text{opt}$, $y_\text{opt}$ (here flat). (\textcolor{blue}{c}) Simulator-based inference (data: observations; latents: model parameter $\theta$). Given data and a prior over $\theta$, we can compute both the posterior over $\theta$ and predictive distribution over unseen data.
    Our method fully amortizes probabilistic conditioning and prediction.}
    \label{fig:intro}
    \vspace{-0.3cm}
\end{figure}

Amortization, or pre-training, is a crucial technique for improving computational efficiency and generalization across many machine learning tasks, from regression \citep{garnelo2018neural} to optimization \citep{amos2022tutorial} and simulation-based inference \citep{cranmer2020frontier}. By training a deep neural network on a large dataset of related problems and solutions, amortization can achieve both fast inference time, solving problems with a single forward pass, and meta-learning, better adapting to new problems by capturing high-level statistical relations \citep{brown2020language}. 
Probabilistic meta-learning models based on the transformer architecture \citep{vaswani2017attention} are the state-of-the-art for amortizing complex predictive data distributions \citep{nguyen2022transformer, muller2022transformers}. 

\begin{tcolorbox}[colback=red!5!white,colframe=red!75!black]
This paper capitalizes on the fact that many machine learning problems reduce to \emph{predicting} data and task-relevant latent variables after \emph{conditioning} on other data and latents \citep{ghahramani2015probabilistic}; see \cref{fig:intro}. 
Moreover, in many scenarios the user has exact or probabilistic information (priors) about task-relevant variables that they would like to leverage, but incorporating such prior knowledge is challenging, requiring dedicated, expensive solutions.
\end{tcolorbox}

For instance, in Bayesian optimization \citep{garnett2023bayesian}, the goal is to find the location $\xopt$ and value $\yopt$ of the global minimum of a function (\cref{fig:intro}b). These are \emph{latent variables}, distinct from the data $\data_N = \{(\x_1, y_1), \ldots, (\x_N, y_N)\}$ consisting of observed function location and values. Following information-theoretical principles, we should query points that would reduce uncertainty (entropy) about the latent optimum's location or value.
However, predictive distributions over $\xopt$ and $\yopt$ are intractable, leading to complex approximation techniques \citep{hennig2012entropy, hernandez2014predictive, wang2017max}.
Another case of interest is when prior information is available, such as knowing $\yopt$ due to the problem formulation (e.g., $\yopt = 0$ for some theoretical reason), or expert knowledge of more likely locations for $\xopt$ -- but injecting such information in current methods is highly nontrivial~\citep{nguyen2020knowing, souza2021bayesian,hvarfner2022pi}. Crucially, if we had access to $p(\xopt, \yopt | \data_N)$, and we could likewise condition on $\xopt$ or $\yopt$ or set priors over them, many challenging tasks that so far have required dedicated heuristics or computationally expensive solutions would become straightforward.
Similar challenges extend to many machine learning tasks, including regression and classification (\cref{fig:intro}a), and simulation-based inference (\cref{fig:intro}c),  all involving predicting, sampling, and probabilistic conditioning on either exact values or distributions (priors) at runtime.

In this work, we address the desiderata above by introducing the \emph{Amortized Conditioning Engine} (ACE), a general amortization framework which extends transformer-based meta-learning architectures \citep{nguyen2022transformer, muller2022transformers} with explicit and flexible probabilistic modeling of task-relevant latent variables.
Our main goal with ACE is to develop a method capable of addressing a variety of tasks that would otherwise require bespoke solutions and approximations. Through the lens of amortized probabilistic conditioning and prediction, we provide a unifying methodological bridge across multiple fields.

\vspace{-0.25em}
\paragraph{Contributions.} Our contributions include: 
\begin{itemize}[nosep]
\item We propose ACE, a transformer-based architecture that simultaneously affords, at inference time, conditioning and autoregressive probabilistic prediction for arbitrary combinations of data and latent variables, both continuous and discrete.
\item We introduce a new technique for allowing the user to provide probabilistic information (priors) over each latent variable at inference time. %
\item We substantiate the generality of our framework through a series of tasks from different fields, including image completion and classification, Bayesian optimization, and simulation-based inference, on both synthetic and real data.
\end{itemize}

\begin{tcolorbox}[colback=red!5!white,colframe=red!75!black]
ACE requires availability of predefined, interpretable latent variables during training (e.g., $\xopt$, $\yopt$). For many tasks, this can be achieved through explicit construction of the generative model, as shown in \cref{sec:experiments}.
\end{tcolorbox}

\section{PRELIMINARIES}
\label{sec:preliminaries}

\vspace{-0.25em}
In this section, we review previous work on transformer-based probabilistic meta-learning models within the framework of prediction maps \citep{foong2020meta, markou2022practical} and Conditional Neural Processes (CNPs; \citealp{garnelo2018conditional}).
We denote with $\x \in \XX \subseteq \mathbb{R}^{D}$ input vectors (covariates) and $y \in \YY \subseteq \mathbb{R}$ scalar output vectors (values). %
\cref{tab:acronyms} in \cref{app:notation} summarizes key acronyms used in the paper.

\vspace{-0.25em}
\paragraph{Prediction maps.}

A \emph{prediction map} $\pi$ is a function that maps (1) a \emph{context set} of input/output pairs $\data_N = \{(\x_1, y_1), \ldots, (\x_N, y_N)\}$ and (2) a collection of \emph{target inputs} $\xt_{1:M} \equiv (\xt_1, \ldots, \xt_M)$ to a distribution over the corresponding \emph{target outputs} $\yt_{1:M} \equiv (\yt_{1}, \ldots, \yt_M)$: 
\begin{equation} 
\label{eq:predictionmap}
 \pi\left(\yt_{1:M} \vert \xt_{1:M}; \data_N \right) = %
p\left(\yt_{1:M}| \r(\xt_{1:M}, \data_N) \right), 
\end{equation}
where $\r$ is a \emph{representation vector} of the context and target sets that parameterizes the predictive distribution. 
Such map should be invariant with respect to permutations of the context set and, separately, of the targets \citep{foong2020meta}. 
The Bayesian posterior is a prediction map, with the Gaussian Process (GP; \citealp{rasmussen2006gaussian}) posterior a special case.

\vspace{-0.25em}
\paragraph{Diagonal prediction maps.}

We call a prediction map \emph{diagonal} if it represents and predicts each target independently:
\begin{equation}
\label{eq:diagonal_map}
 \pi(\yt_{1:M} \vert \xt_{1:M}; \data_N ) =
 \prod_{m=1}^M p\left(\yt_m| \r(\xt_m, \r_{\data}(\data_N))\right)
\end{equation}
where $\r_{\data}$ denotes a representation of the context alone. Diagonal outputs ensure that a prediction map is permutation and marginalization consistent with respect to the targets for a fixed context, necessary conditions for a valid stochastic process \citep{markou2022practical}.
Importantly, while diagonal prediction maps directly model conditional 1D marginals, they can represent \emph{any} conditional joint distribution autoregressively~\citep{bruinsma2023autoregressive}.
CNPs are diagonal prediction maps parameterized by deep neural networks \citep{garnelo2018conditional}. CNPs encode the context set to a \emph{fixed-dimension} vector $\r_{\data}$ via a DeepSet  \citep{zaheer2017deep} to ensure permutation invariance of the context, and each target predictive distribution is a Gaussian whose mean and variance are decoded by a multi-layer perceptron (MLP).
Given the likelihood in \cref{eq:diagonal_map}, CNPs are easily trainable via maximum-likelihood optimization of parameters of encoder and decoder networks by sampling batches of context and target sets.

\vspace{-0.25em}
\paragraph{Transformers.}
Transformers \citep{vaswani2017attention} are deep neural networks based on the attention mechanism, which computes a weighted combination of hidden states of dimension $D_\text{emb}$ through three learnable linear projections: query ($Q$), key ($K$), and value ($V$). The attention operation, $\text{softmax}(QK^T / \sqrt{D_\text{emb}})V$, captures complex relationships between inputs. \emph{Self-attention} computes $Q$, $K$, and $V$ from the same input set, whereas \emph{cross-attention} uses two sets, one for computing the queries and another for keys and values.
A standard transformer architecture consists of multiple stacked self-attention and MLP layers with residual connections and layer normalization \citep{vaswani2017attention}. Without specifically injecting positional information, transformers process inputs in a permutation-equivariant manner. 

\vspace{-0.25em}
\paragraph{Transformer diagonal prediction maps.} We define here a general \emph{transformer prediction map} model family, focusing on its \emph{diagonal} variant (TPM-D), which includes the TNP-D model from \citet{nguyen2022transformer} and \emph{prior-fitted networks} (PFNs; \citealp{muller2022transformers}). TPM-Ds are not strictly CNPs because the context set is encoded by a \emph{variable-size} representation, but they otherwise share many similarities.
In a TPM-D, context data $(\x_n, y_n)_{n=1}^N$ and target inputs $(\xt_m)_{m=1}^M$ are first individually mapped to vector embeddings of size $D_\text{emb}$ via an embedder $f_\text{emb}$, often a linear map or an MLP. The embedded context points are processed together via a series of $B-1$ transformer layers implementing \emph{self-attention} within the context set. We denote by $\E^{(b)} = (\e_1^{(b)}, \ldots, \e_N^{(b)})$ the matrix of output embeddings of the $b$-th transformer layer, with $b=0$ the embedding layer. The encoded context representation is the stacked output of all layers, i.e. $\r_{\data} = (\E^{(0)}, \ldots, \E^{(B-1)})$, whose size is \emph{linear} in the context size $N$.
The decoder is represented by a series of $B$ transformer layers that apply \emph{cross-attention} from the embedded target points to the context set layer-wise, with the $b$-th target transformer layer attending the output $\E^{(b-1)}$ of the previous context transformer layer. The decoder transformer layers operate \emph{in parallel} on each target point. The $M$ outputs of the $B$-th decoder block %
are fed in parallel to an output head yielding the predictive distribution, \cref{eq:diagonal_map}. This shows that indeed TPM-Ds are diagonal prediction maps. The predictive distribution is a single Gaussian in TNP-D \citep{nguyen2022transformer} and a `Riemannian distribution' (a mixture of uniform distributions with fixed bin edges and half-Gaussian tails on the sides) in PFNs \citep{muller2022transformers}.
While in TPM-Ds encoding is mathematically decoupled from decoding, in practice encoding and decoding are commonly implemented in parallel within a single transformer layer via masking \citep{nguyen2022transformer, muller2022transformers}. 

\begin{figure}[t!]
    \centering
    \begin{subfigure}[t]{0.258\linewidth}
        \centering
        \includegraphics[width=\linewidth]{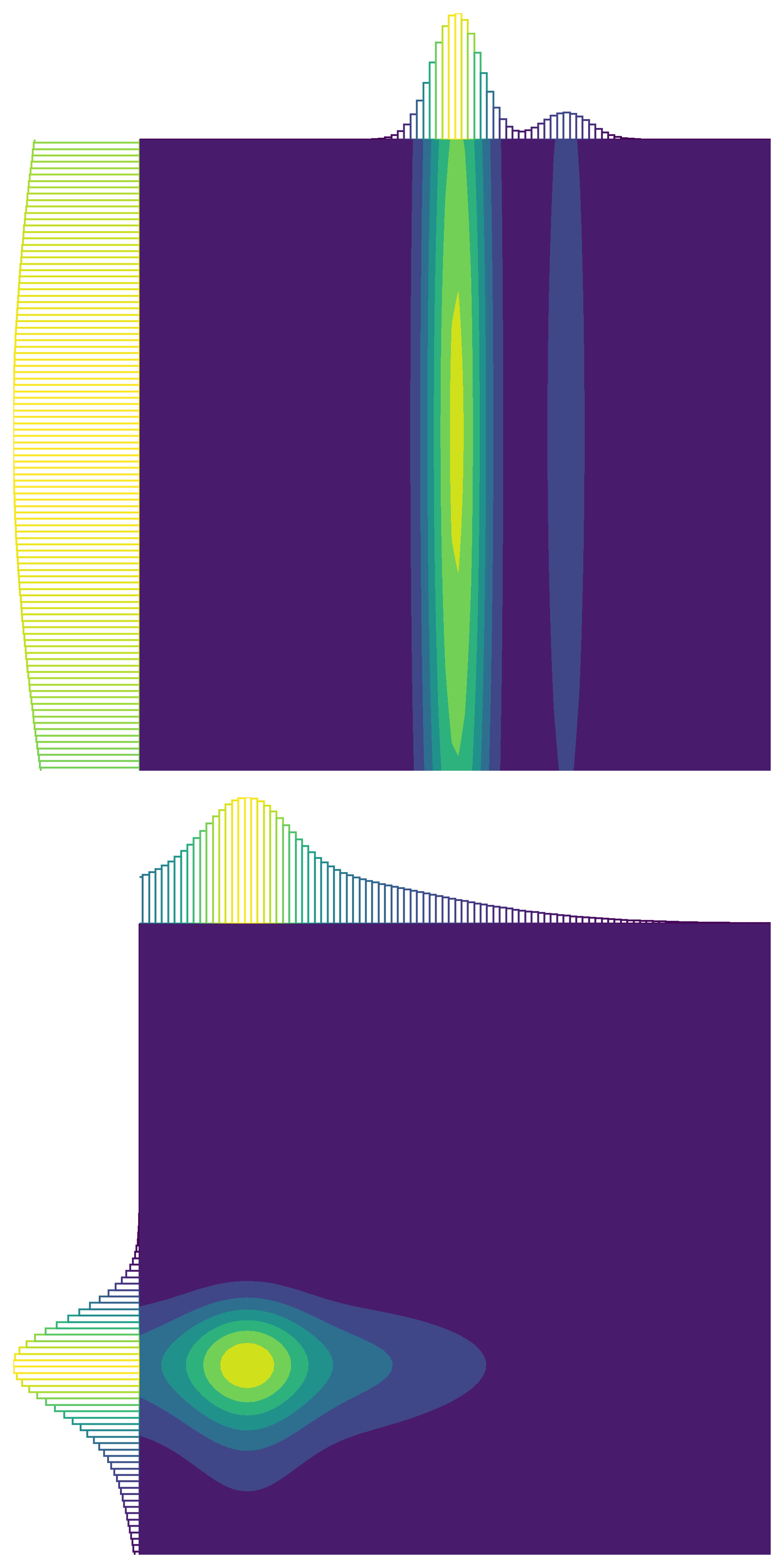}
        \caption{}
        \label{fig:prior_distribution}
        \begin{tikzpicture}[remember picture, overlay]
            \node at (0, 0.76) {$\mu$};
            \node at (-0.9, 2.0) {\textcolor{gray}{$\sigma$}};
            \node at (-0.9, 4.0) {\textcolor{gray}{$\sigma$}};
        \end{tikzpicture}
    \end{subfigure}
    \hfill
    \begin{subfigure}[t]{0.215\linewidth}
        \centering
        \includegraphics[width=\linewidth]{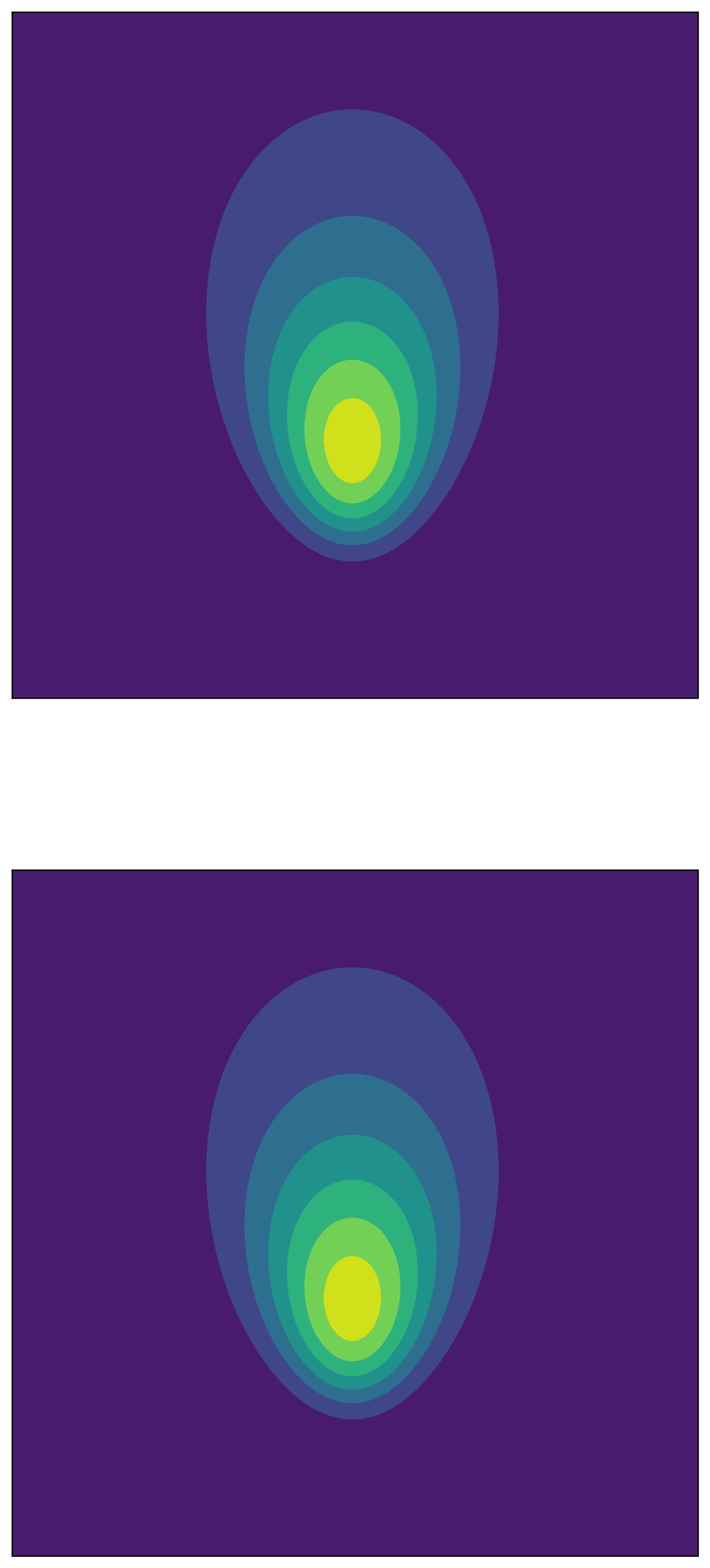}
        \caption{}
        \label{fig:likelihood}
        \begin{tikzpicture}[remember picture, overlay]
            \node at (0, 0.76) {$\mu$};  %
        \end{tikzpicture}
    \end{subfigure}
    \hfill
    \begin{subfigure}[t]{0.215\linewidth}
        \centering
        \includegraphics[width=\linewidth]{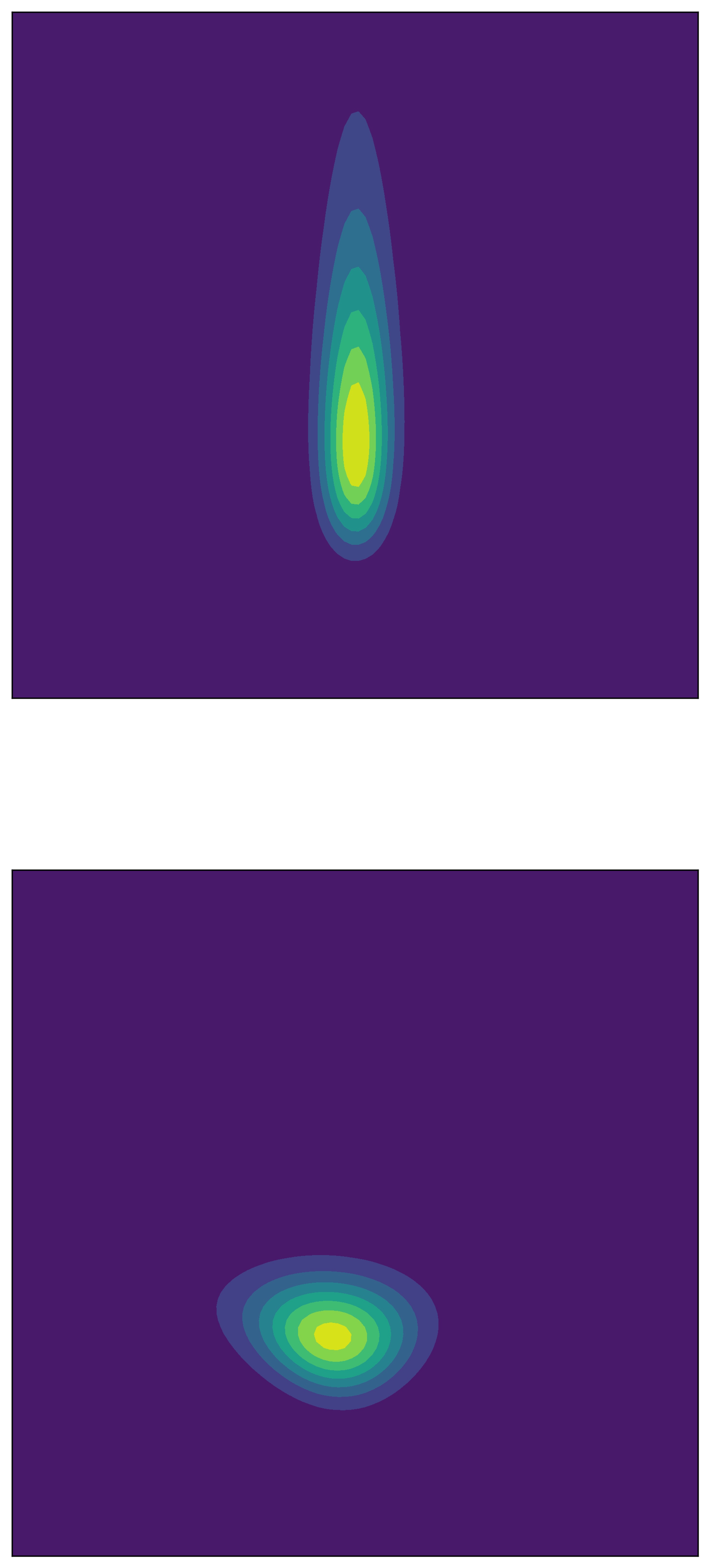}
        \caption{}
        \label{fig:true_posterior}
        \begin{tikzpicture}[remember picture, overlay]
            \node at (0, 0.76) {$\mu$};
        \end{tikzpicture}
    \end{subfigure}
    \hfill
    \begin{subfigure}[t]{0.215\linewidth}
        \centering
        \includegraphics[width=\linewidth]{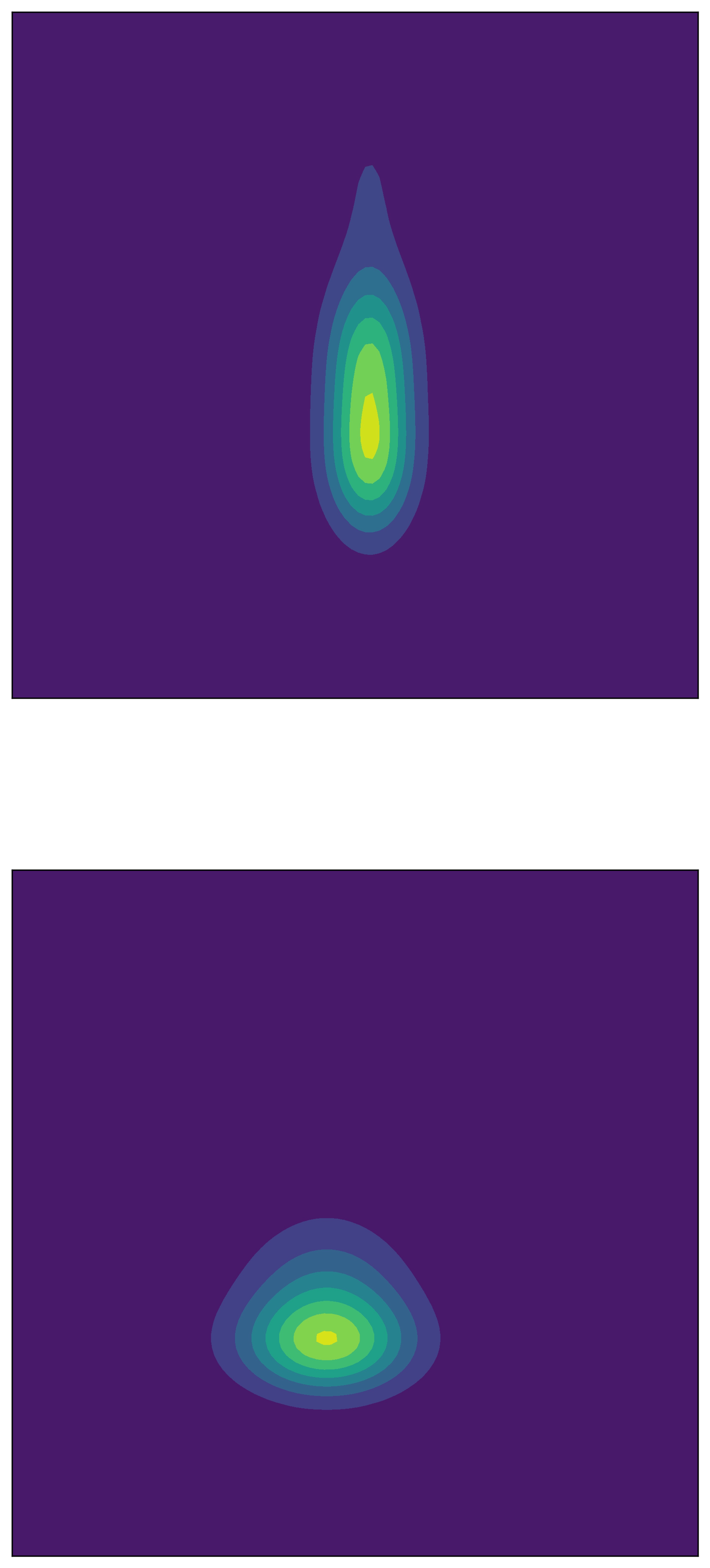}
        \caption{}
        \label{fig:ace_predicted_posterior}
        \begin{tikzpicture}[remember picture, overlay]
            \node at (0, 0.76) {$\mu$};
        \end{tikzpicture}
    \end{subfigure}
    \vspace{-0.5cm}
    \caption{\textbf{Prior amortization.} Two example posterior distributions for the mean $\mu$ and standard deviation $\sigma$ of a 1D Gaussian. (\subref{fig:prior_distribution}) Prior distribution over $\vtheta = (\mu, \sigma)$ set at runtime. (\subref{fig:likelihood}) Likelihood for the observed data. (\subref{fig:true_posterior}) Ground-truth Bayesian posterior. (\subref{fig:ace_predicted_posterior}) ACE's predicted posterior approximates well the true posterior.}
    \label{fig:gaussian_examples_main}
    \vspace{-0.55cm}
\end{figure}

\section{AMORTIZED CONDITIONING ENGINE}
\label{sec:ace}

We describe now our proposed Amortized Conditioning Engine (ACE) architecture, which affords arbitrary probabilistic conditioning and predictions. We assume the problem has $ L$ task-relevant latent variables of interest $ \vtheta = (\theta_1, \ldots, \theta_L) $. ACE amortizes arbitrary conditioning over  latents (in context) and data to predict arbitrary combinations of latents (in target)  and data.
ACE also amortizes conditioning on probabilistic information about unobserved latents, represented by an approximate prior distribution $p(\theta_l)$ for $l \in \{1, \ldots, L\}$; see  \cref{fig:gaussian_examples_main} for an example (details in \cref{app:priors}).

\subsection{ACE encodes latents and priors}
\label{sec:latentspriors}

We demonstrate here that ACE is a new member of the TPM-D family, by extending the prediction map formalism to explicitly accommodate latent variables.
In ACE, we aim to seamlessly manipulate variables that could be either data points $(\x, y)$ or latent variables $\theta_l$, for a finite set of continuous or discrete-valued latents $1 \le l\le L$.
We redefine inputs as $\vxi \in  \XX \cup \{\ell_1, \ldots, \ell_L \}$ where $\XX \subseteq \mathbb{R}^D$ denotes the data input space (covariates) and $\ell_l$ is a marker for the $l$-th latent. We also redefine the values as $z \in \mathcal{Z} \subseteq \mathbb{R}$ where $\mathcal{Z}$ can be continuous or a finite set of integers for discrete-valued output.
Thus,  $(\vxi, z)$ could denote either a  (input, output) data pair or a (index, value) latent pair with either continuous or discrete values. With these new flexible definitions, ACE is indeed a transformer diagonal prediction map (TPM-D). In particular, we can predict any combination of target variables (data or latents) conditioning on any other combination of context data and latents, $\dataplus_N = \left\{(\vxi_1, z_1), \ldots, (\vxi_N, z_N)\right\}$:
\begin{equation} 
\label{eq:ace_map}
 \pi(\zt_{1:M} \vert \vxit_{1:M}; \dataplus_N ) = 
 \prod_{m=1}^M p\left(\zt_m| \r(\vxit_m, \r_{\data}(\dataplus_N))\right).
\end{equation}

\paragraph{Prior encoding.} \label{sec:prior} ACE also allows the user to express probabilistic information over latent variables as prior probability distributions at runtime. Our method affords prior specification separately for each latent, corresponding to a factorized prior $p(\vtheta) = \prod_{l=1}^L p(\theta_l)$. To flexibly approximate a broad class of distributions, we convert each one-dimensional probability density function $p(\theta_l)$ to a normalized histogram of probabilities $\mathbf{p}_l \in [0, 1]^{N_\text{grid}}$ over a predefined grid $\mathcal{G}$ of $N_\text{bins}$ bins uniformly covering the range of values. We can represent this probabilistic conditioning information within the prediction map formalism by extending the context output representation to $z \in \left\{ \mathcal{Z} \cup  [0,1]^{N_\text{bins}} \right\}$, meaning that a context point either takes a specific value or a prior defined on $\mathcal{G}$ (see \cref{app:priors}).

\subsection{ACE architecture}

We detail below how ACE extends the general TPM-D architecture presented in \cref{sec:preliminaries} to implement latent and prior encoding, enabling flexible probabilistic conditioning and prediction. We introduce a novel  embedding layer for latents and priors,  adopt an efficient transformer layer implementation, and provide an output represented by a flexible Gaussian mixture or categorical distribution. See \cref{app:architecture} for an illustration and comparison to the TNP-D architecture.

\vspace{-0.25em}
\paragraph{Embedding layer.}

In ACE, the embedders map context and target data points and latents to the same embedding space of dimension $D_\text{emb}$. The ACE embedders handle discrete and continuous inputs without the need of tokenization.
For the context set, we embed an observed data point $(\x_n, y_n)$ as $f_\x(\x_n) + f_\text{val}(y_n) + \e_\text{data}$, while we embed an observed latent variable $\theta_l$ as $f_\text{val}(\theta_l) + \e_l$, where $\e_\text{data}$ and $\e_l$ for $1\le l \le L$ are learnt vector embeddings, and $f_\x$ and $f_\text{val}$ are learnt nonlinear embedders (MLPs) for the covariates and values, respectively.
For discrete-valued variables (data or latents), $f_\text{val}$ is replaced by a vector embedding matrix $\E_\text{val}$ with a separate row for each discrete value.
Latent variables with a prior $\mathbf{p}_l$ are mapped to the context set as $f_\text{prob}(\mathbf{p}_l) + \e_l$, where $f_\text{prob}$ is a learnt MLP.
In the target set, the \emph{value} of a variable is unknown and needs to be predicted, so we replace the value embedders above with a learnt `unknown' embedding $\e_?$, i.e. $f_\x(\x_n) + \e_? + \e_\text{data}$ for data and $\e_? + \e_l$ for latents.

\vspace{-0.25em}
\paragraph{Transformer layers.}

The embedding layer is followed by $B$ stacked transformer layers, each with a multi-head attention block followed by a MLP~\citep{vaswani2017attention}. Both the attention and MLP blocks are followed by a normalization layer and include skip connections. The attention block combines encoder and decoder in the same step, with self-attention on the context points (encoding) and cross-attention from the target points to the context (decoding). Computing a single masked context + target attention matrix would incur $O((N + M)^2)$ cost \citep{nguyen2022transformer,muller2022transformers}. Instead, by separating the context self-attention and target cross-attention matrices we incur a $O(N^2 + NM)$ cost~\citep{feng2023latent}.

\vspace{-0.25em}
\paragraph{Output heads.} 
A prediction output head is applied in parallel to all target points after the last transformer layer. For a continuous-valued variable, the output head is a Gaussian mixture, consisting of $K$ MLPs that separately output the parameters of $K$ 1D Gaussians, i.e., `raw' weight, mean, standard deviation for each mixture component~\citep{uria2016neural}. A learnt global raw bias term ($3 \times K$ parameters) is added to each raw output, helping the network learn deviations from the global distribution of values. Then weights, means and standard deviations for each Gaussian are obtained through appropriate transformations (softmax, identity, and softplus, respectively). For discrete-valued variables, the output head is a MLP that outputs a softmax categorical distribution over the discrete values.

\subsection{Training and prediction}
\label{sec:training}

\vspace{-0.25em}
ACE is trained via maximum-likelihood on synthetic data consisting of batches of context and target sets, using the Adam optimizer (details in \cref{app:sampling}).

\vspace{-0.25em}
\paragraph{Training.}
We generate each problem instance hierarchically by first sampling the latent variables $\vtheta$, and then data points $(\X, \y)$ according to the generative model of the task. For example, $\vtheta$ could be length scale and output scale of a 1D Gaussian process with a given kernel, and $(\X, \y)$ input locations and function values. Data and latents are randomly split between context and target. For training with probabilistic information $\mathbf{p}_l$, we first sample the priors for each latent variable from a hierarchical model $\mathcal{P}$ which includes mixtures of Gaussians and Uniform distributions (see \cref{app:priors}) and then sample the value of the latent from the chosen prior. During training, we minimize the expected negative log-likelihood of the target set conditioned on the context, $\mathcal{L}\left(\textbf{w}\right)$:
\begin{equation} 
\label{eq:loss}
\begin{split}
\mathbb{E}_\mathcal{\mathbf{p} \sim P} &\left[ \mathbb{E}_{\dataplus_N, \vxi_{1:M},\z_{1:M} \sim \mathbf{p}}\left[-\sum_{m=1}^M \log q\left(z^\star_m|\r_\textbf{w}(\vxi^\star_m, \dataplus_N)\right) \right] \right],
\end{split}
\end{equation}
\vspace{-1.75em}

\noindent where $q$ is our model's prediction (a mixture of Gaussians or categorical), and $\textbf{w}$ are the model parameters. Minimizing \cref{eq:loss} is equivalent to minimizing the Kullback-Leibler (KL) divergence between the data sampled from the generative process and the model. 
Since the generative process is consistent with the provided contextual prior information, training will aim to converge (KL-wise) as close as possible, for the model capacity, to the correct Bayesian posteriors and predictive distributions for the specified generative model and priors \citep{muller2022transformers, elsemueller2024sensitivity}.

\vspace{-0.25em}
\paragraph{Prediction.} ACE is trained via \emph{independent} predictions of target data and latents, \cref{eq:loss}. Given the closed-form likelihood (mixture of Gaussians or categorical), we can easily evaluate or sample from the predictive distribution at any desired target point (data or latent) in parallel, conditioned on the context. Moreover, we can predict \emph{non-independent} joint distributions autoregressively \citep{nguyen2022transformer,bruinsma2023autoregressive}; see \cref{app:autoregressive} for details.

\begin{tcolorbox}[colback=red!5!white,colframe=red!75!black]
\paragraph{Task-specific contributions.} The availability of predictive distributions in closed form allows ACE to simplify tasks or perform new ones. We give an example in \cref{exp:bo}, where ACE facilitates the computation of acquisition functions in Bayesian optimization.
\end{tcolorbox}

\input{image.tex}

\vspace{-0.25em}
\section{EXPERIMENTS}
\label{sec:experiments}

\vspace{-0.25em}
The following section showcases ACE's capabilities as a general framework applicable to diverse machine learning and modeling tasks.\footnote{The code implementation of ACE is available at \href{https://github.com/acerbilab/amortized-conditioning-engine/}{github.com/acerbilab/amortized-conditioning-engine/}.}

Firstly, \cref{exp:image} demonstrates how ACE complements transformer-based meta-learning in image completion and classification. In \cref{exp:bo}, we show how ACE can be applied to Bayesian optimization (BO) by treating the location and value of the global optimum as latent variables. We then move to simulation-based inference (SBI) in \cref{exp:sbi}, where ACE unifies the SBI problem into a single framework, treating parameters as latent variables and affording both forward and inverse modelling. Notably, SBI and BO users may have information about the simulator or target function. ACE affords incorporation of informative priors about latent variables at runtime, as detailed in \cref{sec:prior}, a variant we call ACEP in these experiments. Finally, 
in \cref{app:gp} we provide extra experimental results on Gaussian Processes (GPs) where ACE can accurately predict the kernel, \ie model selection (\cref{fig:GP}), while at the same time learn the hyperparameters, in addition to the common data prediction task.

\vspace{-0.25em}
\subsection{Image completion and classification}
\label{exp:image}

\vspace{-0.25em}
We treat image completion as a regression task \citep{garnelo2018neural}, where the goal is given some limited $\data_N$ of image coordinates and corresponding pixel values to predict the complete image. For the \textsc{MNIST} \citep{deng2012mnist} task, we downsize the images to $16 \times 16$ and likewise for \textsc{CelebA} to $32 \times 32$ \citep{liu2015faceattributes}. We turn the class label into a single discrete latent for \textsc{MNIST} while for \textsc{CelebA}, we feed the full $40$ corresponding binary features (\eg, \textsc{BrownHair, Man, Smiling}). The latents are sampled using the procedure outlined in \cref{app:sampling} and more experimental details of the image completion task can be found in \cref{app:image}. Notably, ACE affords conditional image generation, i.e., predictions of pixels based on latent variables ($\vtheta$) -- such as class labels in \textsc{MNIST} and appearance features in \textsc{CelebA} -- as well as image classification, i.e. the prediction of these latent variables themselves from pixel values.

\textbf{Results.} In \cref{fig:whole_image_celeb_main} we present a snapshot of the results for \textsc{CelebA} and in \cref{app:image} we present the same for \textsc{MNIST}. The more complex output distribution allows ACE to outperform other Transformer NPs convincingly, and the integration of latent information shows a clear improvement. In \cref{app:image}, we present our full results, including predicting $\vtheta$. 

\subsection{Bayesian optimization (BO)}
\label{exp:bo}

\begin{figure}[t!]

  \centerline{\legendTrueFunction \quad \legendObservations}
  \vspace{0.1cm}
  \centering
  \begin{subfigure}{0.50\textwidth}
    \centering
    \setlength{\figurewidth}{1.0\linewidth}
    \setlength{\figureheight}{0.45\linewidth}
    \input{figures/figure_1a}
    \setlength{\abovecaptionskip}{1pt}
    \captionsetup{labelformat=empty}
    \caption{}
    \label{fig:BO_walkthrough_1}
  \end{subfigure}
  
  \vspace{-8pt}
  
  \begin{subfigure}{0.50\textwidth}
    \centering
    \setlength{\figurewidth}{1.0\linewidth}
    \setlength{\figureheight}{0.45\linewidth}
    \input{figures/figure_1b}
    \setlength{\abovecaptionskip}{1pt}
    \captionsetup{labelformat=empty}
    \caption{}
    \label{fig:BO_walkthrough_2}
  \end{subfigure}
  
  \vspace{-5pt}
  
  \caption{Bayesian Optimization example. (\subref{fig:BO_walkthrough_1}) ACE predicts function values (\legendpygivenx) \ as well as latents: optimum location (\legendpxopt) \ and optimum value (\legendpyopt). (\subref{fig:BO_walkthrough_2}) Further conditioning on \legendyoptcond \ (here the true minimum value) leads to updated predictions.}
  \label{fig:BO_walkthrough}
  \vspace{-0.4cm}
\end{figure}

BO aims to find the global minimum $ \yopt = f(\xopt) $  of a black-box function. This is typically achieved iteratively by building a \emph{surrogate model} that approximates the target and optimizing an \emph{acquisition function} $\alpha(\mathbf{x})$ to determine the next query point. ACE provides additional modeling options for the BO loop by affording direct conditioning on, and prediction of, key latents $\xopt$ and $\yopt$, yielding closed-form predictive distributions and samples for $p(\xopt | \data_N)$, $p(\yopt | \data_N)$, and $ p(\xopt | \data_N, \yopt)$; see  \cref{fig:BO_walkthrough}.

For the BO task, we trained ACE on synthetic functions generated from GP samples with RBF and Matérn-($\nicefrac{1}{2}$, $\nicefrac{3}{2}$, $\nicefrac{5}{2}$) kernels and a random global optimum $(\xopt,\yopt)$ within the function domain; see \cref{app:bo} for details.  We leverage ACE's explicit modeling of latents in multiple ways.

\paragraph{Acquisition functions with ACE.}

ACE affords a straightforward implementation of a variant of \textbf{Thompson Sampling} (TS) \citep{dutordoir2023neural, liu2024large}. First, we sample a candidate optimum value, $\yopt^\star$, conditioning on it being below a threshold $\tau$, from the truncated predictive distribution $\yopt^\star \sim p(\yopt | \data_N, \yopt<\tau)$; see \cref{fig:BO_walkthrough_1}. Given $\yopt^\star$, we then sample the query point $\x^\star \sim p(\xopt | \data_N, \yopt^\star)$; \cref{fig:BO_walkthrough_2}.\footnote{Why not sampling directly from $p(\xopt | \data_N)$? The issue is that $p(\xopt | \data_N)$ may reasonably include substantial probability mass at the current optimum, which would curb exploration. The constraint $\yopt < \tau$, with $\tau$ (just) below the current optimum value, ensures continual exploration.} This process is repeated iteratively within the BO loop (see \cref{fig:bo_evolution}). For higher input dimensions ($D > 1$), we sample $\x^\star$ autoregressively; see \cref{sec:ace-bo-alg}. Crucially, ACE's capability of directly modeling $\xopt$ and $\yopt$ bypasses the need for surrogate optimization or grid searches typical of standard TS implementations (\eg, GP-TS or TNP-D based TS).

\input{bo_main}

ACE also easily supports advanced acquisition functions used in BO, such as \textbf{Max-Value Entropy Search} (MES; \citealp{wang2017max}).
For a candidate point $\xt$, MES evaluates the expected gain in mutual information between $\yopt$ and $\xt$:
\begin{equation}
\begin{aligned}
\label{eq:mes}
    \alpha_\text{MES}(\xt) &= \textcolor{red}{H[p(\yt| \xt, \data_N)]}\\ &- \mathbb{E}_{\textcolor{latent2}{p(\yopt| \data_N)}}\left[\textcolor{red}{H[p(\yt| \xt, \data_N, \yopt)]}\right].
\end{aligned}
\end{equation}
With all predictive distributions available in closed form, ACE can readily calculate the expectation and entropies in \cref{eq:mes} via \textcolor{latent2}{Monte Carlo sampling} and fast 1D \textcolor{red}{numerical integration}, unlike other methods that require more laborious approximations. For maximizing $\alpha(\x^\star)$, we obtain a good set of candidate points via Thompson Sampling (see \cref{sec:ace-bo-alg} for details). %

\paragraph{Results.} We compare ACE with gold-standard Gaussian processes (GPs) and the state-of-the-art TNP-D model (\cref{fig:bo_comparisons}). Additionally, we test a setup where prior information is provided about the location of the optimum \citep{souza2021bayesian, hvarfner2022pi, muller2023pfns4bo}; \cref{fig:bo_prior_main_comparisons}. 
Unlike other methods that employ heuristics or complex approximations, ACE's architecture affords seamless incorporation of a prior $p(\xopt)$. 
Here, we consider two types of priors: strong and weak, represented by Gaussians with a standard deviation equal to, respectively, 10\% and 25\% of the optimization box width in each dimension, and mean drawn from a Gaussian centered on the true optimum and same standard deviation (see \cref{sec:bo_prior}).

In \cref{fig:bo_comparisons}, we show the performance of ACE Thompson sampling (ACE-TS) and MES (ACE-MES) with GP-based MES (GP-MES; \citealp{wang2017max}), GP-based Thompson Sampling (GP-TS; \citealp{balandat2020botorch}), and Autoregressive TNP-D based Thompson Sampling (AR-TNPD-TS; \citealp{bruinsma2023autoregressive,nguyen2022transformer}) on several benchmark functions  (see \cref{sec:bo-benchmarks} for details).
ACE-MES frequently outperforms ACE-TS and often matches the gold-standard GP-MES. In the prior setting, we compare ACE without (ACE-TS) and with prior (ACEP-TS) against $\pi$BO-TS, a state-of-the-art heuristic for prior injection in BO \citep{hvarfner2022pi}, as well as the GP-TS baseline. ACEP-TS shows significant improvement over its no-prior variant ACE-TS while showing competitive performance compared to $\pi$BO-TS in both weak and strong prior case (\cref{fig:bo_prior_main_comparisons}).

\begin{figure}[htb]
  \scriptsize
  
  \setlength{\figurewidth}{.19\textwidth}
  \setlength{\figureheight}{.75\figurewidth}

  \centerline{\legendACETS \quad \legendACEPTS \quad \legendGPTS \quad \legendpiBOTS}
  
  \vspace{0.2cm}
  \begin{subfigure}[b]{.19\textwidth}
    \centering
    \input{figures/bop_2d_michalewicz_weak}
  \end{subfigure}
  \hspace{0.8cm}
  \begin{subfigure}[b]{.16\textwidth}
    \centering
    \input{figures/bop_2d_michalewicz_strong}
  \end{subfigure}
  
  \begin{subfigure}[b]{.19\textwidth}
    \centering
    \input{figures/bop_3d_levy_weak}
  \end{subfigure}
  \hspace{0.8cm}
  \begin{subfigure}[b]{.16\textwidth}
    \centering
    \input{figures/bop_3d_levy_strong}
  \end{subfigure}
  \vspace{-0.3cm}
  \caption{\textbf{Bayesian optimization with prior over $\xopt$.} Regret comparison (mean ± standard error) on 2D and 3D optimization benchmarks. Left: Weak Gaussian prior (25\%), Right: Strong prior (10\%). ACEP-TS performs competitively compared to $\pi$BO-TS.}
  \label{fig:bo_prior_main_comparisons}
\end{figure}

\subsection{Simulation-based inference (SBI)}\label{exp:sbi}

\begin{table*}[!t]
\centering
\scalebox{0.88}{
\begin{tabular}{cc|cccc|cc}
\toprule
           &             & NPE & NRE & Simformer & ACE & ACEP$_\text{weak prior}$ & ACEP$_\text{strong prior}$ \\ \cmidrule(lr){3-8}
\multirow{3}{*}{OUP} & $\text{log-probs}_{\theta}$ ($\uparrow$) &    \textbf{1.09}\textcolor{gray}{(0.10)}   &  \textbf{1.07}\textcolor{gray}{(0.13)}  & \textbf{1.03}\textcolor{gray}{(0.04)} & \textbf{1.03}\textcolor{gray}{(0.02)}   &    1.05\textcolor{gray}{(0.02)}               & 1.44\textcolor{gray}{(0.03)}               \\ 
                     & $\text{RMSE}_{\theta}$ ($\downarrow$)    &  \textbf{0.48}\textcolor{gray}{(0.01)}    &   0.49\textcolor{gray}{(0.00)} &  0.50\textcolor{gray}{(0.02)}   &   \textbf{0.48}\textcolor{gray}{(0.00)}     &    0.43\textcolor{gray}{(0.01)} & 0.27\textcolor{gray}{(0.00)}                  \\  
                     & $\text{MMD}_{y}$ ($\downarrow$) &  - & - & \textbf{0.43}\textcolor{gray}{(0.02)}  &  0.51\textcolor{gray}{(0.00)}  &  0.37\textcolor{gray}{(0.00)}   &   0.35\textcolor{gray}{(0.00)}             \\  
                     
\midrule \midrule
\multirow{3}{*}{SIR} & $\text{log-probs}_{\theta}$ ($\uparrow$) &  6.53\textcolor{gray}{(0.11)}  &  6.24\textcolor{gray}{(0.16)}  &  \textbf{6.89}\textcolor{gray}{(0.09)}  &  6.78\textcolor{gray}{(0.02)}   & 6.62\textcolor{gray}{(0.10)}                   &  6.69\textcolor{gray}{(0.10)}                \\ 
                     & $\text{RMSE}_{\theta}$ ($\downarrow$) & \textbf{0.02}\textcolor{gray}{(0.00)}   &  0.03\textcolor{gray}{(0.00)}   &  \textbf{0.02}\textcolor{gray}{(0.00)} & \textbf{0.02}\textcolor{gray}{(0.00)} & 0.02\textcolor{gray}{(0.00)}  &  0.02\textcolor{gray}{(0.00)}                 \\  
                     & $\text{MMD}_{y}$ ($\downarrow$)  & - & - & \textbf{0.02}\textcolor{gray}{(0.00)} &  \textbf{0.02}\textcolor{gray}{(0.00)} & 0.02\textcolor{gray}{(0.00)} & 0.00\textcolor{gray}{(0.00)}   \\ 
\midrule \midrule
\multirow{3}{*}{Turin} & $\text{log-probs}_{\theta}$ ($\uparrow$) &   1.99\textcolor{gray}{(0.05)}   &  2.33\textcolor{gray}{(0.07)}  & \textbf{3.16}\textcolor{gray}{(0.03)}  &  \textbf{3.14}\textcolor{gray}{(0.02)}   &     3.58\textcolor{gray}{(0.04)}  &      4.87\textcolor{gray}{(0.08)}        \\ 
                     & $\text{RMSE}_{\theta}$ ($\downarrow$)      &  0.26\textcolor{gray}{(0.00)}   &  0.28\textcolor{gray}{(0.00)}   &  0.25\textcolor{gray}{(0.00)}   &  \textbf{0.24}\textcolor{gray}{(0.00)}   &  0.21\textcolor{gray}{(0.00)} &  0.13\textcolor{gray}{(0.00)}                  \\  
                     & $\text{MMD}_{y}$ ($\downarrow$)  &  -  &  -   &  \textbf{0.35}\textcolor{gray}{(0.00)}    &   \textbf{0.35}\textcolor{gray}{(0.00)} & 0.35\textcolor{gray}{(0.00)}  & 0.34\textcolor{gray}{(0.00)}       \\ 
\bottomrule
\end{tabular}}

\caption{\textbf{Comparison metrics for SBI models} on parameters ($\vtheta$) and data ($y$) prediction; mean and (\textcolor{gray}{standard deviation}) from 5 runs. \emph{Left}: Statistically significantly (see \cref{app:sbi_cfg}) best results are \textbf{bolded}. ACE shows performance comparable to the other dedicated methods on latents prediction. In the data prediction task, ACE performs similarly to Simformer with much lower sampling cost at runtime (see text). \emph{Right}: ACE can leverage probabilistic information provided at runtime by informative priors (ACEP), yielding improved performance.}
\label{tab:sbi}
\end{table*}

\vspace{-0.25em}
We now apply ACE to simulation-based inference (SBI; \citealp{cranmer2020frontier}). With ACE, we can predict the posterior distribution of (latent) model parameters, simulate data based on parameters, predict missing data given partial observations, and set priors at runtime. We consider two benchmark time-series models, each with two latents: the Ornstein-Uhlenbeck Process (OUP; \citealt{uhlenbeck1930theory}) and the Susceptible-Infectious-Recovered model (SIR; \citealt{kermack1927contribution}); and a third more complex engineering model from the field of radio propagation (Turin;  \citealp{turin1972statistical}), which has four parameters and produces 101-dimensional data representing a radio signal. See \cref{app:sbi} for all model descriptions.

We compare ACE with Neural Posterior Estimation (NPE; \citealt{greenberg2019automatic}), Neural Ratio Estimation (NRE; \citealt{miller2022contrastive}), and Simformer \citep{gloeckler2024all}, from established to state-of-the-art methods in amortized SBI. We evaluate ACE in three different scenarios. For the first one, we use only the observed data as context. For the other two scenarios, we inform ACE with priors over the parameters (ACEP), to assess their impact on posterior prediction. These priors are Gaussians with standard deviation equal to 25\% (weak) or 10\% (strong) of the parameter range, and mean drawn from a Gaussian centered on the true parameter value and the same standard deviation.

We evaluate the performance of posterior estimation using the log probabilities of the ground-truth parameters and the root mean squared error (RMSE) between the true parameters and posterior samples. Since both ACE and Simformer can predict missing data from partial observations -- an ability that previous SBI methods lack -- we also test them on a data prediction task. For each observed dataset, we randomly designate half of the data points as missing and use the remaining half as context for predictions. We then measure performance via the maximum mean discrepancy (MMD) between the true data and the predicted distributions.

\textbf{Results.} Results are reported in \cref{tab:sbi}; see \cref{app:sbi} for details. For ACE, we see that joint training to predict data and latents does not compromise its posterior estimation performance compared to NPE and NRE, even achieving better performance on the Turin model. ACE and Simformer obtain similar results. However, as Simformer uses diffusion, data sampling is substantially slower. For example, we measured the time required to generate 1,000 posterior samples for 100 sets of observations on the OUP model using a CPU (\textcolor{brown}{GPU}) across 5 runs: the average time for Simformer is $\sim$ 130 minutes (\textcolor{brown}{14 minutes}), whereas ACE takes 0.05 seconds (\textcolor{brown}{0.02 s}). 
When we provide ACE with informative priors (ACEP; \cref{tab:sbi} right), performance improves in proportion to the provided information.
Importantly, simulation-based calibration checks \citep{talts2018validating} show that both ACE and ACEP output good approximate posteriors (\cref{app:sbc}).

Finally, we applied ACE to a real-world outbreak dataset \citep{avilov20231978} using an extended, four-parameter version of the SIR model. We show that ACE can handle real data, providing reasonable results under a likely model mismatch (see \cref{app:sbi_real}).

\section{RELATED WORK}
\label{sec:related_work}

\vspace{-0.25em}
Our work combines insights from different fields such as neural processes, meta-learning, and simulation-based inference, providing a new unified and versatile framework for amortized inference.

\vspace{-0.25em}
\paragraph{Neural processes.} 
ACE relates to the broader work on neural processes \citep{garnelo2018neural, garnelo2018conditional, kim2019attentive, gordon2020convolutional, markou2022practical, nguyen2022transformer, huang2023practical} and shares similarities with autoregressive diffusion models~\citep{hoogeboom2022autoregressive}, whose permutation-invariant conditioning set mimics the context set in neural processes.
Unlike previous methods that focused on predictive data distributions conditioned on observed data, ACE also explicitly models and conditions on latent variables of interest. While Latent Neural Processes (LNPs) capture correlations via non-interpretable latent vectors that yield an intractable learning objective \citep{garnelo2018neural}, ACE uses autoregressive sampling to model correlations \citep{bruinsma2023autoregressive}, and `latents' in ACE denote explicitly modelled task variables.

\vspace{-0.25em}
\paragraph{Meta-learning.} Our approach falls within the fields of amortized inference, meta-learning, and pre-trained models, which overlap significantly with neural process literature \citep{finn2017model, finn2018probabilistic}. Prior-Fitted Networks (PFNs) demonstrated the use of transformers for Bayesian inference \citep{muller2022transformers} and optimization \citep{muller2023pfns4bo}, focusing on predictive posterior distributions over data using fixed bins (`Riemannian' distributions). In particular, \citet{muller2023pfns4bo} allow users to indicate a `high-probability' interval for the optimum location at runtime within a set of specified ranges. ACE differs from these methods by allowing the specification of more flexible priors at runtime, using learnable  mixtures of Gaussians for continuous values, and, most importantly, for yielding explicit predictive distributions over both data and task-relevant, interpretable latents. While other works target specific optimization tasks \citep{liu2020task, simpson2021kernel, amos2022tutorial}, we provide a generative model for random functions with known optima, directly amortizing predictive distributions useful for optimization (see \cref{app:bo}). More broadly, ACE relates to efforts in amortized inference over explicit task-relevant latents \citep{mittal2023exploring}, though our focus extends beyond posterior estimation to flexible conditioning and prediction of both data and latents. Additionally, \citet{mittal2024does} show that enforcing a bottleneck in transformers captures task-relevant latents but does not necessarily improve generalization. Unlike their learned latents, ACE conditions on predefined, interpretable task variables, enabling direct control and adaptability. 
By providing methods to condition on interpretable latents as well as injecting probabilistic knowledge (priors), our work aligns with the principles of \emph{informed meta-learning} \citep{kobalczyk2024informed}. Subsequent to our work, \emph{distribution transformers} were recently proposed as amortized meta-learning models for inference with flexible priors and posteriors~\citep{whittle2025distribution}.

\vspace{-0.25em}
\paragraph{Simulation-based inference.} ACE is related to the extensive literature on amortized or neural simulation-based inference (SBI) \citep{cranmer2020frontier}, where deep networks are used to (a) recover the posterior distribution over model parameters \citep{lueckmann2017flexible}, and (b) emulate the simulator or likelihood \citep{papamakarios2019sequential}. While these two steps are often implemented separately, recent work has started to combine them \citep{radev2023jana}. In particular, the recently proposed Simformer architecture allows users to freely condition on observed variables and model parameters, in a manner similar to ACE \citep{gloeckler2024all}. Simformer uses a combination of a transformer architecture and a denoising diffusion model \citep{song2019generative}, which makes it suitable for continuous predictions but does not immediately apply to discrete outputs. Notably, Simformer allows the user to specify \emph{intervals} for variables at runtime, but not (yet) general priors. To our knowledge, before our paper the only work in amortized SBI that afforded some meaningful flexibility in prior specification at runtime was \citet{elsemueller2024sensitivity}. Even so, the choice there is between a limited number of fixed priors or a global prior scale parameter, with the main focus being sensitivity analysis \citep{elsemueller2024sensitivity}. Importantly, all these works focus entirely on SBI settings, while our paper showcases the applicability of ACE to a wider variety of machine learning tasks.

\vspace{-0.25em}
\section{DISCUSSION}
\label{sec:discussion}\label{sec:limitations}

\vspace{-0.25em}
In this paper, we introduced the Amortized Conditioning Engine (ACE), a unified transformer-based architecture that affords arbitrary probabilistic conditioning and prediction over data and task-relevant variables for a wide variety of tasks.
In each tested domain ACE performed on par with state-of-the-art, bespoke solutions, and with greater flexibility. As a key feature, ACE allows users to specify probabilistic information at runtime (priors) without the need for retraining as required instead by the majority of previous amortized inference methods \citep{cranmer2020frontier}.

\vspace{-0.25em}
\paragraph{Limitations and future work.}

As all amortized and machine learning methods, ACE's predictions become unreliable if applied to data unseen during training due to mismatch between simulated and real data. This is an active area of research in terms of developing both more robust training objectives \citep{huang2024learning} and diagnostics \citep{schmitt2023detecting}.

The method's quadratic complexity in context size could benefit from sub-quadratic attention variants \citep{feng2023latent}, while incorporating equivariances \citep{huang2023practical} and variable covariate dimensionality \citep{liu2020task, dutordoir2023neural,muller2023pfns4bo} could further improve performance.

While ACE can learn full joint distributions through 1D marginals \citep{bruinsma2023autoregressive}, scaling to many data points and latents remains challenging. 
Similar scaling challenges affect our prior-setting approach, currently limited to few latents under factorized, smooth priors.
Future work should also explore discovering interpretable latents~\citep{mittal2024does} and handling multiple tasks simultaneously~\citep{kim2022multi,ashman2024context}, extending beyond our current single-task supervised learning approach.

\vspace{-0.25em}
\paragraph{Conclusions.} ACE shows strong promise as a new unified and versatile method for amortized probabilistic conditioning and prediction, able to perform various probabilistic machine learning tasks.

\section*{Acknowledgments}

PC, DH, UR, SK, and LA were supported by the Research Council of Finland (Flagship programme: Finnish Center for Artificial Intelligence FCAI). NL was funded by Business Finland (project 3576/31/2023). LA was also supported by Research Council of Finland grants 358980 and 356498. SK was also supported by the UKRI Turing AI World-Leading Researcher Fellowship, [EP/W002973/1].
The authors wish to thank the Finnish Computing Competence Infrastructure (FCCI), Aalto Science-IT project, and CSC–IT Center for Science, Finland, for
the computational and data storage resources provided, including access to the LUMI supercomputer, owned by the EuroHPC Joint Undertaking, hosted by CSC (Finland) and the LUMI consortium (LUMI project 462000551).

\bibliographystyle{unsrtnat}  %
\section*{Checklist}
 \begin{enumerate}

 \item For all models and algorithms presented, check if you include:
 \begin{enumerate}
   \item A clear description of the mathematical setting, assumptions, algorithm, and/or model.
   
   [\textcolor{red}{Yes}]
   See description in \cref{sec:ace} and further details in \cref{app:methods} and \cref{app:experiments}.
   \item An analysis of the properties and complexity (time, space, sample size) of any algorithm.
   
   [\textcolor{red}{Not Applicable}] Our architecture is based on the standard attention mechanism \citep{vaswani2017attention} which has well-known properties.
   \item (Optional) Anonymized source code, with specification of all dependencies, including external libraries. 
   [\textcolor{red}{Yes}] Our source code is available at \url{https://github.com/acerbilab/amortized-conditioning-engine/}.
 \end{enumerate}

 \item For any theoretical claim, check if you include:
 \begin{enumerate}
   \item Statements of the full set of assumptions of all theoretical results. 
   [\textcolor{red}{Not Applicable}]
   \item Complete proofs of all theoretical results.
   [\textcolor{red}{Not Applicable}]
   \item Clear explanations of any assumptions. 
   [\textcolor{red}{Not Applicable}]
 \end{enumerate}

 \item For all figures and tables that present empirical results, check if you include:
 \begin{enumerate}
   \item The code, data, and instructions needed to reproduce the main experimental results (either in the supplemental material or as a URL).

   [\textcolor{red}{Yes}] The linked GitHub repo contains working examples demonstrating our main results, in addition to the full codebase.
   \item All the training details (e.g., data splits, hyperparameters, how they were chosen). 
   
   [\textcolor{red}{Yes}] In  \cref{app:experiments} we provide all experimental details.
   
     \item A clear definition of the specific measure or statistics and error bars (e.g., with respect to the random seed after running experiments multiple times). 
     
     [\textcolor{red}{Yes}] See \cref{exp:image}, \cref{exp:bo}, \cref{exp:sbi} and \cref{app:experiments}.
    
    \item A description of the computing infrastructure used. (e.g., type of GPUs, internal cluster, or cloud provider). [\textcolor{red}{Yes}] See \cref{app:computation}.
 \end{enumerate}

 \item If you are using existing assets (e.g., code, data, models) or curating/releasing new assets, check if you include:
 \begin{enumerate}
   \item Citations of the creator If your work uses existing assets. [\textcolor{red}{Yes}] See \cref{app:computation}.
   \item The license information of the assets, if applicable. [\textcolor{red}{Yes}] See \cref{app:computation}.
   \item New assets either in the supplemental material or as a URL, if applicable. [\textcolor{red}{Not Applicable}]
   \item Information about consent from data providers/curators. [\textcolor{red}{Not Applicable}]
   \item Discussion of sensible content if applicable, e.g., personally identifiable information or offensive content. [\textcolor{red}{Not Applicable}]
 \end{enumerate}

 \item If you used crowdsourcing or conducted research with human subjects, check if you include:
 \begin{enumerate}
   \item The full text of instructions given to participants and screenshots. [\textcolor{red}{Not Applicable}]
   \item Descriptions of potential participant risks, with links to Institutional Review Board (IRB) approvals if applicable. [\textcolor{red}{Not Applicable}]
   \item The estimated hourly wage paid to participants and the total amount spent on participant compensation. [\textcolor{red}{Not Applicable}]
 \end{enumerate}

 \end{enumerate}

\appendix

\include{appendix}

\end{document}

%% file: image.tex
\begin{figure}[t!]
    \centering
    \begin{subfigure}[b]{0.50\textwidth}  
        \centering
        \begin{subfigure}[b]{0.17\linewidth}  
            \centering
            \includegraphics[width=\linewidth]{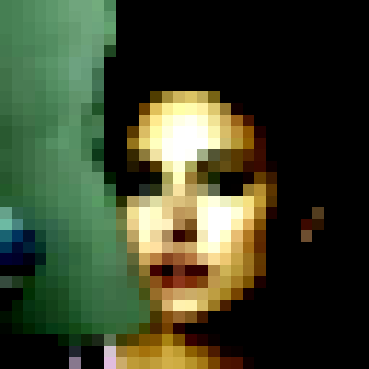}
        \end{subfigure}
        \hspace{-5pt}
        \begin{subfigure}[b]{0.17\linewidth}
            \centering
            \includegraphics[width=\linewidth]{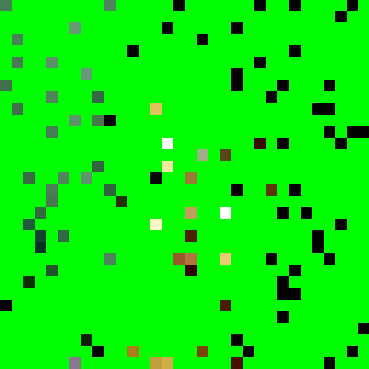}
        \end{subfigure}
        \hspace{-5pt}
        \begin{subfigure}[b]{0.17\linewidth}
            \centering
            \includegraphics[width=\linewidth]{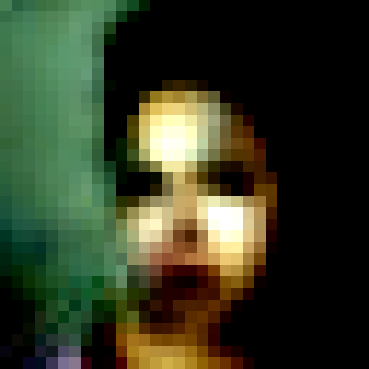}
        \end{subfigure}
        \hspace{-5pt}
        \begin{subfigure}[b]{0.17\linewidth}
            \centering
            \includegraphics[width=\linewidth]{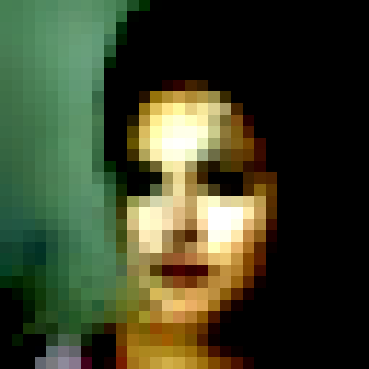}
        \end{subfigure}
        \hspace{-5pt}
        \begin{subfigure}[b]{0.17\linewidth}
            \centering
            \includegraphics[width=\linewidth]{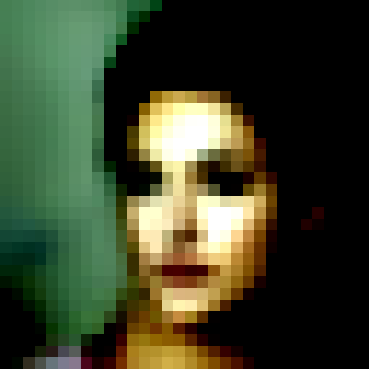}
        \end{subfigure}
        \centering
        \begin{subfigure}[b]{0.17\linewidth}
            \centering
            \includegraphics[width=\linewidth]{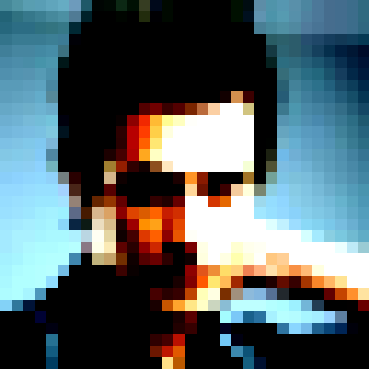}
            \caption{\fontsize{7}{8}\selectfont{Image}}
            
            \label{fig:image_full}
        \end{subfigure}
        \hspace{-5pt}
        \begin{subfigure}[b]{0.17\linewidth}
            \centering
            \includegraphics[width=\linewidth]{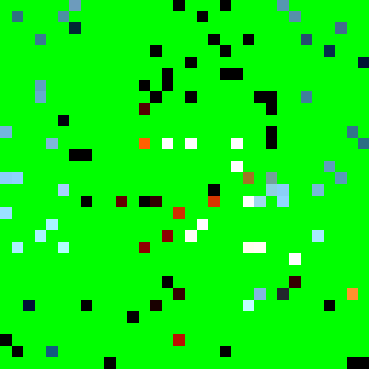}
            \caption{\fontsize{7}{8}\selectfont{$\data_N$}}
            \label{fig:context_celeb}
        \end{subfigure}
        \hspace{-5pt}
        \begin{subfigure}[b]{0.17\linewidth}
            \centering
            \includegraphics[width=\linewidth]{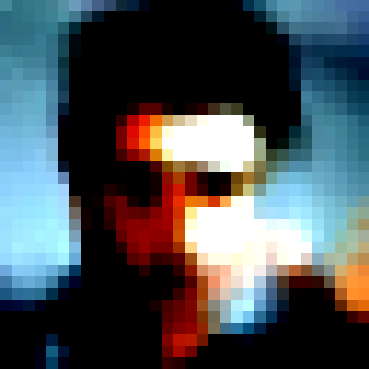}
            \caption{\fontsize{7}{8}\selectfont{TNP-D}}
            \label{fig:tnpd_celeb}
        \end{subfigure}
        \hspace{-5pt}
        \begin{subfigure}[b]{0.17\linewidth}
            \centering
            \includegraphics[width=\linewidth]{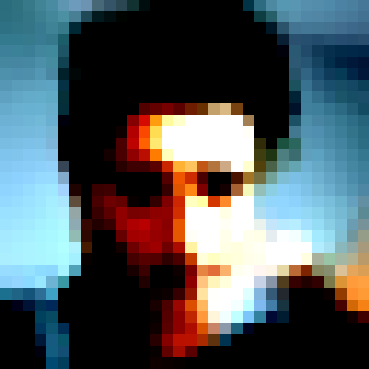}
            \caption{\fontsize{7}{8}\selectfont{ACE}}
            \label{fig:ACE_image_celeb}
        \end{subfigure}
        \hspace{-5pt}
        \begin{subfigure}[b]{0.17\linewidth}
            \centering
            \includegraphics[width=\linewidth]{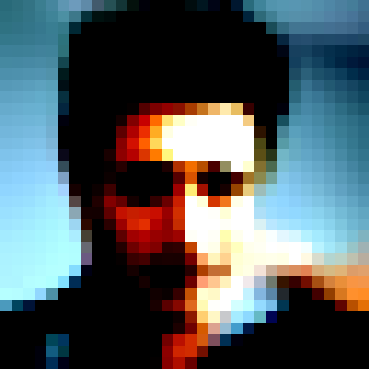}
            \caption{\fontsize{7}{8}\selectfont{ACE-$\vtheta$}}
            \label{fig:ACE_theta_celeb}
        \end{subfigure} 
    \end{subfigure}
    \begin{subfigure}[b]{0.48\textwidth}
        \centering
        \setlength{\figurewidth}{\linewidth}  
        \setlength{\figureheight}{0.50\linewidth}  
        \input{figures/celeba_nlpd}  
        \caption{Negative log-probability density vs. Context}
        \label{fig:celeba-nlpd_main}
    \end{subfigure}
    \caption{\textbf{Image completion.} Image (\subref{fig:image_full}) serves as the reference for the problem, where $10\%$ of the pixels are observed (\subref{fig:context_celeb}). Figures (\subref{fig:tnpd_celeb}) through (\subref{fig:ACE_theta_celeb}) display different models' prediction conditioned on the observed pixels (\subref{fig:context_celeb}). In addition, (\subref{fig:ACE_theta_celeb}) incorporates latent variable $\vtheta$ information for the ACE model. Figure (\subref{fig:celeba-nlpd_main}) illustrates the different models' performance across varying levels of context.}
    \label{fig:whole_image_celeb_main}
\end{figure}

%% file: figures/celeba_nlpd.tex
\begin{tikzpicture}

\definecolor{color0}{rgb}{0.12156862745098,0.466666666666667,0.705882352941177}
\definecolor{color1}{rgb}{1,0.498039215686275,0.0549019607843137}
\definecolor{color2}{rgb}{0.172549019607843,0.627450980392157,0.172549019607843}

\begin{axis}[
height=\figureheight,
legend cell align={left},
legend style={fill opacity=0.8, draw opacity=1, text opacity=1, draw=white!80!black},
tick align=outside,
tick pos=left,
width=\figurewidth,
x grid style={white!69.0196078431373!black},
xlabel={Context \%},
xmajorgrids,
xmin=-0.45, xmax=31.45,
xtick style={color=black},
y grid style={white!69.0196078431373!black},
ymajorgrids,
ymin=-1.37318834066391, ymax=1.1993261218071,
ytick style={color=black}
]
\path [fill=color0, fill opacity=0.2]
(axis cs:1,0.62503057718277)
--(axis cs:1,0.484585016965866)
--(axis cs:2,0.292641758918762)
--(axis cs:3,0.0987067520618439)
--(axis cs:4,-0.0287488773465157)
--(axis cs:5,-0.1368288397789)
--(axis cs:6,-0.262523829936981)
--(axis cs:8,-0.350899547338486)
--(axis cs:10,-0.424698561429977)
--(axis cs:15,-0.67389041185379)
--(axis cs:20,-0.882839322090149)
--(axis cs:25,-0.91611647605896)
--(axis cs:30,-1.04242813587189)
--(axis cs:30,-0.912706911563873)
--(axis cs:30,-0.912706911563873)
--(axis cs:25,-0.807681083679199)
--(axis cs:20,-0.70132052898407)
--(axis cs:15,-0.582760632038116)
--(axis cs:10,-0.359173387289047)
--(axis cs:8,-0.272823601961136)
--(axis cs:6,-0.094575509428978)
--(axis cs:5,-0.0148505643010139)
--(axis cs:4,0.093697302043438)
--(axis cs:3,0.212607622146606)
--(axis cs:2,0.370250701904297)
--(axis cs:1,0.62503057718277)
--cycle;

\path [fill=color1, fill opacity=0.2]
(axis cs:1,0.452571988105774)
--(axis cs:1,0.313966274261475)
--(axis cs:2,0.103770315647125)
--(axis cs:3,-0.106521740555763)
--(axis cs:4,-0.245982736349106)
--(axis cs:5,-0.357124924659729)
--(axis cs:6,-0.500539183616638)
--(axis cs:8,-0.560498297214508)
--(axis cs:10,-0.635713636875153)
--(axis cs:15,-0.873799085617065)
--(axis cs:20,-1.08774971961975)
--(axis cs:25,-1.12952041625977)
--(axis cs:30,-1.25625586509705)
--(axis cs:30,-1.13334798812866)
--(axis cs:30,-1.13334798812866)
--(axis cs:25,-1.03371858596802)
--(axis cs:20,-0.908951222896576)
--(axis cs:15,-0.78762674331665)
--(axis cs:10,-0.554427564144135)
--(axis cs:8,-0.477529346942902)
--(axis cs:6,-0.307096481323242)
--(axis cs:5,-0.225631594657898)
--(axis cs:4,-0.134631365537643)
--(axis cs:3,-0.0155173949897289)
--(axis cs:2,0.167022705078125)
--(axis cs:1,0.452571988105774)
--cycle;

\path [fill=color2, fill opacity=0.2]
(axis cs:1,1.08239364624023)
--(axis cs:1,1.01260638237)
--(axis cs:2,0.781918823719025)
--(axis cs:3,0.626531839370728)
--(axis cs:4,0.522833526134491)
--(axis cs:5,0.405584663152695)
--(axis cs:6,0.32244610786438)
--(axis cs:8,0.18397530913353)
--(axis cs:10,0.076886847615242)
--(axis cs:15,-0.153240695595741)
--(axis cs:20,-0.360990613698959)
--(axis cs:25,-0.442191958427429)
--(axis cs:30,-0.577674269676208)
--(axis cs:30,-0.470009326934814)
--(axis cs:30,-0.470009326934814)
--(axis cs:25,-0.35465681552887)
--(axis cs:20,-0.238179415464401)
--(axis cs:15,-0.101866409182549)
--(axis cs:10,0.116055265069008)
--(axis cs:8,0.244121551513672)
--(axis cs:6,0.383738160133362)
--(axis cs:5,0.483318656682968)
--(axis cs:4,0.582003057003021)
--(axis cs:3,0.694413065910339)
--(axis cs:2,0.858468949794769)
--(axis cs:1,1.08239364624023)
--cycle;

\addplot [semithick, color0, mark=*, mark size=2, mark options={solid}]
table {%
1 0.554807782173157
2 0.33144623041153
3 0.155657187104225
4 0.0324742123484612
5 -0.0758396983146667
6 -0.17854967713356
8 -0.311861574649811
10 -0.391935974359512
15 -0.628325521945953
20 -0.792079925537109
25 -0.86189877986908
30 -0.977567553520203
};
\addlegendentry{\tiny{ACE}}
\addplot [semithick, color1, mark=*, mark size=2, mark options={solid}]
table {%
1 0.383269131183624
2 0.135396510362625
3 -0.0610195696353912
4 -0.190307050943375
5 -0.291378259658813
6 -0.40381783246994
8 -0.519013822078705
10 -0.595070600509644
15 -0.830712914466858
20 -0.998350501060486
25 -1.08161950111389
30 -1.19480192661285
};
\addlegendentry{\tiny{ACE-$\vtheta$}}
\addplot [semithick, color2, mark=*, mark size=2, mark options={solid}]
table {%
1 1.04750001430511
2 0.820193886756897
3 0.660472452640533
4 0.552418291568756
5 0.444451659917831
6 0.353092133998871
8 0.214048430323601
10 0.0964710563421249
15 -0.127553552389145
20 -0.29958501458168
25 -0.398424386978149
30 -0.523841798305511
};
\addlegendentry{\tiny{TNP-D}}
\end{axis}

\end{tikzpicture}

%% file: bo_main.tex

\begin{figure*}[!t]
  \centering
  \scriptsize
 
  \setlength{\figurewidth}{.19\textwidth}
  \setlength{\figureheight}{.75\figurewidth}

  \centerline{\legendACETS \quad \legendACEMES \quad \legendTNPDTS \quad \legendGPTS \quad \legendGPMES \quad \legendRandom}
  \vspace{0.2cm}
  
  \begin{minipage}[b]{\textwidth}  
    \centering
    \begin{subfigure}[b]{.25\textwidth}
      \centering
      \input{figures/bo_1d_gramacy_lee}
    \end{subfigure}
    \begin{subfigure}[b]{.23\textwidth}
      \centering
      \input{figures/bo_2d_branin_scaled}
    \end{subfigure}
    \begin{subfigure}[b]{.23\textwidth}
      \centering
      \input{figures/bo_3d_hartmann3}
    \end{subfigure}
    \begin{subfigure}[b]{.23\textwidth}
      \centering
      \input{figures/bo_4d_rosenbrock4d}
    \end{subfigure}
  \end{minipage}  
  \begin{minipage}[b]{\textwidth}  
    \centering
    \begin{subfigure}[b]{.25\textwidth}
      \centering
      \input{figures/bo_5d_rosenbrock5d}
    \end{subfigure}
    \begin{subfigure}[b]{.23\textwidth}
      \centering
      \input{figures/bo_5d_levy5d}
    \end{subfigure}
    \begin{subfigure}[b]{.23\textwidth}
      \centering
      \input{figures/bo_6d_hartmann6d}
    \end{subfigure}
    \begin{subfigure}[b]{.23\textwidth}
      \centering
      \input{figures/bo_6d_levy_6d}
    \end{subfigure}
  \end{minipage}  
  \vspace{-0.5cm}
  \caption{\textbf{Bayesian optimization results.} Regret comparison (mean ± standard error) for different methods across benchmark tasks.}
  \label{fig:bo_comparisons}
  \vspace{-0.3cm}
\end{figure*}

%% file: figures/bop_2d_michalewicz_weak.tex
\begin{tikzpicture}

\definecolor{color0}{rgb}{0,0,1}
\definecolor{color1}{rgb}{1,0.549019607843137,0}
\definecolor{color2}{rgb}{1,0.647058823529412,0}
\definecolor{color3}{rgb}{0.564705882352941,0.933333333333333,0.564705882352941}

\begin{axis}[axis on top,
enlarge x limits=false,
enlarge y limits=false,
height=\figureheight,
scale only axis,
tick align=outside,
tick pos=left,
tick pos=left,
width=\figurewidth,
xmin=10, xmax=100,
xtick style={color=black},
xtick={-10,0,10,30,50,70,90},
xticklabels={\ensuremath{-}10,0,10,30,50,70,90},
ylabel={Regret},
ymin=-0.05, ymax=1.2,
ytick={0.   , 1.2},
]
\node[anchor=north east] at (rel axis cs:1,1) {Michalewicz 2D (weak)};
\path [draw=blue, fill=blue, opacity=0.3]
(axis cs:10,1.02326834201813)
--(axis cs:10,0.826151013374329)
--(axis cs:11,0.764189541339874)
--(axis cs:12,0.665249347686768)
--(axis cs:13,0.548758447170258)
--(axis cs:14,0.548758447170258)
--(axis cs:15,0.468498706817627)
--(axis cs:16,0.44959083199501)
--(axis cs:17,0.435864716768265)
--(axis cs:18,0.386376172304153)
--(axis cs:19,0.385418117046356)
--(axis cs:20,0.301011860370636)
--(axis cs:21,0.301011860370636)
--(axis cs:22,0.26057305932045)
--(axis cs:23,0.243601024150848)
--(axis cs:24,0.223688691854477)
--(axis cs:25,0.112993977963924)
--(axis cs:26,0.10345321893692)
--(axis cs:27,0.0923125669360161)
--(axis cs:28,0.0902853310108185)
--(axis cs:29,0.0871852561831474)
--(axis cs:30,0.086898647248745)
--(axis cs:31,0.086898647248745)
--(axis cs:32,0.086898647248745)
--(axis cs:33,0.0849867165088654)
--(axis cs:34,0.0807064324617386)
--(axis cs:35,0.0798281729221344)
--(axis cs:36,0.0701713785529137)
--(axis cs:37,0.0573972314596176)
--(axis cs:38,0.0560382455587387)
--(axis cs:39,0.0550645738840103)
--(axis cs:40,0.0550645738840103)
--(axis cs:41,0.0497946590185165)
--(axis cs:42,0.0492888391017914)
--(axis cs:43,0.0350885093212128)
--(axis cs:44,0.0347461849451065)
--(axis cs:45,0.0337554588913918)
--(axis cs:46,0.00898849219083786)
--(axis cs:47,0.00898849219083786)
--(axis cs:48,0.00677929818630219)
--(axis cs:49,0.00677929818630219)
--(axis cs:50,0.00677929818630219)
--(axis cs:51,0.00677929818630219)
--(axis cs:52,0.00677867978811264)
--(axis cs:53,0.00677867978811264)
--(axis cs:54,0.00677230581641197)
--(axis cs:55,0.00586185231804848)
--(axis cs:56,0.00478334166109562)
--(axis cs:57,0.00478127971291542)
--(axis cs:58,0.004709642380476)
--(axis cs:59,0.004709642380476)
--(axis cs:60,0.00469528790563345)
--(axis cs:61,0.0034191426821053)
--(axis cs:62,0.0034191426821053)
--(axis cs:63,0.0034191426821053)
--(axis cs:64,0.0034191426821053)
--(axis cs:65,0.0034191426821053)
--(axis cs:66,0.0034191426821053)
--(axis cs:67,0.0034191426821053)
--(axis cs:68,0.0033401888795197)
--(axis cs:69,0.0033401888795197)
--(axis cs:70,0.0033401888795197)
--(axis cs:71,0.0033401888795197)
--(axis cs:72,0.0033401888795197)
--(axis cs:73,0.0033401888795197)
--(axis cs:74,0.00320510659366846)
--(axis cs:75,0.00320510659366846)
--(axis cs:76,0.00320510659366846)
--(axis cs:77,0.00300374790094793)
--(axis cs:78,0.00300374790094793)
--(axis cs:79,0.00300374790094793)
--(axis cs:80,0.00300374790094793)
--(axis cs:81,0.00300374790094793)
--(axis cs:82,0.00300374790094793)
--(axis cs:83,0.00300374790094793)
--(axis cs:84,0.00300374790094793)
--(axis cs:85,0.00300374790094793)
--(axis cs:86,0.00300374790094793)
--(axis cs:87,0.00300374790094793)
--(axis cs:88,0.00300374790094793)
--(axis cs:89,0.00300374790094793)
--(axis cs:90,0.00300374790094793)
--(axis cs:91,0.00300374790094793)
--(axis cs:92,0.00300374790094793)
--(axis cs:93,0.00300374790094793)
--(axis cs:94,0.00272491970099509)
--(axis cs:95,0.00272491970099509)
--(axis cs:96,0.00272491970099509)
--(axis cs:97,0.00272491970099509)
--(axis cs:98,0.00272491970099509)
--(axis cs:99,0.00272491970099509)
--(axis cs:100,0.00272491970099509)
--(axis cs:101,0.00272491970099509)
--(axis cs:102,0.00272491970099509)
--(axis cs:103,0.00272491970099509)
--(axis cs:104,0.00272491970099509)
--(axis cs:105,0.00272491970099509)
--(axis cs:106,0.00272491970099509)
--(axis cs:107,0.00272491970099509)
--(axis cs:108,0.00272491970099509)
--(axis cs:109,0.00272491970099509)
--(axis cs:110,0.00272491970099509)
--(axis cs:111,0.00272491970099509)
--(axis cs:111,0.00428966479375958)
--(axis cs:111,0.00428966479375958)
--(axis cs:110,0.00428966479375958)
--(axis cs:109,0.00428966479375958)
--(axis cs:108,0.00428966479375958)
--(axis cs:107,0.00428966479375958)
--(axis cs:106,0.00428966479375958)
--(axis cs:105,0.00428966479375958)
--(axis cs:104,0.00428966479375958)
--(axis cs:103,0.00428966479375958)
--(axis cs:102,0.00428966479375958)
--(axis cs:101,0.00428966479375958)
--(axis cs:100,0.00428966479375958)
--(axis cs:99,0.00428966479375958)
--(axis cs:98,0.00428966479375958)
--(axis cs:97,0.00428966479375958)
--(axis cs:96,0.00428966479375958)
--(axis cs:95,0.00428966479375958)
--(axis cs:94,0.00428966479375958)
--(axis cs:93,0.0045623704791069)
--(axis cs:92,0.0045623704791069)
--(axis cs:91,0.0045623704791069)
--(axis cs:90,0.0045623704791069)
--(axis cs:89,0.0045623704791069)
--(axis cs:88,0.0045623704791069)
--(axis cs:87,0.0045623704791069)
--(axis cs:86,0.0045623704791069)
--(axis cs:85,0.0045623704791069)
--(axis cs:84,0.0045623704791069)
--(axis cs:83,0.0045623704791069)
--(axis cs:82,0.0045623704791069)
--(axis cs:81,0.0045623704791069)
--(axis cs:80,0.0045623704791069)
--(axis cs:79,0.0045623704791069)
--(axis cs:78,0.0045623704791069)
--(axis cs:77,0.0045623704791069)
--(axis cs:76,0.00463636126369238)
--(axis cs:75,0.00463636126369238)
--(axis cs:74,0.00463636126369238)
--(axis cs:73,0.0050001940689981)
--(axis cs:72,0.0050001940689981)
--(axis cs:71,0.0050001940689981)
--(axis cs:70,0.0050001940689981)
--(axis cs:69,0.0050001940689981)
--(axis cs:68,0.0050001940689981)
--(axis cs:67,0.0051200813613832)
--(axis cs:66,0.0051200813613832)
--(axis cs:65,0.0051200813613832)
--(axis cs:64,0.0051200813613832)
--(axis cs:63,0.0051200813613832)
--(axis cs:62,0.0051200813613832)
--(axis cs:61,0.0051200813613832)
--(axis cs:60,0.0248472765088081)
--(axis cs:59,0.0500619001686573)
--(axis cs:58,0.0500619001686573)
--(axis cs:57,0.0501195080578327)
--(axis cs:56,0.059897854924202)
--(axis cs:55,0.110437080264091)
--(axis cs:54,0.111178368330002)
--(axis cs:53,0.151487439870834)
--(axis cs:52,0.151487439870834)
--(axis cs:51,0.15712858736515)
--(axis cs:50,0.15712858736515)
--(axis cs:49,0.15712858736515)
--(axis cs:48,0.15712858736515)
--(axis cs:47,0.158941328525543)
--(axis cs:46,0.158941328525543)
--(axis cs:45,0.190311372280121)
--(axis cs:44,0.191041216254234)
--(axis cs:43,0.1912971585989)
--(axis cs:42,0.217930227518082)
--(axis cs:41,0.218290880322456)
--(axis cs:40,0.233166709542274)
--(axis cs:39,0.233166709542274)
--(axis cs:38,0.233897969126701)
--(axis cs:37,0.234854176640511)
--(axis cs:36,0.292614489793777)
--(axis cs:35,0.307650685310364)
--(axis cs:34,0.314028143882751)
--(axis cs:33,0.318763881921768)
--(axis cs:32,0.320137560367584)
--(axis cs:31,0.320137560367584)
--(axis cs:30,0.320137560367584)
--(axis cs:29,0.320342004299164)
--(axis cs:28,0.322635233402252)
--(axis cs:27,0.326067596673965)
--(axis cs:26,0.334868043661118)
--(axis cs:25,0.341961264610291)
--(axis cs:24,0.464788943529129)
--(axis cs:23,0.476096987724304)
--(axis cs:22,0.488581150770187)
--(axis cs:21,0.517304241657257)
--(axis cs:20,0.517304241657257)
--(axis cs:19,0.590207517147064)
--(axis cs:18,0.592429399490356)
--(axis cs:17,0.643261909484863)
--(axis cs:16,0.676230192184448)
--(axis cs:15,0.701371550559998)
--(axis cs:14,0.780378997325897)
--(axis cs:13,0.780378997325897)
--(axis cs:12,0.869368672370911)
--(axis cs:11,0.954290807247162)
--(axis cs:10,1.02326834201813)
--cycle;

\path [draw=color1, fill=color1, opacity=0.3]
(axis cs:10,1.02326841488654)
--(axis cs:10,0.826151056536423)
--(axis cs:11,0.82533599202006)
--(axis cs:12,0.823193404493012)
--(axis cs:13,0.696622444844033)
--(axis cs:14,0.670671764815904)
--(axis cs:15,0.574693546102886)
--(axis cs:16,0.50289858283744)
--(axis cs:17,0.502098333238599)
--(axis cs:18,0.482161206341879)
--(axis cs:19,0.466340728736823)
--(axis cs:20,0.465130049554325)
--(axis cs:21,0.453006216038728)
--(axis cs:22,0.412641909297478)
--(axis cs:23,0.316414296934632)
--(axis cs:24,0.295139595894653)
--(axis cs:25,0.293729850143461)
--(axis cs:26,0.239832687241923)
--(axis cs:27,0.185945633370012)
--(axis cs:28,0.185945633370012)
--(axis cs:29,0.185546412509806)
--(axis cs:30,0.161457883367811)
--(axis cs:31,0.155125623355411)
--(axis cs:32,0.151848663732972)
--(axis cs:33,0.126649837914775)
--(axis cs:34,0.111699784457498)
--(axis cs:35,0.105412063199825)
--(axis cs:36,0.102183056940334)
--(axis cs:37,0.102124471272996)
--(axis cs:38,0.102124471272996)
--(axis cs:39,0.101708211799884)
--(axis cs:40,0.101708211799884)
--(axis cs:41,0.101684316734088)
--(axis cs:42,0.101608625963967)
--(axis cs:43,0.101608625963967)
--(axis cs:44,0.101460051807739)
--(axis cs:45,0.101158972840762)
--(axis cs:46,0.06645258464715)
--(axis cs:47,0.0436031160165498)
--(axis cs:48,0.0436031160165498)
--(axis cs:49,0.0436031160165498)
--(axis cs:50,0.0430079393256916)
--(axis cs:51,0.0430079393256916)
--(axis cs:52,0.0430079393256916)
--(axis cs:53,0.0406631820313099)
--(axis cs:54,0.0406631820313099)
--(axis cs:55,0.0406579042804314)
--(axis cs:56,0.0404255193334093)
--(axis cs:57,0.0401072275541309)
--(axis cs:58,0.039839289320447)
--(axis cs:59,0.039839289320447)
--(axis cs:60,0.039839289320447)
--(axis cs:61,0.0394200426586012)
--(axis cs:62,0.0394200426586012)
--(axis cs:63,0.0394107779821601)
--(axis cs:64,0.0394107779821601)
--(axis cs:65,0.0394107779821601)
--(axis cs:66,0.0394107779821601)
--(axis cs:67,0.0394017724532608)
--(axis cs:68,0.0394017724532608)
--(axis cs:69,0.0394017724532608)
--(axis cs:70,0.0394017724532608)
--(axis cs:71,0.039388150873516)
--(axis cs:72,0.039388150873516)
--(axis cs:73,0.0392876160829674)
--(axis cs:74,0.0392876160829674)
--(axis cs:75,0.0392876160829674)
--(axis cs:76,0.0392708555803614)
--(axis cs:77,0.0392691581992718)
--(axis cs:78,0.0392691581992718)
--(axis cs:79,0.0392691581992718)
--(axis cs:80,0.0392691581992718)
--(axis cs:81,0.0392691581992718)
--(axis cs:82,0.0392299733662513)
--(axis cs:83,0.0392299733662513)
--(axis cs:84,0.0392299733662513)
--(axis cs:85,0.0392299733662513)
--(axis cs:86,0.0392299733662513)
--(axis cs:87,0.0392299733662513)
--(axis cs:88,0.0392299733662513)
--(axis cs:89,0.0392299733662513)
--(axis cs:90,0.0392299733662513)
--(axis cs:91,0.0392299733662513)
--(axis cs:92,0.0392299733662513)
--(axis cs:93,0.0392299733662513)
--(axis cs:94,0.0392299733662513)
--(axis cs:95,0.0392299733662513)
--(axis cs:96,0.0392278475836642)
--(axis cs:97,0.0392179696394978)
--(axis cs:98,0.0392179696394978)
--(axis cs:99,0.0392179696394978)
--(axis cs:100,0.0392179696394978)
--(axis cs:101,0.0392179696394978)
--(axis cs:102,0.0392169509824832)
--(axis cs:103,0.0392169509824832)
--(axis cs:104,0.0392169509824832)
--(axis cs:105,0.0392169509824832)
--(axis cs:106,0.0392169509824832)
--(axis cs:107,0.0392169509824832)
--(axis cs:108,0.0392169509824832)
--(axis cs:109,0.0392169509824832)
--(axis cs:110,0.0392169509824832)
--(axis cs:111,0.0392169509824832)
--(axis cs:111,0.195820016080402)
--(axis cs:111,0.195820016080402)
--(axis cs:110,0.195820016080402)
--(axis cs:109,0.195820016080402)
--(axis cs:108,0.195820016080402)
--(axis cs:107,0.195820016080402)
--(axis cs:106,0.195820016080402)
--(axis cs:105,0.195820016080402)
--(axis cs:104,0.195820016080402)
--(axis cs:103,0.195820016080402)
--(axis cs:102,0.195820016080402)
--(axis cs:101,0.195825108759389)
--(axis cs:100,0.195825108759389)
--(axis cs:99,0.195825108759389)
--(axis cs:98,0.195825108759389)
--(axis cs:97,0.195825108759389)
--(axis cs:96,0.195832165710337)
--(axis cs:95,0.195842794638566)
--(axis cs:94,0.195842794638566)
--(axis cs:93,0.195842794638566)
--(axis cs:92,0.195842794638566)
--(axis cs:91,0.195842794638566)
--(axis cs:90,0.195842794638566)
--(axis cs:89,0.195842794638566)
--(axis cs:88,0.195842794638566)
--(axis cs:87,0.195842794638566)
--(axis cs:86,0.195842794638566)
--(axis cs:85,0.195842794638566)
--(axis cs:84,0.195842794638566)
--(axis cs:83,0.195842794638566)
--(axis cs:82,0.195842794638566)
--(axis cs:81,0.195870790013772)
--(axis cs:80,0.195870790013772)
--(axis cs:79,0.195870790013772)
--(axis cs:78,0.195870790013772)
--(axis cs:77,0.195870790013772)
--(axis cs:76,0.195872003186881)
--(axis cs:75,0.195883984881215)
--(axis cs:74,0.195883984881215)
--(axis cs:73,0.195883984881215)
--(axis cs:72,0.19595584555821)
--(axis cs:71,0.19595584555821)
--(axis cs:70,0.195965576230362)
--(axis cs:69,0.195965576230362)
--(axis cs:68,0.195965576230362)
--(axis cs:67,0.195965576230362)
--(axis cs:66,0.195972004726401)
--(axis cs:65,0.195972004726401)
--(axis cs:64,0.195972004726401)
--(axis cs:63,0.195972004726401)
--(axis cs:62,0.195978619010471)
--(axis cs:61,0.195978619010471)
--(axis cs:60,0.196354041898542)
--(axis cs:59,0.196354041898542)
--(axis cs:58,0.196354041898542)
--(axis cs:57,0.196548022597308)
--(axis cs:56,0.196779953796971)
--(axis cs:55,0.196950322007033)
--(axis cs:54,0.196976742679608)
--(axis cs:53,0.196976742679608)
--(axis cs:52,0.19874216917158)
--(axis cs:51,0.19874216917158)
--(axis cs:50,0.19874216917158)
--(axis cs:49,0.199205747604574)
--(axis cs:48,0.199205747604574)
--(axis cs:47,0.199205747604574)
--(axis cs:46,0.223485933882706)
--(axis cs:45,0.331942547932426)
--(axis cs:44,0.334005479027199)
--(axis cs:43,0.334317407530061)
--(axis cs:42,0.334317407530061)
--(axis cs:41,0.334367046304684)
--(axis cs:40,0.334417326650459)
--(axis cs:39,0.334417326650459)
--(axis cs:38,0.335214676699842)
--(axis cs:37,0.335214676699842)
--(axis cs:36,0.335336874558747)
--(axis cs:35,0.34165821474894)
--(axis cs:34,0.345983295765506)
--(axis cs:33,0.362542486348126)
--(axis cs:32,0.409268654251465)
--(axis cs:31,0.420667127596148)
--(axis cs:30,0.432529582426661)
--(axis cs:29,0.454606106043925)
--(axis cs:28,0.454839862634122)
--(axis cs:27,0.454839862634122)
--(axis cs:26,0.559860763323411)
--(axis cs:25,0.602342215833925)
--(axis cs:24,0.603093864937826)
--(axis cs:23,0.62919396075264)
--(axis cs:22,0.713348344831903)
--(axis cs:21,0.73384635947404)
--(axis cs:20,0.740402240334247)
--(axis cs:19,0.740843459551043)
--(axis cs:18,0.758802192417656)
--(axis cs:17,0.804298303499913)
--(axis cs:16,0.804611165876437)
--(axis cs:15,0.872781588975372)
--(axis cs:14,0.943601342706241)
--(axis cs:13,0.95531989689661)
--(axis cs:12,1.01934627610597)
--(axis cs:11,1.02215975251834)
--(axis cs:10,1.02326841488654)
--cycle;

\path [draw=blue, fill=blue, opacity=0.3]
(axis cs:10,1.02326846122742)
--(axis cs:10,0.826151132583618)
--(axis cs:11,0.820872604846954)
--(axis cs:12,0.786018073558807)
--(axis cs:13,0.747940242290497)
--(axis cs:14,0.732710659503937)
--(axis cs:15,0.732710659503937)
--(axis cs:16,0.684009552001953)
--(axis cs:17,0.684009552001953)
--(axis cs:18,0.684009552001953)
--(axis cs:19,0.664320826530457)
--(axis cs:20,0.639858961105347)
--(axis cs:21,0.617985725402832)
--(axis cs:22,0.503011226654053)
--(axis cs:23,0.493634581565857)
--(axis cs:24,0.450355887413025)
--(axis cs:25,0.43842750787735)
--(axis cs:26,0.380042910575867)
--(axis cs:27,0.372763693332672)
--(axis cs:28,0.372751981019974)
--(axis cs:29,0.372751981019974)
--(axis cs:30,0.265868961811066)
--(axis cs:31,0.265868961811066)
--(axis cs:32,0.26573184132576)
--(axis cs:33,0.182801157236099)
--(axis cs:34,0.124844670295715)
--(axis cs:35,0.124844670295715)
--(axis cs:36,0.105535969138145)
--(axis cs:37,0.105535969138145)
--(axis cs:38,0.105535969138145)
--(axis cs:39,0.104695111513138)
--(axis cs:40,0.0882072597742081)
--(axis cs:41,0.0882072597742081)
--(axis cs:42,0.0882072597742081)
--(axis cs:43,0.0882072597742081)
--(axis cs:44,0.0626280009746552)
--(axis cs:45,0.0622498616576195)
--(axis cs:46,0.0622498616576195)
--(axis cs:47,0.0622498616576195)
--(axis cs:48,0.0597199276089668)
--(axis cs:49,0.0597199276089668)
--(axis cs:50,0.0597199276089668)
--(axis cs:51,0.0597199276089668)
--(axis cs:52,0.053096704185009)
--(axis cs:53,0.053096704185009)
--(axis cs:54,0.0487261153757572)
--(axis cs:55,0.0435517206788063)
--(axis cs:56,0.0346436910331249)
--(axis cs:57,0.0346436910331249)
--(axis cs:58,0.0312480162829161)
--(axis cs:59,0.0312480162829161)
--(axis cs:60,0.0282510109245777)
--(axis cs:61,0.0282510109245777)
--(axis cs:62,0.0281019546091557)
--(axis cs:63,0.0281019546091557)
--(axis cs:64,0.0281019546091557)
--(axis cs:65,0.0206168852746487)
--(axis cs:66,0.0159130282700062)
--(axis cs:67,0.0159130282700062)
--(axis cs:68,0.0159130282700062)
--(axis cs:69,0.0159130282700062)
--(axis cs:70,0.0159130282700062)
--(axis cs:71,0.0159130282700062)
--(axis cs:72,0.0144279291853309)
--(axis cs:73,0.0144279291853309)
--(axis cs:74,0.0144279291853309)
--(axis cs:75,0.0123523082584143)
--(axis cs:76,0.0123523082584143)
--(axis cs:77,0.0123523082584143)
--(axis cs:78,0.0123523082584143)
--(axis cs:79,0.0123523082584143)
--(axis cs:80,0.0117475371807814)
--(axis cs:81,0.0117475371807814)
--(axis cs:82,0.011656267568469)
--(axis cs:83,0.011656267568469)
--(axis cs:84,0.011656267568469)
--(axis cs:85,0.011656267568469)
--(axis cs:86,0.011656267568469)
--(axis cs:87,0.011656267568469)
--(axis cs:88,0.011656267568469)
--(axis cs:89,0.011656267568469)
--(axis cs:90,0.0105899721384048)
--(axis cs:91,0.00815887656062841)
--(axis cs:92,0.00815887656062841)
--(axis cs:93,0.00815887656062841)
--(axis cs:94,0.00815887656062841)
--(axis cs:95,0.00815887656062841)
--(axis cs:96,0.00815887656062841)
--(axis cs:97,0.00815887656062841)
--(axis cs:98,0.00815887656062841)
--(axis cs:99,0.00815887656062841)
--(axis cs:100,0.00774943176656961)
--(axis cs:101,0.00748166348785162)
--(axis cs:102,0.00748166348785162)
--(axis cs:103,0.00748166348785162)
--(axis cs:104,0.00748166348785162)
--(axis cs:105,0.00748166348785162)
--(axis cs:106,0.00748166348785162)
--(axis cs:107,0.00748166348785162)
--(axis cs:108,0.00748166348785162)
--(axis cs:109,0.00748166348785162)
--(axis cs:110,0.00748166348785162)
--(axis cs:111,0.00748166348785162)
--(axis cs:111,0.0150500359013677)
--(axis cs:111,0.0150500359013677)
--(axis cs:110,0.0150500359013677)
--(axis cs:109,0.0150500359013677)
--(axis cs:108,0.0150500359013677)
--(axis cs:107,0.0150500359013677)
--(axis cs:106,0.0150500359013677)
--(axis cs:105,0.0150500359013677)
--(axis cs:104,0.0150500359013677)
--(axis cs:103,0.0150500359013677)
--(axis cs:102,0.0150500359013677)
--(axis cs:101,0.0150500359013677)
--(axis cs:100,0.0166849680244923)
--(axis cs:99,0.0219245664775372)
--(axis cs:98,0.0219245664775372)
--(axis cs:97,0.0219245664775372)
--(axis cs:96,0.0219245664775372)
--(axis cs:95,0.0219245664775372)
--(axis cs:94,0.0219245664775372)
--(axis cs:93,0.0219245664775372)
--(axis cs:92,0.0219245664775372)
--(axis cs:91,0.0219245664775372)
--(axis cs:90,0.0240165330469608)
--(axis cs:89,0.0247230213135481)
--(axis cs:88,0.0247230213135481)
--(axis cs:87,0.0247230213135481)
--(axis cs:86,0.0247230213135481)
--(axis cs:85,0.0247230213135481)
--(axis cs:84,0.0247230213135481)
--(axis cs:83,0.0247230213135481)
--(axis cs:82,0.0247230213135481)
--(axis cs:81,0.0268788952380419)
--(axis cs:80,0.0268788952380419)
--(axis cs:79,0.0273740906268358)
--(axis cs:78,0.0273740906268358)
--(axis cs:77,0.0273740906268358)
--(axis cs:76,0.0273740906268358)
--(axis cs:75,0.0273740906268358)
--(axis cs:74,0.0297053530812263)
--(axis cs:73,0.0297053530812263)
--(axis cs:72,0.0297053530812263)
--(axis cs:71,0.0326932482421398)
--(axis cs:70,0.0326932482421398)
--(axis cs:69,0.0326932482421398)
--(axis cs:68,0.0326932482421398)
--(axis cs:67,0.0326932482421398)
--(axis cs:66,0.0326932482421398)
--(axis cs:65,0.0377894192934036)
--(axis cs:64,0.0548681877553463)
--(axis cs:63,0.0548681877553463)
--(axis cs:62,0.0548681877553463)
--(axis cs:61,0.0552549995481968)
--(axis cs:60,0.0552549995481968)
--(axis cs:59,0.0573530495166779)
--(axis cs:58,0.0573530495166779)
--(axis cs:57,0.0611470080912113)
--(axis cs:56,0.0611470080912113)
--(axis cs:55,0.0691997408866882)
--(axis cs:54,0.0749141573905945)
--(axis cs:53,0.0801870301365852)
--(axis cs:52,0.0801870301365852)
--(axis cs:51,0.0899118855595589)
--(axis cs:50,0.0899118855595589)
--(axis cs:49,0.0899118855595589)
--(axis cs:48,0.0899118855595589)
--(axis cs:47,0.0931477621197701)
--(axis cs:46,0.0931477621197701)
--(axis cs:45,0.0931477621197701)
--(axis cs:44,0.0941135883331299)
--(axis cs:43,0.123901948332787)
--(axis cs:42,0.123901948332787)
--(axis cs:41,0.123901948332787)
--(axis cs:40,0.123901948332787)
--(axis cs:39,0.159470528364182)
--(axis cs:38,0.160136327147484)
--(axis cs:37,0.160136327147484)
--(axis cs:36,0.160136327147484)
--(axis cs:35,0.210111051797867)
--(axis cs:34,0.210111051797867)
--(axis cs:33,0.292888820171356)
--(axis cs:32,0.417070955038071)
--(axis cs:31,0.417767226696014)
--(axis cs:30,0.417767226696014)
--(axis cs:29,0.559287250041962)
--(axis cs:28,0.559287250041962)
--(axis cs:27,0.559314489364624)
--(axis cs:26,0.567831158638)
--(axis cs:25,0.641737163066864)
--(axis cs:24,0.668629050254822)
--(axis cs:23,0.692384958267212)
--(axis cs:22,0.708930850028992)
--(axis cs:21,0.792099595069885)
--(axis cs:20,0.81207013130188)
--(axis cs:19,0.828676342964172)
--(axis cs:18,0.839565992355347)
--(axis cs:17,0.839565992355347)
--(axis cs:16,0.839565992355347)
--(axis cs:15,0.907949030399323)
--(axis cs:14,0.907949030399323)
--(axis cs:13,0.935862362384796)
--(axis cs:12,0.977770149707794)
--(axis cs:11,1.0111186504364)
--(axis cs:10,1.02326846122742)
--cycle;

\path [draw=color1, fill=color1, opacity=0.3]
(axis cs:10,1.02326841488654)
--(axis cs:10,0.826151056536423)
--(axis cs:11,0.803229154768156)
--(axis cs:12,0.711350827443832)
--(axis cs:13,0.49973731983671)
--(axis cs:14,0.49806554639554)
--(axis cs:15,0.490877507678634)
--(axis cs:16,0.436153763794818)
--(axis cs:17,0.409526076208102)
--(axis cs:18,0.409526076208102)
--(axis cs:19,0.346437415302914)
--(axis cs:20,0.296500523276755)
--(axis cs:21,0.279317855414899)
--(axis cs:22,0.267138462213209)
--(axis cs:23,0.200498062439389)
--(axis cs:24,0.118791727493836)
--(axis cs:25,0.118791727493836)
--(axis cs:26,0.102305050016016)
--(axis cs:27,0.0782025510488217)
--(axis cs:28,0.072792380263893)
--(axis cs:29,0.0627019135250071)
--(axis cs:30,0.0623545512822159)
--(axis cs:31,0.0486801635579538)
--(axis cs:32,0.0383026814604392)
--(axis cs:33,0.0383026814604392)
--(axis cs:34,0.0383026814604392)
--(axis cs:35,0.0360196088746233)
--(axis cs:36,0.0360196088746233)
--(axis cs:37,0.0360196088746233)
--(axis cs:38,0.035912580340032)
--(axis cs:39,0.0337705770073371)
--(axis cs:40,0.0337705770073371)
--(axis cs:41,0.0337705770073371)
--(axis cs:42,0.0337705770073371)
--(axis cs:43,0.0281211586921001)
--(axis cs:44,0.0281211586921001)
--(axis cs:45,0.0281211586921001)
--(axis cs:46,0.0281211586921001)
--(axis cs:47,0.0281211586921001)
--(axis cs:48,0.0281211586921001)
--(axis cs:49,0.0281211586921001)
--(axis cs:50,0.0253754721637053)
--(axis cs:51,0.0253754721637053)
--(axis cs:52,0.0253754721637053)
--(axis cs:53,0.018193893882941)
--(axis cs:54,0.0170336763854081)
--(axis cs:55,0.0167061280055009)
--(axis cs:56,0.0167061280055009)
--(axis cs:57,0.0167061280055009)
--(axis cs:58,0.0167061280055009)
--(axis cs:59,0.0167061280055009)
--(axis cs:60,0.0167061280055009)
--(axis cs:61,0.0167061280055009)
--(axis cs:62,0.0150347812282777)
--(axis cs:63,0.0150347812282777)
--(axis cs:64,0.0150347812282777)
--(axis cs:65,0.0150347812282777)
--(axis cs:66,0.0150347812282777)
--(axis cs:67,0.0150347812282777)
--(axis cs:68,0.0150347812282777)
--(axis cs:69,0.0150347812282777)
--(axis cs:70,0.0150347812282777)
--(axis cs:71,0.013517772014873)
--(axis cs:72,0.013517772014873)
--(axis cs:73,0.0131023521908315)
--(axis cs:74,0.0092010238363357)
--(axis cs:75,0.0092010238363357)
--(axis cs:76,0.0092010238363357)
--(axis cs:77,0.0092010238363357)
--(axis cs:78,0.0092010238363357)
--(axis cs:79,0.0092010238363357)
--(axis cs:80,0.0092010238363357)
--(axis cs:81,0.0092010238363357)
--(axis cs:82,0.0092010238363357)
--(axis cs:83,0.0092010238363357)
--(axis cs:84,0.0092010238363357)
--(axis cs:85,0.0092010238363357)
--(axis cs:86,0.0092010238363357)
--(axis cs:87,0.0092010238363357)
--(axis cs:88,0.0092010238363357)
--(axis cs:89,0.0092010238363357)
--(axis cs:90,0.0092010238363357)
--(axis cs:91,0.00920037261027785)
--(axis cs:92,0.00920037261027785)
--(axis cs:93,0.00920037261027785)
--(axis cs:94,0.00920037261027785)
--(axis cs:95,0.00920037261027785)
--(axis cs:96,0.00920037261027785)
--(axis cs:97,0.00920037261027785)
--(axis cs:98,0.00650928759040209)
--(axis cs:99,0.00650928759040209)
--(axis cs:100,0.00650928759040209)
--(axis cs:101,0.00650928759040209)
--(axis cs:102,0.00650928759040209)
--(axis cs:103,0.00650928759040209)
--(axis cs:104,0.00650928759040209)
--(axis cs:105,0.00650928759040209)
--(axis cs:106,0.00650928759040209)
--(axis cs:107,0.00650928759040209)
--(axis cs:108,0.0058083901505237)
--(axis cs:109,0.0058083901505237)
--(axis cs:110,0.0058083901505237)
--(axis cs:111,0.0058083901505237)
--(axis cs:111,0.0116231289318545)
--(axis cs:111,0.0116231289318545)
--(axis cs:110,0.0116231289318545)
--(axis cs:109,0.0116231289318545)
--(axis cs:108,0.0116231289318545)
--(axis cs:107,0.0125629697183285)
--(axis cs:106,0.0125629697183285)
--(axis cs:105,0.0125629697183285)
--(axis cs:104,0.0125629697183285)
--(axis cs:103,0.0125629697183285)
--(axis cs:102,0.0125629697183285)
--(axis cs:101,0.0125629697183285)
--(axis cs:100,0.0125629697183285)
--(axis cs:99,0.0125629697183285)
--(axis cs:98,0.0125629697183285)
--(axis cs:97,0.0170734975404408)
--(axis cs:96,0.0170734975404408)
--(axis cs:95,0.0170734975404408)
--(axis cs:94,0.0170734975404408)
--(axis cs:93,0.0170734975404408)
--(axis cs:92,0.0170734975404408)
--(axis cs:91,0.0170734975404408)
--(axis cs:90,0.017075549485401)
--(axis cs:89,0.017075549485401)
--(axis cs:88,0.017075549485401)
--(axis cs:87,0.017075549485401)
--(axis cs:86,0.017075549485401)
--(axis cs:85,0.017075549485401)
--(axis cs:84,0.017075549485401)
--(axis cs:83,0.017075549485401)
--(axis cs:82,0.017075549485401)
--(axis cs:81,0.017075549485401)
--(axis cs:80,0.017075549485401)
--(axis cs:79,0.017075549485401)
--(axis cs:78,0.017075549485401)
--(axis cs:77,0.017075549485401)
--(axis cs:76,0.017075549485401)
--(axis cs:75,0.017075549485401)
--(axis cs:74,0.017075549485401)
--(axis cs:73,0.0237101769983679)
--(axis cs:72,0.0247490439288469)
--(axis cs:71,0.0247490439288469)
--(axis cs:70,0.028641834818174)
--(axis cs:69,0.028641834818174)
--(axis cs:68,0.028641834818174)
--(axis cs:67,0.028641834818174)
--(axis cs:66,0.028641834818174)
--(axis cs:65,0.028641834818174)
--(axis cs:64,0.028641834818174)
--(axis cs:63,0.028641834818174)
--(axis cs:62,0.028641834818174)
--(axis cs:61,0.0302083003568798)
--(axis cs:60,0.0302083003568798)
--(axis cs:59,0.0302083003568798)
--(axis cs:58,0.0302083003568798)
--(axis cs:57,0.0302083003568798)
--(axis cs:56,0.0302083003568798)
--(axis cs:55,0.0302083003568798)
--(axis cs:54,0.0309082584419092)
--(axis cs:53,0.0315159628629312)
--(axis cs:52,0.154319675097641)
--(axis cs:51,0.154319675097641)
--(axis cs:50,0.154319675097641)
--(axis cs:49,0.156482396746116)
--(axis cs:48,0.156482396746116)
--(axis cs:47,0.156482396746116)
--(axis cs:46,0.156482396746116)
--(axis cs:45,0.156482396746116)
--(axis cs:44,0.156482396746116)
--(axis cs:43,0.156482396746116)
--(axis cs:42,0.16104544205048)
--(axis cs:41,0.16104544205048)
--(axis cs:40,0.16104544205048)
--(axis cs:39,0.16104544205048)
--(axis cs:38,0.162980588817327)
--(axis cs:37,0.20054714918882)
--(axis cs:36,0.20054714918882)
--(axis cs:35,0.20054714918882)
--(axis cs:34,0.202288388128612)
--(axis cs:33,0.202288388128612)
--(axis cs:32,0.202288388128612)
--(axis cs:31,0.212223801446308)
--(axis cs:30,0.230174907108409)
--(axis cs:29,0.23041966120762)
--(axis cs:28,0.238778703232423)
--(axis cs:27,0.242955954557889)
--(axis cs:26,0.266792625300074)
--(axis cs:25,0.279290317863083)
--(axis cs:24,0.279290317863083)
--(axis cs:23,0.417735049154163)
--(axis cs:22,0.491985543355787)
--(axis cs:21,0.504677925572954)
--(axis cs:20,0.52744006566507)
--(axis cs:19,0.59651695077238)
--(axis cs:18,0.671843232214411)
--(axis cs:17,0.671843232214411)
--(axis cs:16,0.692784394778065)
--(axis cs:15,0.732446667026033)
--(axis cs:14,0.751912853705367)
--(axis cs:13,0.752677561587024)
--(axis cs:12,0.922348877177511)
--(axis cs:11,0.985106260240562)
--(axis cs:10,1.02326841488654)
--cycle;

\addplot [semithick, blue, dash dot]
table {%
10 0.924709677696228
11 0.859240174293518
12 0.767309010028839
13 0.664568722248077
14 0.664568722248077
15 0.584935128688812
16 0.562910497188568
17 0.539563298225403
18 0.489402770996094
19 0.48781281709671
20 0.409158051013947
21 0.409158051013947
22 0.374577105045319
23 0.359849005937576
24 0.344238817691803
25 0.227477625012398
26 0.219160631299019
27 0.209190085530281
28 0.206460282206535
29 0.203763633966446
30 0.203518107533455
31 0.203518107533455
32 0.203518107533455
33 0.201875299215317
34 0.197367280721664
35 0.193739429116249
36 0.181392937898636
37 0.146125704050064
38 0.14496810734272
39 0.144115641713142
40 0.144115641713142
41 0.134042769670486
42 0.133609533309937
43 0.113192833960056
44 0.11289370059967
45 0.112033411860466
46 0.0839649066329002
47 0.0839649066329002
48 0.0819539427757263
49 0.0819539427757263
50 0.0819539427757263
51 0.0819539427757263
52 0.0791330561041832
53 0.0791330561041832
54 0.0589753389358521
55 0.0581494681537151
56 0.0323405973613262
57 0.0274503938853741
58 0.0273857712745667
59 0.0273857712745667
60 0.0147712826728821
61 0.00426961202174425
62 0.00426961202174425
63 0.00426961202174425
64 0.00426961202174425
65 0.00426961202174425
66 0.00426961202174425
67 0.00426961202174425
68 0.0041701914742589
69 0.0041701914742589
70 0.0041701914742589
71 0.0041701914742589
72 0.0041701914742589
73 0.0041701914742589
74 0.00392073392868042
75 0.00392073392868042
76 0.00392073392868042
77 0.00378305907361209
78 0.00378305907361209
79 0.00378305907361209
80 0.00378305907361209
81 0.00378305907361209
82 0.00378305907361209
83 0.00378305907361209
84 0.00378305907361209
85 0.00378305907361209
86 0.00378305907361209
87 0.00378305907361209
88 0.00378305907361209
89 0.00378305907361209
90 0.00378305907361209
91 0.00378305907361209
92 0.00378305907361209
93 0.00378305907361209
94 0.00350729236379266
95 0.00350729236379266
96 0.00350729236379266
97 0.00350729236379266
98 0.00350729236379266
99 0.00350729236379266
100 0.00350729236379266
101 0.00350729236379266
102 0.00350729236379266
103 0.00350729236379266
104 0.00350729236379266
105 0.00350729236379266
106 0.00350729236379266
107 0.00350729236379266
108 0.00350729236379266
109 0.00350729236379266
110 0.00350729236379266
111 0.00350729236379266
};
\addplot [semithick, color1]
table {%
10 0.924709735711479
11 0.923747872269201
12 0.92126984029949
13 0.825971170870321
14 0.807136553761072
15 0.723737567539129
16 0.653754874356939
17 0.653198318369256
18 0.620481699379767
19 0.603592094143933
20 0.602766144944286
21 0.593426287756384
22 0.56299512706469
23 0.472804128843636
24 0.449116730416239
25 0.448036032988693
26 0.399846725282667
27 0.320392748002067
28 0.320392748002067
29 0.320076259276865
30 0.296993732897236
31 0.287896375475779
32 0.280558658992219
33 0.244596162131451
34 0.228841540111502
35 0.223535138974382
36 0.218759965749541
37 0.218669573986419
38 0.218669573986419
39 0.218062769225171
40 0.218062769225171
41 0.218025681519386
42 0.217963016747014
43 0.217963016747014
44 0.217732765417469
45 0.216550760386594
46 0.144969259264928
47 0.121404431810562
48 0.121404431810562
49 0.121404431810562
50 0.120875054248636
51 0.120875054248636
52 0.120875054248636
53 0.118819962355459
54 0.118819962355459
55 0.118804113143732
56 0.11860273656519
57 0.11832762507572
58 0.118096665609494
59 0.118096665609494
60 0.118096665609494
61 0.117699330834536
62 0.117699330834536
63 0.11769139135428
64 0.11769139135428
65 0.11769139135428
66 0.11769139135428
67 0.117683674341811
68 0.117683674341811
69 0.117683674341811
70 0.117683674341811
71 0.117671998215863
72 0.117671998215863
73 0.117585800482091
74 0.117585800482091
75 0.117585800482091
76 0.117571429383621
77 0.117569974106522
78 0.117569974106522
79 0.117569974106522
80 0.117569974106522
81 0.117569974106522
82 0.117536384002409
83 0.117536384002409
84 0.117536384002409
85 0.117536384002409
86 0.117536384002409
87 0.117536384002409
88 0.117536384002409
89 0.117536384002409
90 0.117536384002409
91 0.117536384002409
92 0.117536384002409
93 0.117536384002409
94 0.117536384002409
95 0.117536384002409
96 0.117530006647001
97 0.117521539199444
98 0.117521539199444
99 0.117521539199444
100 0.117521539199444
101 0.117521539199444
102 0.117518483531442
103 0.117518483531442
104 0.117518483531442
105 0.117518483531442
106 0.117518483531442
107 0.117518483531442
108 0.117518483531442
109 0.117518483531442
110 0.117518483531442
111 0.117518483531442
};
\addplot [semithick, blue]
table {%
10 0.924709796905518
11 0.915995597839355
12 0.881894111633301
13 0.841901302337646
14 0.82032984495163
15 0.82032984495163
16 0.76178777217865
17 0.76178777217865
18 0.76178777217865
19 0.746498584747314
20 0.725964546203613
21 0.705042660236359
22 0.605971038341522
23 0.593009769916534
24 0.559492468833923
25 0.540082335472107
26 0.473937034606934
27 0.466039091348648
28 0.466019630432129
29 0.466019630432129
30 0.34181809425354
31 0.34181809425354
32 0.341401398181915
33 0.237844988703728
34 0.167477861046791
35 0.167477861046791
36 0.132836148142815
37 0.132836148142815
38 0.132836148142815
39 0.13208281993866
40 0.106054604053497
41 0.106054604053497
42 0.106054604053497
43 0.106054604053497
44 0.0783707946538925
45 0.0776988118886948
46 0.0776988118886948
47 0.0776988118886948
48 0.0748159065842628
49 0.0748159065842628
50 0.0748159065842628
51 0.0748159065842628
52 0.0666418671607971
53 0.0666418671607971
54 0.061820138245821
55 0.0563757307827473
56 0.0478953495621681
57 0.0478953495621681
58 0.0443005338311195
59 0.0443005338311195
60 0.0417530052363873
61 0.0417530052363873
62 0.041485071182251
63 0.041485071182251
64 0.041485071182251
65 0.0292031522840261
66 0.024303138256073
67 0.024303138256073
68 0.024303138256073
69 0.024303138256073
70 0.024303138256073
71 0.024303138256073
72 0.0220666415989399
73 0.0220666415989399
74 0.0220666415989399
75 0.019863199442625
76 0.019863199442625
77 0.019863199442625
78 0.019863199442625
79 0.019863199442625
80 0.0193132162094116
81 0.0193132162094116
82 0.0181896444410086
83 0.0181896444410086
84 0.0181896444410086
85 0.0181896444410086
86 0.0181896444410086
87 0.0181896444410086
88 0.0181896444410086
89 0.0181896444410086
90 0.0173032525926828
91 0.0150417210534215
92 0.0150417210534215
93 0.0150417210534215
94 0.0150417210534215
95 0.0150417210534215
96 0.0150417210534215
97 0.0150417210534215
98 0.0150417210534215
99 0.0150417210534215
100 0.0122171994298697
101 0.0112658496946096
102 0.0112658496946096
103 0.0112658496946096
104 0.0112658496946096
105 0.0112658496946096
106 0.0112658496946096
107 0.0112658496946096
108 0.0112658496946096
109 0.0112658496946096
110 0.0112658496946096
111 0.0112658496946096
};
\addplot [semithick, color1, dash dot]
table {%
10 0.924709735711479
11 0.894167707504359
12 0.816849852310672
13 0.626207440711867
14 0.624989200050453
15 0.611662087352334
16 0.564469079286441
17 0.540684654211257
18 0.540684654211257
19 0.471477183037647
20 0.411970294470912
21 0.391997890493927
22 0.379562002784498
23 0.309116555796776
24 0.19904102267846
25 0.19904102267846
26 0.184548837658045
27 0.160579252803356
28 0.155785541748158
29 0.146560787366314
30 0.146264729195312
31 0.130451982502131
32 0.120295534794525
33 0.120295534794525
34 0.120295534794525
35 0.118283379031722
36 0.118283379031722
37 0.118283379031722
38 0.0994465845786793
39 0.0974080095289085
40 0.0974080095289085
41 0.0974080095289085
42 0.0974080095289085
43 0.0923017777191081
44 0.0923017777191081
45 0.0923017777191081
46 0.0923017777191081
47 0.0923017777191081
48 0.0923017777191081
49 0.0923017777191081
50 0.0898475736306733
51 0.0898475736306733
52 0.0898475736306733
53 0.0248549283729361
54 0.0239709674136587
55 0.0234572141811903
56 0.0234572141811903
57 0.0234572141811903
58 0.0234572141811903
59 0.0234572141811903
60 0.0234572141811903
61 0.0234572141811903
62 0.0218383080232259
63 0.0218383080232259
64 0.0218383080232259
65 0.0218383080232259
66 0.0218383080232259
67 0.0218383080232259
68 0.0218383080232259
69 0.0218383080232259
70 0.0218383080232259
71 0.0191334079718599
72 0.0191334079718599
73 0.0184062645945997
74 0.0131382866608684
75 0.0131382866608684
76 0.0131382866608684
77 0.0131382866608684
78 0.0131382866608684
79 0.0131382866608684
80 0.0131382866608684
81 0.0131382866608684
82 0.0131382866608684
83 0.0131382866608684
84 0.0131382866608684
85 0.0131382866608684
86 0.0131382866608684
87 0.0131382866608684
88 0.0131382866608684
89 0.0131382866608684
90 0.0131382866608684
91 0.0131369350753593
92 0.0131369350753593
93 0.0131369350753593
94 0.0131369350753593
95 0.0131369350753593
96 0.0131369350753593
97 0.0131369350753593
98 0.00953612865436531
99 0.00953612865436531
100 0.00953612865436531
101 0.00953612865436531
102 0.00953612865436531
103 0.00953612865436531
104 0.00953612865436531
105 0.00953612865436531
106 0.00953612865436531
107 0.00953612865436531
108 0.00871575954118911
109 0.00871575954118911
110 0.00871575954118911
111 0.00871575954118911
};
\end{axis}

\end{tikzpicture}

%% file: figures/bop_2d_michalewicz_strong.tex
\begin{tikzpicture}

\definecolor{color0}{rgb}{0,0,1}
\definecolor{color1}{rgb}{1,0.549019607843137,0}
\definecolor{color2}{rgb}{1,0.647058823529412,0}
\definecolor{color3}{rgb}{0.564705882352941,0.933333333333333,0.564705882352941}

\begin{axis}[axis on top,
enlarge x limits=false,
enlarge y limits=false,
height=\figureheight,
scale only axis,
tick align=outside,
tick pos=left,
tick pos=left,
width=\figurewidth,
xmin=10, xmax=100,
xtick style={color=black},
xtick={-10,0,10,30,50,70,90},
xticklabels={\ensuremath{-}10,0,10,30,50,70,90},
ymin=-0.05, ymax=1.2,
ytick={0.   , 1.2},
yticklabels={},
]
\node[anchor=north east] at (rel axis cs:1,1) {Michalewicz 2D (strong)};
\path [draw=blue, fill=blue, opacity=0.3]
(axis cs:10,1.02326834201813)
--(axis cs:10,0.826151013374329)
--(axis cs:11,0.735370635986328)
--(axis cs:12,0.522100925445557)
--(axis cs:13,0.444995254278183)
--(axis cs:14,0.415760099887848)
--(axis cs:15,0.387576878070831)
--(axis cs:16,0.373676329851151)
--(axis cs:17,0.312079906463623)
--(axis cs:18,0.259039640426636)
--(axis cs:19,0.254735052585602)
--(axis cs:20,0.211366236209869)
--(axis cs:21,0.202814549207687)
--(axis cs:22,0.194872409105301)
--(axis cs:23,0.131220191717148)
--(axis cs:24,0.131220191717148)
--(axis cs:25,0.114789195358753)
--(axis cs:26,0.0892666280269623)
--(axis cs:27,0.0416098944842815)
--(axis cs:28,0.0358860194683075)
--(axis cs:29,0.0275784526020288)
--(axis cs:30,0.0251718927174807)
--(axis cs:31,0.0251718927174807)
--(axis cs:32,0.0174946710467339)
--(axis cs:33,0.0174946710467339)
--(axis cs:34,0.0110717266798019)
--(axis cs:35,0.00821941718459129)
--(axis cs:36,0.00821941718459129)
--(axis cs:37,0.00821941718459129)
--(axis cs:38,0.00821941718459129)
--(axis cs:39,0.00821941718459129)
--(axis cs:40,0.00821941718459129)
--(axis cs:41,0.00821941718459129)
--(axis cs:42,0.00821941718459129)
--(axis cs:43,0.00645783077925444)
--(axis cs:44,0.00645783077925444)
--(axis cs:45,0.00645783077925444)
--(axis cs:46,0.00612376723438501)
--(axis cs:47,0.00612376723438501)
--(axis cs:48,0.00556969828903675)
--(axis cs:49,0.00556969828903675)
--(axis cs:50,0.00555101409554482)
--(axis cs:51,0.00555101409554482)
--(axis cs:52,0.00555101409554482)
--(axis cs:53,0.00555101409554482)
--(axis cs:54,0.00478125736117363)
--(axis cs:55,0.00478125736117363)
--(axis cs:56,0.00425870250910521)
--(axis cs:57,0.00425870250910521)
--(axis cs:58,0.00425870250910521)
--(axis cs:59,0.00425870250910521)
--(axis cs:60,0.00425870250910521)
--(axis cs:61,0.00371289229951799)
--(axis cs:62,0.00371289229951799)
--(axis cs:63,0.00371289229951799)
--(axis cs:64,0.00371289229951799)
--(axis cs:65,0.00371289229951799)
--(axis cs:66,0.00371289229951799)
--(axis cs:67,0.00358923617750406)
--(axis cs:68,0.00356340315192938)
--(axis cs:69,0.00356340315192938)
--(axis cs:70,0.00356340315192938)
--(axis cs:71,0.00356340315192938)
--(axis cs:72,0.00356340315192938)
--(axis cs:73,0.00356340315192938)
--(axis cs:74,0.00356340315192938)
--(axis cs:75,0.00356340315192938)
--(axis cs:76,0.00325535284355283)
--(axis cs:77,0.00325535284355283)
--(axis cs:78,0.00325535284355283)
--(axis cs:79,0.00325535284355283)
--(axis cs:80,0.00325535284355283)
--(axis cs:81,0.00325535284355283)
--(axis cs:82,0.0028016516007483)
--(axis cs:83,0.0028016516007483)
--(axis cs:84,0.0028016516007483)
--(axis cs:85,0.0028016516007483)
--(axis cs:86,0.0028016516007483)
--(axis cs:87,0.0028016516007483)
--(axis cs:88,0.0028016516007483)
--(axis cs:89,0.0028016516007483)
--(axis cs:90,0.0028016516007483)
--(axis cs:91,0.0028016516007483)
--(axis cs:92,0.0028016516007483)
--(axis cs:93,0.0028016516007483)
--(axis cs:94,0.0028016516007483)
--(axis cs:95,0.0028016516007483)
--(axis cs:96,0.0028016516007483)
--(axis cs:97,0.0028016516007483)
--(axis cs:98,0.0028016516007483)
--(axis cs:99,0.0028016516007483)
--(axis cs:100,0.0028016516007483)
--(axis cs:101,0.0028016516007483)
--(axis cs:102,0.0028016516007483)
--(axis cs:103,0.0028016516007483)
--(axis cs:104,0.0026522025000304)
--(axis cs:105,0.0026522025000304)
--(axis cs:106,0.0026522025000304)
--(axis cs:107,0.0026522025000304)
--(axis cs:108,0.0026522025000304)
--(axis cs:109,0.0026522025000304)
--(axis cs:110,0.0026522025000304)
--(axis cs:111,0.0026522025000304)
--(axis cs:111,0.00408331304788589)
--(axis cs:111,0.00408331304788589)
--(axis cs:110,0.00408331304788589)
--(axis cs:109,0.00408331304788589)
--(axis cs:108,0.00408331304788589)
--(axis cs:107,0.00408331304788589)
--(axis cs:106,0.00408331304788589)
--(axis cs:105,0.00408331304788589)
--(axis cs:104,0.00408331304788589)
--(axis cs:103,0.00437105214223266)
--(axis cs:102,0.00437105214223266)
--(axis cs:101,0.00437105214223266)
--(axis cs:100,0.00437105214223266)
--(axis cs:99,0.00437105214223266)
--(axis cs:98,0.00437105214223266)
--(axis cs:97,0.00437105214223266)
--(axis cs:96,0.00437105214223266)
--(axis cs:95,0.00437105214223266)
--(axis cs:94,0.00437105214223266)
--(axis cs:93,0.00437105214223266)
--(axis cs:92,0.00437105214223266)
--(axis cs:91,0.00437105214223266)
--(axis cs:90,0.00437105214223266)
--(axis cs:89,0.00437105214223266)
--(axis cs:88,0.00437105214223266)
--(axis cs:87,0.00437105214223266)
--(axis cs:86,0.00437105214223266)
--(axis cs:85,0.00437105214223266)
--(axis cs:84,0.00437105214223266)
--(axis cs:83,0.00437105214223266)
--(axis cs:82,0.00437105214223266)
--(axis cs:81,0.00470372708514333)
--(axis cs:80,0.00470372708514333)
--(axis cs:79,0.00470372708514333)
--(axis cs:78,0.00470372708514333)
--(axis cs:77,0.00470372708514333)
--(axis cs:76,0.00470372708514333)
--(axis cs:75,0.00509171560406685)
--(axis cs:74,0.00509171560406685)
--(axis cs:73,0.00509171560406685)
--(axis cs:72,0.00509171560406685)
--(axis cs:71,0.00509171560406685)
--(axis cs:70,0.00509171560406685)
--(axis cs:69,0.00509171560406685)
--(axis cs:68,0.00509171560406685)
--(axis cs:67,0.00509942788630724)
--(axis cs:66,0.00544030731543899)
--(axis cs:65,0.00544030731543899)
--(axis cs:64,0.00544030731543899)
--(axis cs:63,0.00544030731543899)
--(axis cs:62,0.00544030731543899)
--(axis cs:61,0.00544030731543899)
--(axis cs:60,0.00647087302058935)
--(axis cs:59,0.00647087302058935)
--(axis cs:58,0.00647087302058935)
--(axis cs:57,0.00647087302058935)
--(axis cs:56,0.00647087302058935)
--(axis cs:55,0.00775681808590889)
--(axis cs:54,0.00775681808590889)
--(axis cs:53,0.00830294098705053)
--(axis cs:52,0.00830294098705053)
--(axis cs:51,0.00830294098705053)
--(axis cs:50,0.00830294098705053)
--(axis cs:49,0.00840279832482338)
--(axis cs:48,0.00840279832482338)
--(axis cs:47,0.00983542483299971)
--(axis cs:46,0.00983542483299971)
--(axis cs:45,0.0101793510839343)
--(axis cs:44,0.0101793510839343)
--(axis cs:43,0.0101793510839343)
--(axis cs:42,0.0153118055313826)
--(axis cs:41,0.0153118055313826)
--(axis cs:40,0.0153118055313826)
--(axis cs:39,0.0153118055313826)
--(axis cs:38,0.0153118055313826)
--(axis cs:37,0.0153118055313826)
--(axis cs:36,0.0153118055313826)
--(axis cs:35,0.0153118055313826)
--(axis cs:34,0.0234413892030716)
--(axis cs:33,0.0431935638189316)
--(axis cs:32,0.0431935638189316)
--(axis cs:31,0.0491317585110664)
--(axis cs:30,0.0491317585110664)
--(axis cs:29,0.0509721040725708)
--(axis cs:28,0.0651466175913811)
--(axis cs:27,0.11653645336628)
--(axis cs:26,0.207230240106583)
--(axis cs:25,0.262722164392471)
--(axis cs:24,0.276336491107941)
--(axis cs:23,0.276336491107941)
--(axis cs:22,0.377256244421005)
--(axis cs:21,0.386878997087479)
--(axis cs:20,0.391626119613647)
--(axis cs:19,0.429121255874634)
--(axis cs:18,0.437309682369232)
--(axis cs:17,0.48980987071991)
--(axis cs:16,0.566411554813385)
--(axis cs:15,0.586630821228027)
--(axis cs:14,0.628573954105377)
--(axis cs:13,0.657482504844666)
--(axis cs:12,0.766202688217163)
--(axis cs:11,0.913545370101929)
--(axis cs:10,1.02326834201813)
--cycle;

\path [draw=color1, fill=color1, opacity=0.3]
(axis cs:10,1.02326841488654)
--(axis cs:10,0.826151056536423)
--(axis cs:11,0.82533599202006)
--(axis cs:12,0.823193404493012)
--(axis cs:13,0.696622444844033)
--(axis cs:14,0.670671764815904)
--(axis cs:15,0.574693546102886)
--(axis cs:16,0.50289858283744)
--(axis cs:17,0.501971195749866)
--(axis cs:18,0.482039250414511)
--(axis cs:19,0.46624246064019)
--(axis cs:20,0.465031807234493)
--(axis cs:21,0.452908467282446)
--(axis cs:22,0.412545522538771)
--(axis cs:23,0.322234092734231)
--(axis cs:24,0.321874228788074)
--(axis cs:25,0.320449464969671)
--(axis cs:26,0.312573600777972)
--(axis cs:27,0.25389657172811)
--(axis cs:28,0.204484101322261)
--(axis cs:29,0.186267812975323)
--(axis cs:30,0.170794099299941)
--(axis cs:31,0.164437430660705)
--(axis cs:32,0.161163166726154)
--(axis cs:33,0.136067487799923)
--(axis cs:34,0.119759027501612)
--(axis cs:35,0.105902290556476)
--(axis cs:36,0.102672559541435)
--(axis cs:37,0.102613961796562)
--(axis cs:38,0.102613961796562)
--(axis cs:39,0.101986791876263)
--(axis cs:40,0.101986791876263)
--(axis cs:41,0.101852290254861)
--(axis cs:42,0.101776587925626)
--(axis cs:43,0.101776587925626)
--(axis cs:44,0.101529387895515)
--(axis cs:45,0.101228334640013)
--(axis cs:46,0.0665217443445013)
--(axis cs:47,0.043670451446439)
--(axis cs:48,0.043670451446439)
--(axis cs:49,0.043670451446439)
--(axis cs:50,0.0430752229950804)
--(axis cs:51,0.0430752229950804)
--(axis cs:52,0.0430752229950804)
--(axis cs:53,0.0407302613513408)
--(axis cs:54,0.0406895298552146)
--(axis cs:55,0.0406842521106465)
--(axis cs:56,0.0404518592023036)
--(axis cs:57,0.0401335565172642)
--(axis cs:58,0.0398656091022232)
--(axis cs:59,0.0398656091022232)
--(axis cs:60,0.0398656091022232)
--(axis cs:61,0.0394265412099353)
--(axis cs:62,0.0394265412099353)
--(axis cs:63,0.0394172764549989)
--(axis cs:64,0.0394172764549989)
--(axis cs:65,0.0394172764549989)
--(axis cs:66,0.0394172764549989)
--(axis cs:67,0.0394017724532608)
--(axis cs:68,0.0394017724532608)
--(axis cs:69,0.0394017724532608)
--(axis cs:70,0.0394017724532608)
--(axis cs:71,0.039388150873516)
--(axis cs:72,0.039388150873516)
--(axis cs:73,0.0392876160829674)
--(axis cs:74,0.0392876160829674)
--(axis cs:75,0.0392876160829674)
--(axis cs:76,0.0392708555803614)
--(axis cs:77,0.0392691581992718)
--(axis cs:78,0.0392691581992718)
--(axis cs:79,0.0392691581992718)
--(axis cs:80,0.0392691581992718)
--(axis cs:81,0.0392691581992718)
--(axis cs:82,0.0392299733662513)
--(axis cs:83,0.0392299733662513)
--(axis cs:84,0.0392299733662513)
--(axis cs:85,0.0392299733662513)
--(axis cs:86,0.0392299733662513)
--(axis cs:87,0.0392299733662513)
--(axis cs:88,0.0392299733662513)
--(axis cs:89,0.0392299733662513)
--(axis cs:90,0.0392299733662513)
--(axis cs:91,0.0392299733662513)
--(axis cs:92,0.0392299733662513)
--(axis cs:93,0.0392299733662513)
--(axis cs:94,0.0392299733662513)
--(axis cs:95,0.0392299733662513)
--(axis cs:96,0.0392278475836642)
--(axis cs:97,0.0392179696394978)
--(axis cs:98,0.0392179696394978)
--(axis cs:99,0.0392179696394978)
--(axis cs:100,0.0392179696394978)
--(axis cs:101,0.0392179696394978)
--(axis cs:102,0.0392169509824832)
--(axis cs:103,0.0392169509824832)
--(axis cs:104,0.0392169509824832)
--(axis cs:105,0.0392169509824832)
--(axis cs:106,0.0392169509824832)
--(axis cs:107,0.0392169509824832)
--(axis cs:108,0.0392169509824832)
--(axis cs:109,0.0392169509824832)
--(axis cs:110,0.0392169509824832)
--(axis cs:111,0.0392169509824832)
--(axis cs:111,0.195820016080402)
--(axis cs:111,0.195820016080402)
--(axis cs:110,0.195820016080402)
--(axis cs:109,0.195820016080402)
--(axis cs:108,0.195820016080402)
--(axis cs:107,0.195820016080402)
--(axis cs:106,0.195820016080402)
--(axis cs:105,0.195820016080402)
--(axis cs:104,0.195820016080402)
--(axis cs:103,0.195820016080402)
--(axis cs:102,0.195820016080402)
--(axis cs:101,0.195825108759389)
--(axis cs:100,0.195825108759389)
--(axis cs:99,0.195825108759389)
--(axis cs:98,0.195825108759389)
--(axis cs:97,0.195825108759389)
--(axis cs:96,0.195832165710337)
--(axis cs:95,0.195842794638566)
--(axis cs:94,0.195842794638566)
--(axis cs:93,0.195842794638566)
--(axis cs:92,0.195842794638566)
--(axis cs:91,0.195842794638566)
--(axis cs:90,0.195842794638566)
--(axis cs:89,0.195842794638566)
--(axis cs:88,0.195842794638566)
--(axis cs:87,0.195842794638566)
--(axis cs:86,0.195842794638566)
--(axis cs:85,0.195842794638566)
--(axis cs:84,0.195842794638566)
--(axis cs:83,0.195842794638566)
--(axis cs:82,0.195842794638566)
--(axis cs:81,0.195870790013772)
--(axis cs:80,0.195870790013772)
--(axis cs:79,0.195870790013772)
--(axis cs:78,0.195870790013772)
--(axis cs:77,0.195870790013772)
--(axis cs:76,0.195872003186881)
--(axis cs:75,0.195883984881215)
--(axis cs:74,0.195883984881215)
--(axis cs:73,0.195883984881215)
--(axis cs:72,0.19595584555821)
--(axis cs:71,0.19595584555821)
--(axis cs:70,0.195965576230362)
--(axis cs:69,0.195965576230362)
--(axis cs:68,0.195965576230362)
--(axis cs:67,0.195965576230362)
--(axis cs:66,0.195976644375913)
--(axis cs:65,0.195976644375913)
--(axis cs:64,0.195976644375913)
--(axis cs:63,0.195976644375913)
--(axis cs:62,0.195983258581489)
--(axis cs:61,0.195983258581489)
--(axis cs:60,0.19637281309282)
--(axis cs:59,0.19637281309282)
--(axis cs:58,0.19637281309282)
--(axis cs:57,0.196566784610229)
--(axis cs:56,0.196798704904131)
--(axis cs:55,0.196969065152873)
--(axis cs:54,0.196995485831758)
--(axis cs:53,0.197024482444532)
--(axis cs:52,0.198789704587145)
--(axis cs:51,0.198789704587145)
--(axis cs:50,0.198789704587145)
--(axis cs:49,0.199253231259639)
--(axis cs:48,0.199253231259639)
--(axis cs:47,0.199253231259639)
--(axis cs:46,0.22353159327031)
--(axis cs:45,0.33198800521813)
--(axis cs:44,0.334050962024378)
--(axis cs:43,0.334427647332049)
--(axis cs:42,0.334427647332049)
--(axis cs:41,0.334477274547559)
--(axis cs:40,0.334600277139847)
--(axis cs:39,0.334600277139847)
--(axis cs:38,0.335536458987487)
--(axis cs:37,0.335536458987487)
--(axis cs:36,0.335658644768857)
--(axis cs:35,0.341979260203499)
--(axis cs:34,0.351479480339275)
--(axis cs:33,0.368842777045814)
--(axis cs:32,0.415323291008571)
--(axis cs:31,0.426724460041141)
--(axis cs:30,0.438562506244818)
--(axis cs:29,0.455084847667059)
--(axis cs:28,0.468659578619467)
--(axis cs:27,0.532116472254086)
--(axis cs:26,0.632623417916432)
--(axis cs:25,0.637063102731254)
--(axis cs:24,0.637799733767944)
--(axis cs:23,0.637980318511254)
--(axis cs:22,0.713210845751077)
--(axis cs:21,0.733710222390791)
--(axis cs:20,0.740266596814548)
--(axis cs:19,0.740707841808144)
--(axis cs:18,0.75863887003754)
--(axis cs:17,0.804140162681163)
--(axis cs:16,0.804611165876437)
--(axis cs:15,0.872781588975372)
--(axis cs:14,0.943601342706241)
--(axis cs:13,0.95531989689661)
--(axis cs:12,1.01934627610597)
--(axis cs:11,1.02215975251834)
--(axis cs:10,1.02326841488654)
--cycle;

\path [draw=blue, fill=blue, opacity=0.3]
(axis cs:10,1.02326846122742)
--(axis cs:10,0.826151132583618)
--(axis cs:11,0.820872604846954)
--(axis cs:12,0.786018073558807)
--(axis cs:13,0.747940242290497)
--(axis cs:14,0.732710659503937)
--(axis cs:15,0.732710659503937)
--(axis cs:16,0.684009552001953)
--(axis cs:17,0.684009552001953)
--(axis cs:18,0.615689218044281)
--(axis cs:19,0.59761780500412)
--(axis cs:20,0.574923634529114)
--(axis cs:21,0.554446876049042)
--(axis cs:22,0.409736424684525)
--(axis cs:23,0.400960087776184)
--(axis cs:24,0.361194223165512)
--(axis cs:25,0.350204914808273)
--(axis cs:26,0.297081768512726)
--(axis cs:27,0.29051485657692)
--(axis cs:28,0.29051485657692)
--(axis cs:29,0.29051485657692)
--(axis cs:30,0.198671385645866)
--(axis cs:31,0.198671385645866)
--(axis cs:32,0.198671385645866)
--(axis cs:33,0.16856524348259)
--(axis cs:34,0.111416555941105)
--(axis cs:35,0.111416555941105)
--(axis cs:36,0.0920018330216408)
--(axis cs:37,0.0920018330216408)
--(axis cs:38,0.0920018330216408)
--(axis cs:39,0.0920018330216408)
--(axis cs:40,0.076359435915947)
--(axis cs:41,0.076359435915947)
--(axis cs:42,0.076359435915947)
--(axis cs:43,0.076359435915947)
--(axis cs:44,0.0519911050796509)
--(axis cs:45,0.0516244471073151)
--(axis cs:46,0.0516244471073151)
--(axis cs:47,0.0516244471073151)
--(axis cs:48,0.0492338575422764)
--(axis cs:49,0.0492338575422764)
--(axis cs:50,0.0492338575422764)
--(axis cs:51,0.0492338575422764)
--(axis cs:52,0.0430090203881264)
--(axis cs:53,0.0430090203881264)
--(axis cs:54,0.0389484763145447)
--(axis cs:55,0.0389484763145447)
--(axis cs:56,0.0321785695850849)
--(axis cs:57,0.0321785695850849)
--(axis cs:58,0.0288308970630169)
--(axis cs:59,0.0288308970630169)
--(axis cs:60,0.0258888676762581)
--(axis cs:61,0.0258888676762581)
--(axis cs:62,0.0257400311529636)
--(axis cs:63,0.0257400311529636)
--(axis cs:64,0.0257400311529636)
--(axis cs:65,0.018303319811821)
--(axis cs:66,0.0137093588709831)
--(axis cs:67,0.0137093588709831)
--(axis cs:68,0.0137093588709831)
--(axis cs:69,0.0137093588709831)
--(axis cs:70,0.0137093588709831)
--(axis cs:71,0.0137093588709831)
--(axis cs:72,0.0122562842443585)
--(axis cs:73,0.0122562842443585)
--(axis cs:74,0.0122562842443585)
--(axis cs:75,0.0102367009967566)
--(axis cs:76,0.0102367009967566)
--(axis cs:77,0.0102367009967566)
--(axis cs:78,0.0102367009967566)
--(axis cs:79,0.0102367009967566)
--(axis cs:80,0.00964841619133949)
--(axis cs:81,0.00964841619133949)
--(axis cs:82,0.0095635000616312)
--(axis cs:83,0.0095635000616312)
--(axis cs:84,0.0095635000616312)
--(axis cs:85,0.0095635000616312)
--(axis cs:86,0.0095635000616312)
--(axis cs:87,0.0095635000616312)
--(axis cs:88,0.0095635000616312)
--(axis cs:89,0.0095635000616312)
--(axis cs:90,0.00852932874113321)
--(axis cs:91,0.00617046421393752)
--(axis cs:92,0.00617046421393752)
--(axis cs:93,0.00617046421393752)
--(axis cs:94,0.00617046421393752)
--(axis cs:95,0.00617046421393752)
--(axis cs:96,0.00617046421393752)
--(axis cs:97,0.00617046421393752)
--(axis cs:98,0.00617046421393752)
--(axis cs:99,0.00617046421393752)
--(axis cs:100,0.00585184711962938)
--(axis cs:101,0.0056401751935482)
--(axis cs:102,0.0056401751935482)
--(axis cs:103,0.0056401751935482)
--(axis cs:104,0.0056401751935482)
--(axis cs:105,0.0056401751935482)
--(axis cs:106,0.0056401751935482)
--(axis cs:107,0.0056401751935482)
--(axis cs:108,0.0056401751935482)
--(axis cs:109,0.0056401751935482)
--(axis cs:110,0.0056401751935482)
--(axis cs:111,0.0056401751935482)
--(axis cs:111,0.0130724161863327)
--(axis cs:111,0.0130724161863327)
--(axis cs:110,0.0130724161863327)
--(axis cs:109,0.0130724161863327)
--(axis cs:108,0.0130724161863327)
--(axis cs:107,0.0130724161863327)
--(axis cs:106,0.0130724161863327)
--(axis cs:105,0.0130724161863327)
--(axis cs:104,0.0130724161863327)
--(axis cs:103,0.0130724161863327)
--(axis cs:102,0.0130724161863327)
--(axis cs:101,0.0130724161863327)
--(axis cs:100,0.0147634437307715)
--(axis cs:99,0.0200938694179058)
--(axis cs:98,0.0200938694179058)
--(axis cs:97,0.0200938694179058)
--(axis cs:96,0.0200938694179058)
--(axis cs:95,0.0200938694179058)
--(axis cs:94,0.0200938694179058)
--(axis cs:93,0.0200938694179058)
--(axis cs:92,0.0200938694179058)
--(axis cs:91,0.0200938694179058)
--(axis cs:90,0.0222580693662167)
--(axis cs:89,0.0229966808110476)
--(axis cs:88,0.0229966808110476)
--(axis cs:87,0.0229966808110476)
--(axis cs:86,0.0229966808110476)
--(axis cs:85,0.0229966808110476)
--(axis cs:84,0.0229966808110476)
--(axis cs:83,0.0229966808110476)
--(axis cs:82,0.0229966808110476)
--(axis cs:81,0.0251589082181454)
--(axis cs:80,0.0251589082181454)
--(axis cs:79,0.0256705898791552)
--(axis cs:78,0.0256705898791552)
--(axis cs:77,0.0256705898791552)
--(axis cs:76,0.0256705898791552)
--(axis cs:75,0.0256705898791552)
--(axis cs:74,0.0280578918755054)
--(axis cs:73,0.0280578918755054)
--(axis cs:72,0.0280578918755054)
--(axis cs:71,0.0310778096318245)
--(axis cs:70,0.0310778096318245)
--(axis cs:69,0.0310778096318245)
--(axis cs:68,0.0310778096318245)
--(axis cs:67,0.0310778096318245)
--(axis cs:66,0.0310778096318245)
--(axis cs:65,0.0362838767468929)
--(axis cs:64,0.0534110032021999)
--(axis cs:63,0.0534110032021999)
--(axis cs:62,0.0534110032021999)
--(axis cs:61,0.053798034787178)
--(axis cs:60,0.053798034787178)
--(axis cs:59,0.0559510625898838)
--(axis cs:58,0.0559510625898838)
--(axis cs:57,0.0597930215299129)
--(axis cs:56,0.0597930215299129)
--(axis cs:55,0.0666642338037491)
--(axis cs:54,0.0666642338037491)
--(axis cs:53,0.0722471475601196)
--(axis cs:52,0.0722471475601196)
--(axis cs:51,0.0823703855276108)
--(axis cs:50,0.0823703855276108)
--(axis cs:49,0.0823703855276108)
--(axis cs:48,0.0823703855276108)
--(axis cs:47,0.0857456177473068)
--(axis cs:46,0.0857456177473068)
--(axis cs:45,0.0857456177473068)
--(axis cs:44,0.0867229253053665)
--(axis cs:43,0.117722198367119)
--(axis cs:42,0.117722198367119)
--(axis cs:41,0.117722198367119)
--(axis cs:40,0.117722198367119)
--(axis cs:39,0.152537271380424)
--(axis cs:38,0.152537271380424)
--(axis cs:37,0.152537271380424)
--(axis cs:36,0.152537271380424)
--(axis cs:35,0.202405959367752)
--(axis cs:34,0.202405959367752)
--(axis cs:33,0.28599151968956)
--(axis cs:32,0.325072348117828)
--(axis cs:31,0.325072348117828)
--(axis cs:30,0.325072348117828)
--(axis cs:29,0.481631904840469)
--(axis cs:28,0.481631904840469)
--(axis cs:27,0.481631904840469)
--(axis cs:26,0.490860879421234)
--(axis cs:25,0.570028483867645)
--(axis cs:24,0.597859442234039)
--(axis cs:23,0.625128269195557)
--(axis cs:22,0.642274379730225)
--(axis cs:21,0.747675597667694)
--(axis cs:20,0.76904284954071)
--(axis cs:19,0.787416636943817)
--(axis cs:18,0.799923598766327)
--(axis cs:17,0.839565992355347)
--(axis cs:16,0.839565992355347)
--(axis cs:15,0.907949030399323)
--(axis cs:14,0.907949030399323)
--(axis cs:13,0.935862362384796)
--(axis cs:12,0.977770149707794)
--(axis cs:11,1.0111186504364)
--(axis cs:10,1.02326846122742)
--cycle;

\path [draw=color1, fill=color1, opacity=0.3]
(axis cs:10,1.02326841488654)
--(axis cs:10,0.826151056536423)
--(axis cs:11,0.676578013191154)
--(axis cs:12,0.654145273912285)
--(axis cs:13,0.465008535220775)
--(axis cs:14,0.428202083546036)
--(axis cs:15,0.380787420607002)
--(axis cs:16,0.358327130957515)
--(axis cs:17,0.274974484993432)
--(axis cs:18,0.221401496862925)
--(axis cs:19,0.18655145855825)
--(axis cs:20,0.134462783961012)
--(axis cs:21,0.08711258970047)
--(axis cs:22,0.0592360465574987)
--(axis cs:23,0.049324677738428)
--(axis cs:24,0.049324677738428)
--(axis cs:25,0.0425925095441091)
--(axis cs:26,0.0425925095441091)
--(axis cs:27,0.0425925095441091)
--(axis cs:28,0.0371763545426669)
--(axis cs:29,0.0371763545426669)
--(axis cs:30,0.0304237189626115)
--(axis cs:31,0.0304237189626115)
--(axis cs:32,0.0304237189626115)
--(axis cs:33,0.028699156147151)
--(axis cs:34,0.028699156147151)
--(axis cs:35,0.028699156147151)
--(axis cs:36,0.028699156147151)
--(axis cs:37,0.028699156147151)
--(axis cs:38,0.028699156147151)
--(axis cs:39,0.0270852351413062)
--(axis cs:40,0.0270852351413062)
--(axis cs:41,0.0232159489076826)
--(axis cs:42,0.0232159489076826)
--(axis cs:43,0.0185404277665025)
--(axis cs:44,0.0185404277665025)
--(axis cs:45,0.0183197959344149)
--(axis cs:46,0.0183197959344149)
--(axis cs:47,0.0183197959344149)
--(axis cs:48,0.0180824067329941)
--(axis cs:49,0.0180824067329941)
--(axis cs:50,0.0180824067329941)
--(axis cs:51,0.0180824067329941)
--(axis cs:52,0.0180824067329941)
--(axis cs:53,0.0180824067329941)
--(axis cs:54,0.0168038977367844)
--(axis cs:55,0.0168038977367844)
--(axis cs:56,0.0168038977367844)
--(axis cs:57,0.0168038977367844)
--(axis cs:58,0.0133353700263356)
--(axis cs:59,0.0133353700263356)
--(axis cs:60,0.0132558794857371)
--(axis cs:61,0.0132558794857371)
--(axis cs:62,0.0132558794857371)
--(axis cs:63,0.0132558794857371)
--(axis cs:64,0.0132558794857371)
--(axis cs:65,0.0132558794857371)
--(axis cs:66,0.0132558794857371)
--(axis cs:67,0.0114510062354197)
--(axis cs:68,0.0114510062354197)
--(axis cs:69,0.0114510062354197)
--(axis cs:70,0.0114510062354197)
--(axis cs:71,0.0114510062354197)
--(axis cs:72,0.0114510062354197)
--(axis cs:73,0.0114510062354197)
--(axis cs:74,0.00762116319789817)
--(axis cs:75,0.00762116319789817)
--(axis cs:76,0.00762116319789817)
--(axis cs:77,0.0064568467617377)
--(axis cs:78,0.0064568467617377)
--(axis cs:79,0.0064568467617377)
--(axis cs:80,0.0064568467617377)
--(axis cs:81,0.0064568467617377)
--(axis cs:82,0.0064568467617377)
--(axis cs:83,0.0064568467617377)
--(axis cs:84,0.0064568467617377)
--(axis cs:85,0.0064568467617377)
--(axis cs:86,0.00591048924098885)
--(axis cs:87,0.00591048924098885)
--(axis cs:88,0.00591048924098885)
--(axis cs:89,0.00591048924098885)
--(axis cs:90,0.00591048924098885)
--(axis cs:91,0.00591048924098885)
--(axis cs:92,0.00591048924098885)
--(axis cs:93,0.00591048924098885)
--(axis cs:94,0.00591048924098885)
--(axis cs:95,0.00591048924098885)
--(axis cs:96,0.00591048924098885)
--(axis cs:97,0.00591048924098885)
--(axis cs:98,0.00417298468570029)
--(axis cs:99,0.00417298468570029)
--(axis cs:100,0.00417298468570029)
--(axis cs:101,0.00417298468570029)
--(axis cs:102,0.00417298468570029)
--(axis cs:103,0.00417298468570029)
--(axis cs:104,0.00390654649731902)
--(axis cs:105,0.00390654649731902)
--(axis cs:106,0.00390654649731902)
--(axis cs:107,0.00390654649731902)
--(axis cs:108,0.00354573073653834)
--(axis cs:109,0.00354573073653834)
--(axis cs:110,0.00354573073653834)
--(axis cs:111,0.00354573073653834)
--(axis cs:111,0.00613476083647635)
--(axis cs:111,0.00613476083647635)
--(axis cs:110,0.00613476083647635)
--(axis cs:109,0.00613476083647635)
--(axis cs:108,0.00613476083647635)
--(axis cs:107,0.00744483065341818)
--(axis cs:106,0.00744483065341818)
--(axis cs:105,0.00744483065341818)
--(axis cs:104,0.00744483065341818)
--(axis cs:103,0.0112381407877202)
--(axis cs:102,0.0112381407877202)
--(axis cs:101,0.0112381407877202)
--(axis cs:100,0.0112381407877202)
--(axis cs:99,0.0112381407877202)
--(axis cs:98,0.0112381407877202)
--(axis cs:97,0.0144115237328616)
--(axis cs:96,0.0144115237328616)
--(axis cs:95,0.0144115237328616)
--(axis cs:94,0.0144115237328616)
--(axis cs:93,0.0144115237328616)
--(axis cs:92,0.0144115237328616)
--(axis cs:91,0.0144115237328616)
--(axis cs:90,0.0144115237328616)
--(axis cs:89,0.0144115237328616)
--(axis cs:88,0.0144115237328616)
--(axis cs:87,0.0144115237328616)
--(axis cs:86,0.0144115237328616)
--(axis cs:85,0.0147873027360596)
--(axis cs:84,0.0147873027360596)
--(axis cs:83,0.0147873027360596)
--(axis cs:82,0.0147873027360596)
--(axis cs:81,0.0147873027360596)
--(axis cs:80,0.0147873027360596)
--(axis cs:79,0.0147873027360596)
--(axis cs:78,0.0147873027360596)
--(axis cs:77,0.0147873027360596)
--(axis cs:76,0.0158847690871372)
--(axis cs:75,0.0158847690871372)
--(axis cs:74,0.0158847690871372)
--(axis cs:73,0.0227795848362931)
--(axis cs:72,0.0227795848362931)
--(axis cs:71,0.0227795848362931)
--(axis cs:70,0.0227795848362931)
--(axis cs:69,0.0227795848362931)
--(axis cs:68,0.0227795848362931)
--(axis cs:67,0.0227795848362931)
--(axis cs:66,0.0318052580071243)
--(axis cs:65,0.0318052580071243)
--(axis cs:64,0.0318052580071243)
--(axis cs:63,0.0318052580071243)
--(axis cs:62,0.0318052580071243)
--(axis cs:61,0.0318052580071243)
--(axis cs:60,0.0318052580071243)
--(axis cs:59,0.0326996194295892)
--(axis cs:58,0.0326996194295892)
--(axis cs:57,0.0357278487326278)
--(axis cs:56,0.0357278487326278)
--(axis cs:55,0.0357278487326278)
--(axis cs:54,0.0357278487326278)
--(axis cs:53,0.0364992445440153)
--(axis cs:52,0.0364992445440153)
--(axis cs:51,0.0364992445440153)
--(axis cs:50,0.0364992445440153)
--(axis cs:49,0.0364992445440153)
--(axis cs:48,0.0364992445440153)
--(axis cs:47,0.0396564363597724)
--(axis cs:46,0.0396564363597724)
--(axis cs:45,0.0396564363597724)
--(axis cs:44,0.0400419596838482)
--(axis cs:43,0.0400419596838482)
--(axis cs:42,0.0450587115188743)
--(axis cs:41,0.0450587115188743)
--(axis cs:40,0.0486183884870398)
--(axis cs:39,0.0486183884870398)
--(axis cs:38,0.0517456184693912)
--(axis cs:37,0.0517456184693912)
--(axis cs:36,0.0517456184693912)
--(axis cs:35,0.0517456184693912)
--(axis cs:34,0.0517456184693912)
--(axis cs:33,0.0517456184693912)
--(axis cs:32,0.0723075312066464)
--(axis cs:31,0.0723075312066464)
--(axis cs:30,0.0723075312066464)
--(axis cs:29,0.0828061176711639)
--(axis cs:28,0.0828061176711639)
--(axis cs:27,0.207297158294373)
--(axis cs:26,0.207297158294373)
--(axis cs:25,0.207297158294373)
--(axis cs:24,0.213150200609272)
--(axis cs:23,0.213150200609272)
--(axis cs:22,0.222244881814636)
--(axis cs:21,0.244753051224335)
--(axis cs:20,0.288093442314913)
--(axis cs:19,0.345707596116928)
--(axis cs:18,0.423810182246524)
--(axis cs:17,0.458359103170888)
--(axis cs:16,0.566483293787548)
--(axis cs:15,0.59712934469467)
--(axis cs:14,0.650247866566022)
--(axis cs:13,0.684074907897134)
--(axis cs:12,0.874249129854811)
--(axis cs:11,0.90479245661823)
--(axis cs:10,1.02326841488654)
--cycle;

\addplot [semithick, blue, dash dot]
table {%
10 0.924709677696228
11 0.824458003044128
12 0.64415180683136
13 0.551238894462585
14 0.522167026996613
15 0.487103849649429
16 0.470043957233429
17 0.400944888591766
18 0.348174661397934
19 0.341928154230118
20 0.301496177911758
21 0.294846773147583
22 0.286064326763153
23 0.203778341412544
24 0.203778341412544
25 0.188755676150322
26 0.148248434066772
27 0.0790731757879257
28 0.0505163185298443
29 0.0392752774059772
30 0.037151824682951
31 0.037151824682951
32 0.0303441174328327
33 0.0303441174328327
34 0.0172565579414368
35 0.0117656113579869
36 0.0117656113579869
37 0.0117656113579869
38 0.0117656113579869
39 0.0117656113579869
40 0.0117656113579869
41 0.0117656113579869
42 0.0117656113579869
43 0.00831859093159437
44 0.00831859093159437
45 0.00831859093159437
46 0.00797959603369236
47 0.00797959603369236
48 0.00698624830693007
49 0.00698624830693007
50 0.00692697754129767
51 0.00692697754129767
52 0.00692697754129767
53 0.00692697754129767
54 0.00626903772354126
55 0.00626903772354126
56 0.00536478776484728
57 0.00536478776484728
58 0.00536478776484728
59 0.00536478776484728
60 0.00536478776484728
61 0.00457659969106317
62 0.00457659969106317
63 0.00457659969106317
64 0.00457659969106317
65 0.00457659969106317
66 0.00457659969106317
67 0.00434433203190565
68 0.00432755937799811
69 0.00432755937799811
70 0.00432755937799811
71 0.00432755937799811
72 0.00432755937799811
73 0.00432755937799811
74 0.00432755937799811
75 0.00432755937799811
76 0.00397953996434808
77 0.00397953996434808
78 0.00397953996434808
79 0.00397953996434808
80 0.00397953996434808
81 0.00397953996434808
82 0.00358635187149048
83 0.00358635187149048
84 0.00358635187149048
85 0.00358635187149048
86 0.00358635187149048
87 0.00358635187149048
88 0.00358635187149048
89 0.00358635187149048
90 0.00358635187149048
91 0.00358635187149048
92 0.00358635187149048
93 0.00358635187149048
94 0.00358635187149048
95 0.00358635187149048
96 0.00358635187149048
97 0.00358635187149048
98 0.00358635187149048
99 0.00358635187149048
100 0.00358635187149048
101 0.00358635187149048
102 0.00358635187149048
103 0.00358635187149048
104 0.00336775789037347
105 0.00336775789037347
106 0.00336775789037347
107 0.00336775789037347
108 0.00336775789037347
109 0.00336775789037347
110 0.00336775789037347
111 0.00336775789037347
};
\addplot [semithick, color1]
table {%
10 0.924709735711479
11 0.923747872269201
12 0.92126984029949
13 0.825971170870321
14 0.807136553761072
15 0.723737567539129
16 0.653754874356939
17 0.653055679215514
18 0.620339060226026
19 0.603475151224167
20 0.60264920202452
21 0.593309344836618
22 0.562878184144924
23 0.480107205622743
24 0.479836981278009
25 0.478756283850463
26 0.472598509347202
27 0.393006521991098
28 0.336571839970864
29 0.320676330321191
30 0.304678302772379
31 0.295580945350923
32 0.288243228867362
33 0.252455132422869
34 0.235619253920444
35 0.223940775379988
36 0.219165602155146
37 0.219075210392025
38 0.219075210392025
39 0.218293534508055
40 0.218293534508055
41 0.21816478240121
42 0.218102117628838
43 0.218102117628838
44 0.217790174959946
45 0.216608169929072
46 0.145026668807405
47 0.121461841353039
48 0.121461841353039
49 0.121461841353039
50 0.120932463791113
51 0.120932463791113
52 0.120932463791113
53 0.118877371897936
54 0.118842507843486
55 0.11882665863176
56 0.118625282053218
57 0.118350170563747
58 0.118119211097522
59 0.118119211097522
60 0.118119211097522
61 0.117704899895712
62 0.117704899895712
63 0.117696960415456
64 0.117696960415456
65 0.117696960415456
66 0.117696960415456
67 0.117683674341811
68 0.117683674341811
69 0.117683674341811
70 0.117683674341811
71 0.117671998215863
72 0.117671998215863
73 0.117585800482091
74 0.117585800482091
75 0.117585800482091
76 0.117571429383621
77 0.117569974106522
78 0.117569974106522
79 0.117569974106522
80 0.117569974106522
81 0.117569974106522
82 0.117536384002409
83 0.117536384002409
84 0.117536384002409
85 0.117536384002409
86 0.117536384002409
87 0.117536384002409
88 0.117536384002409
89 0.117536384002409
90 0.117536384002409
91 0.117536384002409
92 0.117536384002409
93 0.117536384002409
94 0.117536384002409
95 0.117536384002409
96 0.117530006647001
97 0.117521539199444
98 0.117521539199444
99 0.117521539199444
100 0.117521539199444
101 0.117521539199444
102 0.117518483531442
103 0.117518483531442
104 0.117518483531442
105 0.117518483531442
106 0.117518483531442
107 0.117518483531442
108 0.117518483531442
109 0.117518483531442
110 0.117518483531442
111 0.117518483531442
};
\addplot [semithick, blue]
table {%
10 0.924709796905518
11 0.915995597839355
12 0.881894111633301
13 0.841901302337646
14 0.82032984495163
15 0.82032984495163
16 0.76178777217865
17 0.76178777217865
18 0.707806408405304
19 0.692517220973969
20 0.671983242034912
21 0.651061236858368
22 0.526005387306213
23 0.51304417848587
24 0.479526817798615
25 0.460116684436798
26 0.39397132396698
27 0.386073380708694
28 0.386073380708694
29 0.386073380708694
30 0.261871874332428
31 0.261871874332428
32 0.261871874332428
33 0.227278381586075
34 0.156911253929138
35 0.156911253929138
36 0.122269548475742
37 0.122269548475742
38 0.122269548475742
39 0.122269548475742
40 0.0970408171415329
41 0.0970408171415329
42 0.0970408171415329
43 0.0970408171415329
44 0.0693570151925087
45 0.0686850324273109
46 0.0686850324273109
47 0.0686850324273109
48 0.0658021196722984
49 0.0658021196722984
50 0.0658021196722984
51 0.0658021196722984
52 0.057628083974123
53 0.057628083974123
54 0.0528063550591469
55 0.0528063550591469
56 0.0459857955574989
57 0.0459857955574989
58 0.0423909798264503
59 0.0423909798264503
60 0.0398434512317181
61 0.0398434512317181
62 0.0395755171775818
63 0.0395755171775818
64 0.0395755171775818
65 0.027293598279357
66 0.0223935842514038
67 0.0223935842514038
68 0.0223935842514038
69 0.0223935842514038
70 0.0223935842514038
71 0.0223935842514038
72 0.0201570875942707
73 0.0201570875942707
74 0.0201570875942707
75 0.0179536454379559
76 0.0179536454379559
77 0.0179536454379559
78 0.0179536454379559
79 0.0179536454379559
80 0.0174036622047424
81 0.0174036622047424
82 0.0162800904363394
83 0.0162800904363394
84 0.0162800904363394
85 0.0162800904363394
86 0.0162800904363394
87 0.0162800904363394
88 0.0162800904363394
89 0.0162800904363394
90 0.0153936985880136
91 0.0131321670487523
92 0.0131321670487523
93 0.0131321670487523
94 0.0131321670487523
95 0.0131321670487523
96 0.0131321670487523
97 0.0131321670487523
98 0.0131321670487523
99 0.0131321670487523
100 0.0103076454252005
101 0.00935629568994045
102 0.00935629568994045
103 0.00935629568994045
104 0.00935629568994045
105 0.00935629568994045
106 0.00935629568994045
107 0.00935629568994045
108 0.00935629568994045
109 0.00935629568994045
110 0.00935629568994045
111 0.00935629568994045
};
\addplot [semithick, color1, dash dot]
table {%
10 0.924709735711479
11 0.790685234904692
12 0.764197201883548
13 0.574541721558954
14 0.539224975056029
15 0.488958382650836
16 0.462405212372532
17 0.36666679408216
18 0.322605839554725
19 0.266129527337589
20 0.211278113137963
21 0.165932820462403
22 0.140740464186067
23 0.13123743917385
24 0.13123743917385
25 0.124944833919241
26 0.124944833919241
27 0.124944833919241
28 0.0599912361069154
29 0.0599912361069154
30 0.051365625084629
31 0.051365625084629
32 0.051365625084629
33 0.0402223873082711
34 0.0402223873082711
35 0.0402223873082711
36 0.0402223873082711
37 0.0402223873082711
38 0.0402223873082711
39 0.037851811814173
40 0.037851811814173
41 0.0341373302132785
42 0.0341373302132785
43 0.0292911937251753
44 0.0292911937251753
45 0.0289881161470936
46 0.0289881161470936
47 0.0289881161470936
48 0.0272908256385047
49 0.0272908256385047
50 0.0272908256385047
51 0.0272908256385047
52 0.0272908256385047
53 0.0272908256385047
54 0.0262658732347061
55 0.0262658732347061
56 0.0262658732347061
57 0.0262658732347061
58 0.0230174947279624
59 0.0230174947279624
60 0.0225305687464307
61 0.0225305687464307
62 0.0225305687464307
63 0.0225305687464307
64 0.0225305687464307
65 0.0225305687464307
66 0.0225305687464307
67 0.0171152955358564
68 0.0171152955358564
69 0.0171152955358564
70 0.0171152955358564
71 0.0171152955358564
72 0.0171152955358564
73 0.0171152955358564
74 0.0117529661425177
75 0.0117529661425177
76 0.0117529661425177
77 0.0106220747488986
78 0.0106220747488986
79 0.0106220747488986
80 0.0106220747488986
81 0.0106220747488986
82 0.0106220747488986
83 0.0106220747488986
84 0.0106220747488986
85 0.0106220747488986
86 0.0101610064869252
87 0.0101610064869252
88 0.0101610064869252
89 0.0101610064869252
90 0.0101610064869252
91 0.0101610064869252
92 0.0101610064869252
93 0.0101610064869252
94 0.0101610064869252
95 0.0101610064869252
96 0.0101610064869252
97 0.0101610064869252
98 0.00770556273671026
99 0.00770556273671026
100 0.00770556273671026
101 0.00770556273671026
102 0.00770556273671026
103 0.00770556273671026
104 0.0056756885753686
105 0.0056756885753686
106 0.0056756885753686
107 0.0056756885753686
108 0.00484024578650735
109 0.00484024578650735
110 0.00484024578650735
111 0.00484024578650735
};
\end{axis}

\end{tikzpicture}

%% file: figures/bop_3d_levy_weak.tex
\begin{tikzpicture}

\definecolor{color0}{rgb}{0,0,1}
\definecolor{color1}{rgb}{1,0.549019607843137,0}
\definecolor{color2}{rgb}{1,0.647058823529412,0}
\definecolor{color3}{rgb}{0.564705882352941,0.933333333333333,0.564705882352941}

\begin{axis}[axis on top,
enlarge x limits=false,
enlarge y limits=false,
height=\figureheight,
scale only axis,
tick align=outside,
tick pos=left,
tick pos=left,
width=\figurewidth,
xlabel={Iteration},
xmin=10, xmax=60,
xtick style={color=black},
xtick={-10,0,10,25,50,75,100},
xticklabels={\ensuremath{-}10,0,10,25,50,75,90},
ylabel={Regret},
ymin=-0.05, ymax=1.3,
ytick style={color=black},
ytick={0.   , 1.3},
]
\node[anchor=north east] at (rel axis cs:1,1) {Levy 3D (weak)};
\path [draw=blue, fill=blue, opacity=0.3]
(axis cs:10,1.241255402565)
--(axis cs:10,1.00074923038483)
--(axis cs:11,0.876969575881958)
--(axis cs:12,0.805118143558502)
--(axis cs:13,0.735591113567352)
--(axis cs:14,0.670308887958527)
--(axis cs:15,0.561274528503418)
--(axis cs:16,0.554334878921509)
--(axis cs:17,0.449187695980072)
--(axis cs:18,0.391276597976685)
--(axis cs:19,0.391276597976685)
--(axis cs:20,0.391276597976685)
--(axis cs:21,0.323309153318405)
--(axis cs:22,0.316483557224274)
--(axis cs:23,0.297248154878616)
--(axis cs:24,0.297248154878616)
--(axis cs:25,0.23415020108223)
--(axis cs:26,0.224957048892975)
--(axis cs:27,0.216140329837799)
--(axis cs:28,0.216140329837799)
--(axis cs:29,0.188582748174667)
--(axis cs:30,0.15810926258564)
--(axis cs:31,0.147104352712631)
--(axis cs:32,0.134197354316711)
--(axis cs:33,0.129070952534676)
--(axis cs:34,0.124609708786011)
--(axis cs:35,0.120345517992973)
--(axis cs:36,0.110876709222794)
--(axis cs:37,0.110876709222794)
--(axis cs:38,0.109105706214905)
--(axis cs:39,0.10908704996109)
--(axis cs:40,0.106293708086014)
--(axis cs:41,0.106293708086014)
--(axis cs:42,0.0824086368083954)
--(axis cs:43,0.0771691054105759)
--(axis cs:44,0.0669306367635727)
--(axis cs:45,0.0662958100438118)
--(axis cs:46,0.0657716765999794)
--(axis cs:47,0.0657716765999794)
--(axis cs:48,0.0644363909959793)
--(axis cs:49,0.0642179250717163)
--(axis cs:50,0.0625239685177803)
--(axis cs:51,0.0625239685177803)
--(axis cs:52,0.0625239685177803)
--(axis cs:53,0.0583559684455395)
--(axis cs:54,0.0559921115636826)
--(axis cs:55,0.0559921115636826)
--(axis cs:56,0.0559921115636826)
--(axis cs:57,0.0559921115636826)
--(axis cs:58,0.0440161228179932)
--(axis cs:59,0.0440161228179932)
--(axis cs:60,0.0440161228179932)
--(axis cs:61,0.0440161228179932)
--(axis cs:62,0.0440161228179932)
--(axis cs:63,0.0440161228179932)
--(axis cs:64,0.0440161228179932)
--(axis cs:65,0.0415075197815895)
--(axis cs:66,0.0415075197815895)
--(axis cs:67,0.0384765826165676)
--(axis cs:68,0.0384765826165676)
--(axis cs:69,0.0369325205683708)
--(axis cs:70,0.0369325205683708)
--(axis cs:71,0.0369325205683708)
--(axis cs:72,0.0369325205683708)
--(axis cs:73,0.0369325205683708)
--(axis cs:74,0.0369325205683708)
--(axis cs:75,0.0369325205683708)
--(axis cs:76,0.0369325205683708)
--(axis cs:77,0.0369325205683708)
--(axis cs:78,0.0369325205683708)
--(axis cs:79,0.0369325205683708)
--(axis cs:80,0.0369325205683708)
--(axis cs:81,0.0365444347262383)
--(axis cs:82,0.0365444347262383)
--(axis cs:83,0.0365444347262383)
--(axis cs:84,0.0365444347262383)
--(axis cs:85,0.0365444347262383)
--(axis cs:86,0.0365444347262383)
--(axis cs:87,0.0363676436245441)
--(axis cs:88,0.0363676436245441)
--(axis cs:89,0.0363676436245441)
--(axis cs:90,0.0363676436245441)
--(axis cs:91,0.0335241109132767)
--(axis cs:92,0.0335241109132767)
--(axis cs:93,0.0328845456242561)
--(axis cs:94,0.0328845456242561)
--(axis cs:95,0.0328845456242561)
--(axis cs:96,0.0328845456242561)
--(axis cs:97,0.0328845493495464)
--(axis cs:98,0.0328845493495464)
--(axis cs:99,0.0328845493495464)
--(axis cs:100,0.0328845493495464)
--(axis cs:101,0.0301262848079205)
--(axis cs:102,0.0301262848079205)
--(axis cs:103,0.0291488990187645)
--(axis cs:104,0.0291488990187645)
--(axis cs:105,0.0291488990187645)
--(axis cs:106,0.0291488990187645)
--(axis cs:107,0.0291488990187645)
--(axis cs:108,0.0281238052994013)
--(axis cs:109,0.0281238052994013)
--(axis cs:110,0.0252609495073557)
--(axis cs:111,0.0252609495073557)
--(axis cs:111,0.0397492237389088)
--(axis cs:111,0.0397492237389088)
--(axis cs:110,0.0397492237389088)
--(axis cs:109,0.0416301004588604)
--(axis cs:108,0.0416301004588604)
--(axis cs:107,0.0437524542212486)
--(axis cs:106,0.0437524542212486)
--(axis cs:105,0.0437524542212486)
--(axis cs:104,0.0437524542212486)
--(axis cs:103,0.0437524542212486)
--(axis cs:102,0.0447445847094059)
--(axis cs:101,0.0447445847094059)
--(axis cs:100,0.0482886545360088)
--(axis cs:99,0.0482886545360088)
--(axis cs:98,0.0482886545360088)
--(axis cs:97,0.0482886545360088)
--(axis cs:96,0.0482886508107185)
--(axis cs:95,0.0482886508107185)
--(axis cs:94,0.0482886508107185)
--(axis cs:93,0.0482886508107185)
--(axis cs:92,0.050602100789547)
--(axis cs:91,0.050602100789547)
--(axis cs:90,0.0548621378839016)
--(axis cs:89,0.0548621378839016)
--(axis cs:88,0.0548621378839016)
--(axis cs:87,0.0548621378839016)
--(axis cs:86,0.0551403239369392)
--(axis cs:85,0.0551403239369392)
--(axis cs:84,0.0551403239369392)
--(axis cs:83,0.0551403239369392)
--(axis cs:82,0.0551403239369392)
--(axis cs:81,0.0551403239369392)
--(axis cs:80,0.0555139109492302)
--(axis cs:79,0.0555139109492302)
--(axis cs:78,0.0555139109492302)
--(axis cs:77,0.0555139109492302)
--(axis cs:76,0.0555139109492302)
--(axis cs:75,0.0555139109492302)
--(axis cs:74,0.0555139109492302)
--(axis cs:73,0.0555139109492302)
--(axis cs:72,0.0555139109492302)
--(axis cs:71,0.0555139109492302)
--(axis cs:70,0.0555139109492302)
--(axis cs:69,0.0555139109492302)
--(axis cs:68,0.063143715262413)
--(axis cs:67,0.063143715262413)
--(axis cs:66,0.0686723440885544)
--(axis cs:65,0.0686723440885544)
--(axis cs:64,0.0699330121278763)
--(axis cs:63,0.0699330121278763)
--(axis cs:62,0.0699330121278763)
--(axis cs:61,0.0699330121278763)
--(axis cs:60,0.0699330121278763)
--(axis cs:59,0.0699330121278763)
--(axis cs:58,0.0699330121278763)
--(axis cs:57,0.0819606930017471)
--(axis cs:56,0.0819606930017471)
--(axis cs:55,0.0819606930017471)
--(axis cs:54,0.0819606930017471)
--(axis cs:53,0.0832188650965691)
--(axis cs:52,0.0866025313735008)
--(axis cs:51,0.0866025313735008)
--(axis cs:50,0.0866025313735008)
--(axis cs:49,0.0956852585077286)
--(axis cs:48,0.0960192978382111)
--(axis cs:47,0.0968696996569633)
--(axis cs:46,0.0968696996569633)
--(axis cs:45,0.0974027141928673)
--(axis cs:44,0.103405579924583)
--(axis cs:43,0.123538419604301)
--(axis cs:42,0.127500802278519)
--(axis cs:41,0.170641243457794)
--(axis cs:40,0.170641243457794)
--(axis cs:39,0.185353934764862)
--(axis cs:38,0.185384333133698)
--(axis cs:37,0.186565905809402)
--(axis cs:36,0.186565905809402)
--(axis cs:35,0.209420725703239)
--(axis cs:34,0.211912840604782)
--(axis cs:33,0.215081080794334)
--(axis cs:32,0.24358657002449)
--(axis cs:31,0.259759277105331)
--(axis cs:30,0.265765517950058)
--(axis cs:29,0.296754211187363)
--(axis cs:28,0.379217922687531)
--(axis cs:27,0.379217922687531)
--(axis cs:26,0.387189924716949)
--(axis cs:25,0.393030792474747)
--(axis cs:24,0.480383723974228)
--(axis cs:23,0.480383723974228)
--(axis cs:22,0.494392991065979)
--(axis cs:21,0.500836193561554)
--(axis cs:20,0.6242356300354)
--(axis cs:19,0.6242356300354)
--(axis cs:18,0.6242356300354)
--(axis cs:17,0.692644536495209)
--(axis cs:16,0.773667812347412)
--(axis cs:15,0.778069376945496)
--(axis cs:14,0.954470455646515)
--(axis cs:13,0.992536127567291)
--(axis cs:12,1.10958433151245)
--(axis cs:11,1.1766791343689)
--(axis cs:10,1.241255402565)
--cycle;

\path [draw=color1, fill=color1, opacity=0.3]
(axis cs:10,1.24125538018958)
--(axis cs:10,1.00074916931375)
--(axis cs:11,0.942357521064722)
--(axis cs:12,0.891845748703846)
--(axis cs:13,0.817123787456793)
--(axis cs:14,0.684293465148453)
--(axis cs:15,0.572821226693481)
--(axis cs:16,0.566921558274605)
--(axis cs:17,0.566921558274605)
--(axis cs:18,0.521996305979786)
--(axis cs:19,0.508683283947353)
--(axis cs:20,0.410145358190978)
--(axis cs:21,0.410145358190978)
--(axis cs:22,0.410145358190978)
--(axis cs:23,0.410145358190978)
--(axis cs:24,0.410145358190978)
--(axis cs:25,0.371681952220337)
--(axis cs:26,0.293842133366977)
--(axis cs:27,0.255917488375228)
--(axis cs:28,0.255917488375228)
--(axis cs:29,0.172122016564057)
--(axis cs:30,0.156297053112766)
--(axis cs:31,0.156297053112766)
--(axis cs:32,0.128819070571654)
--(axis cs:33,0.128146094082678)
--(axis cs:34,0.106954368659691)
--(axis cs:35,0.100917427703841)
--(axis cs:36,0.0988437128348637)
--(axis cs:37,0.0850093059098832)
--(axis cs:38,0.0837385316057173)
--(axis cs:39,0.0637927938692712)
--(axis cs:40,0.0637927938692712)
--(axis cs:41,0.0562104527439424)
--(axis cs:42,0.055367201175154)
--(axis cs:43,0.055367201175154)
--(axis cs:44,0.055367201175154)
--(axis cs:45,0.055367201175154)
--(axis cs:46,0.055367201175154)
--(axis cs:47,0.055367201175154)
--(axis cs:48,0.055367201175154)
--(axis cs:49,0.053158439568907)
--(axis cs:50,0.053158439568907)
--(axis cs:51,0.053158439568907)
--(axis cs:52,0.053158439568907)
--(axis cs:53,0.053158439568907)
--(axis cs:54,0.0526401946289326)
--(axis cs:55,0.0497762424246097)
--(axis cs:56,0.0497762424246097)
--(axis cs:57,0.0497762424246097)
--(axis cs:58,0.0497762424246097)
--(axis cs:59,0.0497762424246097)
--(axis cs:60,0.0497762424246097)
--(axis cs:61,0.0497762424246097)
--(axis cs:62,0.0497762424246097)
--(axis cs:63,0.0497762424246097)
--(axis cs:64,0.0497762424246097)
--(axis cs:65,0.0497762424246097)
--(axis cs:66,0.0497762424246097)
--(axis cs:67,0.0497762424246097)
--(axis cs:68,0.0497762424246097)
--(axis cs:69,0.0497762424246097)
--(axis cs:70,0.0497762424246097)
--(axis cs:71,0.0497762424246097)
--(axis cs:72,0.0497762424246097)
--(axis cs:73,0.0497762424246097)
--(axis cs:74,0.0497762424246097)
--(axis cs:75,0.0489104836964972)
--(axis cs:76,0.0489104836964972)
--(axis cs:77,0.0456455838600914)
--(axis cs:78,0.0456455838600914)
--(axis cs:79,0.0456455838600914)
--(axis cs:80,0.0456102608262886)
--(axis cs:81,0.0456102608262886)
--(axis cs:82,0.0456102608262886)
--(axis cs:83,0.0456102608262886)
--(axis cs:84,0.0456102608262886)
--(axis cs:85,0.0436519157308547)
--(axis cs:86,0.0436519157308547)
--(axis cs:87,0.0436519157308547)
--(axis cs:88,0.0432402656269677)
--(axis cs:89,0.0432402656269677)
--(axis cs:90,0.04160658286531)
--(axis cs:91,0.04160658286531)
--(axis cs:92,0.0406656644700497)
--(axis cs:93,0.0406656644700497)
--(axis cs:94,0.0406656644700497)
--(axis cs:95,0.0406656644700497)
--(axis cs:96,0.0406656644700497)
--(axis cs:97,0.0406656644700497)
--(axis cs:98,0.0326907424214019)
--(axis cs:99,0.0326907424214019)
--(axis cs:100,0.0326907424214019)
--(axis cs:101,0.0283928880242215)
--(axis cs:102,0.0283928880242215)
--(axis cs:103,0.0283928880242215)
--(axis cs:104,0.0283928880242215)
--(axis cs:105,0.0283928880242215)
--(axis cs:106,0.0283928880242215)
--(axis cs:107,0.0283928880242215)
--(axis cs:108,0.0283928880242215)
--(axis cs:109,0.0283928880242215)
--(axis cs:110,0.0283928880242215)
--(axis cs:111,0.0265374240535497)
--(axis cs:111,0.0444981912477927)
--(axis cs:111,0.0444981912477927)
--(axis cs:110,0.0462380579232303)
--(axis cs:109,0.0462380579232303)
--(axis cs:108,0.0462380579232303)
--(axis cs:107,0.0462380579232303)
--(axis cs:106,0.0462380579232303)
--(axis cs:105,0.0462380579232303)
--(axis cs:104,0.0462380579232303)
--(axis cs:103,0.0462380579232303)
--(axis cs:102,0.0462380579232303)
--(axis cs:101,0.0462380579232303)
--(axis cs:100,0.055267718024942)
--(axis cs:99,0.055267718024942)
--(axis cs:98,0.055267718024942)
--(axis cs:97,0.0616288121180601)
--(axis cs:96,0.0616288121180601)
--(axis cs:95,0.0616288121180601)
--(axis cs:94,0.0616288121180601)
--(axis cs:93,0.0616288121180601)
--(axis cs:92,0.0616288121180601)
--(axis cs:91,0.0622620944969833)
--(axis cs:90,0.0622620944969833)
--(axis cs:89,0.0639979202927378)
--(axis cs:88,0.0639979202927378)
--(axis cs:87,0.0646487153815637)
--(axis cs:86,0.0646487153815637)
--(axis cs:85,0.0646487153815637)
--(axis cs:84,0.0676997211121385)
--(axis cs:83,0.0676997211121385)
--(axis cs:82,0.0676997211121385)
--(axis cs:81,0.0676997211121385)
--(axis cs:80,0.0676997211121385)
--(axis cs:79,0.0677542247051631)
--(axis cs:78,0.0677542247051631)
--(axis cs:77,0.0677542247051631)
--(axis cs:76,0.0706671350097134)
--(axis cs:75,0.0706671350097134)
--(axis cs:74,0.0717904447147709)
--(axis cs:73,0.0717904447147709)
--(axis cs:72,0.0717904447147709)
--(axis cs:71,0.0717904447147709)
--(axis cs:70,0.0717904447147709)
--(axis cs:69,0.0717904447147709)
--(axis cs:68,0.0717904447147709)
--(axis cs:67,0.0717904447147709)
--(axis cs:66,0.0717904447147709)
--(axis cs:65,0.0717904447147709)
--(axis cs:64,0.0717904447147709)
--(axis cs:63,0.0717904447147709)
--(axis cs:62,0.0717904447147709)
--(axis cs:61,0.0717904447147709)
--(axis cs:60,0.0717904447147709)
--(axis cs:59,0.0717904447147709)
--(axis cs:58,0.0717904447147709)
--(axis cs:57,0.0717904447147709)
--(axis cs:56,0.0717904447147709)
--(axis cs:55,0.0717904447147709)
--(axis cs:54,0.0821288148818247)
--(axis cs:53,0.0831260542552967)
--(axis cs:52,0.0831260542552967)
--(axis cs:51,0.0831260542552967)
--(axis cs:50,0.0831260542552967)
--(axis cs:49,0.0831260542552967)
--(axis cs:48,0.0869499755665397)
--(axis cs:47,0.0869499755665397)
--(axis cs:46,0.0869499755665397)
--(axis cs:45,0.0869499755665397)
--(axis cs:44,0.0869499755665397)
--(axis cs:43,0.0869499755665397)
--(axis cs:42,0.0869499755665397)
--(axis cs:41,0.0884781755861721)
--(axis cs:40,0.104193742790443)
--(axis cs:39,0.104193742790443)
--(axis cs:38,0.198371970076643)
--(axis cs:37,0.199654838974864)
--(axis cs:36,0.212400620511681)
--(axis cs:35,0.294522272049085)
--(axis cs:34,0.301420753975204)
--(axis cs:33,0.318719863188582)
--(axis cs:32,0.355477228872303)
--(axis cs:31,0.379537807683369)
--(axis cs:30,0.379537807683369)
--(axis cs:29,0.423797499632084)
--(axis cs:28,0.526584892635705)
--(axis cs:27,0.526584892635705)
--(axis cs:26,0.555221287886899)
--(axis cs:25,0.639029559433424)
--(axis cs:24,0.689702705541912)
--(axis cs:23,0.689702705541912)
--(axis cs:22,0.689702705541912)
--(axis cs:21,0.689702705541912)
--(axis cs:20,0.689702705541912)
--(axis cs:19,0.768157099600046)
--(axis cs:18,0.79054918767979)
--(axis cs:17,0.833745774257507)
--(axis cs:16,0.833745774257507)
--(axis cs:15,0.836484322685596)
--(axis cs:14,0.939060655260891)
--(axis cs:13,1.06970075356222)
--(axis cs:12,1.09051764460771)
--(axis cs:11,1.1754970580552)
--(axis cs:10,1.24125538018958)
--cycle;

\path [draw=blue, fill=blue, opacity=0.3]
(axis cs:10,1.241255402565)
--(axis cs:10,1.00074923038483)
--(axis cs:11,0.999580085277557)
--(axis cs:12,0.748606979846954)
--(axis cs:13,0.73406445980072)
--(axis cs:14,0.651569783687592)
--(axis cs:15,0.638103485107422)
--(axis cs:16,0.620693445205688)
--(axis cs:17,0.56354808807373)
--(axis cs:18,0.551998376846313)
--(axis cs:19,0.454945981502533)
--(axis cs:20,0.454945981502533)
--(axis cs:21,0.401033520698547)
--(axis cs:22,0.401033520698547)
--(axis cs:23,0.401033520698547)
--(axis cs:24,0.388648748397827)
--(axis cs:25,0.388648748397827)
--(axis cs:26,0.340034335851669)
--(axis cs:27,0.240012645721436)
--(axis cs:28,0.217757537961006)
--(axis cs:29,0.211496233940125)
--(axis cs:30,0.211496233940125)
--(axis cs:31,0.17584627866745)
--(axis cs:32,0.158419579267502)
--(axis cs:33,0.153859615325928)
--(axis cs:34,0.153859615325928)
--(axis cs:35,0.142365485429764)
--(axis cs:36,0.121059745550156)
--(axis cs:37,0.118939578533173)
--(axis cs:38,0.110578656196594)
--(axis cs:39,0.103164337575436)
--(axis cs:40,0.103164337575436)
--(axis cs:41,0.103164337575436)
--(axis cs:42,0.0885284841060638)
--(axis cs:43,0.083843320608139)
--(axis cs:44,0.0772291943430901)
--(axis cs:45,0.0772291943430901)
--(axis cs:46,0.0772291943430901)
--(axis cs:47,0.0659895241260529)
--(axis cs:48,0.0659895241260529)
--(axis cs:49,0.0659895241260529)
--(axis cs:50,0.0659895241260529)
--(axis cs:51,0.0659895241260529)
--(axis cs:52,0.0599450059235096)
--(axis cs:53,0.0599450059235096)
--(axis cs:54,0.0498202182352543)
--(axis cs:55,0.0498202182352543)
--(axis cs:56,0.0498202182352543)
--(axis cs:57,0.0498202182352543)
--(axis cs:58,0.0462660603225231)
--(axis cs:59,0.0462660603225231)
--(axis cs:60,0.0462660603225231)
--(axis cs:61,0.0462660603225231)
--(axis cs:62,0.0438873544335365)
--(axis cs:63,0.0438873544335365)
--(axis cs:64,0.0438873544335365)
--(axis cs:65,0.0438873544335365)
--(axis cs:66,0.0438873544335365)
--(axis cs:67,0.0428789854049683)
--(axis cs:68,0.0419917739927769)
--(axis cs:69,0.0419917739927769)
--(axis cs:70,0.0419917739927769)
--(axis cs:71,0.0419917739927769)
--(axis cs:72,0.0419917739927769)
--(axis cs:73,0.0419917739927769)
--(axis cs:74,0.0419917739927769)
--(axis cs:75,0.0419917739927769)
--(axis cs:76,0.0419917739927769)
--(axis cs:77,0.0419917739927769)
--(axis cs:78,0.0419917739927769)
--(axis cs:79,0.0419917739927769)
--(axis cs:80,0.0419917739927769)
--(axis cs:81,0.0419917739927769)
--(axis cs:82,0.0419917739927769)
--(axis cs:83,0.0419917739927769)
--(axis cs:84,0.0419917739927769)
--(axis cs:85,0.0419917739927769)
--(axis cs:86,0.0419917739927769)
--(axis cs:87,0.0419917739927769)
--(axis cs:88,0.0419917739927769)
--(axis cs:89,0.0419917739927769)
--(axis cs:90,0.0419917739927769)
--(axis cs:91,0.0419917739927769)
--(axis cs:92,0.0419917739927769)
--(axis cs:93,0.0419917739927769)
--(axis cs:94,0.0419917739927769)
--(axis cs:95,0.0419917739927769)
--(axis cs:96,0.0419917739927769)
--(axis cs:97,0.0419917739927769)
--(axis cs:98,0.0419917739927769)
--(axis cs:99,0.0419917739927769)
--(axis cs:100,0.0419914983212948)
--(axis cs:101,0.0419914983212948)
--(axis cs:102,0.0419914983212948)
--(axis cs:103,0.0419914983212948)
--(axis cs:104,0.0419914983212948)
--(axis cs:105,0.0419914983212948)
--(axis cs:106,0.0419914983212948)
--(axis cs:107,0.0419914983212948)
--(axis cs:108,0.0419914983212948)
--(axis cs:109,0.0419914983212948)
--(axis cs:110,0.0419914983212948)
--(axis cs:111,0.0419914983212948)
--(axis cs:111,0.0569079928100109)
--(axis cs:111,0.0569079928100109)
--(axis cs:110,0.0569079928100109)
--(axis cs:109,0.0569079928100109)
--(axis cs:108,0.0569079928100109)
--(axis cs:107,0.0569079928100109)
--(axis cs:106,0.0569079928100109)
--(axis cs:105,0.0569079928100109)
--(axis cs:104,0.0569079928100109)
--(axis cs:103,0.0569079928100109)
--(axis cs:102,0.0569079928100109)
--(axis cs:101,0.0569079928100109)
--(axis cs:100,0.0569079928100109)
--(axis cs:99,0.0569085888564587)
--(axis cs:98,0.0569085888564587)
--(axis cs:97,0.0569085888564587)
--(axis cs:96,0.0569085888564587)
--(axis cs:95,0.0569085888564587)
--(axis cs:94,0.0569085888564587)
--(axis cs:93,0.0569085888564587)
--(axis cs:92,0.0569085888564587)
--(axis cs:91,0.0569085888564587)
--(axis cs:90,0.0569085888564587)
--(axis cs:89,0.0569085888564587)
--(axis cs:88,0.0569085888564587)
--(axis cs:87,0.0569085888564587)
--(axis cs:86,0.0569085888564587)
--(axis cs:85,0.0569085888564587)
--(axis cs:84,0.0569085888564587)
--(axis cs:83,0.0569085888564587)
--(axis cs:82,0.0569085888564587)
--(axis cs:81,0.0569085888564587)
--(axis cs:80,0.0569085888564587)
--(axis cs:79,0.0569085888564587)
--(axis cs:78,0.0569085888564587)
--(axis cs:77,0.0569085888564587)
--(axis cs:76,0.0569085888564587)
--(axis cs:75,0.0569085888564587)
--(axis cs:74,0.0569085888564587)
--(axis cs:73,0.0569085888564587)
--(axis cs:72,0.0569085888564587)
--(axis cs:71,0.0569085888564587)
--(axis cs:70,0.0569085888564587)
--(axis cs:69,0.0569085888564587)
--(axis cs:68,0.0569085888564587)
--(axis cs:67,0.0593538135290146)
--(axis cs:66,0.0640649273991585)
--(axis cs:65,0.0640649273991585)
--(axis cs:64,0.0640649273991585)
--(axis cs:63,0.0640649273991585)
--(axis cs:62,0.0640649273991585)
--(axis cs:61,0.0694191977381706)
--(axis cs:60,0.0694191977381706)
--(axis cs:59,0.0694191977381706)
--(axis cs:58,0.0694191977381706)
--(axis cs:57,0.0736806765198708)
--(axis cs:56,0.0736806765198708)
--(axis cs:55,0.0736806765198708)
--(axis cs:54,0.0736806765198708)
--(axis cs:53,0.0919819623231888)
--(axis cs:52,0.0919819623231888)
--(axis cs:51,0.112630665302277)
--(axis cs:50,0.112630665302277)
--(axis cs:49,0.112630665302277)
--(axis cs:48,0.112630665302277)
--(axis cs:47,0.112630665302277)
--(axis cs:46,0.123617507517338)
--(axis cs:45,0.123617507517338)
--(axis cs:44,0.123617507517338)
--(axis cs:43,0.127034738659859)
--(axis cs:42,0.138151749968529)
--(axis cs:41,0.163831874728203)
--(axis cs:40,0.163831874728203)
--(axis cs:39,0.163831874728203)
--(axis cs:38,0.210974335670471)
--(axis cs:37,0.221786320209503)
--(axis cs:36,0.248495578765869)
--(axis cs:35,0.30697113275528)
--(axis cs:34,0.314994424581528)
--(axis cs:33,0.314994424581528)
--(axis cs:32,0.319029778242111)
--(axis cs:31,0.345704257488251)
--(axis cs:30,0.380028545856476)
--(axis cs:29,0.380028545856476)
--(axis cs:28,0.384157240390778)
--(axis cs:27,0.409400403499603)
--(axis cs:26,0.541722774505615)
--(axis cs:25,0.623561382293701)
--(axis cs:24,0.623561382293701)
--(axis cs:23,0.656973361968994)
--(axis cs:22,0.656973361968994)
--(axis cs:21,0.656973361968994)
--(axis cs:20,0.711178839206696)
--(axis cs:19,0.711178839206696)
--(axis cs:18,0.844116449356079)
--(axis cs:17,0.868630290031433)
--(axis cs:16,0.905799031257629)
--(axis cs:15,0.922247886657715)
--(axis cs:14,0.93188601732254)
--(axis cs:13,1.01589512825012)
--(axis cs:12,1.0481892824173)
--(axis cs:11,1.23820638656616)
--(axis cs:10,1.241255402565)
--cycle;

\path [draw=color1, fill=color1, opacity=0.3]
(axis cs:10,1.24125538018958)
--(axis cs:10,1.00074916931375)
--(axis cs:11,0.967239681699779)
--(axis cs:12,0.967239681699779)
--(axis cs:13,0.938135883741241)
--(axis cs:14,0.814910772869819)
--(axis cs:15,0.738228342763797)
--(axis cs:16,0.653001500523529)
--(axis cs:17,0.653001500523529)
--(axis cs:18,0.571554652352398)
--(axis cs:19,0.571554652352398)
--(axis cs:20,0.557802998188886)
--(axis cs:21,0.557802998188886)
--(axis cs:22,0.491444438813368)
--(axis cs:23,0.413698287892712)
--(axis cs:24,0.388194514958201)
--(axis cs:25,0.364239282041116)
--(axis cs:26,0.334840502190176)
--(axis cs:27,0.318578592714816)
--(axis cs:28,0.31779162010446)
--(axis cs:29,0.271652496041195)
--(axis cs:30,0.249415845312278)
--(axis cs:31,0.230018488171806)
--(axis cs:32,0.230018488171806)
--(axis cs:33,0.220306795182573)
--(axis cs:34,0.159156683081515)
--(axis cs:35,0.156470434439201)
--(axis cs:36,0.156470434439201)
--(axis cs:37,0.140117408174105)
--(axis cs:38,0.133986444486344)
--(axis cs:39,0.130349808464998)
--(axis cs:40,0.130349808464998)
--(axis cs:41,0.122717545928255)
--(axis cs:42,0.122717545928255)
--(axis cs:43,0.122717545928255)
--(axis cs:44,0.116339588079541)
--(axis cs:45,0.108734970962923)
--(axis cs:46,0.0942634203514159)
--(axis cs:47,0.0856460063911266)
--(axis cs:48,0.0771115261302216)
--(axis cs:49,0.0721893970131167)
--(axis cs:50,0.0701179586654307)
--(axis cs:51,0.0701179586654307)
--(axis cs:52,0.066215501755682)
--(axis cs:53,0.066215501755682)
--(axis cs:54,0.06381601476713)
--(axis cs:55,0.0631609602471258)
--(axis cs:56,0.0511917014920626)
--(axis cs:57,0.0511917014920626)
--(axis cs:58,0.0508542839931662)
--(axis cs:59,0.0508542839931662)
--(axis cs:60,0.0508542839931662)
--(axis cs:61,0.0503734682549043)
--(axis cs:62,0.0482776014455166)
--(axis cs:63,0.0388096385977602)
--(axis cs:64,0.0348198703214882)
--(axis cs:65,0.0348198703214882)
--(axis cs:66,0.0332509217565704)
--(axis cs:67,0.0257308393302977)
--(axis cs:68,0.0171200657462708)
--(axis cs:69,0.0171200657462708)
--(axis cs:70,0.0171200657462708)
--(axis cs:71,0.0168131261457052)
--(axis cs:72,0.0168131261457052)
--(axis cs:73,0.0118830281689806)
--(axis cs:74,0.0118830281689806)
--(axis cs:75,0.0118128669230862)
--(axis cs:76,0.0118128669230862)
--(axis cs:77,0.0118128669230862)
--(axis cs:78,0.0118128669230862)
--(axis cs:79,0.010140778811764)
--(axis cs:80,0.0079823433296055)
--(axis cs:81,0.0079823433296055)
--(axis cs:82,0.0068737677108664)
--(axis cs:83,0.0068737677108664)
--(axis cs:84,0.0068737677108664)
--(axis cs:85,0.00668396969208794)
--(axis cs:86,0.00668396969208794)
--(axis cs:87,0.00668396969208794)
--(axis cs:88,0.00460184225121685)
--(axis cs:89,0.00415132046357608)
--(axis cs:90,0.00415132046357608)
--(axis cs:91,0.00405620281299037)
--(axis cs:92,0.00405620281299037)
--(axis cs:93,0.00405620281299037)
--(axis cs:94,0.00374005253078662)
--(axis cs:95,0.00374005253078662)
--(axis cs:96,0.00374005253078662)
--(axis cs:97,0.00374005253078662)
--(axis cs:98,0.00374005253078662)
--(axis cs:99,0.00374005253078662)
--(axis cs:100,0.00344367057760766)
--(axis cs:101,0.00331205050741953)
--(axis cs:102,0.00331205050741953)
--(axis cs:103,0.00331205050741953)
--(axis cs:104,0.00331205050741953)
--(axis cs:105,0.00331205050741953)
--(axis cs:106,0.00331205050741953)
--(axis cs:107,0.00331205050741953)
--(axis cs:108,0.00331205050741953)
--(axis cs:109,0.00331205050741953)
--(axis cs:110,0.00331205050741953)
--(axis cs:111,0.00331205050741953)
--(axis cs:111,0.010635624338633)
--(axis cs:111,0.010635624338633)
--(axis cs:110,0.010635624338633)
--(axis cs:109,0.010635624338633)
--(axis cs:108,0.010635624338633)
--(axis cs:107,0.010635624338633)
--(axis cs:106,0.010635624338633)
--(axis cs:105,0.010635624338633)
--(axis cs:104,0.010635624338633)
--(axis cs:103,0.010635624338633)
--(axis cs:102,0.010635624338633)
--(axis cs:101,0.010635624338633)
--(axis cs:100,0.0107341201199653)
--(axis cs:99,0.0204265162726125)
--(axis cs:98,0.0204265162726125)
--(axis cs:97,0.0204265162726125)
--(axis cs:96,0.0204265162726125)
--(axis cs:95,0.0204265162726125)
--(axis cs:94,0.0204265162726125)
--(axis cs:93,0.0207026224144495)
--(axis cs:92,0.0207026224144495)
--(axis cs:91,0.0207026224144495)
--(axis cs:90,0.0207896448802597)
--(axis cs:89,0.0207896448802597)
--(axis cs:88,0.0211891189354511)
--(axis cs:87,0.0232229027249591)
--(axis cs:86,0.0232229027249591)
--(axis cs:85,0.0232229027249591)
--(axis cs:84,0.0233906031819596)
--(axis cs:83,0.0233906031819596)
--(axis cs:82,0.0233906031819596)
--(axis cs:81,0.0243883185816053)
--(axis cs:80,0.0243883185816053)
--(axis cs:79,0.0265181488055003)
--(axis cs:78,0.0278129215353053)
--(axis cs:77,0.0278129215353053)
--(axis cs:76,0.0278129215353053)
--(axis cs:75,0.0278129215353053)
--(axis cs:74,0.0278784059955518)
--(axis cs:73,0.0278784059955518)
--(axis cs:72,0.0394322210162988)
--(axis cs:71,0.0394322210162988)
--(axis cs:70,0.0396268867751553)
--(axis cs:69,0.0396268867751553)
--(axis cs:68,0.0396268867751553)
--(axis cs:67,0.0512582453527156)
--(axis cs:66,0.059727763696127)
--(axis cs:65,0.0610677231605638)
--(axis cs:64,0.0610677231605638)
--(axis cs:63,0.075719703850269)
--(axis cs:62,0.0851507261902916)
--(axis cs:61,0.0864077638764324)
--(axis cs:60,0.0866093080014781)
--(axis cs:59,0.0866093080014781)
--(axis cs:58,0.0866093080014781)
--(axis cs:57,0.086773338329917)
--(axis cs:56,0.086773338329917)
--(axis cs:55,0.101472117995397)
--(axis cs:54,0.101717755308142)
--(axis cs:53,0.102689215093723)
--(axis cs:52,0.102689215093723)
--(axis cs:51,0.106647610000485)
--(axis cs:50,0.106647610000485)
--(axis cs:49,0.107402009625308)
--(axis cs:48,0.114785467652086)
--(axis cs:47,0.131005467072523)
--(axis cs:46,0.140259882691591)
--(axis cs:45,0.152216660384908)
--(axis cs:44,0.154713158024023)
--(axis cs:43,0.168444419248259)
--(axis cs:42,0.168444419248259)
--(axis cs:41,0.168444419248259)
--(axis cs:40,0.181384554928469)
--(axis cs:39,0.181384554928469)
--(axis cs:38,0.18336922969744)
--(axis cs:37,0.186388236797905)
--(axis cs:36,0.22055456094773)
--(axis cs:35,0.22055456094773)
--(axis cs:34,0.223714196474802)
--(axis cs:33,0.293542055290176)
--(axis cs:32,0.409696273665244)
--(axis cs:31,0.409696273665244)
--(axis cs:30,0.433452182935337)
--(axis cs:29,0.467070207498914)
--(axis cs:28,0.529024682805707)
--(axis cs:27,0.529630325681452)
--(axis cs:26,0.544223589363344)
--(axis cs:25,0.564427821097834)
--(axis cs:24,0.580672751118639)
--(axis cs:23,0.604352530153143)
--(axis cs:22,0.765728393497996)
--(axis cs:21,0.807542835084093)
--(axis cs:20,0.807542835084093)
--(axis cs:19,0.822454464087742)
--(axis cs:18,0.822454464087742)
--(axis cs:17,0.894630270853251)
--(axis cs:16,0.894630270853251)
--(axis cs:15,0.988250759776194)
--(axis cs:14,1.10719276669695)
--(axis cs:13,1.19299098365925)
--(axis cs:12,1.19821734503193)
--(axis cs:11,1.19821734503193)
--(axis cs:10,1.24125538018958)
--cycle;

\addplot [semithick, blue, dash dot]
table {%
10 1.12100231647491
11 1.02682435512543
12 0.957351207733154
13 0.864063620567322
14 0.812389671802521
15 0.669671952724457
16 0.66400134563446
17 0.57091611623764
18 0.507756114006042
19 0.507756114006042
20 0.507756114006042
21 0.412072658538818
22 0.405438274145126
23 0.388815939426422
24 0.388815939426422
25 0.313590496778488
26 0.306073486804962
27 0.297679126262665
28 0.297679126262665
29 0.242668479681015
30 0.211937382817268
31 0.203431814908981
32 0.188891962170601
33 0.172076016664505
34 0.168261274695396
35 0.164883121848106
36 0.148721307516098
37 0.148721307516098
38 0.147245019674301
39 0.147220492362976
40 0.138467475771904
41 0.138467475771904
42 0.104954719543457
43 0.100353762507439
44 0.0851681083440781
45 0.0818492621183395
46 0.0813206881284714
47 0.0813206881284714
48 0.0802278444170952
49 0.0799515917897224
50 0.0745632499456406
51 0.0745632499456406
52 0.0745632499456406
53 0.0707874149084091
54 0.0689764022827148
55 0.0689764022827148
56 0.0689764022827148
57 0.0689764022827148
58 0.0569745674729347
59 0.0569745674729347
60 0.0569745674729347
61 0.0569745674729347
62 0.0569745674729347
63 0.0569745674729347
64 0.0569745674729347
65 0.0550899319350719
66 0.0550899319350719
67 0.0508101508021355
68 0.0508101508021355
69 0.0462232157588005
70 0.0462232157588005
71 0.0462232157588005
72 0.0462232157588005
73 0.0462232157588005
74 0.0462232157588005
75 0.0462232157588005
76 0.0462232157588005
77 0.0462232157588005
78 0.0462232157588005
79 0.0462232157588005
80 0.0462232157588005
81 0.0458423793315887
82 0.0458423793315887
83 0.0458423793315887
84 0.0458423793315887
85 0.0458423793315887
86 0.0458423793315887
87 0.0456148907542229
88 0.0456148907542229
89 0.0456148907542229
90 0.0456148907542229
91 0.0420631058514118
92 0.0420631058514118
93 0.0405865982174873
94 0.0405865982174873
95 0.0405865982174873
96 0.0405865982174873
97 0.0405866019427776
98 0.0405866019427776
99 0.0405866019427776
100 0.0405866019427776
101 0.0374354347586632
102 0.0374354347586632
103 0.0364506766200066
104 0.0364506766200066
105 0.0364506766200066
106 0.0364506766200066
107 0.0364506766200066
108 0.0348769538104534
109 0.0348769538104534
110 0.0325050875544548
111 0.0325050875544548
};
\addplot [semithick, color1, dash dot]
table {%
10 1.12100227475166
11 1.05892728955996
12 0.991181696655776
13 0.943412270509505
14 0.811677060204672
15 0.704652774689538
16 0.700333666266056
17 0.700333666266056
18 0.656272746829788
19 0.638420191773699
20 0.549924031866445
21 0.549924031866445
22 0.549924031866445
23 0.549924031866445
24 0.549924031866445
25 0.505355755826881
26 0.424531710626938
27 0.391251190505466
28 0.391251190505466
29 0.297959758098071
30 0.267917430398068
31 0.267917430398068
32 0.242148149721978
33 0.22343297863563
34 0.204187561317448
35 0.197719849876463
36 0.155622166673272
37 0.142332072442373
38 0.14105525084118
39 0.0839932683298571
40 0.0839932683298571
41 0.0723443141650572
42 0.0711585883708468
43 0.0711585883708468
44 0.0711585883708468
45 0.0711585883708468
46 0.0711585883708468
47 0.0711585883708468
48 0.0711585883708468
49 0.0681422469121019
50 0.0681422469121019
51 0.0681422469121019
52 0.0681422469121019
53 0.0681422469121019
54 0.0673845047553786
55 0.0607833435696903
56 0.0607833435696903
57 0.0607833435696903
58 0.0607833435696903
59 0.0607833435696903
60 0.0607833435696903
61 0.0607833435696903
62 0.0607833435696903
63 0.0607833435696903
64 0.0607833435696903
65 0.0607833435696903
66 0.0607833435696903
67 0.0607833435696903
68 0.0607833435696903
69 0.0607833435696903
70 0.0607833435696903
71 0.0607833435696903
72 0.0607833435696903
73 0.0607833435696903
74 0.0607833435696903
75 0.0597888093531053
76 0.0597888093531053
77 0.0566999042826272
78 0.0566999042826272
79 0.0566999042826272
80 0.0566549909692135
81 0.0566549909692135
82 0.0566549909692135
83 0.0566549909692135
84 0.0566549909692135
85 0.0541503155562092
86 0.0541503155562092
87 0.0541503155562092
88 0.0536190929598527
89 0.0536190929598527
90 0.0519343386811466
91 0.0519343386811466
92 0.0511472382940549
93 0.0511472382940549
94 0.0511472382940549
95 0.0511472382940549
96 0.0511472382940549
97 0.0511472382940549
98 0.0439792302231719
99 0.0439792302231719
100 0.0439792302231719
101 0.0373154729737259
102 0.0373154729737259
103 0.0373154729737259
104 0.0373154729737259
105 0.0373154729737259
106 0.0373154729737259
107 0.0373154729737259
108 0.0373154729737259
109 0.0373154729737259
110 0.0373154729737259
111 0.0355178076506712
};
\addplot [semithick, blue]
table {%
10 1.12100231647491
11 1.11889326572418
12 0.898398101329803
13 0.874979794025421
14 0.791727900505066
15 0.780175685882568
16 0.763246238231659
17 0.716089189052582
18 0.698057413101196
19 0.583062410354614
20 0.583062410354614
21 0.529003441333771
22 0.529003441333771
23 0.529003441333771
24 0.506105065345764
25 0.506105065345764
26 0.440878570079803
27 0.324706524610519
28 0.300957381725311
29 0.2957623898983
30 0.2957623898983
31 0.26077526807785
32 0.238724678754807
33 0.234427019953728
34 0.234427019953728
35 0.224668309092522
36 0.184777662158012
37 0.170362949371338
38 0.160776495933533
39 0.133498102426529
40 0.133498102426529
41 0.133498102426529
42 0.113340117037296
43 0.105439029633999
44 0.100423350930214
45 0.100423350930214
46 0.100423350930214
47 0.0893100947141647
48 0.0893100947141647
49 0.0893100947141647
50 0.0893100947141647
51 0.0893100947141647
52 0.075963482260704
53 0.075963482260704
54 0.0617504492402077
55 0.0617504492402077
56 0.0617504492402077
57 0.0617504492402077
58 0.057842630892992
59 0.057842630892992
60 0.057842630892992
61 0.057842630892992
62 0.0539761409163475
63 0.0539761409163475
64 0.0539761409163475
65 0.0539761409163475
66 0.0539761409163475
67 0.0511163994669914
68 0.0494501814246178
69 0.0494501814246178
70 0.0494501814246178
71 0.0494501814246178
72 0.0494501814246178
73 0.0494501814246178
74 0.0494501814246178
75 0.0494501814246178
76 0.0494501814246178
77 0.0494501814246178
78 0.0494501814246178
79 0.0494501814246178
80 0.0494501814246178
81 0.0494501814246178
82 0.0494501814246178
83 0.0494501814246178
84 0.0494501814246178
85 0.0494501814246178
86 0.0494501814246178
87 0.0494501814246178
88 0.0494501814246178
89 0.0494501814246178
90 0.0494501814246178
91 0.0494501814246178
92 0.0494501814246178
93 0.0494501814246178
94 0.0494501814246178
95 0.0494501814246178
96 0.0494501814246178
97 0.0494501814246178
98 0.0494501814246178
99 0.0494501814246178
100 0.0494497455656528
101 0.0494497455656528
102 0.0494497455656528
103 0.0494497455656528
104 0.0494497455656528
105 0.0494497455656528
106 0.0494497455656528
107 0.0494497455656528
108 0.0494497455656528
109 0.0494497455656528
110 0.0494497455656528
111 0.0494497455656528
};
\addplot [semithick, color1]
table {%
10 1.12100227475166
11 1.08272851336586
12 1.08272851336586
13 1.06556343370024
14 0.961051769783383
15 0.863239551269996
16 0.77381588568839
17 0.77381588568839
18 0.69700455822007
19 0.69700455822007
20 0.682672916636489
21 0.682672916636489
22 0.628586416155682
23 0.509025409022927
24 0.48443363303842
25 0.464333551569475
26 0.43953204577676
27 0.424104459198134
28 0.423408151455083
29 0.369361351770054
30 0.341434014123808
31 0.319857380918525
32 0.319857380918525
33 0.256924425236375
34 0.191435439778159
35 0.188512497693466
36 0.188512497693466
37 0.163252822486005
38 0.158677837091892
39 0.155867181696733
40 0.155867181696733
41 0.145580982588257
42 0.145580982588257
43 0.145580982588257
44 0.135526373051782
45 0.130475815673916
46 0.117261651521504
47 0.108325736731825
48 0.0959484968911539
49 0.0897957033192123
50 0.0883827843329581
51 0.0883827843329581
52 0.0844523584247026
53 0.0844523584247026
54 0.0827668850376361
55 0.0823165391212614
56 0.0689825199109898
57 0.0689825199109898
58 0.0687317959973221
59 0.0687317959973221
60 0.0687317959973221
61 0.0683906160656683
62 0.0667141638179041
63 0.0572646712240146
64 0.047943796741026
65 0.047943796741026
66 0.0464893427263487
67 0.0384945423415066
68 0.028373476260713
69 0.028373476260713
70 0.028373476260713
71 0.028122673581002
72 0.028122673581002
73 0.0198807170822662
74 0.0198807170822662
75 0.0198128942291957
76 0.0198128942291957
77 0.0198128942291957
78 0.0198128942291957
79 0.0183294638086321
80 0.0161853309556054
81 0.0161853309556054
82 0.015132185446413
83 0.015132185446413
84 0.015132185446413
85 0.0149534362085235
86 0.0149534362085235
87 0.0149534362085235
88 0.012895480593334
89 0.0124704826719179
90 0.0124704826719179
91 0.0123794126137199
92 0.0123794126137199
93 0.0123794126137199
94 0.0120832844016996
95 0.0120832844016996
96 0.0120832844016996
97 0.0120832844016996
98 0.0120832844016996
99 0.0120832844016996
100 0.00708889534878647
101 0.00697383742302626
102 0.00697383742302626
103 0.00697383742302626
104 0.00697383742302626
105 0.00697383742302626
106 0.00697383742302626
107 0.00697383742302626
108 0.00697383742302626
109 0.00697383742302626
110 0.00697383742302626
111 0.00697383742302626
};
\end{axis}

\end{tikzpicture}

%% file: figures/bop_3d_levy_strong.tex
\begin{tikzpicture}

\definecolor{color0}{rgb}{0,0,1}
\definecolor{color1}{rgb}{1,0.549019607843137,0}
\definecolor{color2}{rgb}{1,0.647058823529412,0}
\definecolor{color3}{rgb}{0.564705882352941,0.933333333333333,0.564705882352941}

\begin{axis}[axis on top,
enlarge x limits=false,
enlarge y limits=false,
height=\figureheight,
scale only axis,
tick align=outside,
tick pos=left,
tick pos=left,
width=\figurewidth,
xlabel={Iteration},
xmin=10, xmax=60,
xtick style={color=black},
xtick={-10,0,10,25,50,75,100},
xticklabels={\ensuremath{-}10,0,10,25,50,75,90},
ymin=-0.05, ymax=1.3,
ytick style={color=black},
ytick={0.   , 1.3},
yticklabels={},
]
\node[anchor=north east] at (rel axis cs:1,1) {Levy 3D (strong)};
\path [draw=blue, fill=blue, opacity=0.3]
(axis cs:10,1.241255402565)
--(axis cs:10,1.00074923038483)
--(axis cs:11,0.824178695678711)
--(axis cs:12,0.747050344944)
--(axis cs:13,0.665185272693634)
--(axis cs:14,0.477369606494904)
--(axis cs:15,0.406069815158844)
--(axis cs:16,0.27922049164772)
--(axis cs:17,0.2685686647892)
--(axis cs:18,0.265221685171127)
--(axis cs:19,0.222272112965584)
--(axis cs:20,0.207131773233414)
--(axis cs:21,0.179898321628571)
--(axis cs:22,0.142645791172981)
--(axis cs:23,0.129276722669601)
--(axis cs:24,0.1285190731287)
--(axis cs:25,0.122566297650337)
--(axis cs:26,0.101503141224384)
--(axis cs:27,0.09821567684412)
--(axis cs:28,0.0914435088634491)
--(axis cs:29,0.0914435088634491)
--(axis cs:30,0.0821532160043716)
--(axis cs:31,0.0744810700416565)
--(axis cs:32,0.074143573641777)
--(axis cs:33,0.074143573641777)
--(axis cs:34,0.074143573641777)
--(axis cs:35,0.0673811733722687)
--(axis cs:36,0.0673811733722687)
--(axis cs:37,0.0673811733722687)
--(axis cs:38,0.0673811733722687)
--(axis cs:39,0.049337062984705)
--(axis cs:40,0.049337062984705)
--(axis cs:41,0.049337062984705)
--(axis cs:42,0.049337062984705)
--(axis cs:43,0.0446456931531429)
--(axis cs:44,0.0446456931531429)
--(axis cs:45,0.0295812152326107)
--(axis cs:46,0.0295175425708294)
--(axis cs:47,0.0295175425708294)
--(axis cs:48,0.028259165585041)
--(axis cs:49,0.028259165585041)
--(axis cs:50,0.028259165585041)
--(axis cs:51,0.028259165585041)
--(axis cs:52,0.028259165585041)
--(axis cs:53,0.0276290718466043)
--(axis cs:54,0.0257484614849091)
--(axis cs:55,0.0257484614849091)
--(axis cs:56,0.0257484614849091)
--(axis cs:57,0.0257484614849091)
--(axis cs:58,0.0257484614849091)
--(axis cs:59,0.0212580729275942)
--(axis cs:60,0.0212580729275942)
--(axis cs:61,0.0212580729275942)
--(axis cs:62,0.0212580729275942)
--(axis cs:63,0.0212580729275942)
--(axis cs:64,0.0212580729275942)
--(axis cs:65,0.0212580729275942)
--(axis cs:66,0.0212580729275942)
--(axis cs:67,0.0189619474112988)
--(axis cs:68,0.0189619474112988)
--(axis cs:69,0.0189619474112988)
--(axis cs:70,0.0189619474112988)
--(axis cs:71,0.0189619474112988)
--(axis cs:72,0.0189619474112988)
--(axis cs:73,0.0189619474112988)
--(axis cs:74,0.0189619474112988)
--(axis cs:75,0.0189619474112988)
--(axis cs:76,0.0189619474112988)
--(axis cs:77,0.0189619474112988)
--(axis cs:78,0.0189619474112988)
--(axis cs:79,0.017669441178441)
--(axis cs:80,0.017669441178441)
--(axis cs:81,0.0163892358541489)
--(axis cs:82,0.0163892358541489)
--(axis cs:83,0.0163892358541489)
--(axis cs:84,0.0163892358541489)
--(axis cs:85,0.0163892358541489)
--(axis cs:86,0.0163892358541489)
--(axis cs:87,0.0163892358541489)
--(axis cs:88,0.0163892358541489)
--(axis cs:89,0.0163892358541489)
--(axis cs:90,0.0163892358541489)
--(axis cs:91,0.0163892358541489)
--(axis cs:92,0.0163892358541489)
--(axis cs:93,0.0163892358541489)
--(axis cs:94,0.0163892358541489)
--(axis cs:95,0.0163892358541489)
--(axis cs:96,0.0163892358541489)
--(axis cs:97,0.0163892358541489)
--(axis cs:98,0.0163892358541489)
--(axis cs:99,0.0163892358541489)
--(axis cs:100,0.0163892358541489)
--(axis cs:101,0.0163892358541489)
--(axis cs:102,0.0163892358541489)
--(axis cs:103,0.0163892358541489)
--(axis cs:104,0.0163892358541489)
--(axis cs:105,0.0163892358541489)
--(axis cs:106,0.0163892358541489)
--(axis cs:107,0.0163892358541489)
--(axis cs:108,0.0163892358541489)
--(axis cs:109,0.0163892358541489)
--(axis cs:110,0.0163892358541489)
--(axis cs:111,0.0163892358541489)
--(axis cs:111,0.0327556170523167)
--(axis cs:111,0.0327556170523167)
--(axis cs:110,0.0327556170523167)
--(axis cs:109,0.0327556170523167)
--(axis cs:108,0.0327556170523167)
--(axis cs:107,0.0327556170523167)
--(axis cs:106,0.0327556170523167)
--(axis cs:105,0.0327556170523167)
--(axis cs:104,0.0327556170523167)
--(axis cs:103,0.0327556170523167)
--(axis cs:102,0.0327556170523167)
--(axis cs:101,0.0327556170523167)
--(axis cs:100,0.0327556170523167)
--(axis cs:99,0.0327556170523167)
--(axis cs:98,0.0327556170523167)
--(axis cs:97,0.0327556170523167)
--(axis cs:96,0.0327556170523167)
--(axis cs:95,0.0327556170523167)
--(axis cs:94,0.0327556170523167)
--(axis cs:93,0.0327556170523167)
--(axis cs:92,0.0327556170523167)
--(axis cs:91,0.0327556170523167)
--(axis cs:90,0.0327556170523167)
--(axis cs:89,0.0327556170523167)
--(axis cs:88,0.0327556170523167)
--(axis cs:87,0.0327556170523167)
--(axis cs:86,0.0327556170523167)
--(axis cs:85,0.0327556170523167)
--(axis cs:84,0.0327556170523167)
--(axis cs:83,0.0327556170523167)
--(axis cs:82,0.0327556170523167)
--(axis cs:81,0.0327556170523167)
--(axis cs:80,0.0336051061749458)
--(axis cs:79,0.0336051061749458)
--(axis cs:78,0.03499661013484)
--(axis cs:77,0.03499661013484)
--(axis cs:76,0.03499661013484)
--(axis cs:75,0.03499661013484)
--(axis cs:74,0.03499661013484)
--(axis cs:73,0.03499661013484)
--(axis cs:72,0.03499661013484)
--(axis cs:71,0.03499661013484)
--(axis cs:70,0.03499661013484)
--(axis cs:69,0.03499661013484)
--(axis cs:68,0.03499661013484)
--(axis cs:67,0.03499661013484)
--(axis cs:66,0.0372256487607956)
--(axis cs:65,0.0372256487607956)
--(axis cs:64,0.0372256487607956)
--(axis cs:63,0.0372256487607956)
--(axis cs:62,0.0372256487607956)
--(axis cs:61,0.0372256487607956)
--(axis cs:60,0.0372256487607956)
--(axis cs:59,0.0372256487607956)
--(axis cs:58,0.0411982387304306)
--(axis cs:57,0.0411982387304306)
--(axis cs:56,0.0411982387304306)
--(axis cs:55,0.0411982387304306)
--(axis cs:54,0.0411982387304306)
--(axis cs:53,0.0425042361021042)
--(axis cs:52,0.0432057976722717)
--(axis cs:51,0.0432057976722717)
--(axis cs:50,0.0432057976722717)
--(axis cs:49,0.0432057976722717)
--(axis cs:48,0.0432057976722717)
--(axis cs:47,0.0969254821538925)
--(axis cs:46,0.0969254821538925)
--(axis cs:45,0.114041447639465)
--(axis cs:44,0.127908989787102)
--(axis cs:43,0.127908989787102)
--(axis cs:42,0.131505012512207)
--(axis cs:41,0.131505012512207)
--(axis cs:40,0.131505012512207)
--(axis cs:39,0.131505012512207)
--(axis cs:38,0.145651400089264)
--(axis cs:37,0.145651400089264)
--(axis cs:36,0.145651400089264)
--(axis cs:35,0.145651400089264)
--(axis cs:34,0.150736123323441)
--(axis cs:33,0.150736123323441)
--(axis cs:32,0.150736123323441)
--(axis cs:31,0.151040583848953)
--(axis cs:30,0.159830003976822)
--(axis cs:29,0.167516350746155)
--(axis cs:28,0.167516350746155)
--(axis cs:27,0.172265112400055)
--(axis cs:26,0.175867319107056)
--(axis cs:25,0.207855001091957)
--(axis cs:24,0.214977070689201)
--(axis cs:23,0.217671662569046)
--(axis cs:22,0.239235296845436)
--(axis cs:21,0.270145505666733)
--(axis cs:20,0.328496545553207)
--(axis cs:19,0.368585467338562)
--(axis cs:18,0.419359475374222)
--(axis cs:17,0.423642009496689)
--(axis cs:16,0.435357064008713)
--(axis cs:15,0.616479337215424)
--(axis cs:14,0.671205580234528)
--(axis cs:13,0.945015370845795)
--(axis cs:12,1.0340838432312)
--(axis cs:11,1.09760451316833)
--(axis cs:10,1.241255402565)
--cycle;

\path [draw=color1, fill=color1, opacity=0.3]
(axis cs:10,1.24125538018958)
--(axis cs:10,1.00074916931375)
--(axis cs:11,0.768751540242032)
--(axis cs:12,0.702211476830987)
--(axis cs:13,0.516950687035073)
--(axis cs:14,0.422428808994938)
--(axis cs:15,0.422428808994938)
--(axis cs:16,0.260017788322304)
--(axis cs:17,0.233474758891723)
--(axis cs:18,0.224427379916708)
--(axis cs:19,0.202936835416436)
--(axis cs:20,0.197153529734285)
--(axis cs:21,0.166988687087582)
--(axis cs:22,0.158551925473255)
--(axis cs:23,0.142644143566397)
--(axis cs:24,0.142593318984604)
--(axis cs:25,0.132877287078616)
--(axis cs:26,0.114209814432443)
--(axis cs:27,0.079127618460459)
--(axis cs:28,0.079127618460459)
--(axis cs:29,0.0691416227428864)
--(axis cs:30,0.0508110439278068)
--(axis cs:31,0.0508110439278068)
--(axis cs:32,0.0444078610662756)
--(axis cs:33,0.0444078610662756)
--(axis cs:34,0.0444078610662756)
--(axis cs:35,0.0444078610662756)
--(axis cs:36,0.0444078610662756)
--(axis cs:37,0.0373019483468081)
--(axis cs:38,0.0373019483468081)
--(axis cs:39,0.0373019483468081)
--(axis cs:40,0.0373019483468081)
--(axis cs:41,0.0333206987073508)
--(axis cs:42,0.0314823786886378)
--(axis cs:43,0.0314823786886378)
--(axis cs:44,0.0314823786886378)
--(axis cs:45,0.0314823786886378)
--(axis cs:46,0.0314823786886378)
--(axis cs:47,0.0314823786886378)
--(axis cs:48,0.0314823786886378)
--(axis cs:49,0.0314823786886378)
--(axis cs:50,0.0314823786886378)
--(axis cs:51,0.0314823786886378)
--(axis cs:52,0.0314823786886378)
--(axis cs:53,0.0294709673431068)
--(axis cs:54,0.0294709673431068)
--(axis cs:55,0.0294709673431068)
--(axis cs:56,0.0294709673431068)
--(axis cs:57,0.0234597039753956)
--(axis cs:58,0.0234597039753956)
--(axis cs:59,0.0234597039753956)
--(axis cs:60,0.0234597039753956)
--(axis cs:61,0.0234597039753956)
--(axis cs:62,0.0234597039753956)
--(axis cs:63,0.0234597039753956)
--(axis cs:64,0.0234597039753956)
--(axis cs:65,0.0234597039753956)
--(axis cs:66,0.0234597039753956)
--(axis cs:67,0.0225133155227876)
--(axis cs:68,0.0225133155227876)
--(axis cs:69,0.0225133155227876)
--(axis cs:70,0.0225133155227876)
--(axis cs:71,0.0225133155227876)
--(axis cs:72,0.0225133155227876)
--(axis cs:73,0.0225133155227876)
--(axis cs:74,0.0221822804205987)
--(axis cs:75,0.0221822804205987)
--(axis cs:76,0.0221822804205987)
--(axis cs:77,0.0221822804205987)
--(axis cs:78,0.0207306345005861)
--(axis cs:79,0.0207306345005861)
--(axis cs:80,0.0207306345005861)
--(axis cs:81,0.0193561341160943)
--(axis cs:82,0.0193561341160943)
--(axis cs:83,0.0193561341160943)
--(axis cs:84,0.0193561341160943)
--(axis cs:85,0.0187330300431957)
--(axis cs:86,0.0187330300431957)
--(axis cs:87,0.0174149688069116)
--(axis cs:88,0.0174149688069116)
--(axis cs:89,0.0174149688069116)
--(axis cs:90,0.0174149688069116)
--(axis cs:91,0.0174149688069116)
--(axis cs:92,0.0166213902962168)
--(axis cs:93,0.0166213902962168)
--(axis cs:94,0.0166213902962168)
--(axis cs:95,0.0166213902962168)
--(axis cs:96,0.0159328774304873)
--(axis cs:97,0.0159328774304873)
--(axis cs:98,0.0159328774304873)
--(axis cs:99,0.0159328774304873)
--(axis cs:100,0.0159328774304873)
--(axis cs:101,0.0159328774304873)
--(axis cs:102,0.0159328774304873)
--(axis cs:103,0.0159328774304873)
--(axis cs:104,0.0159328774304873)
--(axis cs:105,0.0159328774304873)
--(axis cs:106,0.0159328774304873)
--(axis cs:107,0.0159328774304873)
--(axis cs:108,0.0159328774304873)
--(axis cs:109,0.0159328774304873)
--(axis cs:110,0.0159328774304873)
--(axis cs:111,0.0159328774304873)
--(axis cs:111,0.0241139641196146)
--(axis cs:111,0.0241139641196146)
--(axis cs:110,0.0241139641196146)
--(axis cs:109,0.0241139641196146)
--(axis cs:108,0.0241139641196146)
--(axis cs:107,0.0241139641196146)
--(axis cs:106,0.0241139641196146)
--(axis cs:105,0.0241139641196146)
--(axis cs:104,0.0241139641196146)
--(axis cs:103,0.0241139641196146)
--(axis cs:102,0.0241139641196146)
--(axis cs:101,0.0241139641196146)
--(axis cs:100,0.0241139641196146)
--(axis cs:99,0.0241139641196146)
--(axis cs:98,0.0241139641196146)
--(axis cs:97,0.0241139641196146)
--(axis cs:96,0.0241139641196146)
--(axis cs:95,0.0254702217253612)
--(axis cs:94,0.0254702217253612)
--(axis cs:93,0.0254702217253612)
--(axis cs:92,0.0254702217253612)
--(axis cs:91,0.0270327940541424)
--(axis cs:90,0.0270327940541424)
--(axis cs:89,0.0270327940541424)
--(axis cs:88,0.0270327940541424)
--(axis cs:87,0.0270327940541424)
--(axis cs:86,0.0282820744923511)
--(axis cs:85,0.0282820744923511)
--(axis cs:84,0.0288338979859886)
--(axis cs:83,0.0288338979859886)
--(axis cs:82,0.0288338979859886)
--(axis cs:81,0.0288338979859886)
--(axis cs:80,0.0305320939505085)
--(axis cs:79,0.0305320939505085)
--(axis cs:78,0.0305320939505085)
--(axis cs:77,0.0349728082275997)
--(axis cs:76,0.0349728082275997)
--(axis cs:75,0.0349728082275997)
--(axis cs:74,0.0349728082275997)
--(axis cs:73,0.037031376789532)
--(axis cs:72,0.037031376789532)
--(axis cs:71,0.037031376789532)
--(axis cs:70,0.037031376789532)
--(axis cs:69,0.037031376789532)
--(axis cs:68,0.037031376789532)
--(axis cs:67,0.037031376789532)
--(axis cs:66,0.0386255777600947)
--(axis cs:65,0.0386255777600947)
--(axis cs:64,0.0386255777600947)
--(axis cs:63,0.0386255777600947)
--(axis cs:62,0.0386255777600947)
--(axis cs:61,0.0386255777600947)
--(axis cs:60,0.0386255777600947)
--(axis cs:59,0.0386255777600947)
--(axis cs:58,0.0386255777600947)
--(axis cs:57,0.0386255777600947)
--(axis cs:56,0.0483454534796598)
--(axis cs:55,0.0483454534796598)
--(axis cs:54,0.0483454534796598)
--(axis cs:53,0.0483454534796598)
--(axis cs:52,0.0500750935312484)
--(axis cs:51,0.0500750935312484)
--(axis cs:50,0.0500750935312484)
--(axis cs:49,0.0500750935312484)
--(axis cs:48,0.0500750935312484)
--(axis cs:47,0.0500750935312484)
--(axis cs:46,0.0500750935312484)
--(axis cs:45,0.0500750935312484)
--(axis cs:44,0.0500750935312484)
--(axis cs:43,0.0500750935312484)
--(axis cs:42,0.0500750935312484)
--(axis cs:41,0.0528058175445722)
--(axis cs:40,0.0594936666449602)
--(axis cs:39,0.0594936666449602)
--(axis cs:38,0.0594936666449602)
--(axis cs:37,0.0594936666449602)
--(axis cs:36,0.0732380052841296)
--(axis cs:35,0.0732380052841296)
--(axis cs:34,0.0732380052841296)
--(axis cs:33,0.0732380052841296)
--(axis cs:32,0.0732380052841296)
--(axis cs:31,0.0835890604217988)
--(axis cs:30,0.0835890604217988)
--(axis cs:29,0.102962604891931)
--(axis cs:28,0.13127344500449)
--(axis cs:27,0.13127344500449)
--(axis cs:26,0.181952390176401)
--(axis cs:25,0.225990903106018)
--(axis cs:24,0.237749204895224)
--(axis cs:23,0.237986563761466)
--(axis cs:22,0.262069619218558)
--(axis cs:21,0.284129183713775)
--(axis cs:20,0.32132215645909)
--(axis cs:19,0.335538886803223)
--(axis cs:18,0.394429954351809)
--(axis cs:17,0.399840749076316)
--(axis cs:16,0.437349596078332)
--(axis cs:15,0.646978741899259)
--(axis cs:14,0.646978741899259)
--(axis cs:13,0.808871376765326)
--(axis cs:12,0.942541208059233)
--(axis cs:11,1.01782337813834)
--(axis cs:10,1.24125538018958)
--cycle;

\path [draw=blue, fill=blue, opacity=0.3]
(axis cs:10,1.241255402565)
--(axis cs:10,1.00074923038483)
--(axis cs:11,0.999580085277557)
--(axis cs:12,0.844946622848511)
--(axis cs:13,0.726391911506653)
--(axis cs:14,0.644058585166931)
--(axis cs:15,0.630677759647369)
--(axis cs:16,0.61331969499588)
--(axis cs:17,0.544993042945862)
--(axis cs:18,0.461352527141571)
--(axis cs:19,0.412569046020508)
--(axis cs:20,0.412569046020508)
--(axis cs:21,0.362006187438965)
--(axis cs:22,0.345583081245422)
--(axis cs:23,0.345583081245422)
--(axis cs:24,0.344801306724548)
--(axis cs:25,0.344801306724548)
--(axis cs:26,0.285745561122894)
--(axis cs:27,0.241023555397987)
--(axis cs:28,0.218768358230591)
--(axis cs:29,0.212496355175972)
--(axis cs:30,0.212496355175972)
--(axis cs:31,0.176824480295181)
--(axis cs:32,0.15942195057869)
--(axis cs:33,0.157897293567657)
--(axis cs:34,0.157897293567657)
--(axis cs:35,0.146343618631363)
--(axis cs:36,0.125006049871445)
--(axis cs:37,0.122938513755798)
--(axis cs:38,0.112037368118763)
--(axis cs:39,0.104808926582336)
--(axis cs:40,0.104808926582336)
--(axis cs:41,0.104808926582336)
--(axis cs:42,0.0963888019323349)
--(axis cs:43,0.0889857709407806)
--(axis cs:44,0.0822618305683136)
--(axis cs:45,0.0822618305683136)
--(axis cs:46,0.0822618305683136)
--(axis cs:47,0.0707437694072723)
--(axis cs:48,0.0707437694072723)
--(axis cs:49,0.0707437694072723)
--(axis cs:50,0.0707437694072723)
--(axis cs:51,0.0707437694072723)
--(axis cs:52,0.0639458894729614)
--(axis cs:53,0.0639458894729614)
--(axis cs:54,0.0527873449027538)
--(axis cs:55,0.0527873449027538)
--(axis cs:56,0.0527873449027538)
--(axis cs:57,0.0527873449027538)
--(axis cs:58,0.0490056797862053)
--(axis cs:59,0.0490056797862053)
--(axis cs:60,0.0490056797862053)
--(axis cs:61,0.0490056797862053)
--(axis cs:62,0.0461421757936478)
--(axis cs:63,0.0461421757936478)
--(axis cs:64,0.0461421757936478)
--(axis cs:65,0.0461421757936478)
--(axis cs:66,0.0461421757936478)
--(axis cs:67,0.0461421757936478)
--(axis cs:68,0.0461421757936478)
--(axis cs:69,0.0461421757936478)
--(axis cs:70,0.0461421757936478)
--(axis cs:71,0.0461421757936478)
--(axis cs:72,0.0461421757936478)
--(axis cs:73,0.0461421757936478)
--(axis cs:74,0.0461421757936478)
--(axis cs:75,0.0461421757936478)
--(axis cs:76,0.0461421757936478)
--(axis cs:77,0.0461421757936478)
--(axis cs:78,0.0451078750193119)
--(axis cs:79,0.0451078750193119)
--(axis cs:80,0.0451078750193119)
--(axis cs:81,0.0451078750193119)
--(axis cs:82,0.0451078750193119)
--(axis cs:83,0.0441576540470123)
--(axis cs:84,0.0441576540470123)
--(axis cs:85,0.0441576540470123)
--(axis cs:86,0.0441576540470123)
--(axis cs:87,0.0441576540470123)
--(axis cs:88,0.0441576540470123)
--(axis cs:89,0.0441576540470123)
--(axis cs:90,0.0441576540470123)
--(axis cs:91,0.0441576540470123)
--(axis cs:92,0.0441576540470123)
--(axis cs:93,0.0441576540470123)
--(axis cs:94,0.0441576540470123)
--(axis cs:95,0.0441576540470123)
--(axis cs:96,0.0387669093906879)
--(axis cs:97,0.0387669093906879)
--(axis cs:98,0.0387669093906879)
--(axis cs:99,0.0387669093906879)
--(axis cs:100,0.0387666486203671)
--(axis cs:101,0.0387666486203671)
--(axis cs:102,0.0387666486203671)
--(axis cs:103,0.0387666486203671)
--(axis cs:104,0.0387666486203671)
--(axis cs:105,0.0387666486203671)
--(axis cs:106,0.0387666486203671)
--(axis cs:107,0.0387666486203671)
--(axis cs:108,0.0387666486203671)
--(axis cs:109,0.0387666486203671)
--(axis cs:110,0.0387666486203671)
--(axis cs:111,0.0387666486203671)
--(axis cs:111,0.0538482330739498)
--(axis cs:111,0.0538482330739498)
--(axis cs:110,0.0538482330739498)
--(axis cs:109,0.0538482330739498)
--(axis cs:108,0.0538482330739498)
--(axis cs:107,0.0538482330739498)
--(axis cs:106,0.0538482330739498)
--(axis cs:105,0.0538482330739498)
--(axis cs:104,0.0538482330739498)
--(axis cs:103,0.0538482330739498)
--(axis cs:102,0.0538482330739498)
--(axis cs:101,0.0538482330739498)
--(axis cs:100,0.0538482330739498)
--(axis cs:99,0.0538488514721394)
--(axis cs:98,0.0538488514721394)
--(axis cs:97,0.0538488514721394)
--(axis cs:96,0.0538488514721394)
--(axis cs:95,0.0639441087841988)
--(axis cs:94,0.0639441087841988)
--(axis cs:93,0.0639441087841988)
--(axis cs:92,0.0639441087841988)
--(axis cs:91,0.0639441087841988)
--(axis cs:90,0.0639441087841988)
--(axis cs:89,0.0639441087841988)
--(axis cs:88,0.0639441087841988)
--(axis cs:87,0.0639441087841988)
--(axis cs:86,0.0639441087841988)
--(axis cs:85,0.0639441087841988)
--(axis cs:84,0.0639441087841988)
--(axis cs:83,0.0639441087841988)
--(axis cs:82,0.0703798234462738)
--(axis cs:81,0.0703798234462738)
--(axis cs:80,0.0703798234462738)
--(axis cs:79,0.0703798234462738)
--(axis cs:78,0.0703798234462738)
--(axis cs:77,0.072638601064682)
--(axis cs:76,0.072638601064682)
--(axis cs:75,0.072638601064682)
--(axis cs:74,0.072638601064682)
--(axis cs:73,0.072638601064682)
--(axis cs:72,0.072638601064682)
--(axis cs:71,0.072638601064682)
--(axis cs:70,0.072638601064682)
--(axis cs:69,0.072638601064682)
--(axis cs:68,0.072638601064682)
--(axis cs:67,0.072638601064682)
--(axis cs:66,0.072638601064682)
--(axis cs:65,0.072638601064682)
--(axis cs:64,0.072638601064682)
--(axis cs:63,0.072638601064682)
--(axis cs:62,0.072638601064682)
--(axis cs:61,0.0775080695748329)
--(axis cs:60,0.0775080695748329)
--(axis cs:59,0.0775080695748329)
--(axis cs:58,0.0775080695748329)
--(axis cs:57,0.081542044878006)
--(axis cs:56,0.081542044878006)
--(axis cs:55,0.081542044878006)
--(axis cs:54,0.081542044878006)
--(axis cs:53,0.0988095551729202)
--(axis cs:52,0.0988095551729202)
--(axis cs:51,0.118704915046692)
--(axis cs:50,0.118704915046692)
--(axis cs:49,0.118704915046692)
--(axis cs:48,0.118704915046692)
--(axis cs:47,0.118704915046692)
--(axis cs:46,0.129413366317749)
--(axis cs:45,0.129413366317749)
--(axis cs:44,0.129413366317749)
--(axis cs:43,0.132720783352852)
--(axis cs:42,0.147337257862091)
--(axis cs:41,0.164234459400177)
--(axis cs:40,0.164234459400177)
--(axis cs:39,0.164234459400177)
--(axis cs:38,0.211562842130661)
--(axis cs:37,0.224385678768158)
--(axis cs:36,0.251147598028183)
--(axis cs:35,0.309591382741928)
--(axis cs:34,0.317555129528046)
--(axis cs:33,0.317555129528046)
--(axis cs:32,0.318761736154556)
--(axis cs:31,0.345460385084152)
--(axis cs:30,0.379762768745422)
--(axis cs:29,0.379762768745422)
--(axis cs:28,0.38388067483902)
--(axis cs:27,0.409123778343201)
--(axis cs:26,0.461376130580902)
--(axis cs:25,0.559968292713165)
--(axis cs:24,0.559968292713165)
--(axis cs:23,0.565273284912109)
--(axis cs:22,0.565273284912109)
--(axis cs:21,0.575784206390381)
--(axis cs:20,0.633339285850525)
--(axis cs:19,0.633339285850525)
--(axis cs:18,0.752982199192047)
--(axis cs:17,0.843712687492371)
--(axis cs:16,0.907344281673431)
--(axis cs:15,0.923845112323761)
--(axis cs:14,0.933568835258484)
--(axis cs:13,1.01773917675018)
--(axis cs:12,1.13459861278534)
--(axis cs:11,1.23820638656616)
--(axis cs:10,1.241255402565)
--cycle;

\path [draw=color1, fill=color1, opacity=0.3]
(axis cs:10,1.24125538018958)
--(axis cs:10,1.00074916931375)
--(axis cs:11,0.967239681699779)
--(axis cs:12,0.967239681699779)
--(axis cs:13,0.938135883741241)
--(axis cs:14,0.814910772869819)
--(axis cs:15,0.785666964755436)
--(axis cs:16,0.696000922724387)
--(axis cs:17,0.514701554451467)
--(axis cs:18,0.443594220295908)
--(axis cs:19,0.443594220295908)
--(axis cs:20,0.431623136947889)
--(axis cs:21,0.431623136947889)
--(axis cs:22,0.370880121639137)
--(axis cs:23,0.341643306008074)
--(axis cs:24,0.341643306008074)
--(axis cs:25,0.339306574838175)
--(axis cs:26,0.310552059322651)
--(axis cs:27,0.291410440054152)
--(axis cs:28,0.270193943020425)
--(axis cs:29,0.270193943020425)
--(axis cs:30,0.247648129808963)
--(axis cs:31,0.228091511103859)
--(axis cs:32,0.228091511103859)
--(axis cs:33,0.212695548774206)
--(axis cs:34,0.130378010717987)
--(axis cs:35,0.127907238642532)
--(axis cs:36,0.127907238642532)
--(axis cs:37,0.113361378735466)
--(axis cs:38,0.10780986812745)
--(axis cs:39,0.0962426953901579)
--(axis cs:40,0.0962426953901579)
--(axis cs:41,0.0902234771499113)
--(axis cs:42,0.0883971265496445)
--(axis cs:43,0.0883971265496445)
--(axis cs:44,0.0883971265496445)
--(axis cs:45,0.0816975262735178)
--(axis cs:46,0.0711131263526385)
--(axis cs:47,0.070197651176121)
--(axis cs:48,0.062179771631066)
--(axis cs:49,0.0571847757710874)
--(axis cs:50,0.0552670248900386)
--(axis cs:51,0.0552670248900386)
--(axis cs:52,0.051675623138445)
--(axis cs:53,0.051675623138445)
--(axis cs:54,0.0494359695504772)
--(axis cs:55,0.048822304894465)
--(axis cs:56,0.046113546080677)
--(axis cs:57,0.046113546080677)
--(axis cs:58,0.0457862487379286)
--(axis cs:59,0.0457862487379286)
--(axis cs:60,0.0457862487379286)
--(axis cs:61,0.0453199676480807)
--(axis cs:62,0.0453199676480807)
--(axis cs:63,0.0359975787252944)
--(axis cs:64,0.0321219625821009)
--(axis cs:65,0.0321219625821009)
--(axis cs:66,0.030585357257497)
--(axis cs:67,0.0241766771465902)
--(axis cs:68,0.015788809239085)
--(axis cs:69,0.015788809239085)
--(axis cs:70,0.015788809239085)
--(axis cs:71,0.0154840415168857)
--(axis cs:72,0.0154840415168857)
--(axis cs:73,0.0110006485917678)
--(axis cs:74,0.0110006485917678)
--(axis cs:75,0.0109321110355865)
--(axis cs:76,0.0109321110355865)
--(axis cs:77,0.0109321110355865)
--(axis cs:78,0.0109321110355865)
--(axis cs:79,0.00927698811919353)
--(axis cs:80,0.00717640411636141)
--(axis cs:81,0.00717640411636141)
--(axis cs:82,0.00608940372824475)
--(axis cs:83,0.00608940372824475)
--(axis cs:84,0.00482842320996481)
--(axis cs:85,0.00468139078851307)
--(axis cs:86,0.00468139078851307)
--(axis cs:87,0.00468139078851307)
--(axis cs:88,0.00360300572265615)
--(axis cs:89,0.00316276699695871)
--(axis cs:90,0.00316276699695871)
--(axis cs:91,0.00313871936207431)
--(axis cs:92,0.00313871936207431)
--(axis cs:93,0.00313871936207431)
--(axis cs:94,0.00301294010377603)
--(axis cs:95,0.00301294010377603)
--(axis cs:96,0.00301294010377603)
--(axis cs:97,0.00301294010377603)
--(axis cs:98,0.00301294010377603)
--(axis cs:99,0.00301294010377603)
--(axis cs:100,0.00275826755878278)
--(axis cs:101,0.00262215135158785)
--(axis cs:102,0.00262215135158785)
--(axis cs:103,0.00238213270121992)
--(axis cs:104,0.00228217476400746)
--(axis cs:105,0.00228217476400746)
--(axis cs:106,0.00228217476400746)
--(axis cs:107,0.00228217476400746)
--(axis cs:108,0.00228217476400746)
--(axis cs:109,0.00228217476400746)
--(axis cs:110,0.00228217476400746)
--(axis cs:111,0.00228217476400746)
--(axis cs:111,0.00349700806573097)
--(axis cs:111,0.00349700806573097)
--(axis cs:110,0.00349700806573097)
--(axis cs:109,0.00349700806573097)
--(axis cs:108,0.00349700806573097)
--(axis cs:107,0.00349700806573097)
--(axis cs:106,0.00349700806573097)
--(axis cs:105,0.00349700806573097)
--(axis cs:104,0.00349700806573097)
--(axis cs:103,0.00360279659326998)
--(axis cs:102,0.00385461225546143)
--(axis cs:101,0.00385461225546143)
--(axis cs:100,0.00394861189978692)
--(axis cs:99,0.00416505655808577)
--(axis cs:98,0.00416505655808577)
--(axis cs:97,0.00416505655808577)
--(axis cs:96,0.00416505655808577)
--(axis cs:95,0.00416505655808577)
--(axis cs:94,0.00416505655808577)
--(axis cs:93,0.00463153372382824)
--(axis cs:92,0.00463153372382824)
--(axis cs:91,0.00463153372382824)
--(axis cs:90,0.00478962620533972)
--(axis cs:89,0.00478962620533972)
--(axis cs:88,0.00599472117029588)
--(axis cs:87,0.00903224733481805)
--(axis cs:86,0.00903224733481805)
--(axis cs:85,0.00903224733481805)
--(axis cs:84,0.0092427133891453)
--(axis cs:83,0.0205516574533165)
--(axis cs:82,0.0205516574533165)
--(axis cs:81,0.0215709480835847)
--(axis cs:80,0.0215709480835847)
--(axis cs:79,0.023758629786806)
--(axis cs:78,0.0250703677115402)
--(axis cs:77,0.0250703677115402)
--(axis cs:76,0.0250703677115402)
--(axis cs:75,0.0250703677115402)
--(axis cs:74,0.0251374758614998)
--(axis cs:73,0.0251374758614998)
--(axis cs:72,0.0371379959338535)
--(axis cs:71,0.0371379959338535)
--(axis cs:70,0.0373348335710763)
--(axis cs:69,0.0373348335710763)
--(axis cs:68,0.0373348335710763)
--(axis cs:67,0.0491890978251584)
--(axis cs:66,0.0572076371026405)
--(axis cs:65,0.0585799398073913)
--(axis cs:64,0.0585799398073913)
--(axis cs:63,0.0733460726301749)
--(axis cs:62,0.0829226688951676)
--(axis cs:61,0.0829226688951676)
--(axis cs:60,0.0831387476686272)
--(axis cs:59,0.0831387476686272)
--(axis cs:58,0.0831387476686272)
--(axis cs:57,0.0833128981532142)
--(axis cs:56,0.0833128981532142)
--(axis cs:55,0.0883295513077554)
--(axis cs:54,0.0886165784844924)
--(axis cs:53,0.0897478716706576)
--(axis cs:52,0.0897478716706576)
--(axis cs:51,0.0940173217355751)
--(axis cs:50,0.0940173217355751)
--(axis cs:49,0.0949254088270347)
--(axis cs:48,0.102882716434771)
--(axis cs:47,0.119619316571058)
--(axis cs:46,0.120090832665699)
--(axis cs:45,0.129961417338709)
--(axis cs:44,0.133362931818315)
--(axis cs:43,0.133362931818315)
--(axis cs:42,0.133362931818315)
--(axis cs:41,0.133982184999367)
--(axis cs:40,0.148535364976073)
--(axis cs:39,0.148535364976073)
--(axis cs:38,0.158334858385452)
--(axis cs:37,0.161933318565661)
--(axis cs:36,0.197906809073517)
--(axis cs:35,0.197906809073517)
--(axis cs:34,0.201281921167448)
--(axis cs:33,0.30602028665567)
--(axis cs:32,0.416490235690318)
--(axis cs:31,0.416490235690318)
--(axis cs:30,0.44008688339578)
--(axis cs:29,0.473395745476811)
--(axis cs:28,0.473395745476811)
--(axis cs:27,0.499642139563724)
--(axis cs:26,0.51619751366145)
--(axis cs:25,0.537046009731356)
--(axis cs:24,0.540426565139138)
--(axis cs:23,0.540426565139138)
--(axis cs:22,0.6217758952191)
--(axis cs:21,0.669205880871962)
--(axis cs:20,0.669205880871962)
--(axis cs:19,0.685898080691105)
--(axis cs:18,0.685898080691105)
--(axis cs:17,0.763069633742483)
--(axis cs:16,0.961945943105827)
--(axis cs:15,1.05112723223799)
--(axis cs:14,1.10719276669695)
--(axis cs:13,1.19299098365925)
--(axis cs:12,1.19821734503193)
--(axis cs:11,1.19821734503193)
--(axis cs:10,1.24125538018958)
--cycle;

\addplot [semithick, blue, dash dot]
table {%
10 1.12100231647491
11 0.960891604423523
12 0.890567123889923
13 0.805100321769714
14 0.574287593364716
15 0.511274576187134
16 0.357288777828217
17 0.346105337142944
18 0.342290580272675
19 0.295428782701492
20 0.267814159393311
21 0.225021913647652
22 0.190940544009209
23 0.173474192619324
24 0.171748071908951
25 0.165210649371147
26 0.13868522644043
27 0.135240390896797
28 0.129479929804802
29 0.129479929804802
30 0.120991609990597
31 0.112760826945305
32 0.112439848482609
33 0.112439848482609
34 0.112439848482609
35 0.106516286730766
36 0.106516286730766
37 0.106516286730766
38 0.106516286730766
39 0.0904210358858109
40 0.0904210358858109
41 0.0904210358858109
42 0.0904210358858109
43 0.0862773433327675
44 0.0862773433327675
45 0.0718113332986832
46 0.0632215142250061
47 0.0632215142250061
48 0.0357324816286564
49 0.0357324816286564
50 0.0357324816286564
51 0.0357324816286564
52 0.0357324816286564
53 0.0350666530430317
54 0.0334733501076698
55 0.0334733501076698
56 0.0334733501076698
57 0.0334733501076698
58 0.0334733501076698
59 0.0292418599128723
60 0.0292418599128723
61 0.0292418599128723
62 0.0292418599128723
63 0.0292418599128723
64 0.0292418599128723
65 0.0292418599128723
66 0.0292418599128723
67 0.0269792787730694
68 0.0269792787730694
69 0.0269792787730694
70 0.0269792787730694
71 0.0269792787730694
72 0.0269792787730694
73 0.0269792787730694
74 0.0269792787730694
75 0.0269792787730694
76 0.0269792787730694
77 0.0269792787730694
78 0.0269792787730694
79 0.0256372727453709
80 0.0256372727453709
81 0.0245724264532328
82 0.0245724264532328
83 0.0245724264532328
84 0.0245724264532328
85 0.0245724264532328
86 0.0245724264532328
87 0.0245724264532328
88 0.0245724264532328
89 0.0245724264532328
90 0.0245724264532328
91 0.0245724264532328
92 0.0245724264532328
93 0.0245724264532328
94 0.0245724264532328
95 0.0245724264532328
96 0.0245724264532328
97 0.0245724264532328
98 0.0245724264532328
99 0.0245724264532328
100 0.0245724264532328
101 0.0245724264532328
102 0.0245724264532328
103 0.0245724264532328
104 0.0245724264532328
105 0.0245724264532328
106 0.0245724264532328
107 0.0245724264532328
108 0.0245724264532328
109 0.0245724264532328
110 0.0245724264532328
111 0.0245724264532328
};
\addplot [semithick, color1, dash dot]
table {%
10 1.12100227475166
11 0.893287459190183
12 0.82237634244511
13 0.662911031900199
14 0.534703775447099
15 0.534703775447099
16 0.348683692200318
17 0.316657753984019
18 0.309428667134258
19 0.269237861109829
20 0.259237843096687
21 0.225558935400679
22 0.210310772345907
23 0.190315353663932
24 0.190171261939914
25 0.179434095092317
26 0.148081102304422
27 0.105200531732474
28 0.105200531732474
29 0.0860521138174085
30 0.0672000521748028
31 0.0672000521748028
32 0.0588229331752026
33 0.0588229331752026
34 0.0588229331752026
35 0.0588229331752026
36 0.0588229331752026
37 0.0483978074958842
38 0.0483978074958842
39 0.0483978074958842
40 0.0483978074958842
41 0.0430632581259615
42 0.0407787361099431
43 0.0407787361099431
44 0.0407787361099431
45 0.0407787361099431
46 0.0407787361099431
47 0.0407787361099431
48 0.0407787361099431
49 0.0407787361099431
50 0.0407787361099431
51 0.0407787361099431
52 0.0407787361099431
53 0.0389082104113833
54 0.0389082104113833
55 0.0389082104113833
56 0.0389082104113833
57 0.0310426408677452
58 0.0310426408677452
59 0.0310426408677452
60 0.0310426408677452
61 0.0310426408677452
62 0.0310426408677452
63 0.0310426408677452
64 0.0310426408677452
65 0.0310426408677452
66 0.0310426408677452
67 0.0297723461561598
68 0.0297723461561598
69 0.0297723461561598
70 0.0297723461561598
71 0.0297723461561598
72 0.0297723461561598
73 0.0297723461561598
74 0.0285775443240992
75 0.0285775443240992
76 0.0285775443240992
77 0.0285775443240992
78 0.0256313642255473
79 0.0256313642255473
80 0.0256313642255473
81 0.0240950160510414
82 0.0240950160510414
83 0.0240950160510414
84 0.0240950160510414
85 0.0235075522677734
86 0.0235075522677734
87 0.022223881430527
88 0.022223881430527
89 0.022223881430527
90 0.022223881430527
91 0.022223881430527
92 0.021045806010789
93 0.021045806010789
94 0.021045806010789
95 0.021045806010789
96 0.0200234207750509
97 0.0200234207750509
98 0.0200234207750509
99 0.0200234207750509
100 0.0200234207750509
101 0.0200234207750509
102 0.0200234207750509
103 0.0200234207750509
104 0.0200234207750509
105 0.0200234207750509
106 0.0200234207750509
107 0.0200234207750509
108 0.0200234207750509
109 0.0200234207750509
110 0.0200234207750509
111 0.0200234207750509
};
\addplot [semithick, blue]
table {%
10 1.12100231647491
11 1.11889326572418
12 0.989772617816925
13 0.872065544128418
14 0.788813710212708
15 0.777261435985565
16 0.760331988334656
17 0.694352865219116
18 0.607167363166809
19 0.522954165935516
20 0.522954165935516
21 0.468895196914673
22 0.455428183078766
23 0.455428183078766
24 0.452384799718857
25 0.452384799718857
26 0.373560845851898
27 0.325073659420013
28 0.301324516534805
29 0.296129554510117
30 0.296129554510117
31 0.261142432689667
32 0.239091843366623
33 0.237726211547852
34 0.237726211547852
35 0.227967500686646
36 0.188076823949814
37 0.173662096261978
38 0.161800101399422
39 0.134521692991257
40 0.134521692991257
41 0.134521692991257
42 0.121863029897213
43 0.110853277146816
44 0.105837598443031
45 0.105837598443031
46 0.105837598443031
47 0.0947243422269821
48 0.0947243422269821
49 0.0947243422269821
50 0.0947243422269821
51 0.0947243422269821
52 0.0813777223229408
53 0.0813777223229408
54 0.0671646967530251
55 0.0671646967530251
56 0.0671646967530251
57 0.0671646967530251
58 0.0632568746805191
59 0.0632568746805191
60 0.0632568746805191
61 0.0632568746805191
62 0.0593903884291649
63 0.0593903884291649
64 0.0593903884291649
65 0.0593903884291649
66 0.0593903884291649
67 0.0593903884291649
68 0.0593903884291649
69 0.0593903884291649
70 0.0593903884291649
71 0.0593903884291649
72 0.0593903884291649
73 0.0593903884291649
74 0.0593903884291649
75 0.0593903884291649
76 0.0593903884291649
77 0.0593903884291649
78 0.0577438473701477
79 0.0577438473701477
80 0.0577438473701477
81 0.0577438473701477
82 0.0577438473701477
83 0.0540508814156055
84 0.0540508814156055
85 0.0540508814156055
86 0.0540508814156055
87 0.0540508814156055
88 0.0540508814156055
89 0.0540508814156055
90 0.0540508814156055
91 0.0540508814156055
92 0.0540508814156055
93 0.0540508814156055
94 0.0540508814156055
95 0.0540508814156055
96 0.0463078804314137
97 0.0463078804314137
98 0.0463078804314137
99 0.0463078804314137
100 0.0463074408471584
101 0.0463074408471584
102 0.0463074408471584
103 0.0463074408471584
104 0.0463074408471584
105 0.0463074408471584
106 0.0463074408471584
107 0.0463074408471584
108 0.0463074408471584
109 0.0463074408471584
110 0.0463074408471584
111 0.0463074408471584
};
\addplot [semithick, color1]
table {%
10 1.12100227475166
11 1.08272851336586
12 1.08272851336586
13 1.06556343370024
14 0.961051769783383
15 0.918397098496712
16 0.828973432915107
17 0.638885594096975
18 0.564746150493506
19 0.564746150493506
20 0.550414508909926
21 0.550414508909926
22 0.496328008429119
23 0.441034935573606
24 0.441034935573606
25 0.438176292284766
26 0.41337478649205
27 0.395526289808938
28 0.371794844248618
29 0.371794844248618
30 0.343867506602371
31 0.322290873397089
32 0.322290873397089
33 0.259357917714938
34 0.165829965942718
35 0.162907023858025
36 0.162907023858025
37 0.137647348650564
38 0.133072363256451
39 0.122389030183116
40 0.122389030183116
41 0.112102831074639
42 0.11088002918398
43 0.11088002918398
44 0.11088002918398
45 0.105829471806113
46 0.0956019795091687
47 0.0949084838735895
48 0.0825312440329184
49 0.0760550922990611
50 0.0746421733128068
51 0.0746421733128068
52 0.0707117474045513
53 0.0707117474045513
54 0.0690262740174848
55 0.0685759281011102
56 0.0647132221169456
57 0.0647132221169456
58 0.0644624982032779
59 0.0644624982032779
60 0.0644624982032779
61 0.0641213182716242
62 0.0641213182716242
63 0.0546718256777347
64 0.0453509511947461
65 0.0453509511947461
66 0.0438964971800688
67 0.0366828874858743
68 0.0265618214050806
69 0.0265618214050806
70 0.0265618214050806
71 0.0263110187253696
72 0.0263110187253696
73 0.0180690622266338
74 0.0180690622266338
75 0.0180012393735633
76 0.0180012393735633
77 0.0180012393735633
78 0.0180012393735633
79 0.0165178089529998
80 0.014373676099973
81 0.014373676099973
82 0.0133205305907806
83 0.0133205305907806
84 0.00703556829955506
85 0.00685681906166556
86 0.00685681906166556
87 0.00685681906166556
88 0.00479886344647602
89 0.00397619660114922
90 0.00397619660114922
91 0.00388512654295128
92 0.00388512654295128
93 0.00388512654295128
94 0.0035889983309309
95 0.0035889983309309
96 0.0035889983309309
97 0.0035889983309309
98 0.0035889983309309
99 0.0035889983309309
100 0.00335343972928485
101 0.00323838180352464
102 0.00323838180352464
103 0.00299246464724495
104 0.00288959141486921
105 0.00288959141486921
106 0.00288959141486921
107 0.00288959141486921
108 0.00288959141486921
109 0.00288959141486921
110 0.00288959141486921
111 0.00288959141486921
};
\end{axis}

\end{tikzpicture}

%% file: appendix.tex
\newpage
\addtocontents{toc}{\protect\setcounter{tocdepth}{2}}

\onecolumn

\setcounter{figure}{0}
\setcounter{table}{0}
\setcounter{equation}{0}

\renewcommand{\thefigure}{S\arabic{figure}}
\renewcommand{\thetable}{S\arabic{table}}
\renewcommand{\theequation}{S\arabic{equation}}

\aistatstitle{Supplementary Material}
\label{app:intro}

\vspace*{-13cm} 

\tableofcontents

\newpage

\section{TABLE OF ACRONYMS}
\label{app:notation}

For ease of reference, \cref{tab:acronyms} reports a list of key acronyms and abbreviations used in the paper.

\begin{table}[htbp]
\centering
\begin{tabular}{llp{6cm}}
\toprule
\textbf{Acronym} & \textbf{Full Name} & \textbf{Description} \\ 
\midrule
\midrule
\multicolumn{3}{l}{\textbf{Architectures}} \\ \midrule
TPM-D & Transformer Prediction Map – Diagonal & Family of transformer architectures for diagonal prediction maps, including all architectures below \\
\midrule
ACE & Amortized Conditioning Engine & Our transformer-based meta-learning model for probabilistic tasks with explicit latent variables \\
ACEP & Amortized Conditioning Engine (with Priors) & ACE variant allowing runtime injection of priors over latent variables  \\
CNP & Conditional Neural Process & Context-to-target mapping with permutation invariance \\
PFN & Prior-Fitted Network & Meta-learning approach using transformers for inference and introducing Riemannian output distributions \\
TNP-D & Transformer Neural Process – Diagonal & Transformer neural process variant with independent target predictions \\
\midrule
\multicolumn{3}{l}{\textbf{Bayesian Optimization (BO) Terms}} \\
\midrule
BO & Bayesian Optimization & Black-box function optimization using surrogate models \\
MES & Max-Value Entropy Search & Acquisition function based on uncertainty over optimum value \\
TS & Thompson Sampling & Optimization via sampling from the posterior over optimum location \\
\(\pi\)BO & Prior-information BO & BO incorporating prior knowledge on optimum location \\
AR-TNP-D-TS & Autoregressive TNP-D Thompson Sampling & TNP extension with autoregressive sampling for BO \\
\midrule
\multicolumn{3}{l}{\textbf{Simulation-Based Inference (SBI) Terms}} \\ 
\midrule
SBI & Simulation-Based Inference & Parameter posterior inference using synthetic data \\
NPE & Neural Posterior Estimation & Direct posterior modeling with neural networks \\
NRE & Neural Ratio Estimation & Likelihood-ratio-based posterior inference \\
OUP & Ornstein–Uhlenbeck Process & Mean-reverting stochastic process \\
SIR & Susceptible–Infectious–Recovered & Epidemiological disease spread model \\ 
\bottomrule
\end{tabular}
\caption{Key acronyms used in the paper, grouped by category.}
\label{tab:acronyms}
\end{table}

\section{METHODS}
\label{app:methods}

This section details several technical aspects of our paper, such as the prior amortization techniques, neural network architecture and general training and inference details.

\subsection{Details and experiments with prior injection}
\label{app:priors}

\paragraph{Prior generative process.} \label{app:prior}

To expose ACE to a wide array of distinct priors during training, we generate priors following a hierarchical approach that generates smooth priors over a bounded range. The process is as follows, separately for each latent variable $\theta_l$, for $1 \le l \le L$:
\begin{itemize}
    \item We first sample the type of priors for the latent variable. Specifically, with 80\% probability, we sample from a mixture of Gaussians to generate a smooth prior, otherwise, we create a flat prior with uniform distribution.
    \item If we sample from a mixture of Gaussians:
    \begin{itemize}
        \item We first sample the number of Gaussian components $K$ from a geometric distribution with $q=0.5$:
        \begin{equation}
            K \sim \text{Geometric}(0.5).
        \end{equation}
        \item If $K > 1$, we randomly choose among three configurations with equal probability: 
        \begin{enumerate}
            \item Same means and different standard deviations.
            \item Different means and same standard deviations.
            \item All different means and standard deviations.
        \end{enumerate}
        \item Given the predefined global priors for mean and standard deviation (uniform distributions whose ranges are determined by the range of the corresponding latent variable), we sample the means and standard deviations for each component from the predefined uniform distributions.
        \item The weights of the mixture components are sampled from a Dirichlet distribution:
        \begin{equation}
            \mathbf{w} \sim \text{Dirichlet}(\alpha_0 = 1).
        \end{equation}
        \item Finally, we convert the mixture of Gaussians into a normalized histogram over a grid $\mathcal{G}$ with $N_\text{bins}$ uniformly-spaced bins. For each bin $b$, we compute the probability mass $\mathbf{p}^{(b)}_l$ by calculating the difference between the cumulative distribution function values at the bin edges. This is done for each Gaussian component and then summed up, weighted by the mixture weights.
        \item We normalize the bin probabilities to ensure a valid probability distribution:
        \begin{equation}
            \mathbf{p}_l = \frac{\mathbf{p}_l}{\sum_{b=1}^{N_\text{bins}} \mathbf{p}^{(b)}_l}.
        \end{equation}
    \end{itemize}
    
    \item If we sample from a uniform distribution:
        \begin{itemize}
            \item We assign equal probability to each bin over the grid:
            \begin{equation}
                \mathbf{p}_l = \frac{1}{N_\text{bins}} \mathbf{1}_{N_\text{bins}},
            \end{equation}
            where $\mathbf{1}_{N_\text{bins}}$ is a vector of ones of length $N_\text{bins}$.
        \end{itemize}
    
\end{itemize}

\begin{figure}[h] 
    \centering
    \includegraphics[width=0.80\linewidth]{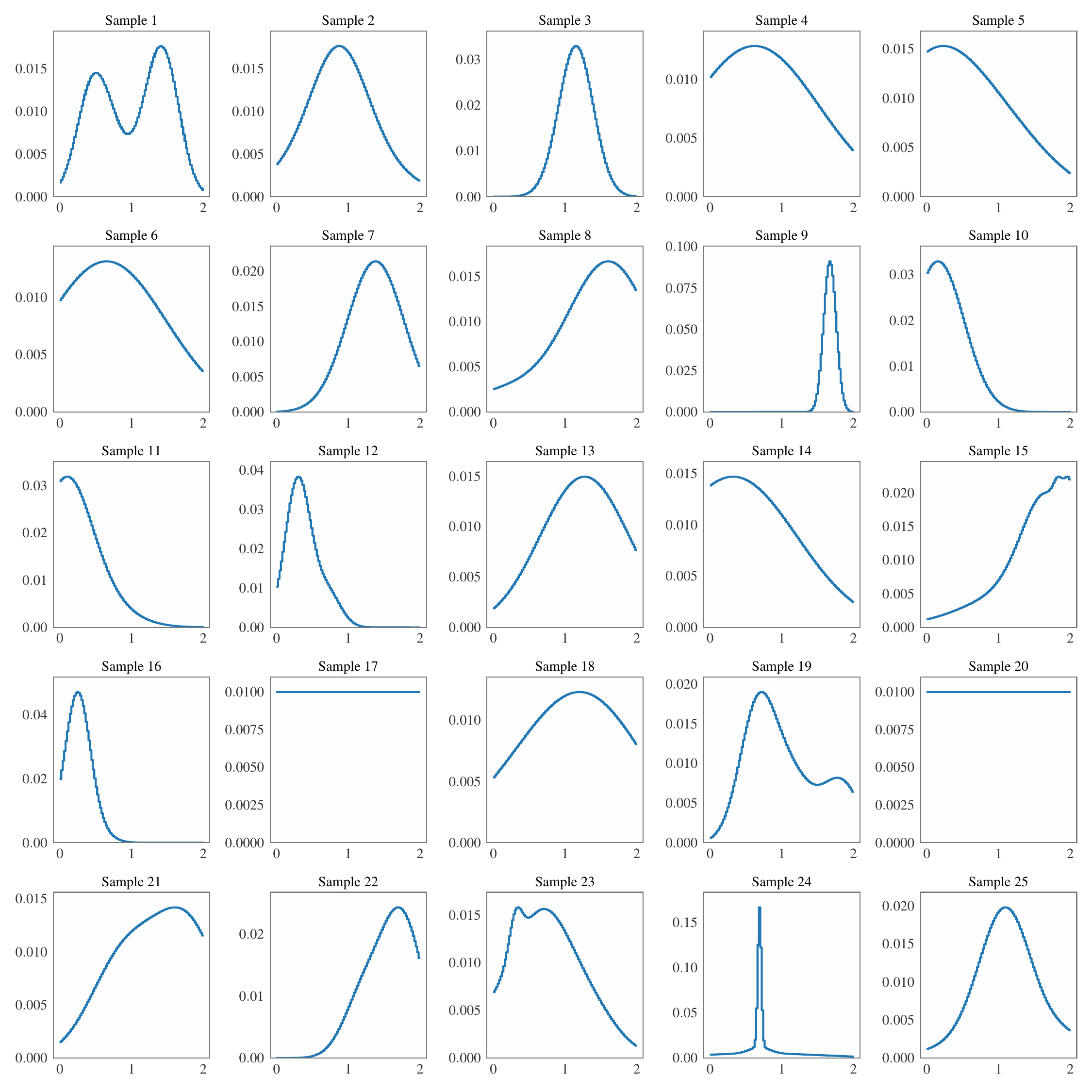}
    \caption{Examples of randomly sampled priors over the range $[0, 2]$. Samples include mixtures of Gaussians and Uniform distributions.} 
    \label{fig:prior_injection_samples}
\end{figure}

For all experiments, we select $N_\text{bins} = 100$ as the number of bins for the prior grid. See \cref{fig:prior_injection_samples} for some examples of sampled priors.

\paragraph{Investigation of prior injection with a Gaussian toy model.} \label{app:gaussian}
To investigate the effect of the injected prior, we test our method with a simple 1D Gaussian model with two latent variables: mean $\mu$ and standard deviation $\sigma$. The data is the samples drawn from this distribution, $\data_N = \{y_n\}_{n=1}^N \sim \mathcal{N}\left(\mu, \sigma^2\right)$. We can numerically compute the exact Bayesian posterior on the predefined grid given the data and any prior, and subsequently compare the ground-truth posterior with the ACE's predicted posterior after injecting the same prior.

\begin{figure}[h] 
    \centering
    \includegraphics[width=0.60\linewidth]{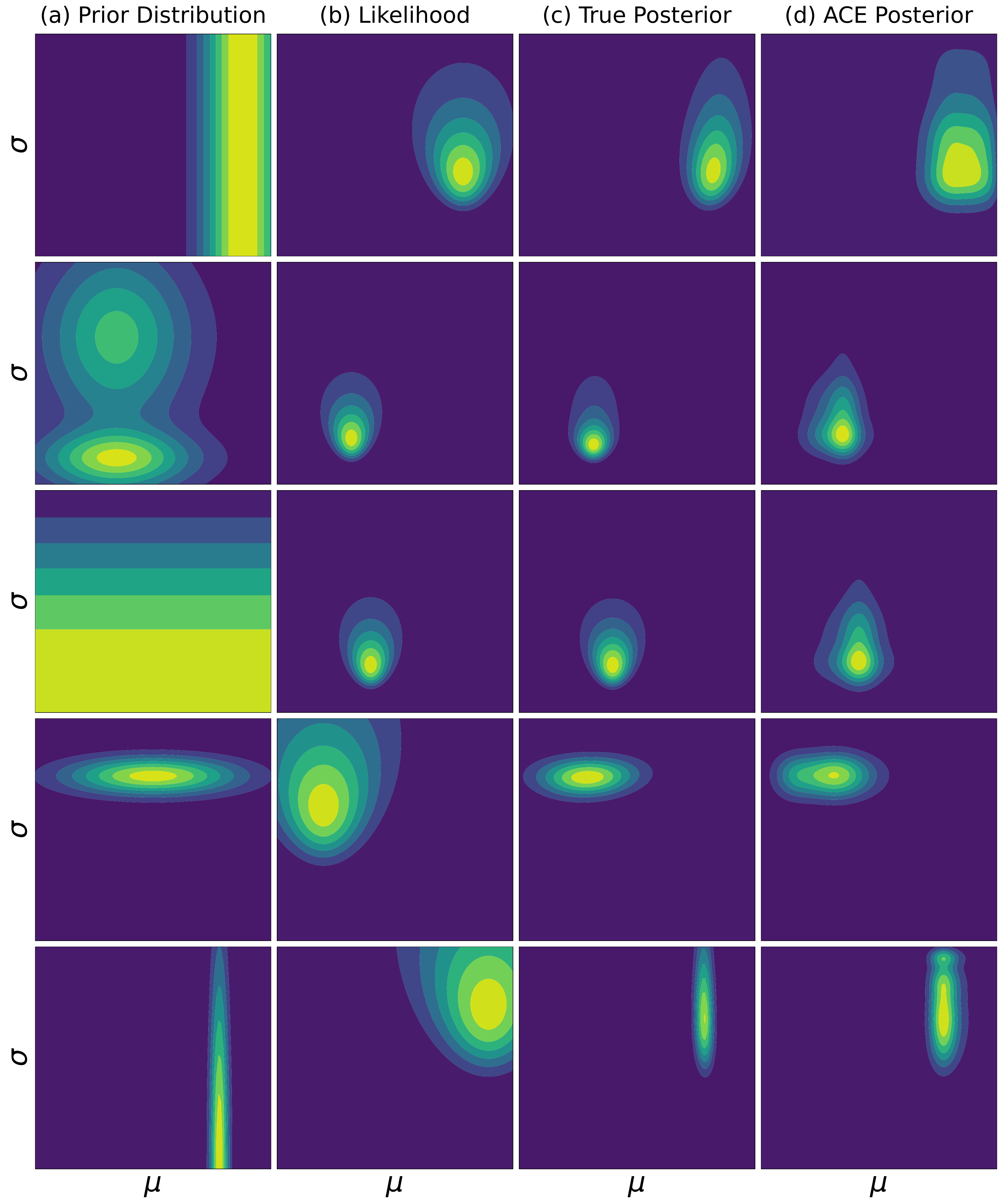}
    \caption{Examples of the true and predicted posterior distributions in the toy 1D Gaussian case. (a) Prior distribution over $\vtheta = (\mu, \sigma)$ set at runtime. (b) Likelihood for the observed data (the data themselves are not shown). (c) Ground-truth Bayesian posterior. (d) ACE's predicted posterior, based on the user-set prior and observed data, approximates well the true posterior.}

    \label{fig:gaussian_examples}
\end{figure}

We first sample random priors using the generative process described above. Then we sample $\mu$ and $\sigma$ from the priors and generate the corresponding data $\data_N$. We pass the data along with the prior to ACE to get the predictive distributions $p(\mu|\data_N)$ and $p(\sigma | \data_N)$ as well as the autoregressive predictions $p(\mu|\sigma, \data_N)$ and $p(\sigma|\mu, \data_N)$. 
With these equations, we can autoregressively compute our model's prediction for $p(\mu, \sigma|\data_N)$ on the grid.
The true posterior is calculated numerically via Bayes rule on the grid. 
See \cref{fig:gaussian_examples} for several examples.

\begin{figure}[h] 
    \centering
    \includegraphics[width=0.60\linewidth]{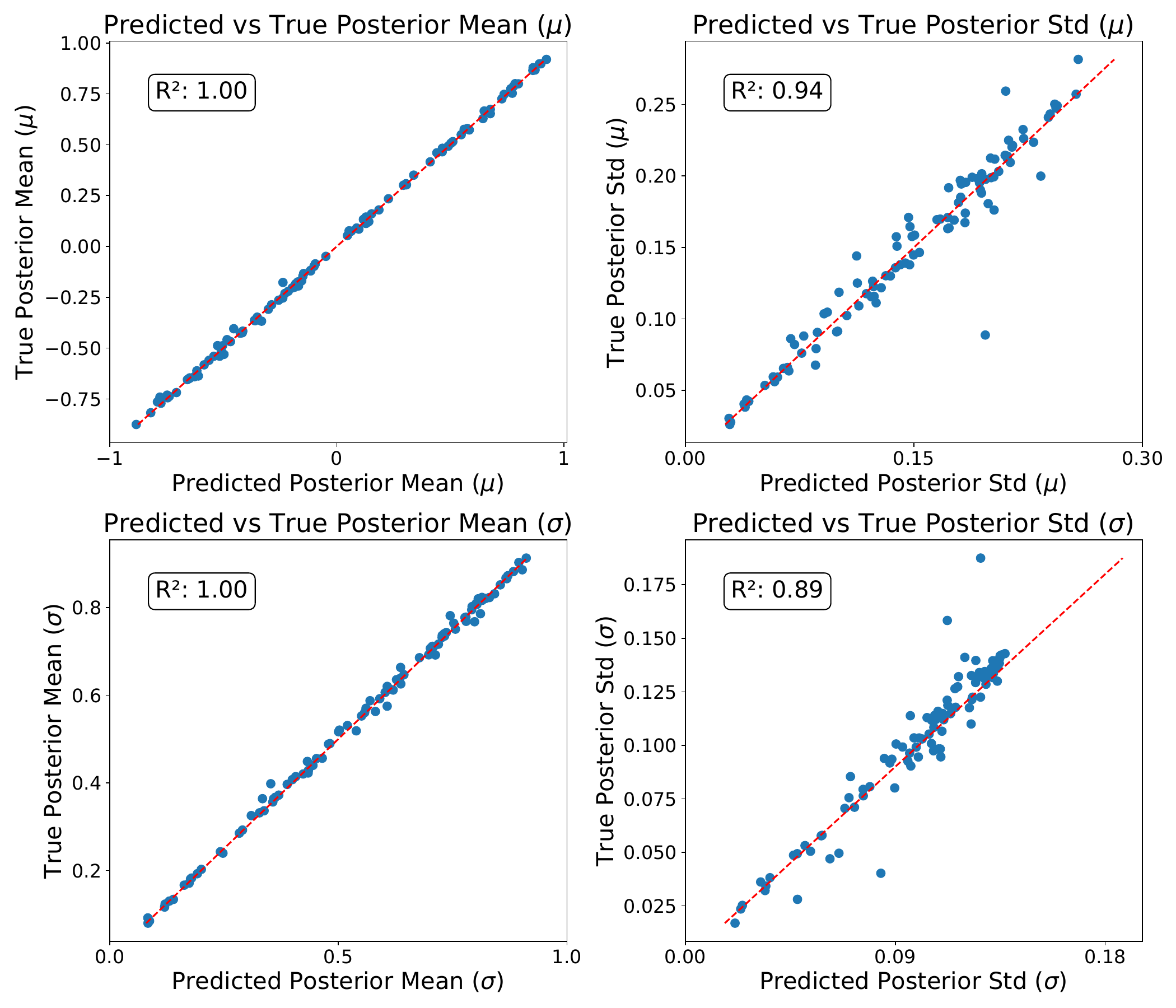}
    \caption{The scatter plots compare the predicted and true posterior mean and standard deviation values for both $\mu$ and $\sigma$ across 100 examples. We can see that the points lie closely along the diagonal red dashed line, indicating that the moments (mean and standard deviation) of the predicted posteriors closely match the true posteriors.} 
    \label{fig:gaussian_scatter}
\end{figure}

To quantitatively assess the quality of our model's predicted posteriors, we compare the posterior mean and standard deviation (i.e., the first two moments\footnote{We prefer standard deviation to variance as it has the same units as the quantity of interest, as opposed to squared units which are less interpretable.}) for $\mu$ and $\sigma$ of predicted vs. true posteriors, visualized in \cref{fig:gaussian_scatter}.
The scatter points are aligned along the diagonal line, indicating that the moments of the predicted posterior closely match the moments of true posterior. These results show that ACE is effectively incorporating the information provided by the prior and adjusts the final posterior accordingly.
In \cref{app:sbc} we perform a more extensive analysis of posterior calibration in ACE with a complex simulator model.

\clearpage

\subsection{Architecture}
\label{app:architecture}

Here we give an overview of two key architectures used in our paper.
First, we show the TNP-D~\citep{nguyen2022transformer}
architecture in \cref{fig:tnpd_architecture}, which our method extends. 
\cref{fig:ace_architecture} shows the ACE architecture introduced in this paper.

\input{architecture/architecture_tnpd}
\input{architecture/architecture_ace}

\subsection{Training batch construction}

\label{app:sampling}

ACE can condition on and predict data, latent variables, and combinations of both. Here, we outline the sampling process used to construct the training batch.

\begin{itemize}
    \item First, we generate our dataset by following the steps outlined for the respective cases (GP, Image Completion, BO, SBI); see \cref{app:experiments}. For example, in the GP emulation case, we draw $n_{\text{data}}$ points from a function sampled from a GP along with its respective latent variables.
    \item Next, we sample the number of context points, $n_{\text{context}}$, uniformly between the minimum and maximum context points, $\text{min\_ctx}$ and $\text{max\_ctx}$. We then split our data based on this $n_{\text{context}}$ value; the remaining data points that are not in the context set are allocated to the target set.
    \item We then determine whether the context includes any latent variables at all with a 50\% probability. If latent variables are to be included in the context set, we sample the number of latents residing in the context set, uniformly from 1 to $n_{\text{latents}}$. All latent variables not in the context set are assigned to the target set.
\end{itemize}

The above steps are applied for each element (dataset) in the training batch. In the implementation, we also ensure that, within each batch, the number of context points remains consistent across all elements, as does the number of target points, to facilitate batch training for our model. However, the number of latents in the context set may vary for each element, introducing variability that improves the model's training process.

\subsection{Autoregressive predictions}
\label{app:autoregressive}

While ACE predicts conditional marginals independently, we can still obtain \emph{joint} predictions over both data and latents autoregressively \citep{nguyen2022transformer,bruinsma2023autoregressive}.
Suppose we want to predict the joint target distribution $p(\z^\star_{1:M}| \vxi^\star_{1:M}, \dataplus_N)$, where we use compact indexing notation. We can write:
\begin{equation}
p(\z^\star_{1:M}| \vxi^\star_{1:M}, \dataplus_N) = 
\prod_{m=1}^M p(z^\star_m | \z^\star_{1:m-1}, \vxi^\star_{1:m},  \dataplus_N) =  
\mathbb{E}_{\vpi} \left[ \prod_{m=1}^M p(z^\star_{\pi_m} | \z^\star_{\pi_1:\pi_{m-1}}, \vxi^\star_{\pi_1:\pi_m}, \dataplus_N) \right],
\end{equation}
where $\vpi$ is a permutation of $(1, \ldots, M)$, i.e. an element of the symmetric group $\mathcal{S}_M$.
The first passage follows from the standard rules of probability and the second passage follows from permutation invariance of the joint distribution with respect to the ordering of the variables $\vxi_{1:M}$. The last expression can be used to enforce permutation invariance and validity of our joint predictions even if sequential predictions of the model are not natively invariant \citep{murphy2019janossy}. In practice, for moderate to large $M$ ($M \gtrsim 4$) we approximate the expectation over permutations via Monte Carlo sampling.

\section{EXPERIMENTAL DETAILS}
\label{app:experiments}
In this section, we show additional experiments to validate our method and provide additional details about sampling, training, and model architecture.

\subsection{Gaussian process (GP) experiments}
\label{app:gp}

\begin{figure}[!h]
  \centering
  \scriptsize
  \setlength{\figurewidth}{.22\columnwidth}
  \setlength{\figureheight}{.9\figurewidth}
  \begin{subfigure}[b]{.31\textwidth}
    \centering
    \input{figures/condition.tex}
    \caption{}
    \label{fig:condition}
  \end{subfigure}
  \begin{subfigure}[b]{.31\textwidth}
    \centering
    \input{figures/accuracy.tex}
    \caption{}
    \label{fig:accuracy}
  \end{subfigure}
  \begin{subfigure}[b]{.31\textwidth}
    \centering
    \input{figures/latent_pred.tex}
    \caption{}
    \label{fig:latent}
  \end{subfigure}
  \vspace{-0.2cm}
  \caption{(a) Conditioning on the latent variable $\vtheta$ (kernel hyperparameters and type) improves predictive performance, approaching the GP upper bound for the log predictive density. (b) ACE can identify the kernel $\kappa$. (c) ACE can learn kernel hyperparameters.}
  \label{fig:GP}
\end{figure}

We now demonstrate the use of ACE for performing amortized inference tasks in the Gaussian processes (GP) model class. GPs are a Bayesian non-parametric method used as priors over functions (see \citealp{rasmussen2006gaussian}). To perform inference with a GP, one must first define a kernel function $\kappa_\vtheta$ parameterized by hyperparameters $\vtheta$ such as lengthscales and output scale. 
As a flexible model of distributions over functions used for regression and classification, GPs are a go-to generative model for meta-learning and feature heavily in the (conditional) neural process literature (CNP; \citealp{garnelo2018conditional}). ACE can handle many of the challenges faced when applying GPs. Firstly, it can accurately amortize the GP predictive distribution as is usually shown in the CNP literature. In addition, ACE can perform other crucial tasks in the GP workflow, such as amortized learning of $\vtheta$, usually found through optimization of the marginal likelihood (Gaussian likelihood) or via approximate inference for non-Gaussian likelihoods (e.g., \citealt{hensman2015scalable}). Furthermore, we can also do kernel selection by treating the kernel as a latent discrete variable, and incorporate prior knowledge about $\vtheta$. 

\textbf{Results.} The main results from our GP regression experiments are displayed in \cref{fig:GP}. We trained ACE on samples from four kernels, the RBF and Matérn-($\nicefrac{1}{2}$, $\nicefrac{3}{2}$, $\nicefrac{5}{2}$), using the architecture described in \cref{sec:ace}; see below for details. In \cref{fig:condition}, we show ACE's ability to condition on provided information: data only, or data and $\vtheta$ (kernel hyperparameters and type). As expected, there is an improvement when conditioning on more information, specifically when the context set $\data_N$ is small. As an upper bound, we show the ground-truth GP predictive performance. The method can accurately predict the kernel, \ie model selection (\cref{fig:accuracy}), while at the same time learn the hyperparameters (\cref{fig:latent}), both improving as a function of the context set size. 

\paragraph{Sampling from a GP.} Both the GP experiments and the Bayesian optimization experiments reported in \cref{exp:bo} and further detailed in \cref{app:bo} use a similar sampling process to generate data. 

\begin{itemize}
\item We first sample the latents. These are kernel hyperparameters, the output scale $\sigma_f$ and lengthscale $\mathbf{\ell}$. Each input dimension of $\x$ is assigned its own lengthscale $\mathbf{\ell} = (\ell^{(1)}, \ell^{(2)},...)$ and a corresponding kernel. For all GP examples the RBF and three Matérn-($\nicefrac{1}{2}$, $\nicefrac{3}{2}$, $\nicefrac{5}{2}$) kernels were used with equal weights. The kernel output scale $\sigma_f\sim U(0.1, 1)$ and each $k$-th lengthscale is $\mathbf{\ell}^{(k)} \sim \mathcal{N}(1/3, 0.75)$ truncated between $[0.05, 2]$.
\item Once all latent information is defined, we draw from a GP prior from a range $[-1, 1]$ for each input dimension. The realizations from the prior form our context data $\mathcal{D}_{N}$ where the size of the context set $N$ is drawn from a discrete uniform distribution ${3,4,....50}$. The target data $(\mathbf{X}^*,\mathbf{y}^*)$ of size $200 - N$  is then drawn from the predictive posterior of the GP conditioned on $\mathcal{D}_{N}$.
\end{itemize}

\paragraph{Architecture.} The ACE model used in the GP experiments had embedding dimension $256$ and $6$ transformer layers.
The attention blocks had $6$ heads and the MLP block had hidden dimension $128$. The output head had $K=20$ MLP components with hidden dimension $256$. The model was trained for $5\times 10^4$ steps with batch size $32$, using learning rate $1 \times 10^{-4}$ with cosine annealing. Following~\citet{nguyen2022transformer}, and unlike the original transformer implementation~\citep{vaswani2017attention}, we do not use dropout in any of our experiments.

\input{image_mnist}

\subsection{Image completion and classification}
\begin{figure}[t]
\label{fig:64x64img}
    \centering
    \begin{subfigure}[b]{0.18\textwidth}
        \centering
        \includegraphics[width=\linewidth]{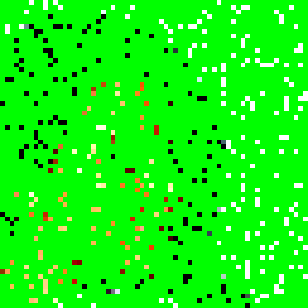}
        \label{fig:image1}
    \end{subfigure}
    \hspace{0.5em} 
    \begin{subfigure}[b]{0.18\textwidth}
        \centering
        \includegraphics[width=\linewidth]{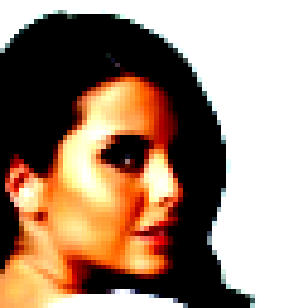}
        \label{fig:image2}
    \end{subfigure}
    \hspace{0.5em}
    \begin{subfigure}[b]{0.18\textwidth}
        \centering
        \includegraphics[width=\linewidth]{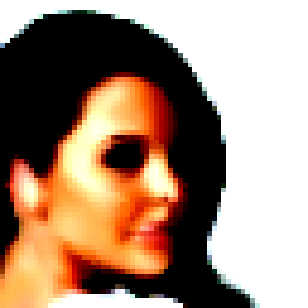}
        \label{fig:image3}
    \end{subfigure}
    \hspace{0.5em}
    \begin{subfigure}[b]{0.18\textwidth}
        \centering
        \includegraphics[width=\linewidth]{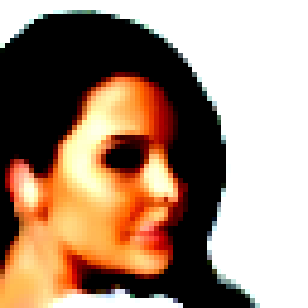}
        \label{fig:image4}
    \end{subfigure}

    \vspace{1em} 

    \begin{subfigure}[b]{0.18\textwidth}
        \centering
        \includegraphics[width=\linewidth]{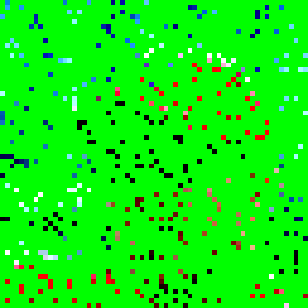}
        \caption{Context}
        \label{fig:image5}
    \end{subfigure}
    \hspace{0.5em}
    \begin{subfigure}[b]{0.18\textwidth}
        \centering
        \includegraphics[width=\linewidth]{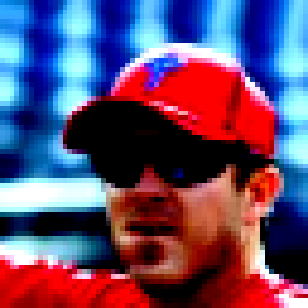}
        \caption{Image}
        \label{fig:image6}
    \end{subfigure}
    \hspace{0.5em}
    \begin{subfigure}[b]{0.18\textwidth}
        \centering
        \includegraphics[width=\linewidth]{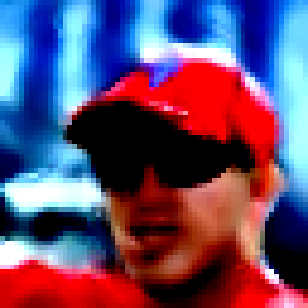}
        \caption{ACE}
        \label{fig:image7}
    \end{subfigure}
    \hspace{0.5em}
    \begin{subfigure}[b]{0.18\textwidth}
        \centering
        \includegraphics[width=\linewidth]{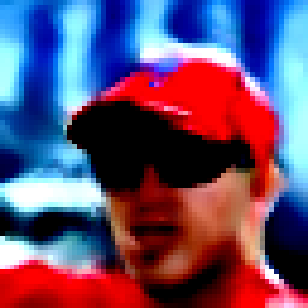}
        \caption{ACE-$\vtheta$}
        \label{fig:image8}
    \end{subfigure}

    \caption{Examples of ACE on 64x64 image size.}
    \label{fig:64_rows_images}
\end{figure}

In this section, we detail the image experiments in \cref{exp:image} as well as report additional experiments. Image completion experiments have long been a benchmark in the neural process literature treating them as regression problems \citep{garnelo2018neural, kim2019attentive}. We use two standard datasets, \textsc{MNIST} \citep{deng2012mnist} and \textsc{CelebA} \citep{liu2015faceattributes}. The \textsc{MNIST} results presented are with the full image size $28 \times 28$, while \textsc{CelebA} results were downsized to $32 \times 32$. However, as shown in \cref{fig:64_rows_images}, ACE can also handle the full image size. All image datasets were normalised based on the complete dataset average and standard deviation. The data input $\mathbf{x}$ for image experiments is the $2$\text{-}D image pixel-coordinates and the data value $y$ for \textsc{MNIST} is one output dimension, while \textsc{CelebA} uses all three RGB channels and thus is a multi-output $\mathbf{y}$.

The experiments on images demonstrate the versatility of the ACE method and its advantages over conventional CNPs. We outperform the current state-of-the-art TNP-D on the standard image completion task (\cref{fig:whole_image_celeb_main}). Given a random sample from the image space as context $\data_N$, the model predicts the remaining $M$ image pixel values at $\xt$. The total number of points $N + M$ for \textsc{MNIST} is thus $784$ points and $1024$ for \textsc{CelebA} where the split is randomly sampled (see below for details). The model is then trained as detailed in \cref{sec:training}. Thus, the final trained model can perform image completion, also sometimes known as in-painting. 

\begin{wrapfigure}{r}{0.5\textwidth}  
\vspace{-0.2cm}
    \centering
    \setlength{\figurewidth}{1\linewidth}
    \setlength{\figureheight}{0.6\linewidth}
    \input{figures/mnist_acc.tex}
    \caption{Classification accuracy for \textsc{MNIST} for varying context size.}
    \label{fig:mnist-accuracy}
    \vspace{-0.8cm}
\end{wrapfigure}

In addition to image completion, our method can condition on and predict latent variables $\vtheta$. For \textsc{MNIST}, we use the class labels as latents, so \textsc{0, 1, 2, ...}, which were encoded into a single discrete variable. Meanwhile, for \textsc{CelebA} we use as latents the $40$ binary features that accompany the dataset, \eg \textsc{BrownHair, Man, Smiling}, trained with the sampling procedure discussed below. We recall that in ACE the latents $\vtheta$ can be both conditioned on and predicted. Thus, we can do conditional generation based on the class or features or, given a sample of an image, predict its class or features, as initially promised in \cref{fig:intro}a.

\begin{figure}[b!]
    \centering
    \setlength{\figurewidth}{0.15\linewidth}
    \setlength{\figureheight}{0.15\linewidth}
    
    \begin{subfigure}{\figurewidth}
        \centering
        \includegraphics[width=\linewidth]{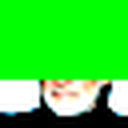}
        \caption{Context}
        \label{fig:mask_context}
    \end{subfigure}%
    \hfill
    \begin{subfigure}{\figurewidth}
        \centering
        \includegraphics[width=\linewidth]{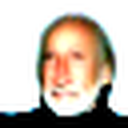}
        \caption{\textsc{Bald} = \textsc{True}}
        \label{fig:bald_pred}
    \end{subfigure}%
    \hfill
    \begin{subfigure}{\figurewidth}
        \centering
        \includegraphics[width=\linewidth]{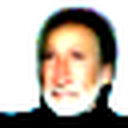}
        \caption{\textsc{Bald} = \textsc{False}}
        \label{fig:no_bald_pred}
    \end{subfigure}%
    \hfill
    \begin{subfigure}{\figurewidth}
        \centering
        \includegraphics[width=\linewidth]{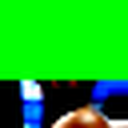}
        \caption{Context}
        \label{fig:mask_context_}
    \end{subfigure}%
    \hfill
    \begin{subfigure}{\figurewidth}
        \centering
        \includegraphics[width=\linewidth]{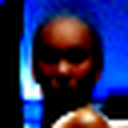}
        \caption{\textsc{Bald} = \textsc{True}}
        \label{fig:bald_pred_}
    \end{subfigure}%
    \hfill
    \begin{subfigure}{\figurewidth}
        \centering
        \includegraphics[width=\linewidth]{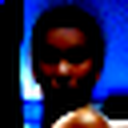}
        \caption{\textsc{Bald} = \textsc{False}}
        \label{fig:no_bald_pred_}
    \end{subfigure}
    
    \caption{Example of ACE conditioning on the value of the \textsc{Bald} feature when the image is masked for the first 22 rows. (\subref{fig:mask_context}) and (\subref{fig:mask_context_}) show the context points used for prediction, where (\subref{fig:bald_pred}) and (\subref{fig:bald_pred_}) show predictions where the \textsc{Bald} feature is conditioned on \textsc{True}. Meanwhile, \subref{fig:no_bald_pred} and \subref{fig:no_bald_pred_} are conditioned on \textsc{False}.}
    \label{fig:masked_celeba_row}
\end{figure}

\begin{figure}[h]
    \centering
    \setlength{\figurewidth}{.33\linewidth}
    \setlength{\figureheight}{.9\figurewidth}
    \begin{subfigure}{.2\linewidth}
        \centering
        \includegraphics[width=\linewidth]{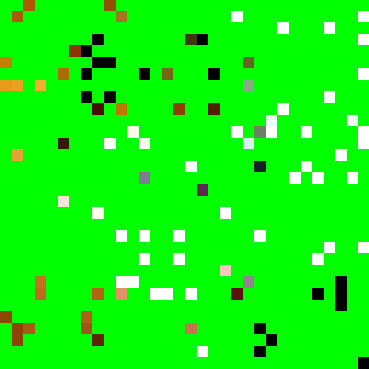}
        \label{fig:plot1}
    \end{subfigure}
    \begin{subfigure}{.2\linewidth}
        \centering
        \includegraphics[width=\linewidth]{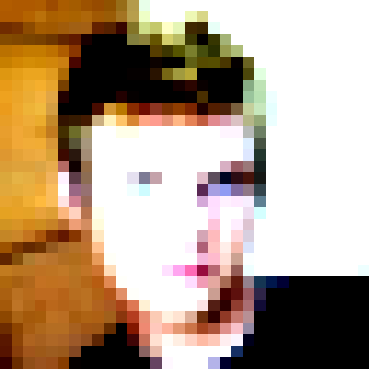}
        \label{fig:plot2}
    \end{subfigure}
    \begin{subfigure}{.45\linewidth}
        \centering
        \scriptsize
        \input{image_files/output_plot_1.tex}
        \label{fig:plot3}
    \end{subfigure}
    
    \setlength{\figurewidth}{.33\linewidth}
    \setlength{\figureheight}{.9\figurewidth}
    \begin{subfigure}{.2\linewidth}
        \centering
        \includegraphics[width=\linewidth]{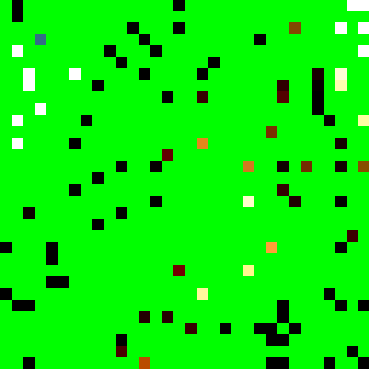}
        \caption{Context}
        \label{fig:plot1_dup}
    \end{subfigure}
    \begin{subfigure}{.2\linewidth}
        \centering
        \includegraphics[width=\linewidth]{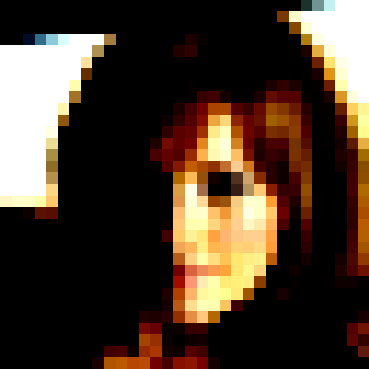}
        \caption{Full image}
        \label{fig:plot2_dup}
    \end{subfigure}
    \begin{subfigure}{.45\linewidth}
        \centering
        \scriptsize
        \input{image_files/output_plot_1_.tex}
        \caption{Classification probability for some features}
        \label{fig:plot3_dup}
    \end{subfigure}

    \caption{An example showing the classification ability of ACE. (\subref{fig:plot1_dup}) is the context available of the full image displayed in the panel (\subref{fig:plot2_dup}). The probabilities for a subset of features are in (\subref{fig:plot3_dup}).}
    \label{fig:image_classification}
\end{figure}

\paragraph{Results.} The main image completion results for the \textsc{CelebA} dataset are shown in \cref{fig:whole_image_celeb_main}, with the same experiment performed on \textsc{MNIST} and displayed in \cref{fig:whole_image}. In both figures, we display some example images and predictions and negative log-probability density for different context sizes (shaded area is 95\% confidence interval). Our method demonstrates a clear improvement over the TNP-D method across all context sizes on both datasets (\cref{fig:whole_image_celeb_main} and \cref{fig:whole_image}). Moreover, incorporating latent information for conditional generation further enhances the performance of our base method. A variation of the image completion experiment is shown in \cref{fig:masked_celeba_row}, where the context is no longer randomly sampled from within the image but instead selected according to a top 22-row image mask. For this example, the latent information \textsc{Bald} is either conditioned on \textsc{True} or \textsc{False}. The results show that the model adjusts its generated output based on the provided latent information, highlighting the potential of conditional generation. Furthermore, in \cref{fig:image_classification}, we show examples of ACE's ability to perform image classification showing a subset of the $40$ features in \textsc{CelebA} dataset. Despite only having $10 \%$ of the image available, ACE can predict most features successfully. Finally, in \cref{fig:mnist-accuracy} the accuracy for predicting the correct class label for \textsc{MNIST} is reported.

\label{app:image}

\paragraph{Sampling for Image experiments.}
\label{app:image_sampling}
For sampling, we use the full available dataset for both \textsc{MNIST} and \textsc{CelebA}, detailed in \cref{app:sampling}. In the \textsc{MNIST} dataset there is one latent class label and for \textsc{CelebA} all 40 features were used. In \cref{fig:masked_celeba_row}, the sampling procedure was adjusted to represent features that would influence the top 22 rows of the images. Therefore, we selected a subset of seven features, which were \textsc{Bald}, \textsc{Bangs}, \textsc{Black\_Hair}, \textsc{Blond\_Hair}, \textsc{Brown\_Hair}, \textsc{Gray\_Hair} and \textsc{Eyeglasses}. The same sampling procedure was performed again but, now on a smaller set of features.

\paragraph{Architecture and training.} For the image experiments, we used the same embedder layer as in the other experiments. Through grid search, we found that a transformer architecture with 8 heads, 6 layers, and a hidden dimension of 128 performed best. For the MLP layer, we used a dimension of 256. Finally, we reduced the number of components for the output head to $K=3$. We trained the model for $80,000$ iterations using a cosine scheduler with \textsc{Adam} \citep{kingma14}, with a learning rate $0.0005$ and a batch size of $64$.

\subsection{Bayesian optimization}
\label{app:bo}

This section presents ACE's Bayesian Optimization (BO) experiments (\cref{exp:bo} in the main paper) in more detail, including the training data generation, algorithms, benchmark functions, and baselines used in this paper.

\subsubsection{Bayesian Optimization dataset, architecture and training details} 
\label{sec:bo-dataset}

\paragraph{Dataset.} The BO datasets are generated by sampling from a GP, following the approach described in \cref{app:gp}. The sampling procedure is adjusted to include the known optimum location and value of the function within the generative process. The detailed dataset generation procedure is outlined as follows:

\begin{enumerate}
    \item Sampling GP hyper-parameters, to determine the base function shape: 
        \begin{itemize}
            \item First, we randomly select a kernel from a set comprising the RBF kernel and three Matérn kernels (Matérn-$1/2$, Matérn-$3/2$, and Matérn-$5/2$) based on predefined weights [$0.35$, $0.1$, $0.2$, $0.35$], corresponding respectively to the RBF kernel and the Matérn kernels in the specified order.
            \item Then, we sample whether the kernel is isotropic or not with $p=0.5$.
            \item The output scale $\sigma_f$ and lengthscales $l^{(k)}$ are sampled following the procedure outlined in \cref{app:gp}.
            \item We assume the GP (constant) mean to be 0 for now.
        \end{itemize}  
    \item Sampling the latent values, $\xopt$ and $\yopt$:
        \begin{itemize}
            \item  We sample the optimum location $\xopt$ uniformly inside $[-1,1]^D$.
            \item We sample the value of the global minimum $\yopt$ from a minimum-value distribution of a zero-mean Gaussian with variance equal to the output variance of the GP. The number of samples for the minimum-value distribution, $N$, is approximated as the number of uncorrelated samples from the GP in the hypercube, determined based on the GP's length scale. This approach ensures that $\yopt$ roughly respects the statistics of optima for the GP hyperparameters.
            \item With $p=0.1$ probability, we add $\Delta y \sim \exp(1)$ to the mean function to represent an “unexpectedly low” optimum.
        \end{itemize}

    \item Sampling from GP posterior to get the context and target sets:
        \begin{itemize}
            \item We build a posterior GP with the above specification and a single observation at $(\mathbf{x}_\text{opt}, y_\text{opt})$.
            \item We sample a total of $100 \cdot D$ (context + target) locations where the number of context points is sampled similarly to the GP dataset generation. The maximum number of context points is 50 for the 1D case and 100 for both 2D and 3D cases. 
            \item Then, the values of this context set are jointly sampled from a GP posterior conditioned on one observation at $(\xopt,\yopt)$. 
            \item Instead, the target points are sampled \emph{independently} from a GP posterior conditioned on $\data_N$ (the previously sampled context points) and $(\xopt,\yopt)$. Independent sampling of the targets speeds up GP data generation and is valid since \emph{during training} we only predict 1D marginal distributions at the target points.
        \end{itemize}

    \item Further adjustment of $y$, and consequently $\yopt$:
        \begin{itemize}
            \item To ensure that the global optimum is at $(\xopt,\yopt)$ we add a convex envelope (a quadratic component). Specifically, we transform the $y$ values of the datasets as $y^\prime_i =|y_i| + \frac{1}{5}||\xopt - \x_i||^2$ where $\x_i$ and $y_i$ are the input and output values of all sampled points. 
            \item Lastly, we add an offset to the $y^\prime$ values of sampled points uniformly drawn from $[-5, 5]$, meaning that $\yopt \in [-5, 5]$.
        \end{itemize}

\end{enumerate}

\begin{figure}[t] 
    \centering
    \includegraphics[width=1.\linewidth]{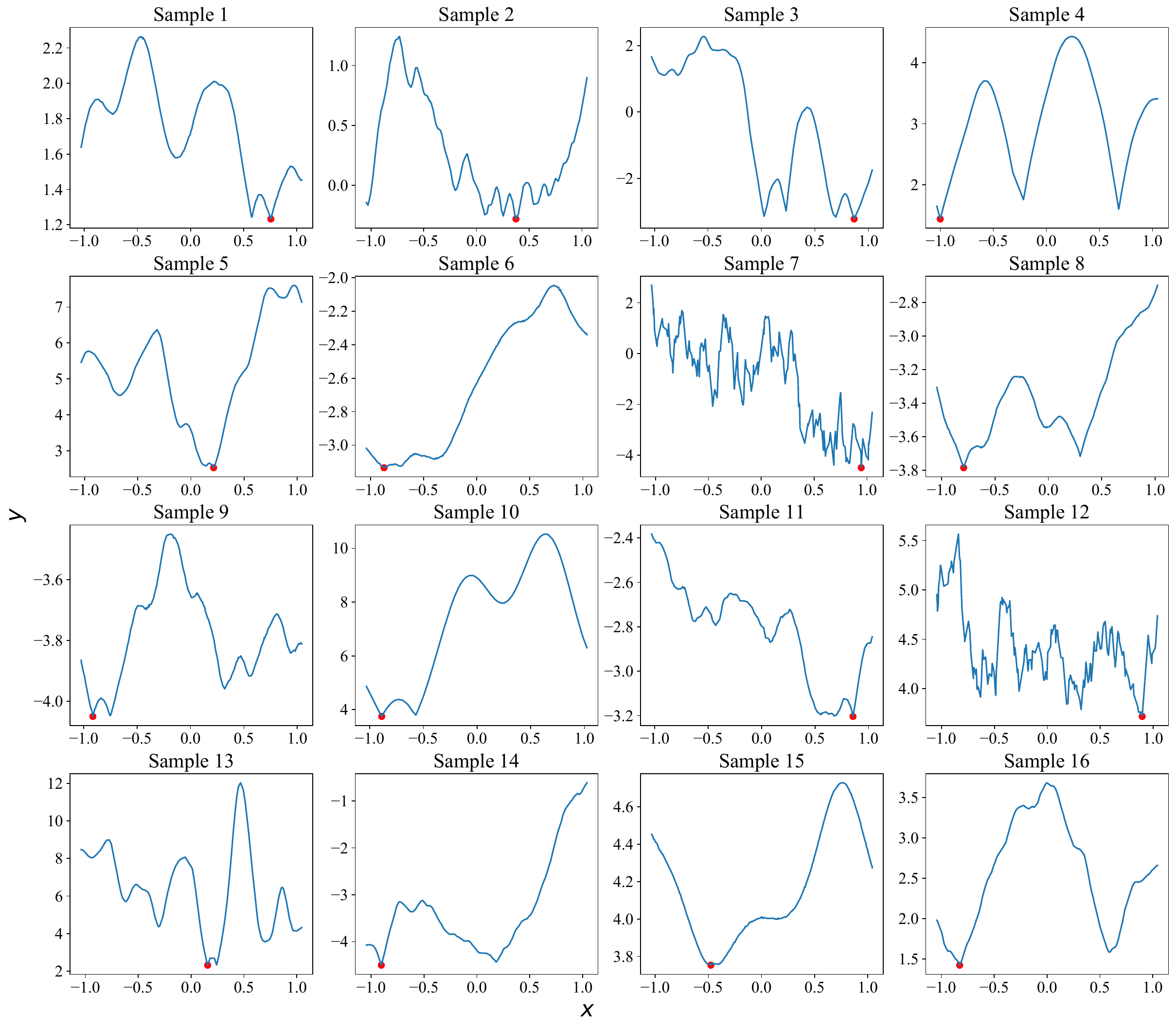}
    \caption{One-dimensional Bayesian optimization dataset samples, with their global optimum (red dot).} 
    \label{fig:bo_dataset_1d}
\end{figure}

\begin{figure}[th!] 
    \centering
    \includegraphics[width=1.\linewidth]{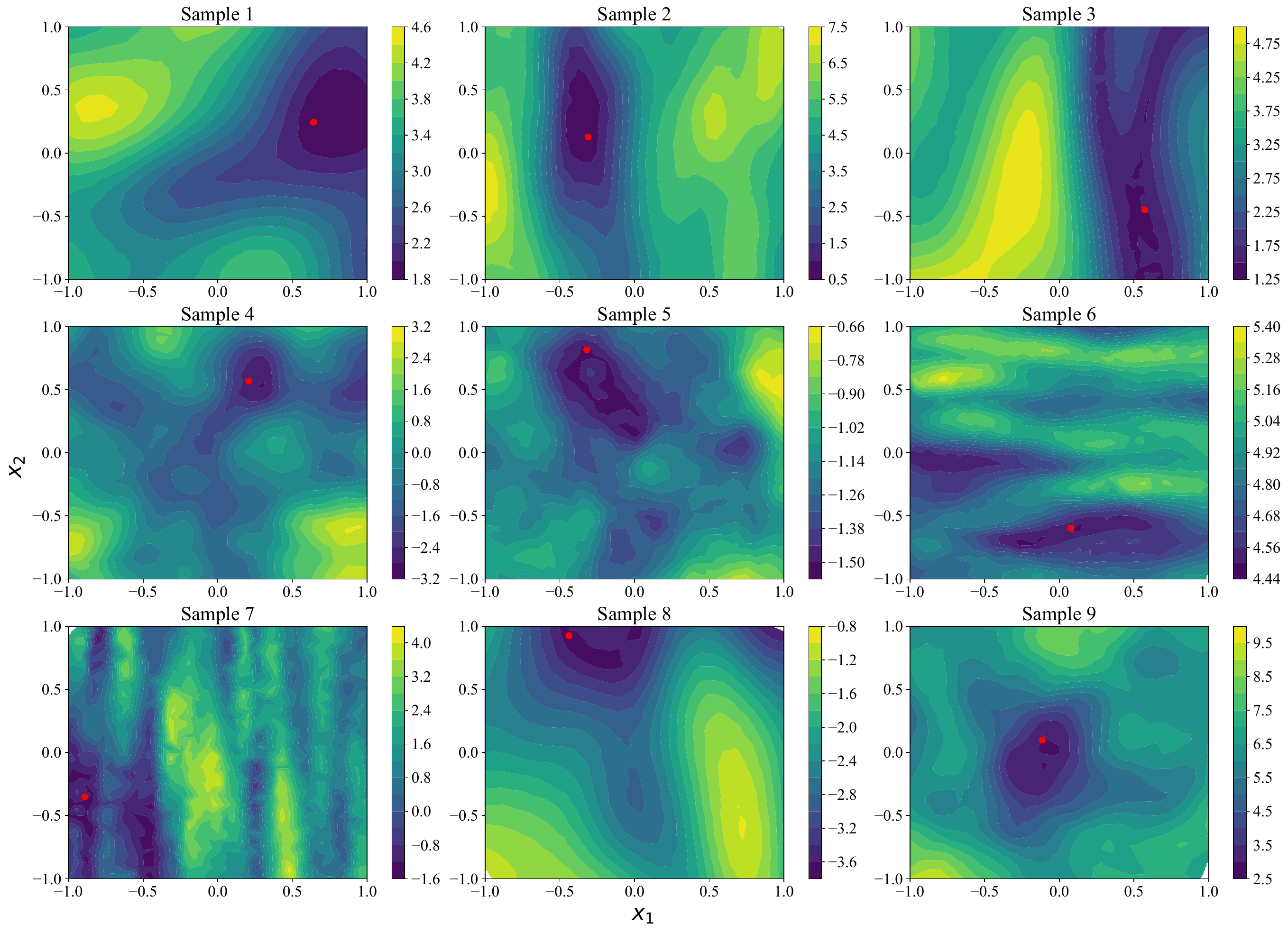}
    \caption{Two-dimensional Bayesian optimization dataset samples, with their optimum (red dot).} 
    \label{fig:bo_dataset_2d}
\end{figure}

One and two-dimensional examples of the sampled functions are illustrated in \cref{fig:bo_dataset_1d} and \cref{fig:bo_dataset_2d}, respectively.

\paragraph{Architecture and training details.}
In the Bayesian Optimization (BO) experiments, the ACE model was configured differently depending on the dimensionality of the problem. 
For the 1-3D cases, the model used an embedding dimension of $ D_\text{emb} = 256$ with six transformer layers. Each attention block had 16 heads, while the MLP block had a hidden dimension of 128. The output head consisted of $K=20$ MLP components, each with a hidden dimension of 128. For the 4-6D cases, the model was configured with embedding dimension of $D_\text{emb} = 128$ while still using six transformer layers. Each attention block had 8 heads, and the MLP block had a hidden dimension of 512. The output head structure remained unchanged, consisting of $K=20$ MLP components, each with a hidden dimension of 128. The model configuration varied with problem dimensionality to balance capacity and efficiency.

The model was trained for $5\times 10^5$ steps with a batch size of 64 for 1-3D cases and $3.5\times 10^5$ steps and 128 batch size for 4-6D cases, using learning rate $5 \times 10^{-4}$ with cosine annealing. We apply loss weighing to give more importance to the latent variables during training. This adjustment accounts for the fact that the number of latent variables, $ n_{\text{latent}}$, is generally much smaller than the number of data points, $(n_{\text{total}}-n_{\text{latent}})$. The weight assigned to the latent loss is calculated as $w_{\text{latent}} =  ({n_{\text{total}} - \nicefrac{1}{2} \left( \text{max\_ctx} + \text{min\_ctx} \right)}/{n_{\text{latent}}})^{T}$
where $T$ is a tunable parameter, $\text{max\_ctx}$ and $\text{max\_ctx}$ are the maximum and minimum number of context points during the dataset generation. We conducted a grid search over $T = 1, \nicefrac{2}{3}, \nicefrac{1}{3}, 0$ to identify the best-performing model. In our experiments, the optimal $T$ values are $T = 1$ for 1D, $T = \nicefrac{2}{3}$ for 2D and 3D, and $T = 0$ for 4D--6D. 
Note that ACE has different models trained with different datasets for each input dimensionality.

\subsubsection{ACE-BO Algorithm}
\label{sec:ace-bo-alg}

\begin{algorithm}
\caption{ACE-Thompson Sample (D>1)}\label{alg:ACE-TS-ND}
    \begin{algorithmic}[1]
    \Require observed data points $\data_N = \{\x_{1:N}, y_{1:N}\}$, improvement parameter $\alpha$, input dimensionality  $D \in \mathbb{N}^{+}$, whether to condition on $\yopt$\ or not flag $c \in \{\text{True}, \text{False}\}$.
    \State \textbf{Initialization} $y_\text{min} \gets \min{y_{1:N}}$, $y_\text{max} \gets \max{y_{1:N}}$.
    \If {$c$ is $\text{True}$}
        \State set threshold value: $\tau \gets y_\text{min} - \alpha \max(1,y_\text{max}-y_\text{min}) $.
        \State sample $\yopt$ from mixture truncated at $\tau$: $\yopt \sim p(\yopt| \data_N, \yopt<\tau)$. 
    \EndIf
    \State randomly permute dimension indices: $(1, \dots, D) \rightarrow (\pi_1, \dots, \pi_D)$. \Comment{$\vpi$ is permutation of $(1, \dots, D)$}
    \For{$i \gets \pi_1, \dots, \pi_D$} 
        \If {$c$ is $\text{True}$}
            \State sample $x^{i}_\text{opt}$ conditioned on $\yopt$, $\data_N$, and already sampled $\xopt$ dimensions if any:
            \State $x^{i}_\text{opt} \sim p(x^{i}_\text{opt}| \data_N, \yopt, x^{0:i-1}_\text{opt})$.
        \Else
            \State sample $x^{i}_\text{opt}$ conditioned on $\data_N$ and already sampled $\xopt$ dimensions if any:
            \State $x^{i}_\text{opt} \sim p(x^{i}_\text{opt}| \data_N, x^{0:i-1}_\text{opt})$.
        \EndIf
    \EndFor
    \State get full value of $\xopt$ using the true indices: $\xopt \gets (x^{1}_\text{opt}, \dots ,x^{D}_\text{opt})$.

    \State \Return $\xopt$
    \end{algorithmic}
\end{algorithm}

\begin{algorithm}
\caption{ACE-MES}\label{alg:ACE-MES}
    \begin{algorithmic}[1]
    \Require observed data points $\data_N = \{x_{1:N}, y_{1:N}\}$, number of candidate points $N_\text{cand}$, Thompson sampling ratio for candidate point $TS_\text{ratio}$.
    \State \textbf{Initialization} $N_{TS1} \gets N_\text{cand} \times TS_\text{ratio}$, $N_{TS2} \gets N_\text{cand} \times (1-TS_\text{ratio})$.
    \State propose $N_\text{cand}$ candidate points $X^*_{1:N_\text{cand}}$ according to $TS_\text{ratio}$: 
    \State sample $X^*_{1:N_{TS1}}$ using ACE-TS with conditioning on $\yopt$ ($c = \text{True}$). 
    \State sample $X^*_{N_{TS1}+1:N_{TS1}+N_{TS2}}$ using ACE-TS without conditioning on $\yopt$ ($c = \text{True}$). 
    \For {$i \gets 1$ \textbf{to} $N_\text{cand}$}:
        \State sample $\yopt$ for conditioning: $\yopt \sim p(\yopt| \data_N)$. 
        \State $\alpha_{(i)}(\mathbf{x}^*_{(i)}) = H[p(y^*_{(i)}| \mathbf{x}^*_{(i)}, \data_N)] - \mathbb{E}(H[p(y^*_{(i)}| \mathbf{x}^*_{(i)}, \data_N, \yopt)])$ \Comment{see \cref{sec:ace-bo-alg} for more detail}
    \EndFor
    \State $\xopt = \text{arg}\,\max \boldsymbol{\alpha}$.
    \State \Return $\xopt$
    \end{algorithmic}
\end{algorithm}

\paragraph{Bayesian optimization with Thompson sampling (ACE-TS).} For Thompson sampling, we sample the query point at each step from $p(\xopt| \data_N, \yopt < \tau)$ where $\tau$ is a threshold lower than the minimum point seen so far. This encourages exploration to sample a new point that is lower than the current optimum. We set $\tau=y_\text{min} - \alpha \max(1, y_\text{max}-y_\text{min})$, where $y_\text{max}$ and $ y_\text{min}$ are the maximum and minimum values currently observed so far, and $\alpha$ a parameter controlling the minimum improvement. We set $\alpha = 0.01$ throughout all experiments. First, we sample $\yopt$ from a truncated mixture of Gaussian obtained from ACE's predictive distribution $p(\yopt| \data_N)$, truncated for $\yopt < \tau$. After that, we sample $\xopt$ conditioned on that sampled $\yopt$ (i.e., sample from $p(\xopt|\data_N, \yopt < \tau)$). 
For higher dimension $(D>1)$ we sample $\xopt$ in an autoregressive manner, one dimension at a time. The order of the dimensions is randomly permuted to mitigate order bias among the dimensions. The detailed pseudocode for ACE-TS (D>1) is presented in Algorithm \cref{alg:ACE-TS-ND}. An example evolution of ACE-TS is reported in \cref{fig:bo_evolution}.

\paragraph{Bayesian optimization with Minimum-value Entropy Search (ACE-MES).}

For Minimum-value Entropy Search (MES; \citealp{wang2017max}), the procedure is as follows:
\begin{enumerate}
\item First, we propose $N_\text{candidate}$ points. We generate these candidate points by sampling 80\% of them using the conditional Thompson sampling approach described earlier, i.e., $p(\xopt| \data_N, \yopt < \tau)$, and the remaining 20\% directly from $p(\xopt|\data_N)$. In our experiments we use $N_\text{candidate} = 20$. 
\item For each candidate point $\xs$, we evaluate the acquisition function, which in our case is the gain in mutual information between the maximum $\yopt$ and the candidate point $\xs$ (\cref{eq:mes}).

\item To compute the first term of the right-hand side of \cref{eq:mes}, for a candidate point $\xs$, we calculate the predictive distribution $p(y^\star| \mathbf{x}^\star, \data_N)$ represented in our model by a mixture of Gaussians. We compute its entropy via numerical integration over a grid.

\item For the second term of the right-hand side of \cref{eq:mes}, we perform Monte Carlo sampling to evaluate the expected entropy. For each candidate point $\xs$, we draw $N_\text{mc}$ samples of $\yopt$ from the predictive distribution $p(\yopt | \data_N)$. We set $N_\text{mc} = 20$ to ensure the procedure remains efficient while maintaining accuracy.
\item For each sampled $\yopt$, we determine the predictive distribution $p(y^\star| \mathbf{x}^\star, \data_N, \yopt)$. Then, for each mixture, we compute the entropy as in step 2. We then average over samples to compute the expectation.
\item To compute the estimated MES value of candidate point $\xs$ we subtract the computed first term to the second term of the equation \cref{eq:mes}.
\item We repeat this procedure for all candidate points $\xs$ and select the point with the highest information gain. This point is expected to yield the lowest uncertainty about the value of the minimum, thus guiding our next query in the Bayesian optimization process.
\end{enumerate}

To illustrate the implementation details of ACE-MES, we present its pseudocode in \cref{alg:ACE-MES}.

\begin{figure*}[t!]
  \centering
  \scriptsize
  \setlength{\figurewidth}{.35\columnwidth}
  \setlength{\figureheight}{.9\figurewidth}
  \begin{subfigure}[b]{.33\textwidth}
    \centering
    \input{figures/acebo1}
  \end{subfigure}
  \hspace{0.01cm}
  \begin{subfigure}[b]{.33\textwidth}
    \centering
    \input{figures/acebo2}
  \end{subfigure}
  \hspace{-0.2cm}
  \begin{subfigure}[b]{.33\textwidth}
    \centering
    \input{figures/acebo3}
  \end{subfigure}
  \caption{\textbf{Bayesian optimization example.} We show here an example evolution of ACE-TS on a 1D function. The orange pdf on the left of each panel is $p(\yopt|\data_N)$, the red pdf at the bottom of each panel is $p(x_\text{opt}|\yopt,\data_N)$, for a sampled $\yopt$ (orange dashed-dot line). The queried point at each iteration is marked with a red asterisk, while black and blue dots represent the observed points. Note how ACE is able to learn complex conditional predictive distributions for $\xopt$ and $\yopt$.}
  \label{fig:bo_evolution}
\end{figure*}

\subsubsection{Bayesian optimization with prior over $\xopt$}
\label{sec:bo_prior}

ACE is capable of injecting a prior over latents when predicting values and latents. In the context of BO, this prior could incorporate information about the location of the optimum, $\xopt$. Several works, such as \citep{souza2021bayesian, hvarfner2022pi, muller2023pfns4bo}, have explored the use of priors in BO to improve predictive performance. In our experiments, we evaluate two types of priors: strong and weak, to assess the robustness of the model under varying levels of prior knowledge. As a baseline, we utilize a $\pi$BO-like procedure \citep{hvarfner2022pi}, as described below, to perform Thompson sampling across all experiments.

\paragraph{Training.} For training, we generate a prior distribution similar to \cref{app:priors}, but with slight adjustments: when sampling the mixture distribution, we include a 50\% chance of adding a uniform component. If present, the uniform distribution weight $w_\text{unif}$ is sampled uniformly from 0.0 to 0.2 (otherwise $w_\text{unif} = 0$). The uniform component is then added as follows:
\begin{align}
   \mathbf{p} = (w_\text{unif} \cdot \mathbf{p}_\text{unif}) + (1 - w_\text{unif}) \cdot \mathbf{p}_\text{mixture}, 
\end{align}
where $\mathbf{p}_\text{unif}$ represents the uniform component, and $\mathbf{p}_\text{mixture}$ is the sampled mixture. The inclusion of a uniform component during training means that the prior can be a mixture of an informative and a non-informative (flat) component, which will be useful later.
Using this binned distribution, we then sample $\xopt$ and $\yopt$, and use these two latent samples to construct our function, as outlined in \cref{sec:bo-dataset}.

\paragraph{Testing.} During the BO testing phase, we consider two scenarios: 
\begin{enumerate}
    \item Strong prior: We first sample a mean for the $\xopt$ prior by drawing from a Gaussian distribution centered on the true $\xopt$ with a standard deviation set to $10\%$ of the domain (in our case $[-1,1]$), resulting in a standard deviation of $0.2$. We use this sampled prior mean and standard deviation to construct the binned prior. 
    \item Weak prior: The same steps are applied to generate the prior, but with a standard deviation of $25\%$, which translates to $0.5$ for our domain. 
\end{enumerate}
In both scenarios, we add a uniform prior component with $w_\text{uniform}=0.1$. The uniform component helps with model and prior misspecification, by allowing the model to explore outside the region indicated by the prior.

We compare ACE with Prior Thompson Sampling (ACEP-TS) to the no-prior ACE-TS and a baseline GP-TS. We also consider a state-of-the-art heuristic for prior injection in BO, $\pi$BO  \citep{hvarfner2022pi}, with the TS acquisition procedure described below ($\pi$BO-TS).  The procedure is repeated 10 times for each case, with different initial points sampled at random.

\paragraph{$\pi$BO-TS.} 

The main technique in $\pi$BO for injecting a prior in the BO procedure consists of rescaling the chosen acquisition function $\alpha(\x)$ by the user-provided prior over the optimum location $\pi(\x)$ (Eq. 6 in \citealp{hvarfner2022pi}), 
\begin{equation} \label{eq:pibo}
\alpha_\text{$\pi$BO}(\x; \alpha) \propto \alpha(\x) \pi(\x)^{\gamma_n},
\end{equation}
where $n$ is the BO iteration and $\gamma_n$ governs the relative influence of the prior with respect to the acquisition function, which is heuristically made to decay over iterations to reflect the increased role of the observed data. As in \citet{hvarfner2022pi}, we set $\gamma_n = \frac{\beta}{n}$ where $\beta$ is a hyperparameter reflecting the user confidence on the prior.

To implement Thompson sampling (TS) with $\pi$BO, we first note that the TS acquisition function $\alpha_\text{TS}(\x)$ corresponds to the current posterior probability over the optimum location, and the TS procedure consists of drawing one sample from this acquisition function (as opposed to optimizing it). Thus, the $\pi$BO variant of TS ($\pi$BO-TS) corresponds to sampling from \cref{eq:pibo}, where the current posterior over the optimum takes the role of $\alpha(\x)$. We sample from \cref{eq:pibo} using a self-normalized importance sampling-resampling approach \citep{robert2004monte}. Namely, we sample $N_\text{TS} = 100$ points from $\alpha_\text{TS}$ using batch GP-TS, then resample one point from this batch using importance sampling weight $w \propto \frac{\alpha_\text{TS}(\x) \pi(\x)^{\beta/n}}{\alpha_\text{TS}(\x)} = \pi(\x)^{\beta/n}$, where all weights are then normalized to sum to 1. Following \citep{hvarfner2022pi}, we set $\beta = 10$, i.e., equal to their setting when running BO experiments with 100 iterations, as in our case.

\subsubsection{Benchmark functions and baselines}
\label{sec:bo-benchmarks}
\paragraph{BO benchmarks.} We use a diverse set of benchmark functions with input dimensions ranging from 1D to 6D to thoroughly evaluate ACE's performance on the BO task. These include (1) the non-convex Ackley function in both 1D and 2D, (2) the 1D Gramacy-Lee function, known for its multiple local minima, (3) the 1D Negative Easom function, characterized by a sharp, narrow global minimum and deceptive flat regions, (4) the non-convex 2D Branin Scaled function with multiple global minima, (5) the 2D Michalewicz function, which features a complex landscape with multiple local minima, (6) the 3D, 5D, and 6D Levy function, with numerous local minima due to its sinusoidal component, (7) the 5D and 6D Griewank function, which is highly multimodal and regularly spaced local minima, but a single smooth global minimum, (8) the 4D and 5D Rosenbrock function, which has a narrow, curved valley containing the global minimum, and (9) the 3D, 4D, and 6D Hartmann function, a widely used standard benchmark. These functions present a range of challenges, allowing us to effectively test the robustness and accuracy of ACE across different scenarios. 

\input{bo_extra}

\paragraph{BO baselines.} For our baselines, we employ three methods: autoregressive Thompson Sampling with TNP-D (AR-TNPD-TS) \citep{nguyen2022transformer}, Gaussian Process-based Bayesian Optimization with the MES acquisition function (GP-MES) \citep{wang2017max}, and Gaussian Process-based Thompson Sampling (GP-TS) with 5000 candidate points. In addition, we use $\pi$BO-TS for the prior injection case as the baseline \cite{hvarfner2022pi} (using the same number of candidate points used in GP-TS). We optimize the acquisition function in GP-MES using the `shotgun' procedure detailed later in this section (with 1000 candidate points for minimum value approximation via Gumbel sampling). Both GP-MES and GP-TS implementations are written using the BOTorch library \citep{balandat2020botorch}. For AR-TNPD-TS, we use the same network architecture configurations as ACE, but with a non-linear embedder and a single Gaussian head  \citep{nguyen2022transformer}. Additionally, AR-TNPD-TS uses autoregressive sampling, as described in \citep{bruinsma2023autoregressive}. 

We conducted our experiments with 100 BO iterations across all benchmark functions. The number of initial points was set to 3 for 1D experiments and 10 for 2D-6D experiments. These initial points were drawn uniformly randomly within the input domain. We evaluated the runtime performance of our methods and baseline algorithms on a local machine equipped with a 13th Gen Intel(R) Core(TM) i5-1335U processor and 15GB of RAM. On average, the runtime for 100 BO iterations was as follows: ACE-TS and ACEP-TS completed in approximately 5 seconds; ACE-MES required about 1.3 minutes; GP-TS and $\pi$BO-TS  took roughly 2 minutes; GP-MES took about 1.4 minutes; and AR-TNPD-TS was the slowest, requiring approximately 10 minutes, largely due to the computational cost of its autoregressive steps.

\paragraph{Shotgun optimizer.} To perform fast optimization in parallel, we first sample 10000 points from a quasirandom grid using the Sobol sequence. Then we pick the point with the highest acquisition function value, referred to as $\mathbf{x}_0$. Subsequently, we sample 1000 points around $\x_0$ using a multivariate normal distribution with diagonal covariance $\sigma^2 \mathbf{I}$, where the initial $\sigma$ is set to the median distance among the points. We re-evaluate the acquisition function over this neighborhood, including $\x_0$, and select the best point. After that, we reduce $\sigma$ by a factor of five and repeat the process, iterating from the current best point. This `shotgun' approach allows us to zoom into a high-valued region of the acquisition function while exploiting large parallel evaluations.

\input{bop_weak}
\input{bop_strong}

\subsubsection{Additional Bayesian optimization results.} 

\paragraph{Standard BO setting additional results.}

Additional results in \cref{fig:bo_comparisons_extra} complement those in \cref{fig:bo_comparisons}. While our method performs generally well across different benchmark functions, we find that it struggles on the Michalewicz function, likely because its sharp, narrow optima and highly irregular landscape differ significantly from the function classes used during training. Conversely, ACE performs competitively on Griewank, where the structured landscape aligns well with our approach. On the 2D Ackley function, the challenge may stem from its highly non-stationary nature, while our method was trained only on draws from stationary kernels. Addressing functions like Michalewicz and Ackley may require extending our relatively simple function generation process and incorporating specialized techniques like input and output warping~\citep{muller2023pfns4bo} to better handle non-stationarity.

\paragraph{BO with prior over $\xopt$ additional results.}
Additional results on the weak prior scenario are presented in \cref{fig:bo_comparisons_weak} and with strong prior in \cref{fig:bo_comparisons_strong}. The results indicate that ACEP-TS consistently outperforms ACE-TS, particularly when using a strong prior. In this case, the model benefits from the prior information, leading to a notable improvement in performance. Specifically, the strong prior allows the model to converge more rapidly toward the optimum.

\subsection{Simulation-based inference} \label{app:sbi}

\subsubsection{Simulators}\label{app:sbi-simulators}

The experiments reported in \cref{exp:sbi} used three time-series models to simulate the training and test data. This section describes the simulators in more details. 

\textbf{Ornstein-Uhlenbeck Process (OUP)} is widely used in financial mathematics and evolutionary biology due to its ability to model mean-reverting stochastic processes \citep{uhlenbeck1930theory}. The model is defined as:
\begin{equation*}
    y_{t+1} = y_t + \Delta y_t, \quad
    \Delta y_t = \theta_1 \left[\exp(\theta_2) - y_t \right] \Delta t + 0.5w, \quad \text{ for } t = 1, \ldots, T,
\end{equation*}
where $ T = 25 $, $\Delta t = 0.2 $, $ x_0 = 10 $, and $w \sim \mathcal{N}(0, \Delta t) $. We use a uniform prior $ U([0, 2] \times [-2, 2]) $ for the latent variables $ \vtheta = (\theta_1, \theta_2)$ to generate the simulated data.

\textbf{Susceptible-Infectious-Recovered (SIR)} is a simple compartmental model used to describe infectious disease outbreaks \citep{kermack1927contribution}. The model divides a population into susceptible (S), infectious (I), and recovered (R) individuals. Assuming population size $N$ and using $S_t$,  $I_t$, and $R_t$ to denote the number of individuals in each compartment at time $t$, $t = 1, \ldots, T$,
the disease outbreak dynamics can be expressed as
\begin{equation*}
\Delta S_t = -\beta\frac{I_t S_t}{N}, \quad
\Delta I_t = \beta\frac{I_t S_t}{N} - \gamma I_t, \quad
\Delta R_t = \gamma I_t,
\end{equation*}
where the parameters $\beta$ and $\gamma$ denote the contact rate and the mean recovery rate.
An observation model with parameters $\phi$ is used to convert the SIR model predictions to observations~$(t, y_t)$. The experiments carried out in this work consider two observation models and simulator setups.


The setups considered in this work are as follows. First, we consider a SIR model with fixed initial condition and $10$ observations $y_t\sim\mathrm{Bin}(1000, I_t/N)$ collected from $T=160$ time points at even interval, as proposed in \citep{lueckmann2021benchmarking}. Here the population size $N=10^6$ and the initial condition is fixed as $S_0=N-1$, $I_0=1$, $R_0=0$. We use uniform priors $\beta\sim U(0.01, 1.5)$ and $\gamma\sim U(0.02, 0.25)$. We used this model version in the main experiments presented in \cref{exp:sbi} and \cref{app:sbi_cfg}.


In addition we consider a setup where $N$ and $I_0$ are unknown and we collect $25$ observations $y_t\sim\mathrm{Poi}(\phi I_t/N)$ from $T=250$ time points at even interval. We use $\beta\sim U(0.5, 3.5)$, $\gamma\sim U(0.0001, 1.5)$, $\phi\sim U(50, 5000)$, and $I_0/N\sim U(0.0001, 0.01)$ with $S_0/N = 1 - I_0/N$ and $R_0/N = 0$ to generate simulated samples. We used this model version in an additional experiment to test ACE on real world data, presented in \cref{app:sbi_real}.

\textbf{Turin} model is a time-series model used to simulate radio propagation phenomena, making it useful for testing and designing wireless communication systems \citep{turin1972statistical, pedersen2019stochastic, bharti2019estimator}. The model generates high-dimensional complex-valued time-series data and is characterized by four key parameters that control different aspects of the radio signal: $G_0$ controls the reverberation gain, $T$ determines the reverberation time, $\nu$ specifies the arrival rate of the point process, and $\sigma^2_W$ represents the noise variance.

The model starts with a frequency bandwidth $B=0.5$ GHz and simulates the transfer function $H_k$ over $N_s = 101$ equidistant frequency points. The measured transfer function at the $k$-th point, $Y_k$, is given by:
\begin{equation*}
Y_k = H_k + W_k, \quad k = 0, 1, \ldots, N_s - 1,
\end{equation*}
where $W_k$ denotes additive zero-mean complex circular symmetric Gaussian noise with variance $\sigma^2_W$. The transfer function $H_k$ is defined as:
\begin{equation*}
H_k = \sum_{l=1}^{N_{\text{points}}} \alpha_l \exp(-j 2 \pi \Delta f k \tau_l),
\end{equation*}
where $\tau_l$ are the time delays sampled from a one-dimensional homogeneous Poisson point process with rate $ \nu$, and $\alpha_l$ are complex gains. The gains $\alpha_l$ are modeled as i.i.d. zero-mean complex Gaussian random variables conditioned on the delays, with a conditional variance:
\begin{equation*}
\mathbb{E}[\vert \alpha_l \vert^2 | \tau_l] = \frac{G_0 \exp(-\tau_l / T)}{\nu}.
\end{equation*}

The time-domain signal $\tilde{y}(t)$ can be obtained by taking the inverse Fourier transform:
\begin{equation*}
\tilde{y}(t) = \frac{1}{N_s} \sum_{k=0}^{N_s - 1} Y_k \exp(j 2 \pi k \Delta f t),
\end{equation*}
with $\Delta f = B / (N_s - 1) $ being the frequency separation. Our final real-valued output is calculated by taking the absolute square of the complex-valued data and applying a logarithmic transformation $y(t) = 10 \log_{10}(|\tilde{y}(t)|^2)$.

The four parameters of the model are sampled from the following uniform priors: $G_0 \sim \mathcal{U}(10^{-9}, 10^{-8})$, $ T \sim \mathcal{U}(10^{-9}, 10^{-8})$, $ \nu \sim \mathcal{U}(10^{7}, 5 \times 10^{9}) $, $ \sigma^2_W \sim \mathcal{U}(10^{-10}, 10^{-9})$.

\subsubsection{Main experiments} \label{app:sbi_cfg}

ACE was trained on examples that included simulated time series data and model parameters divided between target and context.
In these experiments, the time series data were divided into context and target data by sampling $N_d$ data points into the context set and including the rest in the target set. The context size $N_d\sim U(10, 25)$ in the OUP experiments, $N_d\sim U(5, 10)$ in the SIR experiments, and $N_d\sim U(50, 101)$ in the Turin experiments. 
In addition, the model parameters were randomly assigned to either the context or target set.
NPE and NRE cannot handle partial observations and was trained with the full time series data in both cases.

The ACE model used in these experiments had embedding dimension $64$ and $6$ transformer layers.
The attention blocks had $4$ heads and the MLP block had hidden dimension $128$. The output head had $K=20$ MLP components with hidden dimension $128$. The model was trained for $5\times 10^4$ steps with batch size $32$, using learning rate $5 \times 10^{-4}$ with cosine annealing. 

We used the \texttt{sbi} package \citep{tejero-cantero2020sbi} (\url{https://sbi-dev.github.io/sbi/}, Version: 0.22.0, License: Apache 2.0) to implement NPE and NRE. Specifically, we chose the NPE-C \citep{greenberg2019automatic} and NRE-C \citep{miller2022contrastive} with Masked Autoregressive Flow (MAF) \citep{papamakarios2017masked} as the inference network. We used the default configuration with $50$ hidden units and $5$ transforms for MAF, and training with a fixed learning rate $5 \times 10^{-4}$. 
For Simformer \citep{gloeckler2024all}, we used their official package (\url{https://github.com/mackelab/simformer}, Version: 2, License: MIT). We used the same configuration as in our setup for the transformer, while we used their default configuration for the diffusion part. For a fair comparison, we pre-generated $10^4$ parameter-simulation pairs for all methods. We also normalized the parameters of the Turin model when feeding into the networks. For evaluation, we randomly generated 100 observations and assessed each method across 5 runs. For the RMSE evaluation, given $N_{\text{obs}}$ observations, with $N_{\text{post}}$ posterior samples generated for each observation, and $L$ latent parameters, our RMSE metric is calculated as:
\begin{equation*}
    \text{RMSE} = \frac{1}{N_{\text{obs}}}\sum_{i=1}^{N_{\text{obs}}}\sqrt{\frac{1}{L \cdot N_{\text{post}}}\sum_{l=1}^{L}\sum_{j=1}^{N_{\text{post}}}({\theta}_{i,l} - \hat{\theta}_{i,l,j})^2},
\end{equation*}
where $\theta_{i,l}$ represents the true value of the $l$-th latent parameter for the $i$-th observation, and $\hat{\theta}_{i,l,j}$ represents the $j$-th posterior sample of the $l$-th latent parameter for the $i$-th observation. This approach first calculates the RMSE for each observation (averaging across all latent dimensions and posterior samples for that observation), and then averages these observation-specific RMSE values to obtain the final metric. For MMD, we use an exponentiated quadratic kernel with a lengthscale of 1.

\paragraph{Statistical comparisons.} We evaluate models based on their average results across multiple runs and perform pairwise comparisons to identify models with comparable performance. The results from pairwise comparisons are used in \cref{tab:sbi} to highlight in bold the models that are considered best in each experiment. The following procedure is used to determine the best models:
\begin{itemize}
    \item First, we identify the model (A) with the highest empirical mean and highlight it in bold.
    \item For each alternative model (B), we perform $10^5$ bootstrap iterations to resample the mean performance for both model A and model B.
    \item We then calculate the proportion of bootstrap iterations where model B outperforms model A.
    \item If this proportion is larger than the significance level ($\alpha=0.05$), model B is considered statistically indistinguishable from model A.
    \item All models that are not statistically significantly different from the best model are highlighted in bold.
\end{itemize}

\subsubsection{Ablation study: Gaussian vs. mixture-of-Gaussians output heads} \label{app:ablation}

To assess the impact of using a Gaussian versus a mixture-of-Gaussians output head in ACE, we conduct an ablation study on the SBI tasks. In theory, a mixture-of-Gaussians head should improve performance when the predictive or posterior data distributions are non-Gaussian. Table~\ref{tab:ablation_gaussian_gmm} shows the results. As expected, we observe improvements in OUP and Turin when using a mixture-of-Gaussians head. This suggests that more flexible distributional families better capture complex distributions. However, for the SIR task, the performance difference is negligible as the posteriors are largely Gaussian. These findings align with our expectations.

\begin{table*}[h]
\centering
\scalebox{0.88}{
\begin{tabular}{cc|cc}
\toprule
      &     & Gaussian (ablation) & Mixture-of-Gaussians (ACE) \\
           \cmidrule(lr){3-4}
\multirow{3}{*}{OUP} & $\text{log-probs}_{\theta}$ ($\uparrow$) &   0.90\textcolor{gray}{(0.01)} &\textbf{1.03}\textcolor{gray}{(0.02)}   \\ 
                     & $\text{RMSE}_{\theta}$ ($\downarrow$)    &  0.48\textcolor{gray}{(0.01)}    &   0.48\textcolor{gray}{(0.00)}    \\  
                     & $\text{MMD}_{y}$ ($\downarrow$) &  0.52\textcolor{gray}{(0.00)} & \textbf{0.51}\textcolor{gray}{(0.00)} \\  
                     
\midrule \midrule
\multirow{3}{*}{SIR} & $\text{log-probs}_{\theta}$ ($\uparrow$) &    6.80\textcolor{gray}{(0.02)}   &  6.78\textcolor{gray}{(0.02)}  \\ 
                     & $\text{RMSE}_{\theta}$ ($\downarrow$)    &  0.02\textcolor{gray}{(0.00)}    &   0.02\textcolor{gray}{(0.00)}   \\  
                     & $\text{MMD}_{y}$ ($\downarrow$) &  0.02\textcolor{gray}{(0.00)} & 0.02\textcolor{gray}{(0.00)}  \\  
\midrule \midrule
\multirow{3}{*}{Turin} & $\text{log-probs}_{\theta}$ ($\uparrow$) &   2.73\textcolor{gray}{(0.02)}  & \textbf{3.14}\textcolor{gray}{(0.02)}     \\ 
                     & $\text{RMSE}_{\theta}$ ($\downarrow$)    &   0.24\textcolor{gray}{(0.00)}   &  0.24\textcolor{gray}{(0.00)}   \\  
                     & $\text{MMD}_{y}$ ($\downarrow$) &  0.36\textcolor{gray}{(0.00)} & \textbf{0.35}\textcolor{gray}{(0.00)} \\  
\bottomrule
\end{tabular}}

\caption{Ablation study comparing single Gaussian versus mixture-of-Gaussians output heads across SBI tasks. Mean and standard deviation from 5 runs are reported. mixture-of-Gaussians heads benefit complex distributions (OUP and Turin), while maintaining similar performance on simpler tasks (SIR).}
\label{tab:ablation_gaussian_gmm}
\end{table*}

\vspace{-2mm}
\subsubsection{Simulation-based calibration} \label{app:sbc}
To evaluate the calibration of the approximate posteriors obtained by ACE, we apply simulation-based calibration (SBC; \citealt{talts2018validating}) on the Turin model to evaluate whether the approximate posteriors produced by ACE are calibrated.
We recall that SBC checks if a Bayesian inference process is well-calibrated by repeatedly simulating data from parameters drawn from the prior and inferring posteriors under those priors and simulated datasets.
If the inference is calibrated, the average posterior should match the prior. Equivalently, when ranking the true parameters within each posterior, 
the ranks should follow a uniform distribution \citep{talts2018validating}.

We use the following procedure for SBC: for a given prior, we first sample 1000 samples from the prior distribution and generate corresponding simulated data. Then we use ACE to approximate the posteriors and subsequently compare the true parameter values with samples drawn from the inferred posterior distribution. 
To visualize the calibration, we plot the density of the posterior samples against the prior samples. If the model is well-calibrated, the posterior distribution should recover the true posterior, which results in a close match between the density of the posterior samples and the prior. We also present the fractional rank statistic against the ECDF difference \citep{sailynoja2022graphical}. Ideally, the ECDF difference between the rank statistics and the theoretical uniform distribution should remain close to zero, indicating well-calibrated posteriors.

\begin{figure}[ht] 
    \centering
    \includegraphics[width=\linewidth]{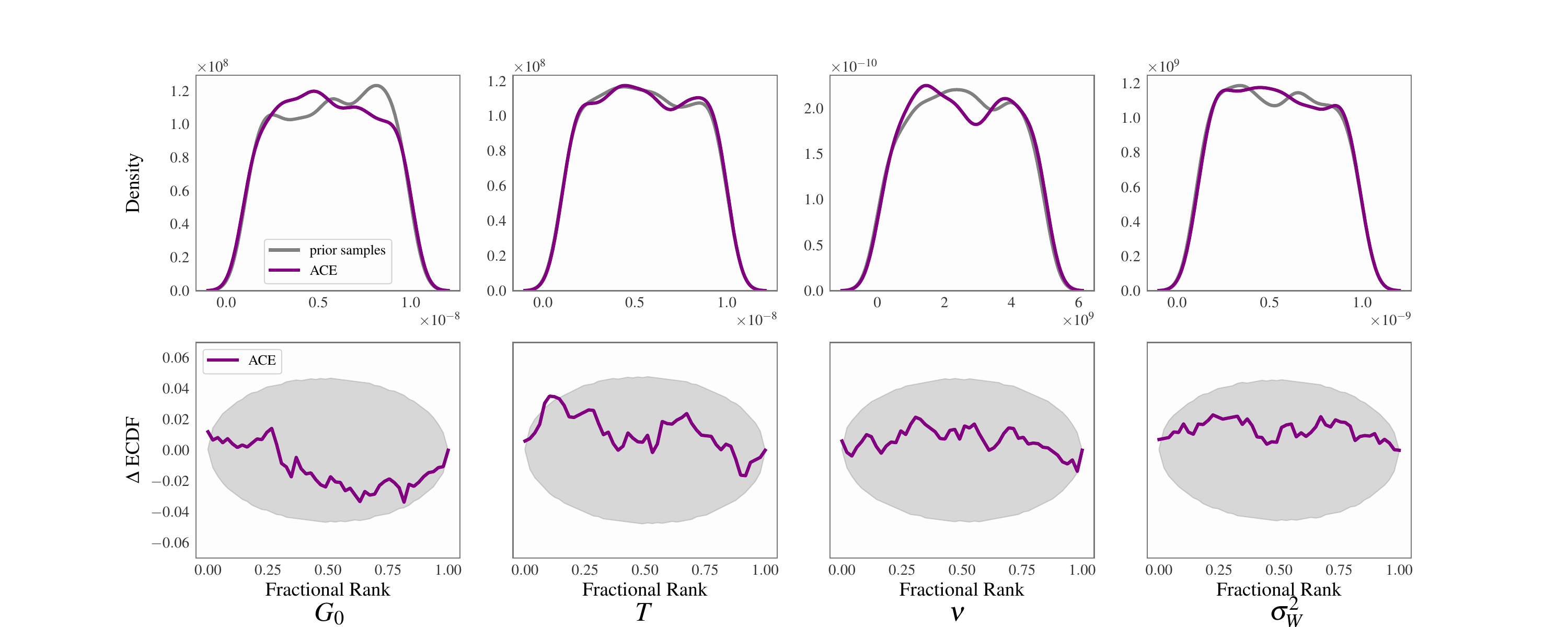}
    \caption{Simulation-based calibration of ACE on the Turin model. The top row shows the density of the posterior samples from ACE compared with the prior samples. The bottom row shows the fractional rank statistic against the ECDF difference with 95\% confidence bands. ACE is well-calibrated.}

    \label{fig:sbc}
\end{figure}

\begin{figure}[ht] 
    \centering
    \includegraphics[width=\linewidth]{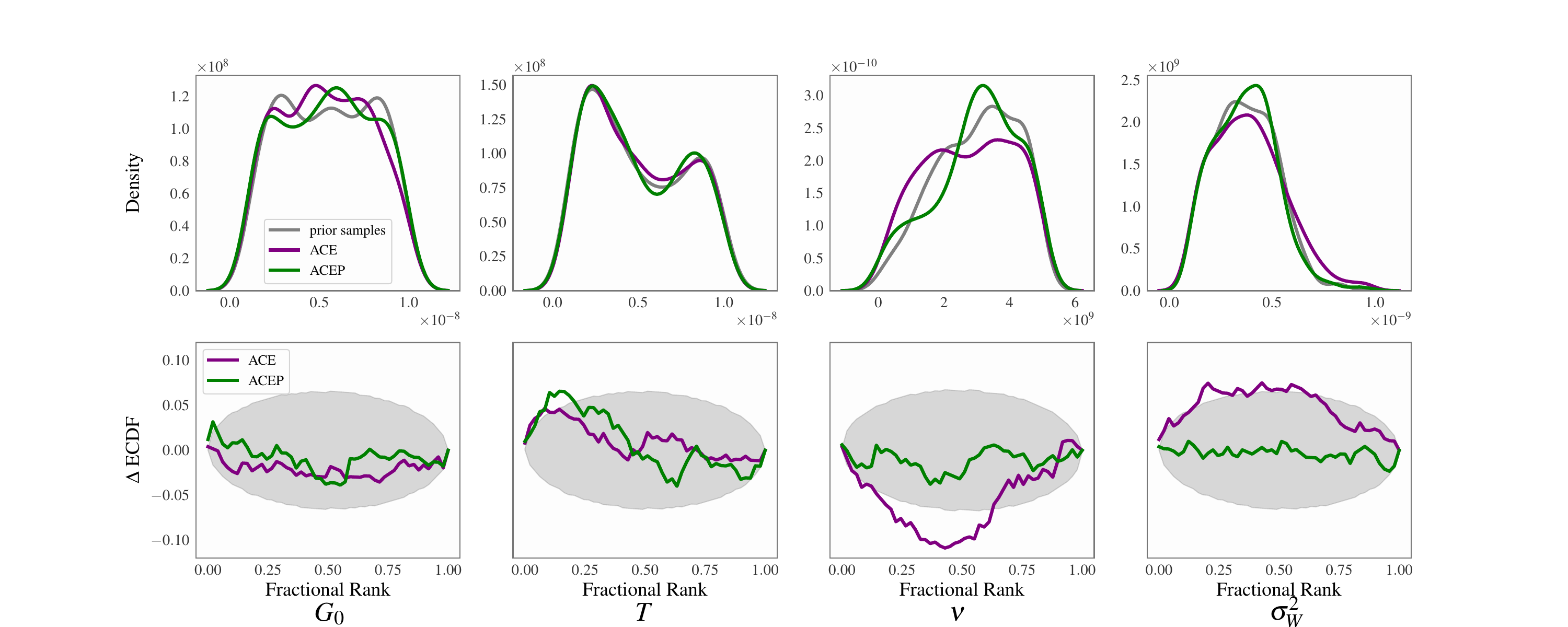}
    \caption{Simulation-based calibration of ACE and ACEP on the Turin model with an example custom prior. ACEP demonstrates improved calibration by closely following the prior distribution and showing lower deviations in the ECDF difference, highlighting its ability to condition on user-specified priors effectively.}

    \label{fig:sbc_pi}
\end{figure}

\cref{fig:sbc} shows that our ACE is well-calibrated with pre-defined uniform priors across all four latents. Since ACEP allows conditioning on different priors at runtime, we also test the calibration of ACEP using randomly generated priors (following \cref{app:prior}). For comparison, we show what happens if we forego prior-injection, using vanilla ACE instead of ACEP.  The visualization on one set of priors is shown in \cref{fig:sbc_pi}. As expected, vanilla ACE (without prior-injection) does not include the correct prior information and shows suboptimal calibration performance, whereas ACEP correctly leverages the provided prior information and shows closer alignment with the prior and lower ECDF deviations. We also calculate the average absolute deviation over 100 randomly sampled priors. In the prior-injection setting, ACEP demonstrates better calibration, with an average deviation of $0.03 \pm 0.01$ compared to $0.10 \pm 0.05$ for ACE without the correct prior.

\subsubsection{Extended SIR model on real-world data}
\label{app:sbi_real}

We present here results obtained by considering an extended four-parameter version of the SIR model then applied to real-world data. We further include details on the training data and model configurations used in the real-data experiment as well as additional evaluation results from experiments carried out with simulated data.
As our real-world data, we used a dataset that describes an influenza outbreak in a boarding school. The dataset is available in the R package \texttt{outbreaks} (\url{https://cran.r-project.org/package=outbreaks}, Version: 1.9.0, License: MIT).

\paragraph{Methods.}

The four-parameter SIR model we used is detailed in \cref{app:sbi-simulators} (last paragraph).
The ACE models were trained with samples constructed based on simulated data as follows. The observations were divided into context and target points by sampling $N_d\sim U(2, 20)$ data points into the context set and $2$ data points into the target set. The examples included $50$\% interpolation tasks where the context and target points were sampled at random (without overlap) and $50$\% forecast tasks where the points were sampled in order. The model parameters were divided between the context and target set by sampling the number to include $N_l\sim U(0, 4)$ and sampling the $N_l$ parameters from the parameter set at random. The parameters were normalized to range $[-1, 1]$ and the observations were square-root compressed and scaled to the approximate range $[-1, 1]$. 

The ACE models had the same architecture as the models used in the main experiment, but the models were trained for $10^5$ steps with batch size $32$. In this experiment, we generated the data online during the training, which means that the models were trained with $3.2 \times 10^6$ samples. The NPE models used in this experiment had the same configuration as the model used in the main experiment, for fair comparison, the models were now trained with $3.2 \times 10^6$ samples. Each sample corresponded to a unique simulation and the full time series was used as the observation data.

To validate model predictions, we note that ground-truth parameter values are not available for real data. Instead, we examined whether running the simulator with parameters sampled from the posterior can replicate the observed data. For reference, we also included MCMC results. The MCMC posterior was sampled with Pyro \citep{bingham2018pyro} (\url{https://pyro.ai/}, Version: 1.9.0, License: Apache 2.0) using the random walk kernel and sampling $4$ chains with $5\times 10^4$ warm-up steps and $5 \times 10^4$ samples. 

\paragraph{Results.}

\begin{figure}[h]
    \raggedleft
    \resizebox{0.9\textwidth}{!}{\input{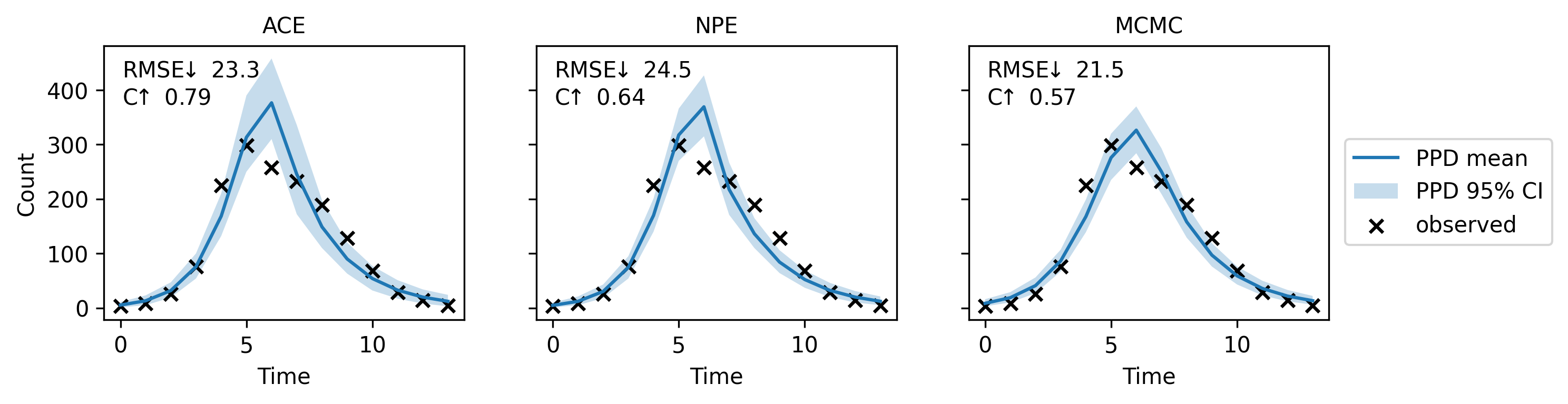}}
    \caption{\textbf{SIR model on a real dataset.} Posterior predictive distributions based on the ACE, NPE, and MCMC posteriors. 
    The dataset is mildly misspecified, in that even MCMC does not fully match the data.}
    \label{fig:sir-example}
    \vspace{-2mm}
\end{figure}

The posterior predictive distributions and log-probabilities for observed data calculated based on ACE, NPE, and MCMC results are shown in \cref{fig:sir-example}. For this visualization, ACE and NPE models were trained once, and simulations were carried out with $5000$ parameters sampled from each posterior distribution. The log-probabilities observed in this experiment are $-64.4$ with ACE, $-64.6$ with NPE. Repeating ACE and NPE training and posterior estimation $10$ times, the average log-probabilities across the $10$ runs were $-65.1$ (standard deviation $0.4$) with ACE and $-65.5$ (standard deviation $0.7$) with NPE, showing a similar performance. The ACE predictions used in this experiment are sampled autoregressively (see \cref{app:autoregressive}). These results show that ACE can handle inference with real data.

\paragraph{Validation on simulated data.}

For completeness, we performed a more extensive validation of ACE and other methods with the extended SIR model using simulated data.
Specifically, we assessed the ACE and NPE models on simulated data and evaluated the same ACE models in a data completion task with the TNP-D baseline. All the training details remain the same as in the real-world experiment for ACE and NPE. The TNP-D models had the same overall architecture as ACE but used a different embedder and output head. The MLP block in the TNP-D embedder had hidden dimension $64$ and the MLP block in the single-component output head hidden dimension $128$. The TNP-D models were trained for $10^5$ steps with batch size $32$. The evaluation set used in these experiments included $1000$ simulations sampled from the training distribution and the evaluation metrics included log-probabilities and coverage probabilities calculated based on 95\% quantile intervals that were estimated based on $5000$ samples.

We start with the posterior estimation task where we used ACE and NPE to predict simulator parameters based on the simulated observations with $25$ observation points. 
The results are reported in \cref{tab:sbi_sir_ext_ace_npe}. We observe that the ACE log-probabilities are on average better than NPE log-probabilities and that both methods have marginal coverage probabilities close to the nominal level $0.95$.

\begin{table}[t]
    \caption{Comparison between ACE and NPE in posterior estimation task in the extended SIR model. The ACE predictions were generated autoregressively so both methods target the joint posterior. The estimated posteriors are compared based on log-probabilities and 95\% marginal coverage probabilities. The evaluation set includes 1000 examples and we report the mean and \textcolor{gray}{(standard deviation)} from $10$ runs. ACE log-probabilities are on average better than NPE log-probabilities and the coverage probabilities are close to the nominal level $0.95$.}
    \label{tab:sbi_sir_ext_ace_npe}
    \centering
    \begin{tabular}{lcccccc}
    \toprule
    & log-probs ($\uparrow$) & cover $\beta$ & cover $\gamma$ & cover $\phi$ & cover $I_0$ & cover ave \\
    \midrule
    NPE & 6.63 \textcolor{gray}{(0.16)} & 0.92 \textcolor{gray}{(0.01)} & 0.94 \textcolor{gray}{(0.01)} & 0.94 \textcolor{gray}{(0.01)} & 0.92 \textcolor{gray}{(0.01)} & 0.93 \textcolor{gray}{(0.01)} \\
    ACE (AR) & 7.38 \textcolor{gray}{(0.04)} & 0.96 \textcolor{gray}{(0.00)} & 0.97 \textcolor{gray}{(0.00)} & 0.97 \textcolor{gray}{(0.00)} & 0.96 \textcolor{gray}{(0.00)} & 0.97 \textcolor{gray}{(0.00)} \\
    \bottomrule
    \end{tabular}
\end{table}

The simulated observations used in the previous experiment were complete with $25$ observation points. Next, we evaluate ACE posteriors estimated based on incomplete data with $5$--$20$ observation points. NPE is not included in this experiment since it cannot handle incomplete observations. Instead, we use this experiment to compare independent and autoregressive ACE predictions. The results are reported in \cref{tab:sbi_sir_ext_ace}. The log-probabilities indicate that both independent and autoregressive predictions improve when more observation points are available while the coverage probabilities are close to the nominal level in all conditions. That autoregressive predictions result in better log-probabilities than independent predictions indicates that ACE is able to use dependencies between simulator parameters.

\begin{table}[t]
    \caption{ACE posterior estimation based on incomplete data with $M$ observation points using either independent or autoregressive predictions. The estimated posteriors are evaluated using (a) log-probabilities and (b) average 95\% marginal coverage probabilities. We report the mean and \textcolor{gray}{(standard deviation)} from $10$ runs. The log-probabilities improve when the context size $M$ increases and when autoregressive predictions are used.}
    \label{tab:sbi_sir_ext_ace}
    \centering
    \begin{tabular}{cc}
    (a) &
    \begin{tabular}{lccccc}
    \toprule
    & $M=25$ & $M=20$ & $M=15$ & $M=10$ & $M=5$ \\
    \midrule
     ACE & 4.94 \textcolor{gray}{(0.04)} & 4.55 \textcolor{gray}{(0.03)} & 3.87 \textcolor{gray}{(0.02)} & 2.82 \textcolor{gray}{(0.03)} & 0.88 \textcolor{gray}{(0.03)}\\
     ACE (AR) & 7.38 \textcolor{gray}{(0.04)} & 6.93 \textcolor{gray}{(0.04)} & 6.21 \textcolor{gray}{(0.04)} & 5.11 \textcolor{gray}{(0.04)} & 2.91 \textcolor{gray}{(0.05)}\\
     \midrule\midrule
    \end{tabular}\\
    (b) &
    \begin{tabular}{lccccc}
     ACE & 0.97 \textcolor{gray}{(0.00)} & 0.96 \textcolor{gray}{(0.00)} & 0.95 \textcolor{gray}{(0.00)} & 0.95 \textcolor{gray}{(0.00)} & 0.96 \textcolor{gray}{(0.00)}\\
     ACE (AR) & 0.97 \textcolor{gray}{(0.00)} & 0.97 \textcolor{gray}{(0.00)} & 0.96 \textcolor{gray}{(0.00)} & 0.96 \textcolor{gray}{(0.00)} & 0.97 \textcolor{gray}{(0.00)}\\
    \bottomrule
    \end{tabular}
    \end{tabular}
\end{table}

\begin{table}[h]
    \caption{Comparison between ACE and TNP-D in data completion task in the extended SIR model. The estimated predictive distributions are compared based on (a) log-probabilities and (a) 95\% coverage probabilities. We report the mean and \textcolor{gray}{(standard deviation)} from $10$ runs. ACE log-probabilities are on average better than TNP-D log-probabilities and improve both when the context size $M$ increases or when predictions are conditioned on the simulator parameters $\theta$. }
    \label{tab:sbi_sir_ext_data_pred}
    \centering
    \begin{tabular}{cc}
    (a) &
    \begin{tabular}{lcccc}
    \toprule
    & $M=20$ & $M=15$ & $M=10$ & $M=5$ \\
    \midrule
     TNP-D & 10.1 \textcolor{gray}{(0.11)} & 9.99 \textcolor{gray}{(0.09)} & 9.44 \textcolor{gray}{(0.10)} & 8.02 \textcolor{gray}{(0.07)} \\
     ACE & 14.2 \textcolor{gray}{(0.31)} & 13.8 \textcolor{gray}{(0.31)} & 13.2 \textcolor{gray}{(0.31)} & 11.4 \textcolor{gray}{(0.28)} \\
     ACE + $\theta$ & 14.7 \textcolor{gray}{(0.31)} & 14.6 \textcolor{gray}{(0.31)} & 14.6 \textcolor{gray}{(0.30)} & 14.3 \textcolor{gray}{(0.30)} \\
     \midrule\midrule
    \end{tabular}\\
    (b) &
    \begin{tabular}{lccccc}
     TNP-D & 0.96 \textcolor{gray}{(0.00)} & 0.96 \textcolor{gray}{(0.00)} & 0.95 \textcolor{gray}{(0.00)} & 0.95 \textcolor{gray}{(0.00)} \\
     ACE & 0.97 \textcolor{gray}{(0.00)} & 0.96 \textcolor{gray}{(0.00)} & 0.96 \textcolor{gray}{(0.00)} & 0.95 \textcolor{gray}{(0.00)} \\
     ACE + $\theta$ & 0.96 \textcolor{gray}{(0.00)} & 0.96 \textcolor{gray}{(0.00)} & 0.96 \textcolor{gray}{(0.00)} & 0.96 \textcolor{gray}{(0.00)} \\
    \bottomrule
    \end{tabular}
    \end{tabular}
\end{table}

The same ACE models that have been evaluated in the posterior estimation (latent prediction) task can also make predictions about the unobserved values in incomplete data. To evaluate ACE in the data completion task, we selected $5$ target observations from each evaluation sample and used $5$--$20$ remaining observations as context. We used ACE to make target predictions either based on the context data alone or based on both context data and the simulator parameters $\theta$. For comparison, we also evaluated data completion with TNP-D.
The results are reported in \cref{tab:sbi_sir_ext_data_pred}. 
We observe that ACE log-probabilities are on average better than TNP-D log-probabilities and improve when simulator parameters are available as context.
In these experiments, both ACE and TNP-D were used to make independent predictions.


\subsection{Computational resources and software}
\label{app:computation}
For the experiments and baselines, we used a GPU cluster containing AMD MI250X GPUs. All experiments can be run using a single GPU with a VRAM of 50GB. Most of the experiments took under 6 hours, with the exception of a few BO experiments that took around 10 hours. The core code base was built using Pytorch \citep{paszke2019pytorch} (\url{https://pytorch.org/} Version: 2.2.0, License: modified BSD license) and based on the Pytorch implementation for TNP \citep{nguyen2022transformer} (\url{https://github.com/tung-nd/TNP-pytorch}, License: MIT). Botorch \citep{balandat2020botorch} (\url{https://github.com/pytorch/botorch} Version: 0.10.0, License: MIT) was used for the implementation of GP-MES, GP-TS, and $\pi$BO-TS.

%% file: architecture/architecture_tnpd.tex
\begin{figure}[H]
    \centering
    \begin{tikzpicture}[
        nodeColor/.style={fill=black!60, text=white, minimum width=1.1cm},
        redTextNode/.style={fill=black!60, text=red, minimum width=1.1cm},
        embedderStyle/.style={draw, rectangle, minimum height=6.5cm, minimum width=2cm, color=black!50, fill=blue!10},
        mhsaStyle/.style={draw, rounded corners, color=purple, inner sep=0.3cm, label={[below, yshift=0.5cm, text=purple]MHSA}},
        caStyle/.style={draw, rectangle, rounded corners, minimum height=2.5cm, minimum width=1cm, color=purple},
        headStyle/.style={draw, rectangle, minimum height=2.5cm, minimum width=1cm, color=black!50, fill=blue!10},
        lossStyle/.style={draw, rectangle, rounded corners, color=orange, fill=none, inner sep=0.15cm, label={[below, yshift=0.5cm, text=orange]Loss}}
    ]
        \node (x1) [nodeColor] at (-10, 6.5) {($x_1,y_1$)};
        \node (x3) [nodeColor, below=0.5cm of x1] {($x_3,y_3$)};
        \node (x5) [nodeColor, below=0.5cm of x3] {($x_5,y_5$)};
        \node (x2) [redTextNode, below=0.5cm of x5] {($x_2$)};
        \node (x4) [redTextNode, below=0.5cm of x2] {($x_4$)};
        \node (x6) [redTextNode, below=0.5cm of x4] {($x_6$)};
        
        \node (embedder) [embedderStyle] at (-6.9, 3.7) {{\Large Embedder}};
        \foreach \i in {x1, x3, x5, x2, x4, x6} {
            \draw[->, thick] (\i.east) -- (embedder.west |- \i);
        }
        
        \node (z1) [nodeColor] at (-2.4, 6.5) {($z_1$)};
        \node (z3) [nodeColor, below=0.5cm of z1] {($z_3$)};
        \node (z5) [nodeColor, below=0.5cm of z3] {($z_5$)};
        \node (z2) [redTextNode, below=0.5cm of z5, xshift=-2cm] {($z_2$)};
        \node (z4) [redTextNode, below=0.5cm of z2] {($z_4$)};
        \node (z6) [redTextNode, below=0.5cm of z4] {($z_6$)};
        
        \foreach \i/\j in {x1/z1, x3/z3, x5/z5, x2/z2, x4/z4, x6/z6} {
            \draw[->, thick] (embedder.east |- \i) -- (\j.west);
        }
        
        \node (mhsa) [mhsaStyle, fit=(z1) (z3) (z5)] {};
        \node (cross-attention) [caStyle, below=0.4cm of mhsa] {CA};
        \foreach \i in {z2, z4, z6} {
            \draw[->, thick] (\i.east) -- (cross-attention.west |- \i);
        }
        \draw[->, thick, color=purple] (mhsa) -- (cross-attention);
        
        \node (mhsa_ca_box) [draw, rounded corners, fit=(mhsa) (cross-attention), inner sep=0.05cm, color=blue, label={[below, yshift=-6.5cm, text=blue]k-blocks}] {};
        
        \node (head) [headStyle, right=1.0cm of cross-attention] {{\Large Head}};
        \foreach \i in {z2, z4, z6} {
            \draw[->, thick] (cross-attention.east |- \i) -- (head.west |- \i);
        }
        
        \node (y2h) [draw, fill=none, text=black, right=5cm of z2] {($\hat{y_2}$)};
        \node (y2) [redTextNode, right=0.5cm of y2h] {($y_2$)};
        \node (y4h) [draw, fill=none, text=black, below=0.5cm of y2h] {($\hat{y_4}$)};
        \node (y4) [redTextNode, right=0.5cm of y4h] {($y_4$)};
        \node (y6h) [draw, fill=none, text=black, below=0.5cm of y4h] {($\hat{y_6}$)};
        \node (y6) [redTextNode, right=0.5cm of y6h] {($y_6$)};
        \foreach \i/\j in {y2h/y2, y4h/y4, y6h/y6} {
            \draw[->, thick] (head.east |- \i) -- (\i.west);
        }
        
        \node (loss) [lossStyle, fit=(y2h) (y2) (y4h) (y4) (y6h) (y6)] {};
    \end{tikzpicture}
\caption{\textbf{A conceptual figure of TNP-D architecture.} The TNP-D architecture can be summarized in the embedding layer, attention layers and output head. The \textcolor{red}{$x$} denotes locations where the output is unknown (target inputs). The $z$ is the embedded data, while MHSA stands for multi head cross attention and CA for cross attention. The head for TNP-D is Gaussian, so it outputs a mean and variance for each target point.}
\label{fig:tnpd_architecture}
\end{figure}

%% file: architecture/architecture_ace.tex
\begin{figure}[H]
    \centering
    \begin{tikzpicture}[
        nodeColor/.style={fill=black!60, text=white, minimum width=1.1cm},
        greenNode/.style={fill=green, text=olive, minimum width=1.1cm},
        redTextNode/.style={fill=black!60, text=red, minimum width=1.1cm},
        embedderStyle/.style={draw, rectangle, minimum height=6.5cm, minimum width=2cm, color=black!50, fill=red!10},
        mhsaStyle/.style={draw, rounded corners, color=purple, inner sep=0.3cm, label={[below, yshift=0.5cm, text=purple]MHSA}},
        caStyle/.style={draw, rectangle, rounded corners, minimum height=2.7cm, minimum width=1cm, color=purple},
        headStyle/.style={draw, rectangle, minimum height=2.5cm, minimum width=1cm, color=black!50, fill=red!10},
        lossStyle/.style={draw, rectangle, rounded corners, color=orange, fill=none, inner sep=0.15cm, label={[below, yshift=0.5cm, text=orange]Loss}}
    ]
        \node (t1) [greenNode] at (-10, 6.5) {\textcolor{olive}{$(\theta_1)$}};
        \node (x3) [nodeColor, below=0.5cm of t1] {($x_3,y_3$)};
        \node (x5) [nodeColor, below=0.5cm of x3] {($x_5,y_5$)};
        \node (t2) [greenNode, below=0.5cm of x5] {\textcolor{olive}{$(?_{\theta_2})$}};
        \node (x4) [redTextNode, below=0.5cm of t2] {($x_4$)};
        \node (x6) [redTextNode, below=0.5cm of x4] {($x_6$)};
        
        \node (embedder) [embedderStyle] at (-6.9, 3.7) {{\Large Embedder}};
        \foreach \i in {t1, x3, x5, t2, x4, x6} {
            \draw[->, thick] (\i.east) -- (embedder.west |- \i);
        }
        
        \node (z1) [nodeColor] at (-2.4, 6.5) {($z_1$)};
        \node (z3) [nodeColor, below=0.5cm of z1] {($z_3$)};
        \node (z5) [nodeColor, below=0.5cm of z3] {($z_5$)};
        \node (z2) [redTextNode, below=0.5cm of z5, xshift=-2cm] {($z_2$)};
        \node (z4) [redTextNode, below=0.5cm of z2] {($z_4$)};
        \node (z6) [redTextNode, below=0.5cm of z4] {($z_6$)};
        
        \foreach \i/\j in {t1/z1, x3/z3, x5/z5, t2/z2, x4/z4, x6/z6} {
            \draw[->, thick] (embedder.east |- \i) -- (\j.west);
        }
        
        \node (mhsa) [mhsaStyle, fit=(z1) (z3) (z5)] {};
        \node (cross-attention) [caStyle, below=0.4cm of mhsa] {CA};
        \foreach \i in {z2, z4, z6} {
            \draw[->, thick] (\i.east) -- (cross-attention.west |- \i);
        }
        \draw[->, thick, color=purple] (mhsa) -- (cross-attention);
        
        \node (mhsa_ca_box) [draw, rounded corners, fit=(mhsa) (cross-attention), inner sep=0.05cm, color=blue, label={[below, yshift=-6.8cm, text=blue]k-blocks}] {};
        
        \node (head) [headStyle, right=1.0cm of cross-attention, align=center] {\Large Head \\ \\ (GMM \\ or \\ Cat)};
        \foreach \i in {z2, z4, z6} {
            \draw[->, thick] (cross-attention.east |- \i) -- (head.west |- \i);
        }
        
        \node (y2h) [draw, fill=none, text=black, right=5cm of z2] {$(\hat{\theta}_2)$};
        \node (y2) [greenNode, right=0.5cm of y2h] {\textcolor{olive}{$(\theta_2)$}};
        \node (y4h) [draw, fill=none, text=black, below=0.5cm of y2h] {($\hat{y_4}$)};
        \node (y4) [nodeColor, right=0.5cm of y4h] {($y_4$)};
        \node (y6h) [draw, fill=none, text=black, below=0.5cm of y4h] {($\hat{y_6}$)};
        \node (y6) [nodeColor, right=0.5cm of y6h] {($y_6$)};
        \foreach \i/\j in {y2h/y2, y4h/y4, y6h/y6} {
            \draw[->, thick] (head.east |- \i) -- (\i.west);
        }
        
        \node (loss) [lossStyle, fit=(y2h) (y2) (y4h) (y4) (y6h) (y6)] {};
    \end{tikzpicture}
    \caption{\textbf{A conceptual figure of ACE architecture.} The diagram shows key differences between ACE and TNP-D. The differences boil down to the embedder layer that now incorporates latents $\theta_l$ (and possibly priors over these) and the output head that is now a Gaussian mixture model (GMM, for continuous variables) or categorical (Cat, for discrete variables). Both latent and data can be of either type.}
    \label{fig:ace_architecture}
\end{figure}

%% file: figures/condition.tex
\begin{tikzpicture}

\definecolor{color0}{rgb}{1,0.701960784313725,0.27843137254902}
\definecolor{color1}{rgb}{0.462745098039216,0.690196078431373,0.254901960784314}
\definecolor{color2}{rgb}{0.12156862745098,0.466666666666667,0.705882352941177}

\begin{axis}[
axis on top,
enlarge x limits=false,
enlarge y limits=false,
height=\figureheight,
scale only axis,
tick align=outside,
tick pos=left,
tick pos=left,
width=\figurewidth,
legend cell align={left},
legend style={fill opacity=0.8, draw opacity=1, text opacity=1, at={(0.23,0.39)}, anchor=north west, draw=none},
outer sep=0pt,
scale only axis,
tick align=outside,
tick pos=left,
tick pos=left,
width=\figurewidth,
x grid style={white!69.0196078431373!black},
xlabel={Size of \(\displaystyle \mathcal{D}_N\)},
xmajorgrids,
xmin=2, xmax=25,
xtick style={color=black},
y grid style={white!69.0196078431373!black},
ylabel={Log predictive density $p(y| \cdot)$},
ymajorgrids,
ymin=-1.5, ymax=2,
ytick style={color=black}
]
\path [draw=color0, fill=color0, opacity=0.1]
(axis cs:2,-1.15163989410242)
--(axis cs:2,-1.68054847374121)
--(axis cs:4,-0.990842836013672)
--(axis cs:6,-0.430565379031441)
--(axis cs:8,-0.0121070820299595)
--(axis cs:10,0.373079639935928)
--(axis cs:15,1.00254884219883)
--(axis cs:20,1.41670055076651)
--(axis cs:25,1.64786510729674)
--(axis cs:25,2.04693965649721)
--(axis cs:25,2.04693965649721)
--(axis cs:20,1.75985484435984)
--(axis cs:15,1.39870719455959)
--(axis cs:10,0.846757441496414)
--(axis cs:8,0.487373538633963)
--(axis cs:6,0.0439439723570557)
--(axis cs:4,-0.517019135841492)
--(axis cs:2,-1.15163989410242)
--cycle;

\path [draw=color1, fill=color1, opacity=0.1]
(axis cs:2,-0.687558286865652)
--(axis cs:2,-1.12515490702778)
--(axis cs:4,-0.521242345477735)
--(axis cs:6,-0.0623614880333684)
--(axis cs:8,0.252451429830826)
--(axis cs:10,0.58584791863042)
--(axis cs:15,1.11968087926794)
--(axis cs:20,1.50328547047324)
--(axis cs:25,1.71680448161759)
--(axis cs:25,2.07778704059921)
--(axis cs:25,2.07778704059921)
--(axis cs:20,1.81285375071816)
--(axis cs:15,1.48593788370203)
--(axis cs:10,1.02193027293605)
--(axis cs:8,0.747302629483902)
--(axis cs:6,0.345082805965378)
--(axis cs:4,-0.104055236671771)
--(axis cs:2,-0.687558286865652)
--cycle;

\path [draw=color2, fill=color2, opacity=0.1]
(axis cs:2,-0.545716309691283)
--(axis cs:2,-0.909037790483315)
--(axis cs:4,-0.321526533766144)
--(axis cs:6,0.142545388674105)
--(axis cs:8,0.446746344717494)
--(axis cs:10,0.772574768604837)
--(axis cs:15,1.23876721667127)
--(axis cs:20,1.55032978354555)
--(axis cs:25,1.69977558004304)
--(axis cs:25,2.06704005606099)
--(axis cs:25,2.06704005606099)
--(axis cs:20,1.84416412822466)
--(axis cs:15,1.55945921841147)
--(axis cs:10,1.11645397235442)
--(axis cs:8,0.875894488392433)
--(axis cs:6,0.511831469690121)
--(axis cs:4,0.0589511283666704)
--(axis cs:2,-0.545716309691283)
--cycle;

\addplot [semithick, color0, opacity=0.9, dashed, mark=*, mark size=1.5, mark options={solid}]
table {%
2 -1.41609418392181
4 -0.753930985927582
6 -0.193310703337193
8 0.237633228302002
10 0.609918540716171
15 1.20062801837921
20 1.58827769756317
25 1.84740238189697
};
\addlegendentry{$p(y|\mathcal{D}_N)$}
\addplot [semithick, color1, opacity=0.9, mark=*, mark size=1.5, mark options={solid}]
table {%
2 -0.906356596946716
4 -0.312648791074753
6 0.141360658966005
8 0.499877029657364
10 0.803889095783234
15 1.30280938148499
20 1.6580696105957
25 1.8972957611084
};
\addlegendentry{$p(y|\vtheta, \mathcal{D}_N)$}
\addplot [semithick, color2, opacity=0.9, mark=*, mark size=1.5, mark options={solid}]
table {%
2 -0.727377050087299
4 -0.131287702699737
6 0.327188429182113
8 0.661320416554963
10 0.944514370479627
15 1.39911321754137
20 1.6972469558851
25 1.88340781805201
};
\addlegendentry{GP predictive}
\end{axis}
\end{tikzpicture}

%% file: figures/accuracy.tex
\begin{tikzpicture}

\definecolor{color0}{rgb}{1,0.701960784313725,0.27843137254902}

\begin{axis}[
axis on top,
enlarge x limits=false,
enlarge y limits=false,
height=\figureheight,
scale only axis,
tick align=outside,
tick pos=left,
tick pos=left,
width=\figurewidth,
x grid style={white!69.0196078431373!black},
xlabel={Size of $\data_N$},
xmajorgrids,
xmin=0.85, xmax=26.15,
xtick style={color=black},
y grid style={white!69.0196078431373!black},
ylabel={Kernel identification accuracy},
ymajorgrids,
ymin=0.35, ymax=1.0,
ytick style={color=black}
]
\path [draw=color0, fill=color0, opacity=0.3]
(axis cs:2,0.363301507491499)
--(axis cs:2,0.337498494129748)
--(axis cs:4,0.443692321967126)
--(axis cs:6,0.571794086403722)
--(axis cs:8,0.644912560609806)
--(axis cs:10,0.713566261909152)
--(axis cs:15,0.83409993351111)
--(axis cs:20,0.880357829839572)
--(axis cs:25,0.918458575635852)
--(axis cs:25,0.935141401857434)
--(axis cs:25,0.935141401857434)
--(axis cs:20,0.898042185991422)
--(axis cs:15,0.860300067061095)
--(axis cs:10,0.748033731796598)
--(axis cs:8,0.667087427946102)
--(axis cs:6,0.613005942397243)
--(axis cs:4,0.501107673455237)
--(axis cs:2,0.363301507491499)
--cycle;

\addplot [semithick, color0, opacity=0.7, mark=*, mark size=1.5, mark options={solid}]
table {%
2 0.350400000810623
4 0.472399997711182
6 0.592400014400482
8 0.655999994277954
10 0.730799996852875
15 0.847200000286102
20 0.889200007915497
25 0.926799988746643
};
\end{axis}
\end{tikzpicture}

%% file: figures/latent_pred.tex
\begin{tikzpicture}

\definecolor{color0}{rgb}{1,0.701960784313725,0.27843137254902}

\begin{axis}[
axis on top,
enlarge x limits=false,
enlarge y limits=false,
height=\figureheight,
scale only axis,
tick align=outside,
tick pos=left,
tick pos=left,
width=\figurewidth,
x grid style={white!69.0196078431373!black},
xlabel={Size of $\data_N$},
xmajorgrids,
xmin=2, xmax=26,
xtick style={color=black},
y grid style={white!69.0196078431373!black},
ylabel={Log predictive density $p(\theta | \mathcal{D}_N)$},
ymajorgrids,
ymin=0, ymax=0.6,
ytick style={color=black}
]
\path [draw=color0, fill=color0, opacity=0.3]
(axis cs:2,0.102959283316908)
--(axis cs:2,0.00810269847760135)
--(axis cs:4,0.127688840088769)
--(axis cs:6,0.229787630061289)
--(axis cs:8,0.284245793536527)
--(axis cs:10,0.324457451319319)
--(axis cs:15,0.400169790299197)
--(axis cs:20,0.465261654067067)
--(axis cs:25,0.47260496213666)
--(axis cs:25,0.568749618222365)
--(axis cs:25,0.568749618222365)
--(axis cs:20,0.551069893973496)
--(axis cs:15,0.486596816811953)
--(axis cs:10,0.414400414014238)
--(axis cs:8,0.372846138326738)
--(axis cs:6,0.315793667379674)
--(axis cs:4,0.235080598373054)
--(axis cs:2,0.102959283316908)
--cycle;

\addplot [semithick, color0, opacity=0.7, mark=*, mark size=1.5, mark options={solid}]
table {%
2 0.0555309908972545
4 0.181384719230912
6 0.272790648720481
8 0.328545965931632
10 0.369428932666779
15 0.443383303555575
20 0.508165774020282
25 0.520677290179513
};
\end{axis}
\end{tikzpicture}

%% file: image_mnist.tex
\begin{figure}[t!]
    \centering
    \begin{subfigure}[b]{0.50\textwidth}  
        \centering
        \begin{subfigure}[b]{0.17\linewidth}  
            \centering
            \includegraphics[width=\linewidth]{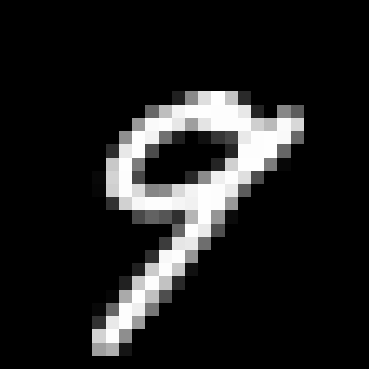}
        \end{subfigure}
        \hspace{-5pt}
        \begin{subfigure}[b]{0.17\linewidth}
            \centering
            \includegraphics[width=\linewidth]{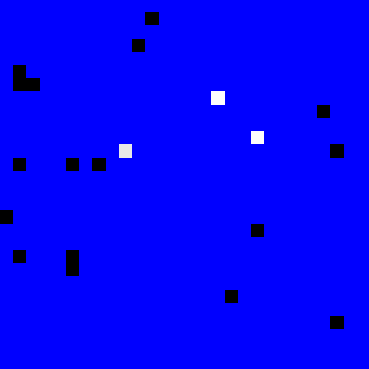}
        \end{subfigure}
        \hspace{-5pt}
        \begin{subfigure}[b]{0.17\linewidth}
            \centering
            \includegraphics[width=\linewidth]{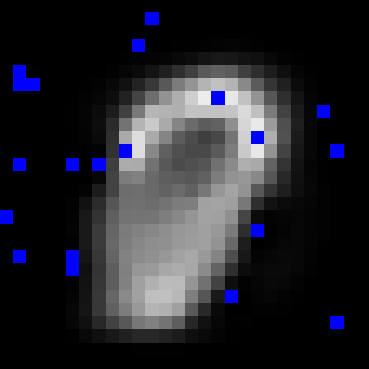}
        \end{subfigure}
        \hspace{-5pt}
        \begin{subfigure}[b]{0.17\linewidth}
            \centering
            \includegraphics[width=\linewidth]{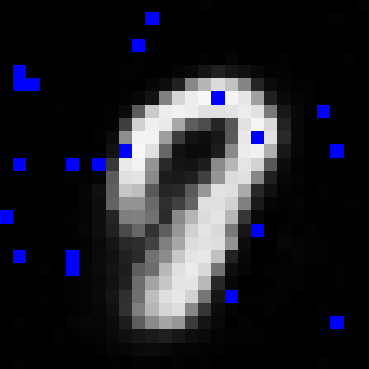}
        \end{subfigure}
        \hspace{-5pt}
        \begin{subfigure}[b]{0.17\linewidth}
            \centering
            \includegraphics[width=\linewidth]{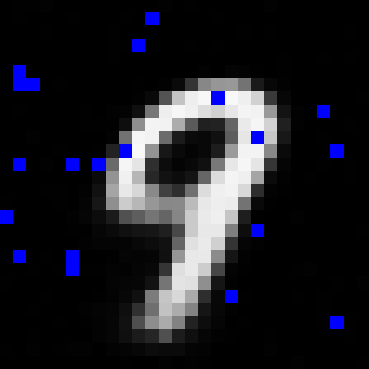}
        \end{subfigure}

        \centering
        \begin{subfigure}[b]{0.17\linewidth}
            \centering
            \includegraphics[width=\linewidth]{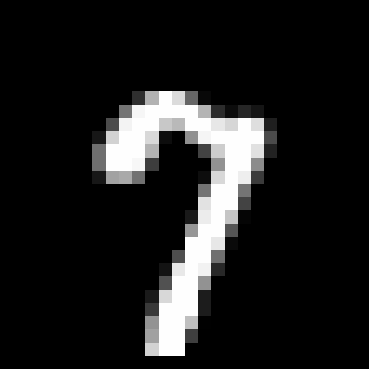}
             \caption{Image}
             \label{fig:image-a}
        \end{subfigure}
        \hspace{-5pt}
        \begin{subfigure}[b]{0.17\linewidth}
            \centering
            \includegraphics[width=\linewidth]{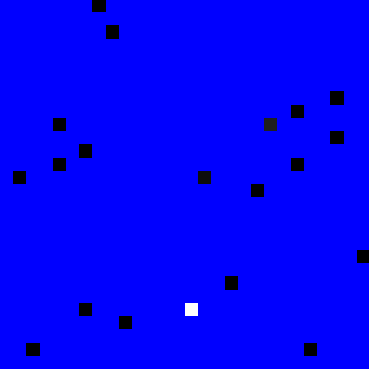}
           \caption{$\data_N$}
            \label{fig:context}
        \end{subfigure}
        \hspace{-5pt}
        \begin{subfigure}[b]{0.17\linewidth}
            \centering
            \includegraphics[width=\linewidth]{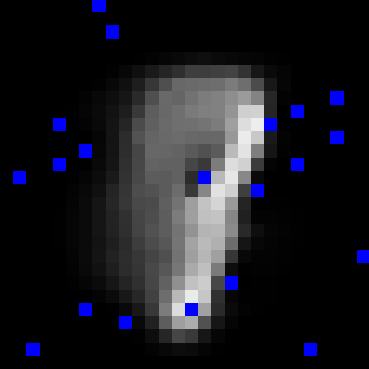}
            \caption{TNPD}
            \label{fig:tnpd}
        \end{subfigure}
        \hspace{-5pt}
        \begin{subfigure}[b]{0.17\linewidth}
            \centering
            \includegraphics[width=\linewidth]{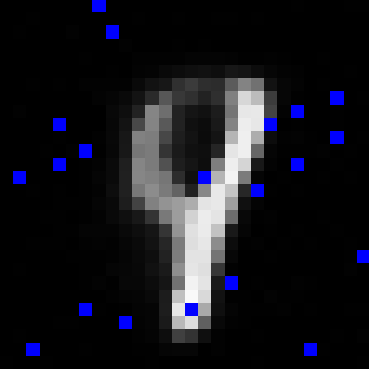}
            \caption{ACE}
            \label{fig:ACE_image}
        \end{subfigure}
        \hspace{-5pt}
        \begin{subfigure}[b]{0.17\linewidth}
            \centering
            \includegraphics[width=\linewidth]{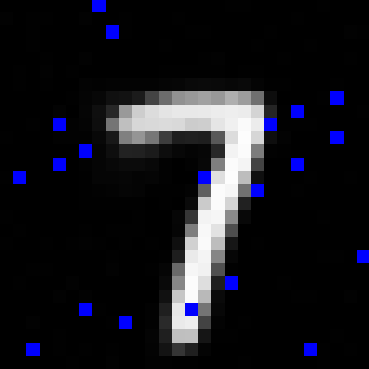}
            \caption{ACE-$\vtheta$}
            \label{fig:ACE_theta}
        \end{subfigure} 
    \end{subfigure}
    \begin{subfigure}[b]{0.40\textwidth}
        \centering
        \setlength{\figurewidth}{0.9\columnwidth}
        \setlength{\figureheight}{.56\columnwidth}
        \input{figures/mnist_nlpd}  
        \caption{NLPD v Context($\%$ of image)}
        \label{fig:celeba-nlpd}
    \end{subfigure}
    \caption{\textbf{Image regression (MNIST).} Image (\subref{fig:image-a}) serves as the reference for the problem, while (\subref{fig:context}) is the context where $10\%$ of the pixels are observed. Figures (\subref{fig:tnpd}) - (\subref{fig:ACE_theta}) are the respective model predictions, while (\subref{fig:celeba-nlpd}) shows performance over varying context (mean and 95\% confidence interval). In (\subref{fig:ACE_theta}) the model is also conditioned on the class label, showing a clear improvement in performance.}
    \label{fig:whole_image}
\end{figure}

%% file: figures/mnist_nlpd.tex
\begin{tikzpicture}

\definecolor{color0}{rgb}{0.12156862745098,0.466666666666667,0.705882352941177}
\definecolor{color1}{rgb}{1,0.498039215686275,0.0549019607843137}
\definecolor{color2}{rgb}{0.172549019607843,0.627450980392157,0.172549019607843}

\begin{axis}[
height=\figureheight,
legend cell align={left},
legend style={fill opacity=0.8, draw opacity=1, text opacity=1, draw=white!80!black},
tick align=outside,
tick pos=left,
width=\figurewidth,
x grid style={white!69.0196078431373!black},
xmin=0, xmax=26,
xtick style={color=black},
y grid style={white!69.0196078431373!black},
ymin=-1.3, ymax=-0.5,
ytick style={color=black}
]
\path [fill=color0, fill opacity=0.2]
(axis cs:2,-0.8878855509212)
--(axis cs:2,-0.978341301018955)
--(axis cs:3,-1.03794191599431)
--(axis cs:5,-1.07214268602077)
--(axis cs:6,-1.09174288360729)
--(axis cs:8,-1.16723618085981)
--(axis cs:12,-1.18785113054544)
--(axis cs:19,-1.24893724982912)
--(axis cs:25,-1.2565231131798)
--(axis cs:25,-1.21431831843794)
--(axis cs:25,-1.21431831843794)
--(axis cs:19,-1.17234718735044)
--(axis cs:12,-1.12879346650809)
--(axis cs:8,-1.0590874952495)
--(axis cs:6,-1.01079672725544)
--(axis cs:5,-0.97977918230351)
--(axis cs:3,-0.943970279688218)
--(axis cs:2,-0.8878855509212)
--cycle;

\path [fill=color1, fill opacity=0.2]
(axis cs:2,-0.945794636037744)
--(axis cs:2,-1.03490721580704)
--(axis cs:3,-1.08532179338664)
--(axis cs:5,-1.10797459370428)
--(axis cs:6,-1.12335887553741)
--(axis cs:8,-1.18946240801156)
--(axis cs:12,-1.20196350200702)
--(axis cs:19,-1.2549914013192)
--(axis cs:25,-1.26224162746303)
--(axis cs:25,-1.2179372532285)
--(axis cs:25,-1.2179372532285)
--(axis cs:19,-1.17978981625588)
--(axis cs:12,-1.15074159519146)
--(axis cs:8,-1.09273966890037)
--(axis cs:6,-1.05952298042725)
--(axis cs:5,-1.03670862429804)
--(axis cs:3,-1.00173143403798)
--(axis cs:2,-0.945794636037744)
--cycle;

\path [fill=color2, fill opacity=0.2]
(axis cs:2,-0.551278770183649)
--(axis cs:2,-0.757901763702307)
--(axis cs:3,-0.819334220967936)
--(axis cs:5,-0.899780629228454)
--(axis cs:6,-0.910900909845268)
--(axis cs:8,-1.08063233646004)
--(axis cs:12,-1.08945337217531)
--(axis cs:19,-1.20609025164615)
--(axis cs:25,-1.21446082877417)
--(axis cs:25,-1.15377475929956)
--(axis cs:25,-1.15377475929956)
--(axis cs:19,-1.07514398888577)
--(axis cs:12,-1.00284618455687)
--(axis cs:8,-0.862847320713114)
--(axis cs:6,-0.790120606477822)
--(axis cs:5,-0.727600290228028)
--(axis cs:3,-0.643813013948751)
--(axis cs:2,-0.551278770183649)
--cycle;

\addplot [semithick, color0, mark=*, mark size=1.5, mark options={solid}]
table {%
2 -0.933113425970078
3 -0.990956097841263
5 -1.02596093416214
6 -1.05126980543137
8 -1.11316183805466
12 -1.15832229852676
19 -1.21064221858978
25 -1.23542071580887
};
\addlegendentry{\tiny{ACE}}
\addplot [semithick, color1, mark=*, mark size=1.5, mark options={solid}]
table {%
2 -0.990350925922394
3 -1.04352661371231
5 -1.07234160900116
6 -1.09144092798233
8 -1.14110103845596
12 -1.17635254859924
19 -1.21739060878754
25 -1.24008944034576
};
\addlegendentry{\tiny{ACE-$\vtheta$}}
\addplot [semithick, color2, mark=*, mark size=1.5, mark options={solid}]
table {%
2 -0.654590266942978
3 -0.731573617458343
5 -0.813690459728241
6 -0.850510758161545
8 -0.971739828586578
12 -1.04614977836609
19 -1.14061712026596
25 -1.18411779403687
};
\addlegendentry{\tiny{TNPD}}
\end{axis}

\end{tikzpicture}

%% file: figures/mnist_acc.tex
\begin{tikzpicture}

\definecolor{color0}{rgb}{0.12156862745098,0.466666666666667,0.705882352941177}

\begin{axis}[
height=\figureheight,
tick align=outside,
tick pos=left,
width=\figurewidth,
x grid style={white!69.0196078431373!black},
xlabel={Context Size \%},
xmajorgrids,
xmin=-3.95, xmax=104.95,
xtick style={color=black},
y grid style={white!69.0196078431373!black},
ylabel={Classification Accuracy},
ymajorgrids,
ymin=0, ymax=1,
ytick style={color=black}
]
\path [fill=color0, fill opacity=0.2]
(axis cs:1,0.3)
--(axis cs:1,0.1625)
--(axis cs:2,0.2875)
--(axis cs:3,0.4375)
--(axis cs:6,0.725)
--(axis cs:8,0.7375)
--(axis cs:12,0.775)
--(axis cs:19,0.875)
--(axis cs:25,0.9375)
--(axis cs:31,0.95)
--(axis cs:38,0.9125)
--(axis cs:63,0.9625)
--(axis cs:76,1)
--(axis cs:100,0.925)
--(axis cs:100,1)
--(axis cs:100,1)
--(axis cs:76,1)
--(axis cs:63,1)
--(axis cs:38,1)
--(axis cs:31,1)
--(axis cs:25,0.9875)
--(axis cs:19,0.9625)
--(axis cs:12,0.875)
--(axis cs:8,0.8625)
--(axis cs:6,0.8)
--(axis cs:3,0.5625)
--(axis cs:2,0.4375)
--(axis cs:1,0.3)
--cycle;

\addplot [semithick, color0, mark=*, mark size=1.5, mark options={solid}]
table {%
1 0.225
2 0.3625
3 0.4875
6 0.7625
8 0.8
12 0.825
19 0.9125
25 0.9625
31 0.975
38 0.9625
63 0.9875
76 1
100 0.975
};
\end{axis}

\end{tikzpicture}

%% file: image_files/output_plot_1.tex
\begin{tikzpicture}

\definecolor{color0}{rgb}{0.12156862745098,0.466666666666667,0.705882352941177}

\begin{axis}[
  clip=false,  
  height=\figureheight,
  legend cell align={left},
  legend style={fill opacity=0.8, draw opacity=1, text opacity=1, draw=white!80!black},
  tick align=outside,
  tick pos=left,
  width=\figurewidth,
  x grid style={white!69.0196078431373!black},
  xmin=-0.52, xmax=0.52,
  xtick={-0.5, -0.25, 0, 0.25, 0.5},  
  xticklabels={0,0.25, 0.5, 0.75, 1.0},  
  y grid style={white!69.0196078431373!black},
  ymin=-0.725, ymax=9.725,
  ytick style={color=black},
  ytick={0,1,2,3,4,5,6,7,8,9},
  yticklabels={No\_Beard,Young,Bangs,Male,Wearing\_Necktie,Big\_Lips,Black\_Hair,Smiling,Gray\_Hair,Bald}
]
\draw[draw=none,fill=color0] (axis cs:0,-0.25) rectangle (axis cs:0.269394755363464,0.25);
\draw[draw=none,fill=color0] (axis cs:0,0.75) rectangle (axis cs:0.209146857261658,1.25);
\draw[draw=none,fill=color0] (axis cs:0,1.75) rectangle (axis cs:0.0127872228622437,2.25);
\draw[draw=none,fill=color0] (axis cs:0,2.75) rectangle (axis cs:-0.0484082698822021,3.25);
\draw[draw=none,fill=color0] (axis cs:0,3.75) rectangle (axis cs:-0.0492472648620605,4.25);
\draw[draw=none,fill=color0] (axis cs:0,4.75) rectangle (axis cs:-0.203940004110336,5.25);
\draw[draw=none,fill=color0] (axis cs:0,5.75) rectangle (axis cs:-0.329178512096405,6.25);
\draw[draw=none,fill=color0] (axis cs:0,6.75) rectangle (axis cs:-0.372095733880997,7.25);
\draw[draw=none,fill=color0] (axis cs:0,7.75) rectangle (axis cs:-0.424515068531036,8.25);
\draw[draw=none,fill=color0] (axis cs:0,8.75) rectangle (axis cs:-0.457507848739624,9.25);
\addplot [red, dashed, forget plot]
table {%
-1.11022302462516e-16 -0.725
-1.11022302462516e-16 9.725
};
\addplot [semithick, red, mark=asterisk, mark size=1.5, mark options={solid}, only marks]
table {%
0.5 0
};
\addlegendentry{Label = 1}
\addplot [semithick, red, mark=asterisk, mark size=1.5, mark options={solid}, only marks, forget plot]
table {%
0.5 1
};
\addplot [semithick, black, mark=x, mark size=1.5, mark options={solid}, only marks]
table {%
-0.5 2
};
\addlegendentry{Label = 0}
\addplot [semithick, red, mark=asterisk, mark size=1.5, mark options={solid}, only marks, forget plot]
table {%
0.5 3
};
\addplot [semithick, black, mark=x, mark size=1.5, mark options={solid}, only marks, forget plot]
table {%
-0.5 4
};
\addplot [semithick, black, mark=x, mark size=1.5, mark options={solid}, only marks, forget plot]
table {%
-0.5 5
};
\addplot [semithick, black, mark=x, mark size=1.5, mark options={solid}, only marks, forget plot]
table {%
-0.5 6
};
\addplot [semithick, black, mark=x, mark size=1.5, mark options={solid}, only marks, forget plot]
table {%
-0.5 7
};
\addplot [semithick, black, mark=x, mark size=1.5, mark options={solid}, only marks, forget plot]
table {%
-0.5 8
};
\addplot [semithick, black, mark=x, mark size=1.5, mark options={solid}, only marks, forget plot]
table {%
-0.5 9
};
\addplot [semithick, black, mark=*, mark size=1.5, mark options={solid}, only marks]
table {%
0.334939955281122 0
};
\addlegendentry{Average}
\addplot [semithick, black, mark=*, mark size=1.5, mark options={solid}, only marks, forget plot]
table {%
0.273616849046639 1
};
\addplot [semithick, black, mark=*, mark size=1.5, mark options={solid}, only marks, forget plot]
table {%
-0.348424720753804 2
};
\addplot [semithick, black, mark=*, mark size=1.5, mark options={solid}, only marks, forget plot]
table {%
-0.083245721844629 3
};
\addplot [semithick, black, mark=*, mark size=1.5, mark options={solid}, only marks, forget plot]
table {%
-0.4272849323047 4
};
\addplot [semithick, black, mark=*, mark size=1.5, mark options={solid}, only marks, forget plot]
table {%
-0.259204142172469 5
};
\addplot [semithick, black, mark=*, mark size=1.5, mark options={solid}, only marks, forget plot]
table {%
-0.260749065888775 6
};
\addplot [semithick, black, mark=*, mark size=1.5, mark options={solid}, only marks, forget plot]
table {%
-0.0179196343516009 7
};
\addplot [semithick, black, mark=*, mark size=1.5, mark options={solid}, only marks, forget plot]
table {%
-0.458050138450831 8
};
\addplot [semithick, black, mark=*, mark size=1.5, mark options={solid}, only marks, forget plot]
table {%
-0.477556651316147 9
};

\end{axis}

\end{tikzpicture}

%% file: image_files/output_plot_1_.tex
\begin{tikzpicture}

\definecolor{color0}{rgb}{0.12156862745098,0.466666666666667,0.705882352941177}

\begin{axis}[
height=\figureheight,
legend cell align={left},
legend style={fill opacity=0.8, draw opacity=1, text opacity=1, draw=white!80!black},
scaled x ticks=manual:{}{\pgfmathparse{#1}},
tick align=outside,
tick pos=left,
width=\figurewidth,
x grid style={white!69.0196078431373!black},
xmin=-0.52, xmax=0.52,
xtick style={color=black},
xtick={-0.5, -0.25, 0, 0.25, 0.5}, 
xticklabels={0,0.25, 0.5, 0.75, 1.0},  
y grid style={white!69.0196078431373!black},
ymin=-0.725, ymax=9.725,
ytick style={color=black},
ytick={0,1,2,3,4,5,6,7,8,9},
yticklabels={Young,No\_Beard,Bangs,Big\_Lips,Smiling,Male,Wearing\_Necktie,Black\_Hair,Gray\_Hair,Bald}
]
\draw[draw=none,fill=color0] (axis cs:0,-0.25) rectangle (axis cs:0.320205390453339,0.25);
\draw[draw=none,fill=color0] (axis cs:0,0.75) rectangle (axis cs:0.272623658180237,1.25);
\draw[draw=none,fill=color0] (axis cs:0,1.75) rectangle (axis cs:0.0684733390808105,2.25);
\draw[draw=none,fill=color0] (axis cs:0,2.75) rectangle (axis cs:-0.103595376014709,3.25);
\draw[draw=none,fill=color0] (axis cs:0,3.75) rectangle (axis cs:-0.161007344722748,4.25);
\draw[draw=none,fill=color0] (axis cs:0,4.75) rectangle (axis cs:-0.169925153255463,5.25);
\draw[draw=none,fill=color0] (axis cs:0,5.75) rectangle (axis cs:-0.227203577756882,6.25);
\draw[draw=none,fill=color0] (axis cs:0,6.75) rectangle (axis cs:-0.263009160757065,7.25);
\draw[draw=none,fill=color0] (axis cs:0,7.75) rectangle (axis cs:-0.442010283470154,8.25);
\draw[draw=none,fill=color0] (axis cs:0,8.75) rectangle (axis cs:-0.44317689538002,9.25);
\addplot [red, dashed, forget plot]
table {%
-1.11022302462516e-16 -0.725
-1.11022302462516e-16 9.725
};
\addplot [semithick, red, mark=asterisk, mark size=1.5, mark options={solid}, only marks]
table {%
0.5 0
};
\addlegendentry{Label = 1}
\addplot [semithick, red, mark=asterisk, mark size=1.5, mark options={solid}, only marks, forget plot]
table {%
0.5 1
};
\addplot [semithick, red, mark=asterisk, mark size=1.5, mark options={solid}, only marks, forget plot]
table {%
0.5 2
};
\addplot [semithick, black, mark=x, mark size=1.5, mark options={solid}, only marks]
table {%
-0.5 3
};
\addlegendentry{Label = 0}
\addplot [semithick, black, mark=x, mark size=1.5, mark options={solid}, only marks, forget plot]
table {%
-0.5 4
};
\addplot [semithick, black, mark=x, mark size=1.5, mark options={solid}, only marks, forget plot]
table {%
-0.5 5
};
\addplot [semithick, black, mark=x, mark size=1.5, mark options={solid}, only marks, forget plot]
table {%
-0.5 6
};
\addplot [semithick, black, mark=x, mark size=1.5, mark options={solid}, only marks, forget plot]
table {%
-0.5 7
};
\addplot [semithick, black, mark=x, mark size=1.5, mark options={solid}, only marks, forget plot]
table {%
-0.5 8
};
\addplot [semithick, black, mark=x, mark size=1.5, mark options={solid}, only marks, forget plot]
table {%
-0.5 9
};
\addplot [semithick, black, mark=*, mark size=1.5, mark options={solid}, only marks]
table {%
0.273616849046639 0
};
\addlegendentry{Average}
\addplot [semithick, black, mark=*, mark size=1.5, mark options={solid}, only marks, forget plot]
table {%
0.334939955281122 1
};
\addplot [semithick, black, mark=*, mark size=1.5, mark options={solid}, only marks, forget plot]
table {%
-0.348424720753804 2
};
\addplot [semithick, black, mark=*, mark size=1.5, mark options={solid}, only marks, forget plot]
table {%
-0.259204142172469 3
};
\addplot [semithick, black, mark=*, mark size=1.5, mark options={solid}, only marks, forget plot]
table {%
-0.0179196343516009 4
};
\addplot [semithick, black, mark=*, mark size=1.5, mark options={solid}, only marks, forget plot]
table {%
-0.083245721844629 5
};
\addplot [semithick, black, mark=*, mark size=1.5, mark options={solid}, only marks, forget plot]
table {%
-0.4272849323047 6
};
\addplot [semithick, black, mark=*, mark size=1.5, mark options={solid}, only marks, forget plot]
table {%
-0.260749065888775 7
};
\addplot [semithick, black, mark=*, mark size=1.5, mark options={solid}, only marks, forget plot]
table {%
-0.458050138450831 8
};
\addplot [semithick, black, mark=*, mark size=1.5, mark options={solid}, only marks, forget plot]
table {%
-0.477556651316147 9
};
\draw (axis cs:-0.5,-1.2) node[
  scale=0.5,
  anchor=south,
  text=black,
  rotate=0.0
]{0.0};
\draw (axis cs:-0.25,-1.2) node[
  scale=0.5,
  anchor=south,
  text=black,
  rotate=0.0
]{0.25};
\draw (axis cs:0,-1.2) node[
  scale=0.5,
  anchor=south,
  text=black,
  rotate=0.0
]{0.5};
\draw (axis cs:0.25,-1.2) node[
  scale=0.5,
  anchor=base,
  text=black,
  rotate=0.0
]{0.75};
\draw (axis cs:0.5,-1.2) node[
  scale=0.5,
  anchor=base,
  text=black,
  rotate=0.0
]{1.0};
\end{axis}

\end{tikzpicture}

%% file: bo_extra.tex

\begin{figure*}[!t]
  \centering
  \scriptsize
 
  \setlength{\figurewidth}{.19\textwidth}
  \setlength{\figureheight}{.75\figurewidth}

  \centerline{\legendACETS \quad \legendACEMES \quad \legendTNPDTS \quad \legendGPTS \quad \legendGPMES \quad \legendRandom}
  \vspace{0.2cm}

  \begin{minipage}[b]{\textwidth}  
    \centering
    \begin{subfigure}[b]{.25\textwidth}
      \centering
      \input{figures/bo_1d_ackley}
    \end{subfigure}
    \begin{subfigure}[b]{.23\textwidth}
      \centering
      \input{figures/bo_1d_negeasom}
    \end{subfigure}
    \begin{subfigure}[b]{.23\textwidth}
      \centering
      \input{figures/bo_2d_michalewicz2d}
    \end{subfigure}
    \begin{subfigure}[b]{.23\textwidth}
      \centering
      \input{figures/bo_2d_ackley}
    \end{subfigure}
  \end{minipage}  
  
  \begin{minipage}[b]{\textwidth}  
    \centering
    \begin{subfigure}[b]{.25\textwidth}
      \centering
      \input{figures/bo_3d_levy}
    \end{subfigure}
    \begin{subfigure}[b]{.23\textwidth}
      \centering
      \input{figures/bo_4d_hartmann4d}
    \end{subfigure}
    \begin{subfigure}[b]{.23\textwidth}
      \centering
      \input{figures/bo_5d_griewank5d}
    \end{subfigure}
    \begin{subfigure}[b]{.23\textwidth}
      \centering
      \input{figures/bo_6d_griewank6d}
    \end{subfigure}
  \end{minipage}  
  \vspace{-0.5cm}
  \caption{\textbf{Bayesian optimization additional results.} Regret comparison (mean ± standard error) on extended BO benchmark results on distinct test functions.}
  \label{fig:bo_comparisons_extra}
  \vspace{-0.3cm}
\end{figure*}

%% file: bop_weak.tex

\begin{figure*}[!t]
  \centering
  \scriptsize
 
  \setlength{\figurewidth}{.21\textwidth}
  \setlength{\figureheight}{.75\figurewidth}
  \centerline{\legendACETS \quad \legendACEPTS \quad \legendGPTS \quad \legendpiBOTS}
  
  \vspace{0.2cm}
  \begin{minipage}[b]{\textwidth}  
    \centering
    \begin{subfigure}[b]{.28\textwidth}
      \centering
      \input{figures/bop_1d_ackley1d_weak}
    \end{subfigure}
    \begin{subfigure}[b]{.25\textwidth}
      \centering
      \input{figures/bop_1d_gramacylee_weak}
    \end{subfigure}
    \begin{subfigure}[b]{.25\textwidth}
      \centering
      \input{figures/bop_1d_negeasom_weak}
    \end{subfigure}
  \end{minipage}  

  \begin{minipage}[b]{\textwidth}  
    \centering
    \begin{subfigure}[b]{.28\textwidth}
      \centering
      \input{figures/bop_2d_braninscaled_weak}
    \end{subfigure}
    \begin{subfigure}[b]{.25\textwidth}
      \centering
      \input{figures/bop_2d_ackley_weak}
    \end{subfigure}
    \begin{subfigure}[b]{.25\textwidth}
      \centering
      \input{figures/bop_3d_hartmann3d_weak}
    \end{subfigure}
  \end{minipage}  
  \vspace{-0.5cm}
  \caption{\textbf{Bayesian optimization with weak prior.} Simple regret (mean ± standard error). Prior injection can improve the performance of ACE, making it perform competitively compared to $\pi$BO-TS.}
  \label{fig:bo_comparisons_weak}
  \vspace{-0.3cm}
\end{figure*}

%% file: figures/bop_1d_ackley1d_weak.tex
\begin{tikzpicture}

\definecolor{color0}{rgb}{0,0,1}
\definecolor{color1}{rgb}{1,0.549019607843137,0}
\definecolor{color2}{rgb}{1,0.647058823529412,0}
\definecolor{color3}{rgb}{0.564705882352941,0.933333333333333,0.564705882352941}

\begin{axis}[axis on top,
enlarge x limits=false,
enlarge y limits=false,
height=\figureheight,
scale only axis,
tick align=outside,
tick pos=left,
tick pos=left,
width=\figurewidth,
xmin=3, xmax=25,
xtick style={color=black},
xtick={-10,0,10,25,50,75,100},
xticklabels={\ensuremath{-}10,0,10,25,50,75,90},
ylabel={Regret},
ymin=-0.2, ymax=4.2,
ytick style={color=black},
ytick={0.   , 4.2},
]
\node[anchor=north east] at (rel axis cs:1,1) {Ackley 1D (weak)};
\path [draw=color1, fill=color1, opacity=0.3]
(axis cs:3,4.02692893971875)
--(axis cs:3,3.24591423044726)
--(axis cs:4,1.77067730890269)
--(axis cs:5,1.76412949836681)
--(axis cs:6,0.901900046752295)
--(axis cs:7,0.641698695153097)
--(axis cs:8,0.309029295790925)
--(axis cs:9,0.194494641339335)
--(axis cs:10,0.186929311105284)
--(axis cs:11,0.153295366726285)
--(axis cs:12,0.117130087915778)
--(axis cs:13,0.102951630894041)
--(axis cs:14,0.102463431888656)
--(axis cs:15,0.0884310277031863)
--(axis cs:16,0.0836861772565924)
--(axis cs:17,0.0794272775879709)
--(axis cs:18,0.0792284572551402)
--(axis cs:19,0.0759855897844494)
--(axis cs:20,0.0759636236866293)
--(axis cs:21,0.0754958686864695)
--(axis cs:22,0.0754958686864695)
--(axis cs:23,0.0738244784337902)
--(axis cs:24,0.0738244784337902)
--(axis cs:25,0.073040267409242)
--(axis cs:26,0.073040267409242)
--(axis cs:27,0.0729969727563917)
--(axis cs:28,0.0726331614431581)
--(axis cs:29,0.0726087055531572)
--(axis cs:30,0.0725385389036556)
--(axis cs:31,0.0725385389036556)
--(axis cs:32,0.072266874337939)
--(axis cs:33,0.0720332077871527)
--(axis cs:34,0.0720332077871527)
--(axis cs:35,0.0719413585829494)
--(axis cs:36,0.0719413585829494)
--(axis cs:37,0.0719413585829494)
--(axis cs:38,0.0715913469944206)
--(axis cs:39,0.0715913469944206)
--(axis cs:40,0.0715913469944206)
--(axis cs:41,0.071250765419455)
--(axis cs:42,0.071250765419455)
--(axis cs:43,0.071250765419455)
--(axis cs:44,0.071250765419455)
--(axis cs:45,0.071250765419455)
--(axis cs:46,0.071250765419455)
--(axis cs:47,0.071250765419455)
--(axis cs:48,0.071250765419455)
--(axis cs:49,0.071250765419455)
--(axis cs:50,0.071250765419455)
--(axis cs:51,0.0710422301800616)
--(axis cs:52,0.0710422301800616)
--(axis cs:53,0.0710422301800616)
--(axis cs:54,0.0710422301800616)
--(axis cs:55,0.0710422301800616)
--(axis cs:56,0.0710422301800616)
--(axis cs:57,0.0710422301800616)
--(axis cs:58,0.0705702930474379)
--(axis cs:59,0.0703763333555568)
--(axis cs:60,0.0703763333555568)
--(axis cs:61,0.0703763333555568)
--(axis cs:62,0.0703763333555568)
--(axis cs:63,0.0703763333555568)
--(axis cs:64,0.0703763333555568)
--(axis cs:65,0.0703763333555568)
--(axis cs:66,0.0703763333555568)
--(axis cs:67,0.0703763333555568)
--(axis cs:68,0.0703763333555568)
--(axis cs:69,0.0703763333555568)
--(axis cs:70,0.0703763333555568)
--(axis cs:71,0.07029700821286)
--(axis cs:72,0.07029700821286)
--(axis cs:73,0.07029700821286)
--(axis cs:74,0.07029700821286)
--(axis cs:75,0.07029700821286)
--(axis cs:76,0.07029700821286)
--(axis cs:77,0.07029700821286)
--(axis cs:78,0.07029700821286)
--(axis cs:79,0.07029700821286)
--(axis cs:80,0.07029700821286)
--(axis cs:81,0.07029700821286)
--(axis cs:82,0.07029700821286)
--(axis cs:83,0.07029700821286)
--(axis cs:84,0.07029700821286)
--(axis cs:85,0.07029700821286)
--(axis cs:86,0.07029700821286)
--(axis cs:87,0.07029700821286)
--(axis cs:88,0.07029700821286)
--(axis cs:89,0.0702924188641426)
--(axis cs:90,0.0700734620720922)
--(axis cs:91,0.0699415283758774)
--(axis cs:92,0.0699415283758774)
--(axis cs:93,0.0698574663780539)
--(axis cs:94,0.0698574663780539)
--(axis cs:95,0.0698574663780539)
--(axis cs:96,0.0698574663780539)
--(axis cs:97,0.0698574663780539)
--(axis cs:98,0.0698574663780539)
--(axis cs:99,0.0698574663780539)
--(axis cs:100,0.0698574663780539)
--(axis cs:101,0.0698574663780539)
--(axis cs:102,0.0698574663780539)
--(axis cs:103,0.0698574663780539)
--(axis cs:104,0.0698574663780539)
--(axis cs:104,0.0700749656229224)
--(axis cs:104,0.0700749656229224)
--(axis cs:103,0.0700749656229224)
--(axis cs:102,0.0700749656229224)
--(axis cs:101,0.0700749656229224)
--(axis cs:100,0.0700749656229224)
--(axis cs:99,0.0700749656229224)
--(axis cs:98,0.0700749656229224)
--(axis cs:97,0.0700749656229224)
--(axis cs:96,0.0700749656229224)
--(axis cs:95,0.0700749656229224)
--(axis cs:94,0.0700749656229224)
--(axis cs:93,0.0700749656229224)
--(axis cs:92,0.0703106131757006)
--(axis cs:91,0.0703106131757006)
--(axis cs:90,0.0705909586544736)
--(axis cs:89,0.0709477242512671)
--(axis cs:88,0.0709522769484284)
--(axis cs:87,0.0709522769484284)
--(axis cs:86,0.0709522769484284)
--(axis cs:85,0.0709522769484284)
--(axis cs:84,0.0709522769484284)
--(axis cs:83,0.0709522769484284)
--(axis cs:82,0.0709522769484284)
--(axis cs:81,0.0709522769484284)
--(axis cs:80,0.0709522769484284)
--(axis cs:79,0.0709522769484284)
--(axis cs:78,0.0709522769484284)
--(axis cs:77,0.0709522769484284)
--(axis cs:76,0.0709522769484284)
--(axis cs:75,0.0709522769484284)
--(axis cs:74,0.0709522769484284)
--(axis cs:73,0.0709522769484284)
--(axis cs:72,0.0709522769484284)
--(axis cs:71,0.0709522769484284)
--(axis cs:70,0.0710235632501925)
--(axis cs:69,0.0710235632501925)
--(axis cs:68,0.0710235632501925)
--(axis cs:67,0.0710235632501925)
--(axis cs:66,0.0710235632501925)
--(axis cs:65,0.0710235632501925)
--(axis cs:64,0.0710235632501925)
--(axis cs:63,0.0710235632501925)
--(axis cs:62,0.0710235632501925)
--(axis cs:61,0.0710235632501925)
--(axis cs:60,0.0710235632501925)
--(axis cs:59,0.0710235632501925)
--(axis cs:58,0.0713556242889386)
--(axis cs:57,0.0728543458326262)
--(axis cs:56,0.0728543458326262)
--(axis cs:55,0.0728543458326262)
--(axis cs:54,0.0728543458326262)
--(axis cs:53,0.0728543458326262)
--(axis cs:52,0.0728543458326262)
--(axis cs:51,0.0728543458326262)
--(axis cs:50,0.0730422000738874)
--(axis cs:49,0.0730422000738874)
--(axis cs:48,0.0730422000738874)
--(axis cs:47,0.0730422000738874)
--(axis cs:46,0.0730422000738874)
--(axis cs:45,0.0730422000738874)
--(axis cs:44,0.0730422000738874)
--(axis cs:43,0.0730422000738874)
--(axis cs:42,0.0730422000738874)
--(axis cs:41,0.0730422000738874)
--(axis cs:40,0.0736839122666617)
--(axis cs:39,0.0736839122666617)
--(axis cs:38,0.0736839122666617)
--(axis cs:37,0.0740724064486756)
--(axis cs:36,0.0740724064486756)
--(axis cs:35,0.0740724064486756)
--(axis cs:34,0.074377830362279)
--(axis cs:33,0.074377830362279)
--(axis cs:32,0.0762184755576035)
--(axis cs:31,0.0764579682362075)
--(axis cs:30,0.0764579682362075)
--(axis cs:29,0.0765310391320598)
--(axis cs:28,0.0765517198056821)
--(axis cs:27,0.0774099719189021)
--(axis cs:26,0.0774539548532842)
--(axis cs:25,0.0774539548532842)
--(axis cs:24,0.0780652202201908)
--(axis cs:23,0.0780652202201908)
--(axis cs:22,0.0821084926925505)
--(axis cs:21,0.0821084926925505)
--(axis cs:20,0.0825255235654226)
--(axis cs:19,0.0825480800089228)
--(axis cs:18,0.0871477920021379)
--(axis cs:17,0.0872414194546463)
--(axis cs:16,0.110691559004632)
--(axis cs:15,0.11558328936136)
--(axis cs:14,0.17025623602934)
--(axis cs:13,0.170687677481004)
--(axis cs:12,0.458662915632257)
--(axis cs:11,0.560436696361466)
--(axis cs:10,0.590209039145273)
--(axis cs:9,0.692185294023852)
--(axis cs:8,1.24872759713536)
--(axis cs:7,1.63116751887993)
--(axis cs:6,1.90867522471984)
--(axis cs:5,3.09060343860587)
--(axis cs:4,3.1070994237873)
--(axis cs:3,4.02692893971875)
--cycle;

\path [draw=color1, fill=color1, opacity=0.3]
(axis cs:3,4.02692893971875)
--(axis cs:3,3.24591423044726)
--(axis cs:4,2.32155921710791)
--(axis cs:5,1.33350408850246)
--(axis cs:6,1.27879100957968)
--(axis cs:7,0.917086138942526)
--(axis cs:8,0.695592374604861)
--(axis cs:9,0.540239827618069)
--(axis cs:10,0.438920275713534)
--(axis cs:11,0.412677384800578)
--(axis cs:12,0.382640144867949)
--(axis cs:13,0.321849876922532)
--(axis cs:14,0.317382686185337)
--(axis cs:15,0.282331675779895)
--(axis cs:16,0.240382488521631)
--(axis cs:17,0.148264472838925)
--(axis cs:18,0.148264472838925)
--(axis cs:19,0.117767655387537)
--(axis cs:20,0.0853902752417036)
--(axis cs:21,0.0736782696444838)
--(axis cs:22,0.0730477559932568)
--(axis cs:23,0.0730057741395088)
--(axis cs:24,0.0714475272876691)
--(axis cs:25,0.0712015510947565)
--(axis cs:26,0.0707607589168566)
--(axis cs:27,0.0707607589168566)
--(axis cs:28,0.0701746050509503)
--(axis cs:29,0.0701134684908355)
--(axis cs:30,0.0701134684908355)
--(axis cs:31,0.0701134684908355)
--(axis cs:32,0.0701134684908355)
--(axis cs:33,0.0701134684908355)
--(axis cs:34,0.0701134684908355)
--(axis cs:35,0.0700936773253813)
--(axis cs:36,0.0700936773253813)
--(axis cs:37,0.0700936773253813)
--(axis cs:38,0.0700743439217147)
--(axis cs:39,0.0700743439217147)
--(axis cs:40,0.0699951731956626)
--(axis cs:41,0.0699739298393459)
--(axis cs:42,0.0699739298393459)
--(axis cs:43,0.0699096021106847)
--(axis cs:44,0.0699096021106847)
--(axis cs:45,0.0697696607451501)
--(axis cs:46,0.0697696607451501)
--(axis cs:47,0.0697465211882078)
--(axis cs:48,0.0696823450235266)
--(axis cs:49,0.0696823450235266)
--(axis cs:50,0.0696823450235266)
--(axis cs:51,0.0696823450235266)
--(axis cs:52,0.0696823450235266)
--(axis cs:53,0.0696823450235266)
--(axis cs:54,0.0696823450235266)
--(axis cs:55,0.0696823450235266)
--(axis cs:56,0.0696795212977115)
--(axis cs:57,0.0696795212977115)
--(axis cs:58,0.0696795212977115)
--(axis cs:59,0.0696795212977115)
--(axis cs:60,0.0696735669474826)
--(axis cs:61,0.0696735669474826)
--(axis cs:62,0.0696735669474826)
--(axis cs:63,0.0696735669474826)
--(axis cs:64,0.0696551725564278)
--(axis cs:65,0.0696551725564278)
--(axis cs:66,0.0696551725564278)
--(axis cs:67,0.0696551725564278)
--(axis cs:68,0.0696551725564278)
--(axis cs:69,0.0696551725564278)
--(axis cs:70,0.0696551725564278)
--(axis cs:71,0.0696551725564278)
--(axis cs:72,0.0696551725564278)
--(axis cs:73,0.0696551725564278)
--(axis cs:74,0.0696551725564278)
--(axis cs:75,0.0696551725564278)
--(axis cs:76,0.0696551725564278)
--(axis cs:77,0.0696551725564278)
--(axis cs:78,0.0696551725564278)
--(axis cs:79,0.0696551725564278)
--(axis cs:80,0.0696551725564278)
--(axis cs:81,0.0696551725564278)
--(axis cs:82,0.0696551725564278)
--(axis cs:83,0.0696551725564278)
--(axis cs:84,0.0696551725564278)
--(axis cs:85,0.0696551725564278)
--(axis cs:86,0.0696551725564278)
--(axis cs:87,0.0696551725564278)
--(axis cs:88,0.0696551725564278)
--(axis cs:89,0.0696551725564278)
--(axis cs:90,0.0696551725564278)
--(axis cs:91,0.0696551725564278)
--(axis cs:92,0.0696425099663787)
--(axis cs:93,0.0696425099663787)
--(axis cs:94,0.0696425099663787)
--(axis cs:95,0.0696425099663787)
--(axis cs:96,0.069641410195386)
--(axis cs:97,0.069641410195386)
--(axis cs:98,0.069641410195386)
--(axis cs:99,0.069641410195386)
--(axis cs:100,0.0696274496209619)
--(axis cs:101,0.0696274496209619)
--(axis cs:102,0.0696274496209619)
--(axis cs:103,0.0696239984926955)
--(axis cs:104,0.0696239984926955)
--(axis cs:104,0.0696556234378365)
--(axis cs:104,0.0696556234378365)
--(axis cs:103,0.0696556234378365)
--(axis cs:102,0.0696612803147752)
--(axis cs:101,0.0696612803147752)
--(axis cs:100,0.0696612803147752)
--(axis cs:99,0.0697320500769526)
--(axis cs:98,0.0697320500769526)
--(axis cs:97,0.0697320500769526)
--(axis cs:96,0.0697320500769526)
--(axis cs:95,0.0697762437440281)
--(axis cs:94,0.0697762437440281)
--(axis cs:93,0.0697762437440281)
--(axis cs:92,0.0697762437440281)
--(axis cs:91,0.0697868557489181)
--(axis cs:90,0.0697868557489181)
--(axis cs:89,0.0697868557489181)
--(axis cs:88,0.0697868557489181)
--(axis cs:87,0.0697868557489181)
--(axis cs:86,0.0697868557489181)
--(axis cs:85,0.0697868557489181)
--(axis cs:84,0.0697868557489181)
--(axis cs:83,0.0697868557489181)
--(axis cs:82,0.0697868557489181)
--(axis cs:81,0.0697868557489181)
--(axis cs:80,0.0697868557489181)
--(axis cs:79,0.0697868557489181)
--(axis cs:78,0.0697868557489181)
--(axis cs:77,0.0697868557489181)
--(axis cs:76,0.0697868557489181)
--(axis cs:75,0.0697868557489181)
--(axis cs:74,0.0697868557489181)
--(axis cs:73,0.0697868557489181)
--(axis cs:72,0.0697868557489181)
--(axis cs:71,0.0697868557489181)
--(axis cs:70,0.0697868557489181)
--(axis cs:69,0.0697868557489181)
--(axis cs:68,0.0697868557489181)
--(axis cs:67,0.0697868557489181)
--(axis cs:66,0.0697868557489181)
--(axis cs:65,0.0697868557489181)
--(axis cs:64,0.0697868557489181)
--(axis cs:63,0.0698092972400554)
--(axis cs:62,0.0698092972400554)
--(axis cs:61,0.0698092972400554)
--(axis cs:60,0.0698092972400554)
--(axis cs:59,0.0698131229788801)
--(axis cs:58,0.0698131229788801)
--(axis cs:57,0.0698131229788801)
--(axis cs:56,0.0698131229788801)
--(axis cs:55,0.0698146827776924)
--(axis cs:54,0.0698146827776924)
--(axis cs:53,0.0698146827776924)
--(axis cs:52,0.0698146827776924)
--(axis cs:51,0.0698146827776924)
--(axis cs:50,0.0698146827776924)
--(axis cs:49,0.0698146827776924)
--(axis cs:48,0.0698146827776924)
--(axis cs:47,0.0699207155139916)
--(axis cs:46,0.0699695647023139)
--(axis cs:45,0.0699695647023139)
--(axis cs:44,0.0703280193548176)
--(axis cs:43,0.0703280193548176)
--(axis cs:42,0.0704497083512155)
--(axis cs:41,0.0704497083512155)
--(axis cs:40,0.0704628791882193)
--(axis cs:39,0.0705627603365646)
--(axis cs:38,0.0705627603365646)
--(axis cs:37,0.0705840296066886)
--(axis cs:36,0.0705840296066886)
--(axis cs:35,0.0705840296066886)
--(axis cs:34,0.070597328867546)
--(axis cs:33,0.070597328867546)
--(axis cs:32,0.070597328867546)
--(axis cs:31,0.070597328867546)
--(axis cs:30,0.070597328867546)
--(axis cs:29,0.070597328867546)
--(axis cs:28,0.0711614197507179)
--(axis cs:27,0.0740163539587488)
--(axis cs:26,0.0740163539587488)
--(axis cs:25,0.0755180088441992)
--(axis cs:24,0.0757107295618235)
--(axis cs:23,0.0792701864068554)
--(axis cs:22,0.0793002380857388)
--(axis cs:21,0.0972156384678027)
--(axis cs:20,0.310261020297767)
--(axis cs:19,0.59023949137341)
--(axis cs:18,0.631366491243948)
--(axis cs:17,0.631366491243948)
--(axis cs:16,0.913157237349568)
--(axis cs:15,1.3127892676717)
--(axis cs:14,1.35387805330781)
--(axis cs:13,1.35713877095893)
--(axis cs:12,1.4046399407892)
--(axis cs:11,1.4273839979377)
--(axis cs:10,1.44752708847341)
--(axis cs:9,1.5230889730454)
--(axis cs:8,1.69149512493273)
--(axis cs:7,1.9425064910224)
--(axis cs:6,2.42664984724055)
--(axis cs:5,2.47110254532285)
--(axis cs:4,3.6056749680516)
--(axis cs:3,4.02692893971875)
--cycle;

\path [draw=blue, fill=blue, opacity=0.3]
(axis cs:3,4.02692890167236)
--(axis cs:3,3.24591422080994)
--(axis cs:4,2.58645415306091)
--(axis cs:5,1.76886117458344)
--(axis cs:6,1.46387791633606)
--(axis cs:7,1.20092153549194)
--(axis cs:8,0.577654838562012)
--(axis cs:9,0.457167714834213)
--(axis cs:10,0.287367522716522)
--(axis cs:11,0.227289706468582)
--(axis cs:12,0.193893566727638)
--(axis cs:13,0.1653693318367)
--(axis cs:14,0.149102240800858)
--(axis cs:15,0.142325952649117)
--(axis cs:16,0.13651679456234)
--(axis cs:17,0.129028290510178)
--(axis cs:18,0.129028290510178)
--(axis cs:19,0.118326999247074)
--(axis cs:20,0.118326999247074)
--(axis cs:21,0.118326999247074)
--(axis cs:22,0.118326999247074)
--(axis cs:23,0.117767155170441)
--(axis cs:24,0.107896737754345)
--(axis cs:25,0.107896737754345)
--(axis cs:26,0.107896737754345)
--(axis cs:27,0.103074133396149)
--(axis cs:28,0.103074133396149)
--(axis cs:29,0.103074133396149)
--(axis cs:30,0.102863490581512)
--(axis cs:31,0.102863490581512)
--(axis cs:32,0.0993673130869865)
--(axis cs:33,0.0976372435688972)
--(axis cs:34,0.0976372435688972)
--(axis cs:35,0.0976372435688972)
--(axis cs:36,0.0976372435688972)
--(axis cs:37,0.0976372435688972)
--(axis cs:38,0.0976372435688972)
--(axis cs:39,0.0928342044353485)
--(axis cs:40,0.0927421003580093)
--(axis cs:41,0.0927421003580093)
--(axis cs:42,0.0927421003580093)
--(axis cs:43,0.0927421003580093)
--(axis cs:44,0.0927421003580093)
--(axis cs:45,0.0927421003580093)
--(axis cs:46,0.0922162979841232)
--(axis cs:47,0.0887466669082642)
--(axis cs:48,0.0887466669082642)
--(axis cs:49,0.0887466669082642)
--(axis cs:50,0.0887466669082642)
--(axis cs:51,0.0887466669082642)
--(axis cs:52,0.0887466669082642)
--(axis cs:53,0.0887466669082642)
--(axis cs:54,0.0873439013957977)
--(axis cs:55,0.0873439013957977)
--(axis cs:56,0.0873439013957977)
--(axis cs:57,0.0873439013957977)
--(axis cs:58,0.0873439013957977)
--(axis cs:59,0.0873439013957977)
--(axis cs:60,0.0838145390152931)
--(axis cs:61,0.0838145390152931)
--(axis cs:62,0.0813532620668411)
--(axis cs:63,0.0813532620668411)
--(axis cs:64,0.0813532620668411)
--(axis cs:65,0.0813532620668411)
--(axis cs:66,0.0813532620668411)
--(axis cs:67,0.0813532620668411)
--(axis cs:68,0.0813532620668411)
--(axis cs:69,0.0813532620668411)
--(axis cs:70,0.0813532620668411)
--(axis cs:71,0.0798036754131317)
--(axis cs:72,0.0798036754131317)
--(axis cs:73,0.0798036754131317)
--(axis cs:74,0.0798036754131317)
--(axis cs:75,0.078035444021225)
--(axis cs:76,0.078035444021225)
--(axis cs:77,0.0778139382600784)
--(axis cs:78,0.0778139382600784)
--(axis cs:79,0.0778139382600784)
--(axis cs:80,0.0778139382600784)
--(axis cs:81,0.0778139382600784)
--(axis cs:82,0.0778139382600784)
--(axis cs:83,0.0778139382600784)
--(axis cs:84,0.0778139382600784)
--(axis cs:85,0.0778139382600784)
--(axis cs:86,0.0778139382600784)
--(axis cs:87,0.0778139382600784)
--(axis cs:88,0.0778139382600784)
--(axis cs:89,0.0778139382600784)
--(axis cs:90,0.0778139382600784)
--(axis cs:91,0.0768444836139679)
--(axis cs:92,0.0768444836139679)
--(axis cs:93,0.0768444836139679)
--(axis cs:94,0.0768444836139679)
--(axis cs:95,0.0768444836139679)
--(axis cs:96,0.0768444836139679)
--(axis cs:97,0.0768444836139679)
--(axis cs:98,0.0768444836139679)
--(axis cs:99,0.0768444836139679)
--(axis cs:100,0.0768444836139679)
--(axis cs:101,0.0768444836139679)
--(axis cs:102,0.0768444836139679)
--(axis cs:103,0.0768444836139679)
--(axis cs:104,0.076112762093544)
--(axis cs:104,0.0808856785297394)
--(axis cs:104,0.0808856785297394)
--(axis cs:103,0.0818942189216614)
--(axis cs:102,0.0818942189216614)
--(axis cs:101,0.0818942189216614)
--(axis cs:100,0.0818942189216614)
--(axis cs:99,0.0818942189216614)
--(axis cs:98,0.0818942189216614)
--(axis cs:97,0.0818942189216614)
--(axis cs:96,0.0818942189216614)
--(axis cs:95,0.0818942189216614)
--(axis cs:94,0.0818942189216614)
--(axis cs:93,0.0818942189216614)
--(axis cs:92,0.0818942189216614)
--(axis cs:91,0.0818942189216614)
--(axis cs:90,0.0825383812189102)
--(axis cs:89,0.0825383812189102)
--(axis cs:88,0.0825383812189102)
--(axis cs:87,0.0825383812189102)
--(axis cs:86,0.0825383812189102)
--(axis cs:85,0.0825383812189102)
--(axis cs:84,0.0825383812189102)
--(axis cs:83,0.0825383812189102)
--(axis cs:82,0.0825383812189102)
--(axis cs:81,0.0825383812189102)
--(axis cs:80,0.0825383812189102)
--(axis cs:79,0.0825383812189102)
--(axis cs:78,0.0825383812189102)
--(axis cs:77,0.0825383812189102)
--(axis cs:76,0.0841460227966309)
--(axis cs:75,0.0841460227966309)
--(axis cs:74,0.0859105885028839)
--(axis cs:73,0.0859105885028839)
--(axis cs:72,0.0859105885028839)
--(axis cs:71,0.0859105885028839)
--(axis cs:70,0.0880955904722214)
--(axis cs:69,0.0880955904722214)
--(axis cs:68,0.0880955904722214)
--(axis cs:67,0.0880955904722214)
--(axis cs:66,0.0880955904722214)
--(axis cs:65,0.0880955904722214)
--(axis cs:64,0.0880955904722214)
--(axis cs:63,0.0880955904722214)
--(axis cs:62,0.0880955904722214)
--(axis cs:61,0.0915039852261543)
--(axis cs:60,0.0915039852261543)
--(axis cs:59,0.0959057658910751)
--(axis cs:58,0.0959057658910751)
--(axis cs:57,0.0959057658910751)
--(axis cs:56,0.0959057658910751)
--(axis cs:55,0.0959057658910751)
--(axis cs:54,0.0959057658910751)
--(axis cs:53,0.0978816747665405)
--(axis cs:52,0.0978816747665405)
--(axis cs:51,0.0978816747665405)
--(axis cs:50,0.0978816747665405)
--(axis cs:49,0.0978816747665405)
--(axis cs:48,0.0978816747665405)
--(axis cs:47,0.0978816747665405)
--(axis cs:46,0.10370235145092)
--(axis cs:45,0.105450496077538)
--(axis cs:44,0.105450496077538)
--(axis cs:43,0.105450496077538)
--(axis cs:42,0.105450496077538)
--(axis cs:41,0.105450496077538)
--(axis cs:40,0.105450496077538)
--(axis cs:39,0.10569866001606)
--(axis cs:38,0.1098597869277)
--(axis cs:37,0.1098597869277)
--(axis cs:36,0.1098597869277)
--(axis cs:35,0.1098597869277)
--(axis cs:34,0.1098597869277)
--(axis cs:33,0.1098597869277)
--(axis cs:32,0.111172698438168)
--(axis cs:31,0.114192023873329)
--(axis cs:30,0.114192023873329)
--(axis cs:29,0.114695876836777)
--(axis cs:28,0.114695876836777)
--(axis cs:27,0.114695876836777)
--(axis cs:26,0.123846121132374)
--(axis cs:25,0.123846121132374)
--(axis cs:24,0.123846121132374)
--(axis cs:23,0.152332872152328)
--(axis cs:22,0.152807205915451)
--(axis cs:21,0.152807205915451)
--(axis cs:20,0.152807205915451)
--(axis cs:19,0.152807205915451)
--(axis cs:18,0.166141539812088)
--(axis cs:17,0.166141539812088)
--(axis cs:16,0.196348741650581)
--(axis cs:15,0.200075939297676)
--(axis cs:14,0.206517577171326)
--(axis cs:13,0.258102506399155)
--(axis cs:12,0.302590429782867)
--(axis cs:11,0.446858078241348)
--(axis cs:10,0.527687966823578)
--(axis cs:9,0.818758845329285)
--(axis cs:8,1.01936435699463)
--(axis cs:7,1.74775409698486)
--(axis cs:6,1.97530674934387)
--(axis cs:5,2.27544164657593)
--(axis cs:4,3.57738184928894)
--(axis cs:3,4.02692890167236)
--cycle;

\path [draw=blue, fill=blue, opacity=0.3]
(axis cs:3,4.02692890167236)
--(axis cs:3,3.24591422080994)
--(axis cs:4,2.44299411773682)
--(axis cs:5,1.65112829208374)
--(axis cs:6,1.58197045326233)
--(axis cs:7,0.939526915550232)
--(axis cs:8,0.734895706176758)
--(axis cs:9,0.402687609195709)
--(axis cs:10,0.264930188655853)
--(axis cs:11,0.121380046010017)
--(axis cs:12,0.116011418402195)
--(axis cs:13,0.0942744314670563)
--(axis cs:14,0.0892305299639702)
--(axis cs:15,0.0876713022589684)
--(axis cs:16,0.0876713022589684)
--(axis cs:17,0.0807694867253304)
--(axis cs:18,0.0807694867253304)
--(axis cs:19,0.0803975388407707)
--(axis cs:20,0.0803975388407707)
--(axis cs:21,0.0773421972990036)
--(axis cs:22,0.0773421972990036)
--(axis cs:23,0.0760211646556854)
--(axis cs:24,0.0760211646556854)
--(axis cs:25,0.0755685642361641)
--(axis cs:26,0.0745559558272362)
--(axis cs:27,0.0738402977585793)
--(axis cs:28,0.0737224593758583)
--(axis cs:29,0.0737224593758583)
--(axis cs:30,0.0733534693717957)
--(axis cs:31,0.0733534693717957)
--(axis cs:32,0.0727358758449554)
--(axis cs:33,0.0727222561836243)
--(axis cs:34,0.0727222561836243)
--(axis cs:35,0.0727222561836243)
--(axis cs:36,0.0727222561836243)
--(axis cs:37,0.0727137997746468)
--(axis cs:38,0.0727137997746468)
--(axis cs:39,0.0727137997746468)
--(axis cs:40,0.0727137997746468)
--(axis cs:41,0.0727137997746468)
--(axis cs:42,0.0727137997746468)
--(axis cs:43,0.0727137997746468)
--(axis cs:44,0.0727137997746468)
--(axis cs:45,0.0727137997746468)
--(axis cs:46,0.0727137997746468)
--(axis cs:47,0.0724543482065201)
--(axis cs:48,0.0719625055789948)
--(axis cs:49,0.0719625055789948)
--(axis cs:50,0.0719625055789948)
--(axis cs:51,0.0719625055789948)
--(axis cs:52,0.0719625055789948)
--(axis cs:53,0.0719625055789948)
--(axis cs:54,0.0719625055789948)
--(axis cs:55,0.0719625055789948)
--(axis cs:56,0.0717416405677795)
--(axis cs:57,0.0715111717581749)
--(axis cs:58,0.0715111717581749)
--(axis cs:59,0.0715111717581749)
--(axis cs:60,0.0715111717581749)
--(axis cs:61,0.0715111717581749)
--(axis cs:62,0.0715111717581749)
--(axis cs:63,0.0715111717581749)
--(axis cs:64,0.0715111717581749)
--(axis cs:65,0.0715111717581749)
--(axis cs:66,0.0715111717581749)
--(axis cs:67,0.0715111717581749)
--(axis cs:68,0.0715111717581749)
--(axis cs:69,0.0715111717581749)
--(axis cs:70,0.0715111717581749)
--(axis cs:71,0.0715111717581749)
--(axis cs:72,0.0715111717581749)
--(axis cs:73,0.0715111717581749)
--(axis cs:74,0.0715111717581749)
--(axis cs:75,0.0715111717581749)
--(axis cs:76,0.0715111717581749)
--(axis cs:77,0.0715111717581749)
--(axis cs:78,0.0715111717581749)
--(axis cs:79,0.0715111717581749)
--(axis cs:80,0.0715111717581749)
--(axis cs:81,0.0714431330561638)
--(axis cs:82,0.0714431330561638)
--(axis cs:83,0.0714431330561638)
--(axis cs:84,0.0714431330561638)
--(axis cs:85,0.0709166154265404)
--(axis cs:86,0.0709166154265404)
--(axis cs:87,0.0709166154265404)
--(axis cs:88,0.0709166154265404)
--(axis cs:89,0.0709166154265404)
--(axis cs:90,0.0709166154265404)
--(axis cs:91,0.0709166154265404)
--(axis cs:92,0.0709166154265404)
--(axis cs:93,0.0709166154265404)
--(axis cs:94,0.0709166154265404)
--(axis cs:95,0.0709166154265404)
--(axis cs:96,0.0709166154265404)
--(axis cs:97,0.0709166154265404)
--(axis cs:98,0.0708341598510742)
--(axis cs:99,0.0708341598510742)
--(axis cs:100,0.0708341598510742)
--(axis cs:101,0.0708341598510742)
--(axis cs:102,0.0708341598510742)
--(axis cs:103,0.0708341598510742)
--(axis cs:104,0.0708341598510742)
--(axis cs:104,0.0713708847761154)
--(axis cs:104,0.0713708847761154)
--(axis cs:103,0.0713708847761154)
--(axis cs:102,0.0713708847761154)
--(axis cs:101,0.0713708847761154)
--(axis cs:100,0.0713708847761154)
--(axis cs:99,0.0713708847761154)
--(axis cs:98,0.0713708847761154)
--(axis cs:97,0.0718324109911919)
--(axis cs:96,0.0718324109911919)
--(axis cs:95,0.0718324109911919)
--(axis cs:94,0.0718324109911919)
--(axis cs:93,0.0718324109911919)
--(axis cs:92,0.0718324109911919)
--(axis cs:91,0.0718324109911919)
--(axis cs:90,0.0718324109911919)
--(axis cs:89,0.0718324109911919)
--(axis cs:88,0.0718324109911919)
--(axis cs:87,0.0718324109911919)
--(axis cs:86,0.0718324109911919)
--(axis cs:85,0.0718324109911919)
--(axis cs:84,0.0731892064213753)
--(axis cs:83,0.0731892064213753)
--(axis cs:82,0.0731892064213753)
--(axis cs:81,0.0731892064213753)
--(axis cs:80,0.075920395553112)
--(axis cs:79,0.075920395553112)
--(axis cs:78,0.075920395553112)
--(axis cs:77,0.075920395553112)
--(axis cs:76,0.075920395553112)
--(axis cs:75,0.075920395553112)
--(axis cs:74,0.075920395553112)
--(axis cs:73,0.075920395553112)
--(axis cs:72,0.075920395553112)
--(axis cs:71,0.075920395553112)
--(axis cs:70,0.075920395553112)
--(axis cs:69,0.075920395553112)
--(axis cs:68,0.075920395553112)
--(axis cs:67,0.075920395553112)
--(axis cs:66,0.075920395553112)
--(axis cs:65,0.075920395553112)
--(axis cs:64,0.075920395553112)
--(axis cs:63,0.075920395553112)
--(axis cs:62,0.075920395553112)
--(axis cs:61,0.075920395553112)
--(axis cs:60,0.075920395553112)
--(axis cs:59,0.075920395553112)
--(axis cs:58,0.075920395553112)
--(axis cs:57,0.075920395553112)
--(axis cs:56,0.0761160254478455)
--(axis cs:55,0.0762903690338135)
--(axis cs:54,0.0762903690338135)
--(axis cs:53,0.0762903690338135)
--(axis cs:52,0.0762903690338135)
--(axis cs:51,0.0762903690338135)
--(axis cs:50,0.0762903690338135)
--(axis cs:49,0.0762903690338135)
--(axis cs:48,0.0762903690338135)
--(axis cs:47,0.0768662542104721)
--(axis cs:46,0.0770641788840294)
--(axis cs:45,0.0770641788840294)
--(axis cs:44,0.0770641788840294)
--(axis cs:43,0.0770641788840294)
--(axis cs:42,0.0770641788840294)
--(axis cs:41,0.0770641788840294)
--(axis cs:40,0.0770641788840294)
--(axis cs:39,0.0770641788840294)
--(axis cs:38,0.0770641788840294)
--(axis cs:37,0.0770641788840294)
--(axis cs:36,0.077331155538559)
--(axis cs:35,0.077331155538559)
--(axis cs:34,0.077331155538559)
--(axis cs:33,0.077331155538559)
--(axis cs:32,0.0778405219316483)
--(axis cs:31,0.0784730017185211)
--(axis cs:30,0.0784730017185211)
--(axis cs:29,0.0787654891610146)
--(axis cs:28,0.0787654891610146)
--(axis cs:27,0.0807976052165031)
--(axis cs:26,0.081835575401783)
--(axis cs:25,0.0832872614264488)
--(axis cs:24,0.083591490983963)
--(axis cs:23,0.083591490983963)
--(axis cs:22,0.0850094109773636)
--(axis cs:21,0.0850094109773636)
--(axis cs:20,0.0886454358696938)
--(axis cs:19,0.0886454358696938)
--(axis cs:18,0.0891855731606483)
--(axis cs:17,0.0891855731606483)
--(axis cs:16,0.0957923606038094)
--(axis cs:15,0.0957923606038094)
--(axis cs:14,0.0964502468705177)
--(axis cs:13,0.119896024465561)
--(axis cs:12,0.145768761634827)
--(axis cs:11,0.231368824839592)
--(axis cs:10,0.806348145008087)
--(axis cs:9,0.918398916721344)
--(axis cs:8,1.21697080135345)
--(axis cs:7,1.73025763034821)
--(axis cs:6,2.28359198570251)
--(axis cs:5,2.3674578666687)
--(axis cs:4,3.21731376647949)
--(axis cs:3,4.02692890167236)
--cycle;

\addplot [semithick, color1, dash dot]
table {%
3 3.63642158508301
4 2.43888836634499
5 2.42736646848634
6 1.40528763573607
7 1.13643310701652
8 0.778878446463143
9 0.443339967681593
10 0.388569175125278
11 0.356866031543876
12 0.287896501774018
13 0.136819654187523
14 0.136359833958998
15 0.102007158532273
16 0.0971888681306123
17 0.0833343485213086
18 0.083188124628639
19 0.0792668348966861
20 0.079244573626026
21 0.07880218068951
22 0.07880218068951
23 0.0759448493269905
24 0.0759448493269905
25 0.0752471111312631
26 0.0752471111312631
27 0.0752034723376469
28 0.0745924406244201
29 0.0745698723426085
30 0.0744982535699315
31 0.0744982535699315
32 0.0742426749477713
33 0.0732055190747158
34 0.0732055190747158
35 0.0730068825158125
36 0.0730068825158125
37 0.0730068825158125
38 0.0726376296305411
39 0.0726376296305411
40 0.0726376296305411
41 0.0721464827466712
42 0.0721464827466712
43 0.0721464827466712
44 0.0721464827466712
45 0.0721464827466712
46 0.0721464827466712
47 0.0721464827466712
48 0.0721464827466712
49 0.0721464827466712
50 0.0721464827466712
51 0.0719482880063439
52 0.0719482880063439
53 0.0719482880063439
54 0.0719482880063439
55 0.0719482880063439
56 0.0719482880063439
57 0.0719482880063439
58 0.0709629586681882
59 0.0706999483028747
60 0.0706999483028747
61 0.0706999483028747
62 0.0706999483028747
63 0.0706999483028747
64 0.0706999483028747
65 0.0706999483028747
66 0.0706999483028747
67 0.0706999483028747
68 0.0706999483028747
69 0.0706999483028747
70 0.0706999483028747
71 0.0706246425806442
72 0.0706246425806442
73 0.0706246425806442
74 0.0706246425806442
75 0.0706246425806442
76 0.0706246425806442
77 0.0706246425806442
78 0.0706246425806442
79 0.0706246425806442
80 0.0706246425806442
81 0.0706246425806442
82 0.0706246425806442
83 0.0706246425806442
84 0.0706246425806442
85 0.0706246425806442
86 0.0706246425806442
87 0.0706246425806442
88 0.0706246425806442
89 0.0706200715577048
90 0.0703322103632829
91 0.070126070775789
92 0.070126070775789
93 0.0699662160004881
94 0.0699662160004881
95 0.0699662160004881
96 0.0699662160004881
97 0.0699662160004881
98 0.0699662160004881
99 0.0699662160004881
100 0.0699662160004881
101 0.0699662160004881
102 0.0699662160004881
103 0.0699662160004881
104 0.0699662160004881
};
\addplot [semithick, color1]
table {%
3 3.63642158508301
4 2.96361709257976
5 1.90230331691265
6 1.85272042841011
7 1.42979631498246
8 1.1935437497688
9 1.03166440033173
10 0.943223682093473
11 0.920030691369137
12 0.893640042828575
13 0.839494323940729
14 0.835630369746572
15 0.797560471725799
16 0.576769862935599
17 0.389815482041436
18 0.389815482041436
19 0.354003573380474
20 0.197825647769735
21 0.0854469540561433
22 0.0761739970394978
23 0.0761379802731821
24 0.0735791284247463
25 0.0733597799694778
26 0.0723885564378027
27 0.0723885564378027
28 0.0706680124008341
29 0.0703553986791907
30 0.0703553986791907
31 0.0703553986791907
32 0.0703553986791907
33 0.0703553986791907
34 0.0703553986791907
35 0.0703388534660349
36 0.0703388534660349
37 0.0703388534660349
38 0.0703185521291396
39 0.0703185521291396
40 0.0702290261919409
41 0.0702118190952807
42 0.0702118190952807
43 0.0701188107327511
44 0.0701188107327511
45 0.069869612723732
46 0.069869612723732
47 0.0698336183510997
48 0.0697485139006095
49 0.0697485139006095
50 0.0697485139006095
51 0.0697485139006095
52 0.0697485139006095
53 0.0697485139006095
54 0.0697485139006095
55 0.0697485139006095
56 0.0697463221382958
57 0.0697463221382958
58 0.0697463221382958
59 0.0697463221382958
60 0.069741432093769
61 0.069741432093769
62 0.069741432093769
63 0.069741432093769
64 0.0697210141526729
65 0.0697210141526729
66 0.0697210141526729
67 0.0697210141526729
68 0.0697210141526729
69 0.0697210141526729
70 0.0697210141526729
71 0.0697210141526729
72 0.0697210141526729
73 0.0697210141526729
74 0.0697210141526729
75 0.0697210141526729
76 0.0697210141526729
77 0.0697210141526729
78 0.0697210141526729
79 0.0697210141526729
80 0.0697210141526729
81 0.0697210141526729
82 0.0697210141526729
83 0.0697210141526729
84 0.0697210141526729
85 0.0697210141526729
86 0.0697210141526729
87 0.0697210141526729
88 0.0697210141526729
89 0.0697210141526729
90 0.0697210141526729
91 0.0697210141526729
92 0.0697093768552034
93 0.0697093768552034
94 0.0697093768552034
95 0.0697093768552034
96 0.0696867301361693
97 0.0696867301361693
98 0.0696867301361693
99 0.0696867301361693
100 0.0696443649678685
101 0.0696443649678685
102 0.0696443649678685
103 0.069639810965266
104 0.069639810965266
};
\addplot [semithick, blue, dash dot]
table {%
3 3.63642168045044
4 3.08191800117493
5 2.02215147018433
6 1.71959233283997
7 1.4743378162384
8 0.79850959777832
9 0.63796329498291
10 0.40752774477005
11 0.337073892354965
12 0.248241990804672
13 0.211735919117928
14 0.177809908986092
15 0.171200945973396
16 0.166432768106461
17 0.147584915161133
18 0.147584915161133
19 0.135567098855972
20 0.135567098855972
21 0.135567098855972
22 0.135567098855972
23 0.135050013661385
24 0.115871429443359
25 0.115871429443359
26 0.115871429443359
27 0.108885005116463
28 0.108885005116463
29 0.108885005116463
30 0.108527757227421
31 0.108527757227421
32 0.105270005762577
33 0.103748515248299
34 0.103748515248299
35 0.103748515248299
36 0.103748515248299
37 0.103748515248299
38 0.103748515248299
39 0.0992664322257042
40 0.0990962982177734
41 0.0990962982177734
42 0.0990962982177734
43 0.0990962982177734
44 0.0990962982177734
45 0.0990962982177734
46 0.0979593247175217
47 0.0933141708374023
48 0.0933141708374023
49 0.0933141708374023
50 0.0933141708374023
51 0.0933141708374023
52 0.0933141708374023
53 0.0933141708374023
54 0.0916248336434364
55 0.0916248336434364
56 0.0916248336434364
57 0.0916248336434364
58 0.0916248336434364
59 0.0916248336434364
60 0.0876592621207237
61 0.0876592621207237
62 0.0847244262695312
63 0.0847244262695312
64 0.0847244262695312
65 0.0847244262695312
66 0.0847244262695312
67 0.0847244262695312
68 0.0847244262695312
69 0.0847244262695312
70 0.0847244262695312
71 0.0828571319580078
72 0.0828571319580078
73 0.0828571319580078
74 0.0828571319580078
75 0.0810907334089279
76 0.0810907334089279
77 0.0801761597394943
78 0.0801761597394943
79 0.0801761597394943
80 0.0801761597394943
81 0.0801761597394943
82 0.0801761597394943
83 0.0801761597394943
84 0.0801761597394943
85 0.0801761597394943
86 0.0801761597394943
87 0.0801761597394943
88 0.0801761597394943
89 0.0801761597394943
90 0.0801761597394943
91 0.0793693512678146
92 0.0793693512678146
93 0.0793693512678146
94 0.0793693512678146
95 0.0793693512678146
96 0.0793693512678146
97 0.0793693512678146
98 0.0793693512678146
99 0.0793693512678146
100 0.0793693512678146
101 0.0793693512678146
102 0.0793693512678146
103 0.0793693512678146
104 0.0784992203116417
};
\addplot [semithick, blue]
table {%
3 3.63642168045044
4 2.83015394210815
5 2.00929307937622
6 1.93278121948242
7 1.33489227294922
8 0.975933253765106
9 0.660543262958527
10 0.53563916683197
11 0.176374435424805
12 0.130890086293221
13 0.107085227966309
14 0.092840388417244
15 0.0917318314313889
16 0.0917318314313889
17 0.0849775299429893
18 0.0849775299429893
19 0.0845214873552322
20 0.0845214873552322
21 0.0811758041381836
22 0.0811758041381836
23 0.0798063278198242
24 0.0798063278198242
25 0.0794279128313065
26 0.0781957656145096
27 0.0773189514875412
28 0.0762439742684364
29 0.0762439742684364
30 0.0759132355451584
31 0.0759132355451584
32 0.0752881988883018
33 0.0750267058610916
34 0.0750267058610916
35 0.0750267058610916
36 0.0750267058610916
37 0.0748889893293381
38 0.0748889893293381
39 0.0748889893293381
40 0.0748889893293381
41 0.0748889893293381
42 0.0748889893293381
43 0.0748889893293381
44 0.0748889893293381
45 0.0748889893293381
46 0.0748889893293381
47 0.0746603012084961
48 0.0741264373064041
49 0.0741264373064041
50 0.0741264373064041
51 0.0741264373064041
52 0.0741264373064041
53 0.0741264373064041
54 0.0741264373064041
55 0.0741264373064041
56 0.0739288330078125
57 0.0737157836556435
58 0.0737157836556435
59 0.0737157836556435
60 0.0737157836556435
61 0.0737157836556435
62 0.0737157836556435
63 0.0737157836556435
64 0.0737157836556435
65 0.0737157836556435
66 0.0737157836556435
67 0.0737157836556435
68 0.0737157836556435
69 0.0737157836556435
70 0.0737157836556435
71 0.0737157836556435
72 0.0737157836556435
73 0.0737157836556435
74 0.0737157836556435
75 0.0737157836556435
76 0.0737157836556435
77 0.0737157836556435
78 0.0737157836556435
79 0.0737157836556435
80 0.0737157836556435
81 0.0723161697387695
82 0.0723161697387695
83 0.0723161697387695
84 0.0723161697387695
85 0.0713745132088661
86 0.0713745132088661
87 0.0713745132088661
88 0.0713745132088661
89 0.0713745132088661
90 0.0713745132088661
91 0.0713745132088661
92 0.0713745132088661
93 0.0713745132088661
94 0.0713745132088661
95 0.0713745132088661
96 0.0713745132088661
97 0.0713745132088661
98 0.0711025223135948
99 0.0711025223135948
100 0.0711025223135948
101 0.0711025223135948
102 0.0711025223135948
103 0.0711025223135948
104 0.0711025223135948
};
\end{axis}

\end{tikzpicture}

%% file: figures/bop_1d_gramacylee_weak.tex
\begin{tikzpicture}

\definecolor{color0}{rgb}{0,0,1}
\definecolor{color1}{rgb}{1,0.549019607843137,0}
\definecolor{color2}{rgb}{1,0.647058823529412,0}
\definecolor{color3}{rgb}{0.564705882352941,0.933333333333333,0.564705882352941}

\begin{axis}[axis on top,
enlarge x limits=false,
enlarge y limits=false,
height=\figureheight,
scale only axis,
tick align=outside,
tick pos=left,
tick pos=left,
width=\figurewidth,
xmin=3, xmax=75,
xtick style={color=black},
xtick={-10,0,10,25,50,75,100},
xticklabels={\ensuremath{-}10,0,10,25,50,75,90},
ymin=-0.03, ymax=0.7,
ytick style={color=black},
ytick={0.   , 0.7},
]
\node[anchor=north east] at (rel axis cs:1,1) {Gramacy Lee 1D (weak)};
\path [draw=color1, fill=color1, opacity=0.3]
(axis cs:3,0.693378502433763)
--(axis cs:3,0.483341099689005)
--(axis cs:4,0.402588220843527)
--(axis cs:5,0.34273555941984)
--(axis cs:6,0.330617495176611)
--(axis cs:7,0.303990856209147)
--(axis cs:8,0.299809374556666)
--(axis cs:9,0.2960195267413)
--(axis cs:10,0.2960195267413)
--(axis cs:11,0.27518191599544)
--(axis cs:12,0.27518191599544)
--(axis cs:13,0.212960180008012)
--(axis cs:14,0.173122082424271)
--(axis cs:15,0.172675792475223)
--(axis cs:16,0.156752221184697)
--(axis cs:17,0.156752221184697)
--(axis cs:18,0.156012186926451)
--(axis cs:19,0.156012186926451)
--(axis cs:20,0.152320927351411)
--(axis cs:21,0.144621941924371)
--(axis cs:22,0.144621941924371)
--(axis cs:23,0.143491811660113)
--(axis cs:24,0.140818008746422)
--(axis cs:25,0.107996361430422)
--(axis cs:26,0.105677560087453)
--(axis cs:27,0.105677560087453)
--(axis cs:28,0.101756634125556)
--(axis cs:29,0.101756634125556)
--(axis cs:30,0.101707575648071)
--(axis cs:31,0.0605125880455264)
--(axis cs:32,0.0358468968792692)
--(axis cs:33,0.0241359407459898)
--(axis cs:34,0.0241359407459898)
--(axis cs:35,0.0112566940803203)
--(axis cs:36,0.00714790029274484)
--(axis cs:37,0.00589368456485641)
--(axis cs:38,0.00589368456485641)
--(axis cs:39,0.00589368456485641)
--(axis cs:40,0.0039236419195707)
--(axis cs:41,0.00366764164225263)
--(axis cs:42,0.00346006982389629)
--(axis cs:43,0.00346006982389629)
--(axis cs:44,0.00275028587533979)
--(axis cs:45,0.00075121229586236)
--(axis cs:46,0.00075121229586236)
--(axis cs:47,0.00075121229586236)
--(axis cs:48,0.000352141705260541)
--(axis cs:49,0.00035207269183014)
--(axis cs:50,0.00035207269183014)
--(axis cs:51,0.000237188220144912)
--(axis cs:52,0.00017686935334589)
--(axis cs:53,0.000175708453153494)
--(axis cs:54,0.000175708175058759)
--(axis cs:55,0.000175706012234707)
--(axis cs:56,0.000175706012234707)
--(axis cs:57,0.000175706012234707)
--(axis cs:58,0.000141387830088359)
--(axis cs:59,0.000141387830088359)
--(axis cs:60,0.000141387830088359)
--(axis cs:61,0.000141387830088359)
--(axis cs:62,0.000103204878336434)
--(axis cs:63,0.000103204878336434)
--(axis cs:64,0.000103204878336434)
--(axis cs:65,0.000103204878336434)
--(axis cs:66,0.000102523264213911)
--(axis cs:67,0.000102523264213911)
--(axis cs:68,0.000102523264213911)
--(axis cs:69,9.97816239772867e-05)
--(axis cs:70,9.97816239772867e-05)
--(axis cs:71,9.97816239772867e-05)
--(axis cs:72,9.97816239772867e-05)
--(axis cs:73,9.1994264074885e-05)
--(axis cs:74,9.1994264074885e-05)
--(axis cs:75,9.1994264074885e-05)
--(axis cs:76,9.1994264074885e-05)
--(axis cs:77,9.1994264074885e-05)
--(axis cs:78,9.1994262100447e-05)
--(axis cs:79,9.1994262100447e-05)
--(axis cs:80,9.1994262100447e-05)
--(axis cs:81,9.1994262100447e-05)
--(axis cs:82,9.1994262100447e-05)
--(axis cs:83,9.1994262100447e-05)
--(axis cs:84,9.1994262100447e-05)
--(axis cs:85,9.19942600525123e-05)
--(axis cs:86,9.19942600525123e-05)
--(axis cs:87,9.19942600525123e-05)
--(axis cs:88,7.61725423138217e-05)
--(axis cs:89,5.97791657489105e-05)
--(axis cs:90,4.74437206921426e-05)
--(axis cs:91,4.74437206921426e-05)
--(axis cs:92,4.65902529398661e-05)
--(axis cs:93,4.57780946659936e-05)
--(axis cs:94,4.57780946659936e-05)
--(axis cs:95,4.57780946659936e-05)
--(axis cs:96,4.57780946659936e-05)
--(axis cs:97,4.57780946659936e-05)
--(axis cs:98,4.57780946659936e-05)
--(axis cs:99,2.67423803356308e-05)
--(axis cs:100,2.67423803356308e-05)
--(axis cs:101,2.67423803356308e-05)
--(axis cs:102,2.67423803356308e-05)
--(axis cs:103,2.67423803356308e-05)
--(axis cs:104,2.67423803356308e-05)
--(axis cs:104,0.0411679532035003)
--(axis cs:104,0.0411679532035003)
--(axis cs:103,0.0411679532035003)
--(axis cs:102,0.0411679532035003)
--(axis cs:101,0.0411679532035003)
--(axis cs:100,0.0411679532035003)
--(axis cs:99,0.0411679532035003)
--(axis cs:98,0.0411831940299007)
--(axis cs:97,0.0411831940299007)
--(axis cs:96,0.0411831940299007)
--(axis cs:95,0.0411831940299007)
--(axis cs:94,0.0411831940299007)
--(axis cs:93,0.0411831940299007)
--(axis cs:92,0.0411838434704584)
--(axis cs:91,0.0411845263475396)
--(axis cs:90,0.0411845263475396)
--(axis cs:89,0.041194394925335)
--(axis cs:88,0.0412075176220323)
--(axis cs:87,0.0412201745385055)
--(axis cs:86,0.0412201745385055)
--(axis cs:85,0.0412201745385055)
--(axis cs:84,0.0412268380164596)
--(axis cs:83,0.0412268380164596)
--(axis cs:82,0.0412268380164596)
--(axis cs:81,0.0412268380164596)
--(axis cs:80,0.0412268380164596)
--(axis cs:79,0.0412268380164596)
--(axis cs:78,0.0412268380164596)
--(axis cs:77,0.0412332643246623)
--(axis cs:76,0.0412332643246623)
--(axis cs:75,0.0412332643246623)
--(axis cs:74,0.0412332643246623)
--(axis cs:73,0.0412332643246623)
--(axis cs:72,0.0412394928531452)
--(axis cs:71,0.0412394928531452)
--(axis cs:70,0.0412394928531452)
--(axis cs:69,0.0412394928531452)
--(axis cs:68,0.041241684255614)
--(axis cs:67,0.041241684255614)
--(axis cs:66,0.041241684255614)
--(axis cs:65,0.0412422289842026)
--(axis cs:64,0.0412422289842026)
--(axis cs:63,0.0412422289842026)
--(axis cs:62,0.0412422289842026)
--(axis cs:61,0.0412728428024383)
--(axis cs:60,0.0412728428024383)
--(axis cs:59,0.0412728428024383)
--(axis cs:58,0.0412728428024383)
--(axis cs:57,0.0413003099722482)
--(axis cs:56,0.0413003099722482)
--(axis cs:55,0.0413003099722482)
--(axis cs:54,0.0429834060662246)
--(axis cs:53,0.0432098165208732)
--(axis cs:52,0.0432107454783088)
--(axis cs:51,0.0432590631688732)
--(axis cs:50,0.0433513606568694)
--(axis cs:49,0.0433513606568694)
--(axis cs:48,0.0559699930819771)
--(axis cs:47,0.0562922247620967)
--(axis cs:46,0.0562922247620967)
--(axis cs:45,0.0562922247620967)
--(axis cs:44,0.0579168961491435)
--(axis cs:43,0.0584991582931183)
--(axis cs:42,0.0584991582931183)
--(axis cs:41,0.0586632727057059)
--(axis cs:40,0.0588640050321866)
--(axis cs:39,0.060579059166154)
--(axis cs:38,0.060579059166154)
--(axis cs:37,0.060579059166154)
--(axis cs:36,0.0627609482908477)
--(axis cs:35,0.0666434942137179)
--(axis cs:34,0.087097267120766)
--(axis cs:33,0.087097267120766)
--(axis cs:32,0.0988222175575144)
--(axis cs:31,0.127038802541558)
--(axis cs:30,0.206288893633258)
--(axis cs:29,0.206486048113879)
--(axis cs:28,0.206486048113879)
--(axis cs:27,0.214197170068076)
--(axis cs:26,0.214197170068076)
--(axis cs:25,0.21966540236873)
--(axis cs:24,0.249745838603798)
--(axis cs:23,0.250959640926785)
--(axis cs:22,0.251537772730546)
--(axis cs:21,0.251537772730546)
--(axis cs:20,0.27508960351549)
--(axis cs:19,0.279125482564191)
--(axis cs:18,0.279125482564191)
--(axis cs:17,0.279451460269643)
--(axis cs:16,0.279451460269643)
--(axis cs:15,0.301242025330032)
--(axis cs:14,0.301434353447913)
--(axis cs:13,0.332519098570778)
--(axis cs:12,0.377956128272937)
--(axis cs:11,0.377956128272937)
--(axis cs:10,0.431681641328048)
--(axis cs:9,0.431681641328048)
--(axis cs:8,0.436401881637937)
--(axis cs:7,0.461431383696981)
--(axis cs:6,0.539413824492648)
--(axis cs:5,0.551397226396514)
--(axis cs:4,0.640549266984346)
--(axis cs:3,0.693378502433763)
--cycle;

\path [draw=blue, fill=blue, opacity=0.3]
(axis cs:3,0.693378567695618)
--(axis cs:3,0.483341127634048)
--(axis cs:4,0.392196148633957)
--(axis cs:5,0.37172594666481)
--(axis cs:6,0.356535345315933)
--(axis cs:7,0.329365253448486)
--(axis cs:8,0.329365253448486)
--(axis cs:9,0.30151703953743)
--(axis cs:10,0.280828684568405)
--(axis cs:11,0.26484602689743)
--(axis cs:12,0.23950120806694)
--(axis cs:13,0.239355161786079)
--(axis cs:14,0.217315018177032)
--(axis cs:15,0.215245917439461)
--(axis cs:16,0.214145109057426)
--(axis cs:17,0.213079571723938)
--(axis cs:18,0.213079571723938)
--(axis cs:19,0.193825498223305)
--(axis cs:20,0.171152800321579)
--(axis cs:21,0.171152800321579)
--(axis cs:22,0.159736469388008)
--(axis cs:23,0.145386308431625)
--(axis cs:24,0.143393591046333)
--(axis cs:25,0.143393591046333)
--(axis cs:26,0.140576094388962)
--(axis cs:27,0.131147295236588)
--(axis cs:28,0.0985710546374321)
--(axis cs:29,0.0891778469085693)
--(axis cs:30,0.0891778469085693)
--(axis cs:31,0.0891778469085693)
--(axis cs:32,0.0879601985216141)
--(axis cs:33,0.0877914801239967)
--(axis cs:34,0.0580025464296341)
--(axis cs:35,0.057048637419939)
--(axis cs:36,0.0564864575862885)
--(axis cs:37,0.0315755866467953)
--(axis cs:38,0.0315755866467953)
--(axis cs:39,0.0315755866467953)
--(axis cs:40,0.0315755866467953)
--(axis cs:41,0.0166341494768858)
--(axis cs:42,0.0163041111081839)
--(axis cs:43,0.0157876536250114)
--(axis cs:44,0.0142642091959715)
--(axis cs:45,0.0142642091959715)
--(axis cs:46,0.0140522159636021)
--(axis cs:47,0.0140522159636021)
--(axis cs:48,0.0140522159636021)
--(axis cs:49,0.0140522159636021)
--(axis cs:50,0.0140522159636021)
--(axis cs:51,0.0140522159636021)
--(axis cs:52,0.0140522159636021)
--(axis cs:53,0.0140522159636021)
--(axis cs:54,0.0140522159636021)
--(axis cs:55,0.0140522159636021)
--(axis cs:56,0.0140522159636021)
--(axis cs:57,0.0140522159636021)
--(axis cs:58,0.0140522159636021)
--(axis cs:59,0.0140522159636021)
--(axis cs:60,0.0138025842607021)
--(axis cs:61,0.0138025842607021)
--(axis cs:62,0.0138025842607021)
--(axis cs:63,0.0138025842607021)
--(axis cs:64,0.0137947741895914)
--(axis cs:65,0.0137947741895914)
--(axis cs:66,0.0137947741895914)
--(axis cs:67,0.0137944445014)
--(axis cs:68,0.0137944445014)
--(axis cs:69,0.0137944445014)
--(axis cs:70,0.0137944445014)
--(axis cs:71,0.0137944445014)
--(axis cs:72,0.0137944445014)
--(axis cs:73,0.0137944445014)
--(axis cs:74,0.0137944445014)
--(axis cs:75,0.0137944445014)
--(axis cs:76,0.0137944445014)
--(axis cs:77,0.0137725099921227)
--(axis cs:78,5.03771007061005e-05)
--(axis cs:79,5.03771007061005e-05)
--(axis cs:80,5.03771007061005e-05)
--(axis cs:81,5.0373375415802e-05)
--(axis cs:82,5.03547489643097e-05)
--(axis cs:83,5.03547489643097e-05)
--(axis cs:84,3.18903475999832e-05)
--(axis cs:85,3.18903475999832e-05)
--(axis cs:86,3.18903475999832e-05)
--(axis cs:87,3.18693928420544e-05)
--(axis cs:88,3.18693928420544e-05)
--(axis cs:89,3.18693928420544e-05)
--(axis cs:90,3.02203916362487e-05)
--(axis cs:91,3.02203916362487e-05)
--(axis cs:92,3.02203916362487e-05)
--(axis cs:93,3.02203916362487e-05)
--(axis cs:94,3.02203916362487e-05)
--(axis cs:95,2.60440356214531e-05)
--(axis cs:96,2.60440356214531e-05)
--(axis cs:97,2.60440356214531e-05)
--(axis cs:98,2.13914099731483e-05)
--(axis cs:99,2.13914099731483e-05)
--(axis cs:100,2.13914099731483e-05)
--(axis cs:101,2.13914099731483e-05)
--(axis cs:102,2.13914099731483e-05)
--(axis cs:103,2.13914099731483e-05)
--(axis cs:104,2.13914099731483e-05)
--(axis cs:104,0.000149161322042346)
--(axis cs:104,0.000149161322042346)
--(axis cs:103,0.000149161322042346)
--(axis cs:102,0.000149161322042346)
--(axis cs:101,0.000149161322042346)
--(axis cs:100,0.000149161322042346)
--(axis cs:99,0.000149161322042346)
--(axis cs:98,0.000149161322042346)
--(axis cs:97,0.000159076065756381)
--(axis cs:96,0.000159076065756381)
--(axis cs:95,0.000159076065756381)
--(axis cs:94,0.000162874814122915)
--(axis cs:93,0.000162874814122915)
--(axis cs:92,0.000162874814122915)
--(axis cs:91,0.000162874814122915)
--(axis cs:90,0.000162874814122915)
--(axis cs:89,0.00845680944621563)
--(axis cs:88,0.00845680944621563)
--(axis cs:87,0.00845680944621563)
--(axis cs:86,0.035905696451664)
--(axis cs:85,0.035905696451664)
--(axis cs:84,0.035905696451664)
--(axis cs:83,0.0359205007553101)
--(axis cs:82,0.0359205007553101)
--(axis cs:81,0.0359205156564713)
--(axis cs:80,0.0412512421607971)
--(axis cs:79,0.0412512421607971)
--(axis cs:78,0.0412512421607971)
--(axis cs:77,0.0686789751052856)
--(axis cs:76,0.0687887296080589)
--(axis cs:75,0.0687887296080589)
--(axis cs:74,0.0687887296080589)
--(axis cs:73,0.0687887296080589)
--(axis cs:72,0.0687887296080589)
--(axis cs:71,0.0687887296080589)
--(axis cs:70,0.0687887296080589)
--(axis cs:69,0.0687887296080589)
--(axis cs:68,0.0687887296080589)
--(axis cs:67,0.0687887296080589)
--(axis cs:66,0.068788968026638)
--(axis cs:65,0.068788968026638)
--(axis cs:64,0.068788968026638)
--(axis cs:63,0.0688281953334808)
--(axis cs:62,0.0688281953334808)
--(axis cs:61,0.0688281953334808)
--(axis cs:60,0.0688281953334808)
--(axis cs:59,0.0690082907676697)
--(axis cs:58,0.0690082907676697)
--(axis cs:57,0.0690082907676697)
--(axis cs:56,0.0690082907676697)
--(axis cs:55,0.0690082907676697)
--(axis cs:54,0.0690082907676697)
--(axis cs:53,0.0690082907676697)
--(axis cs:52,0.0690082907676697)
--(axis cs:51,0.0690082907676697)
--(axis cs:50,0.0690082907676697)
--(axis cs:49,0.0690082907676697)
--(axis cs:48,0.0690082907676697)
--(axis cs:47,0.0690082907676697)
--(axis cs:46,0.0690082907676697)
--(axis cs:45,0.0691603198647499)
--(axis cs:44,0.0691603198647499)
--(axis cs:43,0.0702901184558868)
--(axis cs:42,0.0706660374999046)
--(axis cs:41,0.0709241777658463)
--(axis cs:40,0.0942173004150391)
--(axis cs:39,0.0942173004150391)
--(axis cs:38,0.0942173004150391)
--(axis cs:37,0.0942173004150391)
--(axis cs:36,0.139380216598511)
--(axis cs:35,0.142419725656509)
--(axis cs:34,0.144160762429237)
--(axis cs:33,0.183180212974548)
--(axis cs:32,0.183753058314323)
--(axis cs:31,0.184423118829727)
--(axis cs:30,0.184423118829727)
--(axis cs:29,0.184423118829727)
--(axis cs:28,0.200993627309799)
--(axis cs:27,0.236386388540268)
--(axis cs:26,0.241168707609177)
--(axis cs:25,0.24288584291935)
--(axis cs:24,0.24288584291935)
--(axis cs:23,0.243708819150925)
--(axis cs:22,0.254551261663437)
--(axis cs:21,0.260145783424377)
--(axis cs:20,0.260145783424377)
--(axis cs:19,0.278214037418365)
--(axis cs:18,0.306747078895569)
--(axis cs:17,0.306747078895569)
--(axis cs:16,0.308156937360764)
--(axis cs:15,0.309841722249985)
--(axis cs:14,0.313205480575562)
--(axis cs:13,0.371503531932831)
--(axis cs:12,0.371732443571091)
--(axis cs:11,0.419986605644226)
--(axis cs:10,0.442252784967422)
--(axis cs:9,0.450673788785934)
--(axis cs:8,0.474575161933899)
--(axis cs:7,0.474575161933899)
--(axis cs:6,0.510518848896027)
--(axis cs:5,0.54637622833252)
--(axis cs:4,0.589954316616058)
--(axis cs:3,0.693378567695618)
--cycle;

\path [draw=blue, fill=blue, opacity=0.3]
(axis cs:3,0.693378448486328)
--(axis cs:3,0.483341068029404)
--(axis cs:4,0.435814201831818)
--(axis cs:5,0.428656578063965)
--(axis cs:6,0.330662250518799)
--(axis cs:7,0.299680858850479)
--(axis cs:8,0.243235260248184)
--(axis cs:9,0.222592756152153)
--(axis cs:10,0.204116702079773)
--(axis cs:11,0.190832331776619)
--(axis cs:12,0.184728398919106)
--(axis cs:13,0.147729814052582)
--(axis cs:14,0.106804512441158)
--(axis cs:15,0.106804512441158)
--(axis cs:16,0.103654451668262)
--(axis cs:17,0.10218258947134)
--(axis cs:18,0.101770393550396)
--(axis cs:19,0.101284705102444)
--(axis cs:20,0.100811839103699)
--(axis cs:21,0.100811839103699)
--(axis cs:22,0.100424833595753)
--(axis cs:23,0.100424833595753)
--(axis cs:24,0.079785481095314)
--(axis cs:25,0.0786969661712646)
--(axis cs:26,0.0782840624451637)
--(axis cs:27,0.0782840624451637)
--(axis cs:28,0.0738817825913429)
--(axis cs:29,0.0736120343208313)
--(axis cs:30,0.0733769834041595)
--(axis cs:31,0.0726112052798271)
--(axis cs:32,0.072419174015522)
--(axis cs:33,0.072419174015522)
--(axis cs:34,0.072419174015522)
--(axis cs:35,0.0723841041326523)
--(axis cs:36,0.0723841041326523)
--(axis cs:37,0.0723841041326523)
--(axis cs:38,0.0431708432734013)
--(axis cs:39,0.0431708358228207)
--(axis cs:40,0.0431708358228207)
--(axis cs:41,0.0424252338707447)
--(axis cs:42,0.0423364415764809)
--(axis cs:43,0.0364027172327042)
--(axis cs:44,0.0252194479107857)
--(axis cs:45,0.00945252925157547)
--(axis cs:46,0.00945252925157547)
--(axis cs:47,0.00318620726466179)
--(axis cs:48,0.00318449176847935)
--(axis cs:49,0.000966893509030342)
--(axis cs:50,0.000776054337620735)
--(axis cs:51,0.000776054337620735)
--(axis cs:52,0.000776054337620735)
--(axis cs:53,0.000776054337620735)
--(axis cs:54,0.000776052474975586)
--(axis cs:55,0.000776052474975586)
--(axis cs:56,0.000603828579187393)
--(axis cs:57,0.00041377916932106)
--(axis cs:58,0.000380311161279678)
--(axis cs:59,0.000380311161279678)
--(axis cs:60,0.000380311161279678)
--(axis cs:61,0.00021786242723465)
--(axis cs:62,0.00021786242723465)
--(axis cs:63,0.00021786242723465)
--(axis cs:64,0.00021786242723465)
--(axis cs:65,0.00021786242723465)
--(axis cs:66,0.000189591199159622)
--(axis cs:67,0.000189591199159622)
--(axis cs:68,0.000189591199159622)
--(axis cs:69,0.000189591199159622)
--(axis cs:70,0.000189591199159622)
--(axis cs:71,0.000189591199159622)
--(axis cs:72,0.000177942216396332)
--(axis cs:73,0.000177895650267601)
--(axis cs:74,0.000171367079019547)
--(axis cs:75,0.000171367079019547)
--(axis cs:76,0.000149545259773731)
--(axis cs:77,0.00014530643238686)
--(axis cs:78,0.000129740132251754)
--(axis cs:79,6.94327900419012e-05)
--(axis cs:80,6.94327900419012e-05)
--(axis cs:81,6.94327900419012e-05)
--(axis cs:82,6.94327900419012e-05)
--(axis cs:83,6.94327900419012e-05)
--(axis cs:84,6.94327900419012e-05)
--(axis cs:85,6.77602220093831e-05)
--(axis cs:86,6.77602220093831e-05)
--(axis cs:87,6.77602220093831e-05)
--(axis cs:88,6.77602220093831e-05)
--(axis cs:89,6.77602220093831e-05)
--(axis cs:90,6.77602220093831e-05)
--(axis cs:91,6.77602220093831e-05)
--(axis cs:92,6.77602220093831e-05)
--(axis cs:93,6.48315617581829e-05)
--(axis cs:94,6.48315617581829e-05)
--(axis cs:95,6.48315617581829e-05)
--(axis cs:96,6.48315617581829e-05)
--(axis cs:97,6.48315617581829e-05)
--(axis cs:98,6.48315617581829e-05)
--(axis cs:99,6.3142811995931e-05)
--(axis cs:100,6.3142811995931e-05)
--(axis cs:101,6.3142811995931e-05)
--(axis cs:102,6.3142811995931e-05)
--(axis cs:103,6.3142811995931e-05)
--(axis cs:104,6.3142811995931e-05)
--(axis cs:104,0.00035954947816208)
--(axis cs:104,0.00035954947816208)
--(axis cs:103,0.00035954947816208)
--(axis cs:102,0.00035954947816208)
--(axis cs:101,0.00035954947816208)
--(axis cs:100,0.00035954947816208)
--(axis cs:99,0.00035954947816208)
--(axis cs:98,0.000360805192030966)
--(axis cs:97,0.000360805192030966)
--(axis cs:96,0.000360805192030966)
--(axis cs:95,0.000360805192030966)
--(axis cs:94,0.000360805192030966)
--(axis cs:93,0.000360805192030966)
--(axis cs:92,0.000362990598659962)
--(axis cs:91,0.000362990598659962)
--(axis cs:90,0.000362990598659962)
--(axis cs:89,0.000362990598659962)
--(axis cs:88,0.000362990598659962)
--(axis cs:87,0.000362990598659962)
--(axis cs:86,0.000362990598659962)
--(axis cs:85,0.000362990598659962)
--(axis cs:84,0.000364214822184294)
--(axis cs:83,0.000364214822184294)
--(axis cs:82,0.000364214822184294)
--(axis cs:81,0.000364214822184294)
--(axis cs:80,0.000364214822184294)
--(axis cs:79,0.000364214822184294)
--(axis cs:78,0.000621171086095273)
--(axis cs:77,0.000860998756252229)
--(axis cs:76,0.000864162808284163)
--(axis cs:75,0.00298486580140889)
--(axis cs:74,0.00298486580140889)
--(axis cs:73,0.037811204791069)
--(axis cs:72,0.0413742922246456)
--(axis cs:71,0.0413835979998112)
--(axis cs:70,0.0413835979998112)
--(axis cs:69,0.0413835979998112)
--(axis cs:68,0.0413835979998112)
--(axis cs:67,0.0413835979998112)
--(axis cs:66,0.0413835979998112)
--(axis cs:65,0.0414061956107616)
--(axis cs:64,0.0414061956107616)
--(axis cs:63,0.0414061956107616)
--(axis cs:62,0.0414061956107616)
--(axis cs:61,0.0414061956107616)
--(axis cs:60,0.0415388122200966)
--(axis cs:59,0.0415388122200966)
--(axis cs:58,0.0415388122200966)
--(axis cs:57,0.0415655225515366)
--(axis cs:56,0.0417180545628071)
--(axis cs:55,0.041856050491333)
--(axis cs:54,0.041856050491333)
--(axis cs:53,0.0418751090764999)
--(axis cs:52,0.0418751090764999)
--(axis cs:51,0.0418751090764999)
--(axis cs:50,0.0418751090764999)
--(axis cs:49,0.0507742911577225)
--(axis cs:48,0.0527145564556122)
--(axis cs:47,0.0536310970783234)
--(axis cs:46,0.0599855333566666)
--(axis cs:45,0.0599855333566666)
--(axis cs:44,0.0939262062311172)
--(axis cs:43,0.111077502369881)
--(axis cs:42,0.141206562519073)
--(axis cs:41,0.141265168786049)
--(axis cs:40,0.141765803098679)
--(axis cs:39,0.141765803098679)
--(axis cs:38,0.141765832901001)
--(axis cs:37,0.187237456440926)
--(axis cs:36,0.187237456440926)
--(axis cs:35,0.187237456440926)
--(axis cs:34,0.187259554862976)
--(axis cs:33,0.187259554862976)
--(axis cs:32,0.187259554862976)
--(axis cs:31,0.187375396490097)
--(axis cs:30,0.187845230102539)
--(axis cs:29,0.188767999410629)
--(axis cs:28,0.188929736614227)
--(axis cs:27,0.191818505525589)
--(axis cs:26,0.191818505525589)
--(axis cs:25,0.193491160869598)
--(axis cs:24,0.194162383675575)
--(axis cs:23,0.212449252605438)
--(axis cs:22,0.212449252605438)
--(axis cs:21,0.212656617164612)
--(axis cs:20,0.212656617164612)
--(axis cs:19,0.214499592781067)
--(axis cs:18,0.216228097677231)
--(axis cs:17,0.216447561979294)
--(axis cs:16,0.218306392431259)
--(axis cs:15,0.222323089838028)
--(axis cs:14,0.222323089838028)
--(axis cs:13,0.268481701612473)
--(axis cs:12,0.297914326190948)
--(axis cs:11,0.303757071495056)
--(axis cs:10,0.3110032081604)
--(axis cs:9,0.336302101612091)
--(axis cs:8,0.360644191503525)
--(axis cs:7,0.410761028528214)
--(axis cs:6,0.492849826812744)
--(axis cs:5,0.635971784591675)
--(axis cs:4,0.642317116260529)
--(axis cs:3,0.693378448486328)
--cycle;

\path [draw=color1, fill=color1, opacity=0.3]
(axis cs:3,0.693378502433763)
--(axis cs:3,0.483341099689005)
--(axis cs:4,0.410583678123776)
--(axis cs:5,0.383913189865149)
--(axis cs:6,0.359070095037889)
--(axis cs:7,0.333768907391226)
--(axis cs:8,0.333768907391226)
--(axis cs:9,0.333768907391226)
--(axis cs:10,0.333768907391226)
--(axis cs:11,0.326650645803854)
--(axis cs:12,0.314123553463998)
--(axis cs:13,0.314123553463998)
--(axis cs:14,0.305788350141367)
--(axis cs:15,0.28751698853766)
--(axis cs:16,0.28751698853766)
--(axis cs:17,0.278384358364701)
--(axis cs:18,0.277787516190868)
--(axis cs:19,0.277787516190868)
--(axis cs:20,0.25655890897981)
--(axis cs:21,0.25655890897981)
--(axis cs:22,0.254855950946076)
--(axis cs:23,0.252668856842142)
--(axis cs:24,0.24612995188668)
--(axis cs:25,0.19742976362007)
--(axis cs:26,0.176794650136652)
--(axis cs:27,0.176707731131252)
--(axis cs:28,0.17645577471373)
--(axis cs:29,0.176307400573426)
--(axis cs:30,0.176302888557802)
--(axis cs:31,0.176302888557802)
--(axis cs:32,0.176280733355629)
--(axis cs:33,0.176280733355629)
--(axis cs:34,0.170631685757345)
--(axis cs:35,0.151247015218912)
--(axis cs:36,0.137404235654268)
--(axis cs:37,0.133754152982904)
--(axis cs:38,0.13368911311621)
--(axis cs:39,0.132913528482032)
--(axis cs:40,0.132913528482032)
--(axis cs:41,0.132792791661477)
--(axis cs:42,0.132792791661477)
--(axis cs:43,0.132792791661477)
--(axis cs:44,0.132782946390236)
--(axis cs:45,0.117046745095711)
--(axis cs:46,0.117046745095711)
--(axis cs:47,0.117032751995003)
--(axis cs:48,0.117010509306618)
--(axis cs:49,0.117010509306618)
--(axis cs:50,0.10336508041317)
--(axis cs:51,0.10336508041317)
--(axis cs:52,0.103334563892174)
--(axis cs:53,0.0872169985537977)
--(axis cs:54,0.0841808543135205)
--(axis cs:55,0.0841042549323822)
--(axis cs:56,0.0841042549323822)
--(axis cs:57,0.0840930019713902)
--(axis cs:58,0.0840820314575579)
--(axis cs:59,0.0840820164384157)
--(axis cs:60,0.0840820164384157)
--(axis cs:61,0.0840820164384157)
--(axis cs:62,0.0840815981336926)
--(axis cs:63,0.0635313182756221)
--(axis cs:64,0.0635313182756221)
--(axis cs:65,0.0383715717515767)
--(axis cs:66,0.0383715717515767)
--(axis cs:67,0.0383715717515767)
--(axis cs:68,0.0377225363857999)
--(axis cs:69,0.0377201768753027)
--(axis cs:70,0.0376727807797088)
--(axis cs:71,0.0376660248824457)
--(axis cs:72,0.0376660248824457)
--(axis cs:73,0.0376660248824457)
--(axis cs:74,0.0376660248824457)
--(axis cs:75,0.0376641607631189)
--(axis cs:76,0.0376641607631189)
--(axis cs:77,0.0376479426287543)
--(axis cs:78,0.0376478893941485)
--(axis cs:79,0.0376352410661341)
--(axis cs:80,0.0376352410661341)
--(axis cs:81,0.0376352410661341)
--(axis cs:82,0.0376290021768074)
--(axis cs:83,0.0376290021768074)
--(axis cs:84,0.037628020446236)
--(axis cs:85,0.037628020446236)
--(axis cs:86,0.037628020446236)
--(axis cs:87,0.037628020446236)
--(axis cs:88,0.0376106828742954)
--(axis cs:89,0.0376106828742954)
--(axis cs:90,0.0376106828742954)
--(axis cs:91,0.0376106828742954)
--(axis cs:92,0.0376106828742954)
--(axis cs:93,0.0376106828742954)
--(axis cs:94,0.0376106828742954)
--(axis cs:95,0.0376106828742954)
--(axis cs:96,0.0376106828742954)
--(axis cs:97,0.0376106828742954)
--(axis cs:98,0.0376103614630878)
--(axis cs:99,0.0376103614630878)
--(axis cs:100,0.0376103614630878)
--(axis cs:101,0.0376102158297487)
--(axis cs:102,0.0376096513879191)
--(axis cs:103,0.0376096513879191)
--(axis cs:104,0.0376096513879191)
--(axis cs:104,0.0989495621209605)
--(axis cs:104,0.0989495621209605)
--(axis cs:103,0.0989495621209605)
--(axis cs:102,0.0989495621209605)
--(axis cs:101,0.0989499027003801)
--(axis cs:100,0.0989499905733774)
--(axis cs:99,0.0989499905733774)
--(axis cs:98,0.0989499905733774)
--(axis cs:97,0.0989504221066231)
--(axis cs:96,0.0989504221066231)
--(axis cs:95,0.0989504221066231)
--(axis cs:94,0.0989504221066231)
--(axis cs:93,0.0989504221066231)
--(axis cs:92,0.0989504221066231)
--(axis cs:91,0.0989504221066231)
--(axis cs:90,0.0989504221066231)
--(axis cs:89,0.0989504221066231)
--(axis cs:88,0.0989504221066231)
--(axis cs:87,0.0989608856679402)
--(axis cs:86,0.0989608856679402)
--(axis cs:85,0.0989608856679402)
--(axis cs:84,0.0989608856679402)
--(axis cs:83,0.0989614783591047)
--(axis cs:82,0.0989614783591047)
--(axis cs:81,0.0989652456176156)
--(axis cs:80,0.0989652456176156)
--(axis cs:79,0.0989652456176156)
--(axis cs:78,0.0989728813149773)
--(axis cs:77,0.0989730401613087)
--(axis cs:76,0.0989828376859187)
--(axis cs:75,0.0989828376859187)
--(axis cs:74,0.0989839619629307)
--(axis cs:73,0.0989839619629307)
--(axis cs:72,0.0989839619629307)
--(axis cs:71,0.0989839619629307)
--(axis cs:70,0.098988045669167)
--(axis cs:69,0.0990167356194872)
--(axis cs:68,0.0990181596103972)
--(axis cs:67,0.0994166499792321)
--(axis cs:66,0.0994166499792321)
--(axis cs:65,0.0994166499792321)
--(axis cs:64,0.144038974180198)
--(axis cs:63,0.144038974180198)
--(axis cs:62,0.163709685986942)
--(axis cs:61,0.163709889352614)
--(axis cs:60,0.163709889352614)
--(axis cs:59,0.163709889352614)
--(axis cs:58,0.163709913262477)
--(axis cs:57,0.163720062670756)
--(axis cs:56,0.163725536332733)
--(axis cs:55,0.163725536332733)
--(axis cs:54,0.163775129796988)
--(axis cs:53,0.169144026988027)
--(axis cs:52,0.186381837336439)
--(axis cs:51,0.186486497063313)
--(axis cs:50,0.186486497063313)
--(axis cs:49,0.194413080374541)
--(axis cs:48,0.194413080374541)
--(axis cs:47,0.194421700007824)
--(axis cs:46,0.194432576726019)
--(axis cs:45,0.194432576726019)
--(axis cs:44,0.216451440988008)
--(axis cs:43,0.21645505207041)
--(axis cs:42,0.21645505207041)
--(axis cs:41,0.21645505207041)
--(axis cs:40,0.216499343491428)
--(axis cs:39,0.216499343491428)
--(axis cs:38,0.216785483475848)
--(axis cs:37,0.216862232557618)
--(axis cs:36,0.220714906064943)
--(axis cs:35,0.227421677903895)
--(axis cs:34,0.243363753026858)
--(axis cs:33,0.25329587251186)
--(axis cs:32,0.25329587251186)
--(axis cs:31,0.253316918639023)
--(axis cs:30,0.253316918639023)
--(axis cs:29,0.253326699426655)
--(axis cs:28,0.253528336340907)
--(axis cs:27,0.254034295196113)
--(axis cs:26,0.254116855762404)
--(axis cs:25,0.292993308142531)
--(axis cs:24,0.342196521349317)
--(axis cs:23,0.347212973781894)
--(axis cs:22,0.348912671107017)
--(axis cs:21,0.349328643436047)
--(axis cs:20,0.349328643436047)
--(axis cs:19,0.380683250911543)
--(axis cs:18,0.380683250911543)
--(axis cs:17,0.381651966653318)
--(axis cs:16,0.401303541320672)
--(axis cs:15,0.401303541320672)
--(axis cs:14,0.415139046141717)
--(axis cs:13,0.435381911796229)
--(axis cs:12,0.435381911796229)
--(axis cs:11,0.469789014443473)
--(axis cs:10,0.471671221384892)
--(axis cs:9,0.471671221384892)
--(axis cs:8,0.471671221384892)
--(axis cs:7,0.471671221384892)
--(axis cs:6,0.491308140995376)
--(axis cs:5,0.555520927462133)
--(axis cs:4,0.602929311166228)
--(axis cs:3,0.693378502433763)
--cycle;

\addplot [semithick, color1, dash dot]
table {%
3 0.588359801061384
4 0.521568743913937
5 0.447066392908177
6 0.435015659834629
7 0.382711119953064
8 0.368105628097301
9 0.363850584034674
10 0.363850584034674
11 0.326569022134189
12 0.326569022134189
13 0.272739639289395
14 0.237278217936092
15 0.236958908902628
16 0.21810184072717
17 0.21810184072717
18 0.217568834745321
19 0.217568834745321
20 0.21370526543345
21 0.198079857327458
22 0.198079857327458
23 0.197225726293449
24 0.19528192367511
25 0.163830881899576
26 0.159937365077765
27 0.159937365077765
28 0.154121341119717
29 0.154121341119717
30 0.153998234640664
31 0.0937756952935424
32 0.0673345572183918
33 0.0556166039333779
34 0.0556166039333779
35 0.0389500941470191
36 0.0349544242917963
37 0.0332363718655052
38 0.0332363718655052
39 0.0332363718655052
40 0.0313938234758787
41 0.0311654571739793
42 0.0309796140585073
43 0.0309796140585073
44 0.0303335910122416
45 0.0285217185289795
46 0.0285217185289795
47 0.0285217185289795
48 0.0281610673936188
49 0.0218517166743498
50 0.0218517166743498
51 0.0217481256945091
52 0.0216938074158273
53 0.0216927624870134
54 0.0215795571206417
55 0.0207380079922415
56 0.0207380079922415
57 0.0207380079922415
58 0.0207071153162633
59 0.0207071153162633
60 0.0207071153162633
61 0.0207071153162633
62 0.0206727169312695
63 0.0206727169312695
64 0.0206727169312695
65 0.0206727169312695
66 0.0206721037599139
67 0.0206721037599139
68 0.0206721037599139
69 0.0206696372385612
70 0.0206696372385612
71 0.0206696372385612
72 0.0206696372385612
73 0.0206626292943686
74 0.0206626292943686
75 0.0206626292943686
76 0.0206626292943686
77 0.0206626292943686
78 0.02065941613928
79 0.02065941613928
80 0.02065941613928
81 0.02065941613928
82 0.02065941613928
83 0.02065941613928
84 0.02065941613928
85 0.020656084399279
86 0.020656084399279
87 0.020656084399279
88 0.020641845082173
89 0.020627087045542
90 0.0206159850341159
91 0.0206159850341159
92 0.0206152168616992
93 0.0206144860622834
94 0.0206144860622834
95 0.0206144860622834
96 0.0206144860622834
97 0.0206144860622834
98 0.0206144860622834
99 0.020597347791918
100 0.020597347791918
101 0.020597347791918
102 0.020597347791918
103 0.020597347791918
104 0.020597347791918
};
\addplot [semithick, blue]
table {%
3 0.588359832763672
4 0.491075217723846
5 0.459051072597504
6 0.433527082204819
7 0.401970207691193
8 0.401970207691193
9 0.376095414161682
10 0.361540734767914
11 0.342416316270828
12 0.305616825819016
13 0.305429339408875
14 0.265260249376297
15 0.262543827295303
16 0.261151015758514
17 0.259913325309753
18 0.259913325309753
19 0.236019775271416
20 0.215649291872978
21 0.215649291872978
22 0.207143858075142
23 0.194547563791275
24 0.193139716982841
25 0.193139716982841
26 0.190872400999069
27 0.183766841888428
28 0.149782344698906
29 0.136800482869148
30 0.136800482869148
31 0.136800482869148
32 0.135856628417969
33 0.135485842823982
34 0.101081654429436
35 0.0997341796755791
36 0.0979333370923996
37 0.0628964453935623
38 0.0628964453935623
39 0.0628964453935623
40 0.0628964453935623
41 0.0437791645526886
42 0.0434850752353668
43 0.0430388860404491
44 0.0417122654616833
45 0.0417122654616833
46 0.0415302515029907
47 0.0415302515029907
48 0.0415302515029907
49 0.0415302515029907
50 0.0415302515029907
51 0.0415302515029907
52 0.0415302515029907
53 0.0415302515029907
54 0.0415302515029907
55 0.0415302515029907
56 0.0415302515029907
57 0.0415302515029907
58 0.0415302515029907
59 0.0415302515029907
60 0.0413153879344463
61 0.0413153879344463
62 0.0413153879344463
63 0.0413153879344463
64 0.0412918701767921
65 0.0412918701767921
66 0.0412918701767921
67 0.0412915870547295
68 0.0412915870547295
69 0.0412915870547295
70 0.0412915870547295
71 0.0412915870547295
72 0.0412915870547295
73 0.0412915870547295
74 0.0412915870547295
75 0.0412915870547295
76 0.0412915870547295
77 0.0412257425487041
78 0.0206508096307516
79 0.0206508096307516
80 0.0206508096307516
81 0.0179854445159435
82 0.0179854277521372
83 0.0179854277521372
84 0.0179687924683094
85 0.0179687924683094
86 0.0179687924683094
87 0.00424433965235949
88 0.00424433965235949
89 0.00424433965235949
90 9.65476065175608e-05
91 9.65476065175608e-05
92 9.65476065175608e-05
93 9.65476065175608e-05
94 9.65476065175608e-05
95 9.25600543268956e-05
96 9.25600543268956e-05
97 9.25600543268956e-05
98 8.52763623697683e-05
99 8.52763623697683e-05
100 8.52763623697683e-05
101 8.52763623697683e-05
102 8.52763623697683e-05
103 8.52763623697683e-05
104 8.52763623697683e-05
};
\addplot [semithick, blue, dash dot]
table {%
3 0.588359773159027
4 0.539065659046173
5 0.53231418132782
6 0.411756038665771
7 0.355220943689346
8 0.301939725875854
9 0.279447436332703
10 0.257559955120087
11 0.247294694185257
12 0.241321370005608
13 0.208105757832527
14 0.164563804864883
15 0.164563804864883
16 0.160980418324471
17 0.159315079450607
18 0.158999249339104
19 0.157892152667046
20 0.156734228134155
21 0.156734228134155
22 0.156437039375305
23 0.156437039375305
24 0.136973932385445
25 0.136094063520432
26 0.135051280260086
27 0.135051280260086
28 0.131405755877495
29 0.13119001686573
30 0.130611106753349
31 0.129993304610252
32 0.129839360713959
33 0.129839360713959
34 0.129839360713959
35 0.129810780286789
36 0.129810780286789
37 0.129810780286789
38 0.092468336224556
39 0.0924683213233948
40 0.0924683213233948
41 0.0918451994657516
42 0.0917714983224869
43 0.0737401098012924
44 0.0595728270709515
45 0.034719031304121
46 0.034719031304121
47 0.0284086521714926
48 0.0279495231807232
49 0.0258705914020538
50 0.0213255826383829
51 0.0213255826383829
52 0.0213255826383829
53 0.0213255826383829
54 0.0213160514831543
55 0.0213160514831543
56 0.0211609415709972
57 0.0209896508604288
58 0.0209595616906881
59 0.0209595616906881
60 0.0209595616906881
61 0.0208120290189981
62 0.0208120290189981
63 0.0208120290189981
64 0.0208120290189981
65 0.0208120290189981
66 0.0207865945994854
67 0.0207865945994854
68 0.0207865945994854
69 0.0207865945994854
70 0.0207865945994854
71 0.0207865945994854
72 0.020776117220521
73 0.0189945511519909
74 0.00157811644021422
75 0.00157811644021422
76 0.000506854034028947
77 0.00050315260887146
78 0.000375455623725429
79 0.000216823813389055
80 0.000216823813389055
81 0.000216823813389055
82 0.000216823813389055
83 0.000216823813389055
84 0.000216823813389055
85 0.00021537541761063
86 0.00021537541761063
87 0.00021537541761063
88 0.00021537541761063
89 0.00021537541761063
90 0.00021537541761063
91 0.00021537541761063
92 0.00021537541761063
93 0.000212818384170532
94 0.000212818384170532
95 0.000212818384170532
96 0.000212818384170532
97 0.000212818384170532
98 0.000212818384170532
99 0.000211346152354963
100 0.000211346152354963
101 0.000211346152354963
102 0.000211346152354963
103 0.000211346152354963
104 0.000211346152354963
};
\addplot [semithick, color1]
table {%
3 0.588359801061384
4 0.506756494645002
5 0.469717058663641
6 0.425189118016632
7 0.402720064388059
8 0.402720064388059
9 0.402720064388059
10 0.402720064388059
11 0.398219830123664
12 0.374752732630114
13 0.374752732630114
14 0.360463698141542
15 0.344410264929166
16 0.344410264929166
17 0.33001816250901
18 0.329235383551205
19 0.329235383551205
20 0.302943776207929
21 0.302943776207929
22 0.301884311026546
23 0.299940915312018
24 0.294163236617999
25 0.2452115358813
26 0.215455752949528
27 0.215371013163682
28 0.214992055527318
29 0.214817050000041
30 0.214809903598413
31 0.214809903598413
32 0.214788302933745
33 0.214788302933745
34 0.206997719392101
35 0.189334346561403
36 0.179059570859605
37 0.175308192770261
38 0.175237298296029
39 0.17470643598673
40 0.17470643598673
41 0.174623921865943
42 0.174623921865943
43 0.174623921865943
44 0.174617193689122
45 0.155739660910865
46 0.155739660910865
47 0.155727226001413
48 0.155711794840579
49 0.155711794840579
50 0.144925788738241
51 0.144925788738241
52 0.144858200614307
53 0.128180512770912
54 0.123977992055254
55 0.123914895632558
56 0.123914895632558
57 0.123906532321073
58 0.123895972360017
59 0.123895952895515
60 0.123895952895515
61 0.123895952895515
62 0.123895642060317
63 0.10378514622791
64 0.10378514622791
65 0.0688941108654044
66 0.0688941108654044
67 0.0688941108654044
68 0.0683703479980985
69 0.068368456247395
70 0.0683304132244379
71 0.0683249934226882
72 0.0683249934226882
73 0.0683249934226882
74 0.0683249934226882
75 0.0683234992245188
76 0.0683234992245188
77 0.0683104913950315
78 0.0683103853545629
79 0.0683002433418748
80 0.0683002433418748
81 0.0683002433418748
82 0.068295240267956
83 0.068295240267956
84 0.0682944530570881
85 0.0682944530570881
86 0.0682944530570881
87 0.0682944530570881
88 0.0682805524904593
89 0.0682805524904593
90 0.0682805524904593
91 0.0682805524904593
92 0.0682805524904593
93 0.0682805524904593
94 0.0682805524904593
95 0.0682805524904593
96 0.0682805524904593
97 0.0682805524904593
98 0.0682801760182326
99 0.0682801760182326
100 0.0682801760182326
101 0.0682800592650644
102 0.0682796067544398
103 0.0682796067544398
104 0.0682796067544398
};
\end{axis}

\end{tikzpicture}

%% file: figures/bop_1d_negeasom_weak.tex
\begin{tikzpicture}

\definecolor{color0}{rgb}{0,0,1}
\definecolor{color1}{rgb}{1,0.549019607843137,0}
\definecolor{color2}{rgb}{1,0.647058823529412,0}
\definecolor{color3}{rgb}{0.564705882352941,0.933333333333333,0.564705882352941}

\begin{axis}[axis on top,
enlarge x limits=false,
enlarge y limits=false,
height=\figureheight,
scale only axis,
tick align=outside,
tick pos=left,
tick pos=left,
width=\figurewidth,
xmin=3, xmax=50,
xtick style={color=black},
xtick={-10,0,10,25,50,75,100},
xticklabels={\ensuremath{-}10,0,10,25,50,75,90},
ymin=-0.03, ymax=0.8,
ytick style={color=black},
ytick={0.   , 0.8},
]
\node[anchor=north east] at (rel axis cs:1,1) {Negeasom 1D (weak)};
\path [draw=blue, fill=blue, opacity=0.3]
(axis cs:3,0.75057637691498)
--(axis cs:3,0.49599277973175)
--(axis cs:4,0.48037099838257)
--(axis cs:5,0.35109281539917)
--(axis cs:6,0.20972013473511)
--(axis cs:7,0.20972013473511)
--(axis cs:8,0.1598504781723)
--(axis cs:9,0.13167405128479)
--(axis cs:10,0.02093529701233)
--(axis cs:11,0.01737773418427)
--(axis cs:12,0.01504349708557)
--(axis cs:13,0.00560021400452)
--(axis cs:14,0.0017306804657)
--(axis cs:15,0.00090730190277)
--(axis cs:16,0.00090730190277)
--(axis cs:17,0.00090730190277)
--(axis cs:18,0.00030839443207)
--(axis cs:19,0.0001323223114)
--(axis cs:20,0.00008749961853)
--(axis cs:21,0.00008749961853)
--(axis cs:22,0.00008749961853)
--(axis cs:23,0.00008749961853)
--(axis cs:24,0.00008749961853)
--(axis cs:25,0.00008749961853)
--(axis cs:26,0.00008296966553)
--(axis cs:27,0.00007057189941)
--(axis cs:28,0.00007057189941)
--(axis cs:29,0.00007057189941)
--(axis cs:30,0.00007021427155)
--(axis cs:31,0.00007021427155)
--(axis cs:32,0.00006937980652)
--(axis cs:33,0.00005543231964)
--(axis cs:34,0.00004029273987)
--(axis cs:35,0.00004029273987)
--(axis cs:36,0.00004029273987)
--(axis cs:37,0.00004029273987)
--(axis cs:38,0.00004029273987)
--(axis cs:39,0.00004029273987)
--(axis cs:40,0.00004029273987)
--(axis cs:41,0.00003290176392)
--(axis cs:42,0.00003290176392)
--(axis cs:43,0.00003290176392)
--(axis cs:44,0.00003290176392)
--(axis cs:45,0.00003290176392)
--(axis cs:46,0.00003290176392)
--(axis cs:47,0.00003290176392)
--(axis cs:48,0.00003290176392)
--(axis cs:49,0.00003290176392)
--(axis cs:50,0.00003135204315)
--(axis cs:51,0.00002872943878)
--(axis cs:52,0.00002634525299)
--(axis cs:53,0.00002634525299)
--(axis cs:54,0.00002634525299)
--(axis cs:55,0.0000251531601)
--(axis cs:56,0.0000251531601)
--(axis cs:57,0.00002324581146)
--(axis cs:58,0.00002324581146)
--(axis cs:59,0.00002324581146)
--(axis cs:60,0.00001645088196)
--(axis cs:61,0.00001645088196)
--(axis cs:62,0.00001645088196)
--(axis cs:63,0.00001561641693)
--(axis cs:64,0.00001561641693)
--(axis cs:65,0.00001561641693)
--(axis cs:66,0.00001561641693)
--(axis cs:67,0.00001561641693)
--(axis cs:68,0.00001561641693)
--(axis cs:69,0.00001561641693)
--(axis cs:70,0.00001561641693)
--(axis cs:71,0.00001394748688)
--(axis cs:72,0.00001394748688)
--(axis cs:73,0.00001394748688)
--(axis cs:74,0.00001394748688)
--(axis cs:75,0.00001394748688)
--(axis cs:76,0.00001394748688)
--(axis cs:77,0.00001215934753)
--(axis cs:78,0.00000727176666)
--(axis cs:79,0.00000727176666)
--(axis cs:80,0.00000727176666)
--(axis cs:81,0.00000727176666)
--(axis cs:82,0.00000727176666)
--(axis cs:83,0.00000727176666)
--(axis cs:84,0.00000727176666)
--(axis cs:85,0.00000727176666)
--(axis cs:86,0.00000727176666)
--(axis cs:87,0.00000727176666)
--(axis cs:88,0.00000727176666)
--(axis cs:89,0.00000727176666)
--(axis cs:90,0.00000643730164)
--(axis cs:91,0.00000643730164)
--(axis cs:92,0.00000643730164)
--(axis cs:93,0.00000643730164)
--(axis cs:94,0.00000643730164)
--(axis cs:95,0.00000643730164)
--(axis cs:96,0.00000643730164)
--(axis cs:97,0.00000655651093)
--(axis cs:98,0.00000655651093)
--(axis cs:99,0.00000524520874)
--(axis cs:100,0.00000524520874)
--(axis cs:101,0.00000524520874)
--(axis cs:102,0.00000524520874)
--(axis cs:103,0.00000524520874)
--(axis cs:104,0.00000524520874)
--(axis cs:104,0.00001502037048)
--(axis cs:104,0.00001502037048)
--(axis cs:103,0.00001502037048)
--(axis cs:102,0.00001502037048)
--(axis cs:101,0.00001502037048)
--(axis cs:100,0.00001502037048)
--(axis cs:99,0.00001502037048)
--(axis cs:98,0.0000194311142)
--(axis cs:97,0.0000194311142)
--(axis cs:96,0.00001931190491)
--(axis cs:95,0.00001931190491)
--(axis cs:94,0.00001931190491)
--(axis cs:93,0.00001931190491)
--(axis cs:92,0.00001931190491)
--(axis cs:91,0.00001931190491)
--(axis cs:90,0.00001931190491)
--(axis cs:89,0.00001990795135)
--(axis cs:88,0.00001990795135)
--(axis cs:87,0.00001990795135)
--(axis cs:86,0.00001990795135)
--(axis cs:85,0.00001990795135)
--(axis cs:84,0.00001990795135)
--(axis cs:83,0.00001990795135)
--(axis cs:82,0.00001990795135)
--(axis cs:81,0.00001990795135)
--(axis cs:80,0.00001990795135)
--(axis cs:79,0.00001990795135)
--(axis cs:78,0.00001990795135)
--(axis cs:77,0.00002861022949)
--(axis cs:76,0.0000296831131)
--(axis cs:75,0.0000296831131)
--(axis cs:74,0.0000296831131)
--(axis cs:73,0.0000296831131)
--(axis cs:72,0.0000296831131)
--(axis cs:71,0.0000296831131)
--(axis cs:70,0.00003087520599)
--(axis cs:69,0.00003087520599)
--(axis cs:68,0.00003087520599)
--(axis cs:67,0.00003087520599)
--(axis cs:66,0.00003087520599)
--(axis cs:65,0.00003087520599)
--(axis cs:64,0.00003087520599)
--(axis cs:63,0.00003087520599)
--(axis cs:62,0.00003361701965)
--(axis cs:61,0.00003361701965)
--(axis cs:60,0.00003361701965)
--(axis cs:59,0.00011456012726)
--(axis cs:58,0.00011456012726)
--(axis cs:57,0.00011456012726)
--(axis cs:56,0.00011622905731)
--(axis cs:55,0.00011622905731)
--(axis cs:54,0.00011718273163)
--(axis cs:53,0.00011718273163)
--(axis cs:52,0.00011718273163)
--(axis cs:51,0.00011956691742)
--(axis cs:50,0.00012147426605)
--(axis cs:49,0.00012254714966)
--(axis cs:48,0.00012254714966)
--(axis cs:47,0.00012254714966)
--(axis cs:46,0.00012254714966)
--(axis cs:45,0.00012254714966)
--(axis cs:44,0.00012254714966)
--(axis cs:43,0.00012254714966)
--(axis cs:42,0.00012254714966)
--(axis cs:41,0.00012254714966)
--(axis cs:40,0.0001323223114)
--(axis cs:39,0.0001323223114)
--(axis cs:38,0.0001323223114)
--(axis cs:37,0.0001323223114)
--(axis cs:36,0.0001323223114)
--(axis cs:35,0.0001323223114)
--(axis cs:34,0.0001323223114)
--(axis cs:33,0.00014889240265)
--(axis cs:32,0.00016140937805)
--(axis cs:31,0.0001620054245)
--(axis cs:30,0.0001620054245)
--(axis cs:29,0.00016260147095)
--(axis cs:28,0.00016260147095)
--(axis cs:27,0.00016260147095)
--(axis cs:26,0.00018906593323)
--(axis cs:25,0.00019240379333)
--(axis cs:24,0.00019240379333)
--(axis cs:23,0.00019240379333)
--(axis cs:22,0.00019240379333)
--(axis cs:21,0.00019240379333)
--(axis cs:20,0.00019240379333)
--(axis cs:19,0.00026488304138)
--(axis cs:18,0.00089991092682)
--(axis cs:17,0.00281465053558)
--(axis cs:16,0.00281465053558)
--(axis cs:15,0.00281465053558)
--(axis cs:14,0.04126811027527)
--(axis cs:13,0.12788009643555)
--(axis cs:12,0.17028045654297)
--(axis cs:11,0.1721967458725)
--(axis cs:10,0.17496657371521)
--(axis cs:9,0.358238697052)
--(axis cs:8,0.38286602497101)
--(axis cs:7,0.45791339874268)
--(axis cs:6,0.45791339874268)
--(axis cs:5,0.59837555885315)
--(axis cs:4,0.74207234382629)
--(axis cs:3,0.75057637691498)
--cycle;

\path [draw=color1, fill=color1, opacity=0.3]
(axis cs:3,0.75057640231309)
--(axis cs:3,0.49599281253425)
--(axis cs:4,0.49599281253425)
--(axis cs:5,0.47846499388114)
--(axis cs:6,0.40295814989952)
--(axis cs:7,0.40295810155445)
--(axis cs:8,0.34021333595686)
--(axis cs:9,0.22625149435831)
--(axis cs:10,0.15647605471979)
--(axis cs:11,0.15433052959388)
--(axis cs:12,0.15156556090446)
--(axis cs:13,0.13911924405787)
--(axis cs:14,0.13739274399976)
--(axis cs:15,0.13112049917378)
--(axis cs:16,0.0732474755072)
--(axis cs:17,0.03712593825886)
--(axis cs:18,0.03375379981611)
--(axis cs:19,0.03093772466187)
--(axis cs:20,0.00394488519017)
--(axis cs:21,0.00020483505236)
--(axis cs:22,0.0000621662032)
--(axis cs:23,0.0000621662032)
--(axis cs:24,0.0000621662032)
--(axis cs:25,0.00006164134599)
--(axis cs:26,0.00002598339655)
--(axis cs:27,0.00002598339655)
--(axis cs:28,0.00002598339655)
--(axis cs:29,0.00002598339655)
--(axis cs:30,0.00001955229549)
--(axis cs:31,0.00001955229549)
--(axis cs:32,0.00001955229549)
--(axis cs:33,0.00000631922492)
--(axis cs:34,0.00000624634949)
--(axis cs:35,0.00000624634949)
--(axis cs:36,0.0000046691643)
--(axis cs:37,0.0000046691643)
--(axis cs:38,0.0000046691643)
--(axis cs:39,0.0000046691643)
--(axis cs:40,0.0000046691643)
--(axis cs:41,0.0000046691643)
--(axis cs:42,0.00000410340917)
--(axis cs:43,0.00000346925105)
--(axis cs:44,0.00000346925105)
--(axis cs:45,0.00000346925105)
--(axis cs:46,0.00000346925105)
--(axis cs:47,0.00000346925105)
--(axis cs:48,0.00000346925105)
--(axis cs:49,0.00000346925105)
--(axis cs:50,0.00000346925105)
--(axis cs:51,0.00000346925105)
--(axis cs:52,0.00000346925105)
--(axis cs:53,0.00000346925105)
--(axis cs:54,0.00000346925105)
--(axis cs:55,0.00000346925105)
--(axis cs:56,0.00000346925105)
--(axis cs:57,0.00000336073469)
--(axis cs:58,0.00000336073469)
--(axis cs:59,0.00000326708533)
--(axis cs:60,0.00000326708533)
--(axis cs:61,0.00000240340173)
--(axis cs:62,0.00000240340173)
--(axis cs:63,0.00000240340173)
--(axis cs:64,0.00000240340173)
--(axis cs:65,0.00000240340173)
--(axis cs:66,0.00000168407654)
--(axis cs:67,0.00000168407654)
--(axis cs:68,0.00000168407654)
--(axis cs:69,0.00000168407654)
--(axis cs:70,0.00000168407654)
--(axis cs:71,0.00000168407654)
--(axis cs:72,0.00000168407654)
--(axis cs:73,0.00000168407654)
--(axis cs:74,0.00000168407654)
--(axis cs:75,0.00000168407654)
--(axis cs:76,0.00000168407654)
--(axis cs:77,0.00000168407654)
--(axis cs:78,0.00000168407654)
--(axis cs:79,0.00000168407654)
--(axis cs:80,0.00000168407654)
--(axis cs:81,0.00000168407654)
--(axis cs:82,0.00000168407654)
--(axis cs:83,0.00000168407654)
--(axis cs:84,0.00000168407654)
--(axis cs:85,0.00000168407654)
--(axis cs:86,0.00000168407654)
--(axis cs:87,0.00000168407654)
--(axis cs:88,0.00000168407654)
--(axis cs:89,0.00000168407654)
--(axis cs:90,0.00000161316609)
--(axis cs:91,0.00000161316609)
--(axis cs:92,0.00000161316609)
--(axis cs:93,0.00000161316609)
--(axis cs:94,0.00000161316609)
--(axis cs:95,0.00000161316609)
--(axis cs:96,0.00000161316609)
--(axis cs:97,0.00000161316609)
--(axis cs:98,0.00000161316609)
--(axis cs:99,0.00000161316609)
--(axis cs:100,0.00000161316609)
--(axis cs:101,0.00000161316609)
--(axis cs:102,0.00000161316609)
--(axis cs:103,0.00000161316609)
--(axis cs:104,0.00000110859794)
--(axis cs:104,0.20000066195931)
--(axis cs:104,0.20000066195931)
--(axis cs:103,0.20000106561503)
--(axis cs:102,0.20000106561503)
--(axis cs:101,0.20000106561503)
--(axis cs:100,0.20000106561503)
--(axis cs:99,0.20000106561503)
--(axis cs:98,0.20000106561503)
--(axis cs:97,0.20000106561503)
--(axis cs:96,0.20000106561503)
--(axis cs:95,0.20000106561503)
--(axis cs:94,0.20000106561503)
--(axis cs:93,0.20000106561503)
--(axis cs:92,0.20000106561503)
--(axis cs:91,0.20000106561503)
--(axis cs:90,0.20000106561503)
--(axis cs:89,0.20000112234327)
--(axis cs:88,0.20000112234327)
--(axis cs:87,0.20000112234327)
--(axis cs:86,0.20000112234327)
--(axis cs:85,0.20000112234327)
--(axis cs:84,0.20000112234327)
--(axis cs:83,0.20000112234327)
--(axis cs:82,0.20000112234327)
--(axis cs:81,0.20000112234327)
--(axis cs:80,0.20000112234327)
--(axis cs:79,0.20000112234327)
--(axis cs:78,0.20000112234327)
--(axis cs:77,0.20000112234327)
--(axis cs:76,0.20000112234327)
--(axis cs:75,0.20000112234327)
--(axis cs:74,0.20000112234327)
--(axis cs:73,0.20000112234327)
--(axis cs:72,0.20000112234327)
--(axis cs:71,0.20000112234327)
--(axis cs:70,0.20000112234327)
--(axis cs:69,0.20000112234327)
--(axis cs:68,0.20000112234327)
--(axis cs:67,0.20000112234327)
--(axis cs:66,0.20000112234327)
--(axis cs:65,0.20000169780523)
--(axis cs:64,0.20000169780523)
--(axis cs:63,0.20000169780523)
--(axis cs:62,0.20000169780523)
--(axis cs:61,0.20000169780523)
--(axis cs:60,0.20000238876144)
--(axis cs:59,0.20000238876144)
--(axis cs:58,0.20000246368045)
--(axis cs:57,0.20000246368045)
--(axis cs:56,0.20000255049307)
--(axis cs:55,0.20000255049307)
--(axis cs:54,0.20000255049307)
--(axis cs:53,0.20000255049307)
--(axis cs:52,0.20000255049307)
--(axis cs:51,0.20000255049307)
--(axis cs:50,0.20000255049307)
--(axis cs:49,0.20000255049307)
--(axis cs:48,0.20000255049307)
--(axis cs:47,0.20000255049307)
--(axis cs:46,0.20000255049307)
--(axis cs:45,0.20000255049307)
--(axis cs:44,0.20000255049307)
--(axis cs:43,0.20000255049307)
--(axis cs:42,0.20000305782641)
--(axis cs:41,0.20000351042889)
--(axis cs:40,0.20000351042889)
--(axis cs:39,0.20000362446245)
--(axis cs:38,0.20000362446245)
--(axis cs:37,0.20000366942747)
--(axis cs:36,0.20000366942747)
--(axis cs:35,0.2000049312164)
--(axis cs:34,0.2000049312164)
--(axis cs:33,0.20000498951684)
--(axis cs:32,0.20001557772503)
--(axis cs:31,0.20001557772503)
--(axis cs:30,0.20001557772503)
--(axis cs:29,0.20002072267592)
--(axis cs:28,0.20002072267592)
--(axis cs:27,0.20002072267592)
--(axis cs:26,0.20002072267592)
--(axis cs:25,0.20004926023242)
--(axis cs:24,0.20004971757993)
--(axis cs:23,0.20004971757993)
--(axis cs:22,0.20004971757993)
--(axis cs:21,0.2001639050452)
--(axis cs:20,0.20326420136901)
--(axis cs:19,0.23419058056049)
--(axis cs:18,0.23634121759373)
--(axis cs:17,0.23906066230932)
--(axis cs:16,0.33821627436062)
--(axis cs:15,0.41111394083843)
--(axis cs:14,0.42347284969537)
--(axis cs:13,0.42459396407535)
--(axis cs:12,0.4555398967511)
--(axis cs:11,0.45736804357582)
--(axis cs:10,0.45874012799639)
--(axis cs:9,0.51121534318654)
--(axis cs:8,0.60355076171553)
--(axis cs:7,0.66333711477435)
--(axis cs:6,0.66333722716102)
--(axis cs:5,0.74172989544131)
--(axis cs:4,0.75057640231309)
--(axis cs:3,0.75057640231309)
--cycle;

\path [draw=blue, fill=blue, opacity=0.3]
(axis cs:3,0.75057637691498)
--(axis cs:3,0.49599277973175)
--(axis cs:4,0.49599277973175)
--(axis cs:5,0.37356269359589)
--(axis cs:6,0.3366893529892)
--(axis cs:7,0.18771195411682)
--(axis cs:8,0.16883480548859)
--(axis cs:9,0.08972311019897)
--(axis cs:10,0.02415227890015)
--(axis cs:11,0.00522887706757)
--(axis cs:12,0.00423586368561)
--(axis cs:13,0.00423586368561)
--(axis cs:14,0.0023056268692)
--(axis cs:15,0.00150156021118)
--(axis cs:16,0.00077974796295)
--(axis cs:17,0.00062954425812)
--(axis cs:18,0.00062954425812)
--(axis cs:19,0.00042486190796)
--(axis cs:20,0.00028848648071)
--(axis cs:21,0.00018584728241)
--(axis cs:22,0.00018584728241)
--(axis cs:23,0.00017607212067)
--(axis cs:24,0.00016176700592)
--(axis cs:25,0.00016176700592)
--(axis cs:26,0.00016176700592)
--(axis cs:27,0.00016176700592)
--(axis cs:28,0.0001220703125)
--(axis cs:29,0.00007951259613)
--(axis cs:30,0.00007295608521)
--(axis cs:31,0.0000456571579)
--(axis cs:32,0.00003719329834)
--(axis cs:33,0.00003719329834)
--(axis cs:34,0.00003635883331)
--(axis cs:35,0.00003588199615)
--(axis cs:36,0.00002992153168)
--(axis cs:37,0.00002992153168)
--(axis cs:38,0.00002992153168)
--(axis cs:39,0.00002992153168)
--(axis cs:40,0.00002992153168)
--(axis cs:41,0.00002479553223)
--(axis cs:42,0.00001835823059)
--(axis cs:43,0.00001835823059)
--(axis cs:44,0.00001835823059)
--(axis cs:45,0.00001835823059)
--(axis cs:46,0.00001835823059)
--(axis cs:47,0.00001645088196)
--(axis cs:48,0.00001645088196)
--(axis cs:49,0.00001645088196)
--(axis cs:50,0.00001406669617)
--(axis cs:51,0.00001406669617)
--(axis cs:52,0.00000596046448)
--(axis cs:53,0.00000596046448)
--(axis cs:54,0.00000548362732)
--(axis cs:55,0.00000548362732)
--(axis cs:56,0.00000548362732)
--(axis cs:57,0.00000548362732)
--(axis cs:58,0.00000548362732)
--(axis cs:59,0.00000548362732)
--(axis cs:60,0.00000548362732)
--(axis cs:61,0.00000548362732)
--(axis cs:62,0.00000417232513)
--(axis cs:63,0.00000369548798)
--(axis cs:64,0.00000369548798)
--(axis cs:65,0.00000369548798)
--(axis cs:66,0.00000369548798)
--(axis cs:67,0.00000369548798)
--(axis cs:68,0.00000298023224)
--(axis cs:69,0.00000298023224)
--(axis cs:70,0.00000298023224)
--(axis cs:71,0.00000298023224)
--(axis cs:72,0.00000298023224)
--(axis cs:73,0.00000298023224)
--(axis cs:74,0.00000274181366)
--(axis cs:75,0.00000274181366)
--(axis cs:76,0.00000274181366)
--(axis cs:77,0.00000274181366)
--(axis cs:78,0.00000274181366)
--(axis cs:79,0.00000274181366)
--(axis cs:80,0.00000274181366)
--(axis cs:81,0.00000274181366)
--(axis cs:82,0.00000274181366)
--(axis cs:83,0.00000274181366)
--(axis cs:84,0.00000274181366)
--(axis cs:85,0.00000166893005)
--(axis cs:86,0.00000166893005)
--(axis cs:87,0.00000166893005)
--(axis cs:88,0.00000166893005)
--(axis cs:89,0.00000166893005)
--(axis cs:90,0.00000166893005)
--(axis cs:91,0.00000166893005)
--(axis cs:92,0.00000166893005)
--(axis cs:93,0.00000166893005)
--(axis cs:94,0.00000131130219)
--(axis cs:95,0.00000131130219)
--(axis cs:96,0.00000131130219)
--(axis cs:97,0.00000131130219)
--(axis cs:98,0.00000131130219)
--(axis cs:99,0.00000131130219)
--(axis cs:100,0.00000131130219)
--(axis cs:101,0.00000131130219)
--(axis cs:102,0.00000131130219)
--(axis cs:103,0.00000131130219)
--(axis cs:104,0.00000131130219)
--(axis cs:104,0.00000393390656)
--(axis cs:104,0.00000393390656)
--(axis cs:103,0.00000393390656)
--(axis cs:102,0.00000393390656)
--(axis cs:101,0.00000393390656)
--(axis cs:100,0.00000393390656)
--(axis cs:99,0.00000393390656)
--(axis cs:98,0.00000393390656)
--(axis cs:97,0.00000393390656)
--(axis cs:96,0.00000393390656)
--(axis cs:95,0.00000393390656)
--(axis cs:94,0.00000393390656)
--(axis cs:93,0.00000524520874)
--(axis cs:92,0.00000524520874)
--(axis cs:91,0.00000524520874)
--(axis cs:90,0.00000524520874)
--(axis cs:89,0.00000524520874)
--(axis cs:88,0.00000524520874)
--(axis cs:87,0.00000524520874)
--(axis cs:86,0.00000524520874)
--(axis cs:85,0.00000524520874)
--(axis cs:84,0.0000067949295)
--(axis cs:83,0.0000067949295)
--(axis cs:82,0.0000067949295)
--(axis cs:81,0.0000067949295)
--(axis cs:80,0.0000067949295)
--(axis cs:79,0.0000067949295)
--(axis cs:78,0.0000067949295)
--(axis cs:77,0.0000067949295)
--(axis cs:76,0.0000067949295)
--(axis cs:75,0.0000067949295)
--(axis cs:74,0.0000067949295)
--(axis cs:73,0.00000703334808)
--(axis cs:72,0.00000703334808)
--(axis cs:71,0.00000703334808)
--(axis cs:70,0.00000703334808)
--(axis cs:69,0.00000703334808)
--(axis cs:68,0.00000703334808)
--(axis cs:67,0.00002896785736)
--(axis cs:66,0.00002896785736)
--(axis cs:65,0.00002896785736)
--(axis cs:64,0.00002896785736)
--(axis cs:63,0.00002896785736)
--(axis cs:62,0.00002920627594)
--(axis cs:61,0.00003027915955)
--(axis cs:60,0.00003027915955)
--(axis cs:59,0.00003027915955)
--(axis cs:58,0.00003027915955)
--(axis cs:57,0.00003027915955)
--(axis cs:56,0.00003027915955)
--(axis cs:55,0.00003027915955)
--(axis cs:54,0.00003027915955)
--(axis cs:53,0.0000307559967)
--(axis cs:52,0.0000307559967)
--(axis cs:51,0.00004577636719)
--(axis cs:50,0.00004577636719)
--(axis cs:49,0.00006341934204)
--(axis cs:48,0.00006341934204)
--(axis cs:47,0.00006341934204)
--(axis cs:46,0.00006508827209)
--(axis cs:45,0.00006508827209)
--(axis cs:44,0.00006508827209)
--(axis cs:43,0.00006508827209)
--(axis cs:42,0.00006508827209)
--(axis cs:41,0.00007295608521)
--(axis cs:40,0.00008928775787)
--(axis cs:39,0.00008928775787)
--(axis cs:38,0.00008928775787)
--(axis cs:37,0.00008928775787)
--(axis cs:36,0.00008928775787)
--(axis cs:35,0.00009381771088)
--(axis cs:34,0.00009644031525)
--(axis cs:33,0.0001015663147)
--(axis cs:32,0.0001015663147)
--(axis cs:31,0.00011146068573)
--(axis cs:30,0.00015997886658)
--(axis cs:29,0.0001665353775)
--(axis cs:28,0.00026035308838)
--(axis cs:27,0.00029909610748)
--(axis cs:26,0.00029909610748)
--(axis cs:25,0.00029909610748)
--(axis cs:24,0.00029909610748)
--(axis cs:23,0.00033175945282)
--(axis cs:22,0.00034964084625)
--(axis cs:21,0.00034964084625)
--(axis cs:20,0.00110840797424)
--(axis cs:19,0.00125789642334)
--(axis cs:18,0.00161492824554)
--(axis cs:17,0.00161492824554)
--(axis cs:16,0.00306117534637)
--(axis cs:15,0.00680875778198)
--(axis cs:14,0.00966250896454)
--(axis cs:13,0.04075682163239)
--(axis cs:12,0.04075682163239)
--(axis cs:11,0.18189799785614)
--(axis cs:10,0.19947052001953)
--(axis cs:9,0.35159397125244)
--(axis cs:8,0.46555483341217)
--(axis cs:7,0.47791123390198)
--(axis cs:6,0.62635409832001)
--(axis cs:5,0.64755952358246)
--(axis cs:4,0.75057637691498)
--(axis cs:3,0.75057637691498)
--cycle;

\path [draw=color1, fill=color1, opacity=0.3]
(axis cs:3,0.75057640231309)
--(axis cs:3,0.49599281253425)
--(axis cs:4,0.49599281253425)
--(axis cs:5,0.38280715371981)
--(axis cs:6,0.31575255906939)
--(axis cs:7,0.31133877553069)
--(axis cs:8,0.18243823780558)
--(axis cs:9,0.12039238064058)
--(axis cs:10,0.07903043309967)
--(axis cs:11,0.02427592151452)
--(axis cs:12,0.00368768454923)
--(axis cs:13,0.00095733390679)
--(axis cs:14,0.00019053160552)
--(axis cs:15,0.00015045636274)
--(axis cs:16,0.00008694534365)
--(axis cs:17,0.00005564617314)
--(axis cs:18,0.00002790831506)
--(axis cs:19,0.00002790831506)
--(axis cs:20,0.00002479882183)
--(axis cs:21,0.00002375102977)
--(axis cs:22,0.00001053344703)
--(axis cs:23,0.00000812268168)
--(axis cs:24,0.00000812268168)
--(axis cs:25,0.00000373156631)
--(axis cs:26,0.00000325874174)
--(axis cs:27,0.00000222537698)
--(axis cs:28,0.00000222537698)
--(axis cs:29,0.00000159191972)
--(axis cs:30,0.00000131058724)
--(axis cs:31,0.00000131058724)
--(axis cs:32,0.00000131058724)
--(axis cs:33,0.00000087826638)
--(axis cs:34,0.00000087826638)
--(axis cs:35,0.00000087826638)
--(axis cs:36,0.00000087698123)
--(axis cs:37,0.00000087005893)
--(axis cs:38,0.00000087005893)
--(axis cs:39,0.00000079659244)
--(axis cs:40,0.00000079107376)
--(axis cs:41,0.00000079107376)
--(axis cs:42,0.00000079107376)
--(axis cs:43,0.00000079107376)
--(axis cs:44,0.00000074803752)
--(axis cs:45,0.00000072519959)
--(axis cs:46,0.00000072519959)
--(axis cs:47,0.00000072519959)
--(axis cs:48,0.00000018464839)
--(axis cs:49,0.00000018464839)
--(axis cs:50,0.00000013671966)
--(axis cs:51,0.00000013671966)
--(axis cs:52,0.00000013671966)
--(axis cs:53,0.00000013671966)
--(axis cs:54,0.00000013671966)
--(axis cs:55,0.00000013671966)
--(axis cs:56,0.00000013671966)
--(axis cs:57,0.00000013671966)
--(axis cs:58,0.00000013671966)
--(axis cs:59,0.00000013671966)
--(axis cs:60,0.00000013671966)
--(axis cs:61,0.00000013671966)
--(axis cs:62,0.00000013671966)
--(axis cs:63,0.00000013671966)
--(axis cs:64,0.00000013671966)
--(axis cs:65,0.00000013671966)
--(axis cs:66,0.00000013040736)
--(axis cs:67,0.00000013040736)
--(axis cs:68,0.00000013040736)
--(axis cs:69,0.00000013040736)
--(axis cs:70,0.00000013040736)
--(axis cs:71,0.00000013040736)
--(axis cs:72,0.00000013040736)
--(axis cs:73,0.00000013040736)
--(axis cs:74,0.00000013040736)
--(axis cs:75,0.00000009106075)
--(axis cs:76,0.00000009106075)
--(axis cs:77,0.00000009106075)
--(axis cs:78,0.00000009106075)
--(axis cs:79,0.00000008309164)
--(axis cs:80,0.00000008309164)
--(axis cs:81,0.00000008309164)
--(axis cs:82,0.00000008309164)
--(axis cs:83,0.00000008131926)
--(axis cs:84,0.00000008131926)
--(axis cs:85,0.00000006748716)
--(axis cs:86,0.00000006748716)
--(axis cs:87,0.00000006748716)
--(axis cs:88,0.00000006748716)
--(axis cs:89,0.00000006748716)
--(axis cs:90,0.00000006748716)
--(axis cs:91,0.00000003927302)
--(axis cs:92,0.00000003927302)
--(axis cs:93,0.00000003927302)
--(axis cs:94,0.00000003927302)
--(axis cs:95,0.0000000319584)
--(axis cs:96,0.0000000319584)
--(axis cs:97,0.0000000319584)
--(axis cs:98,0.0000000319584)
--(axis cs:99,0.0000000319584)
--(axis cs:100,0.0000000319584)
--(axis cs:101,0.00000001976218)
--(axis cs:102,0.00000001976218)
--(axis cs:103,0.00000001976218)
--(axis cs:104,0.00000001976218)
--(axis cs:104,0.00000004796743)
--(axis cs:104,0.00000004796743)
--(axis cs:103,0.00000004796743)
--(axis cs:102,0.00000004796743)
--(axis cs:101,0.00000004796743)
--(axis cs:100,0.00000010090514)
--(axis cs:99,0.00000010090514)
--(axis cs:98,0.00000010090514)
--(axis cs:97,0.00000010090514)
--(axis cs:96,0.00000010090514)
--(axis cs:95,0.00000010090514)
--(axis cs:94,0.00000010825112)
--(axis cs:93,0.00000010825112)
--(axis cs:92,0.00000010825112)
--(axis cs:91,0.00000010825112)
--(axis cs:90,0.00000016401575)
--(axis cs:89,0.00000016401575)
--(axis cs:88,0.00000016401575)
--(axis cs:87,0.00000016401575)
--(axis cs:86,0.00000016401575)
--(axis cs:85,0.00000016401575)
--(axis cs:84,0.00000017675402)
--(axis cs:83,0.00000017675402)
--(axis cs:82,0.00000018789812)
--(axis cs:81,0.00000018789812)
--(axis cs:80,0.00000018789812)
--(axis cs:79,0.00000018789812)
--(axis cs:78,0.0000001989907)
--(axis cs:77,0.0000001989907)
--(axis cs:76,0.0000001989907)
--(axis cs:75,0.0000001989907)
--(axis cs:74,0.00000024825161)
--(axis cs:73,0.00000024825161)
--(axis cs:72,0.00000024825161)
--(axis cs:71,0.00000024825161)
--(axis cs:70,0.00000024825161)
--(axis cs:69,0.00000024825161)
--(axis cs:68,0.00000024825161)
--(axis cs:67,0.00000024825161)
--(axis cs:66,0.00000024825161)
--(axis cs:65,0.00000027110047)
--(axis cs:64,0.00000027110047)
--(axis cs:63,0.00000027110047)
--(axis cs:62,0.00000027110047)
--(axis cs:61,0.00000027110047)
--(axis cs:60,0.00000027110047)
--(axis cs:59,0.00000027110047)
--(axis cs:58,0.00000027110047)
--(axis cs:57,0.00000027110047)
--(axis cs:56,0.00000027110047)
--(axis cs:55,0.00000027110047)
--(axis cs:54,0.00000027110047)
--(axis cs:53,0.00000027110047)
--(axis cs:52,0.00000027110047)
--(axis cs:51,0.00000027110047)
--(axis cs:50,0.00000027110047)
--(axis cs:49,0.00000157667385)
--(axis cs:48,0.00000157667385)
--(axis cs:47,0.00000227883222)
--(axis cs:46,0.00000227883222)
--(axis cs:45,0.00000227883222)
--(axis cs:44,0.00000230157042)
--(axis cs:43,0.00000283713623)
--(axis cs:42,0.00000283713623)
--(axis cs:41,0.00000283713623)
--(axis cs:40,0.00000283713623)
--(axis cs:39,0.00000284086242)
--(axis cs:38,0.0000029008356)
--(axis cs:37,0.0000029008356)
--(axis cs:36,0.0000029054587)
--(axis cs:35,0.00000290632281)
--(axis cs:34,0.00000290632281)
--(axis cs:33,0.00000290632281)
--(axis cs:32,0.00000327661916)
--(axis cs:31,0.00000327661916)
--(axis cs:30,0.00000327661916)
--(axis cs:29,0.00000360495184)
--(axis cs:28,0.00000879347075)
--(axis cs:27,0.00000879347075)
--(axis cs:26,0.00001104138856)
--(axis cs:25,0.00001150951209)
--(axis cs:24,0.00003209393459)
--(axis cs:23,0.00003209393459)
--(axis cs:22,0.00003409568025)
--(axis cs:21,0.00006063144323)
--(axis cs:20,0.00007330964376)
--(axis cs:19,0.00007637803517)
--(axis cs:18,0.00007637803517)
--(axis cs:17,0.00026597392095)
--(axis cs:16,0.00029218625375)
--(axis cs:15,0.0004689776938)
--(axis cs:14,0.00238055150491)
--(axis cs:13,0.00734684270321)
--(axis cs:12,0.02598315732656)
--(axis cs:11,0.12615201496075)
--(axis cs:10,0.2528996053755)
--(axis cs:9,0.30130540613662)
--(axis cs:8,0.40547124993532)
--(axis cs:7,0.56961660556601)
--(axis cs:6,0.57523462475051)
--(axis cs:5,0.64868288107057)
--(axis cs:4,0.75057640231309)
--(axis cs:3,0.75057640231309)
--cycle;

\addplot [semithick, blue]
table {%
3 0.62328457832336
4 0.61122167110443
5 0.47473418712616
6 0.33381676673889
7 0.33381676673889
8 0.27135825157166
9 0.2449563741684
10 0.09795093536377
11 0.09478724002838
12 0.09266197681427
13 0.06674015522003
14 0.02149939537048
15 0.00186097621918
16 0.00186097621918
17 0.00186097621918
18 0.00060415267944
19 0.00019860267639
20 0.00013995170593
21 0.00013995170593
22 0.00013995170593
23 0.00013995170593
24 0.00013995170593
25 0.00013995170593
26 0.00013601779938
27 0.00011658668518
28 0.00011658668518
29 0.00011658668518
30 0.00011610984802
31 0.00011610984802
32 0.00011539459229
33 0.00010216236115
34 0.00008630752563
35 0.00008630752563
36 0.00008630752563
37 0.00008630752563
38 0.00008630752563
39 0.00008630752563
40 0.00008630752563
41 0.00007772445679
42 0.00007772445679
43 0.00007772445679
44 0.00007772445679
45 0.00007772445679
46 0.00007772445679
47 0.00007772445679
48 0.00007772445679
49 0.00007772445679
50 0.0000764131546
51 0.0000741481781
52 0.00007176399231
53 0.00007176399231
54 0.00007176399231
55 0.0000706911087
56 0.0000706911087
57 0.00006890296936
58 0.00006890296936
59 0.00006890296936
60 0.00002503395081
61 0.00002503395081
62 0.00002503395081
63 0.00002324581146
64 0.00002324581146
65 0.00002324581146
66 0.00002324581146
67 0.00002324581146
68 0.00002324581146
69 0.00002324581146
70 0.00002324581146
71 0.00002181529999
72 0.00002181529999
73 0.00002181529999
74 0.00002181529999
75 0.00002181529999
76 0.00002181529999
77 0.00002038478851
78 0.00001358985901
79 0.00001358985901
80 0.00001358985901
81 0.00001358985901
82 0.00001358985901
83 0.00001358985901
84 0.00001358985901
85 0.00001358985901
86 0.00001358985901
87 0.00001358985901
88 0.00001358985901
89 0.00001358985901
90 0.00001287460327
91 0.00001287460327
92 0.00001287460327
93 0.00001287460327
94 0.00001287460327
95 0.00001287460327
96 0.00001287460327
97 0.00001299381256
98 0.00001299381256
99 0.00001013278961
100 0.00001013278961
101 0.00001013278961
102 0.00001013278961
103 0.00001013278961
104 0.00001013278961
};
\addplot [semithick, color1, dash dot]
table {%
3 0.62328460742367
4 0.62328460742367
5 0.61009744466122
6 0.53314768853027
7 0.5331476081644
8 0.47188204883619
9 0.36873341877242
10 0.30760809135809
11 0.30584928658485
12 0.30355272882778
13 0.28185660406661
14 0.28043279684756
15 0.27111722000611
16 0.20573187493391
17 0.13809330028409
18 0.13504750870492
19 0.13256415261118
20 0.10360454327959
21 0.10018437004878
22 0.10005594189156
23 0.10005594189156
24 0.10005594189156
25 0.1000554507892
26 0.10002335303623
27 0.10002335303623
28 0.10002335303623
29 0.10002335303623
30 0.10001756501026
31 0.10001756501026
32 0.10001756501026
33 0.10000565437088
34 0.10000558878295
35 0.10000558878295
36 0.10000416929589
37 0.10000416929589
38 0.10000414681337
39 0.10000414681337
40 0.10000408979659
41 0.10000408979659
42 0.10000358061779
43 0.10000300987206
44 0.10000300987206
45 0.10000300987206
46 0.10000300987206
47 0.10000300987206
48 0.10000300987206
49 0.10000300987206
50 0.10000300987206
51 0.10000300987206
52 0.10000300987206
53 0.10000300987206
54 0.10000300987206
55 0.10000300987206
56 0.10000300987206
57 0.10000291220757
58 0.10000291220757
59 0.10000282792339
60 0.10000282792339
61 0.10000205060348
62 0.10000205060348
63 0.10000205060348
64 0.10000205060348
65 0.10000205060348
66 0.1000014032099
67 0.1000014032099
68 0.1000014032099
69 0.1000014032099
70 0.1000014032099
71 0.1000014032099
72 0.1000014032099
73 0.1000014032099
74 0.1000014032099
75 0.1000014032099
76 0.1000014032099
77 0.1000014032099
78 0.1000014032099
79 0.1000014032099
80 0.1000014032099
81 0.1000014032099
82 0.1000014032099
83 0.1000014032099
84 0.1000014032099
85 0.1000014032099
86 0.1000014032099
87 0.1000014032099
88 0.1000014032099
89 0.1000014032099
90 0.10000133939056
91 0.10000133939056
92 0.10000133939056
93 0.10000133939056
94 0.10000133939056
95 0.10000133939056
96 0.10000133939056
97 0.10000133939056
98 0.10000133939056
99 0.10000133939056
100 0.10000133939056
101 0.10000133939056
102 0.10000133939056
103 0.10000133939056
104 0.10000088527863
};
\addplot [semithick, blue, dash dot]
table {%
3 0.62328457832336
4 0.62328457832336
5 0.51056110858917
6 0.4815217256546
7 0.3328115940094
8 0.31719481945038
9 0.22065854072571
10 0.11181139945984
11 0.09356343746185
12 0.022496342659
13 0.022496342659
14 0.00598406791687
15 0.00415515899658
16 0.00192046165466
17 0.00112223625183
18 0.00112223625183
19 0.00084137916565
20 0.00069844722748
21 0.00026774406433
22 0.00026774406433
23 0.00025391578674
24 0.0002304315567
25 0.0002304315567
26 0.0002304315567
27 0.0002304315567
28 0.00019121170044
29 0.00012302398682
30 0.00011646747589
31 0.00007855892181
32 0.00006937980652
33 0.00006937980652
34 0.00006639957428
35 0.00006484985352
36 0.00005960464478
37 0.00005960464478
38 0.00005960464478
39 0.00005960464478
40 0.00005960464478
41 0.00004887580872
42 0.00004172325134
43 0.00004172325134
44 0.00004172325134
45 0.00004172325134
46 0.00004172325134
47 0.000039935112
48 0.000039935112
49 0.000039935112
50 0.00002992153168
51 0.00002992153168
52 0.00001835823059
53 0.00001835823059
54 0.00001788139343
55 0.00001788139343
56 0.00001788139343
57 0.00001788139343
58 0.00001788139343
59 0.00001788139343
60 0.00001788139343
61 0.00001788139343
62 0.00001668930054
63 0.00001633167267
64 0.00001633167267
65 0.00001633167267
66 0.00001633167267
67 0.00001633167267
68 0.00000500679016
69 0.00000500679016
70 0.00000500679016
71 0.00000500679016
72 0.00000500679016
73 0.00000500679016
74 0.00000476837158
75 0.00000476837158
76 0.00000476837158
77 0.00000476837158
78 0.00000476837158
79 0.00000476837158
80 0.00000476837158
81 0.00000476837158
82 0.00000476837158
83 0.00000476837158
84 0.00000476837158
85 0.0000034570694
86 0.0000034570694
87 0.0000034570694
88 0.0000034570694
89 0.0000034570694
90 0.0000034570694
91 0.0000034570694
92 0.0000034570694
93 0.0000034570694
94 0.00000262260437
95 0.00000262260437
96 0.00000262260437
97 0.00000262260437
98 0.00000262260437
99 0.00000262260437
100 0.00000262260437
101 0.00000262260437
102 0.00000262260437
103 0.00000262260437
104 0.00000262260437
};
\addplot [semithick, color1]
table {%
3 0.62328460742367
4 0.62328460742367
5 0.51574501739519
6 0.44549359190995
7 0.44047769054835
8 0.29395474387045
9 0.2108488933886
10 0.16596501923758
11 0.07521396823763
12 0.0148354209379
13 0.004152088305
14 0.00128554155521
15 0.00030971702827
16 0.0001895657987
17 0.00016081004704
18 0.00005214317511
19 0.00005214317511
20 0.0000490542328
21 0.0000421912365
22 0.00002231456364
23 0.00002010830813
24 0.00002010830813
25 0.0000076205392
26 0.00000715006515
27 0.00000550942386
28 0.00000550942386
29 0.00000259843578
30 0.0000022936032
31 0.0000022936032
32 0.0000022936032
33 0.0000018922946
34 0.0000018922946
35 0.0000018922946
36 0.00000189121997
37 0.00000188544726
38 0.00000188544726
39 0.00000181872743
40 0.000001814105
41 0.000001814105
42 0.000001814105
43 0.000001814105
44 0.00000152480397
45 0.0000015020159
46 0.0000015020159
47 0.0000015020159
48 0.00000088066112
49 0.00000088066112
50 0.00000020391006
51 0.00000020391006
52 0.00000020391006
53 0.00000020391006
54 0.00000020391006
55 0.00000020391006
56 0.00000020391006
57 0.00000020391006
58 0.00000020391006
59 0.00000020391006
60 0.00000020391006
61 0.00000020391006
62 0.00000020391006
63 0.00000020391006
64 0.00000020391006
65 0.00000020391006
66 0.00000018932948
67 0.00000018932948
68 0.00000018932948
69 0.00000018932948
70 0.00000018932948
71 0.00000018932948
72 0.00000018932948
73 0.00000018932948
74 0.00000018932948
75 0.00000014502573
76 0.00000014502573
77 0.00000014502573
78 0.00000014502573
79 0.00000013549488
80 0.00000013549488
81 0.00000013549488
82 0.00000013549488
83 0.00000012903664
84 0.00000012903664
85 0.00000011575146
86 0.00000011575146
87 0.00000011575146
88 0.00000011575146
89 0.00000011575146
90 0.00000011575146
91 0.00000007376207
92 0.00000007376207
93 0.00000007376207
94 0.00000007376207
95 0.00000006643177
96 0.00000006643177
97 0.00000006643177
98 0.00000006643177
99 0.00000006643177
100 0.00000006643177
101 0.0000000338648
102 0.0000000338648
103 0.0000000338648
104 0.0000000338648
};
\end{axis}

\end{tikzpicture}

%% file: figures/bop_2d_braninscaled_weak.tex
\begin{tikzpicture}

\definecolor{color0}{rgb}{0,0,1}
\definecolor{color1}{rgb}{1,0.549019607843137,0}
\definecolor{color2}{rgb}{1,0.647058823529412,0}
\definecolor{color3}{rgb}{0.564705882352941,0.933333333333333,0.564705882352941}

\begin{axis}[axis on top,
enlarge x limits=false,
enlarge y limits=false,
height=\figureheight,
scale only axis,
tick align=outside,
tick pos=left,
tick pos=left,
width=\figurewidth,
xlabel={Iteration},
xmin=10, xmax=75,
xtick style={color=black},
xtick={-10,0,10,25,50,75,100},
xticklabels={\ensuremath{-}10,0,10,25,50,75,90},
ylabel={Regret},
ymin=-0.005, ymax=0.13,
ytick style={color=black},
ytick={0.   , 0.13},
]
\node[anchor=north east] at (rel axis cs:1,1) {Branin 2D (weak)};
\path [draw=color1, fill=color1, opacity=0.3]
(axis cs:10,0.108571740173539)
--(axis cs:10,0.0428720774551092)
--(axis cs:11,0.0314381522871823)
--(axis cs:12,0.0314381522871823)
--(axis cs:13,0.0311899656704787)
--(axis cs:14,0.0287260401088036)
--(axis cs:15,0.0245210122488913)
--(axis cs:16,0.0212664598017752)
--(axis cs:17,0.0212664598017752)
--(axis cs:18,0.02116392888657)
--(axis cs:19,0.0175036867802915)
--(axis cs:20,0.0171827095494242)
--(axis cs:21,0.0133797369674256)
--(axis cs:22,0.00947376782879302)
--(axis cs:23,0.00714401949684943)
--(axis cs:24,0.00463034479914797)
--(axis cs:25,0.00319560596159751)
--(axis cs:26,0.00293857534154119)
--(axis cs:27,0.00185754369603897)
--(axis cs:28,0.00171565174267335)
--(axis cs:29,0.00150040421633303)
--(axis cs:30,0.0012948001932565)
--(axis cs:31,0.000854321899702696)
--(axis cs:32,0.000682581952250951)
--(axis cs:33,0.000579685305596125)
--(axis cs:34,0.000579685305596125)
--(axis cs:35,0.000548205423199155)
--(axis cs:36,0.000505215720152609)
--(axis cs:37,0.000408902354972378)
--(axis cs:38,0.000333996007003622)
--(axis cs:39,0.000238798599118422)
--(axis cs:40,0.000238798599118422)
--(axis cs:41,0.000213872516845863)
--(axis cs:42,0.000210725964305264)
--(axis cs:43,0.000163661126041008)
--(axis cs:44,0.000127542555812623)
--(axis cs:45,0.000119806744412761)
--(axis cs:46,0.000119806744412761)
--(axis cs:47,9.9666477271715e-05)
--(axis cs:48,9.9666477271715e-05)
--(axis cs:49,9.9666477271715e-05)
--(axis cs:50,9.9666477271715e-05)
--(axis cs:51,8.58084699435221e-05)
--(axis cs:52,8.2863479563767e-05)
--(axis cs:53,8.2863479563767e-05)
--(axis cs:54,8.2863479563767e-05)
--(axis cs:55,8.2863479563767e-05)
--(axis cs:56,7.81494856437297e-05)
--(axis cs:57,7.81494856437297e-05)
--(axis cs:58,7.81494856437297e-05)
--(axis cs:59,7.81494856437297e-05)
--(axis cs:60,7.81494856437297e-05)
--(axis cs:61,7.81494856437297e-05)
--(axis cs:62,7.81494856437297e-05)
--(axis cs:63,7.81494856437297e-05)
--(axis cs:64,5.96237500824572e-05)
--(axis cs:65,5.96237500824572e-05)
--(axis cs:66,5.92547104905305e-05)
--(axis cs:67,4.93713624426847e-05)
--(axis cs:68,4.93713624426847e-05)
--(axis cs:69,4.93713624426847e-05)
--(axis cs:70,4.93713624426847e-05)
--(axis cs:71,4.76693935880749e-05)
--(axis cs:72,4.76693935880749e-05)
--(axis cs:73,4.76693935880749e-05)
--(axis cs:74,4.76693935880749e-05)
--(axis cs:75,4.76693935880749e-05)
--(axis cs:76,4.55267100803453e-05)
--(axis cs:77,4.27889285334674e-05)
--(axis cs:78,2.66128248529893e-05)
--(axis cs:79,2.66128248529893e-05)
--(axis cs:80,2.66128248529893e-05)
--(axis cs:81,2.66128248529893e-05)
--(axis cs:82,2.66128248529893e-05)
--(axis cs:83,2.66128248529893e-05)
--(axis cs:84,2.66128248529893e-05)
--(axis cs:85,2.66128248529893e-05)
--(axis cs:86,2.66128248529893e-05)
--(axis cs:87,2.66128248529893e-05)
--(axis cs:88,2.66128248529893e-05)
--(axis cs:89,2.66128248529893e-05)
--(axis cs:90,2.22659057208924e-05)
--(axis cs:91,2.22659057208924e-05)
--(axis cs:92,2.22659057208924e-05)
--(axis cs:93,2.22659057208924e-05)
--(axis cs:94,2.22659057208924e-05)
--(axis cs:95,2.22659057208924e-05)
--(axis cs:96,2.22659057208924e-05)
--(axis cs:97,2.22659057208924e-05)
--(axis cs:98,2.08715658475132e-05)
--(axis cs:99,2.08715658475132e-05)
--(axis cs:100,2.08715658475132e-05)
--(axis cs:101,2.08715658475132e-05)
--(axis cs:102,1.96941223813336e-05)
--(axis cs:103,1.96941223813336e-05)
--(axis cs:104,1.96941223813336e-05)
--(axis cs:105,1.96941223813336e-05)
--(axis cs:106,1.96941223813336e-05)
--(axis cs:107,1.96941223813336e-05)
--(axis cs:108,1.96941223813336e-05)
--(axis cs:109,1.96941223813336e-05)
--(axis cs:110,1.96941223813336e-05)
--(axis cs:111,1.96941223813336e-05)
--(axis cs:111,3.59192040484484e-05)
--(axis cs:111,3.59192040484484e-05)
--(axis cs:110,3.59192040484484e-05)
--(axis cs:109,3.59192040484484e-05)
--(axis cs:108,3.59192040484484e-05)
--(axis cs:107,3.59192040484484e-05)
--(axis cs:106,3.59192040484484e-05)
--(axis cs:105,3.59192040484484e-05)
--(axis cs:104,3.59192040484484e-05)
--(axis cs:103,3.59192040484484e-05)
--(axis cs:102,3.59192040484484e-05)
--(axis cs:101,4.22669840744648e-05)
--(axis cs:100,4.22669840744648e-05)
--(axis cs:99,4.22669840744648e-05)
--(axis cs:98,4.22669840744648e-05)
--(axis cs:97,6.9919702743045e-05)
--(axis cs:96,6.9919702743045e-05)
--(axis cs:95,6.9919702743045e-05)
--(axis cs:94,6.9919702743045e-05)
--(axis cs:93,6.9919702743045e-05)
--(axis cs:92,6.9919702743045e-05)
--(axis cs:91,6.9919702743045e-05)
--(axis cs:90,6.9919702743045e-05)
--(axis cs:89,7.4829818215388e-05)
--(axis cs:88,7.4829818215388e-05)
--(axis cs:87,7.4829818215388e-05)
--(axis cs:86,7.4829818215388e-05)
--(axis cs:85,7.4829818215388e-05)
--(axis cs:84,7.4829818215388e-05)
--(axis cs:83,7.4829818215388e-05)
--(axis cs:82,7.4829818215388e-05)
--(axis cs:81,7.4829818215388e-05)
--(axis cs:80,7.4829818215388e-05)
--(axis cs:79,7.4829818215388e-05)
--(axis cs:78,7.4829818215388e-05)
--(axis cs:77,9.66278356254573e-05)
--(axis cs:76,0.000100030867025808)
--(axis cs:75,0.000101213004589279)
--(axis cs:74,0.000101213004589279)
--(axis cs:73,0.000101213004589279)
--(axis cs:72,0.000101213004589279)
--(axis cs:71,0.000101213004589279)
--(axis cs:70,0.00010228890565049)
--(axis cs:69,0.00010228890565049)
--(axis cs:68,0.00010228890565049)
--(axis cs:67,0.00010228890565049)
--(axis cs:66,0.000111844910714256)
--(axis cs:65,0.000113887424409664)
--(axis cs:64,0.000113887424409664)
--(axis cs:63,0.000127356436298275)
--(axis cs:62,0.000127356436298275)
--(axis cs:61,0.000127356436298275)
--(axis cs:60,0.000127356436298275)
--(axis cs:59,0.000127356436298275)
--(axis cs:58,0.000127356436298275)
--(axis cs:57,0.000127356436298275)
--(axis cs:56,0.000127356436298275)
--(axis cs:55,0.000133585309577731)
--(axis cs:54,0.000133585309577731)
--(axis cs:53,0.000133585309577731)
--(axis cs:52,0.000133585309577731)
--(axis cs:51,0.000137753730639785)
--(axis cs:50,0.000171975524799626)
--(axis cs:49,0.000171975524799626)
--(axis cs:48,0.000171975524799626)
--(axis cs:47,0.000171975524799626)
--(axis cs:46,0.000247287184599838)
--(axis cs:45,0.000247287184599838)
--(axis cs:44,0.000251770500242165)
--(axis cs:43,0.000288834013391825)
--(axis cs:42,0.000418872808409693)
--(axis cs:41,0.000422582495069722)
--(axis cs:40,0.000448261998945187)
--(axis cs:39,0.000448261998945187)
--(axis cs:38,0.000582008376012837)
--(axis cs:37,0.00065755493194103)
--(axis cs:36,0.000846462910999803)
--(axis cs:35,0.00089745011696795)
--(axis cs:34,0.00119457554897847)
--(axis cs:33,0.00119457554897847)
--(axis cs:32,0.00133937773220362)
--(axis cs:31,0.00294773343490305)
--(axis cs:30,0.00467718708855366)
--(axis cs:29,0.00485447343757876)
--(axis cs:28,0.00503556391305213)
--(axis cs:27,0.00518222021375704)
--(axis cs:26,0.00658524675372097)
--(axis cs:25,0.00677123035667966)
--(axis cs:24,0.00840219185414705)
--(axis cs:23,0.011664115610554)
--(axis cs:22,0.016435803746643)
--(axis cs:21,0.0194610743025069)
--(axis cs:20,0.0260157647860625)
--(axis cs:19,0.0300616247533489)
--(axis cs:18,0.0344359847056999)
--(axis cs:17,0.0353690878269783)
--(axis cs:16,0.0353690878269783)
--(axis cs:15,0.047653761548673)
--(axis cs:14,0.0543350460223794)
--(axis cs:13,0.0879314611811201)
--(axis cs:12,0.098199460395853)
--(axis cs:11,0.098199460395853)
--(axis cs:10,0.108571740173539)
--cycle;

\path [draw=blue, fill=blue, opacity=0.3]
(axis cs:10,0.108571708202362)
--(axis cs:10,0.0428720451891422)
--(axis cs:11,0.0394763275980949)
--(axis cs:12,0.0394763275980949)
--(axis cs:13,0.0394763275980949)
--(axis cs:14,0.0390365421772003)
--(axis cs:15,0.0390365421772003)
--(axis cs:16,0.0382666885852814)
--(axis cs:17,0.0382666885852814)
--(axis cs:18,0.0265496876090765)
--(axis cs:19,0.0238866694271564)
--(axis cs:20,0.0238866694271564)
--(axis cs:21,0.0213665254414082)
--(axis cs:22,0.0213665254414082)
--(axis cs:23,0.0185194499790668)
--(axis cs:24,0.0185194499790668)
--(axis cs:25,0.0185194499790668)
--(axis cs:26,0.0185194499790668)
--(axis cs:27,0.0167573057115078)
--(axis cs:28,0.0167573057115078)
--(axis cs:29,0.016325056552887)
--(axis cs:30,0.016325056552887)
--(axis cs:31,0.0102417171001434)
--(axis cs:32,0.0102417171001434)
--(axis cs:33,0.0102117955684662)
--(axis cs:34,0.0102117955684662)
--(axis cs:35,0.0102117955684662)
--(axis cs:36,0.00607288908213377)
--(axis cs:37,0.00607288908213377)
--(axis cs:38,0.00607288908213377)
--(axis cs:39,0.00607288908213377)
--(axis cs:40,0.00607288908213377)
--(axis cs:41,0.00525319389998913)
--(axis cs:42,0.00525319389998913)
--(axis cs:43,0.00525319389998913)
--(axis cs:44,0.00525319389998913)
--(axis cs:45,0.00525319389998913)
--(axis cs:46,0.00525319389998913)
--(axis cs:47,0.00525319389998913)
--(axis cs:48,0.00525319389998913)
--(axis cs:49,0.00525319389998913)
--(axis cs:50,0.00525319389998913)
--(axis cs:51,0.00479436200112104)
--(axis cs:52,0.00479436200112104)
--(axis cs:53,0.00479420134797692)
--(axis cs:54,0.00463061267510056)
--(axis cs:55,0.00463061267510056)
--(axis cs:56,0.00460127927362919)
--(axis cs:57,0.00454916944727302)
--(axis cs:58,0.0040155160240829)
--(axis cs:59,0.0040155160240829)
--(axis cs:60,0.00285367434844375)
--(axis cs:61,0.00241067819297314)
--(axis cs:62,0.00219598575495183)
--(axis cs:63,0.00219598575495183)
--(axis cs:64,0.00219598575495183)
--(axis cs:65,0.00219598575495183)
--(axis cs:66,0.00219598575495183)
--(axis cs:67,0.00219598575495183)
--(axis cs:68,0.00219598575495183)
--(axis cs:69,0.00219598575495183)
--(axis cs:70,0.00219598575495183)
--(axis cs:71,0.00219598575495183)
--(axis cs:72,0.00219598575495183)
--(axis cs:73,0.00219598575495183)
--(axis cs:74,0.00219598575495183)
--(axis cs:75,0.00209680129773915)
--(axis cs:76,0.00209680129773915)
--(axis cs:77,0.00209680129773915)
--(axis cs:78,0.00179878156632185)
--(axis cs:79,0.00179878156632185)
--(axis cs:80,0.00179878156632185)
--(axis cs:81,0.00179878156632185)
--(axis cs:82,0.00179878156632185)
--(axis cs:83,0.00179878156632185)
--(axis cs:84,0.00179878156632185)
--(axis cs:85,0.00179878156632185)
--(axis cs:86,0.00179878156632185)
--(axis cs:87,0.00179878156632185)
--(axis cs:88,0.00179878156632185)
--(axis cs:89,0.00179878156632185)
--(axis cs:90,0.00170115474611521)
--(axis cs:91,0.00168472225777805)
--(axis cs:92,0.00168472225777805)
--(axis cs:93,0.00168472225777805)
--(axis cs:94,0.00168472225777805)
--(axis cs:95,0.00168472225777805)
--(axis cs:96,0.00168472225777805)
--(axis cs:97,0.00168472225777805)
--(axis cs:98,0.00168472225777805)
--(axis cs:99,0.00152010156307369)
--(axis cs:100,0.00152010156307369)
--(axis cs:101,0.00152010156307369)
--(axis cs:102,0.00152010156307369)
--(axis cs:103,0.00152010156307369)
--(axis cs:104,0.00152010156307369)
--(axis cs:105,0.00152010156307369)
--(axis cs:106,0.00152010156307369)
--(axis cs:107,0.00152010156307369)
--(axis cs:108,0.00152010156307369)
--(axis cs:109,0.00152010156307369)
--(axis cs:110,0.00152010156307369)
--(axis cs:111,0.00152010156307369)
--(axis cs:111,0.00282102404162288)
--(axis cs:111,0.00282102404162288)
--(axis cs:110,0.00282102404162288)
--(axis cs:109,0.00282102404162288)
--(axis cs:108,0.00282102404162288)
--(axis cs:107,0.00282102404162288)
--(axis cs:106,0.00282102404162288)
--(axis cs:105,0.00282102404162288)
--(axis cs:104,0.00282102404162288)
--(axis cs:103,0.00282102404162288)
--(axis cs:102,0.00282102404162288)
--(axis cs:101,0.00282102404162288)
--(axis cs:100,0.00282102404162288)
--(axis cs:99,0.00282102404162288)
--(axis cs:98,0.00295686372555792)
--(axis cs:97,0.00295686372555792)
--(axis cs:96,0.00295686372555792)
--(axis cs:95,0.00295686372555792)
--(axis cs:94,0.00295686372555792)
--(axis cs:93,0.00295686372555792)
--(axis cs:92,0.00295686372555792)
--(axis cs:91,0.00295686372555792)
--(axis cs:90,0.00300615513697267)
--(axis cs:89,0.00312461517751217)
--(axis cs:88,0.00312461517751217)
--(axis cs:87,0.00312461517751217)
--(axis cs:86,0.00312461517751217)
--(axis cs:85,0.00312461517751217)
--(axis cs:84,0.00312461517751217)
--(axis cs:83,0.00312461517751217)
--(axis cs:82,0.00312461517751217)
--(axis cs:81,0.00312461517751217)
--(axis cs:80,0.00312461517751217)
--(axis cs:79,0.00312461517751217)
--(axis cs:78,0.00312461517751217)
--(axis cs:77,0.00341233680956066)
--(axis cs:76,0.00341233680956066)
--(axis cs:75,0.00341233680956066)
--(axis cs:74,0.00359785067848861)
--(axis cs:73,0.00359785067848861)
--(axis cs:72,0.00359785067848861)
--(axis cs:71,0.00359785067848861)
--(axis cs:70,0.00359785067848861)
--(axis cs:69,0.00359785067848861)
--(axis cs:68,0.00359785067848861)
--(axis cs:67,0.00359785067848861)
--(axis cs:66,0.00359785067848861)
--(axis cs:65,0.00359785067848861)
--(axis cs:64,0.00359785067848861)
--(axis cs:63,0.00359785067848861)
--(axis cs:62,0.00359785067848861)
--(axis cs:61,0.00371487811207771)
--(axis cs:60,0.00555903138592839)
--(axis cs:59,0.00691043911501765)
--(axis cs:58,0.00691043911501765)
--(axis cs:57,0.0076840347610414)
--(axis cs:56,0.00773110706359148)
--(axis cs:55,0.00780360540375113)
--(axis cs:54,0.00780360540375113)
--(axis cs:53,0.00831789243966341)
--(axis cs:52,0.00831802375614643)
--(axis cs:51,0.00831802375614643)
--(axis cs:50,0.0112819019705057)
--(axis cs:49,0.0112819019705057)
--(axis cs:48,0.0112819019705057)
--(axis cs:47,0.0112819019705057)
--(axis cs:46,0.0112819019705057)
--(axis cs:45,0.0112819019705057)
--(axis cs:44,0.0112819019705057)
--(axis cs:43,0.0112819019705057)
--(axis cs:42,0.0112819019705057)
--(axis cs:41,0.0112819019705057)
--(axis cs:40,0.012111640535295)
--(axis cs:39,0.012111640535295)
--(axis cs:38,0.012111640535295)
--(axis cs:37,0.012111640535295)
--(axis cs:36,0.012111640535295)
--(axis cs:35,0.0232194289565086)
--(axis cs:34,0.0232194289565086)
--(axis cs:33,0.0232194289565086)
--(axis cs:32,0.0232389941811562)
--(axis cs:31,0.0232389941811562)
--(axis cs:30,0.0347821116447449)
--(axis cs:29,0.0347821116447449)
--(axis cs:28,0.0350107587873936)
--(axis cs:27,0.0350107587873936)
--(axis cs:26,0.0361970849335194)
--(axis cs:25,0.0361970849335194)
--(axis cs:24,0.0361970849335194)
--(axis cs:23,0.0361970849335194)
--(axis cs:22,0.0380856096744537)
--(axis cs:21,0.0380856096744537)
--(axis cs:20,0.0405683629214764)
--(axis cs:19,0.0405683629214764)
--(axis cs:18,0.0462207347154617)
--(axis cs:17,0.068044476211071)
--(axis cs:16,0.068044476211071)
--(axis cs:15,0.069063663482666)
--(axis cs:14,0.069063663482666)
--(axis cs:13,0.0697760581970215)
--(axis cs:12,0.0697760581970215)
--(axis cs:11,0.0697760581970215)
--(axis cs:10,0.108571708202362)
--cycle;

\path [draw=blue, fill=blue, opacity=0.3]
(axis cs:10,0.108571708202362)
--(axis cs:10,0.0428720451891422)
--(axis cs:11,0.0363190695643425)
--(axis cs:12,0.0296233482658863)
--(axis cs:13,0.0294356793165207)
--(axis cs:14,0.0252363868057728)
--(axis cs:15,0.0245496444404125)
--(axis cs:16,0.0206421390175819)
--(axis cs:17,0.0181445069611073)
--(axis cs:18,0.0146937798708677)
--(axis cs:19,0.0111925005912781)
--(axis cs:20,0.0102438349276781)
--(axis cs:21,0.00978771597146988)
--(axis cs:22,0.00978771597146988)
--(axis cs:23,0.00978771597146988)
--(axis cs:24,0.00714214239269495)
--(axis cs:25,0.00517761707305908)
--(axis cs:26,0.00488229934126139)
--(axis cs:27,0.00488229934126139)
--(axis cs:28,0.00488229934126139)
--(axis cs:29,0.00427801255136728)
--(axis cs:30,0.00278471922501922)
--(axis cs:31,0.00278471922501922)
--(axis cs:32,0.00253304513171315)
--(axis cs:33,0.00253304513171315)
--(axis cs:34,0.00253304513171315)
--(axis cs:35,0.00238807918503881)
--(axis cs:36,0.00219240575097501)
--(axis cs:37,0.00205667829141021)
--(axis cs:38,0.00205667829141021)
--(axis cs:39,0.00205667829141021)
--(axis cs:40,0.00205667829141021)
--(axis cs:41,0.00205667829141021)
--(axis cs:42,0.00198624678887427)
--(axis cs:43,0.00153257045894861)
--(axis cs:44,0.00153257045894861)
--(axis cs:45,0.00128548638895154)
--(axis cs:46,0.00128548638895154)
--(axis cs:47,0.00128548638895154)
--(axis cs:48,0.00128548638895154)
--(axis cs:49,0.00124461704399437)
--(axis cs:50,0.00103820580989122)
--(axis cs:51,0.00103820580989122)
--(axis cs:52,0.00103820580989122)
--(axis cs:53,0.00103820580989122)
--(axis cs:54,0.00103820580989122)
--(axis cs:55,0.00103820580989122)
--(axis cs:56,0.00103820580989122)
--(axis cs:57,0.00103820580989122)
--(axis cs:58,0.00103820580989122)
--(axis cs:59,0.00103820580989122)
--(axis cs:60,0.00103820580989122)
--(axis cs:61,0.00103820580989122)
--(axis cs:62,0.00103820580989122)
--(axis cs:63,0.00103820580989122)
--(axis cs:64,0.00103820580989122)
--(axis cs:65,0.00103820580989122)
--(axis cs:66,0.00103820580989122)
--(axis cs:67,0.00103820580989122)
--(axis cs:68,0.00103820580989122)
--(axis cs:69,0.00103820580989122)
--(axis cs:70,0.00103820580989122)
--(axis cs:71,0.00103820580989122)
--(axis cs:72,0.00103820580989122)
--(axis cs:73,0.00103820580989122)
--(axis cs:74,0.00103820580989122)
--(axis cs:75,0.00103820580989122)
--(axis cs:76,0.000949075678363442)
--(axis cs:77,0.000949075678363442)
--(axis cs:78,0.000949075678363442)
--(axis cs:79,0.000949075678363442)
--(axis cs:80,0.000949075678363442)
--(axis cs:81,0.000949075678363442)
--(axis cs:82,0.000949075678363442)
--(axis cs:83,0.000949075678363442)
--(axis cs:84,0.000949075678363442)
--(axis cs:85,0.000949075678363442)
--(axis cs:86,0.000949075678363442)
--(axis cs:87,0.000861761160194874)
--(axis cs:88,0.000861761160194874)
--(axis cs:89,0.000861761160194874)
--(axis cs:90,0.000861761160194874)
--(axis cs:91,0.000861761160194874)
--(axis cs:92,0.000861761160194874)
--(axis cs:93,0.000689644657541066)
--(axis cs:94,0.000689644657541066)
--(axis cs:95,0.000689644657541066)
--(axis cs:96,0.000689644657541066)
--(axis cs:97,0.000689644657541066)
--(axis cs:98,0.000689644657541066)
--(axis cs:99,0.000689644657541066)
--(axis cs:100,0.000689644657541066)
--(axis cs:101,0.000689644657541066)
--(axis cs:102,0.000689644657541066)
--(axis cs:103,0.000689644657541066)
--(axis cs:104,0.000689644657541066)
--(axis cs:105,0.000689644657541066)
--(axis cs:106,0.000689644657541066)
--(axis cs:107,0.000689644657541066)
--(axis cs:108,0.000689644657541066)
--(axis cs:109,0.000689644657541066)
--(axis cs:110,0.000689644657541066)
--(axis cs:111,0.000689644657541066)
--(axis cs:111,0.0012893620878458)
--(axis cs:111,0.0012893620878458)
--(axis cs:110,0.0012893620878458)
--(axis cs:109,0.0012893620878458)
--(axis cs:108,0.0012893620878458)
--(axis cs:107,0.0012893620878458)
--(axis cs:106,0.0012893620878458)
--(axis cs:105,0.0012893620878458)
--(axis cs:104,0.0012893620878458)
--(axis cs:103,0.0012893620878458)
--(axis cs:102,0.0012893620878458)
--(axis cs:101,0.0012893620878458)
--(axis cs:100,0.0012893620878458)
--(axis cs:99,0.0012893620878458)
--(axis cs:98,0.0012893620878458)
--(axis cs:97,0.0012893620878458)
--(axis cs:96,0.0012893620878458)
--(axis cs:95,0.0012893620878458)
--(axis cs:94,0.0012893620878458)
--(axis cs:93,0.0012893620878458)
--(axis cs:92,0.0015048875939101)
--(axis cs:91,0.0015048875939101)
--(axis cs:90,0.0015048875939101)
--(axis cs:89,0.0015048875939101)
--(axis cs:88,0.0015048875939101)
--(axis cs:87,0.0015048875939101)
--(axis cs:86,0.00156708806753159)
--(axis cs:85,0.00156708806753159)
--(axis cs:84,0.00156708806753159)
--(axis cs:83,0.00156708806753159)
--(axis cs:82,0.00156708806753159)
--(axis cs:81,0.00156708806753159)
--(axis cs:80,0.00156708806753159)
--(axis cs:79,0.00156708806753159)
--(axis cs:78,0.00156708806753159)
--(axis cs:77,0.00156708806753159)
--(axis cs:76,0.00156708806753159)
--(axis cs:75,0.00160638615489006)
--(axis cs:74,0.00160638615489006)
--(axis cs:73,0.00160638615489006)
--(axis cs:72,0.00160638615489006)
--(axis cs:71,0.00160638615489006)
--(axis cs:70,0.00160638615489006)
--(axis cs:69,0.00160638615489006)
--(axis cs:68,0.00160638615489006)
--(axis cs:67,0.00160638615489006)
--(axis cs:66,0.00160638615489006)
--(axis cs:65,0.00160638615489006)
--(axis cs:64,0.00160638615489006)
--(axis cs:63,0.00160638615489006)
--(axis cs:62,0.00160638615489006)
--(axis cs:61,0.00160638615489006)
--(axis cs:60,0.00160638615489006)
--(axis cs:59,0.00160638615489006)
--(axis cs:58,0.00160638615489006)
--(axis cs:57,0.00160638615489006)
--(axis cs:56,0.00160638615489006)
--(axis cs:55,0.00160638615489006)
--(axis cs:54,0.00160638615489006)
--(axis cs:53,0.00160638615489006)
--(axis cs:52,0.00160638615489006)
--(axis cs:51,0.00160638615489006)
--(axis cs:50,0.00160638615489006)
--(axis cs:49,0.00211133621633053)
--(axis cs:48,0.00303330738097429)
--(axis cs:47,0.00303330738097429)
--(axis cs:46,0.00303330738097429)
--(axis cs:45,0.00303330738097429)
--(axis cs:44,0.00327084865421057)
--(axis cs:43,0.00327084865421057)
--(axis cs:42,0.00363598135299981)
--(axis cs:41,0.00366910267621279)
--(axis cs:40,0.00366910267621279)
--(axis cs:39,0.00366910267621279)
--(axis cs:38,0.00366910267621279)
--(axis cs:37,0.00366910267621279)
--(axis cs:36,0.00375942257232964)
--(axis cs:35,0.00392622500658035)
--(axis cs:34,0.00481848558411002)
--(axis cs:33,0.00481848558411002)
--(axis cs:32,0.00481848558411002)
--(axis cs:31,0.00508680241182446)
--(axis cs:30,0.00508680241182446)
--(axis cs:29,0.00842792261391878)
--(axis cs:28,0.00897519756108522)
--(axis cs:27,0.00897519756108522)
--(axis cs:26,0.00897519756108522)
--(axis cs:25,0.00912362989038229)
--(axis cs:24,0.0119350263848901)
--(axis cs:23,0.0174142550677061)
--(axis cs:22,0.0174142550677061)
--(axis cs:21,0.0174142550677061)
--(axis cs:20,0.0180593375116587)
--(axis cs:19,0.01911386474967)
--(axis cs:18,0.0223832223564386)
--(axis cs:17,0.0358803868293762)
--(axis cs:16,0.0413246788084507)
--(axis cs:15,0.0453149043023586)
--(axis cs:14,0.0483066029846668)
--(axis cs:13,0.0521054193377495)
--(axis cs:12,0.0522687546908855)
--(axis cs:11,0.060817152261734)
--(axis cs:10,0.108571708202362)
--cycle;

\path [draw=color1, fill=color1, opacity=0.3]
(axis cs:10,0.108571740173539)
--(axis cs:10,0.0428720774551092)
--(axis cs:11,0.0428177185722702)
--(axis cs:12,0.0395877688902789)
--(axis cs:13,0.0340180875197878)
--(axis cs:14,0.0241163320300674)
--(axis cs:15,0.0194015691646447)
--(axis cs:16,0.0117559038693959)
--(axis cs:17,0.0117559038693959)
--(axis cs:18,0.00961974677133226)
--(axis cs:19,0.00711025294061094)
--(axis cs:20,0.00711025294061094)
--(axis cs:21,0.00711025294061094)
--(axis cs:22,0.00711025294061094)
--(axis cs:23,0.00711025294061094)
--(axis cs:24,0.00652521757209206)
--(axis cs:25,0.00395283368259295)
--(axis cs:26,0.00395283368259295)
--(axis cs:27,0.00316853572378208)
--(axis cs:28,0.00232704601971888)
--(axis cs:29,0.00232704601971888)
--(axis cs:30,0.0021578597130032)
--(axis cs:31,0.00143744213843646)
--(axis cs:32,0.00115841565239813)
--(axis cs:33,0.00100456708487308)
--(axis cs:34,0.000938657858661558)
--(axis cs:35,0.000653745864562669)
--(axis cs:36,0.000653745864562669)
--(axis cs:37,0.000653745864562669)
--(axis cs:38,0.000653745864562669)
--(axis cs:39,0.000653745864562669)
--(axis cs:40,0.000653745864562669)
--(axis cs:41,0.000584892864808668)
--(axis cs:42,0.000584892864808668)
--(axis cs:43,0.000584892864808668)
--(axis cs:44,0.000584892864808668)
--(axis cs:45,0.000584892864808668)
--(axis cs:46,0.000584892864808668)
--(axis cs:47,0.000502110640327112)
--(axis cs:48,0.000480703318177057)
--(axis cs:49,0.00046662756492544)
--(axis cs:50,0.000458788365942624)
--(axis cs:51,0.000458788365942624)
--(axis cs:52,0.000458788365942624)
--(axis cs:53,0.000458788365942624)
--(axis cs:54,0.000458788365942624)
--(axis cs:55,0.000457635341484168)
--(axis cs:56,0.000457635341484168)
--(axis cs:57,0.00038963187509706)
--(axis cs:58,0.00038963187509706)
--(axis cs:59,0.00038963187509706)
--(axis cs:60,0.0003323841130547)
--(axis cs:61,0.0003323841130547)
--(axis cs:62,0.0003323841130547)
--(axis cs:63,0.000305301574750306)
--(axis cs:64,0.000305301574750306)
--(axis cs:65,0.000305301574750306)
--(axis cs:66,0.000305253880774668)
--(axis cs:67,0.000305253880774668)
--(axis cs:68,0.000305253880774668)
--(axis cs:69,0.000296086960308749)
--(axis cs:70,0.000296086960308749)
--(axis cs:71,0.000296086960308749)
--(axis cs:72,0.000296086960308749)
--(axis cs:73,0.000242848596654665)
--(axis cs:74,0.000242848596654665)
--(axis cs:75,0.000242848596654665)
--(axis cs:76,0.000242848596654665)
--(axis cs:77,0.000242848596654665)
--(axis cs:78,0.00022339313242121)
--(axis cs:79,0.00022339313242121)
--(axis cs:80,0.00022339313242121)
--(axis cs:81,0.00022339313242121)
--(axis cs:82,0.000196866300253192)
--(axis cs:83,0.000196866300253192)
--(axis cs:84,0.000196866300253192)
--(axis cs:85,0.000196866300253192)
--(axis cs:86,0.000196866300253192)
--(axis cs:87,0.000196866300253192)
--(axis cs:88,0.000196866300253192)
--(axis cs:89,0.000196866300253192)
--(axis cs:90,0.000196866300253192)
--(axis cs:91,0.000196866300253192)
--(axis cs:92,0.000196866300253192)
--(axis cs:93,0.000161107686317992)
--(axis cs:94,0.000161107686317992)
--(axis cs:95,0.000161107686317992)
--(axis cs:96,0.000161107686317992)
--(axis cs:97,0.000161107686317992)
--(axis cs:98,0.000161107686317992)
--(axis cs:99,0.000157118599707738)
--(axis cs:100,0.000157118599707738)
--(axis cs:101,0.000157118599707738)
--(axis cs:102,0.000157118599707738)
--(axis cs:103,0.000157118599707738)
--(axis cs:104,0.000157118599707738)
--(axis cs:105,0.000153208987404904)
--(axis cs:106,0.000153208987404904)
--(axis cs:107,0.000153208987404904)
--(axis cs:108,0.000148072602019758)
--(axis cs:109,0.000148072602019758)
--(axis cs:110,0.000148072602019758)
--(axis cs:111,0.000148072602019758)
--(axis cs:111,0.000230437269091686)
--(axis cs:111,0.000230437269091686)
--(axis cs:110,0.000230437269091686)
--(axis cs:109,0.000230437269091686)
--(axis cs:108,0.000230437269091686)
--(axis cs:107,0.000234607847338017)
--(axis cs:106,0.000234607847338017)
--(axis cs:105,0.000234607847338017)
--(axis cs:104,0.000266678512387864)
--(axis cs:103,0.000266678512387864)
--(axis cs:102,0.000266678512387864)
--(axis cs:101,0.000266678512387864)
--(axis cs:100,0.000266678512387864)
--(axis cs:99,0.000266678512387864)
--(axis cs:98,0.000269346742554421)
--(axis cs:97,0.000269346742554421)
--(axis cs:96,0.000269346742554421)
--(axis cs:95,0.000269346742554421)
--(axis cs:94,0.000269346742554421)
--(axis cs:93,0.000269346742554421)
--(axis cs:92,0.000371492335379012)
--(axis cs:91,0.000371492335379012)
--(axis cs:90,0.000371492335379012)
--(axis cs:89,0.000371492335379012)
--(axis cs:88,0.000371492335379012)
--(axis cs:87,0.000371492335379012)
--(axis cs:86,0.000371492335379012)
--(axis cs:85,0.000371492335379012)
--(axis cs:84,0.000371492335379012)
--(axis cs:83,0.000371492335379012)
--(axis cs:82,0.000371492335379012)
--(axis cs:81,0.000394983321020427)
--(axis cs:80,0.000394983321020427)
--(axis cs:79,0.000394983321020427)
--(axis cs:78,0.000394983321020427)
--(axis cs:77,0.000467617869941562)
--(axis cs:76,0.000467617869941562)
--(axis cs:75,0.000467617869941562)
--(axis cs:74,0.000467617869941562)
--(axis cs:73,0.000467617869941562)
--(axis cs:72,0.000519372179768009)
--(axis cs:71,0.000519372179768009)
--(axis cs:70,0.000519372179768009)
--(axis cs:69,0.000519372179768009)
--(axis cs:68,0.000549937631776879)
--(axis cs:67,0.000549937631776879)
--(axis cs:66,0.000549937631776879)
--(axis cs:65,0.000549956332648786)
--(axis cs:64,0.000549956332648786)
--(axis cs:63,0.000549956332648786)
--(axis cs:62,0.000603517266777982)
--(axis cs:61,0.000603517266777982)
--(axis cs:60,0.000603517266777982)
--(axis cs:59,0.000649624566478729)
--(axis cs:58,0.000649624566478729)
--(axis cs:57,0.000649624566478729)
--(axis cs:56,0.00068450723774174)
--(axis cs:55,0.00068450723774174)
--(axis cs:54,0.000687506815090837)
--(axis cs:53,0.000687506815090837)
--(axis cs:52,0.000687506815090837)
--(axis cs:51,0.000687506815090837)
--(axis cs:50,0.000687506815090837)
--(axis cs:49,0.000711244686439383)
--(axis cs:48,0.000753659383162851)
--(axis cs:47,0.000955087425078266)
--(axis cs:46,0.00106589539953802)
--(axis cs:45,0.00106589539953802)
--(axis cs:44,0.00106589539953802)
--(axis cs:43,0.00106589539953802)
--(axis cs:42,0.00106589539953802)
--(axis cs:41,0.00106589539953802)
--(axis cs:40,0.00118822597685317)
--(axis cs:39,0.00118822597685317)
--(axis cs:38,0.00118822597685317)
--(axis cs:37,0.00118822597685317)
--(axis cs:36,0.00118822597685317)
--(axis cs:35,0.00118822597685317)
--(axis cs:34,0.00184968222839744)
--(axis cs:33,0.00297081808292824)
--(axis cs:32,0.00414587835651454)
--(axis cs:31,0.00442218577051218)
--(axis cs:30,0.00513497173014198)
--(axis cs:29,0.00538124019838564)
--(axis cs:28,0.00538124019838564)
--(axis cs:27,0.00935934202025622)
--(axis cs:26,0.0100260427739325)
--(axis cs:25,0.0100260427739325)
--(axis cs:24,0.0136477294080606)
--(axis cs:23,0.0151803329570669)
--(axis cs:22,0.0151803329570669)
--(axis cs:21,0.0151803329570669)
--(axis cs:20,0.0151803329570669)
--(axis cs:19,0.0151803329570669)
--(axis cs:18,0.0175959637781743)
--(axis cs:17,0.0195047761073279)
--(axis cs:16,0.0195047761073279)
--(axis cs:15,0.04464104585183)
--(axis cs:14,0.0506462142675425)
--(axis cs:13,0.0658726300899421)
--(axis cs:12,0.0702818124912828)
--(axis cs:11,0.10701582302575)
--(axis cs:10,0.108571740173539)
--cycle;

\addplot [semithick, color1]
table {%
10 0.0757219088143243
11 0.0648188063415176
12 0.0648188063415176
13 0.0595607134257994
14 0.0415305430655915
15 0.0360873868987821
16 0.0283177738143767
17 0.0283177738143767
18 0.027799956796135
19 0.0237826557668202
20 0.0215992371677434
21 0.0164204056349662
22 0.012954785787718
23 0.00940406755370174
24 0.00651626832664751
25 0.00498341815913858
26 0.00476191104763108
27 0.003519881954898
28 0.00337560782786274
29 0.00317743882695589
30 0.00298599364090508
31 0.00190102766730287
32 0.00101097984222729
33 0.0008871304272873
34 0.0008871304272873
35 0.000722827770083552
36 0.000675839315576206
37 0.000533228643456704
38 0.00045800219150823
39 0.000343530299031805
40 0.000343530299031805
41 0.000318227505957792
42 0.000314799386357478
43 0.000226247569716416
44 0.000189656528027394
45 0.000183546964506299
46 0.000183546964506299
47 0.00013582100103567
48 0.00013582100103567
49 0.00013582100103567
50 0.00013582100103567
51 0.000111781100291654
52 0.000108224394570749
53 0.000108224394570749
54 0.000108224394570749
55 0.000108224394570749
56 0.000102752960971003
57 0.000102752960971003
58 0.000102752960971003
59 0.000102752960971003
60 0.000102752960971003
61 0.000102752960971003
62 0.000102752960971003
63 0.000102752960971003
64 8.67555872460605e-05
65 8.67555872460605e-05
66 8.55498106023934e-05
67 7.58301340465876e-05
68 7.58301340465876e-05
69 7.58301340465876e-05
70 7.58301340465876e-05
71 7.44411990886768e-05
72 7.44411990886768e-05
73 7.44411990886768e-05
74 7.44411990886768e-05
75 7.44411990886768e-05
76 7.27787885530764e-05
77 6.97083820794623e-05
78 5.07213215341886e-05
79 5.07213215341886e-05
80 5.07213215341886e-05
81 5.07213215341886e-05
82 5.07213215341886e-05
83 5.07213215341886e-05
84 5.07213215341886e-05
85 5.07213215341886e-05
86 5.07213215341886e-05
87 5.07213215341886e-05
88 5.07213215341886e-05
89 5.07213215341886e-05
90 4.60928042319687e-05
91 4.60928042319687e-05
92 4.60928042319687e-05
93 4.60928042319687e-05
94 4.60928042319687e-05
95 4.60928042319687e-05
96 4.60928042319687e-05
97 4.60928042319687e-05
98 3.1569274960989e-05
99 3.1569274960989e-05
100 3.1569274960989e-05
101 3.1569274960989e-05
102 2.7806663214891e-05
103 2.7806663214891e-05
104 2.7806663214891e-05
105 2.7806663214891e-05
106 2.7806663214891e-05
107 2.7806663214891e-05
108 2.7806663214891e-05
109 2.7806663214891e-05
110 2.7806663214891e-05
111 2.7806663214891e-05
};
\addplot [semithick, blue]
table {%
10 0.075721874833107
11 0.0546261928975582
12 0.0546261928975582
13 0.0546261928975582
14 0.0540501028299332
15 0.0540501028299332
16 0.0531555823981762
17 0.0531555823981762
18 0.0363852120935917
19 0.0322275161743164
20 0.0322275161743164
21 0.0297260675579309
22 0.0297260675579309
23 0.0273582674562931
24 0.0273582674562931
25 0.0273582674562931
26 0.0273582674562931
27 0.0258840322494507
28 0.0258840322494507
29 0.0255535840988159
30 0.0255535840988159
31 0.0167403556406498
32 0.0167403556406498
33 0.0167156122624874
34 0.0167156122624874
35 0.0167156122624874
36 0.00909226480871439
37 0.00909226480871439
38 0.00909226480871439
39 0.00909226480871439
40 0.00909226480871439
41 0.00826754793524742
42 0.00826754793524742
43 0.00826754793524742
44 0.00826754793524742
45 0.00826754793524742
46 0.00826754793524742
47 0.00826754793524742
48 0.00826754793524742
49 0.00826754793524742
50 0.00826754793524742
51 0.00655619287863374
52 0.00655619287863374
53 0.00655604712665081
54 0.00621710903942585
55 0.00621710903942585
56 0.00616619316861033
57 0.00611660210415721
58 0.00546297756955028
59 0.00546297756955028
60 0.00420635286718607
61 0.00306277815252542
62 0.00289691821672022
63 0.00289691821672022
64 0.00289691821672022
65 0.00289691821672022
66 0.00289691821672022
67 0.00289691821672022
68 0.00289691821672022
69 0.00289691821672022
70 0.00289691821672022
71 0.00289691821672022
72 0.00289691821672022
73 0.00289691821672022
74 0.00289691821672022
75 0.0027545690536499
76 0.0027545690536499
77 0.0027545690536499
78 0.00246169837191701
79 0.00246169837191701
80 0.00246169837191701
81 0.00246169837191701
82 0.00246169837191701
83 0.00246169837191701
84 0.00246169837191701
85 0.00246169837191701
86 0.00246169837191701
87 0.00246169837191701
88 0.00246169837191701
89 0.00246169837191701
90 0.00235365494154394
91 0.00232079299166799
92 0.00232079299166799
93 0.00232079299166799
94 0.00232079299166799
95 0.00232079299166799
96 0.00232079299166799
97 0.00232079299166799
98 0.00232079299166799
99 0.00217056274414062
100 0.00217056274414062
101 0.00217056274414062
102 0.00217056274414062
103 0.00217056274414062
104 0.00217056274414062
105 0.00217056274414062
106 0.00217056274414062
107 0.00217056274414062
108 0.00217056274414062
109 0.00217056274414062
110 0.00217056274414062
111 0.00217056274414062
};
\addplot [semithick, blue, dash dot]
table {%
10 0.075721874833107
11 0.0485681109130383
12 0.0409460514783859
13 0.0407705493271351
14 0.0367714948952198
15 0.0349322743713856
16 0.0309834089130163
17 0.0270124468952417
18 0.0185385011136532
19 0.0151531826704741
20 0.0141515862196684
21 0.013600985519588
22 0.013600985519588
23 0.013600985519588
24 0.00953858438879251
25 0.00715062348172069
26 0.00692874845117331
27 0.00692874845117331
28 0.00692874845117331
29 0.00635296758264303
30 0.00393576081842184
31 0.00393576081842184
32 0.00367576535791159
33 0.00367576535791159
34 0.00367576535791159
35 0.00315715209580958
36 0.00297591416165233
37 0.0028628904838115
38 0.0028628904838115
39 0.0028628904838115
40 0.0028628904838115
41 0.0028628904838115
42 0.00281111407093704
43 0.00240170955657959
44 0.00240170955657959
45 0.00215939688496292
46 0.00215939688496292
47 0.00215939688496292
48 0.00215939688496292
49 0.00167797668837011
50 0.00132229598239064
51 0.00132229598239064
52 0.00132229598239064
53 0.00132229598239064
54 0.00132229598239064
55 0.00132229598239064
56 0.00132229598239064
57 0.00132229598239064
58 0.00132229598239064
59 0.00132229598239064
60 0.00132229598239064
61 0.00132229598239064
62 0.00132229598239064
63 0.00132229598239064
64 0.00132229598239064
65 0.00132229598239064
66 0.00132229598239064
67 0.00132229598239064
68 0.00132229598239064
69 0.00132229598239064
70 0.00132229598239064
71 0.00132229598239064
72 0.00132229598239064
73 0.00132229598239064
74 0.00132229598239064
75 0.00132229598239064
76 0.00125808187294751
77 0.00125808187294751
78 0.00125808187294751
79 0.00125808187294751
80 0.00125808187294751
81 0.00125808187294751
82 0.00125808187294751
83 0.00125808187294751
84 0.00125808187294751
85 0.00125808187294751
86 0.00125808187294751
87 0.00118332437705249
88 0.00118332437705249
89 0.00118332437705249
90 0.00118332437705249
91 0.00118332437705249
92 0.00118332437705249
93 0.000989503343589604
94 0.000989503343589604
95 0.000989503343589604
96 0.000989503343589604
97 0.000989503343589604
98 0.000989503343589604
99 0.000989503343589604
100 0.000989503343589604
101 0.000989503343589604
102 0.000989503343589604
103 0.000989503343589604
104 0.000989503343589604
105 0.000989503343589604
106 0.000989503343589604
107 0.000989503343589604
108 0.000989503343589604
109 0.000989503343589604
110 0.000989503343589604
111 0.000989503343589604
};
\addplot [semithick, color1, dash dot]
table {%
10 0.0757219088143243
11 0.07491677079901
12 0.0549347906907809
13 0.049945358804865
14 0.0373812731488049
15 0.0320213075082373
16 0.0156303399883619
17 0.0156303399883619
18 0.0136078552747533
19 0.0111452929488389
20 0.0111452929488389
21 0.0111452929488389
22 0.0111452929488389
23 0.0111452929488389
24 0.0100864734900763
25 0.0069894382282627
26 0.0069894382282627
27 0.00626393887201915
28 0.00385414310905226
29 0.00385414310905226
30 0.00364641572157259
31 0.00292981395447432
32 0.00265214700445634
33 0.00198769258390066
34 0.0013941700435295
35 0.000920985920707921
36 0.000920985920707921
37 0.000920985920707921
38 0.000920985920707921
39 0.000920985920707921
40 0.000920985920707921
41 0.000825394132173344
42 0.000825394132173344
43 0.000825394132173344
44 0.000825394132173344
45 0.000825394132173344
46 0.000825394132173344
47 0.000728599032702689
48 0.000617181350669954
49 0.000588936125682412
50 0.00057314759051673
51 0.00057314759051673
52 0.00057314759051673
53 0.00057314759051673
54 0.00057314759051673
55 0.000571071289612954
56 0.000571071289612954
57 0.000519628220787895
58 0.000519628220787895
59 0.000519628220787895
60 0.000467950689916341
61 0.000467950689916341
62 0.000467950689916341
63 0.000427628953699546
64 0.000427628953699546
65 0.000427628953699546
66 0.000427595756275773
67 0.000427595756275773
68 0.000427595756275773
69 0.000407729570038379
70 0.000407729570038379
71 0.000407729570038379
72 0.000407729570038379
73 0.000355233233298113
74 0.000355233233298113
75 0.000355233233298113
76 0.000355233233298113
77 0.000355233233298113
78 0.000309188226720818
79 0.000309188226720818
80 0.000309188226720818
81 0.000309188226720818
82 0.000284179317816102
83 0.000284179317816102
84 0.000284179317816102
85 0.000284179317816102
86 0.000284179317816102
87 0.000284179317816102
88 0.000284179317816102
89 0.000284179317816102
90 0.000284179317816102
91 0.000284179317816102
92 0.000284179317816102
93 0.000215227214436206
94 0.000215227214436206
95 0.000215227214436206
96 0.000215227214436206
97 0.000215227214436206
98 0.000215227214436206
99 0.000211898556047801
100 0.000211898556047801
101 0.000211898556047801
102 0.000211898556047801
103 0.000211898556047801
104 0.000211898556047801
105 0.00019390841737146
106 0.00019390841737146
107 0.00019390841737146
108 0.000189254935555722
109 0.000189254935555722
110 0.000189254935555722
111 0.000189254935555722
};
\end{axis}

\end{tikzpicture}

%% file: figures/bop_2d_ackley_weak.tex
\begin{tikzpicture}

\definecolor{color0}{rgb}{0,0,1}
\definecolor{color1}{rgb}{1,0.549019607843137,0}
\definecolor{color2}{rgb}{1,0.647058823529412,0}
\definecolor{color3}{rgb}{0.564705882352941,0.933333333333333,0.564705882352941}

\begin{axis}[axis on top,
enlarge x limits=false,
enlarge y limits=false,
height=\figureheight,
scale only axis,
tick align=outside,
tick pos=left,
tick pos=left,
width=\figurewidth,
xlabel={Iteration},
xmin=10, xmax=100,
xtick style={color=black},
xtick={-10,0,10,25,50,75,100},
xticklabels={\ensuremath{-}10,0,10,25,50,75,90},
ymin=-0.25, ymax=4.5,
ytick style={color=black},
ytick={0.   , 4.5},
]
\node[anchor=north east] at (rel axis cs:1,1) {Ackley 2D (weak)};
\path [draw=blue, fill=blue, opacity=0.3]
(axis cs:10,5.34454870223999)
--(axis cs:10,4.39542245864868)
--(axis cs:11,4.08587026596069)
--(axis cs:12,3.33263683319092)
--(axis cs:13,2.96586465835571)
--(axis cs:14,2.6212842464447)
--(axis cs:15,2.21294975280762)
--(axis cs:16,2.09616684913635)
--(axis cs:17,1.80236744880676)
--(axis cs:18,1.80236744880676)
--(axis cs:19,1.64800214767456)
--(axis cs:20,1.61689651012421)
--(axis cs:21,1.6044180393219)
--(axis cs:22,1.42957389354706)
--(axis cs:23,1.32837343215942)
--(axis cs:24,1.18002343177795)
--(axis cs:25,0.866900444030762)
--(axis cs:26,0.803474843502045)
--(axis cs:27,0.587673306465149)
--(axis cs:28,0.489222407341003)
--(axis cs:29,0.395839929580688)
--(axis cs:30,0.352102160453796)
--(axis cs:31,0.352102160453796)
--(axis cs:32,0.339052498340607)
--(axis cs:33,0.339052498340607)
--(axis cs:34,0.33711251616478)
--(axis cs:35,0.33711251616478)
--(axis cs:36,0.290567070245743)
--(axis cs:37,0.290567070245743)
--(axis cs:38,0.256258189678192)
--(axis cs:39,0.235832944512367)
--(axis cs:40,0.211744785308838)
--(axis cs:41,0.211744785308838)
--(axis cs:42,0.211744785308838)
--(axis cs:43,0.199182823300362)
--(axis cs:44,0.199182823300362)
--(axis cs:45,0.188017949461937)
--(axis cs:46,0.176367312669754)
--(axis cs:47,0.176367312669754)
--(axis cs:48,0.176367312669754)
--(axis cs:49,0.176367312669754)
--(axis cs:50,0.175570294260979)
--(axis cs:51,0.169259116053581)
--(axis cs:52,0.16644075512886)
--(axis cs:53,0.16644075512886)
--(axis cs:54,0.16644075512886)
--(axis cs:55,0.161825001239777)
--(axis cs:56,0.161825001239777)
--(axis cs:57,0.161825001239777)
--(axis cs:58,0.161825001239777)
--(axis cs:59,0.161825001239777)
--(axis cs:60,0.161825001239777)
--(axis cs:61,0.161825001239777)
--(axis cs:62,0.161825001239777)
--(axis cs:63,0.161825001239777)
--(axis cs:64,0.161825001239777)
--(axis cs:65,0.161825001239777)
--(axis cs:66,0.161825001239777)
--(axis cs:67,0.144872009754181)
--(axis cs:68,0.144872009754181)
--(axis cs:69,0.144872009754181)
--(axis cs:70,0.144872009754181)
--(axis cs:71,0.144872009754181)
--(axis cs:72,0.144872009754181)
--(axis cs:73,0.144872009754181)
--(axis cs:74,0.135744333267212)
--(axis cs:75,0.135744333267212)
--(axis cs:76,0.135744333267212)
--(axis cs:77,0.135744333267212)
--(axis cs:78,0.135744333267212)
--(axis cs:79,0.135744333267212)
--(axis cs:80,0.128451749682426)
--(axis cs:81,0.128451749682426)
--(axis cs:82,0.128451749682426)
--(axis cs:83,0.128451749682426)
--(axis cs:84,0.128451749682426)
--(axis cs:85,0.128451749682426)
--(axis cs:86,0.128451749682426)
--(axis cs:87,0.119231633841991)
--(axis cs:88,0.117910504341125)
--(axis cs:89,0.117910504341125)
--(axis cs:90,0.117085233330727)
--(axis cs:91,0.117085233330727)
--(axis cs:92,0.117085233330727)
--(axis cs:93,0.117085233330727)
--(axis cs:94,0.115179650485516)
--(axis cs:95,0.115179650485516)
--(axis cs:96,0.115179650485516)
--(axis cs:97,0.0954716876149178)
--(axis cs:98,0.0954716876149178)
--(axis cs:99,0.0954716876149178)
--(axis cs:100,0.0954716876149178)
--(axis cs:101,0.0954716876149178)
--(axis cs:102,0.0954716876149178)
--(axis cs:103,0.0954716876149178)
--(axis cs:104,0.0954716876149178)
--(axis cs:105,0.0954716876149178)
--(axis cs:106,0.0954716876149178)
--(axis cs:107,0.0954716876149178)
--(axis cs:108,0.0954716876149178)
--(axis cs:109,0.0951995402574539)
--(axis cs:110,0.0951995402574539)
--(axis cs:111,0.0911349132657051)
--(axis cs:111,0.147823840379715)
--(axis cs:111,0.147823840379715)
--(axis cs:110,0.160970345139503)
--(axis cs:109,0.160970345139503)
--(axis cs:108,0.161249041557312)
--(axis cs:107,0.161249041557312)
--(axis cs:106,0.161249041557312)
--(axis cs:105,0.161249041557312)
--(axis cs:104,0.161249041557312)
--(axis cs:103,0.161249041557312)
--(axis cs:102,0.161249041557312)
--(axis cs:101,0.161249041557312)
--(axis cs:100,0.161249041557312)
--(axis cs:99,0.161249041557312)
--(axis cs:98,0.161249041557312)
--(axis cs:97,0.161249041557312)
--(axis cs:96,0.177779942750931)
--(axis cs:95,0.177779942750931)
--(axis cs:94,0.177779942750931)
--(axis cs:93,0.179396852850914)
--(axis cs:92,0.179396852850914)
--(axis cs:91,0.179396852850914)
--(axis cs:90,0.179396852850914)
--(axis cs:89,0.180847436189651)
--(axis cs:88,0.180847436189651)
--(axis cs:87,0.18207374215126)
--(axis cs:86,0.219017580151558)
--(axis cs:85,0.219017580151558)
--(axis cs:84,0.219017580151558)
--(axis cs:83,0.219017580151558)
--(axis cs:82,0.219017580151558)
--(axis cs:81,0.219017580151558)
--(axis cs:80,0.219017580151558)
--(axis cs:79,0.223546743392944)
--(axis cs:78,0.223546743392944)
--(axis cs:77,0.223546743392944)
--(axis cs:76,0.223546743392944)
--(axis cs:75,0.223546743392944)
--(axis cs:74,0.223546743392944)
--(axis cs:73,0.233057677745819)
--(axis cs:72,0.233057677745819)
--(axis cs:71,0.233057677745819)
--(axis cs:70,0.233057677745819)
--(axis cs:69,0.233057677745819)
--(axis cs:68,0.233057677745819)
--(axis cs:67,0.233057677745819)
--(axis cs:66,0.24194011092186)
--(axis cs:65,0.24194011092186)
--(axis cs:64,0.24194011092186)
--(axis cs:63,0.24194011092186)
--(axis cs:62,0.24194011092186)
--(axis cs:61,0.24194011092186)
--(axis cs:60,0.24194011092186)
--(axis cs:59,0.24194011092186)
--(axis cs:58,0.24194011092186)
--(axis cs:57,0.24194011092186)
--(axis cs:56,0.24194011092186)
--(axis cs:55,0.24194011092186)
--(axis cs:54,0.245249003171921)
--(axis cs:53,0.245249003171921)
--(axis cs:52,0.245249003171921)
--(axis cs:51,0.24822898209095)
--(axis cs:50,0.252369284629822)
--(axis cs:49,0.252899795770645)
--(axis cs:48,0.252899795770645)
--(axis cs:47,0.252899795770645)
--(axis cs:46,0.252899795770645)
--(axis cs:45,0.272873014211655)
--(axis cs:44,0.28814172744751)
--(axis cs:43,0.28814172744751)
--(axis cs:42,0.307729244232178)
--(axis cs:41,0.307729244232178)
--(axis cs:40,0.307729244232178)
--(axis cs:39,0.333890557289124)
--(axis cs:38,0.341635346412659)
--(axis cs:37,0.369792908430099)
--(axis cs:36,0.369792908430099)
--(axis cs:35,0.408000856637955)
--(axis cs:34,0.408000856637955)
--(axis cs:33,0.40994119644165)
--(axis cs:32,0.40994119644165)
--(axis cs:31,0.52309775352478)
--(axis cs:30,0.52309775352478)
--(axis cs:29,0.582825243473053)
--(axis cs:28,0.748485088348389)
--(axis cs:27,0.898876786231995)
--(axis cs:26,1.20087325572968)
--(axis cs:25,1.45362567901611)
--(axis cs:24,1.82276582717896)
--(axis cs:23,1.94572782516479)
--(axis cs:22,2.08311080932617)
--(axis cs:21,2.22965097427368)
--(axis cs:20,2.26629662513733)
--(axis cs:19,2.28965663909912)
--(axis cs:18,2.51376557350159)
--(axis cs:17,2.51376557350159)
--(axis cs:16,2.7420756816864)
--(axis cs:15,2.84814929962158)
--(axis cs:14,3.30727219581604)
--(axis cs:13,3.77549743652344)
--(axis cs:12,4.21727180480957)
--(axis cs:11,4.86336946487427)
--(axis cs:10,5.34454870223999)
--cycle;

\path [draw=color1, fill=color1, opacity=0.3]
(axis cs:10,5.34454898518408)
--(axis cs:10,4.39542274790918)
--(axis cs:11,3.6289684241036)
--(axis cs:12,2.67040828688908)
--(axis cs:13,1.74288130318632)
--(axis cs:14,1.578569085962)
--(axis cs:15,1.56566143117706)
--(axis cs:16,1.48925416184582)
--(axis cs:17,1.3512262974579)
--(axis cs:18,1.22266555372149)
--(axis cs:19,1.00348159960789)
--(axis cs:20,1.00133630091561)
--(axis cs:21,0.847953057632869)
--(axis cs:22,0.820865529032643)
--(axis cs:23,0.604033324062207)
--(axis cs:24,0.506977023089549)
--(axis cs:25,0.506783345271204)
--(axis cs:26,0.506783345271204)
--(axis cs:27,0.506783345271204)
--(axis cs:28,0.505245092737852)
--(axis cs:29,0.505245092737852)
--(axis cs:30,0.505245092737852)
--(axis cs:31,0.503262741829055)
--(axis cs:32,0.422960343062214)
--(axis cs:33,0.422960343062214)
--(axis cs:34,0.420030771192855)
--(axis cs:35,0.420030771192855)
--(axis cs:36,0.420030771192855)
--(axis cs:37,0.404742084997465)
--(axis cs:38,0.372319683944699)
--(axis cs:39,0.372319683944699)
--(axis cs:40,0.372319683944699)
--(axis cs:41,0.372319683944699)
--(axis cs:42,0.372319683944699)
--(axis cs:43,0.372319683944699)
--(axis cs:44,0.372319683944699)
--(axis cs:45,0.372319683944699)
--(axis cs:46,0.372319683944699)
--(axis cs:47,0.372319683944699)
--(axis cs:48,0.372319683944699)
--(axis cs:49,0.372319683944699)
--(axis cs:50,0.372319683944699)
--(axis cs:51,0.372319683944699)
--(axis cs:52,0.372319683944699)
--(axis cs:53,0.372319683944699)
--(axis cs:54,0.372319683944699)
--(axis cs:55,0.347098719942263)
--(axis cs:56,0.332176202763212)
--(axis cs:57,0.315075626359923)
--(axis cs:58,0.278889466401122)
--(axis cs:59,0.278889466401122)
--(axis cs:60,0.278889466401122)
--(axis cs:61,0.278889466401122)
--(axis cs:62,0.270482923766481)
--(axis cs:63,0.270482923766481)
--(axis cs:64,0.270482923766481)
--(axis cs:65,0.270482923766481)
--(axis cs:66,0.270482923766481)
--(axis cs:67,0.270482923766481)
--(axis cs:68,0.270482923766481)
--(axis cs:69,0.270482923766481)
--(axis cs:70,0.270482923766481)
--(axis cs:71,0.270482923766481)
--(axis cs:72,0.270482923766481)
--(axis cs:73,0.270482923766481)
--(axis cs:74,0.231820322441947)
--(axis cs:75,0.231820322441947)
--(axis cs:76,0.231820322441947)
--(axis cs:77,0.231820322441947)
--(axis cs:78,0.231820322441947)
--(axis cs:79,0.231820322441947)
--(axis cs:80,0.231820322441947)
--(axis cs:81,0.231820322441947)
--(axis cs:82,0.231820322441947)
--(axis cs:83,0.231820322441947)
--(axis cs:84,0.231820322441947)
--(axis cs:85,0.231820322441947)
--(axis cs:86,0.231820322441947)
--(axis cs:87,0.231820322441947)
--(axis cs:88,0.231820322441947)
--(axis cs:89,0.204535943851707)
--(axis cs:90,0.204535943851707)
--(axis cs:91,0.204535943851707)
--(axis cs:92,0.204535943851707)
--(axis cs:93,0.204535943851707)
--(axis cs:94,0.204535943851707)
--(axis cs:95,0.204535943851707)
--(axis cs:96,0.204535943851707)
--(axis cs:97,0.204535943851707)
--(axis cs:98,0.183062314753404)
--(axis cs:99,0.183062314753404)
--(axis cs:100,0.183062314753404)
--(axis cs:101,0.183062314753404)
--(axis cs:102,0.183062314753404)
--(axis cs:103,0.183062314753404)
--(axis cs:104,0.178615198907888)
--(axis cs:105,0.178615198907888)
--(axis cs:106,0.178615198907888)
--(axis cs:107,0.178615198907888)
--(axis cs:108,0.178615198907888)
--(axis cs:109,0.170314391250086)
--(axis cs:110,0.170314391250086)
--(axis cs:111,0.170314391250086)
--(axis cs:111,0.249484894608873)
--(axis cs:111,0.249484894608873)
--(axis cs:110,0.249484894608873)
--(axis cs:109,0.249484894608873)
--(axis cs:108,0.258748363413119)
--(axis cs:107,0.258748363413119)
--(axis cs:106,0.258748363413119)
--(axis cs:105,0.258748363413119)
--(axis cs:104,0.258748363413119)
--(axis cs:103,0.276441266720562)
--(axis cs:102,0.276441266720562)
--(axis cs:101,0.276441266720562)
--(axis cs:100,0.276441266720562)
--(axis cs:99,0.276441266720562)
--(axis cs:98,0.276441266720562)
--(axis cs:97,0.293486711692242)
--(axis cs:96,0.293486711692242)
--(axis cs:95,0.293486711692242)
--(axis cs:94,0.293486711692242)
--(axis cs:93,0.293486711692242)
--(axis cs:92,0.293486711692242)
--(axis cs:91,0.293486711692242)
--(axis cs:90,0.293486711692242)
--(axis cs:89,0.293486711692242)
--(axis cs:88,0.324901922804794)
--(axis cs:87,0.324901922804794)
--(axis cs:86,0.324901922804794)
--(axis cs:85,0.324901922804794)
--(axis cs:84,0.324901922804794)
--(axis cs:83,0.324901922804794)
--(axis cs:82,0.324901922804794)
--(axis cs:81,0.324901922804794)
--(axis cs:80,0.324901922804794)
--(axis cs:79,0.324901922804794)
--(axis cs:78,0.324901922804794)
--(axis cs:77,0.324901922804794)
--(axis cs:76,0.324901922804794)
--(axis cs:75,0.324901922804794)
--(axis cs:74,0.324901922804794)
--(axis cs:73,0.386166448411359)
--(axis cs:72,0.386166448411359)
--(axis cs:71,0.386166448411359)
--(axis cs:70,0.386166448411359)
--(axis cs:69,0.386166448411359)
--(axis cs:68,0.386166448411359)
--(axis cs:67,0.386166448411359)
--(axis cs:66,0.386166448411359)
--(axis cs:65,0.386166448411359)
--(axis cs:64,0.386166448411359)
--(axis cs:63,0.386166448411359)
--(axis cs:62,0.386166448411359)
--(axis cs:61,0.394504391376847)
--(axis cs:60,0.394504391376847)
--(axis cs:59,0.394504391376847)
--(axis cs:58,0.394504391376847)
--(axis cs:57,0.429537235876386)
--(axis cs:56,0.61472514168542)
--(axis cs:55,0.632885477886103)
--(axis cs:54,0.660959008943248)
--(axis cs:53,0.660959008943248)
--(axis cs:52,0.660959008943248)
--(axis cs:51,0.660959008943248)
--(axis cs:50,0.660959008943248)
--(axis cs:49,0.660959008943248)
--(axis cs:48,0.660959008943248)
--(axis cs:47,0.660959008943248)
--(axis cs:46,0.660959008943248)
--(axis cs:45,0.660959008943248)
--(axis cs:44,0.660959008943248)
--(axis cs:43,0.660959008943248)
--(axis cs:42,0.660959008943248)
--(axis cs:41,0.660959008943248)
--(axis cs:40,0.660959008943248)
--(axis cs:39,0.660959008943248)
--(axis cs:38,0.660959008943248)
--(axis cs:37,0.679647863839193)
--(axis cs:36,0.692499963356627)
--(axis cs:35,0.692499963356627)
--(axis cs:34,0.692499963356627)
--(axis cs:33,0.773017917053565)
--(axis cs:32,0.773017917053565)
--(axis cs:31,0.859495986526636)
--(axis cs:30,0.862419654689817)
--(axis cs:29,0.862419654689817)
--(axis cs:28,0.862419654689817)
--(axis cs:27,0.863343870398705)
--(axis cs:26,0.863343870398705)
--(axis cs:25,0.863343870398705)
--(axis cs:24,0.866259861644085)
--(axis cs:23,1.02386528882509)
--(axis cs:22,1.34831319036524)
--(axis cs:21,1.38569978697793)
--(axis cs:20,1.50384680344297)
--(axis cs:19,1.51845306948585)
--(axis cs:18,1.95894375794927)
--(axis cs:17,2.11202425455636)
--(axis cs:16,2.39887462956592)
--(axis cs:15,2.4534485707276)
--(axis cs:14,2.49522411190336)
--(axis cs:13,2.6610692949812)
--(axis cs:12,3.45698848749798)
--(axis cs:11,4.62052352892509)
--(axis cs:10,5.34454898518408)
--cycle;

\path [draw=blue, fill=blue, opacity=0.3]
(axis cs:10,5.34454870223999)
--(axis cs:10,4.39542245864868)
--(axis cs:11,3.25274872779846)
--(axis cs:12,3.20437002182007)
--(axis cs:13,2.74756813049316)
--(axis cs:14,2.65860295295715)
--(axis cs:15,2.5920078754425)
--(axis cs:16,2.09870982170105)
--(axis cs:17,2.00992155075073)
--(axis cs:18,1.56103312969208)
--(axis cs:19,1.56103312969208)
--(axis cs:20,1.43367409706116)
--(axis cs:21,1.36366438865662)
--(axis cs:22,1.36366438865662)
--(axis cs:23,1.20738458633423)
--(axis cs:24,1.20738458633423)
--(axis cs:25,1.20738458633423)
--(axis cs:26,1.14181280136108)
--(axis cs:27,0.956552922725677)
--(axis cs:28,0.902479708194733)
--(axis cs:29,0.902479708194733)
--(axis cs:30,0.83898138999939)
--(axis cs:31,0.83898138999939)
--(axis cs:32,0.782922744750977)
--(axis cs:33,0.782922744750977)
--(axis cs:34,0.782922744750977)
--(axis cs:35,0.699203252792358)
--(axis cs:36,0.699203252792358)
--(axis cs:37,0.699203252792358)
--(axis cs:38,0.699203252792358)
--(axis cs:39,0.699203252792358)
--(axis cs:40,0.699203252792358)
--(axis cs:41,0.699203252792358)
--(axis cs:42,0.670929312705994)
--(axis cs:43,0.670929312705994)
--(axis cs:44,0.670929312705994)
--(axis cs:45,0.670929312705994)
--(axis cs:46,0.670929312705994)
--(axis cs:47,0.670929312705994)
--(axis cs:48,0.670929312705994)
--(axis cs:49,0.670929312705994)
--(axis cs:50,0.670929312705994)
--(axis cs:51,0.670929312705994)
--(axis cs:52,0.653396308422089)
--(axis cs:53,0.653396308422089)
--(axis cs:54,0.653396308422089)
--(axis cs:55,0.653396308422089)
--(axis cs:56,0.653396308422089)
--(axis cs:57,0.63684219121933)
--(axis cs:58,0.63684219121933)
--(axis cs:59,0.612308502197266)
--(axis cs:60,0.573007822036743)
--(axis cs:61,0.573007822036743)
--(axis cs:62,0.573007822036743)
--(axis cs:63,0.573007822036743)
--(axis cs:64,0.573007822036743)
--(axis cs:65,0.573007822036743)
--(axis cs:66,0.573007822036743)
--(axis cs:67,0.573007822036743)
--(axis cs:68,0.573007822036743)
--(axis cs:69,0.573007822036743)
--(axis cs:70,0.540430307388306)
--(axis cs:71,0.50381338596344)
--(axis cs:72,0.50381338596344)
--(axis cs:73,0.50381338596344)
--(axis cs:74,0.50381338596344)
--(axis cs:75,0.503037750720978)
--(axis cs:76,0.503037750720978)
--(axis cs:77,0.503037750720978)
--(axis cs:78,0.470024585723877)
--(axis cs:79,0.444889068603516)
--(axis cs:80,0.444889068603516)
--(axis cs:81,0.444889068603516)
--(axis cs:82,0.444889068603516)
--(axis cs:83,0.413557022809982)
--(axis cs:84,0.413557022809982)
--(axis cs:85,0.379786342382431)
--(axis cs:86,0.379786342382431)
--(axis cs:87,0.379786342382431)
--(axis cs:88,0.379786342382431)
--(axis cs:89,0.379786342382431)
--(axis cs:90,0.379786342382431)
--(axis cs:91,0.379786342382431)
--(axis cs:92,0.379786342382431)
--(axis cs:93,0.379786342382431)
--(axis cs:94,0.379786342382431)
--(axis cs:95,0.357746243476868)
--(axis cs:96,0.357746243476868)
--(axis cs:97,0.357746243476868)
--(axis cs:98,0.357746243476868)
--(axis cs:99,0.357746243476868)
--(axis cs:100,0.357746243476868)
--(axis cs:101,0.357746243476868)
--(axis cs:102,0.357746243476868)
--(axis cs:103,0.356353104114532)
--(axis cs:104,0.356353104114532)
--(axis cs:105,0.353591084480286)
--(axis cs:106,0.353591084480286)
--(axis cs:107,0.353591084480286)
--(axis cs:108,0.353591084480286)
--(axis cs:109,0.353591084480286)
--(axis cs:110,0.351911097764969)
--(axis cs:111,0.351911097764969)
--(axis cs:111,0.697235226631165)
--(axis cs:111,0.697235226631165)
--(axis cs:110,0.697235226631165)
--(axis cs:109,0.747981309890747)
--(axis cs:108,0.747981309890747)
--(axis cs:107,0.747981309890747)
--(axis cs:106,0.747981309890747)
--(axis cs:105,0.747981309890747)
--(axis cs:104,0.750645339488983)
--(axis cs:103,0.750645339488983)
--(axis cs:102,0.804701089859009)
--(axis cs:101,0.804701089859009)
--(axis cs:100,0.804701089859009)
--(axis cs:99,0.804701089859009)
--(axis cs:98,0.804701089859009)
--(axis cs:97,0.804701089859009)
--(axis cs:96,0.804701089859009)
--(axis cs:95,0.804701089859009)
--(axis cs:94,0.824463963508606)
--(axis cs:93,0.824463963508606)
--(axis cs:92,0.824463963508606)
--(axis cs:91,0.824463963508606)
--(axis cs:90,0.824463963508606)
--(axis cs:89,0.824463963508606)
--(axis cs:88,0.824463963508606)
--(axis cs:87,0.824463963508606)
--(axis cs:86,0.824463963508606)
--(axis cs:85,0.824463963508606)
--(axis cs:84,0.857893705368042)
--(axis cs:83,0.857893705368042)
--(axis cs:82,0.900437116622925)
--(axis cs:81,0.900437116622925)
--(axis cs:80,0.900437116622925)
--(axis cs:79,0.900437116622925)
--(axis cs:78,0.917502164840698)
--(axis cs:77,0.937764227390289)
--(axis cs:76,0.937764227390289)
--(axis cs:75,0.937764227390289)
--(axis cs:74,0.938430547714233)
--(axis cs:73,0.938430547714233)
--(axis cs:72,0.938430547714233)
--(axis cs:71,0.938430547714233)
--(axis cs:70,0.958979964256287)
--(axis cs:69,0.990558743476868)
--(axis cs:68,0.990558743476868)
--(axis cs:67,0.990558743476868)
--(axis cs:66,0.990558743476868)
--(axis cs:65,0.990558743476868)
--(axis cs:64,0.990558743476868)
--(axis cs:63,0.990558743476868)
--(axis cs:62,0.990558743476868)
--(axis cs:61,0.990558743476868)
--(axis cs:60,0.990558743476868)
--(axis cs:59,1.0201553106308)
--(axis cs:58,1.04479849338531)
--(axis cs:57,1.04479849338531)
--(axis cs:56,1.07045555114746)
--(axis cs:55,1.07045555114746)
--(axis cs:54,1.07045555114746)
--(axis cs:53,1.07045555114746)
--(axis cs:52,1.07045555114746)
--(axis cs:51,1.09023177623749)
--(axis cs:50,1.09023177623749)
--(axis cs:49,1.09023177623749)
--(axis cs:48,1.09023177623749)
--(axis cs:47,1.09023177623749)
--(axis cs:46,1.09023177623749)
--(axis cs:45,1.09023177623749)
--(axis cs:44,1.09023177623749)
--(axis cs:43,1.09023177623749)
--(axis cs:42,1.09023177623749)
--(axis cs:41,1.10772347450256)
--(axis cs:40,1.10772347450256)
--(axis cs:39,1.10772347450256)
--(axis cs:38,1.10772347450256)
--(axis cs:37,1.10772347450256)
--(axis cs:36,1.10772347450256)
--(axis cs:35,1.10772347450256)
--(axis cs:34,1.26304578781128)
--(axis cs:33,1.26304578781128)
--(axis cs:32,1.26304578781128)
--(axis cs:31,1.30831933021545)
--(axis cs:30,1.30831933021545)
--(axis cs:29,1.42933654785156)
--(axis cs:28,1.42933654785156)
--(axis cs:27,1.49171280860901)
--(axis cs:26,1.72978520393372)
--(axis cs:25,1.89570140838623)
--(axis cs:24,1.89570140838623)
--(axis cs:23,1.89570140838623)
--(axis cs:22,2.01567673683167)
--(axis cs:21,2.01567673683167)
--(axis cs:20,2.12208008766174)
--(axis cs:19,2.23354005813599)
--(axis cs:18,2.23354005813599)
--(axis cs:17,2.55242300033569)
--(axis cs:16,2.6235773563385)
--(axis cs:15,3.05139708518982)
--(axis cs:14,3.15842127799988)
--(axis cs:13,3.45379447937012)
--(axis cs:12,4.11530637741089)
--(axis cs:11,4.1632285118103)
--(axis cs:10,5.34454870223999)
--cycle;

\path [draw=color1, fill=color1, opacity=0.3]
(axis cs:10,5.34454898518408)
--(axis cs:10,4.39542274790918)
--(axis cs:11,3.79801829122048)
--(axis cs:12,3.10696664092773)
--(axis cs:13,2.64762193809687)
--(axis cs:14,2.31898617193081)
--(axis cs:15,1.7597124123334)
--(axis cs:16,1.52327282774938)
--(axis cs:17,1.1214806164081)
--(axis cs:18,0.946946091436208)
--(axis cs:19,0.931903697305733)
--(axis cs:20,0.906054400155405)
--(axis cs:21,0.695256814183608)
--(axis cs:22,0.548791802576466)
--(axis cs:23,0.490233209991641)
--(axis cs:24,0.490233209991641)
--(axis cs:25,0.490233209991641)
--(axis cs:26,0.410050374598355)
--(axis cs:27,0.373687956133197)
--(axis cs:28,0.372047510702013)
--(axis cs:29,0.320057653262002)
--(axis cs:30,0.309158779950114)
--(axis cs:31,0.291164567880433)
--(axis cs:32,0.287103606838863)
--(axis cs:33,0.207639671016859)
--(axis cs:34,0.181547616100683)
--(axis cs:35,0.137184006832133)
--(axis cs:36,0.129090147651805)
--(axis cs:37,0.0800872065958552)
--(axis cs:38,0.0680583715138268)
--(axis cs:39,0.0583345527450068)
--(axis cs:40,0.0558991842226313)
--(axis cs:41,0.0558991842226313)
--(axis cs:42,0.0558991842226313)
--(axis cs:43,0.0558991842226313)
--(axis cs:44,0.0413559512744793)
--(axis cs:45,0.0413559512744793)
--(axis cs:46,0.0406896647494604)
--(axis cs:47,0.0406896647494604)
--(axis cs:48,0.0406896647494604)
--(axis cs:49,0.0406896647494604)
--(axis cs:50,0.0406896647494604)
--(axis cs:51,0.0406896647494604)
--(axis cs:52,0.0406896647494604)
--(axis cs:53,0.0406896647494604)
--(axis cs:54,0.0406896647494604)
--(axis cs:55,0.0355938420202196)
--(axis cs:56,0.0321831699030099)
--(axis cs:57,0.0321831699030099)
--(axis cs:58,0.0300843483870104)
--(axis cs:59,0.0300843483870104)
--(axis cs:60,0.0275216076559612)
--(axis cs:61,0.0275216076559612)
--(axis cs:62,0.0275216076559612)
--(axis cs:63,0.0275216076559612)
--(axis cs:64,0.0275216076559612)
--(axis cs:65,0.0275216076559612)
--(axis cs:66,0.0275216076559612)
--(axis cs:67,0.0275216076559612)
--(axis cs:68,0.0275216076559612)
--(axis cs:69,0.0275216076559612)
--(axis cs:70,0.0275216076559612)
--(axis cs:71,0.0275216076559612)
--(axis cs:72,0.0275216076559612)
--(axis cs:73,0.0275216076559612)
--(axis cs:74,0.0275216076559612)
--(axis cs:75,0.0275216076559612)
--(axis cs:76,0.0275216076559612)
--(axis cs:77,0.0275216076559612)
--(axis cs:78,0.0254744741161235)
--(axis cs:79,0.0254744741161235)
--(axis cs:80,0.0254744741161235)
--(axis cs:81,0.0254744741161235)
--(axis cs:82,0.0254744741161235)
--(axis cs:83,0.0254744741161235)
--(axis cs:84,0.0254744741161235)
--(axis cs:85,0.0254744741161235)
--(axis cs:86,0.0254744741161235)
--(axis cs:87,0.0254744741161235)
--(axis cs:88,0.0254744741161235)
--(axis cs:89,0.0254744741161235)
--(axis cs:90,0.0254744741161235)
--(axis cs:91,0.0254744741161235)
--(axis cs:92,0.0254744741161235)
--(axis cs:93,0.0254744741161235)
--(axis cs:94,0.0254744741161235)
--(axis cs:95,0.0226335551013367)
--(axis cs:96,0.0226335551013367)
--(axis cs:97,0.0226335551013367)
--(axis cs:98,0.0224770405713156)
--(axis cs:99,0.0224770405713156)
--(axis cs:100,0.0224770405713156)
--(axis cs:101,0.0224770405713156)
--(axis cs:102,0.0224770405713156)
--(axis cs:103,0.0224770405713156)
--(axis cs:104,0.0224770405713156)
--(axis cs:105,0.0224770405713156)
--(axis cs:106,0.0224770405713156)
--(axis cs:107,0.0224770405713156)
--(axis cs:108,0.0224770405713156)
--(axis cs:109,0.0224770405713156)
--(axis cs:110,0.0224770405713156)
--(axis cs:111,0.0224770405713156)
--(axis cs:111,0.0281037607267119)
--(axis cs:111,0.0281037607267119)
--(axis cs:110,0.0281037607267119)
--(axis cs:109,0.0281037607267119)
--(axis cs:108,0.0281037607267119)
--(axis cs:107,0.0281037607267119)
--(axis cs:106,0.0281037607267119)
--(axis cs:105,0.0281037607267119)
--(axis cs:104,0.0281037607267119)
--(axis cs:103,0.0281037607267119)
--(axis cs:102,0.0281037607267119)
--(axis cs:101,0.0281037607267119)
--(axis cs:100,0.0281037607267119)
--(axis cs:99,0.0281037607267119)
--(axis cs:98,0.0281037607267119)
--(axis cs:97,0.028488835220956)
--(axis cs:96,0.028488835220956)
--(axis cs:95,0.028488835220956)
--(axis cs:94,0.0302936478063377)
--(axis cs:93,0.0302936478063377)
--(axis cs:92,0.0302936478063377)
--(axis cs:91,0.0302936478063377)
--(axis cs:90,0.0302936478063377)
--(axis cs:89,0.0302936478063377)
--(axis cs:88,0.0302936478063377)
--(axis cs:87,0.0302936478063377)
--(axis cs:86,0.0302936478063377)
--(axis cs:85,0.0302936478063377)
--(axis cs:84,0.0302936478063377)
--(axis cs:83,0.0302936478063377)
--(axis cs:82,0.0302936478063377)
--(axis cs:81,0.0302936478063377)
--(axis cs:80,0.0302936478063377)
--(axis cs:79,0.0302936478063377)
--(axis cs:78,0.0302936478063377)
--(axis cs:77,0.0375230359480721)
--(axis cs:76,0.0375230359480721)
--(axis cs:75,0.0375230359480721)
--(axis cs:74,0.0375230359480721)
--(axis cs:73,0.0375230359480721)
--(axis cs:72,0.0375230359480721)
--(axis cs:71,0.0375230359480721)
--(axis cs:70,0.0375230359480721)
--(axis cs:69,0.0375230359480721)
--(axis cs:68,0.0375230359480721)
--(axis cs:67,0.0375230359480721)
--(axis cs:66,0.0375230359480721)
--(axis cs:65,0.0375230359480721)
--(axis cs:64,0.0375230359480721)
--(axis cs:63,0.0375230359480721)
--(axis cs:62,0.0375230359480721)
--(axis cs:61,0.0375230359480721)
--(axis cs:60,0.0375230359480721)
--(axis cs:59,0.0415566773647441)
--(axis cs:58,0.0415566773647441)
--(axis cs:57,0.0451927331009955)
--(axis cs:56,0.0451927331009955)
--(axis cs:55,0.0484775125937898)
--(axis cs:54,0.0602537361802732)
--(axis cs:53,0.0602537361802732)
--(axis cs:52,0.0602537361802732)
--(axis cs:51,0.0602537361802732)
--(axis cs:50,0.0602537361802732)
--(axis cs:49,0.0602537361802732)
--(axis cs:48,0.0602537361802732)
--(axis cs:47,0.0602537361802732)
--(axis cs:46,0.0602537361802732)
--(axis cs:45,0.0606326979051477)
--(axis cs:44,0.0606326979051477)
--(axis cs:43,0.0964090944048002)
--(axis cs:42,0.0964090944048002)
--(axis cs:41,0.0964090944048002)
--(axis cs:40,0.0964090944048002)
--(axis cs:39,0.105741284298052)
--(axis cs:38,0.529219398862469)
--(axis cs:37,0.539512302936807)
--(axis cs:36,0.714275674460843)
--(axis cs:35,0.721309799074853)
--(axis cs:34,0.761974244673456)
--(axis cs:33,0.786132230164434)
--(axis cs:32,0.874540438042826)
--(axis cs:31,0.878485119807859)
--(axis cs:30,0.891762148947316)
--(axis cs:29,0.899237189764412)
--(axis cs:28,0.946295515846937)
--(axis cs:27,0.967413198444871)
--(axis cs:26,0.99974464609225)
--(axis cs:25,1.11383597889163)
--(axis cs:24,1.11383597889163)
--(axis cs:23,1.11383597889163)
--(axis cs:22,1.1626486668628)
--(axis cs:21,1.27644852263513)
--(axis cs:20,1.48453551765438)
--(axis cs:19,1.50385540699392)
--(axis cs:18,1.53905072244142)
--(axis cs:17,1.66397377101484)
--(axis cs:16,2.19465740149047)
--(axis cs:15,2.86275170165322)
--(axis cs:14,3.39214668581982)
--(axis cs:13,3.75751493692192)
--(axis cs:12,4.27205830737735)
--(axis cs:11,4.92516478553489)
--(axis cs:10,5.34454898518408)
--cycle;

\addplot [semithick, blue, dash dot]
table {%
10 4.86998558044434
11 4.47461986541748
12 3.77495431900024
13 3.37068104743958
14 2.96427822113037
15 2.5305495262146
16 2.41912126541138
17 2.15806651115417
18 2.15806651115417
19 1.96882939338684
20 1.94159662723541
21 1.91703450679779
22 1.75634229183197
23 1.63705062866211
24 1.50139462947845
25 1.16026306152344
26 1.00217401981354
27 0.743275046348572
28 0.618853747844696
29 0.489332586526871
30 0.437599956989288
31 0.437599956989288
32 0.374496847391129
33 0.374496847391129
34 0.372556686401367
35 0.372556686401367
36 0.330179989337921
37 0.330179989337921
38 0.298946768045425
39 0.284861743450165
40 0.259737014770508
41 0.259737014770508
42 0.259737014770508
43 0.243662267923355
44 0.243662267923355
45 0.230445474386215
46 0.2146335542202
47 0.2146335542202
48 0.2146335542202
49 0.2146335542202
50 0.213969796895981
51 0.208744049072266
52 0.205844879150391
53 0.205844879150391
54 0.205844879150391
55 0.201882556080818
56 0.201882556080818
57 0.201882556080818
58 0.201882556080818
59 0.201882556080818
60 0.201882556080818
61 0.201882556080818
62 0.201882556080818
63 0.201882556080818
64 0.201882556080818
65 0.201882556080818
66 0.201882556080818
67 0.18896484375
68 0.18896484375
69 0.18896484375
70 0.18896484375
71 0.18896484375
72 0.18896484375
73 0.18896484375
74 0.179645538330078
75 0.179645538330078
76 0.179645538330078
77 0.179645538330078
78 0.179645538330078
79 0.179645538330078
80 0.173734664916992
81 0.173734664916992
82 0.173734664916992
83 0.173734664916992
84 0.173734664916992
85 0.173734664916992
86 0.173734664916992
87 0.150652691721916
88 0.149378970265388
89 0.149378970265388
90 0.14824104309082
91 0.14824104309082
92 0.14824104309082
93 0.14824104309082
94 0.146479800343513
95 0.146479800343513
96 0.146479800343513
97 0.128360360860825
98 0.128360360860825
99 0.128360360860825
100 0.128360360860825
101 0.128360360860825
102 0.128360360860825
103 0.128360360860825
104 0.128360360860825
105 0.128360360860825
106 0.128360360860825
107 0.128360360860825
108 0.128360360860825
109 0.128084942698479
110 0.128084942698479
111 0.11947937309742
};
\addplot [semithick, color1, dash dot]
table {%
10 4.86998586654663
11 4.12474597651434
12 3.06369838719353
13 2.20197529908376
14 2.03689659893268
15 2.00955500095233
16 1.94406439570587
17 1.73162527600713
18 1.59080465583538
19 1.26096733454687
20 1.25259155217929
21 1.1168264223054
22 1.08458935969894
23 0.813949306443646
24 0.686618442366817
25 0.685063607834954
26 0.685063607834954
27 0.685063607834954
28 0.683832373713834
29 0.683832373713834
30 0.683832373713834
31 0.681379364177845
32 0.59798913005789
33 0.59798913005789
34 0.556265367274741
35 0.556265367274741
36 0.556265367274741
37 0.542194974418329
38 0.516639346443974
39 0.516639346443974
40 0.516639346443974
41 0.516639346443974
42 0.516639346443974
43 0.516639346443974
44 0.516639346443974
45 0.516639346443974
46 0.516639346443974
47 0.516639346443974
48 0.516639346443974
49 0.516639346443974
50 0.516639346443974
51 0.516639346443974
52 0.516639346443974
53 0.516639346443974
54 0.516639346443974
55 0.489992098914183
56 0.473450672224316
57 0.372306431118155
58 0.336696928888984
59 0.336696928888984
60 0.336696928888984
61 0.336696928888984
62 0.32832468608892
63 0.32832468608892
64 0.32832468608892
65 0.32832468608892
66 0.32832468608892
67 0.32832468608892
68 0.32832468608892
69 0.32832468608892
70 0.32832468608892
71 0.32832468608892
72 0.32832468608892
73 0.32832468608892
74 0.27836112262337
75 0.27836112262337
76 0.27836112262337
77 0.27836112262337
78 0.27836112262337
79 0.27836112262337
80 0.27836112262337
81 0.27836112262337
82 0.27836112262337
83 0.27836112262337
84 0.27836112262337
85 0.27836112262337
86 0.27836112262337
87 0.27836112262337
88 0.27836112262337
89 0.249011327771974
90 0.249011327771974
91 0.249011327771974
92 0.249011327771974
93 0.249011327771974
94 0.249011327771974
95 0.249011327771974
96 0.249011327771974
97 0.249011327771974
98 0.229751790736983
99 0.229751790736983
100 0.229751790736983
101 0.229751790736983
102 0.229751790736983
103 0.229751790736983
104 0.218681781160504
105 0.218681781160504
106 0.218681781160504
107 0.218681781160504
108 0.218681781160504
109 0.20989964292948
110 0.20989964292948
111 0.20989964292948
};
\addplot [semithick, blue]
table {%
10 4.86998558044434
11 3.70798873901367
12 3.65983819961548
13 3.10068130493164
14 2.90851211547852
15 2.82170248031616
16 2.36114358901978
17 2.28117227554321
18 1.89728665351868
19 1.89728665351868
20 1.77787709236145
21 1.68967056274414
22 1.68967056274414
23 1.55154299736023
24 1.55154299736023
25 1.55154299736023
26 1.4357990026474
27 1.22413289546967
28 1.16590809822083
29 1.16590809822083
30 1.07365036010742
31 1.07365036010742
32 1.02298426628113
33 1.02298426628113
34 1.02298426628113
35 0.903463363647461
36 0.903463363647461
37 0.903463363647461
38 0.903463363647461
39 0.903463363647461
40 0.903463363647461
41 0.903463363647461
42 0.880580544471741
43 0.880580544471741
44 0.880580544471741
45 0.880580544471741
46 0.880580544471741
47 0.880580544471741
48 0.880580544471741
49 0.880580544471741
50 0.880580544471741
51 0.880580544471741
52 0.861925899982452
53 0.861925899982452
54 0.861925899982452
55 0.861925899982452
56 0.861925899982452
57 0.8408203125
58 0.8408203125
59 0.816231906414032
60 0.781783282756805
61 0.781783282756805
62 0.781783282756805
63 0.781783282756805
64 0.781783282756805
65 0.781783282756805
66 0.781783282756805
67 0.781783282756805
68 0.781783282756805
69 0.781783282756805
70 0.749705135822296
71 0.721121966838837
72 0.721121966838837
73 0.721121966838837
74 0.721121966838837
75 0.720400989055634
76 0.720400989055634
77 0.720400989055634
78 0.693763375282288
79 0.67266309261322
80 0.67266309261322
81 0.67266309261322
82 0.67266309261322
83 0.635725378990173
84 0.635725378990173
85 0.60212516784668
86 0.60212516784668
87 0.60212516784668
88 0.60212516784668
89 0.60212516784668
90 0.60212516784668
91 0.60212516784668
92 0.60212516784668
93 0.60212516784668
94 0.60212516784668
95 0.581223666667938
96 0.581223666667938
97 0.581223666667938
98 0.581223666667938
99 0.581223666667938
100 0.581223666667938
101 0.581223666667938
102 0.581223666667938
103 0.553499221801758
104 0.553499221801758
105 0.550786197185516
106 0.550786197185516
107 0.550786197185516
108 0.550786197185516
109 0.550786197185516
110 0.524573147296906
111 0.524573147296906
};
\addplot [semithick, color1]
table {%
10 4.86998586654663
11 4.36159153837768
12 3.68951247415254
13 3.20256843750939
14 2.85556642887532
15 2.31123205699331
16 1.85896511461993
17 1.39272719371147
18 1.24299840693882
19 1.21787955214983
20 1.19529495890489
21 0.985852668409371
22 0.855720234719631
23 0.802034594441634
24 0.802034594441634
25 0.802034594441634
26 0.704897510345302
27 0.670550577289034
28 0.659171513274475
29 0.609647421513207
30 0.600460464448715
31 0.584824843844146
32 0.580822022440844
33 0.496885950590647
34 0.47176093038707
35 0.429246902953493
36 0.421682911056324
37 0.309799754766331
38 0.298638885188148
39 0.0820379185215295
40 0.0761541393137158
41 0.0761541393137158
42 0.0761541393137158
43 0.0761541393137158
44 0.0509943245898135
45 0.0509943245898135
46 0.0504717004648668
47 0.0504717004648668
48 0.0504717004648668
49 0.0504717004648668
50 0.0504717004648668
51 0.0504717004648668
52 0.0504717004648668
53 0.0504717004648668
54 0.0504717004648668
55 0.0420356773070047
56 0.0386879515020027
57 0.0386879515020027
58 0.0358205128758772
59 0.0358205128758772
60 0.0325223218020167
61 0.0325223218020167
62 0.0325223218020167
63 0.0325223218020167
64 0.0325223218020167
65 0.0325223218020167
66 0.0325223218020167
67 0.0325223218020167
68 0.0325223218020167
69 0.0325223218020167
70 0.0325223218020167
71 0.0325223218020167
72 0.0325223218020167
73 0.0325223218020167
74 0.0325223218020167
75 0.0325223218020167
76 0.0325223218020167
77 0.0325223218020167
78 0.0278840609612306
79 0.0278840609612306
80 0.0278840609612306
81 0.0278840609612306
82 0.0278840609612306
83 0.0278840609612306
84 0.0278840609612306
85 0.0278840609612306
86 0.0278840609612306
87 0.0278840609612306
88 0.0278840609612306
89 0.0278840609612306
90 0.0278840609612306
91 0.0278840609612306
92 0.0278840609612306
93 0.0278840609612306
94 0.0278840609612306
95 0.0255611951611463
96 0.0255611951611463
97 0.0255611951611463
98 0.0252904006490137
99 0.0252904006490137
100 0.0252904006490137
101 0.0252904006490137
102 0.0252904006490137
103 0.0252904006490137
104 0.0252904006490137
105 0.0252904006490137
106 0.0252904006490137
107 0.0252904006490137
108 0.0252904006490137
109 0.0252904006490137
110 0.0252904006490137
111 0.0252904006490137
};
\end{axis}

\end{tikzpicture}

%% file: figures/bop_3d_hartmann3d_weak.tex
\begin{tikzpicture}

\definecolor{color0}{rgb}{0,0,1}
\definecolor{color1}{rgb}{1,0.549019607843137,0}
\definecolor{color2}{rgb}{1,0.647058823529412,0}
\definecolor{color3}{rgb}{0.564705882352941,0.933333333333333,0.564705882352941}

\begin{axis}[axis on top,
enlarge x limits=false,
enlarge y limits=false,
height=\figureheight,
scale only axis,
tick align=outside,
tick pos=left,
tick pos=left,
width=\figurewidth,
xlabel={Iteration},
xmin=10, xmax=60,
xtick style={color=black},
xtick={-10,0,10,25,50,75,100},
xticklabels={\ensuremath{-}10,0,10,25,50,75,90},
ymin=-0.05, ymax=1.3,
ytick style={color=black},
ytick={0.   , 1.3},
]
\node[anchor=north east] at (rel axis cs:1,1) {Hartmann 3D (weak)};
\path [draw=color1, fill=color1, opacity=0.3]
(axis cs:10,1.10511935245571)
--(axis cs:10,0.709856007041358)
--(axis cs:11,0.543981649440431)
--(axis cs:12,0.524567429844632)
--(axis cs:13,0.365760927965146)
--(axis cs:14,0.27936565912546)
--(axis cs:15,0.24316025651909)
--(axis cs:16,0.227321726038996)
--(axis cs:17,0.191558000505132)
--(axis cs:18,0.19002139881381)
--(axis cs:19,0.17395512706942)
--(axis cs:20,0.145707525122188)
--(axis cs:21,0.102037217314454)
--(axis cs:22,0.0910417873022544)
--(axis cs:23,0.0787488025809452)
--(axis cs:24,0.0771803163187426)
--(axis cs:25,0.0718403518778849)
--(axis cs:26,0.0688543880378161)
--(axis cs:27,0.0480798864508352)
--(axis cs:28,0.0371246953817608)
--(axis cs:29,0.0371246953817608)
--(axis cs:30,0.0347730626331979)
--(axis cs:31,0.0321891203573622)
--(axis cs:32,0.0320588911938692)
--(axis cs:33,0.0320588911938692)
--(axis cs:34,0.0320588911938692)
--(axis cs:35,0.0299091130520541)
--(axis cs:36,0.0273671589397508)
--(axis cs:37,0.0269930254720982)
--(axis cs:38,0.0250361207156598)
--(axis cs:39,0.0210120700394524)
--(axis cs:40,0.0210120700394524)
--(axis cs:41,0.0176427841242067)
--(axis cs:42,0.0176427841242067)
--(axis cs:43,0.0176427841242067)
--(axis cs:44,0.0176427841242067)
--(axis cs:45,0.0176427841242067)
--(axis cs:46,0.0165803805545353)
--(axis cs:47,0.0160035560152241)
--(axis cs:48,0.0160035560152241)
--(axis cs:49,0.0160035560152241)
--(axis cs:50,0.0153352511680341)
--(axis cs:51,0.0153352511680341)
--(axis cs:52,0.0153352511680341)
--(axis cs:53,0.0153352511680341)
--(axis cs:54,0.0153352511680341)
--(axis cs:55,0.0153352511680341)
--(axis cs:56,0.0153352511680341)
--(axis cs:57,0.0153352511680341)
--(axis cs:58,0.0153352511680341)
--(axis cs:59,0.0153352511680341)
--(axis cs:60,0.0153352511680341)
--(axis cs:61,0.0153352511680341)
--(axis cs:62,0.0153352511680341)
--(axis cs:63,0.0153352511680341)
--(axis cs:64,0.0152442623672333)
--(axis cs:65,0.0152442623672333)
--(axis cs:66,0.0152442623672333)
--(axis cs:67,0.0148075461378697)
--(axis cs:68,0.0148075461378697)
--(axis cs:69,0.0148075461378697)
--(axis cs:70,0.0148075461378697)
--(axis cs:71,0.0148075461378697)
--(axis cs:72,0.0148075461378697)
--(axis cs:73,0.0148075461378697)
--(axis cs:74,0.0148075461378697)
--(axis cs:75,0.0148075461378697)
--(axis cs:76,0.0148075461378697)
--(axis cs:77,0.014469595344809)
--(axis cs:78,0.014469595344809)
--(axis cs:79,0.014469595344809)
--(axis cs:80,0.014469595344809)
--(axis cs:81,0.014469595344809)
--(axis cs:82,0.014469595344809)
--(axis cs:83,0.014469595344809)
--(axis cs:84,0.014469595344809)
--(axis cs:85,0.014469595344809)
--(axis cs:86,0.014469595344809)
--(axis cs:87,0.014469595344809)
--(axis cs:88,0.014469595344809)
--(axis cs:89,0.014469595344809)
--(axis cs:90,0.014469595344809)
--(axis cs:91,0.014469595344809)
--(axis cs:92,0.014469595344809)
--(axis cs:93,0.014469595344809)
--(axis cs:94,0.014469595344809)
--(axis cs:95,0.014469595344809)
--(axis cs:96,0.014469595344809)
--(axis cs:97,0.014469595344809)
--(axis cs:98,0.014469595344809)
--(axis cs:99,0.014469595344809)
--(axis cs:100,0.014469595344809)
--(axis cs:101,0.014469595344809)
--(axis cs:102,0.014469595344809)
--(axis cs:103,0.014469595344809)
--(axis cs:104,0.014469595344809)
--(axis cs:105,0.014469595344809)
--(axis cs:106,0.014469595344809)
--(axis cs:107,0.014469595344809)
--(axis cs:108,0.014469595344809)
--(axis cs:109,0.014469595344809)
--(axis cs:110,0.014469595344809)
--(axis cs:111,0.0133057211563445)
--(axis cs:111,0.0178702743281856)
--(axis cs:111,0.0178702743281856)
--(axis cs:110,0.0182587931371595)
--(axis cs:109,0.0182587931371595)
--(axis cs:108,0.0182587931371595)
--(axis cs:107,0.0182587931371595)
--(axis cs:106,0.0182587931371595)
--(axis cs:105,0.0182587931371595)
--(axis cs:104,0.0182587931371595)
--(axis cs:103,0.0182587931371595)
--(axis cs:102,0.0182587931371595)
--(axis cs:101,0.0182587931371595)
--(axis cs:100,0.0182587931371595)
--(axis cs:99,0.0182587931371595)
--(axis cs:98,0.0182587931371595)
--(axis cs:97,0.0182587931371595)
--(axis cs:96,0.0182587931371595)
--(axis cs:95,0.0182587931371595)
--(axis cs:94,0.0182587931371595)
--(axis cs:93,0.0182587931371595)
--(axis cs:92,0.0182587931371595)
--(axis cs:91,0.0182587931371595)
--(axis cs:90,0.0182587931371595)
--(axis cs:89,0.0182587931371595)
--(axis cs:88,0.0182587931371595)
--(axis cs:87,0.0182587931371595)
--(axis cs:86,0.0182587931371595)
--(axis cs:85,0.0182587931371595)
--(axis cs:84,0.0182587931371595)
--(axis cs:83,0.0182587931371595)
--(axis cs:82,0.0182587931371595)
--(axis cs:81,0.0182587931371595)
--(axis cs:80,0.0182587931371595)
--(axis cs:79,0.0182587931371595)
--(axis cs:78,0.0182587931371595)
--(axis cs:77,0.0182587931371595)
--(axis cs:76,0.0184989275661793)
--(axis cs:75,0.0184989275661793)
--(axis cs:74,0.0184989275661793)
--(axis cs:73,0.0184989275661793)
--(axis cs:72,0.0184989275661793)
--(axis cs:71,0.0184989275661793)
--(axis cs:70,0.0184989275661793)
--(axis cs:69,0.0184989275661793)
--(axis cs:68,0.0184989275661793)
--(axis cs:67,0.0184989275661793)
--(axis cs:66,0.019229532948628)
--(axis cs:65,0.019229532948628)
--(axis cs:64,0.019229532948628)
--(axis cs:63,0.0193042441526136)
--(axis cs:62,0.0193042441526136)
--(axis cs:61,0.0193042441526136)
--(axis cs:60,0.0193042441526136)
--(axis cs:59,0.0193042441526136)
--(axis cs:58,0.0193042441526136)
--(axis cs:57,0.0193042441526136)
--(axis cs:56,0.0193042441526136)
--(axis cs:55,0.0193042441526136)
--(axis cs:54,0.0193042441526136)
--(axis cs:53,0.0193042441526136)
--(axis cs:52,0.0193042441526136)
--(axis cs:51,0.0193042441526136)
--(axis cs:50,0.0193042441526136)
--(axis cs:49,0.0226437528671147)
--(axis cs:48,0.0226437528671147)
--(axis cs:47,0.0226437528671147)
--(axis cs:46,0.0230669853903284)
--(axis cs:45,0.0250118040666411)
--(axis cs:44,0.0250118040666411)
--(axis cs:43,0.0250118040666411)
--(axis cs:42,0.0250118040666411)
--(axis cs:41,0.0250118040666411)
--(axis cs:40,0.034328875305683)
--(axis cs:39,0.034328875305683)
--(axis cs:38,0.0382489872483113)
--(axis cs:37,0.0396326455865423)
--(axis cs:36,0.0403428528538641)
--(axis cs:35,0.0439975379597806)
--(axis cs:34,0.0468431018344423)
--(axis cs:33,0.0468431018344423)
--(axis cs:32,0.0468431018344423)
--(axis cs:31,0.047354411731313)
--(axis cs:30,0.0490215780873851)
--(axis cs:29,0.0507631720495766)
--(axis cs:28,0.0507631720495766)
--(axis cs:27,0.0727917102034381)
--(axis cs:26,0.0987416777268406)
--(axis cs:25,0.106064826092734)
--(axis cs:24,0.110232756651883)
--(axis cs:23,0.116735176934616)
--(axis cs:22,0.150018771900021)
--(axis cs:21,0.172173389166157)
--(axis cs:20,0.287514476344651)
--(axis cs:19,0.320521042805438)
--(axis cs:18,0.331362188613408)
--(axis cs:17,0.332455220675824)
--(axis cs:16,0.370054807344314)
--(axis cs:15,0.389932194764202)
--(axis cs:14,0.418175188662491)
--(axis cs:13,0.620026398570146)
--(axis cs:12,0.758163491148858)
--(axis cs:11,0.890848326419269)
--(axis cs:10,1.10511935245571)
--cycle;

\path [draw=blue, fill=blue, opacity=0.3]
(axis cs:10,1.10511946678162)
--(axis cs:10,0.70985621213913)
--(axis cs:11,0.497344940900803)
--(axis cs:12,0.477982103824615)
--(axis cs:13,0.349312037229538)
--(axis cs:14,0.288065791130066)
--(axis cs:15,0.231376111507416)
--(axis cs:16,0.220121800899506)
--(axis cs:17,0.190316170454025)
--(axis cs:18,0.180567741394043)
--(axis cs:19,0.17525714635849)
--(axis cs:20,0.150213181972504)
--(axis cs:21,0.149086952209473)
--(axis cs:22,0.149086952209473)
--(axis cs:23,0.138051465153694)
--(axis cs:24,0.127912685275078)
--(axis cs:25,0.113370083272457)
--(axis cs:26,0.0943606644868851)
--(axis cs:27,0.0923316031694412)
--(axis cs:28,0.0923316031694412)
--(axis cs:29,0.0821646451950073)
--(axis cs:30,0.0758992061018944)
--(axis cs:31,0.0667785778641701)
--(axis cs:32,0.0667785778641701)
--(axis cs:33,0.0599892511963844)
--(axis cs:34,0.0526719316840172)
--(axis cs:35,0.0426463484764099)
--(axis cs:36,0.0390123948454857)
--(axis cs:37,0.034311942756176)
--(axis cs:38,0.0298611838370562)
--(axis cs:39,0.02405602671206)
--(axis cs:40,0.02405602671206)
--(axis cs:41,0.0226277858018875)
--(axis cs:42,0.0224328190088272)
--(axis cs:43,0.0200613643974066)
--(axis cs:44,0.0200613643974066)
--(axis cs:45,0.0197501499205828)
--(axis cs:46,0.0197501499205828)
--(axis cs:47,0.0189632289111614)
--(axis cs:48,0.0189632289111614)
--(axis cs:49,0.0156927239149809)
--(axis cs:50,0.0156927239149809)
--(axis cs:51,0.0156927239149809)
--(axis cs:52,0.0156927239149809)
--(axis cs:53,0.0156927239149809)
--(axis cs:54,0.0156927239149809)
--(axis cs:55,0.0142632992938161)
--(axis cs:56,0.0142632992938161)
--(axis cs:57,0.0142632992938161)
--(axis cs:58,0.0142632992938161)
--(axis cs:59,0.0142632992938161)
--(axis cs:60,0.0142632992938161)
--(axis cs:61,0.0142632992938161)
--(axis cs:62,0.0135753871873021)
--(axis cs:63,0.0120918229222298)
--(axis cs:64,0.0120918229222298)
--(axis cs:65,0.0111049124971032)
--(axis cs:66,0.0111049124971032)
--(axis cs:67,0.0111049124971032)
--(axis cs:68,0.00878499262034893)
--(axis cs:69,0.00878499262034893)
--(axis cs:70,0.00878499262034893)
--(axis cs:71,0.00878499262034893)
--(axis cs:72,0.00822488032281399)
--(axis cs:73,0.00822488032281399)
--(axis cs:74,0.00822488032281399)
--(axis cs:75,0.00822488032281399)
--(axis cs:76,0.00822488032281399)
--(axis cs:77,0.00822488032281399)
--(axis cs:78,0.00822488032281399)
--(axis cs:79,0.00822488032281399)
--(axis cs:80,0.00822488032281399)
--(axis cs:81,0.00822488032281399)
--(axis cs:82,0.00822488032281399)
--(axis cs:83,0.00822488032281399)
--(axis cs:84,0.00822488032281399)
--(axis cs:85,0.00822488032281399)
--(axis cs:86,0.00822488032281399)
--(axis cs:87,0.00822488032281399)
--(axis cs:88,0.00807350222021341)
--(axis cs:89,0.00807350222021341)
--(axis cs:90,0.00658612884581089)
--(axis cs:91,0.00658612884581089)
--(axis cs:92,0.00633103027939796)
--(axis cs:93,0.00627751555293798)
--(axis cs:94,0.00627751555293798)
--(axis cs:95,0.00627751555293798)
--(axis cs:96,0.00627751555293798)
--(axis cs:97,0.00627751555293798)
--(axis cs:98,0.00627751555293798)
--(axis cs:99,0.00627751555293798)
--(axis cs:100,0.00627751555293798)
--(axis cs:101,0.00627751555293798)
--(axis cs:102,0.00627751555293798)
--(axis cs:103,0.00627751555293798)
--(axis cs:104,0.00627751555293798)
--(axis cs:105,0.00627751555293798)
--(axis cs:106,0.00627751555293798)
--(axis cs:107,0.00627751555293798)
--(axis cs:108,0.00627751555293798)
--(axis cs:109,0.00627751555293798)
--(axis cs:110,0.00627751555293798)
--(axis cs:111,0.00627751555293798)
--(axis cs:111,0.0104571329429746)
--(axis cs:111,0.0104571329429746)
--(axis cs:110,0.0104571329429746)
--(axis cs:109,0.0104571329429746)
--(axis cs:108,0.0104571329429746)
--(axis cs:107,0.0104571329429746)
--(axis cs:106,0.0104571329429746)
--(axis cs:105,0.0104571329429746)
--(axis cs:104,0.0104571329429746)
--(axis cs:103,0.0104571329429746)
--(axis cs:102,0.0104571329429746)
--(axis cs:101,0.0104571329429746)
--(axis cs:100,0.0104571329429746)
--(axis cs:99,0.0104571329429746)
--(axis cs:98,0.0104571329429746)
--(axis cs:97,0.0104571329429746)
--(axis cs:96,0.0104571329429746)
--(axis cs:95,0.0104571329429746)
--(axis cs:94,0.0104571329429746)
--(axis cs:93,0.0104571329429746)
--(axis cs:92,0.0105627384036779)
--(axis cs:91,0.0112033300101757)
--(axis cs:90,0.0112033300101757)
--(axis cs:89,0.0134206963703036)
--(axis cs:88,0.0134206963703036)
--(axis cs:87,0.0139809977263212)
--(axis cs:86,0.0139809977263212)
--(axis cs:85,0.0139809977263212)
--(axis cs:84,0.0139809977263212)
--(axis cs:83,0.0139809977263212)
--(axis cs:82,0.0139809977263212)
--(axis cs:81,0.0139809977263212)
--(axis cs:80,0.0139809977263212)
--(axis cs:79,0.0139809977263212)
--(axis cs:78,0.0139809977263212)
--(axis cs:77,0.0139809977263212)
--(axis cs:76,0.0139809977263212)
--(axis cs:75,0.0139809977263212)
--(axis cs:74,0.0139809977263212)
--(axis cs:73,0.0139809977263212)
--(axis cs:72,0.0139809977263212)
--(axis cs:71,0.0148468185216188)
--(axis cs:70,0.0148468185216188)
--(axis cs:69,0.0148468185216188)
--(axis cs:68,0.0148468185216188)
--(axis cs:67,0.0169444121420383)
--(axis cs:66,0.0169444121420383)
--(axis cs:65,0.0169444121420383)
--(axis cs:64,0.0199874937534332)
--(axis cs:63,0.0199874937534332)
--(axis cs:62,0.0212822202593088)
--(axis cs:61,0.0216207448393106)
--(axis cs:60,0.0216207448393106)
--(axis cs:59,0.0216207448393106)
--(axis cs:58,0.0216207448393106)
--(axis cs:57,0.0216207448393106)
--(axis cs:56,0.0216207448393106)
--(axis cs:55,0.0216207448393106)
--(axis cs:54,0.0231448505073786)
--(axis cs:53,0.0231448505073786)
--(axis cs:52,0.0231448505073786)
--(axis cs:51,0.0231448505073786)
--(axis cs:50,0.0231448505073786)
--(axis cs:49,0.0231448505073786)
--(axis cs:48,0.0282886624336243)
--(axis cs:47,0.0282886624336243)
--(axis cs:46,0.029424587264657)
--(axis cs:45,0.029424587264657)
--(axis cs:44,0.0309398043900728)
--(axis cs:43,0.0309398043900728)
--(axis cs:42,0.043407566845417)
--(axis cs:41,0.0435210168361664)
--(axis cs:40,0.0450709164142609)
--(axis cs:39,0.0450709164142609)
--(axis cs:38,0.0577479675412178)
--(axis cs:37,0.0606918036937714)
--(axis cs:36,0.0669608041644096)
--(axis cs:35,0.0701922550797462)
--(axis cs:34,0.0952416732907295)
--(axis cs:33,0.100082360208035)
--(axis cs:32,0.104645244777203)
--(axis cs:31,0.104645244777203)
--(axis cs:30,0.116365678608418)
--(axis cs:29,0.129239201545715)
--(axis cs:28,0.13752393424511)
--(axis cs:27,0.13752393424511)
--(axis cs:26,0.140312746167183)
--(axis cs:25,0.16696685552597)
--(axis cs:24,0.175758019089699)
--(axis cs:23,0.192538872361183)
--(axis cs:22,0.227220714092255)
--(axis cs:21,0.227220714092255)
--(axis cs:20,0.229226559400558)
--(axis cs:19,0.249638378620148)
--(axis cs:18,0.257541328668594)
--(axis cs:17,0.285073459148407)
--(axis cs:16,0.332626938819885)
--(axis cs:15,0.340072691440582)
--(axis cs:14,0.655151188373566)
--(axis cs:13,0.711981058120728)
--(axis cs:12,0.826257407665253)
--(axis cs:11,0.839098572731018)
--(axis cs:10,1.10511946678162)
--cycle;

\path [draw=color1, fill=color1, opacity=0.3]
(axis cs:10,1.10511935245571)
--(axis cs:10,0.709856007041358)
--(axis cs:11,0.709856007041358)
--(axis cs:12,0.609489774785933)
--(axis cs:13,0.538594004198679)
--(axis cs:14,0.418142191578086)
--(axis cs:15,0.415429464008913)
--(axis cs:16,0.282715531090004)
--(axis cs:17,0.199287957315256)
--(axis cs:18,0.130587828057952)
--(axis cs:19,0.0920949212944619)
--(axis cs:20,0.065428509220475)
--(axis cs:21,0.0587049833407587)
--(axis cs:22,0.0461496736432424)
--(axis cs:23,0.0401133483953571)
--(axis cs:24,0.0213626227670111)
--(axis cs:25,0.0199746179716721)
--(axis cs:26,0.0181523855500456)
--(axis cs:27,0.013610772212464)
--(axis cs:28,0.0119063836820695)
--(axis cs:29,0.00990690583137368)
--(axis cs:30,0.00903396745976523)
--(axis cs:31,0.00783940267219162)
--(axis cs:32,0.0067357163081806)
--(axis cs:33,0.00520700465023994)
--(axis cs:34,0.00488022809513833)
--(axis cs:35,0.0042995918456326)
--(axis cs:36,0.00364072686737585)
--(axis cs:37,0.00351446204483324)
--(axis cs:38,0.00266911116410286)
--(axis cs:39,0.00231089430210125)
--(axis cs:40,0.00225758558998345)
--(axis cs:41,0.00213750834356387)
--(axis cs:42,0.00181402644005912)
--(axis cs:43,0.00149993094147047)
--(axis cs:44,0.00149993094147047)
--(axis cs:45,0.00149993094147047)
--(axis cs:46,0.00149993094147047)
--(axis cs:47,0.00120099500852978)
--(axis cs:48,0.00120099500852978)
--(axis cs:49,0.00120099500852978)
--(axis cs:50,0.00120099500852978)
--(axis cs:51,0.000908634325522101)
--(axis cs:52,0.0008175254333121)
--(axis cs:53,0.0008175254333121)
--(axis cs:54,0.0008175254333121)
--(axis cs:55,0.0008175254333121)
--(axis cs:56,0.00075621516926468)
--(axis cs:57,0.00075621516926468)
--(axis cs:58,0.000693738986109948)
--(axis cs:59,0.000617044973114808)
--(axis cs:60,0.000549337082282405)
--(axis cs:61,0.000549337082282405)
--(axis cs:62,0.000549337082282405)
--(axis cs:63,0.000543320689319686)
--(axis cs:64,0.000543320689319686)
--(axis cs:65,0.000543320689319686)
--(axis cs:66,0.000543320689319686)
--(axis cs:67,0.000543320689319686)
--(axis cs:68,0.000543320689319686)
--(axis cs:69,0.000543320689319686)
--(axis cs:70,0.000543320689319686)
--(axis cs:71,0.000543320689319686)
--(axis cs:72,0.000543320689319686)
--(axis cs:73,0.000543320689319686)
--(axis cs:74,0.000543320689319686)
--(axis cs:75,0.000543320689319686)
--(axis cs:76,0.000543320689319686)
--(axis cs:77,0.000543320689319686)
--(axis cs:78,0.000543320689319686)
--(axis cs:79,0.000482348794858188)
--(axis cs:80,0.000482348794858188)
--(axis cs:81,0.000482348794858188)
--(axis cs:82,0.000482348794858188)
--(axis cs:83,0.000482348794858188)
--(axis cs:84,0.000482348794858188)
--(axis cs:85,0.000482348794858188)
--(axis cs:86,0.000394280387028209)
--(axis cs:87,0.000394280387028209)
--(axis cs:88,0.000394280387028209)
--(axis cs:89,0.000394280387028209)
--(axis cs:90,0.000394280387028209)
--(axis cs:91,0.000394280387028209)
--(axis cs:92,0.000394280387028209)
--(axis cs:93,0.000394280387028209)
--(axis cs:94,0.000394280387028209)
--(axis cs:95,0.000394280387028209)
--(axis cs:96,0.000394280387028209)
--(axis cs:97,0.00036605964149399)
--(axis cs:98,0.00036605964149399)
--(axis cs:99,0.00036605964149399)
--(axis cs:100,0.00036605964149399)
--(axis cs:101,0.00036605964149399)
--(axis cs:102,0.00036605964149399)
--(axis cs:103,0.00036605964149399)
--(axis cs:104,0.00036605964149399)
--(axis cs:105,0.00036605964149399)
--(axis cs:106,0.000365349906245193)
--(axis cs:107,0.000365349906245193)
--(axis cs:108,0.000365349906245193)
--(axis cs:109,0.000365349906245193)
--(axis cs:110,0.000365349906245193)
--(axis cs:111,0.000365349906245193)
--(axis cs:111,0.000654936795387396)
--(axis cs:111,0.000654936795387396)
--(axis cs:110,0.000654936795387396)
--(axis cs:109,0.000654936795387396)
--(axis cs:108,0.000654936795387396)
--(axis cs:107,0.000654936795387396)
--(axis cs:106,0.000654936795387396)
--(axis cs:105,0.000659645618743607)
--(axis cs:104,0.000659645618743607)
--(axis cs:103,0.000659645618743607)
--(axis cs:102,0.000659645618743607)
--(axis cs:101,0.000659645618743607)
--(axis cs:100,0.000659645618743607)
--(axis cs:99,0.000659645618743607)
--(axis cs:98,0.000659645618743607)
--(axis cs:97,0.000659645618743607)
--(axis cs:96,0.00067764045232869)
--(axis cs:95,0.00067764045232869)
--(axis cs:94,0.00067764045232869)
--(axis cs:93,0.00067764045232869)
--(axis cs:92,0.00067764045232869)
--(axis cs:91,0.00067764045232869)
--(axis cs:90,0.00067764045232869)
--(axis cs:89,0.00067764045232869)
--(axis cs:88,0.00067764045232869)
--(axis cs:87,0.00067764045232869)
--(axis cs:86,0.00067764045232869)
--(axis cs:85,0.000889629813114036)
--(axis cs:84,0.000889629813114036)
--(axis cs:83,0.000889629813114036)
--(axis cs:82,0.000889629813114036)
--(axis cs:81,0.000889629813114036)
--(axis cs:80,0.000889629813114036)
--(axis cs:79,0.000889629813114036)
--(axis cs:78,0.00101387793688481)
--(axis cs:77,0.00101387793688481)
--(axis cs:76,0.00101387793688481)
--(axis cs:75,0.00101387793688481)
--(axis cs:74,0.00101387793688481)
--(axis cs:73,0.00101387793688481)
--(axis cs:72,0.00101387793688481)
--(axis cs:71,0.00101387793688481)
--(axis cs:70,0.00101387793688481)
--(axis cs:69,0.00101387793688481)
--(axis cs:68,0.00101387793688481)
--(axis cs:67,0.00101387793688481)
--(axis cs:66,0.00101387793688481)
--(axis cs:65,0.00101387793688481)
--(axis cs:64,0.00101387793688481)
--(axis cs:63,0.00101387793688481)
--(axis cs:62,0.00101726478009065)
--(axis cs:61,0.00101726478009065)
--(axis cs:60,0.00101726478009065)
--(axis cs:59,0.0010761247450277)
--(axis cs:58,0.00118495595419993)
--(axis cs:57,0.00121742094456644)
--(axis cs:56,0.00121742094456644)
--(axis cs:55,0.00125005514859756)
--(axis cs:54,0.00125005514859756)
--(axis cs:53,0.00125005514859756)
--(axis cs:52,0.00125005514859756)
--(axis cs:51,0.00134206351813329)
--(axis cs:50,0.00182696654242248)
--(axis cs:49,0.00182696654242248)
--(axis cs:48,0.00182696654242248)
--(axis cs:47,0.00182696654242248)
--(axis cs:46,0.00230067693771539)
--(axis cs:45,0.00230067693771539)
--(axis cs:44,0.00230067693771539)
--(axis cs:43,0.00230067693771539)
--(axis cs:42,0.00287680465915166)
--(axis cs:41,0.00321992734090975)
--(axis cs:40,0.00338092969019489)
--(axis cs:39,0.00359350674461569)
--(axis cs:38,0.00487101838903079)
--(axis cs:37,0.00581492825206135)
--(axis cs:36,0.00606038246037003)
--(axis cs:35,0.00687435613209754)
--(axis cs:34,0.00742311173207053)
--(axis cs:33,0.0100508438979898)
--(axis cs:32,0.0129154689187239)
--(axis cs:31,0.0139448502976961)
--(axis cs:30,0.0167933295775949)
--(axis cs:29,0.0202179033155464)
--(axis cs:28,0.0302452627007889)
--(axis cs:27,0.0317830267673551)
--(axis cs:26,0.0369203860318253)
--(axis cs:25,0.0382040155820297)
--(axis cs:24,0.0393062436063888)
--(axis cs:23,0.0812237545224906)
--(axis cs:22,0.0933433582461063)
--(axis cs:21,0.106794791930307)
--(axis cs:20,0.13972882155646)
--(axis cs:19,0.194499431519615)
--(axis cs:18,0.229689805309955)
--(axis cs:17,0.360808513501336)
--(axis cs:16,0.44531404647844)
--(axis cs:15,0.601711367366481)
--(axis cs:14,0.609488258883826)
--(axis cs:13,0.860998514851823)
--(axis cs:12,0.925990622418656)
--(axis cs:11,1.10511935245571)
--(axis cs:10,1.10511935245571)
--cycle;

\path [draw=blue, fill=blue, opacity=0.3]
(axis cs:10,1.10511946678162)
--(axis cs:10,0.70985609292984)
--(axis cs:11,0.546315312385559)
--(axis cs:12,0.381701707839966)
--(axis cs:13,0.328166514635086)
--(axis cs:14,0.174432992935181)
--(axis cs:15,0.174432992935181)
--(axis cs:16,0.150034725666046)
--(axis cs:17,0.133072018623352)
--(axis cs:18,0.122517012059689)
--(axis cs:19,0.109856978058815)
--(axis cs:20,0.0942640528082848)
--(axis cs:21,0.0910769701004028)
--(axis cs:22,0.0770667940378189)
--(axis cs:23,0.0621199160814285)
--(axis cs:24,0.054646797478199)
--(axis cs:25,0.054646797478199)
--(axis cs:26,0.054646797478199)
--(axis cs:27,0.0540038757026196)
--(axis cs:28,0.0524984300136566)
--(axis cs:29,0.0524984300136566)
--(axis cs:30,0.0516223162412643)
--(axis cs:31,0.039860412478447)
--(axis cs:32,0.039860412478447)
--(axis cs:33,0.039860412478447)
--(axis cs:34,0.0381783731281757)
--(axis cs:35,0.0356736145913601)
--(axis cs:36,0.0355454310774803)
--(axis cs:37,0.0355454310774803)
--(axis cs:38,0.0355454310774803)
--(axis cs:39,0.0296934973448515)
--(axis cs:40,0.0285938773304224)
--(axis cs:41,0.0284059848636389)
--(axis cs:42,0.0274775847792625)
--(axis cs:43,0.0230553187429905)
--(axis cs:44,0.0194671917706728)
--(axis cs:45,0.0191413443535566)
--(axis cs:46,0.0191413443535566)
--(axis cs:47,0.0185127947479486)
--(axis cs:48,0.0185127947479486)
--(axis cs:49,0.0185127947479486)
--(axis cs:50,0.0185127947479486)
--(axis cs:51,0.0185127947479486)
--(axis cs:52,0.0171948280185461)
--(axis cs:53,0.0171948280185461)
--(axis cs:54,0.0171948280185461)
--(axis cs:55,0.0171948280185461)
--(axis cs:56,0.0171948280185461)
--(axis cs:57,0.0171948280185461)
--(axis cs:58,0.016653249040246)
--(axis cs:59,0.016653249040246)
--(axis cs:60,0.016653249040246)
--(axis cs:61,0.016653249040246)
--(axis cs:62,0.0130551196634769)
--(axis cs:63,0.0130551196634769)
--(axis cs:64,0.0130551196634769)
--(axis cs:65,0.0130551196634769)
--(axis cs:66,0.0127553949132562)
--(axis cs:67,0.0121950963512063)
--(axis cs:68,0.0121950963512063)
--(axis cs:69,0.0121950963512063)
--(axis cs:70,0.0121950963512063)
--(axis cs:71,0.0121950963512063)
--(axis cs:72,0.0121950963512063)
--(axis cs:73,0.0121950963512063)
--(axis cs:74,0.0121950963512063)
--(axis cs:75,0.0121950963512063)
--(axis cs:76,0.0121950963512063)
--(axis cs:77,0.0121950963512063)
--(axis cs:78,0.0121950963512063)
--(axis cs:79,0.0121950963512063)
--(axis cs:80,0.0121950963512063)
--(axis cs:81,0.0121950963512063)
--(axis cs:82,0.0121950963512063)
--(axis cs:83,0.0121950963512063)
--(axis cs:84,0.0121950963512063)
--(axis cs:85,0.0121950963512063)
--(axis cs:86,0.01133350469172)
--(axis cs:87,0.0102448910474777)
--(axis cs:88,0.0102448910474777)
--(axis cs:89,0.0102448910474777)
--(axis cs:90,0.0102448910474777)
--(axis cs:91,0.0101083964109421)
--(axis cs:92,0.0101083964109421)
--(axis cs:93,0.0101083964109421)
--(axis cs:94,0.0101083964109421)
--(axis cs:95,0.0101083964109421)
--(axis cs:96,0.0101083964109421)
--(axis cs:97,0.0101083964109421)
--(axis cs:98,0.0101083964109421)
--(axis cs:99,0.0101083964109421)
--(axis cs:100,0.00983458198606968)
--(axis cs:101,0.00980239361524582)
--(axis cs:102,0.00980239361524582)
--(axis cs:103,0.00980239361524582)
--(axis cs:104,0.00947150960564613)
--(axis cs:105,0.00947150960564613)
--(axis cs:106,0.00947150960564613)
--(axis cs:107,0.00947150960564613)
--(axis cs:108,0.00947150960564613)
--(axis cs:109,0.00947150960564613)
--(axis cs:110,0.00947150960564613)
--(axis cs:111,0.00947150960564613)
--(axis cs:111,0.0151692889630795)
--(axis cs:111,0.0151692889630795)
--(axis cs:110,0.0151692889630795)
--(axis cs:109,0.0151692889630795)
--(axis cs:108,0.0151692889630795)
--(axis cs:107,0.0151692889630795)
--(axis cs:106,0.0151692889630795)
--(axis cs:105,0.0151692889630795)
--(axis cs:104,0.0151692889630795)
--(axis cs:103,0.0160808991640806)
--(axis cs:102,0.0160808991640806)
--(axis cs:101,0.0160808991640806)
--(axis cs:100,0.0161872319877148)
--(axis cs:99,0.0163737088441849)
--(axis cs:98,0.0163737088441849)
--(axis cs:97,0.0163737088441849)
--(axis cs:96,0.0163737088441849)
--(axis cs:95,0.0163737088441849)
--(axis cs:94,0.0163737088441849)
--(axis cs:93,0.0163737088441849)
--(axis cs:92,0.0163737088441849)
--(axis cs:91,0.0163737088441849)
--(axis cs:90,0.0165747664868832)
--(axis cs:89,0.0165747664868832)
--(axis cs:88,0.0165747664868832)
--(axis cs:87,0.0165747664868832)
--(axis cs:86,0.0172542650252581)
--(axis cs:85,0.0192786827683449)
--(axis cs:84,0.0192786827683449)
--(axis cs:83,0.0192786827683449)
--(axis cs:82,0.0192786827683449)
--(axis cs:81,0.0192786827683449)
--(axis cs:80,0.0192786827683449)
--(axis cs:79,0.0192786827683449)
--(axis cs:78,0.0192786827683449)
--(axis cs:77,0.0192786827683449)
--(axis cs:76,0.0192786827683449)
--(axis cs:75,0.0192786827683449)
--(axis cs:74,0.0192786827683449)
--(axis cs:73,0.0192786827683449)
--(axis cs:72,0.0192786827683449)
--(axis cs:71,0.0192786827683449)
--(axis cs:70,0.0192786827683449)
--(axis cs:69,0.0192786827683449)
--(axis cs:68,0.0192786827683449)
--(axis cs:67,0.0192786827683449)
--(axis cs:66,0.0196153130382299)
--(axis cs:65,0.0210273899137974)
--(axis cs:64,0.0210273899137974)
--(axis cs:63,0.0210273899137974)
--(axis cs:62,0.0210273899137974)
--(axis cs:61,0.0256962794810534)
--(axis cs:60,0.0256962794810534)
--(axis cs:59,0.0256962794810534)
--(axis cs:58,0.0256962794810534)
--(axis cs:57,0.0266004223376513)
--(axis cs:56,0.0266004223376513)
--(axis cs:55,0.0266004223376513)
--(axis cs:54,0.0266004223376513)
--(axis cs:53,0.0266004223376513)
--(axis cs:52,0.0266004223376513)
--(axis cs:51,0.0275676045566797)
--(axis cs:50,0.0275676045566797)
--(axis cs:49,0.0275676045566797)
--(axis cs:48,0.0275676045566797)
--(axis cs:47,0.0275676045566797)
--(axis cs:46,0.0290510598570108)
--(axis cs:45,0.0290510598570108)
--(axis cs:44,0.0300621222704649)
--(axis cs:43,0.0373750180006027)
--(axis cs:42,0.0446767956018448)
--(axis cs:41,0.0455236062407494)
--(axis cs:40,0.0458059310913086)
--(axis cs:39,0.047833688557148)
--(axis cs:38,0.0534341931343079)
--(axis cs:37,0.0534341931343079)
--(axis cs:36,0.0534341931343079)
--(axis cs:35,0.0535696558654308)
--(axis cs:34,0.0568444989621639)
--(axis cs:33,0.0590996071696281)
--(axis cs:32,0.0590996071696281)
--(axis cs:31,0.0590996071696281)
--(axis cs:30,0.0856538563966751)
--(axis cs:29,0.0865871012210846)
--(axis cs:28,0.0865871012210846)
--(axis cs:27,0.0889284461736679)
--(axis cs:26,0.0894164368510246)
--(axis cs:25,0.0894164368510246)
--(axis cs:24,0.0894164368510246)
--(axis cs:23,0.0989199131727219)
--(axis cs:22,0.119883641600609)
--(axis cs:21,0.155952483415604)
--(axis cs:20,0.161822944879532)
--(axis cs:19,0.177270039916039)
--(axis cs:18,0.187479734420776)
--(axis cs:17,0.218013346195221)
--(axis cs:16,0.236350536346436)
--(axis cs:15,0.270654141902924)
--(axis cs:14,0.270654141902924)
--(axis cs:13,0.501963436603546)
--(axis cs:12,0.576623320579529)
--(axis cs:11,0.819665551185608)
--(axis cs:10,1.10511946678162)
--cycle;

\addplot [semithick, color1, dash dot]
table {%
10 0.907487679748535
11 0.71741498792985
12 0.641365460496745
13 0.492893663267646
14 0.348770423893975
15 0.316546225641646
16 0.298688266691655
17 0.262006610590478
18 0.260691793713609
19 0.247238084937429
20 0.216611000733419
21 0.137105303240306
22 0.120530279601137
23 0.0977419897577808
24 0.0937065364853129
25 0.0889525889853094
26 0.0837980328823284
27 0.0604357983271367
28 0.0439439337156687
29 0.0439439337156687
30 0.0418973203602915
31 0.0397717660443376
32 0.0394509965141558
33 0.0394509965141558
34 0.0394509965141558
35 0.0369533255059173
36 0.0338550058968075
37 0.0333128355293202
38 0.0316425539819856
39 0.0276704726725677
40 0.0276704726725677
41 0.0213272940954239
42 0.0213272940954239
43 0.0213272940954239
44 0.0213272940954239
45 0.0213272940954239
46 0.0198236829724319
47 0.0193236544411694
48 0.0193236544411694
49 0.0193236544411694
50 0.0173197476603238
51 0.0173197476603238
52 0.0173197476603238
53 0.0173197476603238
54 0.0173197476603238
55 0.0173197476603238
56 0.0173197476603238
57 0.0173197476603238
58 0.0173197476603238
59 0.0173197476603238
60 0.0173197476603238
61 0.0173197476603238
62 0.0173197476603238
63 0.0173197476603238
64 0.0172368976579306
65 0.0172368976579306
66 0.0172368976579306
67 0.0166532368520245
68 0.0166532368520245
69 0.0166532368520245
70 0.0166532368520245
71 0.0166532368520245
72 0.0166532368520245
73 0.0166532368520245
74 0.0166532368520245
75 0.0166532368520245
76 0.0166532368520245
77 0.0163641942409842
78 0.0163641942409842
79 0.0163641942409842
80 0.0163641942409842
81 0.0163641942409842
82 0.0163641942409842
83 0.0163641942409842
84 0.0163641942409842
85 0.0163641942409842
86 0.0163641942409842
87 0.0163641942409842
88 0.0163641942409842
89 0.0163641942409842
90 0.0163641942409842
91 0.0163641942409842
92 0.0163641942409842
93 0.0163641942409842
94 0.0163641942409842
95 0.0163641942409842
96 0.0163641942409842
97 0.0163641942409842
98 0.0163641942409842
99 0.0163641942409842
100 0.0163641942409842
101 0.0163641942409842
102 0.0163641942409842
103 0.0163641942409842
104 0.0163641942409842
105 0.0163641942409842
106 0.0163641942409842
107 0.0163641942409842
108 0.0163641942409842
109 0.0163641942409842
110 0.0163641942409842
111 0.015587997742265
};
\addplot [semithick, blue]
table {%
10 0.907487869262695
11 0.668221771717072
12 0.652119755744934
13 0.530646562576294
14 0.471608489751816
15 0.285724401473999
16 0.276374369859695
17 0.237694814801216
18 0.219054535031319
19 0.212447762489319
20 0.189719870686531
21 0.188153833150864
22 0.188153833150864
23 0.165295168757439
24 0.151835352182388
25 0.140168473124504
26 0.117336705327034
27 0.114927768707275
28 0.114927768707275
29 0.105701923370361
30 0.0961324423551559
31 0.0857119113206863
32 0.0857119113206863
33 0.0800358057022095
34 0.0739568024873734
35 0.0564193017780781
36 0.0529865995049477
37 0.0475018732249737
38 0.0438045747578144
39 0.0345634706318378
40 0.0345634706318378
41 0.0330744013190269
42 0.0329201929271221
43 0.0255005843937397
44 0.0255005843937397
45 0.0245873685926199
46 0.0245873685926199
47 0.0236259456723928
48 0.0236259456723928
49 0.0194187872111797
50 0.0194187872111797
51 0.0194187872111797
52 0.0194187872111797
53 0.0194187872111797
54 0.0194187872111797
55 0.0179420225322247
56 0.0179420225322247
57 0.0179420225322247
58 0.0179420225322247
59 0.0179420225322247
60 0.0179420225322247
61 0.0179420225322247
62 0.0174288041889668
63 0.0160396583378315
64 0.0160396583378315
65 0.0140246627852321
66 0.0140246627852321
67 0.0140246627852321
68 0.0118159055709839
69 0.0118159055709839
70 0.0118159055709839
71 0.0118159055709839
72 0.0111029390245676
73 0.0111029390245676
74 0.0111029390245676
75 0.0111029390245676
76 0.0111029390245676
77 0.0111029390245676
78 0.0111029390245676
79 0.0111029390245676
80 0.0111029390245676
81 0.0111029390245676
82 0.0111029390245676
83 0.0111029390245676
84 0.0111029390245676
85 0.0111029390245676
86 0.0111029390245676
87 0.0111029390245676
88 0.0107470992952585
89 0.0107470992952585
90 0.0088947294279933
91 0.0088947294279933
92 0.00844688434153795
93 0.00836732424795628
94 0.00836732424795628
95 0.00836732424795628
96 0.00836732424795628
97 0.00836732424795628
98 0.00836732424795628
99 0.00836732424795628
100 0.00836732424795628
101 0.00836732424795628
102 0.00836732424795628
103 0.00836732424795628
104 0.00836732424795628
105 0.00836732424795628
106 0.00836732424795628
107 0.00836732424795628
108 0.00836732424795628
109 0.00836732424795628
110 0.00836732424795628
111 0.00836732424795628
};
\addplot [semithick, color1]
table {%
10 0.907487679748535
11 0.907487679748535
12 0.767740198602294
13 0.699796259525251
14 0.513815225230956
15 0.508570415687697
16 0.364014788784222
17 0.280048235408296
18 0.180138816683954
19 0.143297176407039
20 0.102578665388467
21 0.0827498876355326
22 0.0697465159446743
23 0.0606685514589238
24 0.0303344331867
25 0.0290893167768509
26 0.0275363857909354
27 0.0226968994899096
28 0.0210758231914292
29 0.01506240457346
30 0.0129136485186801
31 0.0108921264849438
32 0.00982559261345224
33 0.00762892427411486
34 0.00615166991360443
35 0.00558697398886507
36 0.00485055466387294
37 0.0046646951484473
38 0.00377006477656683
39 0.00295220052335847
40 0.00281925764008917
41 0.00267871784223681
42 0.00234541554960539
43 0.00190030393959293
44 0.00190030393959293
45 0.00190030393959293
46 0.00190030393959293
47 0.00151398077547613
48 0.00151398077547613
49 0.00151398077547613
50 0.00151398077547613
51 0.0011253489218277
52 0.00103379029095483
53 0.00103379029095483
54 0.00103379029095483
55 0.00103379029095483
56 0.000986818056915562
57 0.000986818056915562
58 0.000939347470154939
59 0.000846584859071253
60 0.000783300931186526
61 0.000783300931186526
62 0.000783300931186526
63 0.000778599313102246
64 0.000778599313102246
65 0.000778599313102246
66 0.000778599313102246
67 0.000778599313102246
68 0.000778599313102246
69 0.000778599313102246
70 0.000778599313102246
71 0.000778599313102246
72 0.000778599313102246
73 0.000778599313102246
74 0.000778599313102246
75 0.000778599313102246
76 0.000778599313102246
77 0.000778599313102246
78 0.000778599313102246
79 0.000685989303986112
80 0.000685989303986112
81 0.000685989303986112
82 0.000685989303986112
83 0.000685989303986112
84 0.000685989303986112
85 0.000685989303986112
86 0.00053596041967845
87 0.00053596041967845
88 0.00053596041967845
89 0.00053596041967845
90 0.00053596041967845
91 0.00053596041967845
92 0.00053596041967845
93 0.00053596041967845
94 0.00053596041967845
95 0.00053596041967845
96 0.00053596041967845
97 0.000512852630118799
98 0.000512852630118799
99 0.000512852630118799
100 0.000512852630118799
101 0.000512852630118799
102 0.000512852630118799
103 0.000512852630118799
104 0.000512852630118799
105 0.000512852630118799
106 0.000510143350816294
107 0.000510143350816294
108 0.000510143350816294
109 0.000510143350816294
110 0.000510143350816294
111 0.000510143350816294
};
\addplot [semithick, blue, dash dot]
table {%
10 0.907487750053406
11 0.682990431785583
12 0.479162514209747
13 0.415064960718155
14 0.222543567419052
15 0.222543567419052
16 0.193192631006241
17 0.175542682409286
18 0.154998376965523
19 0.143563508987427
20 0.128043502569199
21 0.123514726758003
22 0.0984752178192139
23 0.0805199146270752
24 0.0720316171646118
25 0.0720316171646118
26 0.0720316171646118
27 0.0714661628007889
28 0.0695427656173706
29 0.0695427656173706
30 0.0686380863189697
31 0.0494800098240376
32 0.0494800098240376
33 0.0494800098240376
34 0.0475114360451698
35 0.0446216352283955
36 0.0444898121058941
37 0.0444898121058941
38 0.0444898121058941
39 0.0387635938823223
40 0.0371999032795429
41 0.0369647964835167
42 0.0360771901905537
43 0.0302151683717966
44 0.0247646570205688
45 0.0240962021052837
46 0.0240962021052837
47 0.0230401996523142
48 0.0230401996523142
49 0.0230401996523142
50 0.0230401996523142
51 0.0230401996523142
52 0.0218976251780987
53 0.0218976251780987
54 0.0218976251780987
55 0.0218976251780987
56 0.0218976251780987
57 0.0218976251780987
58 0.0211747642606497
59 0.0211747642606497
60 0.0211747642606497
61 0.0211747642606497
62 0.0170412547886372
63 0.0170412547886372
64 0.0170412547886372
65 0.0170412547886372
66 0.0161853544414043
67 0.0157368890941143
68 0.0157368890941143
69 0.0157368890941143
70 0.0157368890941143
71 0.0157368890941143
72 0.0157368890941143
73 0.0157368890941143
74 0.0157368890941143
75 0.0157368890941143
76 0.0157368890941143
77 0.0157368890941143
78 0.0157368890941143
79 0.0157368890941143
80 0.0157368890941143
81 0.0157368890941143
82 0.0157368890941143
83 0.0157368890941143
84 0.0157368890941143
85 0.0157368890941143
86 0.014293884858489
87 0.0134098287671804
88 0.0134098287671804
89 0.0134098287671804
90 0.0134098287671804
91 0.0132410526275635
92 0.0132410526275635
93 0.0132410526275635
94 0.0132410526275635
95 0.0132410526275635
96 0.0132410526275635
97 0.0132410526275635
98 0.0132410526275635
99 0.0132410526275635
100 0.0130109069868922
101 0.0129416463896632
102 0.0129416463896632
103 0.0129416463896632
104 0.0123203992843628
105 0.0123203992843628
106 0.0123203992843628
107 0.0123203992843628
108 0.0123203992843628
109 0.0123203992843628
110 0.0123203992843628
111 0.0123203992843628
};
\end{axis}

\end{tikzpicture}

%% file: bop_strong.tex

\begin{figure*}[!ht]
  \centering
  \scriptsize
 
  \setlength{\figurewidth}{.21\textwidth}
  \setlength{\figureheight}{.75\figurewidth}
  \centerline{\legendACETS \quad \legendACEPTS \quad \legendGPTS \quad \legendpiBOTS}
  
  \vspace{0.2cm}
  \begin{minipage}[b]{\textwidth}  
    \centering
    \begin{subfigure}[b]{.28\textwidth}
      \centering
      \input{figures/bop_1d_ackley1d_strong}
    \end{subfigure}
    \begin{subfigure}[b]{.25\textwidth}
      \centering
      \input{figures/bop_1d_gramacylee_strong}
    \end{subfigure}
    \begin{subfigure}[b]{.25\textwidth}
      \centering
      \input{figures/bop_1d_negeasom_strong}
    \end{subfigure}
  \end{minipage}  

  \begin{minipage}[b]{\textwidth}  
    \centering
    \begin{subfigure}[b]{.28\textwidth}
      \centering
      \input{figures/bop_2d_braninscaled_strong}
    \end{subfigure}
    \begin{subfigure}[b]{.25\textwidth}
      \centering
      \input{figures/bop_2d_ackley_strong}
    \end{subfigure}
    \begin{subfigure}[b]{.25\textwidth}
      \centering
      \input{figures/bop_3d_hartmann3d_strong}
    \end{subfigure}
  \end{minipage}  
  \vspace{-0.5cm}
  \caption{\textbf{Bayesian optimization with strong prior.} Simple regret (mean ± standard error). When strong priors are used, the gap between ACE-TS and ACEP-TS is more evident compared to weak priors.}
  \label{fig:bo_comparisons_strong}
  \vspace{-0.3cm}
\end{figure*}

%% file: figures/bop_1d_ackley1d_strong.tex
\begin{tikzpicture}

\definecolor{color0}{rgb}{0,0,1}
\definecolor{color1}{rgb}{1,0.549019607843137,0}
\definecolor{color2}{rgb}{1,0.647058823529412,0}
\definecolor{color3}{rgb}{0.564705882352941,0.933333333333333,0.564705882352941}

\begin{axis}[axis on top,
enlarge x limits=false,
enlarge y limits=false,
height=\figureheight,
scale only axis,
tick align=outside,
tick pos=left,
tick pos=left,
width=\figurewidth,
xmin=3, xmax=25,
xtick style={color=black},
xtick={-10,0,10,25,50,75,100},
xticklabels={\ensuremath{-}10,0,10,25,50,75,90},
ylabel={Regret},
ymin=-0.2, ymax=4.2,
ytick style={color=black},
ytick={0.   , 4.2},
]
\node[anchor=north east] at (rel axis cs:1,1) {Ackley 1D (strong)};
\path [draw=color1, fill=color1, opacity=0.3]
(axis cs:3,4.02692893971875)
--(axis cs:3,3.24591423044726)
--(axis cs:4,1.61744687735837)
--(axis cs:5,1.61721753209596)
--(axis cs:6,0.641160844724154)
--(axis cs:7,0.453542466505456)
--(axis cs:8,0.303405317564538)
--(axis cs:9,0.207774856120005)
--(axis cs:10,0.194409048965137)
--(axis cs:11,0.128829961215399)
--(axis cs:12,0.104413181862479)
--(axis cs:13,0.0964897535404206)
--(axis cs:14,0.0942621270138992)
--(axis cs:15,0.0885791397318907)
--(axis cs:16,0.0848179083862628)
--(axis cs:17,0.0848179083862628)
--(axis cs:18,0.0846038464344521)
--(axis cs:19,0.0783358572959854)
--(axis cs:20,0.0783120198625782)
--(axis cs:21,0.0778106181905141)
--(axis cs:22,0.0757098468859014)
--(axis cs:23,0.0740378100376961)
--(axis cs:24,0.0740378100376961)
--(axis cs:25,0.0732272799231552)
--(axis cs:26,0.0732272799231552)
--(axis cs:27,0.0731836519091741)
--(axis cs:28,0.0728189676158646)
--(axis cs:29,0.0727942809138637)
--(axis cs:30,0.0727942809138637)
--(axis cs:31,0.0727942809138637)
--(axis cs:32,0.0718922760648874)
--(axis cs:33,0.0718922760648874)
--(axis cs:34,0.0718922760648874)
--(axis cs:35,0.0718922760648874)
--(axis cs:36,0.0718922760648874)
--(axis cs:37,0.0718922760648874)
--(axis cs:38,0.071551574006764)
--(axis cs:39,0.071551574006764)
--(axis cs:40,0.071551574006764)
--(axis cs:41,0.071551574006764)
--(axis cs:42,0.071551574006764)
--(axis cs:43,0.071551574006764)
--(axis cs:44,0.071551574006764)
--(axis cs:45,0.071551574006764)
--(axis cs:46,0.071551574006764)
--(axis cs:47,0.071551574006764)
--(axis cs:48,0.071551574006764)
--(axis cs:49,0.071551574006764)
--(axis cs:50,0.071551574006764)
--(axis cs:51,0.071551574006764)
--(axis cs:52,0.071551574006764)
--(axis cs:53,0.071551574006764)
--(axis cs:54,0.071551574006764)
--(axis cs:55,0.071551574006764)
--(axis cs:56,0.071551574006764)
--(axis cs:57,0.071551574006764)
--(axis cs:58,0.0709380407930774)
--(axis cs:59,0.0706919561203712)
--(axis cs:60,0.0706919561203712)
--(axis cs:61,0.0706919561203712)
--(axis cs:62,0.0703845294194178)
--(axis cs:63,0.0703845294194178)
--(axis cs:64,0.0703845294194178)
--(axis cs:65,0.0703845294194178)
--(axis cs:66,0.0703845294194178)
--(axis cs:67,0.0703845294194178)
--(axis cs:68,0.0703845294194178)
--(axis cs:69,0.0703843269736697)
--(axis cs:70,0.0703843269736697)
--(axis cs:71,0.0703843269736697)
--(axis cs:72,0.0703843269736697)
--(axis cs:73,0.0703843269736697)
--(axis cs:74,0.0703843269736697)
--(axis cs:75,0.0703843269736697)
--(axis cs:76,0.0703843269736697)
--(axis cs:77,0.0703843269736697)
--(axis cs:78,0.0703843269736697)
--(axis cs:79,0.0703843269736697)
--(axis cs:80,0.0703843269736697)
--(axis cs:81,0.0703843269736697)
--(axis cs:82,0.0703843269736697)
--(axis cs:83,0.0703843269736697)
--(axis cs:84,0.0703772702086481)
--(axis cs:85,0.0703772702086481)
--(axis cs:86,0.0703772702086481)
--(axis cs:87,0.0703213130022841)
--(axis cs:88,0.0703213130022841)
--(axis cs:89,0.0703213130022841)
--(axis cs:90,0.0701000121315479)
--(axis cs:91,0.0701000121315479)
--(axis cs:92,0.0701000121315479)
--(axis cs:93,0.0699688882599301)
--(axis cs:94,0.0699688882599301)
--(axis cs:95,0.0699688882599301)
--(axis cs:96,0.0699688882599301)
--(axis cs:97,0.0699688882599301)
--(axis cs:98,0.0699688882599301)
--(axis cs:99,0.0699688882599301)
--(axis cs:100,0.0699688882599301)
--(axis cs:101,0.0699688882599301)
--(axis cs:102,0.0699688882599301)
--(axis cs:103,0.0699688882599301)
--(axis cs:104,0.0699688882599301)
--(axis cs:104,0.070427220337766)
--(axis cs:104,0.070427220337766)
--(axis cs:103,0.070427220337766)
--(axis cs:102,0.070427220337766)
--(axis cs:101,0.070427220337766)
--(axis cs:100,0.070427220337766)
--(axis cs:99,0.070427220337766)
--(axis cs:98,0.070427220337766)
--(axis cs:97,0.070427220337766)
--(axis cs:96,0.070427220337766)
--(axis cs:95,0.070427220337766)
--(axis cs:94,0.070427220337766)
--(axis cs:93,0.070427220337766)
--(axis cs:92,0.07061580601675)
--(axis cs:91,0.07061580601675)
--(axis cs:90,0.07061580601675)
--(axis cs:89,0.0709702275348575)
--(axis cs:88,0.0709702275348575)
--(axis cs:87,0.0709702275348575)
--(axis cs:86,0.0711526336177445)
--(axis cs:85,0.0711526336177445)
--(axis cs:84,0.0711526336177445)
--(axis cs:83,0.07116118271095)
--(axis cs:82,0.07116118271095)
--(axis cs:81,0.07116118271095)
--(axis cs:80,0.07116118271095)
--(axis cs:79,0.07116118271095)
--(axis cs:78,0.07116118271095)
--(axis cs:77,0.07116118271095)
--(axis cs:76,0.07116118271095)
--(axis cs:75,0.07116118271095)
--(axis cs:74,0.07116118271095)
--(axis cs:73,0.07116118271095)
--(axis cs:72,0.07116118271095)
--(axis cs:71,0.07116118271095)
--(axis cs:70,0.07116118271095)
--(axis cs:69,0.07116118271095)
--(axis cs:68,0.0711613181423208)
--(axis cs:67,0.0711613181423208)
--(axis cs:66,0.0711613181423208)
--(axis cs:65,0.0711613181423208)
--(axis cs:64,0.0711613181423208)
--(axis cs:63,0.0711613181423208)
--(axis cs:62,0.0711613181423208)
--(axis cs:61,0.0717676072240523)
--(axis cs:60,0.0717676072240523)
--(axis cs:59,0.0717676072240523)
--(axis cs:58,0.0720475432819734)
--(axis cs:57,0.0734046687445981)
--(axis cs:56,0.0734046687445981)
--(axis cs:55,0.0734046687445981)
--(axis cs:54,0.0734046687445981)
--(axis cs:53,0.0734046687445981)
--(axis cs:52,0.0734046687445981)
--(axis cs:51,0.0734046687445981)
--(axis cs:50,0.0734046687445981)
--(axis cs:49,0.0734046687445981)
--(axis cs:48,0.0734046687445981)
--(axis cs:47,0.0734046687445981)
--(axis cs:46,0.0734046687445981)
--(axis cs:45,0.0734046687445981)
--(axis cs:44,0.0734046687445981)
--(axis cs:43,0.0734046687445981)
--(axis cs:42,0.0734046687445981)
--(axis cs:41,0.0734046687445981)
--(axis cs:40,0.0734046687445981)
--(axis cs:39,0.0734046687445981)
--(axis cs:38,0.0734046687445981)
--(axis cs:37,0.0738024724570175)
--(axis cs:36,0.0738024724570175)
--(axis cs:35,0.0738024724570175)
--(axis cs:34,0.0738024724570175)
--(axis cs:33,0.0738024724570175)
--(axis cs:32,0.0738024724570175)
--(axis cs:31,0.0766490454035509)
--(axis cs:30,0.0766490454035509)
--(axis cs:29,0.0766490454035509)
--(axis cs:28,0.0766694952651732)
--(axis cs:27,0.0775268743983173)
--(axis cs:26,0.0775705239715686)
--(axis cs:25,0.0775705239715686)
--(axis cs:24,0.078191258497788)
--(axis cs:23,0.078191258497788)
--(axis cs:22,0.0822338843746217)
--(axis cs:21,0.0852257579569352)
--(axis cs:20,0.0856091421579031)
--(axis cs:19,0.0856298272658161)
--(axis cs:18,0.0936047871771668)
--(axis cs:17,0.0936831730106951)
--(axis cs:16,0.0936831730106951)
--(axis cs:15,0.0995585224683891)
--(axis cs:14,0.10750188593425)
--(axis cs:13,0.110329051031639)
--(axis cs:12,0.12487482889322)
--(axis cs:11,0.182630734808914)
--(axis cs:10,0.308195306093359)
--(axis cs:9,0.369167169537652)
--(axis cs:8,0.484337813055363)
--(axis cs:7,0.931933006457806)
--(axis cs:6,1.26423375789488)
--(axis cs:5,2.83759236065433)
--(axis cs:4,2.838315448052)
--(axis cs:3,4.02692893971875)
--cycle;

\path [draw=color1, fill=color1, opacity=0.3]
(axis cs:3,4.02692893971875)
--(axis cs:3,3.24591423044726)
--(axis cs:4,2.12591617839677)
--(axis cs:5,1.19931001009678)
--(axis cs:6,1.14690517467796)
--(axis cs:7,0.828767754522432)
--(axis cs:8,0.631058880395918)
--(axis cs:9,0.492178078716041)
--(axis cs:10,0.390571937736771)
--(axis cs:11,0.364446167633503)
--(axis cs:12,0.33444092815105)
--(axis cs:13,0.274173861270599)
--(axis cs:14,0.185578109406224)
--(axis cs:15,0.160382197054706)
--(axis cs:16,0.0745736749112644)
--(axis cs:17,0.0738963271862617)
--(axis cs:18,0.0738963271862617)
--(axis cs:19,0.0733253978243857)
--(axis cs:20,0.0716348401276818)
--(axis cs:21,0.0716144692165654)
--(axis cs:22,0.0715330770820099)
--(axis cs:23,0.0714930259184115)
--(axis cs:24,0.0708359549789784)
--(axis cs:25,0.0706055866873039)
--(axis cs:26,0.0704130546195663)
--(axis cs:27,0.0704130546195663)
--(axis cs:28,0.0701970174469814)
--(axis cs:29,0.0701280956951619)
--(axis cs:30,0.0701280956951619)
--(axis cs:31,0.0701280956951619)
--(axis cs:32,0.0700415381775725)
--(axis cs:33,0.0700415381775725)
--(axis cs:34,0.0700415381775725)
--(axis cs:35,0.0700222073271752)
--(axis cs:36,0.0700222073271752)
--(axis cs:37,0.0700222073271752)
--(axis cs:38,0.0700039023356184)
--(axis cs:39,0.0700039023356184)
--(axis cs:40,0.0699302252757689)
--(axis cs:41,0.0699092942740131)
--(axis cs:42,0.0699092942740131)
--(axis cs:43,0.0698549165994371)
--(axis cs:44,0.0698549165994371)
--(axis cs:45,0.0698237325772185)
--(axis cs:46,0.0698237325772185)
--(axis cs:47,0.0697971197920841)
--(axis cs:48,0.0697231963536957)
--(axis cs:49,0.0697231963536957)
--(axis cs:50,0.0697231963536957)
--(axis cs:51,0.0697231963536957)
--(axis cs:52,0.0697231963536957)
--(axis cs:53,0.0697231963536957)
--(axis cs:54,0.0697231963536957)
--(axis cs:55,0.0697231963536957)
--(axis cs:56,0.0697203480770026)
--(axis cs:57,0.0697203480770026)
--(axis cs:58,0.0697203480770026)
--(axis cs:59,0.0697203480770026)
--(axis cs:60,0.0697142610025255)
--(axis cs:61,0.0697142610025255)
--(axis cs:62,0.0696695071393999)
--(axis cs:63,0.0696695071393999)
--(axis cs:64,0.0696512293631799)
--(axis cs:65,0.0696512293631799)
--(axis cs:66,0.0696512293631799)
--(axis cs:67,0.0696512293631799)
--(axis cs:68,0.0696512293631799)
--(axis cs:69,0.0696512293631799)
--(axis cs:70,0.0696512293631799)
--(axis cs:71,0.0696512293631799)
--(axis cs:72,0.0696512293631799)
--(axis cs:73,0.0696512293631799)
--(axis cs:74,0.0696512293631799)
--(axis cs:75,0.0696512293631799)
--(axis cs:76,0.0696512293631799)
--(axis cs:77,0.0696512293631799)
--(axis cs:78,0.0696512293631799)
--(axis cs:79,0.0696512293631799)
--(axis cs:80,0.0696512293631799)
--(axis cs:81,0.0696512293631799)
--(axis cs:82,0.0696512293631799)
--(axis cs:83,0.0696512293631799)
--(axis cs:84,0.0696512293631799)
--(axis cs:85,0.0696512293631799)
--(axis cs:86,0.0696512293631799)
--(axis cs:87,0.0696512293631799)
--(axis cs:88,0.0696512293631799)
--(axis cs:89,0.0696512293631799)
--(axis cs:90,0.0696512293631799)
--(axis cs:91,0.0696512293631799)
--(axis cs:92,0.0696386418746001)
--(axis cs:93,0.0696386418746001)
--(axis cs:94,0.0696386418746001)
--(axis cs:95,0.0696386418746001)
--(axis cs:96,0.0696376730538561)
--(axis cs:97,0.0696376730538561)
--(axis cs:98,0.0696376730538561)
--(axis cs:99,0.0696376730538561)
--(axis cs:100,0.0696247297633686)
--(axis cs:101,0.0696247297633686)
--(axis cs:102,0.0696247297633686)
--(axis cs:103,0.0696239984926955)
--(axis cs:104,0.0696239984926955)
--(axis cs:104,0.0696556234378365)
--(axis cs:104,0.0696556234378365)
--(axis cs:103,0.0696556234378365)
--(axis cs:102,0.0696565787091741)
--(axis cs:101,0.0696565787091741)
--(axis cs:100,0.0696565787091741)
--(axis cs:99,0.0697283657552882)
--(axis cs:98,0.0697283657552882)
--(axis cs:97,0.0697283657552882)
--(axis cs:96,0.0697283657552882)
--(axis cs:95,0.0697726903726123)
--(axis cs:94,0.0697726903726123)
--(axis cs:93,0.0697726903726123)
--(axis cs:92,0.0697726903726123)
--(axis cs:91,0.0697833774789716)
--(axis cs:90,0.0697833774789716)
--(axis cs:89,0.0697833774789716)
--(axis cs:88,0.0697833774789716)
--(axis cs:87,0.0697833774789716)
--(axis cs:86,0.0697833774789716)
--(axis cs:85,0.0697833774789716)
--(axis cs:84,0.0697833774789716)
--(axis cs:83,0.0697833774789716)
--(axis cs:82,0.0697833774789716)
--(axis cs:81,0.0697833774789716)
--(axis cs:80,0.0697833774789716)
--(axis cs:79,0.0697833774789716)
--(axis cs:78,0.0697833774789716)
--(axis cs:77,0.0697833774789716)
--(axis cs:76,0.0697833774789716)
--(axis cs:75,0.0697833774789716)
--(axis cs:74,0.0697833774789716)
--(axis cs:73,0.0697833774789716)
--(axis cs:72,0.0697833774789716)
--(axis cs:71,0.0697833774789716)
--(axis cs:70,0.0697833774789716)
--(axis cs:69,0.0697833774789716)
--(axis cs:68,0.0697833774789716)
--(axis cs:67,0.0697833774789716)
--(axis cs:66,0.0697833774789716)
--(axis cs:65,0.0697833774789716)
--(axis cs:64,0.0697833774789716)
--(axis cs:63,0.0698059355849437)
--(axis cs:62,0.0698059355849437)
--(axis cs:61,0.0698905405455647)
--(axis cs:60,0.0698905405455647)
--(axis cs:59,0.0698942335601412)
--(axis cs:58,0.0698942335601412)
--(axis cs:57,0.0698942335601412)
--(axis cs:56,0.0698942335601412)
--(axis cs:55,0.0698957688080756)
--(axis cs:54,0.0698957688080756)
--(axis cs:53,0.0698957688080756)
--(axis cs:52,0.0698957688080756)
--(axis cs:51,0.0698957688080756)
--(axis cs:50,0.0698957688080756)
--(axis cs:49,0.0698957688080756)
--(axis cs:48,0.0698957688080756)
--(axis cs:47,0.0699920542706676)
--(axis cs:46,0.0700374302307978)
--(axis cs:45,0.0700374302307978)
--(axis cs:44,0.070197713289335)
--(axis cs:43,0.070197713289335)
--(axis cs:42,0.0703293523398181)
--(axis cs:41,0.0703293523398181)
--(axis cs:40,0.0703428355313828)
--(axis cs:39,0.0704482103459306)
--(axis cs:38,0.0704482103459306)
--(axis cs:37,0.0704705080281645)
--(axis cs:36,0.0704705080281645)
--(axis cs:35,0.0704705080281645)
--(axis cs:34,0.0704842676040787)
--(axis cs:33,0.0704842676040787)
--(axis cs:32,0.0704842676040787)
--(axis cs:31,0.0706356814605958)
--(axis cs:30,0.0706356814605958)
--(axis cs:29,0.0706356814605958)
--(axis cs:28,0.071191987152063)
--(axis cs:27,0.0713883145131116)
--(axis cs:26,0.0713883145131116)
--(axis cs:25,0.0720960337743115)
--(axis cs:24,0.0723043623931738)
--(axis cs:23,0.0767649951506125)
--(axis cs:22,0.0767969775196454)
--(axis cs:21,0.0952614994183808)
--(axis cs:20,0.297908069669322)
--(axis cs:19,0.546946984667202)
--(axis cs:18,0.547403670740952)
--(axis cs:17,0.547403670740952)
--(axis cs:16,0.547945017245394)
--(axis cs:15,0.64826433227989)
--(axis cs:14,0.699208215969917)
--(axis cs:13,0.941718896468799)
--(axis cs:12,0.98974326736404)
--(axis cs:11,1.01251932496271)
--(axis cs:10,1.03277953630811)
--(axis cs:9,1.1081344219376)
--(axis cs:8,1.2936239132481)
--(axis cs:7,1.5677213802525)
--(axis cs:6,2.09423155660431)
--(axis cs:5,2.13489785683722)
--(axis cs:4,3.33091923987144)
--(axis cs:3,4.02692893971875)
--cycle;

\path [draw=blue, fill=blue, opacity=0.3]
(axis cs:3,4.02692890167236)
--(axis cs:3,3.24591422080994)
--(axis cs:4,2.31809329986572)
--(axis cs:5,1.67434418201447)
--(axis cs:6,1.37629556655884)
--(axis cs:7,0.997538685798645)
--(axis cs:8,0.447340697050095)
--(axis cs:9,0.258528709411621)
--(axis cs:10,0.160151034593582)
--(axis cs:11,0.125659823417664)
--(axis cs:12,0.125659823417664)
--(axis cs:13,0.103180669248104)
--(axis cs:14,0.103180669248104)
--(axis cs:15,0.0925922691822052)
--(axis cs:16,0.0925922691822052)
--(axis cs:17,0.0925922691822052)
--(axis cs:18,0.0925922691822052)
--(axis cs:19,0.0916850417852402)
--(axis cs:20,0.0906670540571213)
--(axis cs:21,0.0906670540571213)
--(axis cs:22,0.0881931334733963)
--(axis cs:23,0.0881931334733963)
--(axis cs:24,0.0881931334733963)
--(axis cs:25,0.0881931334733963)
--(axis cs:26,0.0881931334733963)
--(axis cs:27,0.0881931334733963)
--(axis cs:28,0.0877843275666237)
--(axis cs:29,0.0877843275666237)
--(axis cs:30,0.0842249915003777)
--(axis cs:31,0.083881564438343)
--(axis cs:32,0.083881564438343)
--(axis cs:33,0.083881564438343)
--(axis cs:34,0.0826556757092476)
--(axis cs:35,0.0826556757092476)
--(axis cs:36,0.0819377228617668)
--(axis cs:37,0.0819377228617668)
--(axis cs:38,0.0819377228617668)
--(axis cs:39,0.0819377228617668)
--(axis cs:40,0.0819377228617668)
--(axis cs:41,0.0819377228617668)
--(axis cs:42,0.0819377228617668)
--(axis cs:43,0.0819377228617668)
--(axis cs:44,0.0815583094954491)
--(axis cs:45,0.0815583094954491)
--(axis cs:46,0.0815583094954491)
--(axis cs:47,0.0815583094954491)
--(axis cs:48,0.0815583094954491)
--(axis cs:49,0.0813597962260246)
--(axis cs:50,0.0813597962260246)
--(axis cs:51,0.0813597962260246)
--(axis cs:52,0.0811691284179688)
--(axis cs:53,0.0811691284179688)
--(axis cs:54,0.0811691284179688)
--(axis cs:55,0.0811691284179688)
--(axis cs:56,0.0811691284179688)
--(axis cs:57,0.0811691284179688)
--(axis cs:58,0.0811691284179688)
--(axis cs:59,0.0811691284179688)
--(axis cs:60,0.0811691284179688)
--(axis cs:61,0.0811691284179688)
--(axis cs:62,0.0805334150791168)
--(axis cs:63,0.0805334150791168)
--(axis cs:64,0.0805334150791168)
--(axis cs:65,0.0805334150791168)
--(axis cs:66,0.0805334150791168)
--(axis cs:67,0.0805334150791168)
--(axis cs:68,0.0805334150791168)
--(axis cs:69,0.0805334150791168)
--(axis cs:70,0.0805334150791168)
--(axis cs:71,0.0805334150791168)
--(axis cs:72,0.0805334150791168)
--(axis cs:73,0.0805334150791168)
--(axis cs:74,0.0805334150791168)
--(axis cs:75,0.0805334150791168)
--(axis cs:76,0.0805334150791168)
--(axis cs:77,0.0793914422392845)
--(axis cs:78,0.0793914422392845)
--(axis cs:79,0.0793914422392845)
--(axis cs:80,0.0793914422392845)
--(axis cs:81,0.0793914422392845)
--(axis cs:82,0.0793914422392845)
--(axis cs:83,0.0793914422392845)
--(axis cs:84,0.0793882682919502)
--(axis cs:85,0.0793882682919502)
--(axis cs:86,0.0793882682919502)
--(axis cs:87,0.0793882682919502)
--(axis cs:88,0.0793882682919502)
--(axis cs:89,0.0793882682919502)
--(axis cs:90,0.0793882682919502)
--(axis cs:91,0.0793882682919502)
--(axis cs:92,0.0793882682919502)
--(axis cs:93,0.0793882682919502)
--(axis cs:94,0.0793882682919502)
--(axis cs:95,0.0793882682919502)
--(axis cs:96,0.0793882682919502)
--(axis cs:97,0.0793882682919502)
--(axis cs:98,0.0793882682919502)
--(axis cs:99,0.0793882682919502)
--(axis cs:100,0.0793882682919502)
--(axis cs:101,0.0793882682919502)
--(axis cs:102,0.0775830745697021)
--(axis cs:103,0.0775830745697021)
--(axis cs:104,0.0775830745697021)
--(axis cs:104,0.0832792818546295)
--(axis cs:104,0.0832792818546295)
--(axis cs:103,0.0832792818546295)
--(axis cs:102,0.0832792818546295)
--(axis cs:101,0.0849454626441002)
--(axis cs:100,0.0849454626441002)
--(axis cs:99,0.0849454626441002)
--(axis cs:98,0.0849454626441002)
--(axis cs:97,0.0849454626441002)
--(axis cs:96,0.0849454626441002)
--(axis cs:95,0.0849454626441002)
--(axis cs:94,0.0849454626441002)
--(axis cs:93,0.0849454626441002)
--(axis cs:92,0.0849454626441002)
--(axis cs:91,0.0849454626441002)
--(axis cs:90,0.0849454626441002)
--(axis cs:89,0.0849454626441002)
--(axis cs:88,0.0849454626441002)
--(axis cs:87,0.0849454626441002)
--(axis cs:86,0.0849454626441002)
--(axis cs:85,0.0849454626441002)
--(axis cs:84,0.0849454626441002)
--(axis cs:83,0.0849514380097389)
--(axis cs:82,0.0849514380097389)
--(axis cs:81,0.0849514380097389)
--(axis cs:80,0.0849514380097389)
--(axis cs:79,0.0849514380097389)
--(axis cs:78,0.0849514380097389)
--(axis cs:77,0.0849514380097389)
--(axis cs:76,0.0856687426567078)
--(axis cs:75,0.0856687426567078)
--(axis cs:74,0.0856687426567078)
--(axis cs:73,0.0856687426567078)
--(axis cs:72,0.0856687426567078)
--(axis cs:71,0.0856687426567078)
--(axis cs:70,0.0856687426567078)
--(axis cs:69,0.0856687426567078)
--(axis cs:68,0.0856687426567078)
--(axis cs:67,0.0856687426567078)
--(axis cs:66,0.0856687426567078)
--(axis cs:65,0.0856687426567078)
--(axis cs:64,0.0856687426567078)
--(axis cs:63,0.0856687426567078)
--(axis cs:62,0.0856687426567078)
--(axis cs:61,0.0858665406703949)
--(axis cs:60,0.0858665406703949)
--(axis cs:59,0.0858665406703949)
--(axis cs:58,0.0858665406703949)
--(axis cs:57,0.0858665406703949)
--(axis cs:56,0.0858665406703949)
--(axis cs:55,0.0858665406703949)
--(axis cs:54,0.0858665406703949)
--(axis cs:53,0.0858665406703949)
--(axis cs:52,0.0858665406703949)
--(axis cs:51,0.0859409943223)
--(axis cs:50,0.0859409943223)
--(axis cs:49,0.0859409943223)
--(axis cs:48,0.0861209109425545)
--(axis cs:47,0.0861209109425545)
--(axis cs:46,0.0861209109425545)
--(axis cs:45,0.0861209109425545)
--(axis cs:44,0.0861209109425545)
--(axis cs:43,0.0871906951069832)
--(axis cs:42,0.0871906951069832)
--(axis cs:41,0.0871906951069832)
--(axis cs:40,0.0871906951069832)
--(axis cs:39,0.0871906951069832)
--(axis cs:38,0.0871906951069832)
--(axis cs:37,0.0871906951069832)
--(axis cs:36,0.0871906951069832)
--(axis cs:35,0.0899811163544655)
--(axis cs:34,0.0899811163544655)
--(axis cs:33,0.095549963414669)
--(axis cs:32,0.095549963414669)
--(axis cs:31,0.095549963414669)
--(axis cs:30,0.095904253423214)
--(axis cs:29,0.101551301777363)
--(axis cs:28,0.101551301777363)
--(axis cs:27,0.103525921702385)
--(axis cs:26,0.103525921702385)
--(axis cs:25,0.103525921702385)
--(axis cs:24,0.103525921702385)
--(axis cs:23,0.103525921702385)
--(axis cs:22,0.103525921702385)
--(axis cs:21,0.105911925435066)
--(axis cs:20,0.105911925435066)
--(axis cs:19,0.106471315026283)
--(axis cs:18,0.114139527082443)
--(axis cs:17,0.114139527082443)
--(axis cs:16,0.114139527082443)
--(axis cs:15,0.114139527082443)
--(axis cs:14,0.144792973995209)
--(axis cs:13,0.144792973995209)
--(axis cs:12,0.191076010465622)
--(axis cs:11,0.191076010465622)
--(axis cs:10,0.226716488599777)
--(axis cs:9,0.490164399147034)
--(axis cs:8,1.00446891784668)
--(axis cs:7,1.68604242801666)
--(axis cs:6,2.06967830657959)
--(axis cs:5,2.3374981880188)
--(axis cs:4,2.98376369476318)
--(axis cs:3,4.02692890167236)
--cycle;

\path [draw=blue, fill=blue, opacity=0.3]
(axis cs:3,4.02692890167236)
--(axis cs:3,3.24591422080994)
--(axis cs:4,2.42414212226868)
--(axis cs:5,1.65107309818268)
--(axis cs:6,1.58191621303558)
--(axis cs:7,0.950271368026733)
--(axis cs:8,0.745534956455231)
--(axis cs:9,0.454385101795197)
--(axis cs:10,0.273345768451691)
--(axis cs:11,0.129838421940804)
--(axis cs:12,0.124326050281525)
--(axis cs:13,0.0937928855419159)
--(axis cs:14,0.0887052118778229)
--(axis cs:15,0.0877968370914459)
--(axis cs:16,0.0877968370914459)
--(axis cs:17,0.0808785930275917)
--(axis cs:18,0.0808065161108971)
--(axis cs:19,0.0804344117641449)
--(axis cs:20,0.0804344117641449)
--(axis cs:21,0.0773770213127136)
--(axis cs:22,0.0773770213127136)
--(axis cs:23,0.0760549455881119)
--(axis cs:24,0.0760549455881119)
--(axis cs:25,0.0756019279360771)
--(axis cs:26,0.0745885819196701)
--(axis cs:27,0.0738723576068878)
--(axis cs:28,0.0737547129392624)
--(axis cs:29,0.0737547129392624)
--(axis cs:30,0.0733852609992027)
--(axis cs:31,0.0733852609992027)
--(axis cs:32,0.072766900062561)
--(axis cs:33,0.0727532580494881)
--(axis cs:34,0.0727532580494881)
--(axis cs:35,0.0727532580494881)
--(axis cs:36,0.0727532580494881)
--(axis cs:37,0.0727447867393494)
--(axis cs:38,0.0727447867393494)
--(axis cs:39,0.0727447867393494)
--(axis cs:40,0.0727447867393494)
--(axis cs:41,0.0727447867393494)
--(axis cs:42,0.0727447867393494)
--(axis cs:43,0.0727447867393494)
--(axis cs:44,0.0727447867393494)
--(axis cs:45,0.0727447867393494)
--(axis cs:46,0.0727447867393494)
--(axis cs:47,0.0724849626421928)
--(axis cs:48,0.0719923973083496)
--(axis cs:49,0.0719923973083496)
--(axis cs:50,0.0719923973083496)
--(axis cs:51,0.0719923973083496)
--(axis cs:52,0.0719923973083496)
--(axis cs:53,0.0719923973083496)
--(axis cs:54,0.0719923973083496)
--(axis cs:55,0.0719923973083496)
--(axis cs:56,0.0719923973083496)
--(axis cs:57,0.0717592760920525)
--(axis cs:58,0.0717592760920525)
--(axis cs:59,0.0717592760920525)
--(axis cs:60,0.0717592760920525)
--(axis cs:61,0.0717592760920525)
--(axis cs:62,0.0717592760920525)
--(axis cs:63,0.0717592760920525)
--(axis cs:64,0.0717592760920525)
--(axis cs:65,0.0717592760920525)
--(axis cs:66,0.0717592760920525)
--(axis cs:67,0.0717592760920525)
--(axis cs:68,0.0717592760920525)
--(axis cs:69,0.0717592760920525)
--(axis cs:70,0.0717592760920525)
--(axis cs:71,0.0717592760920525)
--(axis cs:72,0.0717592760920525)
--(axis cs:73,0.0717592760920525)
--(axis cs:74,0.0717592760920525)
--(axis cs:75,0.0717592760920525)
--(axis cs:76,0.0717592760920525)
--(axis cs:77,0.0717592760920525)
--(axis cs:78,0.0717592760920525)
--(axis cs:79,0.0717592760920525)
--(axis cs:80,0.0717592760920525)
--(axis cs:81,0.0716851726174355)
--(axis cs:82,0.0716851726174355)
--(axis cs:83,0.0716851726174355)
--(axis cs:84,0.0716851726174355)
--(axis cs:85,0.0711221247911453)
--(axis cs:86,0.0711221247911453)
--(axis cs:87,0.0711221247911453)
--(axis cs:88,0.0711221247911453)
--(axis cs:89,0.0711221247911453)
--(axis cs:90,0.0711221247911453)
--(axis cs:91,0.0711221247911453)
--(axis cs:92,0.0711221247911453)
--(axis cs:93,0.0711221247911453)
--(axis cs:94,0.0711221247911453)
--(axis cs:95,0.0711221247911453)
--(axis cs:96,0.0711221247911453)
--(axis cs:97,0.0711221247911453)
--(axis cs:98,0.0710048973560333)
--(axis cs:99,0.0710048973560333)
--(axis cs:100,0.0710048973560333)
--(axis cs:101,0.0710048973560333)
--(axis cs:102,0.0710048973560333)
--(axis cs:103,0.0710048973560333)
--(axis cs:104,0.0710048973560333)
--(axis cs:104,0.071651428937912)
--(axis cs:104,0.071651428937912)
--(axis cs:103,0.071651428937912)
--(axis cs:102,0.071651428937912)
--(axis cs:101,0.071651428937912)
--(axis cs:100,0.071651428937912)
--(axis cs:99,0.071651428937912)
--(axis cs:98,0.071651428937912)
--(axis cs:97,0.0720781832933426)
--(axis cs:96,0.0720781832933426)
--(axis cs:95,0.0720781832933426)
--(axis cs:94,0.0720781832933426)
--(axis cs:93,0.0720781832933426)
--(axis cs:92,0.0720781832933426)
--(axis cs:91,0.0720781832933426)
--(axis cs:90,0.0720781832933426)
--(axis cs:89,0.0720781832933426)
--(axis cs:88,0.0720781832933426)
--(axis cs:87,0.0720781832933426)
--(axis cs:86,0.0720781832933426)
--(axis cs:85,0.0720781832933426)
--(axis cs:84,0.0733984485268593)
--(axis cs:83,0.0733984485268593)
--(axis cs:82,0.0733984485268593)
--(axis cs:81,0.0733984485268593)
--(axis cs:80,0.0761235728859901)
--(axis cs:79,0.0761235728859901)
--(axis cs:78,0.0761235728859901)
--(axis cs:77,0.0761235728859901)
--(axis cs:76,0.0761235728859901)
--(axis cs:75,0.0761235728859901)
--(axis cs:74,0.0761235728859901)
--(axis cs:73,0.0761235728859901)
--(axis cs:72,0.0761235728859901)
--(axis cs:71,0.0761235728859901)
--(axis cs:70,0.0761235728859901)
--(axis cs:69,0.0761235728859901)
--(axis cs:68,0.0761235728859901)
--(axis cs:67,0.0761235728859901)
--(axis cs:66,0.0761235728859901)
--(axis cs:65,0.0761235728859901)
--(axis cs:64,0.0761235728859901)
--(axis cs:63,0.0761235728859901)
--(axis cs:62,0.0761235728859901)
--(axis cs:61,0.0761235728859901)
--(axis cs:60,0.0761235728859901)
--(axis cs:59,0.0761235728859901)
--(axis cs:58,0.0761235728859901)
--(axis cs:57,0.0761235728859901)
--(axis cs:56,0.0763165503740311)
--(axis cs:55,0.0763165503740311)
--(axis cs:54,0.0763165503740311)
--(axis cs:53,0.0763165503740311)
--(axis cs:52,0.0763165503740311)
--(axis cs:51,0.0763165503740311)
--(axis cs:50,0.0763165503740311)
--(axis cs:49,0.0763165503740311)
--(axis cs:48,0.0763165503740311)
--(axis cs:47,0.0768917128443718)
--(axis cs:46,0.0770892798900604)
--(axis cs:45,0.0770892798900604)
--(axis cs:44,0.0770892798900604)
--(axis cs:43,0.0770892798900604)
--(axis cs:42,0.0770892798900604)
--(axis cs:41,0.0770892798900604)
--(axis cs:40,0.0770892798900604)
--(axis cs:39,0.0770892798900604)
--(axis cs:38,0.0770892798900604)
--(axis cs:37,0.0770892798900604)
--(axis cs:36,0.0773562267422676)
--(axis cs:35,0.0773562267422676)
--(axis cs:34,0.0773562267422676)
--(axis cs:33,0.0773562267422676)
--(axis cs:32,0.0778655707836151)
--(axis cs:31,0.0784972980618477)
--(axis cs:30,0.0784972980618477)
--(axis cs:29,0.0787893086671829)
--(axis cs:28,0.0787893086671829)
--(axis cs:27,0.0808216333389282)
--(axis cs:26,0.0818590223789215)
--(axis cs:25,0.0833099707961082)
--(axis cs:24,0.083613783121109)
--(axis cs:23,0.083613783121109)
--(axis cs:22,0.085030660033226)
--(axis cs:21,0.085030660033226)
--(axis cs:20,0.088664636015892)
--(axis cs:19,0.088664636015892)
--(axis cs:18,0.0892046168446541)
--(axis cs:17,0.0892424061894417)
--(axis cs:16,0.0958327651023865)
--(axis cs:15,0.0958327651023865)
--(axis cs:14,0.0961840152740479)
--(axis cs:13,0.119586020708084)
--(axis cs:12,0.152567565441132)
--(axis cs:11,0.238023892045021)
--(axis cs:10,0.812305271625519)
--(axis cs:9,0.955603301525116)
--(axis cs:8,1.22366583347321)
--(axis cs:7,1.73684763908386)
--(axis cs:6,2.28357124328613)
--(axis cs:5,2.36743712425232)
--(axis cs:4,3.20135474205017)
--(axis cs:3,4.02692890167236)
--cycle;

\addplot [semithick, color1, dash dot]
table {%
3 3.63642158508301
4 2.22788116270519
5 2.22740494637515
6 0.952697301309517
7 0.692737736481631
8 0.39387156530995
9 0.288471012828828
10 0.251302177529248
11 0.155730348012157
12 0.11464400537785
13 0.10340940228603
14 0.100882006474075
15 0.0940688311001399
16 0.089250540698479
17 0.089250540698479
18 0.0891043168058094
19 0.0819828422809008
20 0.0819605810102407
21 0.0815181880737247
22 0.0789718656302616
23 0.0761145342677421
24 0.0761145342677421
25 0.0753989019473619
26 0.0753989019473619
27 0.0753552631537457
28 0.0747442314405189
29 0.0747216631587073
30 0.0747216631587073
31 0.0747216631587073
32 0.0728473742609524
33 0.0728473742609524
34 0.0728473742609524
35 0.0728473742609524
36 0.0728473742609524
37 0.0728473742609524
38 0.0724781213756811
39 0.0724781213756811
40 0.0724781213756811
41 0.0724781213756811
42 0.0724781213756811
43 0.0724781213756811
44 0.0724781213756811
45 0.0724781213756811
46 0.0724781213756811
47 0.0724781213756811
48 0.0724781213756811
49 0.0724781213756811
50 0.0724781213756811
51 0.0724781213756811
52 0.0724781213756811
53 0.0724781213756811
54 0.0724781213756811
55 0.0724781213756811
56 0.0724781213756811
57 0.0724781213756811
58 0.0714927920375254
59 0.0712297816722118
60 0.0712297816722118
61 0.0712297816722118
62 0.0707729237808693
63 0.0707729237808693
64 0.0707729237808693
65 0.0707729237808693
66 0.0707729237808693
67 0.0707729237808693
68 0.0707729237808693
69 0.0707727548423098
70 0.0707727548423098
71 0.0707727548423098
72 0.0707727548423098
73 0.0707727548423098
74 0.0707727548423098
75 0.0707727548423098
76 0.0707727548423098
77 0.0707727548423098
78 0.0707727548423098
79 0.0707727548423098
80 0.0707727548423098
81 0.0707727548423098
82 0.0707727548423098
83 0.0707727548423098
84 0.0707649519131963
85 0.0707649519131963
86 0.0707649519131963
87 0.0706457702685708
88 0.0706457702685708
89 0.0706457702685708
90 0.0703579090741489
91 0.0703579090741489
92 0.0703579090741489
93 0.0701980542988481
94 0.0701980542988481
95 0.0701980542988481
96 0.0701980542988481
97 0.0701980542988481
98 0.0701980542988481
99 0.0701980542988481
100 0.0701980542988481
101 0.0701980542988481
102 0.0701980542988481
103 0.0701980542988481
104 0.0701980542988481
};
\addplot [semithick, color1]
table {%
3 3.63642158508301
4 2.7284177091341
5 1.667103933467
6 1.62056836564114
7 1.19824456738746
8 0.962341396822011
9 0.800156250326819
10 0.711675737022443
11 0.688482746298107
12 0.662092097757545
13 0.607946378869699
14 0.44239316268807
15 0.404323264667298
16 0.311259346078329
17 0.310649998963607
18 0.310649998963607
19 0.310136191245794
20 0.184771454898502
21 0.0834379843174731
22 0.0741650273008276
23 0.074129010534512
24 0.0715701586860761
25 0.0713508102308077
26 0.070900684566339
27 0.070900684566339
28 0.0706945022995222
29 0.0703818885778788
30 0.0703818885778788
31 0.0703818885778788
32 0.0702629028908256
33 0.0702629028908256
34 0.0702629028908256
35 0.0702463576776698
36 0.0702463576776698
37 0.0702463576776698
38 0.0702260563407745
39 0.0702260563407745
40 0.0701365304035758
41 0.0701193233069156
42 0.0701193233069156
43 0.070026314944386
44 0.070026314944386
45 0.0699305814040081
46 0.0699305814040081
47 0.0698945870313758
48 0.0698094825808856
49 0.0698094825808856
50 0.0698094825808856
51 0.0698094825808856
52 0.0698094825808856
53 0.0698094825808856
54 0.0698094825808856
55 0.0698094825808856
56 0.0698072908185719
57 0.0698072908185719
58 0.0698072908185719
59 0.0698072908185719
60 0.0698024007740451
61 0.0698024007740451
62 0.0697377213621718
63 0.0697377213621718
64 0.0697173034210757
65 0.0697173034210757
66 0.0697173034210757
67 0.0697173034210757
68 0.0697173034210757
69 0.0697173034210757
70 0.0697173034210757
71 0.0697173034210757
72 0.0697173034210757
73 0.0697173034210757
74 0.0697173034210757
75 0.0697173034210757
76 0.0697173034210757
77 0.0697173034210757
78 0.0697173034210757
79 0.0697173034210757
80 0.0697173034210757
81 0.0697173034210757
82 0.0697173034210757
83 0.0697173034210757
84 0.0697173034210757
85 0.0697173034210757
86 0.0697173034210757
87 0.0697173034210757
88 0.0697173034210757
89 0.0697173034210757
90 0.0697173034210757
91 0.0697173034210757
92 0.0697056661236062
93 0.0697056661236062
94 0.0697056661236062
95 0.0697056661236062
96 0.0696830194045721
97 0.0696830194045721
98 0.0696830194045721
99 0.0696830194045721
100 0.0696406542362713
101 0.0696406542362713
102 0.0696406542362713
103 0.069639810965266
104 0.069639810965266
};
\addplot [semithick, blue, dash dot]
table {%
3 3.63642168045044
4 2.65092849731445
5 2.00592112541199
6 1.72298693656921
7 1.34179055690765
8 0.725904822349548
9 0.374346554279327
10 0.19343376159668
11 0.158367916941643
12 0.158367916941643
13 0.123986817896366
14 0.123986817896366
15 0.103365898132324
16 0.103365898132324
17 0.103365898132324
18 0.103365898132324
19 0.0990781784057617
20 0.0982894897460938
21 0.0982894897460938
22 0.0958595275878906
23 0.0958595275878906
24 0.0958595275878906
25 0.0958595275878906
26 0.0958595275878906
27 0.0958595275878906
28 0.0946678146719933
29 0.0946678146719933
30 0.0900646224617958
31 0.089715763926506
32 0.089715763926506
33 0.089715763926506
34 0.0863183960318565
35 0.0863183960318565
36 0.084564208984375
37 0.084564208984375
38 0.084564208984375
39 0.084564208984375
40 0.084564208984375
41 0.084564208984375
42 0.084564208984375
43 0.084564208984375
44 0.0838396102190018
45 0.0838396102190018
46 0.0838396102190018
47 0.0838396102190018
48 0.0838396102190018
49 0.0836503952741623
50 0.0836503952741623
51 0.0836503952741623
52 0.0835178345441818
53 0.0835178345441818
54 0.0835178345441818
55 0.0835178345441818
56 0.0835178345441818
57 0.0835178345441818
58 0.0835178345441818
59 0.0835178345441818
60 0.0835178345441818
61 0.0835178345441818
62 0.0831010788679123
63 0.0831010788679123
64 0.0831010788679123
65 0.0831010788679123
66 0.0831010788679123
67 0.0831010788679123
68 0.0831010788679123
69 0.0831010788679123
70 0.0831010788679123
71 0.0831010788679123
72 0.0831010788679123
73 0.0831010788679123
74 0.0831010788679123
75 0.0831010788679123
76 0.0831010788679123
77 0.0821714401245117
78 0.0821714401245117
79 0.0821714401245117
80 0.0821714401245117
81 0.0821714401245117
82 0.0821714401245117
83 0.0821714401245117
84 0.0821668654680252
85 0.0821668654680252
86 0.0821668654680252
87 0.0821668654680252
88 0.0821668654680252
89 0.0821668654680252
90 0.0821668654680252
91 0.0821668654680252
92 0.0821668654680252
93 0.0821668654680252
94 0.0821668654680252
95 0.0821668654680252
96 0.0821668654680252
97 0.0821668654680252
98 0.0821668654680252
99 0.0821668654680252
100 0.0821668654680252
101 0.0821668654680252
102 0.0804311782121658
103 0.0804311782121658
104 0.0804311782121658
};
\addplot [semithick, blue]
table {%
3 3.63642168045044
4 2.81274843215942
5 2.00925517082214
6 1.9327437877655
7 1.3435595035553
8 0.984600424766541
9 0.704994201660156
10 0.542825520038605
11 0.183931156992912
12 0.138446807861328
13 0.106689453125
14 0.0924446135759354
15 0.0918148010969162
16 0.0918148010969162
17 0.0850604996085167
18 0.0850055664777756
19 0.0845495238900185
20 0.0845495238900185
21 0.0812038406729698
22 0.0812038406729698
23 0.0798343643546104
24 0.0798343643546104
25 0.0794559493660927
26 0.0782238021492958
27 0.077346995472908
28 0.0762720108032227
29 0.0762720108032227
30 0.0759412795305252
31 0.0759412795305252
32 0.0753162354230881
33 0.0750547423958778
34 0.0750547423958778
35 0.0750547423958778
36 0.0750547423958778
37 0.0749170333147049
38 0.0749170333147049
39 0.0749170333147049
40 0.0749170333147049
41 0.0749170333147049
42 0.0749170333147049
43 0.0749170333147049
44 0.0749170333147049
45 0.0749170333147049
46 0.0749170333147049
47 0.0746883377432823
48 0.0741544738411903
49 0.0741544738411903
50 0.0741544738411903
51 0.0741544738411903
52 0.0741544738411903
53 0.0741544738411903
54 0.0741544738411903
55 0.0741544738411903
56 0.0741544738411903
57 0.0739414244890213
58 0.0739414244890213
59 0.0739414244890213
60 0.0739414244890213
61 0.0739414244890213
62 0.0739414244890213
63 0.0739414244890213
64 0.0739414244890213
65 0.0739414244890213
66 0.0739414244890213
67 0.0739414244890213
68 0.0739414244890213
69 0.0739414244890213
70 0.0739414244890213
71 0.0739414244890213
72 0.0739414244890213
73 0.0739414244890213
74 0.0739414244890213
75 0.0739414244890213
76 0.0739414244890213
77 0.0739414244890213
78 0.0739414244890213
79 0.0739414244890213
80 0.0739414244890213
81 0.0725418105721474
82 0.0725418105721474
83 0.0725418105721474
84 0.0725418105721474
85 0.071600154042244
86 0.071600154042244
87 0.071600154042244
88 0.071600154042244
89 0.071600154042244
90 0.071600154042244
91 0.071600154042244
92 0.071600154042244
93 0.071600154042244
94 0.071600154042244
95 0.071600154042244
96 0.071600154042244
97 0.071600154042244
98 0.0713281631469727
99 0.0713281631469727
100 0.0713281631469727
101 0.0713281631469727
102 0.0713281631469727
103 0.0713281631469727
104 0.0713281631469727
};
\end{axis}

\end{tikzpicture}

%% file: figures/bop_1d_gramacylee_strong.tex
\begin{tikzpicture}

\definecolor{color0}{rgb}{0,0,1}
\definecolor{color1}{rgb}{1,0.549019607843137,0}
\definecolor{color2}{rgb}{1,0.647058823529412,0}
\definecolor{color3}{rgb}{0.564705882352941,0.933333333333333,0.564705882352941}

\begin{axis}[axis on top,
enlarge x limits=false,
enlarge y limits=false,
height=\figureheight,
scale only axis,
tick align=outside,
tick pos=left,
tick pos=left,
width=\figurewidth,
xmin=3, xmax=75,
xtick style={color=black},
xtick={-10,0,10,25,50,75,100},
xticklabels={\ensuremath{-}10,0,10,25,50,75,90},
ymin=-0.03, ymax=0.7,
ytick style={color=black},
ytick={0.   , 0.7},
]
\node[anchor=north east] at (rel axis cs:1,1) {Gramacy Lee 1D (strong)};
\path [draw=color1, fill=color1, opacity=0.3]
(axis cs:3,0.693378502433763)
--(axis cs:3,0.483341099689005)
--(axis cs:4,0.398795574465956)
--(axis cs:5,0.320885975515929)
--(axis cs:6,0.274033091472781)
--(axis cs:7,0.270690132915736)
--(axis cs:8,0.257716957238609)
--(axis cs:9,0.244828918960579)
--(axis cs:10,0.244828918960579)
--(axis cs:11,0.229109989666479)
--(axis cs:12,0.218828340156898)
--(axis cs:13,0.185648152652743)
--(axis cs:14,0.137737432824625)
--(axis cs:15,0.136085899265937)
--(axis cs:16,0.114567531175806)
--(axis cs:17,0.114567531175806)
--(axis cs:18,0.113801646161203)
--(axis cs:19,0.113801646161203)
--(axis cs:20,0.113801646161203)
--(axis cs:21,0.113801646161203)
--(axis cs:22,0.113801646161203)
--(axis cs:23,0.11334991490107)
--(axis cs:24,0.110637576778646)
--(axis cs:25,0.0808796917567541)
--(axis cs:26,0.0567290483009618)
--(axis cs:27,0.0567290483009618)
--(axis cs:28,0.0567290483009618)
--(axis cs:29,0.0567290483009618)
--(axis cs:30,0.0567290483009618)
--(axis cs:31,0.0561187871033751)
--(axis cs:32,0.0561187871033751)
--(axis cs:33,0.0561187871033751)
--(axis cs:34,0.0558601811516231)
--(axis cs:35,0.0556249402241382)
--(axis cs:36,0.0556249402241382)
--(axis cs:37,0.0555657213346068)
--(axis cs:38,0.0555657213346068)
--(axis cs:39,0.0555657213346068)
--(axis cs:40,0.0540701226598937)
--(axis cs:41,0.0539865923673137)
--(axis cs:42,0.0390735523785183)
--(axis cs:43,0.0390735523785183)
--(axis cs:44,0.0390532267064841)
--(axis cs:45,0.0366618434547425)
--(axis cs:46,0.0366618434547425)
--(axis cs:47,0.0366609587073076)
--(axis cs:48,0.0366609587073076)
--(axis cs:49,0.0365500005048597)
--(axis cs:50,0.0365500005048597)
--(axis cs:51,0.0365208934934115)
--(axis cs:52,0.0364776900844839)
--(axis cs:53,0.0364772207626232)
--(axis cs:54,0.0364772207626232)
--(axis cs:55,0.0364772207626232)
--(axis cs:56,0.0364772207626232)
--(axis cs:57,0.0364772207626232)
--(axis cs:58,0.0280990744304467)
--(axis cs:59,0.0280990744304467)
--(axis cs:60,0.0280990744304467)
--(axis cs:61,0.0236118992698775)
--(axis cs:62,0.0235979636356581)
--(axis cs:63,0.0235979636356581)
--(axis cs:64,0.0235979636356581)
--(axis cs:65,0.0151687651630501)
--(axis cs:66,0.0149037763575374)
--(axis cs:67,0.0149037763575374)
--(axis cs:68,0.0149037763575374)
--(axis cs:69,0.0149008862592472)
--(axis cs:70,0.0146206443468619)
--(axis cs:71,0.0146206443468619)
--(axis cs:72,0.0146206443468619)
--(axis cs:73,0.0146124431622147)
--(axis cs:74,0.0146124431622147)
--(axis cs:75,0.0146124431622147)
--(axis cs:76,0.0146124431622147)
--(axis cs:77,0.0137853091411391)
--(axis cs:78,0.013784806586536)
--(axis cs:79,0.013784806586536)
--(axis cs:80,0.013784806586536)
--(axis cs:81,0.013784806586536)
--(axis cs:82,0.013784806586536)
--(axis cs:83,0.013784806586536)
--(axis cs:84,0.013784806586536)
--(axis cs:85,0.013784806586536)
--(axis cs:86,0.013784806586536)
--(axis cs:87,0.013784806586536)
--(axis cs:88,0.013784806586536)
--(axis cs:89,0.013784806586536)
--(axis cs:90,0.0137836235738508)
--(axis cs:91,0.0137836235738508)
--(axis cs:92,0.0137836235738508)
--(axis cs:93,0.0137650060215813)
--(axis cs:94,0.0137650060215813)
--(axis cs:95,0.0137650060215813)
--(axis cs:96,0.0137650060215813)
--(axis cs:97,0.0137650060215813)
--(axis cs:98,0.0137650060215813)
--(axis cs:99,0.0137650060215813)
--(axis cs:100,0.0137650060215813)
--(axis cs:101,0.0137650060215813)
--(axis cs:102,0.0137649430654365)
--(axis cs:103,0.0137649430654365)
--(axis cs:104,0.0137649430654365)
--(axis cs:104,0.0686129109183642)
--(axis cs:104,0.0686129109183642)
--(axis cs:103,0.0686129109183642)
--(axis cs:102,0.0686129109183642)
--(axis cs:101,0.0686129558679349)
--(axis cs:100,0.0686129558679349)
--(axis cs:99,0.0686129558679349)
--(axis cs:98,0.0686129558679349)
--(axis cs:97,0.0686129558679349)
--(axis cs:96,0.0686129558679349)
--(axis cs:95,0.0686129558679349)
--(axis cs:94,0.0686129558679349)
--(axis cs:93,0.0686129558679349)
--(axis cs:92,0.0686262562342289)
--(axis cs:91,0.0686262562342289)
--(axis cs:90,0.0686262562342289)
--(axis cs:89,0.0686271007618005)
--(axis cs:88,0.0686271007618005)
--(axis cs:87,0.0686271007618005)
--(axis cs:86,0.0686271007618005)
--(axis cs:85,0.0686271007618005)
--(axis cs:84,0.0686271007618005)
--(axis cs:83,0.0686271007618005)
--(axis cs:82,0.0686271007618005)
--(axis cs:81,0.0686271007618005)
--(axis cs:80,0.0686271007618005)
--(axis cs:79,0.0686271007618005)
--(axis cs:78,0.0686271007618005)
--(axis cs:77,0.0686274595131848)
--(axis cs:76,0.0692337149316177)
--(axis cs:75,0.0692337149316177)
--(axis cs:74,0.0692337149316177)
--(axis cs:73,0.0692337149316177)
--(axis cs:72,0.0692395296353558)
--(axis cs:71,0.0692395296353558)
--(axis cs:70,0.0692395296353558)
--(axis cs:69,0.0694522843001386)
--(axis cs:68,0.0694543272445538)
--(axis cs:67,0.0694543272445538)
--(axis cs:66,0.0694543272445538)
--(axis cs:65,0.0696590787553863)
--(axis cs:64,0.0777576715662958)
--(axis cs:63,0.0777576715662958)
--(axis cs:62,0.0777576715662958)
--(axis cs:61,0.0777668259974793)
--(axis cs:60,0.0815683792961252)
--(axis cs:59,0.0815683792961252)
--(axis cs:58,0.0815683792961252)
--(axis cs:57,0.0979958010791732)
--(axis cs:56,0.0979958010791732)
--(axis cs:55,0.0979958010791732)
--(axis cs:54,0.0979958010791732)
--(axis cs:53,0.0979958010791732)
--(axis cs:52,0.097996086981474)
--(axis cs:51,0.098022464860861)
--(axis cs:50,0.0980402655453999)
--(axis cs:49,0.0980402655453999)
--(axis cs:48,0.0981080134502459)
--(axis cs:47,0.0981080134502459)
--(axis cs:46,0.0981106702474758)
--(axis cs:45,0.0981106702474758)
--(axis cs:44,0.100394139155484)
--(axis cs:43,0.100406274125143)
--(axis cs:42,0.100406274125143)
--(axis cs:41,0.131437838114971)
--(axis cs:40,0.131486350364081)
--(axis cs:39,0.132381693928554)
--(axis cs:38,0.132381693928554)
--(axis cs:37,0.132381693928554)
--(axis cs:36,0.132415759940002)
--(axis cs:35,0.132415759940002)
--(axis cs:34,0.132558388878634)
--(axis cs:33,0.132718599035896)
--(axis cs:32,0.132718599035896)
--(axis cs:31,0.132718599035896)
--(axis cs:30,0.1330957589569)
--(axis cs:29,0.1330957589569)
--(axis cs:28,0.1330957589569)
--(axis cs:27,0.1330957589569)
--(axis cs:26,0.1330957589569)
--(axis cs:25,0.154517436049626)
--(axis cs:24,0.188480072316221)
--(axis cs:23,0.189655339430476)
--(axis cs:22,0.18986299437697)
--(axis cs:21,0.18986299437697)
--(axis cs:20,0.18986299437697)
--(axis cs:19,0.18986299437697)
--(axis cs:18,0.18986299437697)
--(axis cs:17,0.190163121326064)
--(axis cs:16,0.190163121326064)
--(axis cs:15,0.233168293724311)
--(axis cs:14,0.234715866501472)
--(axis cs:13,0.272921130220768)
--(axis cs:12,0.300922904721621)
--(axis cs:11,0.318699398179914)
--(axis cs:10,0.377543592686784)
--(axis cs:9,0.377543592686784)
--(axis cs:8,0.395387270930375)
--(axis cs:7,0.435939360193756)
--(axis cs:6,0.465904979953256)
--(axis cs:5,0.517083489805542)
--(axis cs:4,0.642446857363406)
--(axis cs:3,0.693378502433763)
--cycle;

\path [draw=blue, fill=blue, opacity=0.3]
(axis cs:3,0.693378567695618)
--(axis cs:3,0.483341127634048)
--(axis cs:4,0.390137106180191)
--(axis cs:5,0.377283990383148)
--(axis cs:6,0.361825942993164)
--(axis cs:7,0.316047966480255)
--(axis cs:8,0.316047966480255)
--(axis cs:9,0.288683176040649)
--(axis cs:10,0.268315643072128)
--(axis cs:11,0.25261253118515)
--(axis cs:12,0.227877467870712)
--(axis cs:13,0.224256485700607)
--(axis cs:14,0.203465461730957)
--(axis cs:15,0.201505035161972)
--(axis cs:16,0.201505035161972)
--(axis cs:17,0.200482070446014)
--(axis cs:18,0.200482070446014)
--(axis cs:19,0.182180136442184)
--(axis cs:20,0.160126447677612)
--(axis cs:21,0.160126447677612)
--(axis cs:22,0.148839741945267)
--(axis cs:23,0.134798496961594)
--(axis cs:24,0.132815763354301)
--(axis cs:25,0.132815763354301)
--(axis cs:26,0.130036905407906)
--(axis cs:27,0.120688408613205)
--(axis cs:28,0.119942456483841)
--(axis cs:29,0.110094450414181)
--(axis cs:30,0.110094450414181)
--(axis cs:31,0.109829477965832)
--(axis cs:32,0.109829477965832)
--(axis cs:33,0.109650872647762)
--(axis cs:34,0.0782420411705971)
--(axis cs:35,0.0772195160388947)
--(axis cs:36,0.0765645951032639)
--(axis cs:37,0.0492665879428387)
--(axis cs:38,0.0492665879428387)
--(axis cs:39,0.0492665879428387)
--(axis cs:40,0.0492665879428387)
--(axis cs:41,0.0326010808348656)
--(axis cs:42,0.0322547778487206)
--(axis cs:43,0.028442332521081)
--(axis cs:44,0.0210039857774973)
--(axis cs:45,0.0141945797950029)
--(axis cs:46,0.0139826368540525)
--(axis cs:47,0.0137676876038313)
--(axis cs:48,0.0137676876038313)
--(axis cs:49,0.0137676876038313)
--(axis cs:50,0.0137676876038313)
--(axis cs:51,0.0137676876038313)
--(axis cs:52,0.0137676876038313)
--(axis cs:53,0.0137676876038313)
--(axis cs:54,0.0137676876038313)
--(axis cs:55,0.013766823336482)
--(axis cs:56,0.013766823336482)
--(axis cs:57,0.013766823336482)
--(axis cs:58,0.013766823336482)
--(axis cs:59,0.013766823336482)
--(axis cs:60,0.013766823336482)
--(axis cs:61,0.013766823336482)
--(axis cs:62,0.013766823336482)
--(axis cs:63,0.013766823336482)
--(axis cs:64,0.0137590114027262)
--(axis cs:65,0.0137590114027262)
--(axis cs:66,0.0137590114027262)
--(axis cs:67,0.0137586817145348)
--(axis cs:68,0.0137586817145348)
--(axis cs:69,0.0137586817145348)
--(axis cs:70,0.0137586817145348)
--(axis cs:71,0.0137586817145348)
--(axis cs:72,0.0137586817145348)
--(axis cs:73,0.0137586817145348)
--(axis cs:74,0.0137586817145348)
--(axis cs:75,0.0137586817145348)
--(axis cs:76,0.0137586817145348)
--(axis cs:77,0.0137367490679026)
--(axis cs:78,1.63242220878601e-05)
--(axis cs:79,1.63242220878601e-05)
--(axis cs:80,1.63242220878601e-05)
--(axis cs:81,1.63223594427109e-05)
--(axis cs:82,1.63037329912186e-05)
--(axis cs:83,1.63037329912186e-05)
--(axis cs:84,1.63037329912186e-05)
--(axis cs:85,1.63037329912186e-05)
--(axis cs:86,1.63037329912186e-05)
--(axis cs:87,1.62962824106216e-05)
--(axis cs:88,1.62962824106216e-05)
--(axis cs:89,1.62962824106216e-05)
--(axis cs:90,1.57369431690313e-05)
--(axis cs:91,1.57369431690313e-05)
--(axis cs:92,1.57369431690313e-05)
--(axis cs:93,1.57369431690313e-05)
--(axis cs:94,1.57369431690313e-05)
--(axis cs:95,1.16547598736361e-05)
--(axis cs:96,1.16547598736361e-05)
--(axis cs:97,1.16480296128429e-05)
--(axis cs:98,7.15771602699533e-06)
--(axis cs:99,7.15771602699533e-06)
--(axis cs:100,7.15771602699533e-06)
--(axis cs:101,7.15771602699533e-06)
--(axis cs:102,7.15771602699533e-06)
--(axis cs:103,7.15771602699533e-06)
--(axis cs:104,7.15771602699533e-06)
--(axis cs:104,0.00013532122829929)
--(axis cs:104,0.00013532122829929)
--(axis cs:103,0.00013532122829929)
--(axis cs:102,0.00013532122829929)
--(axis cs:101,0.00013532122829929)
--(axis cs:100,0.00013532122829929)
--(axis cs:99,0.00013532122829929)
--(axis cs:98,0.00013532122829929)
--(axis cs:97,0.000145398284075782)
--(axis cs:96,0.000145403479109518)
--(axis cs:95,0.000145403479109518)
--(axis cs:94,0.000149296392919496)
--(axis cs:93,0.000149296392919496)
--(axis cs:92,0.000149296392919496)
--(axis cs:91,0.000149296392919496)
--(axis cs:90,0.000149296392919496)
--(axis cs:89,0.00844432041049004)
--(axis cs:88,0.00844432041049004)
--(axis cs:87,0.00844432041049004)
--(axis cs:86,0.0358932167291641)
--(axis cs:85,0.0358932167291641)
--(axis cs:84,0.0358932167291641)
--(axis cs:83,0.0358932167291641)
--(axis cs:82,0.0358932167291641)
--(axis cs:81,0.0358932316303253)
--(axis cs:80,0.0412239618599415)
--(axis cs:79,0.0412239618599415)
--(axis cs:78,0.0412239618599415)
--(axis cs:77,0.068653404712677)
--(axis cs:76,0.0687631592154503)
--(axis cs:75,0.0687631592154503)
--(axis cs:74,0.0687631592154503)
--(axis cs:73,0.0687631592154503)
--(axis cs:72,0.0687631592154503)
--(axis cs:71,0.0687631592154503)
--(axis cs:70,0.0687631592154503)
--(axis cs:69,0.0687631592154503)
--(axis cs:68,0.0687631592154503)
--(axis cs:67,0.0687631592154503)
--(axis cs:66,0.0687633976340294)
--(axis cs:65,0.0687633976340294)
--(axis cs:64,0.0687633976340294)
--(axis cs:63,0.0688026174902916)
--(axis cs:62,0.0688026174902916)
--(axis cs:61,0.0688026174902916)
--(axis cs:60,0.0688026174902916)
--(axis cs:59,0.0688026174902916)
--(axis cs:58,0.0688026174902916)
--(axis cs:57,0.0688026174902916)
--(axis cs:56,0.0688026174902916)
--(axis cs:55,0.0688026174902916)
--(axis cs:54,0.0688032358884811)
--(axis cs:53,0.0688032358884811)
--(axis cs:52,0.0688032358884811)
--(axis cs:51,0.0688032358884811)
--(axis cs:50,0.0688032358884811)
--(axis cs:49,0.0688032358884811)
--(axis cs:48,0.0688032358884811)
--(axis cs:47,0.0688032358884811)
--(axis cs:46,0.0689578056335449)
--(axis cs:45,0.0691099017858505)
--(axis cs:44,0.075366735458374)
--(axis cs:43,0.0853531658649445)
--(axis cs:42,0.0947774574160576)
--(axis cs:41,0.0950193330645561)
--(axis cs:40,0.11658838391304)
--(axis cs:39,0.11658838391304)
--(axis cs:38,0.11658838391304)
--(axis cs:37,0.11658838391304)
--(axis cs:36,0.159364178776741)
--(axis cs:35,0.16231095790863)
--(axis cs:34,0.163983374834061)
--(axis cs:33,0.201382905244827)
--(axis cs:32,0.201945871114731)
--(axis cs:31,0.201945871114731)
--(axis cs:30,0.202288001775742)
--(axis cs:29,0.202288001775742)
--(axis cs:28,0.218403697013855)
--(axis cs:27,0.220127791166306)
--(axis cs:26,0.22499044239521)
--(axis cs:25,0.226746216416359)
--(axis cs:24,0.226746216416359)
--(axis cs:23,0.227579206228256)
--(axis cs:22,0.238730549812317)
--(axis cs:21,0.244454711675644)
--(axis cs:20,0.244454711675644)
--(axis cs:19,0.263141989707947)
--(axis cs:18,0.292627155780792)
--(axis cs:17,0.292627155780792)
--(axis cs:16,0.294079542160034)
--(axis cs:15,0.294079542160034)
--(axis cs:14,0.297551989555359)
--(axis cs:13,0.357099205255508)
--(axis cs:12,0.360113948583603)
--(axis cs:11,0.408977806568146)
--(axis cs:10,0.431523531675339)
--(axis cs:9,0.440265357494354)
--(axis cs:8,0.464650213718414)
--(axis cs:7,0.464650213718414)
--(axis cs:6,0.517099022865295)
--(axis cs:5,0.552688837051392)
--(axis cs:4,0.587581872940063)
--(axis cs:3,0.693378567695618)
--cycle;

\path [draw=blue, fill=blue, opacity=0.3]
(axis cs:3,0.693378448486328)
--(axis cs:3,0.483341068029404)
--(axis cs:4,0.374360859394073)
--(axis cs:5,0.294243693351746)
--(axis cs:6,0.216424688696861)
--(axis cs:7,0.193649247288704)
--(axis cs:8,0.174874678254128)
--(axis cs:9,0.173277392983437)
--(axis cs:10,0.144913017749786)
--(axis cs:11,0.140645802021027)
--(axis cs:12,0.115284457802773)
--(axis cs:13,0.114326819777489)
--(axis cs:14,0.114292711019516)
--(axis cs:15,0.108178690075874)
--(axis cs:16,0.0934213697910309)
--(axis cs:17,0.0695681571960449)
--(axis cs:18,0.0693304687738419)
--(axis cs:19,0.0670902580022812)
--(axis cs:20,0.0553748607635498)
--(axis cs:21,0.0520738698542118)
--(axis cs:22,0.0520738698542118)
--(axis cs:23,0.0520256049931049)
--(axis cs:24,0.044444739818573)
--(axis cs:25,0.044444739818573)
--(axis cs:26,0.044444739818573)
--(axis cs:27,0.0439899042248726)
--(axis cs:28,0.038065429776907)
--(axis cs:29,0.0377819836139679)
--(axis cs:30,0.0377819836139679)
--(axis cs:31,0.0377819836139679)
--(axis cs:32,0.0191503129899502)
--(axis cs:33,0.0189749263226986)
--(axis cs:34,0.0189749263226986)
--(axis cs:35,0.0189158767461777)
--(axis cs:36,0.0189158767461777)
--(axis cs:37,0.0189158767461777)
--(axis cs:38,0.0189144387841225)
--(axis cs:39,0.000861598178744316)
--(axis cs:40,0.000861598178744316)
--(axis cs:41,0.000860974192619324)
--(axis cs:42,0.000860974192619324)
--(axis cs:43,0.00084350211545825)
--(axis cs:44,0.00084350211545825)
--(axis cs:45,0.000757695641368628)
--(axis cs:46,0.000518966931849718)
--(axis cs:47,0.000410987180657685)
--(axis cs:48,0.000410987180657685)
--(axis cs:49,0.000410987180657685)
--(axis cs:50,0.000207011922611855)
--(axis cs:51,0.000207011922611855)
--(axis cs:52,0.000207011922611855)
--(axis cs:53,0.000207011922611855)
--(axis cs:54,0.000207011922611855)
--(axis cs:55,0.000207011922611855)
--(axis cs:56,0.000207011922611855)
--(axis cs:57,0.000207011922611855)
--(axis cs:58,0.000207011922611855)
--(axis cs:59,0.000206953205633909)
--(axis cs:60,0.000206953205633909)
--(axis cs:61,0.000206953205633909)
--(axis cs:62,0.000206953205633909)
--(axis cs:63,0.000206953205633909)
--(axis cs:64,0.000206953205633909)
--(axis cs:65,0.000206953205633909)
--(axis cs:66,0.000206953205633909)
--(axis cs:67,0.000206953205633909)
--(axis cs:68,0.000206953205633909)
--(axis cs:69,0.000206953205633909)
--(axis cs:70,0.000206953205633909)
--(axis cs:71,0.000172897707670927)
--(axis cs:72,0.000172897707670927)
--(axis cs:73,0.000172897707670927)
--(axis cs:74,0.000171721650986001)
--(axis cs:75,0.000145702535519376)
--(axis cs:76,0.000145645492011681)
--(axis cs:77,0.000145645492011681)
--(axis cs:78,0.000145645492011681)
--(axis cs:79,0.000145645492011681)
--(axis cs:80,0.000145645492011681)
--(axis cs:81,0.000145645492011681)
--(axis cs:82,0.000145645492011681)
--(axis cs:83,0.000145645492011681)
--(axis cs:84,0.000145645492011681)
--(axis cs:85,0.000145645492011681)
--(axis cs:86,0.000145645492011681)
--(axis cs:87,0.000145645492011681)
--(axis cs:88,0.000145645492011681)
--(axis cs:89,0.000145645492011681)
--(axis cs:90,0.000145645492011681)
--(axis cs:91,0.000145645492011681)
--(axis cs:92,0.000145645492011681)
--(axis cs:93,0.000145645492011681)
--(axis cs:94,0.000145645492011681)
--(axis cs:95,0.000145645492011681)
--(axis cs:96,0.00014236597053241)
--(axis cs:97,0.00014236597053241)
--(axis cs:98,0.00014236597053241)
--(axis cs:99,0.000138213523314334)
--(axis cs:100,0.000138213523314334)
--(axis cs:101,0.000138213523314334)
--(axis cs:102,0.000138213523314334)
--(axis cs:103,0.000138213523314334)
--(axis cs:104,0.000138213523314334)
--(axis cs:104,0.000503442424815148)
--(axis cs:104,0.000503442424815148)
--(axis cs:103,0.000503442424815148)
--(axis cs:102,0.000503442424815148)
--(axis cs:101,0.000503442424815148)
--(axis cs:100,0.000503442424815148)
--(axis cs:99,0.000503442424815148)
--(axis cs:98,0.000522166199516505)
--(axis cs:97,0.000522166199516505)
--(axis cs:96,0.000522166199516505)
--(axis cs:95,0.000524441828019917)
--(axis cs:94,0.000524441828019917)
--(axis cs:93,0.000524441828019917)
--(axis cs:92,0.000524441828019917)
--(axis cs:91,0.000524441828019917)
--(axis cs:90,0.000524441828019917)
--(axis cs:89,0.000524441828019917)
--(axis cs:88,0.000524441828019917)
--(axis cs:87,0.000524441828019917)
--(axis cs:86,0.000524441828019917)
--(axis cs:85,0.000524441828019917)
--(axis cs:84,0.000524441828019917)
--(axis cs:83,0.000524441828019917)
--(axis cs:82,0.000524441828019917)
--(axis cs:81,0.000524441828019917)
--(axis cs:80,0.000524441828019917)
--(axis cs:79,0.000524441828019917)
--(axis cs:78,0.000524441828019917)
--(axis cs:77,0.000524441828019917)
--(axis cs:76,0.000524441828019917)
--(axis cs:75,0.000524480128660798)
--(axis cs:74,0.000546323601156473)
--(axis cs:73,0.000547090661711991)
--(axis cs:72,0.000547090661711991)
--(axis cs:71,0.000547090661711991)
--(axis cs:70,0.000586527690757066)
--(axis cs:69,0.000586527690757066)
--(axis cs:68,0.000586527690757066)
--(axis cs:67,0.000586527690757066)
--(axis cs:66,0.000586527690757066)
--(axis cs:65,0.000586527690757066)
--(axis cs:64,0.000586527690757066)
--(axis cs:63,0.000586527690757066)
--(axis cs:62,0.000586527690757066)
--(axis cs:61,0.000586527690757066)
--(axis cs:60,0.000586527690757066)
--(axis cs:59,0.000586527690757066)
--(axis cs:58,0.000586564303375781)
--(axis cs:57,0.000586564303375781)
--(axis cs:56,0.000586564303375781)
--(axis cs:55,0.000586564303375781)
--(axis cs:54,0.000586564303375781)
--(axis cs:53,0.000586564303375781)
--(axis cs:52,0.000586564303375781)
--(axis cs:51,0.000586564303375781)
--(axis cs:50,0.000586564303375781)
--(axis cs:49,0.00141335604712367)
--(axis cs:48,0.00141335604712367)
--(axis cs:47,0.00141335604712367)
--(axis cs:46,0.00159515021368861)
--(axis cs:45,0.00347087252885103)
--(axis cs:44,0.011352775618434)
--(axis cs:43,0.011352775618434)
--(axis cs:42,0.0419404059648514)
--(axis cs:41,0.0419404059648514)
--(axis cs:40,0.0419409051537514)
--(axis cs:39,0.0419409051537514)
--(axis cs:38,0.110675893723965)
--(axis cs:37,0.110676944255829)
--(axis cs:36,0.110676944255829)
--(axis cs:35,0.110676944255829)
--(axis cs:34,0.110720336437225)
--(axis cs:33,0.110720336437225)
--(axis cs:32,0.110848560929298)
--(axis cs:31,0.132386803627014)
--(axis cs:30,0.132386803627014)
--(axis cs:29,0.132386803627014)
--(axis cs:28,0.132578387856483)
--(axis cs:27,0.137076586484909)
--(axis cs:26,0.137466087937355)
--(axis cs:25,0.137466087937355)
--(axis cs:24,0.137466087937355)
--(axis cs:23,0.145547270774841)
--(axis cs:22,0.145604893565178)
--(axis cs:21,0.145604893565178)
--(axis cs:20,0.147945165634155)
--(axis cs:19,0.162059038877487)
--(axis cs:18,0.163494929671288)
--(axis cs:17,0.16524863243103)
--(axis cs:16,0.19334414601326)
--(axis cs:15,0.202385410666466)
--(axis cs:14,0.21050950884819)
--(axis cs:13,0.210674956440926)
--(axis cs:12,0.213808491826057)
--(axis cs:11,0.242673516273499)
--(axis cs:10,0.252018362283707)
--(axis cs:9,0.281720042228699)
--(axis cs:8,0.282313615083694)
--(axis cs:7,0.30894410610199)
--(axis cs:6,0.335179448127747)
--(axis cs:5,0.47712367773056)
--(axis cs:4,0.554347693920135)
--(axis cs:3,0.693378448486328)
--cycle;

\path [draw=color1, fill=color1, opacity=0.3]
(axis cs:3,0.693378502433763)
--(axis cs:3,0.483341099689005)
--(axis cs:4,0.433446434515946)
--(axis cs:5,0.398350991775583)
--(axis cs:6,0.372804780567102)
--(axis cs:7,0.347004117793766)
--(axis cs:8,0.347004117793766)
--(axis cs:9,0.340805651311389)
--(axis cs:10,0.340805651311389)
--(axis cs:11,0.333630970815473)
--(axis cs:12,0.320844371120055)
--(axis cs:13,0.320844371120055)
--(axis cs:14,0.312304144367173)
--(axis cs:15,0.293831994995028)
--(axis cs:16,0.293831994995028)
--(axis cs:17,0.284442770871315)
--(axis cs:18,0.283831970012075)
--(axis cs:19,0.283831970012075)
--(axis cs:20,0.262103911545066)
--(axis cs:21,0.262103911545066)
--(axis cs:22,0.260401069903143)
--(axis cs:23,0.258190022313113)
--(axis cs:24,0.251582370767827)
--(axis cs:25,0.202128523362325)
--(axis cs:26,0.180484971768554)
--(axis cs:27,0.18039668270183)
--(axis cs:28,0.180128377876276)
--(axis cs:29,0.179974830819188)
--(axis cs:30,0.17928250314637)
--(axis cs:31,0.17928250314637)
--(axis cs:32,0.179260064554242)
--(axis cs:33,0.179260064554242)
--(axis cs:34,0.173358562318383)
--(axis cs:35,0.153827062751881)
--(axis cs:36,0.140071324426855)
--(axis cs:37,0.136364334373119)
--(axis cs:38,0.136297970281644)
--(axis cs:39,0.135530512836559)
--(axis cs:40,0.135530512836559)
--(axis cs:41,0.135411042838148)
--(axis cs:42,0.135411042838148)
--(axis cs:43,0.135411042838148)
--(axis cs:44,0.135401300795558)
--(axis cs:45,0.119185530690732)
--(axis cs:46,0.0917208622683713)
--(axis cs:47,0.0917078902258459)
--(axis cs:48,0.0916861318955108)
--(axis cs:49,0.0916861318955108)
--(axis cs:50,0.073749578197281)
--(axis cs:51,0.073749578197281)
--(axis cs:52,0.073749578197281)
--(axis cs:53,0.0595261899190988)
--(axis cs:54,0.057224055764595)
--(axis cs:55,0.057151798880415)
--(axis cs:56,0.057151798880415)
--(axis cs:57,0.0571408658924272)
--(axis cs:58,0.0571310512204214)
--(axis cs:59,0.0571310395398827)
--(axis cs:60,0.0567894735711325)
--(axis cs:61,0.0567894735711325)
--(axis cs:62,0.0567890676479951)
--(axis cs:63,0.038586442761533)
--(axis cs:64,0.038586442761533)
--(axis cs:65,0.038586442761533)
--(axis cs:66,0.038586442761533)
--(axis cs:67,0.038586442761533)
--(axis cs:68,0.0379369059896532)
--(axis cs:69,0.0379345446546924)
--(axis cs:70,0.0379345446546924)
--(axis cs:71,0.0376634787033189)
--(axis cs:72,0.0376634787033189)
--(axis cs:73,0.0376634787033189)
--(axis cs:74,0.0376634787033189)
--(axis cs:75,0.0376616146011297)
--(axis cs:76,0.0376470868792573)
--(axis cs:77,0.0376469966954035)
--(axis cs:78,0.0376316367194748)
--(axis cs:79,0.0376189891335463)
--(axis cs:80,0.0376189891335463)
--(axis cs:81,0.0376189891335463)
--(axis cs:82,0.0376189891335463)
--(axis cs:83,0.0376189891335463)
--(axis cs:84,0.0376189891335463)
--(axis cs:85,0.037617933178209)
--(axis cs:86,0.037617933178209)
--(axis cs:87,0.037617933178209)
--(axis cs:88,0.0376115005531532)
--(axis cs:89,0.0376115005531532)
--(axis cs:90,0.0376115005531532)
--(axis cs:91,0.0376115005531532)
--(axis cs:92,0.0376115005531532)
--(axis cs:93,0.0376115005531532)
--(axis cs:94,0.0376115005531532)
--(axis cs:95,0.0376104490050482)
--(axis cs:96,0.0376104490050482)
--(axis cs:97,0.0376104490050482)
--(axis cs:98,0.037610127594013)
--(axis cs:99,0.037610127594013)
--(axis cs:100,0.037610127594013)
--(axis cs:101,0.0376099819607969)
--(axis cs:102,0.0376099819607969)
--(axis cs:103,0.0376099819607969)
--(axis cs:104,0.0376099819607969)
--(axis cs:104,0.0989497615846159)
--(axis cs:104,0.0989497615846159)
--(axis cs:103,0.0989497615846159)
--(axis cs:102,0.0989497615846159)
--(axis cs:101,0.0989497615846159)
--(axis cs:100,0.0989498494577361)
--(axis cs:99,0.0989498494577361)
--(axis cs:98,0.0989498494577361)
--(axis cs:97,0.0989502809911544)
--(axis cs:96,0.0989502809911544)
--(axis cs:95,0.0989502809911544)
--(axis cs:94,0.09895091550329)
--(axis cs:93,0.09895091550329)
--(axis cs:92,0.09895091550329)
--(axis cs:91,0.09895091550329)
--(axis cs:90,0.09895091550329)
--(axis cs:89,0.09895091550329)
--(axis cs:88,0.09895091550329)
--(axis cs:87,0.0989547975365491)
--(axis cs:86,0.0989547975365491)
--(axis cs:85,0.0989547975365491)
--(axis cs:84,0.0989554347038216)
--(axis cs:83,0.0989554347038216)
--(axis cs:82,0.0989554347038216)
--(axis cs:81,0.0989554347038216)
--(axis cs:80,0.0989554347038216)
--(axis cs:79,0.0989554347038216)
--(axis cs:78,0.0989630711432692)
--(axis cs:77,0.0989724689639878)
--(axis cs:76,0.0989725233612935)
--(axis cs:75,0.0989812989661245)
--(axis cs:74,0.098982423260274)
--(axis cs:73,0.098982423260274)
--(axis cs:72,0.098982423260274)
--(axis cs:71,0.098982423260274)
--(axis cs:70,0.0991473924020206)
--(axis cs:69,0.0991473924020206)
--(axis cs:68,0.0991488145684669)
--(axis cs:67,0.0995468035311989)
--(axis cs:66,0.0995468035311989)
--(axis cs:65,0.0995468035311989)
--(axis cs:64,0.0995468035311989)
--(axis cs:63,0.0995468035311989)
--(axis cs:62,0.121565170309551)
--(axis cs:61,0.121565386056809)
--(axis cs:60,0.121565386056809)
--(axis cs:59,0.121751806577487)
--(axis cs:58,0.121751833825953)
--(axis cs:57,0.121763139076058)
--(axis cs:56,0.12176893271104)
--(axis cs:55,0.12176893271104)
--(axis cs:54,0.121822868672253)
--(axis cs:53,0.127925775949066)
--(axis cs:52,0.147057763357672)
--(axis cs:51,0.147057763357672)
--(axis cs:50,0.147057763357672)
--(axis cs:49,0.158388262002814)
--(axis cs:48,0.158388262002814)
--(axis cs:47,0.158397365994147)
--(axis cs:46,0.158409263770525)
--(axis cs:45,0.203068706978477)
--(axis cs:44,0.224608002430166)
--(axis cs:43,0.224611716741218)
--(axis cs:42,0.224611716741218)
--(axis cs:41,0.224611716741218)
--(axis cs:40,0.22465727498438)
--(axis cs:39,0.22465727498438)
--(axis cs:38,0.224951542157893)
--(axis cs:37,0.225026967014881)
--(axis cs:36,0.228822733139835)
--(axis cs:35,0.235616546218405)
--(axis cs:34,0.251411792313299)
--(axis cs:33,0.261091457160726)
--(axis cs:32,0.261091457160726)
--(axis cs:31,0.261112219897935)
--(axis cs:30,0.261112219897935)
--(axis cs:29,0.263253648364006)
--(axis cs:28,0.263450112361474)
--(axis cs:27,0.263939722808647)
--(axis cs:26,0.264020913313615)
--(axis cs:25,0.301888927583389)
--(axis cs:24,0.350338481651283)
--(axis cs:23,0.355286187494037)
--(axis cs:22,0.356961931333063)
--(axis cs:21,0.357378020053905)
--(axis cs:20,0.357378020053905)
--(axis cs:19,0.388233176273449)
--(axis cs:18,0.388233176273449)
--(axis cs:17,0.389187933329817)
--(axis cs:16,0.408582914046417)
--(axis cs:15,0.408582914046417)
--(axis cs:14,0.422217631099024)
--(axis cs:13,0.442255473323286)
--(axis cs:12,0.442255473323286)
--(axis cs:11,0.476403068614967)
--(axis cs:10,0.478228856647842)
--(axis cs:9,0.478228856647842)
--(axis cs:8,0.485247601394396)
--(axis cs:7,0.485247601394396)
--(axis cs:6,0.504385045878206)
--(axis cs:5,0.567894715963743)
--(axis cs:4,0.622790900094017)
--(axis cs:3,0.693378502433763)
--cycle;

\addplot [semithick, color1, dash dot]
table {%
3 0.588359801061384
4 0.520621215914681
5 0.418984732660735
6 0.369969035713018
7 0.353314746554746
8 0.326552114084492
9 0.311186255823681
10 0.311186255823681
11 0.273904693923197
12 0.25987562243926
13 0.229284641436756
14 0.186226649663049
15 0.184627096495124
16 0.152365326250935
17 0.152365326250935
18 0.151832320269086
19 0.151832320269086
20 0.151832320269086
21 0.151832320269086
22 0.151832320269086
23 0.151502627165773
24 0.149558824547433
25 0.11769856390319
26 0.0949124036289308
27 0.0949124036289308
28 0.0949124036289308
29 0.0949124036289308
30 0.0949124036289308
31 0.0944186930696357
32 0.0944186930696357
33 0.0944186930696357
34 0.0942092850151288
35 0.09402035008207
36 0.09402035008207
37 0.0939737076315802
38 0.0939737076315802
39 0.0939737076315802
40 0.0927782365119872
41 0.0927122152411423
42 0.0697399132518306
43 0.0697399132518306
44 0.0697236829309843
45 0.0673862568511091
46 0.0673862568511091
47 0.0673844860787767
48 0.0673844860787767
49 0.0672951330251298
50 0.0672951330251298
51 0.0672716791771363
52 0.0672368885329789
53 0.0672365109208982
54 0.0672365109208982
55 0.0672365109208982
56 0.0672365109208982
57 0.0672365109208982
58 0.0548337268632859
59 0.0548337268632859
60 0.0548337268632859
61 0.0506893626336784
62 0.050677817600977
63 0.050677817600977
64 0.050677817600977
65 0.0424139219592182
66 0.0421790518010456
67 0.0421790518010456
68 0.0421790518010456
69 0.0421765852796929
70 0.0419300869911088
71 0.0419300869911088
72 0.0419300869911088
73 0.0419230790469162
74 0.0419230790469162
75 0.0419230790469162
76 0.0419230790469162
77 0.0412063843271619
78 0.0412059536741682
79 0.0412059536741682
80 0.0412059536741682
81 0.0412059536741682
82 0.0412059536741682
83 0.0412059536741682
84 0.0412059536741682
85 0.0412059536741682
86 0.0412059536741682
87 0.0412059536741682
88 0.0412059536741682
89 0.0412059536741682
90 0.0412049399040398
91 0.0412049399040398
92 0.0412049399040398
93 0.0411889809447581
94 0.0411889809447581
95 0.0411889809447581
96 0.0411889809447581
97 0.0411889809447581
98 0.0411889809447581
99 0.0411889809447581
100 0.0411889809447581
101 0.0411889809447581
102 0.0411889269919003
103 0.0411889269919003
104 0.0411889269919003
};
\addplot [semithick, blue]
table {%
3 0.588359832763672
4 0.488859504461288
5 0.46498641371727
6 0.43946248292923
7 0.390349090099335
8 0.390349090099335
9 0.364474266767502
10 0.349919587373734
11 0.330795168876648
12 0.293995708227158
13 0.290677845478058
14 0.250508725643158
15 0.247792288661003
16 0.247792288661003
17 0.246554613113403
18 0.246554613113403
19 0.222661063075066
20 0.202290579676628
21 0.202290579676628
22 0.193785145878792
23 0.181188851594925
24 0.17978098988533
25 0.17978098988533
26 0.177513673901558
27 0.170408099889755
28 0.169173076748848
29 0.156191229820251
30 0.156191229820251
31 0.155887678265572
32 0.155887678265572
33 0.155516892671585
34 0.121112704277039
35 0.119765236973763
36 0.117964386940002
37 0.0829274877905846
38 0.0829274877905846
39 0.0829274877905846
40 0.0829274877905846
41 0.0638102069497108
42 0.0635161176323891
43 0.0568977482616901
44 0.0481853596866131
45 0.0416522398591042
46 0.0414702221751213
47 0.0412854626774788
48 0.0412854626774788
49 0.0412854626774788
50 0.0412854626774788
51 0.0412854626774788
52 0.0412854626774788
53 0.0412854626774788
54 0.0412854626774788
55 0.0412847213447094
56 0.0412847213447094
57 0.0412847213447094
58 0.0412847213447094
59 0.0412847213447094
60 0.0412847213447094
61 0.0412847213447094
62 0.0412847213447094
63 0.0412847213447094
64 0.0412612035870552
65 0.0412612035870552
66 0.0412612035870552
67 0.0412609204649925
68 0.0412609204649925
69 0.0412609204649925
70 0.0412609204649925
71 0.0412609204649925
72 0.0412609204649925
73 0.0412609204649925
74 0.0412609204649925
75 0.0412609204649925
76 0.0412609204649925
77 0.0411950759589672
78 0.0206201430410147
79 0.0206201430410147
80 0.0206201430410147
81 0.0179547779262066
82 0.0179547611624002
83 0.0179547611624002
84 0.0179547611624002
85 0.0179547611624002
86 0.0179547611624002
87 0.00423030834645033
88 0.00423030834645033
89 0.00423030834645033
90 8.25166716822423e-05
91 8.25166716822423e-05
92 8.25166716822423e-05
93 8.25166716822423e-05
94 8.25166716822423e-05
95 7.85291194915771e-05
96 7.85291194915771e-05
97 7.85231604822911e-05
98 7.12394685251638e-05
99 7.12394685251638e-05
100 7.12394685251638e-05
101 7.12394685251638e-05
102 7.12394685251638e-05
103 7.12394685251638e-05
104 7.12394685251638e-05
};
\addplot [semithick, blue, dash dot]
table {%
3 0.588359773159027
4 0.464354276657104
5 0.385683685541153
6 0.275802075862885
7 0.251296669244766
8 0.228594154119492
9 0.227498725056648
10 0.198465690016747
11 0.191659659147263
12 0.164546474814415
13 0.162500888109207
14 0.162401109933853
15 0.15528205037117
16 0.143382757902145
17 0.117408394813538
18 0.116412699222565
19 0.114574648439884
20 0.101660013198853
21 0.0988393798470497
22 0.0988393798470497
23 0.098786436021328
24 0.090955413877964
25 0.090955413877964
26 0.090955413877964
27 0.0905332490801811
28 0.0853219106793404
29 0.085084393620491
30 0.085084393620491
31 0.085084393620491
32 0.0649994388222694
33 0.0648476332426071
34 0.0648476332426071
35 0.0647964105010033
36 0.0647964105010033
37 0.0647964105010033
38 0.0647951662540436
39 0.0214012507349253
40 0.0214012507349253
41 0.0214006900787354
42 0.0214006900787354
43 0.00609813909977674
44 0.00609813909977674
45 0.00211428408510983
46 0.00105705857276917
47 0.000912171613890678
48 0.000912171613890678
49 0.000912171613890678
50 0.000396788120269775
51 0.000396788120269775
52 0.000396788120269775
53 0.000396788120269775
54 0.000396788120269775
55 0.000396788120269775
56 0.000396788120269775
57 0.000396788120269775
58 0.000396788120269775
59 0.000396740448195487
60 0.000396740448195487
61 0.000396740448195487
62 0.000396740448195487
63 0.000396740448195487
64 0.000396740448195487
65 0.000396740448195487
66 0.000396740448195487
67 0.000396740448195487
68 0.000396740448195487
69 0.000396740448195487
70 0.000396740448195487
71 0.000359994184691459
72 0.000359994184691459
73 0.000359994184691459
74 0.000359022611519322
75 0.000335091346642002
76 0.000335043674567714
77 0.000335043674567714
78 0.000335043674567714
79 0.000335043674567714
80 0.000335043674567714
81 0.000335043674567714
82 0.000335043674567714
83 0.000335043674567714
84 0.000335043674567714
85 0.000335043674567714
86 0.000335043674567714
87 0.000335043674567714
88 0.000335043674567714
89 0.000335043674567714
90 0.000335043674567714
91 0.000335043674567714
92 0.000335043674567714
93 0.000335043674567714
94 0.000335043674567714
95 0.000335043674567714
96 0.000332266092300415
97 0.000332266092300415
98 0.000332266092300415
99 0.000320827966788784
100 0.000320827966788784
101 0.000320827966788784
102 0.000320827966788784
103 0.000320827966788784
104 0.000320827966788784
};
\addplot [semithick, color1]
table {%
3 0.588359801061384
4 0.528118667304981
5 0.483122853869663
6 0.438594913222654
7 0.416125859594081
8 0.416125859594081
9 0.409517253979616
10 0.409517253979616
11 0.40501701971522
12 0.38154992222167
13 0.38154992222167
14 0.367260887733098
15 0.351207454520723
16 0.351207454520723
17 0.336815352100566
18 0.336032573142762
19 0.336032573142762
20 0.309740965799485
21 0.309740965799485
22 0.308681500618103
23 0.306738104903575
24 0.300960426209555
25 0.252008725472857
26 0.222252942541084
27 0.222168202755239
28 0.221789245118875
29 0.221614239591597
30 0.220197361522152
31 0.220197361522152
32 0.220175760857484
33 0.220175760857484
34 0.212385177315841
35 0.194721804485143
36 0.184447028783345
37 0.180695650694
38 0.180624756219768
39 0.18009389391047
40 0.18009389391047
41 0.180011379789683
42 0.180011379789683
43 0.180011379789683
44 0.180004651612862
45 0.161127118834604
46 0.125065063019448
47 0.125052628109997
48 0.125037196949163
49 0.125037196949163
50 0.110403670777476
51 0.110403670777476
52 0.110403670777476
53 0.0937259829340821
54 0.0895234622184239
55 0.0894603657957275
56 0.0894603657957275
57 0.0894520024842428
58 0.0894414425231873
59 0.0894414230586848
60 0.0891774298139708
61 0.0891774298139708
62 0.0891771189787732
63 0.069066623146366
64 0.069066623146366
65 0.069066623146366
66 0.069066623146366
67 0.069066623146366
68 0.0685428602790601
69 0.0685409685283565
70 0.0685409685283565
71 0.0683229509817965
72 0.0683229509817965
73 0.0683229509817965
74 0.0683229509817965
75 0.0683214567836271
76 0.0683098051202754
77 0.0683097328296957
78 0.068297353931372
79 0.0682872119186839
80 0.0682872119186839
81 0.0682872119186839
82 0.0682872119186839
83 0.0682872119186839
84 0.0682872119186839
85 0.068286365357379
86 0.068286365357379
87 0.068286365357379
88 0.0682812080282216
89 0.0682812080282216
90 0.0682812080282216
91 0.0682812080282216
92 0.0682812080282216
93 0.0682812080282216
94 0.0682812080282216
95 0.0682803649981013
96 0.0682803649981013
97 0.0682803649981013
98 0.0682799885258746
99 0.0682799885258746
100 0.0682799885258746
101 0.0682798717727064
102 0.0682798717727064
103 0.0682798717727064
104 0.0682798717727064
};
\end{axis}

\end{tikzpicture}

%% file: figures/bop_1d_negeasom_strong.tex
\begin{tikzpicture}

\definecolor{color0}{rgb}{0,0,1}
\definecolor{color1}{rgb}{1,0.549019607843137,0}
\definecolor{color2}{rgb}{1,0.647058823529412,0}
\definecolor{color3}{rgb}{0.564705882352941,0.933333333333333,0.564705882352941}

\begin{axis}[axis on top,
enlarge x limits=false,
enlarge y limits=false,
height=\figureheight,
scale only axis,
tick align=outside,
tick pos=left,
tick pos=left,
width=\figurewidth,
xmin=3, xmax=50,
xtick style={color=black},
xtick={-10,0,10,25,50,75,100},
xticklabels={\ensuremath{-}10,0,10,25,50,75,90},
ymin=-0.03, ymax=0.8,
ytick style={color=black},
ytick={0.   , 0.8},
]
\node[anchor=north east] at (rel axis cs:1,1) {Negeasom 1D (strong)};
\path [draw=blue, fill=blue, opacity=0.3]
(axis cs:3,0.75057637691498)
--(axis cs:3,0.49599277973175)
--(axis cs:4,0.48037099838257)
--(axis cs:5,0.35109281539917)
--(axis cs:6,0.29682850837708)
--(axis cs:7,0.29682850837708)
--(axis cs:8,0.2404693365097)
--(axis cs:9,0.21034264564514)
--(axis cs:10,0.07814860343933)
--(axis cs:11,0.07639145851135)
--(axis cs:12,0.0231522321701)
--(axis cs:13,0.01352620124817)
--(axis cs:14,0.00870168209076)
--(axis cs:15,0.00200831890106)
--(axis cs:16,0.00200831890106)
--(axis cs:17,0.00137591362)
--(axis cs:18,0.00034272670746)
--(axis cs:19,0.00016498565674)
--(axis cs:20,0.00011718273163)
--(axis cs:21,0.00011718273163)
--(axis cs:22,0.00011718273163)
--(axis cs:23,0.0000878572464)
--(axis cs:24,0.0000878572464)
--(axis cs:25,0.0000878572464)
--(axis cs:26,0.00008320808411)
--(axis cs:27,0.00007092952728)
--(axis cs:28,0.00007092952728)
--(axis cs:29,0.00007092952728)
--(axis cs:30,0.00007045269012)
--(axis cs:31,0.00007045269012)
--(axis cs:32,0.0000696182251)
--(axis cs:33,0.00005567073822)
--(axis cs:34,0.00004053115845)
--(axis cs:35,0.00004053115845)
--(axis cs:36,0.00004053115845)
--(axis cs:37,0.00004053115845)
--(axis cs:38,0.00004053115845)
--(axis cs:39,0.00004053115845)
--(axis cs:40,0.00004053115845)
--(axis cs:41,0.0000331401825)
--(axis cs:42,0.0000331401825)
--(axis cs:43,0.0000331401825)
--(axis cs:44,0.0000331401825)
--(axis cs:45,0.0000331401825)
--(axis cs:46,0.0000331401825)
--(axis cs:47,0.0000331401825)
--(axis cs:48,0.0000331401825)
--(axis cs:49,0.0000331401825)
--(axis cs:50,0.00003147125244)
--(axis cs:51,0.00002896785736)
--(axis cs:52,0.00002658367157)
--(axis cs:53,0.00002658367157)
--(axis cs:54,0.00002658367157)
--(axis cs:55,0.00002539157867)
--(axis cs:56,0.00002539157867)
--(axis cs:57,0.00002348423004)
--(axis cs:58,0.00002348423004)
--(axis cs:59,0.00002348423004)
--(axis cs:60,0.00001668930054)
--(axis cs:61,0.00001668930054)
--(axis cs:62,0.00001668930054)
--(axis cs:63,0.00001585483551)
--(axis cs:64,0.00001585483551)
--(axis cs:65,0.00001585483551)
--(axis cs:66,0.00001585483551)
--(axis cs:67,0.00001585483551)
--(axis cs:68,0.00001585483551)
--(axis cs:69,0.00001585483551)
--(axis cs:70,0.00001585483551)
--(axis cs:71,0.00001418590546)
--(axis cs:72,0.00001418590546)
--(axis cs:73,0.00001418590546)
--(axis cs:74,0.00001418590546)
--(axis cs:75,0.00001418590546)
--(axis cs:76,0.00001418590546)
--(axis cs:77,0.00001263618469)
--(axis cs:78,0.00000751018524)
--(axis cs:79,0.00000751018524)
--(axis cs:80,0.00000751018524)
--(axis cs:81,0.00000751018524)
--(axis cs:82,0.00000751018524)
--(axis cs:83,0.00000751018524)
--(axis cs:84,0.00000751018524)
--(axis cs:85,0.00000751018524)
--(axis cs:86,0.00000751018524)
--(axis cs:87,0.00000751018524)
--(axis cs:88,0.00000751018524)
--(axis cs:89,0.00000751018524)
--(axis cs:90,0.0000067949295)
--(axis cs:91,0.0000067949295)
--(axis cs:92,0.0000067949295)
--(axis cs:93,0.0000067949295)
--(axis cs:94,0.0000067949295)
--(axis cs:95,0.0000067949295)
--(axis cs:96,0.0000067949295)
--(axis cs:97,0.0000067949295)
--(axis cs:98,0.0000067949295)
--(axis cs:99,0.00000536441803)
--(axis cs:100,0.00000536441803)
--(axis cs:101,0.00000536441803)
--(axis cs:102,0.00000536441803)
--(axis cs:103,0.00000536441803)
--(axis cs:104,0.00000536441803)
--(axis cs:104,0.00001513957977)
--(axis cs:104,0.00001513957977)
--(axis cs:103,0.00001513957977)
--(axis cs:102,0.00001513957977)
--(axis cs:101,0.00001513957977)
--(axis cs:100,0.00001513957977)
--(axis cs:99,0.00001513957977)
--(axis cs:98,0.0000194311142)
--(axis cs:97,0.0000194311142)
--(axis cs:96,0.0000194311142)
--(axis cs:95,0.0000194311142)
--(axis cs:94,0.0000194311142)
--(axis cs:93,0.0000194311142)
--(axis cs:92,0.0000194311142)
--(axis cs:91,0.0000194311142)
--(axis cs:90,0.0000194311142)
--(axis cs:89,0.00002014636993)
--(axis cs:88,0.00002014636993)
--(axis cs:87,0.00002014636993)
--(axis cs:86,0.00002014636993)
--(axis cs:85,0.00002014636993)
--(axis cs:84,0.00002014636993)
--(axis cs:83,0.00002014636993)
--(axis cs:82,0.00002014636993)
--(axis cs:81,0.00002014636993)
--(axis cs:80,0.00002014636993)
--(axis cs:79,0.00002014636993)
--(axis cs:78,0.00002014636993)
--(axis cs:77,0.00002884864807)
--(axis cs:76,0.00002992153168)
--(axis cs:75,0.00002992153168)
--(axis cs:74,0.00002992153168)
--(axis cs:73,0.00002992153168)
--(axis cs:72,0.00002992153168)
--(axis cs:71,0.00002992153168)
--(axis cs:70,0.00003111362457)
--(axis cs:69,0.00003111362457)
--(axis cs:68,0.00003111362457)
--(axis cs:67,0.00003111362457)
--(axis cs:66,0.00003111362457)
--(axis cs:65,0.00003111362457)
--(axis cs:64,0.00003111362457)
--(axis cs:63,0.00003111362457)
--(axis cs:62,0.00003361701965)
--(axis cs:61,0.00003361701965)
--(axis cs:60,0.00003361701965)
--(axis cs:59,0.00011479854584)
--(axis cs:58,0.00011479854584)
--(axis cs:57,0.00011479854584)
--(axis cs:56,0.00011622905731)
--(axis cs:55,0.00011622905731)
--(axis cs:54,0.00011718273163)
--(axis cs:53,0.00011718273163)
--(axis cs:52,0.00011718273163)
--(axis cs:51,0.000119805336)
--(axis cs:50,0.00012159347534)
--(axis cs:49,0.00012278556824)
--(axis cs:48,0.00012278556824)
--(axis cs:47,0.00012278556824)
--(axis cs:46,0.00012278556824)
--(axis cs:45,0.00012278556824)
--(axis cs:44,0.00012278556824)
--(axis cs:43,0.00012278556824)
--(axis cs:42,0.00012278556824)
--(axis cs:41,0.00012278556824)
--(axis cs:40,0.00013256072998)
--(axis cs:39,0.00013256072998)
--(axis cs:38,0.00013256072998)
--(axis cs:37,0.00013256072998)
--(axis cs:36,0.00013256072998)
--(axis cs:35,0.00013256072998)
--(axis cs:34,0.00013256072998)
--(axis cs:33,0.00014913082123)
--(axis cs:32,0.00016164779663)
--(axis cs:31,0.00016224384308)
--(axis cs:30,0.00016224384308)
--(axis cs:29,0.00016272068024)
--(axis cs:28,0.00016272068024)
--(axis cs:27,0.00016272068024)
--(axis cs:26,0.00018906593323)
--(axis cs:25,0.00019252300262)
--(axis cs:24,0.00019252300262)
--(axis cs:23,0.00019252300262)
--(axis cs:22,0.00022089481354)
--(axis cs:21,0.00022089481354)
--(axis cs:20,0.00022089481354)
--(axis cs:19,0.00029063224792)
--(axis cs:18,0.00092375278473)
--(axis cs:17,0.00332069396973)
--(axis cs:16,0.01666271686554)
--(axis cs:15,0.01666271686554)
--(axis cs:14,0.04924595355988)
--(axis cs:13,0.13490319252014)
--(axis cs:12,0.17712104320526)
--(axis cs:11,0.31223893165588)
--(axis cs:10,0.31346487998962)
--(axis cs:9,0.47528171539307)
--(axis cs:8,0.49795830249786)
--(axis cs:7,0.56651663780212)
--(axis cs:6,0.56651663780212)
--(axis cs:5,0.59837555885315)
--(axis cs:4,0.74207234382629)
--(axis cs:3,0.75057637691498)
--cycle;

\path [draw=color1, fill=color1, opacity=0.3]
(axis cs:3,0.75057640231309)
--(axis cs:3,0.49599281253425)
--(axis cs:4,0.49599281253425)
--(axis cs:5,0.47846499388114)
--(axis cs:6,0.40295818296841)
--(axis cs:7,0.40295818296841)
--(axis cs:8,0.35875512088879)
--(axis cs:9,0.31485860899506)
--(axis cs:10,0.24384661519482)
--(axis cs:11,0.24060097519838)
--(axis cs:12,0.24000530390525)
--(axis cs:13,0.23779279176865)
--(axis cs:14,0.23779278429239)
--(axis cs:15,0.17860648289656)
--(axis cs:16,0.14770020551522)
--(axis cs:17,0.1009609163705)
--(axis cs:18,0.10096089899058)
--(axis cs:19,0.10082512653391)
--(axis cs:20,0.07061630325382)
--(axis cs:21,0.0679285654586)
--(axis cs:22,0.06770545077485)
--(axis cs:23,0.06680857336858)
--(axis cs:24,0.06680857336858)
--(axis cs:25,0.06680835020129)
--(axis cs:26,0.06677090513524)
--(axis cs:27,0.06677090513524)
--(axis cs:28,0.06673381361113)
--(axis cs:29,0.06673381361113)
--(axis cs:30,0.06672668996216)
--(axis cs:31,0.06672668996216)
--(axis cs:32,0.06672668996216)
--(axis cs:33,0.06672668996216)
--(axis cs:34,0.06667818497533)
--(axis cs:35,0.06667818497533)
--(axis cs:36,0.06667818497533)
--(axis cs:37,0.06667818497533)
--(axis cs:38,0.06667817146998)
--(axis cs:39,0.06667817146998)
--(axis cs:40,0.06667816718036)
--(axis cs:41,0.06667261368659)
--(axis cs:42,0.06667261368659)
--(axis cs:43,0.06667261368659)
--(axis cs:44,0.06666962763682)
--(axis cs:45,0.06666962763682)
--(axis cs:46,0.06666962227846)
--(axis cs:47,0.06666962227846)
--(axis cs:48,0.06666962227846)
--(axis cs:49,0.06666962227846)
--(axis cs:50,0.06666962227846)
--(axis cs:51,0.06666962227846)
--(axis cs:52,0.06666962227846)
--(axis cs:53,0.06666901972361)
--(axis cs:54,0.06666901972361)
--(axis cs:55,0.06666901972361)
--(axis cs:56,0.06666901972361)
--(axis cs:57,0.06666901972361)
--(axis cs:58,0.06666901972361)
--(axis cs:59,0.06666899998746)
--(axis cs:60,0.06666899998746)
--(axis cs:61,0.06666899998746)
--(axis cs:62,0.06666899998746)
--(axis cs:63,0.06666899998746)
--(axis cs:64,0.06666899998746)
--(axis cs:65,0.06666899998746)
--(axis cs:66,0.06666824469564)
--(axis cs:67,0.06666824469564)
--(axis cs:68,0.06666824469564)
--(axis cs:69,0.06666824469564)
--(axis cs:70,0.06666785796592)
--(axis cs:71,0.06666785796592)
--(axis cs:72,0.06666785796592)
--(axis cs:73,0.06666785796592)
--(axis cs:74,0.06666785796592)
--(axis cs:75,0.06666785796592)
--(axis cs:76,0.06666785796592)
--(axis cs:77,0.06666785796592)
--(axis cs:78,0.06666785796592)
--(axis cs:79,0.06666785796592)
--(axis cs:80,0.06666785796592)
--(axis cs:81,0.06666785796592)
--(axis cs:82,0.06666785796592)
--(axis cs:83,0.06666785796592)
--(axis cs:84,0.06666785796592)
--(axis cs:85,0.06666785796592)
--(axis cs:86,0.06666785796592)
--(axis cs:87,0.06666785796592)
--(axis cs:88,0.06666785796592)
--(axis cs:89,0.06666785796592)
--(axis cs:90,0.06666778350999)
--(axis cs:91,0.06666778350999)
--(axis cs:92,0.06666778350999)
--(axis cs:93,0.06666778350999)
--(axis cs:94,0.06666778350999)
--(axis cs:95,0.06666778350999)
--(axis cs:96,0.06666778350999)
--(axis cs:97,0.06666778350999)
--(axis cs:98,0.06666778350999)
--(axis cs:99,0.06666778350999)
--(axis cs:100,0.06666778350999)
--(axis cs:101,0.06666778350999)
--(axis cs:102,0.06666778350999)
--(axis cs:103,0.06666778350999)
--(axis cs:104,0.06666778350999)
--(axis cs:104,0.33333384614257)
--(axis cs:104,0.33333384614257)
--(axis cs:103,0.33333384614257)
--(axis cs:102,0.33333384614257)
--(axis cs:101,0.33333384614257)
--(axis cs:100,0.33333384614257)
--(axis cs:99,0.33333384614257)
--(axis cs:98,0.33333384614257)
--(axis cs:97,0.33333384614257)
--(axis cs:96,0.33333384614257)
--(axis cs:95,0.33333384614257)
--(axis cs:94,0.33333384614257)
--(axis cs:93,0.33333384614257)
--(axis cs:92,0.33333384614257)
--(axis cs:91,0.33333384614257)
--(axis cs:90,0.33333384614257)
--(axis cs:89,0.33333389932533)
--(axis cs:88,0.33333389932533)
--(axis cs:87,0.33333389932533)
--(axis cs:86,0.33333389932533)
--(axis cs:85,0.33333389932533)
--(axis cs:84,0.33333389932533)
--(axis cs:83,0.33333389932533)
--(axis cs:82,0.33333389932533)
--(axis cs:81,0.33333389932533)
--(axis cs:80,0.33333389932533)
--(axis cs:79,0.33333389932533)
--(axis cs:78,0.33333389932533)
--(axis cs:77,0.33333389932533)
--(axis cs:76,0.33333389932533)
--(axis cs:75,0.33333389932533)
--(axis cs:74,0.33333389932533)
--(axis cs:73,0.33333389932533)
--(axis cs:72,0.33333389932533)
--(axis cs:71,0.33333389932533)
--(axis cs:70,0.33333389932533)
--(axis cs:69,0.33333417556123)
--(axis cs:68,0.33333417556123)
--(axis cs:67,0.33333417556123)
--(axis cs:66,0.33333417556123)
--(axis cs:65,0.33333471505656)
--(axis cs:64,0.33333471505656)
--(axis cs:63,0.33333471505656)
--(axis cs:62,0.33333471505656)
--(axis cs:61,0.33333471505656)
--(axis cs:60,0.33333471505656)
--(axis cs:59,0.33333471505656)
--(axis cs:58,0.33333472915375)
--(axis cs:57,0.33333472915375)
--(axis cs:56,0.33333472915375)
--(axis cs:55,0.33333472915375)
--(axis cs:54,0.33333472915375)
--(axis cs:53,0.33333472915375)
--(axis cs:52,0.33333515954984)
--(axis cs:51,0.33333515954984)
--(axis cs:50,0.33333515954984)
--(axis cs:49,0.33333515954984)
--(axis cs:48,0.33333515954984)
--(axis cs:47,0.33333515954984)
--(axis cs:46,0.33333515954984)
--(axis cs:45,0.33333518634167)
--(axis cs:44,0.33333518634167)
--(axis cs:43,0.3333373192637)
--(axis cs:42,0.3333373192637)
--(axis cs:41,0.3333373192637)
--(axis cs:40,0.33334128631733)
--(axis cs:39,0.33334130776544)
--(axis cs:38,0.33334130776544)
--(axis cs:37,0.33334137529222)
--(axis cs:36,0.33334137529222)
--(axis cs:35,0.33334137529222)
--(axis cs:34,0.33334137529222)
--(axis cs:33,0.33337603205174)
--(axis cs:32,0.33337603205174)
--(axis cs:31,0.33337603205174)
--(axis cs:30,0.33337603205174)
--(axis cs:29,0.33338112006292)
--(axis cs:28,0.33338112006292)
--(axis cs:27,0.33340761721127)
--(axis cs:26,0.33340761721127)
--(axis cs:25,0.33343436765116)
--(axis cs:24,0.33343455381507)
--(axis cs:23,0.33343455381507)
--(axis cs:22,0.33407908360473)
--(axis cs:21,0.33423924368065)
--(axis cs:20,0.33621850914223)
--(axis cs:19,0.36447369563355)
--(axis cs:18,0.36456508080278)
--(axis cs:17,0.36456516183452)
--(axis cs:16,0.45304295381554)
--(axis cs:15,0.47684425043458)
--(axis cs:14,0.56392511099354)
--(axis cs:13,0.56392512878544)
--(axis cs:12,0.56520116342933)
--(axis cs:11,0.56554359431769)
--(axis cs:10,0.56742356018183)
--(axis cs:9,0.6148195499226)
--(axis cs:8,0.63882360784609)
--(axis cs:7,0.66333730403543)
--(axis cs:6,0.66333730403543)
--(axis cs:5,0.74172989544131)
--(axis cs:4,0.75057640231309)
--(axis cs:3,0.75057640231309)
--cycle;

\path [draw=blue, fill=blue, opacity=0.3]
(axis cs:3,0.75057637691498)
--(axis cs:3,0.49599277973175)
--(axis cs:4,0.47969174385071)
--(axis cs:5,0.45600819587708)
--(axis cs:6,0.26394581794739)
--(axis cs:7,0.14276623725891)
--(axis cs:8,0.09845769405365)
--(axis cs:9,0.03353261947632)
--(axis cs:10,0.00829148292542)
--(axis cs:11,0.00276482105255)
--(axis cs:12,0.001011967659)
--(axis cs:13,0.00053012371063)
--(axis cs:14,0.00028729438782)
--(axis cs:15,0.00028729438782)
--(axis cs:16,0.00028729438782)
--(axis cs:17,0.0002738237381)
--(axis cs:18,0.00026655197144)
--(axis cs:19,0.00026655197144)
--(axis cs:20,0.00026416778564)
--(axis cs:21,0.00008189678192)
--(axis cs:22,0.00008189678192)
--(axis cs:23,0.00007247924805)
--(axis cs:24,0.00007247924805)
--(axis cs:25,0.00004982948303)
--(axis cs:26,0.00004982948303)
--(axis cs:27,0.00004971027374)
--(axis cs:28,0.00004971027374)
--(axis cs:29,0.00004971027374)
--(axis cs:30,0.00004839897156)
--(axis cs:31,0.00004839897156)
--(axis cs:32,0.00004839897156)
--(axis cs:33,0.00002682209015)
--(axis cs:34,0.00002682209015)
--(axis cs:35,0.00002062320709)
--(axis cs:36,0.00002062320709)
--(axis cs:37,0.00000822544098)
--(axis cs:38,0.00000488758087)
--(axis cs:39,0.00000488758087)
--(axis cs:40,0.00000488758087)
--(axis cs:41,0.00000488758087)
--(axis cs:42,0.00000488758087)
--(axis cs:43,0.00000488758087)
--(axis cs:44,0.00000488758087)
--(axis cs:45,0.00000488758087)
--(axis cs:46,0.00000488758087)
--(axis cs:47,0.000004529953)
--(axis cs:48,0.000004529953)
--(axis cs:49,0.000004529953)
--(axis cs:50,0.000004529953)
--(axis cs:51,0.000004529953)
--(axis cs:52,0.000004529953)
--(axis cs:53,0.000004529953)
--(axis cs:54,0.000004529953)
--(axis cs:55,0.000004529953)
--(axis cs:56,0.00000429153442)
--(axis cs:57,0.00000429153442)
--(axis cs:58,0.00000429153442)
--(axis cs:59,0.00000429153442)
--(axis cs:60,0.00000429153442)
--(axis cs:61,0.00000429153442)
--(axis cs:62,0.00000429153442)
--(axis cs:63,0.00000429153442)
--(axis cs:64,0.00000429153442)
--(axis cs:65,0.00000429153442)
--(axis cs:66,0.00000429153442)
--(axis cs:67,0.00000429153442)
--(axis cs:68,0.00000429153442)
--(axis cs:69,0.00000393390656)
--(axis cs:70,0.00000393390656)
--(axis cs:71,0.00000309944153)
--(axis cs:72,0.00000309944153)
--(axis cs:73,0.00000309944153)
--(axis cs:74,0.00000309944153)
--(axis cs:75,0.00000309944153)
--(axis cs:76,0.00000309944153)
--(axis cs:77,0.00000309944153)
--(axis cs:78,0.00000309944153)
--(axis cs:79,0.00000309944153)
--(axis cs:80,0.00000286102295)
--(axis cs:81,0.00000286102295)
--(axis cs:82,0.00000286102295)
--(axis cs:83,0.00000286102295)
--(axis cs:84,0.00000286102295)
--(axis cs:85,0.00000286102295)
--(axis cs:86,0.00000286102295)
--(axis cs:87,0.00000286102295)
--(axis cs:88,0.00000286102295)
--(axis cs:89,0.00000274181366)
--(axis cs:90,0.00000274181366)
--(axis cs:91,0.00000274181366)
--(axis cs:92,0.00000274181366)
--(axis cs:93,0.00000274181366)
--(axis cs:94,0.00000274181366)
--(axis cs:95,0.00000274181366)
--(axis cs:96,0.00000274181366)
--(axis cs:97,0.00000286102295)
--(axis cs:98,0.00000286102295)
--(axis cs:99,0.00000286102295)
--(axis cs:100,0.00000286102295)
--(axis cs:101,0.00000286102295)
--(axis cs:102,0.00000286102295)
--(axis cs:103,0.00000286102295)
--(axis cs:104,0.00000286102295)
--(axis cs:104,0.00000905990601)
--(axis cs:104,0.00000905990601)
--(axis cs:103,0.00000905990601)
--(axis cs:102,0.00000905990601)
--(axis cs:101,0.00000905990601)
--(axis cs:100,0.00000905990601)
--(axis cs:99,0.00000905990601)
--(axis cs:98,0.00000905990601)
--(axis cs:97,0.00000905990601)
--(axis cs:96,0.00000894069672)
--(axis cs:95,0.00000894069672)
--(axis cs:94,0.00000894069672)
--(axis cs:93,0.00000894069672)
--(axis cs:92,0.00000894069672)
--(axis cs:91,0.00000894069672)
--(axis cs:90,0.00000894069672)
--(axis cs:89,0.00000894069672)
--(axis cs:88,0.00000905990601)
--(axis cs:87,0.00000905990601)
--(axis cs:86,0.00000905990601)
--(axis cs:85,0.00000905990601)
--(axis cs:84,0.00000905990601)
--(axis cs:83,0.00000905990601)
--(axis cs:82,0.00000905990601)
--(axis cs:81,0.00000905990601)
--(axis cs:80,0.00000905990601)
--(axis cs:79,0.00000929832458)
--(axis cs:78,0.00000929832458)
--(axis cs:77,0.00000929832458)
--(axis cs:76,0.00000929832458)
--(axis cs:75,0.00000929832458)
--(axis cs:74,0.00000929832458)
--(axis cs:73,0.00000929832458)
--(axis cs:72,0.00000929832458)
--(axis cs:71,0.00000929832458)
--(axis cs:70,0.00000989437103)
--(axis cs:69,0.00000989437103)
--(axis cs:68,0.00001072883606)
--(axis cs:67,0.00001072883606)
--(axis cs:66,0.00001072883606)
--(axis cs:65,0.00001072883606)
--(axis cs:64,0.00001072883606)
--(axis cs:63,0.00001072883606)
--(axis cs:62,0.00001072883606)
--(axis cs:61,0.00001072883606)
--(axis cs:60,0.00001072883606)
--(axis cs:59,0.00001072883606)
--(axis cs:58,0.00001072883606)
--(axis cs:57,0.00001072883606)
--(axis cs:56,0.00001072883606)
--(axis cs:55,0.00001072883606)
--(axis cs:54,0.00001072883606)
--(axis cs:53,0.00001072883606)
--(axis cs:52,0.00001072883606)
--(axis cs:51,0.00001072883606)
--(axis cs:50,0.00001072883606)
--(axis cs:49,0.00001072883606)
--(axis cs:48,0.00001072883606)
--(axis cs:47,0.00001072883606)
--(axis cs:46,0.00001108646393)
--(axis cs:45,0.00001108646393)
--(axis cs:44,0.00001108646393)
--(axis cs:43,0.00001108646393)
--(axis cs:42,0.00001108646393)
--(axis cs:41,0.00001108646393)
--(axis cs:40,0.00001108646393)
--(axis cs:39,0.00001108646393)
--(axis cs:38,0.00001108646393)
--(axis cs:37,0.00003445148468)
--(axis cs:36,0.00005710124969)
--(axis cs:35,0.00005710124969)
--(axis cs:34,0.00006282329559)
--(axis cs:33,0.00006282329559)
--(axis cs:32,0.00027108192444)
--(axis cs:31,0.00027108192444)
--(axis cs:30,0.00027108192444)
--(axis cs:29,0.00495636463165)
--(axis cs:28,0.00700509548187)
--(axis cs:27,0.06435072422028)
--(axis cs:26,0.13904023170471)
--(axis cs:25,0.20003962516785)
--(axis cs:24,0.20005774497986)
--(axis cs:23,0.20005774497986)
--(axis cs:22,0.2000652551651)
--(axis cs:21,0.2000652551651)
--(axis cs:20,0.20021152496338)
--(axis cs:19,0.20021343231201)
--(axis cs:18,0.20021343231201)
--(axis cs:17,0.2002192735672)
--(axis cs:16,0.20023012161255)
--(axis cs:15,0.20023012161255)
--(axis cs:14,0.20023012161255)
--(axis cs:13,0.20042455196381)
--(axis cs:12,0.20081079006195)
--(axis cs:11,0.20221889019012)
--(axis cs:10,0.20671510696411)
--(axis cs:9,0.23228025436401)
--(axis cs:8,0.3569473028183)
--(axis cs:7,0.39308404922485)
--(axis cs:6,0.53377056121826)
--(axis cs:5,0.71067261695862)
--(axis cs:4,0.72347378730774)
--(axis cs:3,0.75057637691498)
--cycle;

\path [draw=color1, fill=color1, opacity=0.3]
(axis cs:3,0.75057640231309)
--(axis cs:3,0.49599281253425)
--(axis cs:4,0.49599281253425)
--(axis cs:5,0.38497801625232)
--(axis cs:6,0.31767962701686)
--(axis cs:7,0.31324334244661)
--(axis cs:8,0.26468069451763)
--(axis cs:9,0.19344414929123)
--(axis cs:10,0.14767898959424)
--(axis cs:11,0.07020928881613)
--(axis cs:12,0.01561209646448)
--(axis cs:13,0.00456146117067)
--(axis cs:14,0.00142070040971)
--(axis cs:15,0.00034223339339)
--(axis cs:16,0.00020911983761)
--(axis cs:17,0.0001771657858)
--(axis cs:18,0.00005647593484)
--(axis cs:19,0.00005647593484)
--(axis cs:20,0.00005304375746)
--(axis cs:21,0.00004541941999)
--(axis cs:22,0.00002333518733)
--(axis cs:23,0.00002088376582)
--(axis cs:24,0.00002088376582)
--(axis cs:25,0.0000083611773)
--(axis cs:26,0.0000078384279)
--(axis cs:27,0.00000601551384)
--(axis cs:28,0.00000601551384)
--(axis cs:29,0.00000278113039)
--(axis cs:30,0.0000024424277)
--(axis cs:31,0.0000024424277)
--(axis cs:32,0.0000024424277)
--(axis cs:33,0.00000209182259)
--(axis cs:34,0.00000209182259)
--(axis cs:35,0.00000209182259)
--(axis cs:36,0.00000209062856)
--(axis cs:37,0.00000208421443)
--(axis cs:38,0.00000208421443)
--(axis cs:39,0.0000020100812)
--(axis cs:40,0.00000200494515)
--(axis cs:41,0.00000200494515)
--(axis cs:42,0.00000200494515)
--(axis cs:43,0.00000200494515)
--(axis cs:44,0.00000168350173)
--(axis cs:45,0.00000165818165)
--(axis cs:46,0.00000165818165)
--(axis cs:47,0.00000165818165)
--(axis cs:48,0.00000096778795)
--(axis cs:49,0.00000096778795)
--(axis cs:50,0.0000002158447)
--(axis cs:51,0.0000002158447)
--(axis cs:52,0.0000002158447)
--(axis cs:53,0.0000002158447)
--(axis cs:54,0.0000002158447)
--(axis cs:55,0.0000002158447)
--(axis cs:56,0.0000002158447)
--(axis cs:57,0.0000002158447)
--(axis cs:58,0.0000002158447)
--(axis cs:59,0.0000002158447)
--(axis cs:60,0.0000002158447)
--(axis cs:61,0.0000002158447)
--(axis cs:62,0.0000002158447)
--(axis cs:63,0.0000002158447)
--(axis cs:64,0.0000002158447)
--(axis cs:65,0.0000002158447)
--(axis cs:66,0.00000019964406)
--(axis cs:67,0.00000019964406)
--(axis cs:68,0.00000019964406)
--(axis cs:69,0.00000019964406)
--(axis cs:70,0.00000019964406)
--(axis cs:71,0.00000019964406)
--(axis cs:72,0.00000019964406)
--(axis cs:73,0.00000019964406)
--(axis cs:74,0.00000019964406)
--(axis cs:75,0.00000015041767)
--(axis cs:76,0.00000015041767)
--(axis cs:77,0.00000015041767)
--(axis cs:78,0.00000015041767)
--(axis cs:79,0.00000013982784)
--(axis cs:80,0.00000013982784)
--(axis cs:81,0.00000013982784)
--(axis cs:82,0.00000013982784)
--(axis cs:83,0.00000013265202)
--(axis cs:84,0.00000013265202)
--(axis cs:85,0.0000001178907)
--(axis cs:86,0.0000001178907)
--(axis cs:87,0.0000001178907)
--(axis cs:88,0.0000001178907)
--(axis cs:89,0.0000001178907)
--(axis cs:90,0.0000001178907)
--(axis cs:91,0.00000007123583)
--(axis cs:92,0.00000007123583)
--(axis cs:93,0.00000007123583)
--(axis cs:94,0.00000007123583)
--(axis cs:95,0.00000007123583)
--(axis cs:96,0.00000007123583)
--(axis cs:97,0.00000007123583)
--(axis cs:98,0.00000007123583)
--(axis cs:99,0.00000007123583)
--(axis cs:100,0.00000007123583)
--(axis cs:101,0.00000003505032)
--(axis cs:102,0.00000003505032)
--(axis cs:103,0.00000003505032)
--(axis cs:104,0.00000003505032)
--(axis cs:104,0.19999996174402)
--(axis cs:104,0.19999996174402)
--(axis cs:103,0.19999996174402)
--(axis cs:102,0.19999996174402)
--(axis cs:101,0.19999996174402)
--(axis cs:100,0.19999999069244)
--(axis cs:99,0.19999999069244)
--(axis cs:98,0.19999999069244)
--(axis cs:97,0.19999999069244)
--(axis cs:96,0.19999999069244)
--(axis cs:95,0.19999999069244)
--(axis cs:94,0.19999999069244)
--(axis cs:93,0.19999999069244)
--(axis cs:92,0.19999999069244)
--(axis cs:91,0.19999999069244)
--(axis cs:90,0.20000002801634)
--(axis cs:89,0.20000002801634)
--(axis cs:88,0.20000002801634)
--(axis cs:87,0.20000002801634)
--(axis cs:86,0.20000002801634)
--(axis cs:85,0.20000002801634)
--(axis cs:84,0.20000003982539)
--(axis cs:83,0.20000003982539)
--(axis cs:82,0.20000004556606)
--(axis cs:81,0.20000004556606)
--(axis cs:80,0.20000004556606)
--(axis cs:79,0.20000004556606)
--(axis cs:78,0.20000005403792)
--(axis cs:77,0.20000005403792)
--(axis cs:76,0.20000005403792)
--(axis cs:75,0.20000005403792)
--(axis cs:74,0.20000009341904)
--(axis cs:73,0.20000009341904)
--(axis cs:72,0.20000009341904)
--(axis cs:71,0.20000009341904)
--(axis cs:70,0.20000009341904)
--(axis cs:69,0.20000009341904)
--(axis cs:68,0.20000009341904)
--(axis cs:67,0.20000009341904)
--(axis cs:66,0.20000009341904)
--(axis cs:65,0.20000010637956)
--(axis cs:64,0.20000010637956)
--(axis cs:63,0.20000010637956)
--(axis cs:62,0.20000010637956)
--(axis cs:61,0.20000010637956)
--(axis cs:60,0.20000010637956)
--(axis cs:59,0.20000010637956)
--(axis cs:58,0.20000010637956)
--(axis cs:57,0.20000010637956)
--(axis cs:56,0.20000010637956)
--(axis cs:55,0.20000010637956)
--(axis cs:54,0.20000010637956)
--(axis cs:53,0.20000010637956)
--(axis cs:52,0.20000010637956)
--(axis cs:51,0.20000010637956)
--(axis cs:50,0.20000010637956)
--(axis cs:49,0.20000070793842)
--(axis cs:48,0.20000070793842)
--(axis cs:47,0.2000012602543)
--(axis cs:46,0.2000012602543)
--(axis cs:45,0.2000012602543)
--(axis cs:44,0.20000128051035)
--(axis cs:43,0.20000153766898)
--(axis cs:42,0.20000153766898)
--(axis cs:41,0.20000153766898)
--(axis cs:40,0.20000153766898)
--(axis cs:39,0.2000015417778)
--(axis cs:38,0.20000160108423)
--(axis cs:37,0.20000160108423)
--(axis cs:36,0.20000160621551)
--(axis cs:35,0.20000160717074)
--(axis cs:34,0.20000160717074)
--(axis cs:33,0.20000160717074)
--(axis cs:32,0.20000188765443)
--(axis cs:31,0.20000188765443)
--(axis cs:30,0.20000188765443)
--(axis cs:29,0.2000021586169)
--(axis cs:28,0.20000474620962)
--(axis cs:27,0.20000474620962)
--(axis cs:26,0.20000620457813)
--(axis cs:25,0.20000662277682)
--(axis cs:24,0.20001664200398)
--(axis cs:23,0.20001664200398)
--(axis cs:22,0.20001860309349)
--(axis cs:21,0.20003627220654)
--(axis cs:20,0.20004237386168)
--(axis cs:19,0.20004511956892)
--(axis cs:18,0.20004511956892)
--(axis cs:17,0.20014176346182)
--(axis cs:16,0.20016732091334)
--(axis cs:15,0.20027393918331)
--(axis cs:14,0.20114712122089)
--(axis cs:13,0.20373945395949)
--(axis cs:12,0.21360516014014)
--(axis cs:11,0.27939028967594)
--(axis cs:10,0.38342269089773)
--(axis cs:9,0.42635133973142)
--(axis cs:8,0.51848626864353)
--(axis cs:7,0.57478888684168)
--(axis cs:6,0.58038440499464)
--(axis cs:5,0.65358890368396)
--(axis cs:4,0.75057640231309)
--(axis cs:3,0.75057640231309)
--cycle;

\addplot [semithick, blue]
table {%
3 0.62328457832336
4 0.61122167110443
5 0.47473418712616
6 0.4316725730896
7 0.4316725730896
8 0.36921381950378
9 0.3428121805191
10 0.19580674171448
11 0.19431519508362
12 0.10013663768768
13 0.07421469688416
14 0.02897381782532
15 0.0093355178833
16 0.0093355178833
17 0.00234830379486
18 0.00063323974609
19 0.00022780895233
20 0.00016903877258
21 0.00016903877258
22 0.00016903877258
23 0.00014019012451
24 0.00014019012451
25 0.00014019012451
26 0.00013613700867
27 0.00011682510376
28 0.00011682510376
29 0.00011682510376
30 0.0001163482666
31 0.0001163482666
32 0.00011563301086
33 0.00010240077972
34 0.00008654594421
35 0.00008654594421
36 0.00008654594421
37 0.00008654594421
38 0.00008654594421
39 0.00008654594421
40 0.00008654594421
41 0.00007796287537
42 0.00007796287537
43 0.00007796287537
44 0.00007796287537
45 0.00007796287537
46 0.00007796287537
47 0.00007796287537
48 0.00007796287537
49 0.00007796287537
50 0.00007653236389
51 0.00007438659668
52 0.0000718832016
53 0.0000718832016
54 0.0000718832016
55 0.00007081031799
56 0.00007081031799
57 0.00006914138794
58 0.00006914138794
59 0.00006914138794
60 0.0000251531601
61 0.0000251531601
62 0.0000251531601
63 0.00002348423004
64 0.00002348423004
65 0.00002348423004
66 0.00002348423004
67 0.00002348423004
68 0.00002348423004
69 0.00002348423004
70 0.00002348423004
71 0.00002205371857
72 0.00002205371857
73 0.00002205371857
74 0.00002205371857
75 0.00002205371857
76 0.00002205371857
77 0.00002074241638
78 0.00001382827759
79 0.00001382827759
80 0.00001382827759
81 0.00001382827759
82 0.00001382827759
83 0.00001382827759
84 0.00001382827759
85 0.00001382827759
86 0.00001382827759
87 0.00001382827759
88 0.00001382827759
89 0.00001382827759
90 0.00001311302185
91 0.00001311302185
92 0.00001311302185
93 0.00001311302185
94 0.00001311302185
95 0.00001311302185
96 0.00001311302185
97 0.00001311302185
98 0.00001311302185
99 0.0000102519989
100 0.0000102519989
101 0.0000102519989
102 0.0000102519989
103 0.0000102519989
104 0.0000102519989
};
\addplot [semithick, color1, dash dot]
table {%
3 0.62328460742367
4 0.62328460742367
5 0.61009744466122
6 0.53314774350192
7 0.53314774350192
8 0.49878936436744
9 0.46483907945883
10 0.40563508768833
11 0.40307228475803
12 0.40260323366729
13 0.40085896027704
14 0.40085894764296
15 0.32772536666557
16 0.30037157966538
17 0.23276303910251
18 0.23276298989668
19 0.23264941108373
20 0.20341740619802
21 0.20108390456963
22 0.20089226718979
23 0.20012156359183
24 0.20012156359183
25 0.20012135892623
26 0.20008926117326
27 0.20008926117326
28 0.20005746683703
29 0.20005746683703
30 0.20005136100695
31 0.20005136100695
32 0.20005136100695
33 0.20005136100695
34 0.20000978013378
35 0.20000978013378
36 0.20000978013378
37 0.20000978013378
38 0.20000973961771
39 0.20000973961771
40 0.20000972674885
41 0.20000496647514
42 0.20000496647514
43 0.20000496647514
44 0.20000240698925
45 0.20000240698925
46 0.20000239091415
47 0.20000239091415
48 0.20000239091415
49 0.20000239091415
50 0.20000239091415
51 0.20000239091415
52 0.20000239091415
53 0.20000187443868
54 0.20000187443868
55 0.20000187443868
56 0.20000187443868
57 0.20000187443868
58 0.20000187443868
59 0.20000185752201
60 0.20000185752201
61 0.20000185752201
62 0.20000185752201
63 0.20000185752201
64 0.20000185752201
65 0.20000185752201
66 0.20000121012843
67 0.20000121012843
68 0.20000121012843
69 0.20000121012843
70 0.20000087864563
71 0.20000087864563
72 0.20000087864563
73 0.20000087864563
74 0.20000087864563
75 0.20000087864563
76 0.20000087864563
77 0.20000087864563
78 0.20000087864563
79 0.20000087864563
80 0.20000087864563
81 0.20000087864563
82 0.20000087864563
83 0.20000087864563
84 0.20000087864563
85 0.20000087864563
86 0.20000087864563
87 0.20000087864563
88 0.20000087864563
89 0.20000087864563
90 0.20000081482628
91 0.20000081482628
92 0.20000081482628
93 0.20000081482628
94 0.20000081482628
95 0.20000081482628
96 0.20000081482628
97 0.20000081482628
98 0.20000081482628
99 0.20000081482628
100 0.20000081482628
101 0.20000081482628
102 0.20000081482628
103 0.20000081482628
104 0.20000081482628
};
\addplot [semithick, blue, dash dot]
table {%
3 0.62328457832336
4 0.60158276557922
5 0.58334040641785
6 0.39885818958282
7 0.26792514324188
8 0.22770249843597
9 0.13290643692017
10 0.10750329494476
11 0.10249185562134
12 0.10091137886047
13 0.10047733783722
14 0.10025870800018
15 0.10025870800018
16 0.10025870800018
17 0.10024654865265
18 0.10023999214172
19 0.10023999214172
20 0.10023784637451
21 0.10007357597351
22 0.10007357597351
23 0.10006511211395
24 0.10006511211395
25 0.10004472732544
26 0.06954503059387
27 0.03220021724701
28 0.00352740287781
29 0.0025030374527
30 0.000159740448
31 0.000159740448
32 0.000159740448
33 0.00004482269287
34 0.00004482269287
35 0.00003886222839
36 0.00003886222839
37 0.00002133846283
38 0.0000079870224
39 0.0000079870224
40 0.0000079870224
41 0.0000079870224
42 0.0000079870224
43 0.0000079870224
44 0.0000079870224
45 0.0000079870224
46 0.0000079870224
47 0.00000762939453
48 0.00000762939453
49 0.00000762939453
50 0.00000762939453
51 0.00000762939453
52 0.00000762939453
53 0.00000762939453
54 0.00000762939453
55 0.00000762939453
56 0.00000751018524
57 0.00000751018524
58 0.00000751018524
59 0.00000751018524
60 0.00000751018524
61 0.00000751018524
62 0.00000751018524
63 0.00000751018524
64 0.00000751018524
65 0.00000751018524
66 0.00000751018524
67 0.00000751018524
68 0.00000751018524
69 0.00000691413879
70 0.00000691413879
71 0.00000619888306
72 0.00000619888306
73 0.00000619888306
74 0.00000619888306
75 0.00000619888306
76 0.00000619888306
77 0.00000619888306
78 0.00000619888306
79 0.00000619888306
80 0.00000596046448
81 0.00000596046448
82 0.00000596046448
83 0.00000596046448
84 0.00000596046448
85 0.00000596046448
86 0.00000596046448
87 0.00000596046448
88 0.00000596046448
89 0.00000584125519
90 0.00000584125519
91 0.00000584125519
92 0.00000584125519
93 0.00000584125519
94 0.00000584125519
95 0.00000584125519
96 0.00000584125519
97 0.00000596046448
98 0.00000596046448
99 0.00000596046448
100 0.00000596046448
101 0.00000596046448
102 0.00000596046448
103 0.00000596046448
104 0.00000596046448
};
\addplot [semithick, color1]
table {%
3 0.62328460742367
4 0.62328460742367
5 0.51928345996814
6 0.44903201600575
7 0.44401611464415
8 0.39158348158058
9 0.30989774451133
10 0.26555084024598
11 0.17479978924603
12 0.11460862830231
13 0.10415045756508
14 0.1012839108153
15 0.10030808628835
16 0.10018822037547
17 0.10015946462381
18 0.10005079775188
19 0.10005079775188
20 0.10004770880957
21 0.10004084581327
22 0.10002096914041
23 0.1000187628849
24 0.1000187628849
25 0.10000749197706
26 0.10000702150302
27 0.10000538086173
28 0.10000538086173
29 0.10000246987364
30 0.10000216504107
31 0.10000216504107
32 0.10000216504107
33 0.10000184949667
34 0.10000184949667
35 0.10000184949667
36 0.10000184842203
37 0.10000184264933
38 0.10000184264933
39 0.1000017759295
40 0.10000177130706
41 0.10000177130706
42 0.10000177130706
43 0.10000177130706
44 0.10000148200604
45 0.10000145921797
46 0.10000145921797
47 0.10000145921797
48 0.10000083786319
49 0.10000083786319
50 0.10000016111213
51 0.10000016111213
52 0.10000016111213
53 0.10000016111213
54 0.10000016111213
55 0.10000016111213
56 0.10000016111213
57 0.10000016111213
58 0.10000016111213
59 0.10000016111213
60 0.10000016111213
61 0.10000016111213
62 0.10000016111213
63 0.10000016111213
64 0.10000016111213
65 0.10000016111213
66 0.10000014653155
67 0.10000014653155
68 0.10000014653155
69 0.10000014653155
70 0.10000014653155
71 0.10000014653155
72 0.10000014653155
73 0.10000014653155
74 0.10000014653155
75 0.1000001022278
76 0.1000001022278
77 0.1000001022278
78 0.1000001022278
79 0.10000009269695
80 0.10000009269695
81 0.10000009269695
82 0.10000009269695
83 0.10000008623871
84 0.10000008623871
85 0.10000007295352
86 0.10000007295352
87 0.10000007295352
88 0.10000007295352
89 0.10000007295352
90 0.10000007295352
91 0.10000003096414
92 0.10000003096414
93 0.10000003096414
94 0.10000003096414
95 0.10000003096414
96 0.10000003096414
97 0.10000003096414
98 0.10000003096414
99 0.10000003096414
100 0.10000003096414
101 0.09999999839717
102 0.09999999839717
103 0.09999999839717
104 0.09999999839717
};
\end{axis}

\end{tikzpicture}

%% file: figures/bop_2d_braninscaled_strong.tex
\begin{tikzpicture}

\definecolor{color0}{rgb}{0,0,1}
\definecolor{color1}{rgb}{1,0.549019607843137,0}
\definecolor{color2}{rgb}{1,0.647058823529412,0}
\definecolor{color3}{rgb}{0.564705882352941,0.933333333333333,0.564705882352941}

\begin{axis}[axis on top,
enlarge x limits=false,
enlarge y limits=false,
height=\figureheight,
scale only axis,
tick align=outside,
tick pos=left,
tick pos=left,
width=\figurewidth,
xlabel={Iteration},
xmin=10, xmax=75,
xtick style={color=black},
xtick={-10,0,10,25,50,75,100},
xticklabels={\ensuremath{-}10,0,10,25,50,75,90},
ylabel={Regret},
ymin=-0.005, ymax=0.13,
ytick style={color=black},
ytick={0.   , 0.13},
]
\node[anchor=north east] at (rel axis cs:1,1) {Branin 2D (strong)};
\path [draw=color1, fill=color1, opacity=0.3]
(axis cs:10,0.0790769466867583)
--(axis cs:10,0.0426653284187181)
--(axis cs:11,0.0306546828769693)
--(axis cs:12,0.0306546828769693)
--(axis cs:13,0.0306546828769693)
--(axis cs:14,0.0293053285944014)
--(axis cs:15,0.0221028894295163)
--(axis cs:16,0.0158706268492518)
--(axis cs:17,0.0142175826092757)
--(axis cs:18,0.0142175826092757)
--(axis cs:19,0.0142175826092757)
--(axis cs:20,0.0142175826092757)
--(axis cs:21,0.0142175826092757)
--(axis cs:22,0.0142175826092757)
--(axis cs:23,0.0107553272324083)
--(axis cs:24,0.00777086295004233)
--(axis cs:25,0.00583077940172814)
--(axis cs:26,0.00397016219536023)
--(axis cs:27,0.0026354857928213)
--(axis cs:28,0.0026354857928213)
--(axis cs:29,0.00154106104000021)
--(axis cs:30,0.000995267030959056)
--(axis cs:31,0.000641329560865455)
--(axis cs:32,0.00053525466357823)
--(axis cs:33,0.000389581575921349)
--(axis cs:34,0.000389581575921349)
--(axis cs:35,0.000389581575921349)
--(axis cs:36,0.000389581575921349)
--(axis cs:37,0.000368466762199426)
--(axis cs:38,0.000368466762199426)
--(axis cs:39,0.000368466762199426)
--(axis cs:40,0.000368466762199426)
--(axis cs:41,0.000365312528858806)
--(axis cs:42,0.000347321167729089)
--(axis cs:43,0.000236842413287356)
--(axis cs:44,0.000189840101019434)
--(axis cs:45,0.000181095099486039)
--(axis cs:46,0.000181095099486039)
--(axis cs:47,0.000144000285670434)
--(axis cs:48,0.000129579480441725)
--(axis cs:49,0.000129579480441725)
--(axis cs:50,0.000129579480441725)
--(axis cs:51,0.000107850484256383)
--(axis cs:52,9.36019568106066e-05)
--(axis cs:53,9.36019568106066e-05)
--(axis cs:54,9.36019568106066e-05)
--(axis cs:55,9.36019568106066e-05)
--(axis cs:56,8.79466765590913e-05)
--(axis cs:57,8.79466765590913e-05)
--(axis cs:58,8.79466765590913e-05)
--(axis cs:59,7.86976588557927e-05)
--(axis cs:60,7.86976588557927e-05)
--(axis cs:61,7.86976588557927e-05)
--(axis cs:62,7.86976588557927e-05)
--(axis cs:63,7.86976588557927e-05)
--(axis cs:64,7.86976588557927e-05)
--(axis cs:65,7.59688072380996e-05)
--(axis cs:66,7.59688072380996e-05)
--(axis cs:67,6.38793922556885e-05)
--(axis cs:68,6.38793922556885e-05)
--(axis cs:69,6.01962813835207e-05)
--(axis cs:70,6.01962813835207e-05)
--(axis cs:71,5.79231198375472e-05)
--(axis cs:72,5.79231198375472e-05)
--(axis cs:73,5.79231198375472e-05)
--(axis cs:74,5.79231198375472e-05)
--(axis cs:75,5.79231198375472e-05)
--(axis cs:76,5.79231198375472e-05)
--(axis cs:77,5.43216062820915e-05)
--(axis cs:78,4.54693127748051e-05)
--(axis cs:79,4.54693127748051e-05)
--(axis cs:80,4.54693127748051e-05)
--(axis cs:81,4.54693127748051e-05)
--(axis cs:82,4.54693127748051e-05)
--(axis cs:83,4.54693127748051e-05)
--(axis cs:84,4.54693127748051e-05)
--(axis cs:85,4.54693127748051e-05)
--(axis cs:86,4.54693127748051e-05)
--(axis cs:87,4.54693127748051e-05)
--(axis cs:88,4.54693127748051e-05)
--(axis cs:89,4.54693127748051e-05)
--(axis cs:90,4.03446798540428e-05)
--(axis cs:91,4.03446798540428e-05)
--(axis cs:92,4.03446798540428e-05)
--(axis cs:93,4.03446798540428e-05)
--(axis cs:94,4.03446798540428e-05)
--(axis cs:95,4.03446798540428e-05)
--(axis cs:96,4.03446798540428e-05)
--(axis cs:97,4.03446798540428e-05)
--(axis cs:98,4.03446798540428e-05)
--(axis cs:99,4.03446798540428e-05)
--(axis cs:100,4.03446798540428e-05)
--(axis cs:101,4.03446798540428e-05)
--(axis cs:102,3.61255314862911e-05)
--(axis cs:103,3.61255314862911e-05)
--(axis cs:104,3.61255314862911e-05)
--(axis cs:105,3.61255314862911e-05)
--(axis cs:106,3.61255314862911e-05)
--(axis cs:107,3.61255314862911e-05)
--(axis cs:108,3.61255314862911e-05)
--(axis cs:109,3.61255314862911e-05)
--(axis cs:110,3.61255314862911e-05)
--(axis cs:111,2.59796958265891e-05)
--(axis cs:111,4.35237362553075e-05)
--(axis cs:111,4.35237362553075e-05)
--(axis cs:110,5.89081714855785e-05)
--(axis cs:109,5.89081714855785e-05)
--(axis cs:108,5.89081714855785e-05)
--(axis cs:107,5.89081714855785e-05)
--(axis cs:106,5.89081714855785e-05)
--(axis cs:105,5.89081714855785e-05)
--(axis cs:104,5.89081714855785e-05)
--(axis cs:103,5.89081714855785e-05)
--(axis cs:102,5.89081714855785e-05)
--(axis cs:101,7.11678859540601e-05)
--(axis cs:100,7.11678859540601e-05)
--(axis cs:99,7.11678859540601e-05)
--(axis cs:98,7.11678859540601e-05)
--(axis cs:97,7.11678859540601e-05)
--(axis cs:96,7.11678859540601e-05)
--(axis cs:95,7.11678859540601e-05)
--(axis cs:94,7.11678859540601e-05)
--(axis cs:93,7.11678859540601e-05)
--(axis cs:92,7.11678859540601e-05)
--(axis cs:91,7.11678859540601e-05)
--(axis cs:90,7.11678859540601e-05)
--(axis cs:89,7.64574169632926e-05)
--(axis cs:88,7.64574169632926e-05)
--(axis cs:87,7.64574169632926e-05)
--(axis cs:86,7.64574169632926e-05)
--(axis cs:85,7.64574169632926e-05)
--(axis cs:84,7.64574169632926e-05)
--(axis cs:83,7.64574169632926e-05)
--(axis cs:82,7.64574169632926e-05)
--(axis cs:81,7.64574169632926e-05)
--(axis cs:80,7.64574169632926e-05)
--(axis cs:79,7.64574169632926e-05)
--(axis cs:78,7.64574169632926e-05)
--(axis cs:77,8.0991871843529e-05)
--(axis cs:76,8.78646836882626e-05)
--(axis cs:75,8.78646836882626e-05)
--(axis cs:74,8.78646836882626e-05)
--(axis cs:73,8.78646836882626e-05)
--(axis cs:72,8.78646836882626e-05)
--(axis cs:71,8.78646836882626e-05)
--(axis cs:70,8.87166257975884e-05)
--(axis cs:69,8.87166257975884e-05)
--(axis cs:68,0.000120356111794368)
--(axis cs:67,0.000120356111794368)
--(axis cs:66,0.00013013596906252)
--(axis cs:65,0.00013013596906252)
--(axis cs:64,0.000133846150494137)
--(axis cs:63,0.000133846150494137)
--(axis cs:62,0.000133846150494137)
--(axis cs:61,0.000133846150494137)
--(axis cs:60,0.000133846150494137)
--(axis cs:59,0.000133846150494137)
--(axis cs:58,0.000141480643523873)
--(axis cs:57,0.000141480643523873)
--(axis cs:56,0.000141480643523873)
--(axis cs:55,0.000148136088871787)
--(axis cs:54,0.000148136088871787)
--(axis cs:53,0.000148136088871787)
--(axis cs:52,0.000148136088871787)
--(axis cs:51,0.000195416822243438)
--(axis cs:50,0.000227777602732134)
--(axis cs:49,0.000227777602732134)
--(axis cs:48,0.000227777602732134)
--(axis cs:47,0.000269823594863747)
--(axis cs:46,0.000476185410998209)
--(axis cs:45,0.000476185410998209)
--(axis cs:44,0.000481186927387279)
--(axis cs:43,0.000516514458919655)
--(axis cs:42,0.000657576604979979)
--(axis cs:41,0.000805139443669547)
--(axis cs:40,0.000807446910025156)
--(axis cs:39,0.000807446910025156)
--(axis cs:38,0.000807446910025156)
--(axis cs:37,0.000807446910025156)
--(axis cs:36,0.000826366202363805)
--(axis cs:35,0.000826366202363805)
--(axis cs:34,0.000826366202363805)
--(axis cs:33,0.000826366202363805)
--(axis cs:32,0.00110226483253051)
--(axis cs:31,0.00299879754166336)
--(axis cs:30,0.00485206012687026)
--(axis cs:29,0.00543907618172381)
--(axis cs:28,0.00660044755147615)
--(axis cs:27,0.00660044755147615)
--(axis cs:26,0.00806033660758665)
--(axis cs:25,0.0116156169744201)
--(axis cs:24,0.0130699159480929)
--(axis cs:23,0.015374081441606)
--(axis cs:22,0.0205193196839911)
--(axis cs:21,0.0205193196839911)
--(axis cs:20,0.0205193196839911)
--(axis cs:19,0.0205193196839911)
--(axis cs:18,0.0205193196839911)
--(axis cs:17,0.0205193196839911)
--(axis cs:16,0.0315385958978431)
--(axis cs:15,0.0386185398392212)
--(axis cs:14,0.0573221503852342)
--(axis cs:13,0.0665556116646921)
--(axis cs:12,0.0665556116646921)
--(axis cs:11,0.0665556116646921)
--(axis cs:10,0.0790769466867583)
--cycle;

\path [draw=blue, fill=blue, opacity=0.3]
(axis cs:10,0.0790769159793854)
--(axis cs:10,0.0426652915775776)
--(axis cs:11,0.0426652915775776)
--(axis cs:12,0.0426652915775776)
--(axis cs:13,0.0426652915775776)
--(axis cs:14,0.0421311482787132)
--(axis cs:15,0.0421311482787132)
--(axis cs:16,0.0421311482787132)
--(axis cs:17,0.0421311482787132)
--(axis cs:18,0.0268724337220192)
--(axis cs:19,0.0268724337220192)
--(axis cs:20,0.0268724337220192)
--(axis cs:21,0.0268724337220192)
--(axis cs:22,0.0246674343943596)
--(axis cs:23,0.0213195011019707)
--(axis cs:24,0.0213195011019707)
--(axis cs:25,0.0213195011019707)
--(axis cs:26,0.0213195011019707)
--(axis cs:27,0.0165764540433884)
--(axis cs:28,0.0165764540433884)
--(axis cs:29,0.0163340214639902)
--(axis cs:30,0.0160277225077152)
--(axis cs:31,0.0102095287293196)
--(axis cs:32,0.0102095287293196)
--(axis cs:33,0.0102095287293196)
--(axis cs:34,0.0102095287293196)
--(axis cs:35,0.00898562744259834)
--(axis cs:36,0.00890415161848068)
--(axis cs:37,0.00890415161848068)
--(axis cs:38,0.00890415161848068)
--(axis cs:39,0.00890415161848068)
--(axis cs:40,0.00890415161848068)
--(axis cs:41,0.00890415161848068)
--(axis cs:42,0.00890415161848068)
--(axis cs:43,0.00828212965279818)
--(axis cs:44,0.00828212965279818)
--(axis cs:45,0.00828212965279818)
--(axis cs:46,0.00828212965279818)
--(axis cs:47,0.00759370950981975)
--(axis cs:48,0.00759370950981975)
--(axis cs:49,0.00759370950981975)
--(axis cs:50,0.00759370950981975)
--(axis cs:51,0.00626007467508316)
--(axis cs:52,0.0060425316914916)
--(axis cs:53,0.00585738755762577)
--(axis cs:54,0.00585738755762577)
--(axis cs:55,0.00585738755762577)
--(axis cs:56,0.00585738755762577)
--(axis cs:57,0.00585738755762577)
--(axis cs:58,0.00507774064317346)
--(axis cs:59,0.00507774064317346)
--(axis cs:60,0.00507774064317346)
--(axis cs:61,0.0037662866525352)
--(axis cs:62,0.00370398117229342)
--(axis cs:63,0.00352203100919724)
--(axis cs:64,0.00352203100919724)
--(axis cs:65,0.00352203100919724)
--(axis cs:66,0.00352203100919724)
--(axis cs:67,0.00352203100919724)
--(axis cs:68,0.00352203100919724)
--(axis cs:69,0.0031837101560086)
--(axis cs:70,0.0031837101560086)
--(axis cs:71,0.003110685152933)
--(axis cs:72,0.003110685152933)
--(axis cs:73,0.003110685152933)
--(axis cs:74,0.00288679543882608)
--(axis cs:75,0.00288679543882608)
--(axis cs:76,0.00288679543882608)
--(axis cs:77,0.00288679543882608)
--(axis cs:78,0.00288679543882608)
--(axis cs:79,0.00288679543882608)
--(axis cs:80,0.00288679543882608)
--(axis cs:81,0.00288679543882608)
--(axis cs:82,0.00288679543882608)
--(axis cs:83,0.00288679543882608)
--(axis cs:84,0.00259434850886464)
--(axis cs:85,0.00259434850886464)
--(axis cs:86,0.00259434850886464)
--(axis cs:87,0.00259434850886464)
--(axis cs:88,0.00259434850886464)
--(axis cs:89,0.00259434850886464)
--(axis cs:90,0.00247061904519796)
--(axis cs:91,0.00245873769745231)
--(axis cs:92,0.00245873769745231)
--(axis cs:93,0.00245873769745231)
--(axis cs:94,0.00245873769745231)
--(axis cs:95,0.00245873769745231)
--(axis cs:96,0.00245873769745231)
--(axis cs:97,0.00245873769745231)
--(axis cs:98,0.00245873769745231)
--(axis cs:99,0.00223315856419504)
--(axis cs:100,0.00223315856419504)
--(axis cs:101,0.00223315856419504)
--(axis cs:102,0.00223315856419504)
--(axis cs:103,0.00223315856419504)
--(axis cs:104,0.00223315856419504)
--(axis cs:105,0.00223315856419504)
--(axis cs:106,0.00223315856419504)
--(axis cs:107,0.00223315856419504)
--(axis cs:108,0.00223315856419504)
--(axis cs:109,0.00223315856419504)
--(axis cs:110,0.00223315856419504)
--(axis cs:111,0.00223315856419504)
--(axis cs:111,0.00317313172854483)
--(axis cs:111,0.00317313172854483)
--(axis cs:110,0.00317313172854483)
--(axis cs:109,0.00317313172854483)
--(axis cs:108,0.00317313172854483)
--(axis cs:107,0.00317313172854483)
--(axis cs:106,0.00317313172854483)
--(axis cs:105,0.00317313172854483)
--(axis cs:104,0.00317313172854483)
--(axis cs:103,0.00317313172854483)
--(axis cs:102,0.00317313172854483)
--(axis cs:101,0.00317313172854483)
--(axis cs:100,0.00317313172854483)
--(axis cs:99,0.00317313172854483)
--(axis cs:98,0.0032855705358088)
--(axis cs:97,0.0032855705358088)
--(axis cs:96,0.0032855705358088)
--(axis cs:95,0.0032855705358088)
--(axis cs:94,0.0032855705358088)
--(axis cs:93,0.0032855705358088)
--(axis cs:92,0.0032855705358088)
--(axis cs:91,0.0032855705358088)
--(axis cs:90,0.00334762874990702)
--(axis cs:89,0.00346699682995677)
--(axis cs:88,0.00346699682995677)
--(axis cs:87,0.00346699682995677)
--(axis cs:86,0.00346699682995677)
--(axis cs:85,0.00346699682995677)
--(axis cs:84,0.00346699682995677)
--(axis cs:83,0.00429309066385031)
--(axis cs:82,0.00429309066385031)
--(axis cs:81,0.00429309066385031)
--(axis cs:80,0.00429309066385031)
--(axis cs:79,0.00429309066385031)
--(axis cs:78,0.00429309066385031)
--(axis cs:77,0.00429309066385031)
--(axis cs:76,0.00429309066385031)
--(axis cs:75,0.00429309066385031)
--(axis cs:74,0.00429309066385031)
--(axis cs:73,0.0044379155151546)
--(axis cs:72,0.0044379155151546)
--(axis cs:71,0.0044379155151546)
--(axis cs:70,0.00450689811259508)
--(axis cs:69,0.00450689811259508)
--(axis cs:68,0.00525522977113724)
--(axis cs:67,0.00525522977113724)
--(axis cs:66,0.00525522977113724)
--(axis cs:65,0.00525522977113724)
--(axis cs:64,0.00525522977113724)
--(axis cs:63,0.00525522977113724)
--(axis cs:62,0.00733706401661038)
--(axis cs:61,0.0108415596187115)
--(axis cs:60,0.0138947162777185)
--(axis cs:59,0.0138947162777185)
--(axis cs:58,0.0138947162777185)
--(axis cs:57,0.0145857241004705)
--(axis cs:56,0.0145857241004705)
--(axis cs:55,0.0145857241004705)
--(axis cs:54,0.0145857241004705)
--(axis cs:53,0.0145857241004705)
--(axis cs:52,0.0147025967016816)
--(axis cs:51,0.0148509368300438)
--(axis cs:50,0.0173678509891033)
--(axis cs:49,0.0173678509891033)
--(axis cs:48,0.0173678509891033)
--(axis cs:47,0.0173678509891033)
--(axis cs:46,0.017951425164938)
--(axis cs:45,0.017951425164938)
--(axis cs:44,0.017951425164938)
--(axis cs:43,0.017951425164938)
--(axis cs:42,0.0183906629681587)
--(axis cs:41,0.0183906629681587)
--(axis cs:40,0.0183906629681587)
--(axis cs:39,0.0183906629681587)
--(axis cs:38,0.0183906629681587)
--(axis cs:37,0.0183906629681587)
--(axis cs:36,0.0183906629681587)
--(axis cs:35,0.0190260224044323)
--(axis cs:34,0.0202506203204393)
--(axis cs:33,0.0202506203204393)
--(axis cs:32,0.0202506203204393)
--(axis cs:31,0.0202506203204393)
--(axis cs:30,0.03426219150424)
--(axis cs:29,0.0348071381449699)
--(axis cs:28,0.0349270105361938)
--(axis cs:27,0.0349270105361938)
--(axis cs:26,0.0429999455809593)
--(axis cs:25,0.0429999455809593)
--(axis cs:24,0.0429999455809593)
--(axis cs:23,0.0429999455809593)
--(axis cs:22,0.044979564845562)
--(axis cs:21,0.0484875813126564)
--(axis cs:20,0.0484875813126564)
--(axis cs:19,0.0484875813126564)
--(axis cs:18,0.0484875813126564)
--(axis cs:17,0.0783148482441902)
--(axis cs:16,0.0783148482441902)
--(axis cs:15,0.0783148482441902)
--(axis cs:14,0.0783148482441902)
--(axis cs:13,0.0790769159793854)
--(axis cs:12,0.0790769159793854)
--(axis cs:11,0.0790769159793854)
--(axis cs:10,0.0790769159793854)
--cycle;

\path [draw=blue, fill=blue, opacity=0.3]
(axis cs:10,0.0790769159793854)
--(axis cs:10,0.0426652915775776)
--(axis cs:11,0.0332706943154335)
--(axis cs:12,0.026028610765934)
--(axis cs:13,0.0199037604033947)
--(axis cs:14,0.0167500823736191)
--(axis cs:15,0.0167500823736191)
--(axis cs:16,0.0157139245420694)
--(axis cs:17,0.013634811155498)
--(axis cs:18,0.0130574814975262)
--(axis cs:19,0.0110860057175159)
--(axis cs:20,0.0095641752704978)
--(axis cs:21,0.0095641752704978)
--(axis cs:22,0.00818609166890383)
--(axis cs:23,0.00692386459559202)
--(axis cs:24,0.00499430019408464)
--(axis cs:25,0.00499430019408464)
--(axis cs:26,0.00353463855572045)
--(axis cs:27,0.00353463855572045)
--(axis cs:28,0.00353463855572045)
--(axis cs:29,0.00325878383591771)
--(axis cs:30,0.00232037575915456)
--(axis cs:31,0.00232037575915456)
--(axis cs:32,0.00179158244282007)
--(axis cs:33,0.00179158244282007)
--(axis cs:34,0.00179158244282007)
--(axis cs:35,0.00179158244282007)
--(axis cs:36,0.00179158244282007)
--(axis cs:37,0.0014443164691329)
--(axis cs:38,0.0014443164691329)
--(axis cs:39,0.00105189986061305)
--(axis cs:40,0.00105189986061305)
--(axis cs:41,0.00105189986061305)
--(axis cs:42,0.00105189986061305)
--(axis cs:43,0.00105189986061305)
--(axis cs:44,0.00105189986061305)
--(axis cs:45,0.00105189986061305)
--(axis cs:46,0.00105189986061305)
--(axis cs:47,0.00105189986061305)
--(axis cs:48,0.00105189986061305)
--(axis cs:49,0.00105155690107495)
--(axis cs:50,0.00105155690107495)
--(axis cs:51,0.00105155690107495)
--(axis cs:52,0.00105155690107495)
--(axis cs:53,0.00105155690107495)
--(axis cs:54,0.00105155690107495)
--(axis cs:55,0.00105155690107495)
--(axis cs:56,0.00105155690107495)
--(axis cs:57,0.00105155690107495)
--(axis cs:58,0.00105155690107495)
--(axis cs:59,0.00105155690107495)
--(axis cs:60,0.00105155690107495)
--(axis cs:61,0.00105155690107495)
--(axis cs:62,0.00105155690107495)
--(axis cs:63,0.00105155690107495)
--(axis cs:64,0.00105155690107495)
--(axis cs:65,0.00105155690107495)
--(axis cs:66,0.00105155690107495)
--(axis cs:67,0.00084292225074023)
--(axis cs:68,0.00084292225074023)
--(axis cs:69,0.00084292225074023)
--(axis cs:70,0.00084292225074023)
--(axis cs:71,0.00084292225074023)
--(axis cs:72,0.00084292225074023)
--(axis cs:73,0.00084292225074023)
--(axis cs:74,0.00084292225074023)
--(axis cs:75,0.00084292225074023)
--(axis cs:76,0.00084292225074023)
--(axis cs:77,0.00084292225074023)
--(axis cs:78,0.00084292225074023)
--(axis cs:79,0.00084292225074023)
--(axis cs:80,0.00084292225074023)
--(axis cs:81,0.00084292225074023)
--(axis cs:82,0.00084292225074023)
--(axis cs:83,0.00084292225074023)
--(axis cs:84,0.000758859212510288)
--(axis cs:85,0.000758859212510288)
--(axis cs:86,0.000758859212510288)
--(axis cs:87,0.000758859212510288)
--(axis cs:88,0.000758859212510288)
--(axis cs:89,0.000758859212510288)
--(axis cs:90,0.000758859212510288)
--(axis cs:91,0.000758859212510288)
--(axis cs:92,0.000758859212510288)
--(axis cs:93,0.000758859212510288)
--(axis cs:94,0.000758859212510288)
--(axis cs:95,0.000758859212510288)
--(axis cs:96,0.000758859212510288)
--(axis cs:97,0.000758859212510288)
--(axis cs:98,0.000758859212510288)
--(axis cs:99,0.000758859212510288)
--(axis cs:100,0.000758859212510288)
--(axis cs:101,0.000758859212510288)
--(axis cs:102,0.000758859212510288)
--(axis cs:103,0.000702486839145422)
--(axis cs:104,0.000623468542471528)
--(axis cs:105,0.000623468542471528)
--(axis cs:106,0.000453645014204085)
--(axis cs:107,0.000453645014204085)
--(axis cs:108,0.000453645014204085)
--(axis cs:109,0.000446052697952837)
--(axis cs:110,0.000446052697952837)
--(axis cs:111,0.000446052697952837)
--(axis cs:111,0.00143328169360757)
--(axis cs:111,0.00143328169360757)
--(axis cs:110,0.00143328169360757)
--(axis cs:109,0.00143328169360757)
--(axis cs:108,0.00284439930692315)
--(axis cs:107,0.00284439930692315)
--(axis cs:106,0.00284439930692315)
--(axis cs:105,0.00300120911560953)
--(axis cs:104,0.00300120911560953)
--(axis cs:103,0.00309105077758431)
--(axis cs:102,0.00312903244048357)
--(axis cs:101,0.00312903244048357)
--(axis cs:100,0.00312903244048357)
--(axis cs:99,0.00312903244048357)
--(axis cs:98,0.00312903244048357)
--(axis cs:97,0.00312903244048357)
--(axis cs:96,0.00312903244048357)
--(axis cs:95,0.00312903244048357)
--(axis cs:94,0.00312903244048357)
--(axis cs:93,0.00312903244048357)
--(axis cs:92,0.00312903244048357)
--(axis cs:91,0.00312903244048357)
--(axis cs:90,0.00312903244048357)
--(axis cs:89,0.00312903244048357)
--(axis cs:88,0.00312903244048357)
--(axis cs:87,0.00312903244048357)
--(axis cs:86,0.00312903244048357)
--(axis cs:85,0.00312903244048357)
--(axis cs:84,0.00312903244048357)
--(axis cs:83,0.00323925074189901)
--(axis cs:82,0.00323925074189901)
--(axis cs:81,0.00323925074189901)
--(axis cs:80,0.00323925074189901)
--(axis cs:79,0.00323925074189901)
--(axis cs:78,0.00323925074189901)
--(axis cs:77,0.00323925074189901)
--(axis cs:76,0.00323925074189901)
--(axis cs:75,0.00323925074189901)
--(axis cs:74,0.00323925074189901)
--(axis cs:73,0.00323925074189901)
--(axis cs:72,0.00323925074189901)
--(axis cs:71,0.00323925074189901)
--(axis cs:70,0.00323925074189901)
--(axis cs:69,0.00323925074189901)
--(axis cs:68,0.00323925074189901)
--(axis cs:67,0.00323925074189901)
--(axis cs:66,0.00341134099289775)
--(axis cs:65,0.00341134099289775)
--(axis cs:64,0.00341134099289775)
--(axis cs:63,0.00341134099289775)
--(axis cs:62,0.00341134099289775)
--(axis cs:61,0.00341134099289775)
--(axis cs:60,0.00341134099289775)
--(axis cs:59,0.00341134099289775)
--(axis cs:58,0.00341134099289775)
--(axis cs:57,0.00341134099289775)
--(axis cs:56,0.00341134099289775)
--(axis cs:55,0.00341134099289775)
--(axis cs:54,0.00341134099289775)
--(axis cs:53,0.00341134099289775)
--(axis cs:52,0.00341134099289775)
--(axis cs:51,0.00341134099289775)
--(axis cs:50,0.00341134099289775)
--(axis cs:49,0.00341134099289775)
--(axis cs:48,0.00341156404465437)
--(axis cs:47,0.00341156404465437)
--(axis cs:46,0.00341156404465437)
--(axis cs:45,0.00341156404465437)
--(axis cs:44,0.00341156404465437)
--(axis cs:43,0.00341156404465437)
--(axis cs:42,0.00341156404465437)
--(axis cs:41,0.00341156404465437)
--(axis cs:40,0.00341156404465437)
--(axis cs:39,0.00341156404465437)
--(axis cs:38,0.00397645775228739)
--(axis cs:37,0.00397645775228739)
--(axis cs:36,0.00528170075267553)
--(axis cs:35,0.00528170075267553)
--(axis cs:34,0.00528170075267553)
--(axis cs:33,0.00528170075267553)
--(axis cs:32,0.00528170075267553)
--(axis cs:31,0.00600040284916759)
--(axis cs:30,0.00600040284916759)
--(axis cs:29,0.00686992285773158)
--(axis cs:28,0.00701544340699911)
--(axis cs:27,0.00701544340699911)
--(axis cs:26,0.00701544340699911)
--(axis cs:25,0.00805080216377974)
--(axis cs:24,0.00805080216377974)
--(axis cs:23,0.00976889301091433)
--(axis cs:22,0.0132379634305835)
--(axis cs:21,0.0155064621940255)
--(axis cs:20,0.0155064621940255)
--(axis cs:19,0.0168715231120586)
--(axis cs:18,0.0358583070337772)
--(axis cs:17,0.0430350936949253)
--(axis cs:16,0.0467922911047935)
--(axis cs:15,0.0476115345954895)
--(axis cs:14,0.0476115345954895)
--(axis cs:13,0.0541232563555241)
--(axis cs:12,0.0621579065918922)
--(axis cs:11,0.0712786093354225)
--(axis cs:10,0.0790769159793854)
--cycle;

\path [draw=color1, fill=color1, opacity=0.3]
(axis cs:10,0.0790769466867583)
--(axis cs:10,0.0426653284187181)
--(axis cs:11,0.0426653284187181)
--(axis cs:12,0.0366065214036576)
--(axis cs:13,0.0258265141716468)
--(axis cs:14,0.0222826737959492)
--(axis cs:15,0.0206309813074722)
--(axis cs:16,0.0127559092086345)
--(axis cs:17,0.0122310364093959)
--(axis cs:18,0.0122310364093959)
--(axis cs:19,0.0122310364093959)
--(axis cs:20,0.00804647764777953)
--(axis cs:21,0.00448538868384963)
--(axis cs:22,0.0043171108910826)
--(axis cs:23,0.00314323432367899)
--(axis cs:24,0.00314323432367899)
--(axis cs:25,0.00306713529382294)
--(axis cs:26,0.00306713529382294)
--(axis cs:27,0.00306713529382294)
--(axis cs:28,0.00255130903380553)
--(axis cs:29,0.000787602673643321)
--(axis cs:30,0.000787602673643321)
--(axis cs:31,0.000787602673643321)
--(axis cs:32,0.0007542862858961)
--(axis cs:33,0.000646912981981392)
--(axis cs:34,0.000620655500926645)
--(axis cs:35,0.000620655500926645)
--(axis cs:36,0.000614437743851618)
--(axis cs:37,0.000603781276955257)
--(axis cs:38,0.000603781276955257)
--(axis cs:39,0.000603781276955257)
--(axis cs:40,0.000541107949493316)
--(axis cs:41,0.000541107949493316)
--(axis cs:42,0.000541107949493316)
--(axis cs:43,0.00052614447540775)
--(axis cs:44,0.00052614447540775)
--(axis cs:45,0.00052614447540775)
--(axis cs:46,0.000379059865808188)
--(axis cs:47,0.000379059865808188)
--(axis cs:48,0.000241469207858456)
--(axis cs:49,0.000241469207858456)
--(axis cs:50,0.000241469207858456)
--(axis cs:51,0.000241469207858456)
--(axis cs:52,0.000240201934006592)
--(axis cs:53,0.000154154821633898)
--(axis cs:54,0.000153387189713388)
--(axis cs:55,0.000142720239862199)
--(axis cs:56,0.000142720239862199)
--(axis cs:57,0.000142720239862199)
--(axis cs:58,0.000142720239862199)
--(axis cs:59,0.000142720239862199)
--(axis cs:60,0.000134094171909123)
--(axis cs:61,0.000134094171909123)
--(axis cs:62,0.000134094171909123)
--(axis cs:63,0.000134094171909123)
--(axis cs:64,0.000134094171909123)
--(axis cs:65,0.000134094171909123)
--(axis cs:66,0.000134094171909123)
--(axis cs:67,0.000134094171909123)
--(axis cs:68,0.000134094171909123)
--(axis cs:69,0.000131643061196825)
--(axis cs:70,0.000131643061196825)
--(axis cs:71,0.000131643061196825)
--(axis cs:72,9.89319534660742e-05)
--(axis cs:73,7.42494120145719e-05)
--(axis cs:74,7.42494120145719e-05)
--(axis cs:75,7.42494120145719e-05)
--(axis cs:76,7.42494120145719e-05)
--(axis cs:77,7.42494120145719e-05)
--(axis cs:78,7.42494120145719e-05)
--(axis cs:79,7.42494120145719e-05)
--(axis cs:80,7.42494120145719e-05)
--(axis cs:81,7.42494120145719e-05)
--(axis cs:82,7.42494120145719e-05)
--(axis cs:83,7.42494120145719e-05)
--(axis cs:84,7.42494120145719e-05)
--(axis cs:85,7.42494120145719e-05)
--(axis cs:86,7.42494120145719e-05)
--(axis cs:87,7.42494120145719e-05)
--(axis cs:88,7.42494120145719e-05)
--(axis cs:89,7.42494120145719e-05)
--(axis cs:90,5.59583609431721e-05)
--(axis cs:91,5.59583609431721e-05)
--(axis cs:92,5.59583609431721e-05)
--(axis cs:93,5.59583609431721e-05)
--(axis cs:94,5.59583609431721e-05)
--(axis cs:95,5.59583609431721e-05)
--(axis cs:96,5.59583609431721e-05)
--(axis cs:97,5.59583609431721e-05)
--(axis cs:98,5.59583609431721e-05)
--(axis cs:99,5.59583609431721e-05)
--(axis cs:100,5.59583609431721e-05)
--(axis cs:101,5.59583609431721e-05)
--(axis cs:102,5.59583609431721e-05)
--(axis cs:103,5.59583609431721e-05)
--(axis cs:104,5.59583609431721e-05)
--(axis cs:105,5.59583609431721e-05)
--(axis cs:106,5.59583609431721e-05)
--(axis cs:107,5.59583609431721e-05)
--(axis cs:108,4.76085319037217e-05)
--(axis cs:109,4.76085319037217e-05)
--(axis cs:110,4.47739036237187e-05)
--(axis cs:111,4.1530988485192e-05)
--(axis cs:111,7.55602569372262e-05)
--(axis cs:111,7.55602569372262e-05)
--(axis cs:110,8.7986886255786e-05)
--(axis cs:109,9.62061868384751e-05)
--(axis cs:108,9.62061868384751e-05)
--(axis cs:107,0.000255498326965657)
--(axis cs:106,0.000255498326965657)
--(axis cs:105,0.000255498326965657)
--(axis cs:104,0.000255498326965657)
--(axis cs:103,0.000255498326965657)
--(axis cs:102,0.000255498326965657)
--(axis cs:101,0.000255498326965657)
--(axis cs:100,0.000255498326965657)
--(axis cs:99,0.000255498326965657)
--(axis cs:98,0.000255498326965657)
--(axis cs:97,0.000255498326965657)
--(axis cs:96,0.000255498326965657)
--(axis cs:95,0.000255498326965657)
--(axis cs:94,0.000255498326965657)
--(axis cs:93,0.000255498326965657)
--(axis cs:92,0.000255498326965657)
--(axis cs:91,0.000255498326965657)
--(axis cs:90,0.000255498326965657)
--(axis cs:89,0.00027444353264393)
--(axis cs:88,0.00027444353264393)
--(axis cs:87,0.00027444353264393)
--(axis cs:86,0.00027444353264393)
--(axis cs:85,0.00027444353264393)
--(axis cs:84,0.00027444353264393)
--(axis cs:83,0.00027444353264393)
--(axis cs:82,0.00027444353264393)
--(axis cs:81,0.00027444353264393)
--(axis cs:80,0.00027444353264393)
--(axis cs:79,0.00027444353264393)
--(axis cs:78,0.00027444353264393)
--(axis cs:77,0.00027444353264393)
--(axis cs:76,0.00027444353264393)
--(axis cs:75,0.00027444353264393)
--(axis cs:74,0.00027444353264393)
--(axis cs:73,0.00027444353264393)
--(axis cs:72,0.000296502855627968)
--(axis cs:71,0.000341271823227255)
--(axis cs:70,0.000341271823227255)
--(axis cs:69,0.000341271823227255)
--(axis cs:68,0.000342704137234643)
--(axis cs:67,0.000342704137234643)
--(axis cs:66,0.000342704137234643)
--(axis cs:65,0.000342704137234643)
--(axis cs:64,0.000342704137234643)
--(axis cs:63,0.000342704137234643)
--(axis cs:62,0.000342704137234643)
--(axis cs:61,0.000342704137234643)
--(axis cs:60,0.000342704137234643)
--(axis cs:59,0.000352475994211149)
--(axis cs:58,0.000352475994211149)
--(axis cs:57,0.000352475994211149)
--(axis cs:56,0.000352475994211149)
--(axis cs:55,0.000352475994211149)
--(axis cs:54,0.000563721905944698)
--(axis cs:53,0.000564375136512799)
--(axis cs:52,0.000644926412703582)
--(axis cs:51,0.000672877725173484)
--(axis cs:50,0.000672877725173484)
--(axis cs:49,0.000672877725173484)
--(axis cs:48,0.000672877725173484)
--(axis cs:47,0.000876772494207674)
--(axis cs:46,0.000876772494207674)
--(axis cs:45,0.00101015814922392)
--(axis cs:44,0.00101015814922392)
--(axis cs:43,0.00101015814922392)
--(axis cs:42,0.00103377327684688)
--(axis cs:41,0.00103377327684688)
--(axis cs:40,0.00103377327684688)
--(axis cs:39,0.00118117159091369)
--(axis cs:38,0.00118117159091369)
--(axis cs:37,0.00118117159091369)
--(axis cs:36,0.00123087247364767)
--(axis cs:35,0.00123505499877376)
--(axis cs:34,0.00123505499877376)
--(axis cs:33,0.00125441198048048)
--(axis cs:32,0.00235982664321597)
--(axis cs:31,0.00238506234548101)
--(axis cs:30,0.00238506234548101)
--(axis cs:29,0.00238506234548101)
--(axis cs:28,0.00570234203511936)
--(axis cs:27,0.00603560612306827)
--(axis cs:26,0.00603560612306827)
--(axis cs:25,0.00603560612306827)
--(axis cs:24,0.00606643733144017)
--(axis cs:23,0.00606643733144017)
--(axis cs:22,0.00721410594548578)
--(axis cs:21,0.00792332021945868)
--(axis cs:20,0.0143029908889599)
--(axis cs:19,0.0191666455511421)
--(axis cs:18,0.0191666455511421)
--(axis cs:17,0.0191666455511421)
--(axis cs:16,0.019725936080997)
--(axis cs:15,0.0434598385226567)
--(axis cs:14,0.0492237266633668)
--(axis cs:13,0.0515142366742416)
--(axis cs:12,0.0736207647046327)
--(axis cs:11,0.0790769466867583)
--(axis cs:10,0.0790769466867583)
--cycle;

\addplot [semithick, color1]
table {%
10 0.0608711375527382
11 0.0486051472708307
12 0.0486051472708307
13 0.0486051472708307
14 0.0433137394898178
15 0.0303607146343687
16 0.0237046113735475
17 0.0173684511466334
18 0.0173684511466334
19 0.0173684511466334
20 0.0173684511466334
21 0.0173684511466334
22 0.0173684511466334
23 0.0130647043370071
24 0.0104203894490676
25 0.00872319818807413
26 0.00601524940147344
27 0.00461796667214873
28 0.00461796667214873
29 0.00349006861086201
30 0.00292366357891466
31 0.00182006355126441
32 0.000818759748054371
33 0.000607973889142577
34 0.000607973889142577
35 0.000607973889142577
36 0.000607973889142577
37 0.000587956836112291
38 0.000587956836112291
39 0.000587956836112291
40 0.000587956836112291
41 0.000585225986264176
42 0.000502448886354534
43 0.000376678436103506
44 0.000335513514203356
45 0.000328640255242124
46 0.000328640255242124
47 0.00020691194026709
48 0.00017867854158693
49 0.00017867854158693
50 0.00017867854158693
51 0.000151633653249911
52 0.000120869022841197
53 0.000120869022841197
54 0.000120869022841197
55 0.000120869022841197
56 0.000114713660041482
57 0.000114713660041482
58 0.000114713660041482
59 0.000106271904674965
60 0.000106271904674965
61 0.000106271904674965
62 0.000106271904674965
63 0.000106271904674965
64 0.000106271904674965
65 0.00010305238815031
66 0.00010305238815031
67 9.21177520250283e-05
68 9.21177520250283e-05
69 7.44564535905545e-05
70 7.44564535905545e-05
71 7.28939017629049e-05
72 7.28939017629049e-05
73 7.28939017629049e-05
74 7.28939017629049e-05
75 7.28939017629049e-05
76 7.28939017629049e-05
77 6.76567390628102e-05
78 6.09633648690489e-05
79 6.09633648690489e-05
80 6.09633648690489e-05
81 6.09633648690489e-05
82 6.09633648690489e-05
83 6.09633648690489e-05
84 6.09633648690489e-05
85 6.09633648690489e-05
86 6.09633648690489e-05
87 6.09633648690489e-05
88 6.09633648690489e-05
89 6.09633648690489e-05
90 5.57562829040514e-05
91 5.57562829040514e-05
92 5.57562829040514e-05
93 5.57562829040514e-05
94 5.57562829040514e-05
95 5.57562829040514e-05
96 5.57562829040514e-05
97 5.57562829040514e-05
98 5.57562829040514e-05
99 5.57562829040514e-05
100 5.57562829040514e-05
101 5.57562829040514e-05
102 4.75168514859348e-05
103 4.75168514859348e-05
104 4.75168514859348e-05
105 4.75168514859348e-05
106 4.75168514859348e-05
107 4.75168514859348e-05
108 4.75168514859348e-05
109 4.75168514859348e-05
110 4.75168514859348e-05
111 3.47517160409483e-05
};
\addplot [semithick, blue]
table {%
10 0.0608711019158363
11 0.0608711019158363
12 0.0608711019158363
13 0.0608711019158363
14 0.0602229982614517
15 0.0602229982614517
16 0.0602229982614517
17 0.0602229982614517
18 0.0376800075173378
19 0.0376800075173378
20 0.0376800075173378
21 0.0376800075173378
22 0.0348234996199608
23 0.032159723341465
24 0.032159723341465
25 0.032159723341465
26 0.032159723341465
27 0.0257517322897911
28 0.0257517322897911
29 0.0255705788731575
30 0.0251449570059776
31 0.0152300745248795
32 0.0152300745248795
33 0.0152300745248795
34 0.0152300745248795
35 0.0140058249235153
36 0.0136474072933197
37 0.0136474072933197
38 0.0136474072933197
39 0.0136474072933197
40 0.0136474072933197
41 0.0136474072933197
42 0.0136474072933197
43 0.0131167769432068
44 0.0131167769432068
45 0.0131167769432068
46 0.0131167769432068
47 0.0124807804822922
48 0.0124807804822922
49 0.0124807804822922
50 0.0124807804822922
51 0.0105555057525635
52 0.0103725641965866
53 0.0102215558290482
54 0.0102215558290482
55 0.0102215558290482
56 0.0102215558290482
57 0.0102215558290482
58 0.00948622822761536
59 0.00948622822761536
60 0.00948622822761536
61 0.00730392336845398
62 0.0055205225944519
63 0.00438863039016724
64 0.00438863039016724
65 0.00438863039016724
66 0.00438863039016724
67 0.00438863039016724
68 0.00438863039016724
69 0.00384530425071716
70 0.00384530425071716
71 0.00377430021762848
72 0.00377430021762848
73 0.00377430021762848
74 0.0035899430513382
75 0.0035899430513382
76 0.0035899430513382
77 0.0035899430513382
78 0.0035899430513382
79 0.0035899430513382
80 0.0035899430513382
81 0.0035899430513382
82 0.0035899430513382
83 0.0035899430513382
84 0.00303067266941071
85 0.00303067266941071
86 0.00303067266941071
87 0.00303067266941071
88 0.00303067266941071
89 0.00303067266941071
90 0.00290912389755249
91 0.00287215411663055
92 0.00287215411663055
93 0.00287215411663055
94 0.00287215411663055
95 0.00287215411663055
96 0.00287215411663055
97 0.00287215411663055
98 0.00287215411663055
99 0.00270314514636993
100 0.00270314514636993
101 0.00270314514636993
102 0.00270314514636993
103 0.00270314514636993
104 0.00270314514636993
105 0.00270314514636993
106 0.00270314514636993
107 0.00270314514636993
108 0.00270314514636993
109 0.00270314514636993
110 0.00270314514636993
111 0.00270314514636993
};
\addplot [semithick, blue, dash dot]
table {%
10 0.0608711019158363
11 0.052274651825428
12 0.0440932586789131
13 0.0370135083794594
14 0.0321808084845543
15 0.0321808084845543
16 0.0312531068921089
17 0.028334952890873
18 0.0244578942656517
19 0.0139787644147873
20 0.0125353187322617
21 0.0125353187322617
22 0.0107120275497437
23 0.00834637880325317
24 0.00652255117893219
25 0.00652255117893219
26 0.00527504086494446
27 0.00527504086494446
28 0.00527504086494446
29 0.00506435334682465
30 0.00416038930416107
31 0.00416038930416107
32 0.0035366415977478
33 0.0035366415977478
34 0.0035366415977478
35 0.0035366415977478
36 0.0035366415977478
37 0.00271038711071014
38 0.00271038711071014
39 0.00223173201084137
40 0.00223173201084137
41 0.00223173201084137
42 0.00223173201084137
43 0.00223173201084137
44 0.00223173201084137
45 0.00223173201084137
46 0.00223173201084137
47 0.00223173201084137
48 0.00223173201084137
49 0.00223144888877869
50 0.00223144888877869
51 0.00223144888877869
52 0.00223144888877869
53 0.00223144888877869
54 0.00223144888877869
55 0.00223144888877869
56 0.00223144888877869
57 0.00223144888877869
58 0.00223144888877869
59 0.00223144888877869
60 0.00223144888877869
61 0.00223144888877869
62 0.00223144888877869
63 0.00223144888877869
64 0.00223144888877869
65 0.00223144888877869
66 0.00223144888877869
67 0.00204108655452728
68 0.00204108655452728
69 0.00204108655452728
70 0.00204108655452728
71 0.00204108655452728
72 0.00204108655452728
73 0.00204108655452728
74 0.00204108655452728
75 0.00204108655452728
76 0.00204108655452728
77 0.00204108655452728
78 0.00204108655452728
79 0.00204108655452728
80 0.00204108655452728
81 0.00204108655452728
82 0.00204108655452728
83 0.00204108655452728
84 0.00194394588470459
85 0.00194394588470459
86 0.00194394588470459
87 0.00194394588470459
88 0.00194394588470459
89 0.00194394588470459
90 0.00194394588470459
91 0.00194394588470459
92 0.00194394588470459
93 0.00194394588470459
94 0.00194394588470459
95 0.00194394588470459
96 0.00194394588470459
97 0.00194394588470459
98 0.00194394588470459
99 0.00194394588470459
100 0.00194394588470459
101 0.00194394588470459
102 0.00194394588470459
103 0.00189676880836487
104 0.00181233882904053
105 0.00181233882904053
106 0.00164902210235596
107 0.00164902210235596
108 0.00164902210235596
109 0.000939667224884033
110 0.000939667224884033
111 0.000939667224884033
};
\addplot [semithick, color1, dash dot]
table {%
10 0.0608711375527382
11 0.0608711375527382
12 0.0551136430541452
13 0.0386703754229442
14 0.035753200229658
15 0.0320454099150645
16 0.0162409226448158
17 0.015698840980269
18 0.015698840980269
19 0.015698840980269
20 0.0111747342683697
21 0.00620435445165415
22 0.00576560841828419
23 0.00460483582755958
24 0.00460483582755958
25 0.00455137070844561
26 0.00455137070844561
27 0.00455137070844561
28 0.00412682553446245
29 0.00158633250956217
30 0.00158633250956217
31 0.00158633250956217
32 0.00155705646455603
33 0.000950662481230935
34 0.000927855249850201
35 0.000927855249850201
36 0.000922655108749643
37 0.000892476433934475
38 0.000892476433934475
39 0.000892476433934475
40 0.000787440613170098
41 0.000787440613170098
42 0.000787440613170098
43 0.000768151312315835
44 0.000768151312315835
45 0.000768151312315835
46 0.000627916180007931
47 0.000627916180007931
48 0.00045717346651597
49 0.00045717346651597
50 0.00045717346651597
51 0.00045717346651597
52 0.000442564173355087
53 0.000359264979073348
54 0.000358554547829043
55 0.000247598117036674
56 0.000247598117036674
57 0.000247598117036674
58 0.000247598117036674
59 0.000247598117036674
60 0.000238399154571883
61 0.000238399154571883
62 0.000238399154571883
63 0.000238399154571883
64 0.000238399154571883
65 0.000238399154571883
66 0.000238399154571883
67 0.000238399154571883
68 0.000238399154571883
69 0.00023645744221204
70 0.00023645744221204
71 0.00023645744221204
72 0.000197717404547021
73 0.000174346472329251
74 0.000174346472329251
75 0.000174346472329251
76 0.000174346472329251
77 0.000174346472329251
78 0.000174346472329251
79 0.000174346472329251
80 0.000174346472329251
81 0.000174346472329251
82 0.000174346472329251
83 0.000174346472329251
84 0.000174346472329251
85 0.000174346472329251
86 0.000174346472329251
87 0.000174346472329251
88 0.000174346472329251
89 0.000174346472329251
90 0.000155728343954414
91 0.000155728343954414
92 0.000155728343954414
93 0.000155728343954414
94 0.000155728343954414
95 0.000155728343954414
96 0.000155728343954414
97 0.000155728343954414
98 0.000155728343954414
99 0.000155728343954414
100 0.000155728343954414
101 0.000155728343954414
102 0.000155728343954414
103 0.000155728343954414
104 0.000155728343954414
105 0.000155728343954414
106 0.000155728343954414
107 0.000155728343954414
108 7.19073593710984e-05
109 7.19073593710984e-05
110 6.63803949397523e-05
111 5.85456227112091e-05
};
\end{axis}

\end{tikzpicture}

%% file: figures/bop_2d_ackley_strong.tex
\begin{tikzpicture}

\definecolor{color0}{rgb}{0,0,1}
\definecolor{color1}{rgb}{1,0.549019607843137,0}
\definecolor{color2}{rgb}{1,0.647058823529412,0}
\definecolor{color3}{rgb}{0.564705882352941,0.933333333333333,0.564705882352941}

\begin{axis}[axis on top,
enlarge x limits=false,
enlarge y limits=false,
height=\figureheight,
scale only axis,
tick align=outside,
tick pos=left,
tick pos=left,
width=\figurewidth,
xlabel={Iteration},
xmin=10, xmax=100,
xtick style={color=black},
xtick={-10,0,10,25,50,75,100},
xticklabels={\ensuremath{-}10,0,10,25,50,75,90},
ymin=-0.25, ymax=4.5,
ytick style={color=black},
ytick={0.   , 4.5},
]
\node[anchor=north east] at (rel axis cs:1,1) {Ackley 2D (strong)};
\path [draw=blue, fill=blue, opacity=0.3]
(axis cs:10,5.34454870223999)
--(axis cs:10,4.39542245864868)
--(axis cs:11,3.88091397285461)
--(axis cs:12,3.1576292514801)
--(axis cs:13,2.74383449554443)
--(axis cs:14,2.33305978775024)
--(axis cs:15,2.31176948547363)
--(axis cs:16,1.97702074050903)
--(axis cs:17,1.83676362037659)
--(axis cs:18,1.65795373916626)
--(axis cs:19,1.51583170890808)
--(axis cs:20,1.3749315738678)
--(axis cs:21,1.2121490240097)
--(axis cs:22,1.15717649459839)
--(axis cs:23,0.891646444797516)
--(axis cs:24,0.891646444797516)
--(axis cs:25,0.822120785713196)
--(axis cs:26,0.723312199115753)
--(axis cs:27,0.682594299316406)
--(axis cs:28,0.682594299316406)
--(axis cs:29,0.668160617351532)
--(axis cs:30,0.668160617351532)
--(axis cs:31,0.645785093307495)
--(axis cs:32,0.450593054294586)
--(axis cs:33,0.450593054294586)
--(axis cs:34,0.442295640707016)
--(axis cs:35,0.442295640707016)
--(axis cs:36,0.433308303356171)
--(axis cs:37,0.433308303356171)
--(axis cs:38,0.412064731121063)
--(axis cs:39,0.397277146577835)
--(axis cs:40,0.342329561710358)
--(axis cs:41,0.342329561710358)
--(axis cs:42,0.331745445728302)
--(axis cs:43,0.294486403465271)
--(axis cs:44,0.294486403465271)
--(axis cs:45,0.223338931798935)
--(axis cs:46,0.214090541005135)
--(axis cs:47,0.214090541005135)
--(axis cs:48,0.214090541005135)
--(axis cs:49,0.214090541005135)
--(axis cs:50,0.214090541005135)
--(axis cs:51,0.214090541005135)
--(axis cs:52,0.210031047463417)
--(axis cs:53,0.210031047463417)
--(axis cs:54,0.210031047463417)
--(axis cs:55,0.210031047463417)
--(axis cs:56,0.210031047463417)
--(axis cs:57,0.210031047463417)
--(axis cs:58,0.210031047463417)
--(axis cs:59,0.187263935804367)
--(axis cs:60,0.187263935804367)
--(axis cs:61,0.187263935804367)
--(axis cs:62,0.187263935804367)
--(axis cs:63,0.187263935804367)
--(axis cs:64,0.187263935804367)
--(axis cs:65,0.187263935804367)
--(axis cs:66,0.174298867583275)
--(axis cs:67,0.174298867583275)
--(axis cs:68,0.174298867583275)
--(axis cs:69,0.174298867583275)
--(axis cs:70,0.174298867583275)
--(axis cs:71,0.174298867583275)
--(axis cs:72,0.174298867583275)
--(axis cs:73,0.167514115571976)
--(axis cs:74,0.165840819478035)
--(axis cs:75,0.143277704715729)
--(axis cs:76,0.143277704715729)
--(axis cs:77,0.143277704715729)
--(axis cs:78,0.143277704715729)
--(axis cs:79,0.143277704715729)
--(axis cs:80,0.143277704715729)
--(axis cs:81,0.143277704715729)
--(axis cs:82,0.116174265742302)
--(axis cs:83,0.116174265742302)
--(axis cs:84,0.116174265742302)
--(axis cs:85,0.116174265742302)
--(axis cs:86,0.116174265742302)
--(axis cs:87,0.116174265742302)
--(axis cs:88,0.116174265742302)
--(axis cs:89,0.116174265742302)
--(axis cs:90,0.116174265742302)
--(axis cs:91,0.116174265742302)
--(axis cs:92,0.116174265742302)
--(axis cs:93,0.116174265742302)
--(axis cs:94,0.116174265742302)
--(axis cs:95,0.116174265742302)
--(axis cs:96,0.116174265742302)
--(axis cs:97,0.116174265742302)
--(axis cs:98,0.116174265742302)
--(axis cs:99,0.116174265742302)
--(axis cs:100,0.116174265742302)
--(axis cs:101,0.116174265742302)
--(axis cs:102,0.10993529856205)
--(axis cs:103,0.0883582085371017)
--(axis cs:104,0.0883582085371017)
--(axis cs:105,0.0883582085371017)
--(axis cs:106,0.0883582085371017)
--(axis cs:107,0.0883582085371017)
--(axis cs:108,0.0883582085371017)
--(axis cs:109,0.0883582085371017)
--(axis cs:110,0.0883582085371017)
--(axis cs:111,0.0883582085371017)
--(axis cs:111,0.116655826568604)
--(axis cs:111,0.116655826568604)
--(axis cs:110,0.116655826568604)
--(axis cs:109,0.116655826568604)
--(axis cs:108,0.116655826568604)
--(axis cs:107,0.116655826568604)
--(axis cs:106,0.116655826568604)
--(axis cs:105,0.116655826568604)
--(axis cs:104,0.116655826568604)
--(axis cs:103,0.116655826568604)
--(axis cs:102,0.1765026897192)
--(axis cs:101,0.183400020003319)
--(axis cs:100,0.183400020003319)
--(axis cs:99,0.183400020003319)
--(axis cs:98,0.183400020003319)
--(axis cs:97,0.183400020003319)
--(axis cs:96,0.183400020003319)
--(axis cs:95,0.183400020003319)
--(axis cs:94,0.183400020003319)
--(axis cs:93,0.183400020003319)
--(axis cs:92,0.183400020003319)
--(axis cs:91,0.183400020003319)
--(axis cs:90,0.183400020003319)
--(axis cs:89,0.183400020003319)
--(axis cs:88,0.183400020003319)
--(axis cs:87,0.183400020003319)
--(axis cs:86,0.183400020003319)
--(axis cs:85,0.183400020003319)
--(axis cs:84,0.183400020003319)
--(axis cs:83,0.183400020003319)
--(axis cs:82,0.183400020003319)
--(axis cs:81,0.216449767351151)
--(axis cs:80,0.216449767351151)
--(axis cs:79,0.216449767351151)
--(axis cs:78,0.216449767351151)
--(axis cs:77,0.216449767351151)
--(axis cs:76,0.216449767351151)
--(axis cs:75,0.216449767351151)
--(axis cs:74,0.236724182963371)
--(axis cs:73,0.238404780626297)
--(axis cs:72,0.24199141561985)
--(axis cs:71,0.24199141561985)
--(axis cs:70,0.24199141561985)
--(axis cs:69,0.24199141561985)
--(axis cs:68,0.24199141561985)
--(axis cs:67,0.24199141561985)
--(axis cs:66,0.24199141561985)
--(axis cs:65,0.275599211454391)
--(axis cs:64,0.275599211454391)
--(axis cs:63,0.275599211454391)
--(axis cs:62,0.275599211454391)
--(axis cs:61,0.275599211454391)
--(axis cs:60,0.275599211454391)
--(axis cs:59,0.275599211454391)
--(axis cs:58,0.301079630851746)
--(axis cs:57,0.301079630851746)
--(axis cs:56,0.301079630851746)
--(axis cs:55,0.301079630851746)
--(axis cs:54,0.301079630851746)
--(axis cs:53,0.301079630851746)
--(axis cs:52,0.301079630851746)
--(axis cs:51,0.312937378883362)
--(axis cs:50,0.312937378883362)
--(axis cs:49,0.312937378883362)
--(axis cs:48,0.312937378883362)
--(axis cs:47,0.312937378883362)
--(axis cs:46,0.312937378883362)
--(axis cs:45,0.367740780115128)
--(axis cs:44,0.440508842468262)
--(axis cs:43,0.440508842468262)
--(axis cs:42,0.477627277374268)
--(axis cs:41,0.483801126480103)
--(axis cs:40,0.483801126480103)
--(axis cs:39,0.57536917924881)
--(axis cs:38,0.592392385005951)
--(axis cs:37,0.610099852085114)
--(axis cs:36,0.610099852085114)
--(axis cs:35,0.617206752300262)
--(axis cs:34,0.617206752300262)
--(axis cs:33,0.625095903873444)
--(axis cs:32,0.625095903873444)
--(axis cs:31,0.805643081665039)
--(axis cs:30,0.823435604572296)
--(axis cs:29,0.823435604572296)
--(axis cs:28,0.960858941078186)
--(axis cs:27,0.960858941078186)
--(axis cs:26,1.03337490558624)
--(axis cs:25,1.40919029712677)
--(axis cs:24,1.47080183029175)
--(axis cs:23,1.47080183029175)
--(axis cs:22,1.8884334564209)
--(axis cs:21,1.92933881282806)
--(axis cs:20,2.0491406917572)
--(axis cs:19,2.26987075805664)
--(axis cs:18,2.36845779418945)
--(axis cs:17,2.55872750282288)
--(axis cs:16,2.68847513198853)
--(axis cs:15,2.93001747131348)
--(axis cs:14,2.97110223770142)
--(axis cs:13,3.54079532623291)
--(axis cs:12,3.73759388923645)
--(axis cs:11,4.6047306060791)
--(axis cs:10,5.34454870223999)
--cycle;

\path [draw=color1, fill=color1, opacity=0.3]
(axis cs:10,5.34454898518408)
--(axis cs:10,4.39542274790918)
--(axis cs:11,3.41045067881566)
--(axis cs:12,2.7454080206527)
--(axis cs:13,1.9267474489697)
--(axis cs:14,1.73488600735811)
--(axis cs:15,1.57892945436848)
--(axis cs:16,1.42673877662146)
--(axis cs:17,1.29529392032417)
--(axis cs:18,1.23355881429876)
--(axis cs:19,1.19355978010195)
--(axis cs:20,1.11075406521434)
--(axis cs:21,1.00310591558059)
--(axis cs:22,0.987451362734366)
--(axis cs:23,0.945369972182245)
--(axis cs:24,0.898182650255731)
--(axis cs:25,0.839965539411208)
--(axis cs:26,0.760526038238884)
--(axis cs:27,0.760526038238884)
--(axis cs:28,0.661928837542589)
--(axis cs:29,0.533617815638524)
--(axis cs:30,0.533617815638524)
--(axis cs:31,0.533617815638524)
--(axis cs:32,0.450356667609864)
--(axis cs:33,0.450356667609864)
--(axis cs:34,0.450356667609864)
--(axis cs:35,0.381022665058909)
--(axis cs:36,0.332232589722052)
--(axis cs:37,0.332232589722052)
--(axis cs:38,0.240466944059575)
--(axis cs:39,0.240466944059575)
--(axis cs:40,0.240466944059575)
--(axis cs:41,0.240466944059575)
--(axis cs:42,0.240466944059575)
--(axis cs:43,0.19049513643236)
--(axis cs:44,0.19049513643236)
--(axis cs:45,0.19049513643236)
--(axis cs:46,0.19049513643236)
--(axis cs:47,0.19049513643236)
--(axis cs:48,0.19049513643236)
--(axis cs:49,0.19049513643236)
--(axis cs:50,0.19049513643236)
--(axis cs:51,0.19049513643236)
--(axis cs:52,0.19049513643236)
--(axis cs:53,0.19049513643236)
--(axis cs:54,0.19049513643236)
--(axis cs:55,0.175666736552703)
--(axis cs:56,0.175666736552703)
--(axis cs:57,0.175666736552703)
--(axis cs:58,0.172734619179699)
--(axis cs:59,0.172734619179699)
--(axis cs:60,0.162323561837829)
--(axis cs:61,0.162323561837829)
--(axis cs:62,0.162323561837829)
--(axis cs:63,0.162323561837829)
--(axis cs:64,0.162323561837829)
--(axis cs:65,0.162323561837829)
--(axis cs:66,0.162323561837829)
--(axis cs:67,0.162323561837829)
--(axis cs:68,0.162323561837829)
--(axis cs:69,0.162323561837829)
--(axis cs:70,0.162323561837829)
--(axis cs:71,0.162323561837829)
--(axis cs:72,0.162323561837829)
--(axis cs:73,0.159032792250397)
--(axis cs:74,0.135353266078631)
--(axis cs:75,0.135353266078631)
--(axis cs:76,0.132292925764049)
--(axis cs:77,0.132292925764049)
--(axis cs:78,0.132292925764049)
--(axis cs:79,0.132292925764049)
--(axis cs:80,0.132292925764049)
--(axis cs:81,0.132292925764049)
--(axis cs:82,0.132292925764049)
--(axis cs:83,0.132292925764049)
--(axis cs:84,0.132292925764049)
--(axis cs:85,0.132292925764049)
--(axis cs:86,0.124850076888104)
--(axis cs:87,0.124850076888104)
--(axis cs:88,0.124850076888104)
--(axis cs:89,0.114276672659087)
--(axis cs:90,0.0939117162034636)
--(axis cs:91,0.0939117162034636)
--(axis cs:92,0.0939117162034636)
--(axis cs:93,0.0939117162034636)
--(axis cs:94,0.0939117162034636)
--(axis cs:95,0.0939117162034636)
--(axis cs:96,0.0939117162034636)
--(axis cs:97,0.0939117162034636)
--(axis cs:98,0.0939117162034636)
--(axis cs:99,0.0939117162034636)
--(axis cs:100,0.0939117162034636)
--(axis cs:101,0.0939117162034636)
--(axis cs:102,0.0939117162034636)
--(axis cs:103,0.0939117162034636)
--(axis cs:104,0.0939117162034636)
--(axis cs:105,0.0939117162034636)
--(axis cs:106,0.0939117162034636)
--(axis cs:107,0.0939117162034636)
--(axis cs:108,0.0939117162034636)
--(axis cs:109,0.0939117162034636)
--(axis cs:110,0.0939117162034636)
--(axis cs:111,0.0939117162034636)
--(axis cs:111,0.13707336023586)
--(axis cs:111,0.13707336023586)
--(axis cs:110,0.13707336023586)
--(axis cs:109,0.13707336023586)
--(axis cs:108,0.13707336023586)
--(axis cs:107,0.13707336023586)
--(axis cs:106,0.13707336023586)
--(axis cs:105,0.13707336023586)
--(axis cs:104,0.13707336023586)
--(axis cs:103,0.13707336023586)
--(axis cs:102,0.13707336023586)
--(axis cs:101,0.13707336023586)
--(axis cs:100,0.13707336023586)
--(axis cs:99,0.13707336023586)
--(axis cs:98,0.13707336023586)
--(axis cs:97,0.13707336023586)
--(axis cs:96,0.13707336023586)
--(axis cs:95,0.13707336023586)
--(axis cs:94,0.13707336023586)
--(axis cs:93,0.13707336023586)
--(axis cs:92,0.13707336023586)
--(axis cs:91,0.13707336023586)
--(axis cs:90,0.13707336023586)
--(axis cs:89,0.164068277867525)
--(axis cs:88,0.184006700722958)
--(axis cs:87,0.184006700722958)
--(axis cs:86,0.184006700722958)
--(axis cs:85,0.192543153249833)
--(axis cs:84,0.192543153249833)
--(axis cs:83,0.192543153249833)
--(axis cs:82,0.192543153249833)
--(axis cs:81,0.192543153249833)
--(axis cs:80,0.192543153249833)
--(axis cs:79,0.192543153249833)
--(axis cs:78,0.192543153249833)
--(axis cs:77,0.192543153249833)
--(axis cs:76,0.192543153249833)
--(axis cs:75,0.201508004739374)
--(axis cs:74,0.201508004739374)
--(axis cs:73,0.274910016718368)
--(axis cs:72,0.278732497622703)
--(axis cs:71,0.278732497622703)
--(axis cs:70,0.278732497622703)
--(axis cs:69,0.278732497622703)
--(axis cs:68,0.278732497622703)
--(axis cs:67,0.278732497622703)
--(axis cs:66,0.278732497622703)
--(axis cs:65,0.278732497622703)
--(axis cs:64,0.278732497622703)
--(axis cs:63,0.278732497622703)
--(axis cs:62,0.278732497622703)
--(axis cs:61,0.278732497622703)
--(axis cs:60,0.278732497622703)
--(axis cs:59,0.296509202899174)
--(axis cs:58,0.296509202899174)
--(axis cs:57,0.298917338778298)
--(axis cs:56,0.298917338778298)
--(axis cs:55,0.298917338778298)
--(axis cs:54,0.35009775520927)
--(axis cs:53,0.35009775520927)
--(axis cs:52,0.35009775520927)
--(axis cs:51,0.35009775520927)
--(axis cs:50,0.35009775520927)
--(axis cs:49,0.35009775520927)
--(axis cs:48,0.35009775520927)
--(axis cs:47,0.35009775520927)
--(axis cs:46,0.35009775520927)
--(axis cs:45,0.35009775520927)
--(axis cs:44,0.35009775520927)
--(axis cs:43,0.35009775520927)
--(axis cs:42,0.441057238854729)
--(axis cs:41,0.441057238854729)
--(axis cs:40,0.441057238854729)
--(axis cs:39,0.441057238854729)
--(axis cs:38,0.441057238854729)
--(axis cs:37,0.677173908611807)
--(axis cs:36,0.677173908611807)
--(axis cs:35,0.721637174291168)
--(axis cs:34,0.788454145104753)
--(axis cs:33,0.788454145104753)
--(axis cs:32,0.788454145104753)
--(axis cs:31,0.922600193997987)
--(axis cs:30,0.922600193997987)
--(axis cs:29,0.922600193997987)
--(axis cs:28,1.02206415459087)
--(axis cs:27,1.10204259145367)
--(axis cs:26,1.10204259145367)
--(axis cs:25,1.15850326376631)
--(axis cs:24,1.25820972793353)
--(axis cs:23,1.40706690527927)
--(axis cs:22,1.48963495472051)
--(axis cs:21,1.5037373410628)
--(axis cs:20,1.56785827862184)
--(axis cs:19,1.67320633111403)
--(axis cs:18,1.78289684888948)
--(axis cs:17,1.87606344358266)
--(axis cs:16,2.03773432776584)
--(axis cs:15,2.15844408003502)
--(axis cs:14,2.34703409760354)
--(axis cs:13,2.56926883824648)
--(axis cs:12,3.16822886501207)
--(axis cs:11,4.14606812271821)
--(axis cs:10,5.34454898518408)
--cycle;

\path [draw=blue, fill=blue, opacity=0.3]
(axis cs:10,5.34454870223999)
--(axis cs:10,4.39542245864868)
--(axis cs:11,3.25274872779846)
--(axis cs:12,3.20437002182007)
--(axis cs:13,2.74756813049316)
--(axis cs:14,2.65860295295715)
--(axis cs:15,2.5920078754425)
--(axis cs:16,2.09870982170105)
--(axis cs:17,2.00992155075073)
--(axis cs:18,1.56103312969208)
--(axis cs:19,1.56103312969208)
--(axis cs:20,1.43367409706116)
--(axis cs:21,1.36366438865662)
--(axis cs:22,1.36366438865662)
--(axis cs:23,1.20738458633423)
--(axis cs:24,1.20738458633423)
--(axis cs:25,1.20738458633423)
--(axis cs:26,1.14181280136108)
--(axis cs:27,0.956552922725677)
--(axis cs:28,0.902479708194733)
--(axis cs:29,0.902479708194733)
--(axis cs:30,0.83898138999939)
--(axis cs:31,0.83898138999939)
--(axis cs:32,0.782922744750977)
--(axis cs:33,0.782922744750977)
--(axis cs:34,0.782922744750977)
--(axis cs:35,0.699203252792358)
--(axis cs:36,0.699203252792358)
--(axis cs:37,0.699203252792358)
--(axis cs:38,0.699203252792358)
--(axis cs:39,0.699203252792358)
--(axis cs:40,0.699203252792358)
--(axis cs:41,0.699203252792358)
--(axis cs:42,0.670929312705994)
--(axis cs:43,0.670929312705994)
--(axis cs:44,0.670929312705994)
--(axis cs:45,0.670929312705994)
--(axis cs:46,0.670929312705994)
--(axis cs:47,0.670929312705994)
--(axis cs:48,0.670929312705994)
--(axis cs:49,0.670929312705994)
--(axis cs:50,0.670929312705994)
--(axis cs:51,0.670929312705994)
--(axis cs:52,0.653396308422089)
--(axis cs:53,0.653396308422089)
--(axis cs:54,0.653396308422089)
--(axis cs:55,0.653396308422089)
--(axis cs:56,0.653396308422089)
--(axis cs:57,0.63684219121933)
--(axis cs:58,0.63684219121933)
--(axis cs:59,0.612308502197266)
--(axis cs:60,0.573007822036743)
--(axis cs:61,0.573007822036743)
--(axis cs:62,0.573007822036743)
--(axis cs:63,0.573007822036743)
--(axis cs:64,0.573007822036743)
--(axis cs:65,0.573007822036743)
--(axis cs:66,0.573007822036743)
--(axis cs:67,0.573007822036743)
--(axis cs:68,0.573007822036743)
--(axis cs:69,0.573007822036743)
--(axis cs:70,0.540430307388306)
--(axis cs:71,0.50381338596344)
--(axis cs:72,0.50381338596344)
--(axis cs:73,0.50381338596344)
--(axis cs:74,0.50381338596344)
--(axis cs:75,0.503037750720978)
--(axis cs:76,0.503037750720978)
--(axis cs:77,0.503037750720978)
--(axis cs:78,0.470024585723877)
--(axis cs:79,0.444889068603516)
--(axis cs:80,0.444889068603516)
--(axis cs:81,0.444889068603516)
--(axis cs:82,0.444889068603516)
--(axis cs:83,0.413557022809982)
--(axis cs:84,0.413557022809982)
--(axis cs:85,0.379786342382431)
--(axis cs:86,0.379786342382431)
--(axis cs:87,0.379786342382431)
--(axis cs:88,0.379786342382431)
--(axis cs:89,0.379786342382431)
--(axis cs:90,0.379786342382431)
--(axis cs:91,0.379786342382431)
--(axis cs:92,0.379786342382431)
--(axis cs:93,0.379786342382431)
--(axis cs:94,0.379786342382431)
--(axis cs:95,0.357746243476868)
--(axis cs:96,0.357746243476868)
--(axis cs:97,0.357746243476868)
--(axis cs:98,0.357746243476868)
--(axis cs:99,0.357746243476868)
--(axis cs:100,0.357746243476868)
--(axis cs:101,0.357746243476868)
--(axis cs:102,0.357746243476868)
--(axis cs:103,0.356353104114532)
--(axis cs:104,0.356353104114532)
--(axis cs:105,0.353591084480286)
--(axis cs:106,0.353591084480286)
--(axis cs:107,0.353591084480286)
--(axis cs:108,0.353591084480286)
--(axis cs:109,0.353591084480286)
--(axis cs:110,0.351911097764969)
--(axis cs:111,0.351911097764969)
--(axis cs:111,0.697235226631165)
--(axis cs:111,0.697235226631165)
--(axis cs:110,0.697235226631165)
--(axis cs:109,0.747981309890747)
--(axis cs:108,0.747981309890747)
--(axis cs:107,0.747981309890747)
--(axis cs:106,0.747981309890747)
--(axis cs:105,0.747981309890747)
--(axis cs:104,0.750645339488983)
--(axis cs:103,0.750645339488983)
--(axis cs:102,0.804701089859009)
--(axis cs:101,0.804701089859009)
--(axis cs:100,0.804701089859009)
--(axis cs:99,0.804701089859009)
--(axis cs:98,0.804701089859009)
--(axis cs:97,0.804701089859009)
--(axis cs:96,0.804701089859009)
--(axis cs:95,0.804701089859009)
--(axis cs:94,0.824463963508606)
--(axis cs:93,0.824463963508606)
--(axis cs:92,0.824463963508606)
--(axis cs:91,0.824463963508606)
--(axis cs:90,0.824463963508606)
--(axis cs:89,0.824463963508606)
--(axis cs:88,0.824463963508606)
--(axis cs:87,0.824463963508606)
--(axis cs:86,0.824463963508606)
--(axis cs:85,0.824463963508606)
--(axis cs:84,0.857893705368042)
--(axis cs:83,0.857893705368042)
--(axis cs:82,0.900437116622925)
--(axis cs:81,0.900437116622925)
--(axis cs:80,0.900437116622925)
--(axis cs:79,0.900437116622925)
--(axis cs:78,0.917502164840698)
--(axis cs:77,0.937764227390289)
--(axis cs:76,0.937764227390289)
--(axis cs:75,0.937764227390289)
--(axis cs:74,0.938430547714233)
--(axis cs:73,0.938430547714233)
--(axis cs:72,0.938430547714233)
--(axis cs:71,0.938430547714233)
--(axis cs:70,0.958979964256287)
--(axis cs:69,0.990558743476868)
--(axis cs:68,0.990558743476868)
--(axis cs:67,0.990558743476868)
--(axis cs:66,0.990558743476868)
--(axis cs:65,0.990558743476868)
--(axis cs:64,0.990558743476868)
--(axis cs:63,0.990558743476868)
--(axis cs:62,0.990558743476868)
--(axis cs:61,0.990558743476868)
--(axis cs:60,0.990558743476868)
--(axis cs:59,1.0201553106308)
--(axis cs:58,1.04479849338531)
--(axis cs:57,1.04479849338531)
--(axis cs:56,1.07045555114746)
--(axis cs:55,1.07045555114746)
--(axis cs:54,1.07045555114746)
--(axis cs:53,1.07045555114746)
--(axis cs:52,1.07045555114746)
--(axis cs:51,1.09023177623749)
--(axis cs:50,1.09023177623749)
--(axis cs:49,1.09023177623749)
--(axis cs:48,1.09023177623749)
--(axis cs:47,1.09023177623749)
--(axis cs:46,1.09023177623749)
--(axis cs:45,1.09023177623749)
--(axis cs:44,1.09023177623749)
--(axis cs:43,1.09023177623749)
--(axis cs:42,1.09023177623749)
--(axis cs:41,1.10772347450256)
--(axis cs:40,1.10772347450256)
--(axis cs:39,1.10772347450256)
--(axis cs:38,1.10772347450256)
--(axis cs:37,1.10772347450256)
--(axis cs:36,1.10772347450256)
--(axis cs:35,1.10772347450256)
--(axis cs:34,1.26304578781128)
--(axis cs:33,1.26304578781128)
--(axis cs:32,1.26304578781128)
--(axis cs:31,1.30831933021545)
--(axis cs:30,1.30831933021545)
--(axis cs:29,1.42933654785156)
--(axis cs:28,1.42933654785156)
--(axis cs:27,1.49171280860901)
--(axis cs:26,1.72978520393372)
--(axis cs:25,1.89570140838623)
--(axis cs:24,1.89570140838623)
--(axis cs:23,1.89570140838623)
--(axis cs:22,2.01567673683167)
--(axis cs:21,2.01567673683167)
--(axis cs:20,2.12208008766174)
--(axis cs:19,2.23354005813599)
--(axis cs:18,2.23354005813599)
--(axis cs:17,2.55242300033569)
--(axis cs:16,2.6235773563385)
--(axis cs:15,3.05139708518982)
--(axis cs:14,3.15842127799988)
--(axis cs:13,3.45379447937012)
--(axis cs:12,4.11530637741089)
--(axis cs:11,4.1632285118103)
--(axis cs:10,5.34454870223999)
--cycle;

\path [draw=color1, fill=color1, opacity=0.3]
(axis cs:10,5.34454898518408)
--(axis cs:10,4.39542274790918)
--(axis cs:11,3.79801829122048)
--(axis cs:12,3.10696664092773)
--(axis cs:13,2.64762193809687)
--(axis cs:14,2.31898617193081)
--(axis cs:15,1.7597124123334)
--(axis cs:16,1.52327282774938)
--(axis cs:17,1.1214806164081)
--(axis cs:18,0.946946091436208)
--(axis cs:19,0.931903697305733)
--(axis cs:20,0.906054400155405)
--(axis cs:21,0.695256814183608)
--(axis cs:22,0.548791802576466)
--(axis cs:23,0.490233209991641)
--(axis cs:24,0.490233209991641)
--(axis cs:25,0.490233209991641)
--(axis cs:26,0.410050374598355)
--(axis cs:27,0.373687956133197)
--(axis cs:28,0.372047510702013)
--(axis cs:29,0.320057653262002)
--(axis cs:30,0.309158779950114)
--(axis cs:31,0.291164567880433)
--(axis cs:32,0.287103606838863)
--(axis cs:33,0.207639671016859)
--(axis cs:34,0.181547616100683)
--(axis cs:35,0.137184006832133)
--(axis cs:36,0.129090147651805)
--(axis cs:37,0.0800872065958552)
--(axis cs:38,0.0680583715138268)
--(axis cs:39,0.0583345527450068)
--(axis cs:40,0.0558991842226313)
--(axis cs:41,0.0558991842226313)
--(axis cs:42,0.0558991842226313)
--(axis cs:43,0.0558991842226313)
--(axis cs:44,0.0413559512744793)
--(axis cs:45,0.0413559512744793)
--(axis cs:46,0.0406896647494604)
--(axis cs:47,0.0406896647494604)
--(axis cs:48,0.0406896647494604)
--(axis cs:49,0.0406896647494604)
--(axis cs:50,0.0406896647494604)
--(axis cs:51,0.0406896647494604)
--(axis cs:52,0.0406896647494604)
--(axis cs:53,0.0406896647494604)
--(axis cs:54,0.0406896647494604)
--(axis cs:55,0.0355938420202196)
--(axis cs:56,0.0321831699030099)
--(axis cs:57,0.0321831699030099)
--(axis cs:58,0.0300843483870104)
--(axis cs:59,0.0300843483870104)
--(axis cs:60,0.0275216076559612)
--(axis cs:61,0.0275216076559612)
--(axis cs:62,0.0275216076559612)
--(axis cs:63,0.0275216076559612)
--(axis cs:64,0.0275216076559612)
--(axis cs:65,0.0275216076559612)
--(axis cs:66,0.0275216076559612)
--(axis cs:67,0.0275216076559612)
--(axis cs:68,0.0275216076559612)
--(axis cs:69,0.0275216076559612)
--(axis cs:70,0.0275216076559612)
--(axis cs:71,0.0275216076559612)
--(axis cs:72,0.0275216076559612)
--(axis cs:73,0.0275216076559612)
--(axis cs:74,0.0275216076559612)
--(axis cs:75,0.0275216076559612)
--(axis cs:76,0.0275216076559612)
--(axis cs:77,0.0275216076559612)
--(axis cs:78,0.0254744741161235)
--(axis cs:79,0.0254744741161235)
--(axis cs:80,0.0254744741161235)
--(axis cs:81,0.0254744741161235)
--(axis cs:82,0.0254744741161235)
--(axis cs:83,0.0254744741161235)
--(axis cs:84,0.0254744741161235)
--(axis cs:85,0.0254744741161235)
--(axis cs:86,0.0254744741161235)
--(axis cs:87,0.0254744741161235)
--(axis cs:88,0.0254744741161235)
--(axis cs:89,0.0254744741161235)
--(axis cs:90,0.0254744741161235)
--(axis cs:91,0.0254744741161235)
--(axis cs:92,0.0254744741161235)
--(axis cs:93,0.0254744741161235)
--(axis cs:94,0.0254744741161235)
--(axis cs:95,0.0226335551013367)
--(axis cs:96,0.0226335551013367)
--(axis cs:97,0.0226335551013367)
--(axis cs:98,0.0224770405713156)
--(axis cs:99,0.0224770405713156)
--(axis cs:100,0.0224770405713156)
--(axis cs:101,0.0224770405713156)
--(axis cs:102,0.0224770405713156)
--(axis cs:103,0.0224770405713156)
--(axis cs:104,0.0224770405713156)
--(axis cs:105,0.0224770405713156)
--(axis cs:106,0.0224770405713156)
--(axis cs:107,0.0224770405713156)
--(axis cs:108,0.0224770405713156)
--(axis cs:109,0.0224770405713156)
--(axis cs:110,0.0224770405713156)
--(axis cs:111,0.0224770405713156)
--(axis cs:111,0.0281037607267119)
--(axis cs:111,0.0281037607267119)
--(axis cs:110,0.0281037607267119)
--(axis cs:109,0.0281037607267119)
--(axis cs:108,0.0281037607267119)
--(axis cs:107,0.0281037607267119)
--(axis cs:106,0.0281037607267119)
--(axis cs:105,0.0281037607267119)
--(axis cs:104,0.0281037607267119)
--(axis cs:103,0.0281037607267119)
--(axis cs:102,0.0281037607267119)
--(axis cs:101,0.0281037607267119)
--(axis cs:100,0.0281037607267119)
--(axis cs:99,0.0281037607267119)
--(axis cs:98,0.0281037607267119)
--(axis cs:97,0.028488835220956)
--(axis cs:96,0.028488835220956)
--(axis cs:95,0.028488835220956)
--(axis cs:94,0.0302936478063377)
--(axis cs:93,0.0302936478063377)
--(axis cs:92,0.0302936478063377)
--(axis cs:91,0.0302936478063377)
--(axis cs:90,0.0302936478063377)
--(axis cs:89,0.0302936478063377)
--(axis cs:88,0.0302936478063377)
--(axis cs:87,0.0302936478063377)
--(axis cs:86,0.0302936478063377)
--(axis cs:85,0.0302936478063377)
--(axis cs:84,0.0302936478063377)
--(axis cs:83,0.0302936478063377)
--(axis cs:82,0.0302936478063377)
--(axis cs:81,0.0302936478063377)
--(axis cs:80,0.0302936478063377)
--(axis cs:79,0.0302936478063377)
--(axis cs:78,0.0302936478063377)
--(axis cs:77,0.0375230359480721)
--(axis cs:76,0.0375230359480721)
--(axis cs:75,0.0375230359480721)
--(axis cs:74,0.0375230359480721)
--(axis cs:73,0.0375230359480721)
--(axis cs:72,0.0375230359480721)
--(axis cs:71,0.0375230359480721)
--(axis cs:70,0.0375230359480721)
--(axis cs:69,0.0375230359480721)
--(axis cs:68,0.0375230359480721)
--(axis cs:67,0.0375230359480721)
--(axis cs:66,0.0375230359480721)
--(axis cs:65,0.0375230359480721)
--(axis cs:64,0.0375230359480721)
--(axis cs:63,0.0375230359480721)
--(axis cs:62,0.0375230359480721)
--(axis cs:61,0.0375230359480721)
--(axis cs:60,0.0375230359480721)
--(axis cs:59,0.0415566773647441)
--(axis cs:58,0.0415566773647441)
--(axis cs:57,0.0451927331009955)
--(axis cs:56,0.0451927331009955)
--(axis cs:55,0.0484775125937898)
--(axis cs:54,0.0602537361802732)
--(axis cs:53,0.0602537361802732)
--(axis cs:52,0.0602537361802732)
--(axis cs:51,0.0602537361802732)
--(axis cs:50,0.0602537361802732)
--(axis cs:49,0.0602537361802732)
--(axis cs:48,0.0602537361802732)
--(axis cs:47,0.0602537361802732)
--(axis cs:46,0.0602537361802732)
--(axis cs:45,0.0606326979051477)
--(axis cs:44,0.0606326979051477)
--(axis cs:43,0.0964090944048002)
--(axis cs:42,0.0964090944048002)
--(axis cs:41,0.0964090944048002)
--(axis cs:40,0.0964090944048002)
--(axis cs:39,0.105741284298052)
--(axis cs:38,0.529219398862469)
--(axis cs:37,0.539512302936807)
--(axis cs:36,0.714275674460843)
--(axis cs:35,0.721309799074853)
--(axis cs:34,0.761974244673456)
--(axis cs:33,0.786132230164434)
--(axis cs:32,0.874540438042826)
--(axis cs:31,0.878485119807859)
--(axis cs:30,0.891762148947316)
--(axis cs:29,0.899237189764412)
--(axis cs:28,0.946295515846937)
--(axis cs:27,0.967413198444871)
--(axis cs:26,0.99974464609225)
--(axis cs:25,1.11383597889163)
--(axis cs:24,1.11383597889163)
--(axis cs:23,1.11383597889163)
--(axis cs:22,1.1626486668628)
--(axis cs:21,1.27644852263513)
--(axis cs:20,1.48453551765438)
--(axis cs:19,1.50385540699392)
--(axis cs:18,1.53905072244142)
--(axis cs:17,1.66397377101484)
--(axis cs:16,2.19465740149047)
--(axis cs:15,2.86275170165322)
--(axis cs:14,3.39214668581982)
--(axis cs:13,3.75751493692192)
--(axis cs:12,4.27205830737735)
--(axis cs:11,4.92516478553489)
--(axis cs:10,5.34454898518408)
--cycle;

\addplot [semithick, blue, dash dot]
table {%
10 4.86998558044434
11 4.24282217025757
12 3.44761157035828
13 3.14231491088867
14 2.65208101272583
15 2.62089347839355
16 2.33274793624878
17 2.19774556159973
18 2.01320576667786
19 1.89285123348236
20 1.7120361328125
21 1.57074391841888
22 1.52280497550964
23 1.18122410774231
24 1.18122410774231
25 1.11565554141998
26 0.87834358215332
27 0.821726620197296
28 0.821726620197296
29 0.745798110961914
30 0.745798110961914
31 0.725714087486267
32 0.537844479084015
33 0.537844479084015
34 0.529751181602478
35 0.529751181602478
36 0.521704077720642
37 0.521704077720642
38 0.502228558063507
39 0.486323177814484
40 0.41306534409523
41 0.41306534409523
42 0.404686361551285
43 0.367497622966766
44 0.367497622966766
45 0.295539855957031
46 0.263513952493668
47 0.263513952493668
48 0.263513952493668
49 0.263513952493668
50 0.263513952493668
51 0.263513952493668
52 0.255555331707001
53 0.255555331707001
54 0.255555331707001
55 0.255555331707001
56 0.255555331707001
57 0.255555331707001
58 0.255555331707001
59 0.231431573629379
60 0.231431573629379
61 0.231431573629379
62 0.231431573629379
63 0.231431573629379
64 0.231431573629379
65 0.231431573629379
66 0.208145141601562
67 0.208145141601562
68 0.208145141601562
69 0.208145141601562
70 0.208145141601562
71 0.208145141601562
72 0.208145141601562
73 0.202959448099136
74 0.201282501220703
75 0.17986373603344
76 0.17986373603344
77 0.17986373603344
78 0.17986373603344
79 0.17986373603344
80 0.17986373603344
81 0.17986373603344
82 0.14978714287281
83 0.14978714287281
84 0.14978714287281
85 0.14978714287281
86 0.14978714287281
87 0.14978714287281
88 0.14978714287281
89 0.14978714287281
90 0.14978714287281
91 0.14978714287281
92 0.14978714287281
93 0.14978714287281
94 0.14978714287281
95 0.14978714287281
96 0.14978714287281
97 0.14978714287281
98 0.14978714287281
99 0.14978714287281
100 0.14978714287281
101 0.14978714287281
102 0.143218994140625
103 0.102507017552853
104 0.102507017552853
105 0.102507017552853
106 0.102507017552853
107 0.102507017552853
108 0.102507017552853
109 0.102507017552853
110 0.102507017552853
111 0.102507017552853
};
\addplot [semithick, color1, dash dot]
table {%
10 4.86998586654663
11 3.77825940076693
12 2.95681844283239
13 2.24800814360809
14 2.04096005248083
15 1.86868676720175
16 1.73223655219365
17 1.58567868195341
18 1.50822783159412
19 1.43338305560799
20 1.33930617191809
21 1.25342162832169
22 1.23854315872744
23 1.17621843873076
24 1.07819618909463
25 0.999234401588761
26 0.931284314846277
27 0.931284314846277
28 0.84199649606673
29 0.728109004818256
30 0.728109004818256
31 0.728109004818256
32 0.619405406357308
33 0.619405406357308
34 0.619405406357308
35 0.551329919675039
36 0.50470324916693
37 0.50470324916693
38 0.340762091457152
39 0.340762091457152
40 0.340762091457152
41 0.340762091457152
42 0.340762091457152
43 0.270296445820815
44 0.270296445820815
45 0.270296445820815
46 0.270296445820815
47 0.270296445820815
48 0.270296445820815
49 0.270296445820815
50 0.270296445820815
51 0.270296445820815
52 0.270296445820815
53 0.270296445820815
54 0.270296445820815
55 0.2372920376655
56 0.2372920376655
57 0.2372920376655
58 0.234621911039437
59 0.234621911039437
60 0.220528029730266
61 0.220528029730266
62 0.220528029730266
63 0.220528029730266
64 0.220528029730266
65 0.220528029730266
66 0.220528029730266
67 0.220528029730266
68 0.220528029730266
69 0.220528029730266
70 0.220528029730266
71 0.220528029730266
72 0.220528029730266
73 0.216971404484382
74 0.168430635409002
75 0.168430635409002
76 0.162418039506941
77 0.162418039506941
78 0.162418039506941
79 0.162418039506941
80 0.162418039506941
81 0.162418039506941
82 0.162418039506941
83 0.162418039506941
84 0.162418039506941
85 0.162418039506941
86 0.154428388805531
87 0.154428388805531
88 0.154428388805531
89 0.139172475263306
90 0.115492538219662
91 0.115492538219662
92 0.115492538219662
93 0.115492538219662
94 0.115492538219662
95 0.115492538219662
96 0.115492538219662
97 0.115492538219662
98 0.115492538219662
99 0.115492538219662
100 0.115492538219662
101 0.115492538219662
102 0.115492538219662
103 0.115492538219662
104 0.115492538219662
105 0.115492538219662
106 0.115492538219662
107 0.115492538219662
108 0.115492538219662
109 0.115492538219662
110 0.115492538219662
111 0.115492538219662
};
\addplot [semithick, blue]
table {%
10 4.86998558044434
11 3.70798873901367
12 3.65983819961548
13 3.10068130493164
14 2.90851211547852
15 2.82170248031616
16 2.36114358901978
17 2.28117227554321
18 1.89728665351868
19 1.89728665351868
20 1.77787709236145
21 1.68967056274414
22 1.68967056274414
23 1.55154299736023
24 1.55154299736023
25 1.55154299736023
26 1.4357990026474
27 1.22413289546967
28 1.16590809822083
29 1.16590809822083
30 1.07365036010742
31 1.07365036010742
32 1.02298426628113
33 1.02298426628113
34 1.02298426628113
35 0.903463363647461
36 0.903463363647461
37 0.903463363647461
38 0.903463363647461
39 0.903463363647461
40 0.903463363647461
41 0.903463363647461
42 0.880580544471741
43 0.880580544471741
44 0.880580544471741
45 0.880580544471741
46 0.880580544471741
47 0.880580544471741
48 0.880580544471741
49 0.880580544471741
50 0.880580544471741
51 0.880580544471741
52 0.861925899982452
53 0.861925899982452
54 0.861925899982452
55 0.861925899982452
56 0.861925899982452
57 0.8408203125
58 0.8408203125
59 0.816231906414032
60 0.781783282756805
61 0.781783282756805
62 0.781783282756805
63 0.781783282756805
64 0.781783282756805
65 0.781783282756805
66 0.781783282756805
67 0.781783282756805
68 0.781783282756805
69 0.781783282756805
70 0.749705135822296
71 0.721121966838837
72 0.721121966838837
73 0.721121966838837
74 0.721121966838837
75 0.720400989055634
76 0.720400989055634
77 0.720400989055634
78 0.693763375282288
79 0.67266309261322
80 0.67266309261322
81 0.67266309261322
82 0.67266309261322
83 0.635725378990173
84 0.635725378990173
85 0.60212516784668
86 0.60212516784668
87 0.60212516784668
88 0.60212516784668
89 0.60212516784668
90 0.60212516784668
91 0.60212516784668
92 0.60212516784668
93 0.60212516784668
94 0.60212516784668
95 0.581223666667938
96 0.581223666667938
97 0.581223666667938
98 0.581223666667938
99 0.581223666667938
100 0.581223666667938
101 0.581223666667938
102 0.581223666667938
103 0.553499221801758
104 0.553499221801758
105 0.550786197185516
106 0.550786197185516
107 0.550786197185516
108 0.550786197185516
109 0.550786197185516
110 0.524573147296906
111 0.524573147296906
};
\addplot [semithick, color1]
table {%
10 4.86998586654663
11 4.36159153837768
12 3.68951247415254
13 3.20256843750939
14 2.85556642887532
15 2.31123205699331
16 1.85896511461993
17 1.39272719371147
18 1.24299840693882
19 1.21787955214983
20 1.19529495890489
21 0.985852668409371
22 0.855720234719631
23 0.802034594441634
24 0.802034594441634
25 0.802034594441634
26 0.704897510345302
27 0.670550577289034
28 0.659171513274475
29 0.609647421513207
30 0.600460464448715
31 0.584824843844146
32 0.580822022440844
33 0.496885950590647
34 0.47176093038707
35 0.429246902953493
36 0.421682911056324
37 0.309799754766331
38 0.298638885188148
39 0.0820379185215295
40 0.0761541393137158
41 0.0761541393137158
42 0.0761541393137158
43 0.0761541393137158
44 0.0509943245898135
45 0.0509943245898135
46 0.0504717004648668
47 0.0504717004648668
48 0.0504717004648668
49 0.0504717004648668
50 0.0504717004648668
51 0.0504717004648668
52 0.0504717004648668
53 0.0504717004648668
54 0.0504717004648668
55 0.0420356773070047
56 0.0386879515020027
57 0.0386879515020027
58 0.0358205128758772
59 0.0358205128758772
60 0.0325223218020167
61 0.0325223218020167
62 0.0325223218020167
63 0.0325223218020167
64 0.0325223218020167
65 0.0325223218020167
66 0.0325223218020167
67 0.0325223218020167
68 0.0325223218020167
69 0.0325223218020167
70 0.0325223218020167
71 0.0325223218020167
72 0.0325223218020167
73 0.0325223218020167
74 0.0325223218020167
75 0.0325223218020167
76 0.0325223218020167
77 0.0325223218020167
78 0.0278840609612306
79 0.0278840609612306
80 0.0278840609612306
81 0.0278840609612306
82 0.0278840609612306
83 0.0278840609612306
84 0.0278840609612306
85 0.0278840609612306
86 0.0278840609612306
87 0.0278840609612306
88 0.0278840609612306
89 0.0278840609612306
90 0.0278840609612306
91 0.0278840609612306
92 0.0278840609612306
93 0.0278840609612306
94 0.0278840609612306
95 0.0255611951611463
96 0.0255611951611463
97 0.0255611951611463
98 0.0252904006490137
99 0.0252904006490137
100 0.0252904006490137
101 0.0252904006490137
102 0.0252904006490137
103 0.0252904006490137
104 0.0252904006490137
105 0.0252904006490137
106 0.0252904006490137
107 0.0252904006490137
108 0.0252904006490137
109 0.0252904006490137
110 0.0252904006490137
111 0.0252904006490137
};
\end{axis}

\end{tikzpicture}

%% file: figures/bop_3d_hartmann3d_strong.tex
\begin{tikzpicture}

\definecolor{color0}{rgb}{0,0,1}
\definecolor{color1}{rgb}{1,0.549019607843137,0}
\definecolor{color2}{rgb}{1,0.647058823529412,0}
\definecolor{color3}{rgb}{0.564705882352941,0.933333333333333,0.564705882352941}

\begin{axis}[axis on top,
enlarge x limits=false,
enlarge y limits=false,
height=\figureheight,
scale only axis,
tick align=outside,
tick pos=left,
tick pos=left,
width=\figurewidth,
xlabel={Iteration},
xmin=10, xmax=60,
xtick style={color=black},
xtick={-10,0,10,25,50,75,100},
xticklabels={\ensuremath{-}10,0,10,25,50,75,90},
ymin=-0.05, ymax=1.3,
ytick style={color=black},
ytick={0.   , 1.3},
]
\node[anchor=north east] at (rel axis cs:1,1) {Hartmann 3D (strong)};
\path [draw=color1, fill=color1, opacity=0.3]
(axis cs:10,1.12330900415056)
--(axis cs:10,0.681536299601072)
--(axis cs:11,0.458576673586893)
--(axis cs:12,0.44787186238199)
--(axis cs:13,0.279389896984433)
--(axis cs:14,0.228644848002986)
--(axis cs:15,0.155639005933857)
--(axis cs:16,0.143398826511833)
--(axis cs:17,0.0810196449249413)
--(axis cs:18,0.0730697276083673)
--(axis cs:19,0.0724347187811829)
--(axis cs:20,0.0557470216122867)
--(axis cs:21,0.0459372520301701)
--(axis cs:22,0.0421851812607016)
--(axis cs:23,0.0368143251275772)
--(axis cs:24,0.0368143251275772)
--(axis cs:25,0.0368143251275772)
--(axis cs:26,0.0368143251275772)
--(axis cs:27,0.0355754966909672)
--(axis cs:28,0.035434572179604)
--(axis cs:29,0.035434572179604)
--(axis cs:30,0.0352602581623652)
--(axis cs:31,0.0352602581623652)
--(axis cs:32,0.0304300117703394)
--(axis cs:33,0.0295832055776615)
--(axis cs:34,0.0282262747191583)
--(axis cs:35,0.0282262747191583)
--(axis cs:36,0.0282262747191583)
--(axis cs:37,0.0282262747191583)
--(axis cs:38,0.0264218484027718)
--(axis cs:39,0.024322931093863)
--(axis cs:40,0.0241262692986071)
--(axis cs:41,0.0180684903267788)
--(axis cs:42,0.0180445478522457)
--(axis cs:43,0.0180445478522457)
--(axis cs:44,0.0180445478522457)
--(axis cs:45,0.0180445478522457)
--(axis cs:46,0.0180445478522457)
--(axis cs:47,0.0128966452709309)
--(axis cs:48,0.0128966452709309)
--(axis cs:49,0.0128966452709309)
--(axis cs:50,0.0103447472615269)
--(axis cs:51,0.0103447472615269)
--(axis cs:52,0.00833430329993882)
--(axis cs:53,0.00833430329993882)
--(axis cs:54,0.00833430329993882)
--(axis cs:55,0.00832259782179422)
--(axis cs:56,0.00832259782179422)
--(axis cs:57,0.00832259782179422)
--(axis cs:58,0.00795866396689235)
--(axis cs:59,0.00795866396689235)
--(axis cs:60,0.00795866396689235)
--(axis cs:61,0.00795866396689235)
--(axis cs:62,0.00795866396689235)
--(axis cs:63,0.00795866396689235)
--(axis cs:64,0.00795866396689235)
--(axis cs:65,0.00795866396689235)
--(axis cs:66,0.00795866396689235)
--(axis cs:67,0.00795866396689235)
--(axis cs:68,0.00795866396689235)
--(axis cs:69,0.00795866396689235)
--(axis cs:70,0.00795866396689235)
--(axis cs:71,0.00795866396689235)
--(axis cs:72,0.00795866396689235)
--(axis cs:73,0.00795866396689235)
--(axis cs:74,0.00795866396689235)
--(axis cs:75,0.00795866396689235)
--(axis cs:76,0.00795866396689235)
--(axis cs:77,0.00795866396689235)
--(axis cs:78,0.00795866396689235)
--(axis cs:79,0.00795866396689235)
--(axis cs:80,0.00795866396689235)
--(axis cs:81,0.00795866396689235)
--(axis cs:82,0.00795866396689235)
--(axis cs:83,0.00795866396689235)
--(axis cs:84,0.00795866396689235)
--(axis cs:85,0.00795866396689235)
--(axis cs:86,0.00795866396689235)
--(axis cs:87,0.00795866396689235)
--(axis cs:88,0.00795866396689235)
--(axis cs:89,0.00795866396689235)
--(axis cs:90,0.00795866396689235)
--(axis cs:91,0.00795866396689235)
--(axis cs:92,0.00795866396689235)
--(axis cs:93,0.00795866396689235)
--(axis cs:94,0.00795866396689235)
--(axis cs:95,0.00795866396689235)
--(axis cs:96,0.00795866396689235)
--(axis cs:97,0.00795866396689235)
--(axis cs:98,0.00795866396689235)
--(axis cs:99,0.00795866396689235)
--(axis cs:100,0.00795866396689235)
--(axis cs:101,0.00795866396689235)
--(axis cs:102,0.00795866396689235)
--(axis cs:103,0.00795866396689235)
--(axis cs:104,0.00795866396689235)
--(axis cs:105,0.00795866396689235)
--(axis cs:106,0.00795866396689235)
--(axis cs:107,0.00795866396689235)
--(axis cs:108,0.00795866396689235)
--(axis cs:109,0.00795866396689235)
--(axis cs:110,0.00795866396689235)
--(axis cs:111,0.00795866396689235)
--(axis cs:111,0.0120271579151833)
--(axis cs:111,0.0120271579151833)
--(axis cs:110,0.0120271579151833)
--(axis cs:109,0.0120271579151833)
--(axis cs:108,0.0120271579151833)
--(axis cs:107,0.0120271579151833)
--(axis cs:106,0.0120271579151833)
--(axis cs:105,0.0120271579151833)
--(axis cs:104,0.0120271579151833)
--(axis cs:103,0.0120271579151833)
--(axis cs:102,0.0120271579151833)
--(axis cs:101,0.0120271579151833)
--(axis cs:100,0.0120271579151833)
--(axis cs:99,0.0120271579151833)
--(axis cs:98,0.0120271579151833)
--(axis cs:97,0.0120271579151833)
--(axis cs:96,0.0120271579151833)
--(axis cs:95,0.0120271579151833)
--(axis cs:94,0.0120271579151833)
--(axis cs:93,0.0120271579151833)
--(axis cs:92,0.0120271579151833)
--(axis cs:91,0.0120271579151833)
--(axis cs:90,0.0120271579151833)
--(axis cs:89,0.0120271579151833)
--(axis cs:88,0.0120271579151833)
--(axis cs:87,0.0120271579151833)
--(axis cs:86,0.0120271579151833)
--(axis cs:85,0.0120271579151833)
--(axis cs:84,0.0120271579151833)
--(axis cs:83,0.0120271579151833)
--(axis cs:82,0.0120271579151833)
--(axis cs:81,0.0120271579151833)
--(axis cs:80,0.0120271579151833)
--(axis cs:79,0.0120271579151833)
--(axis cs:78,0.0120271579151833)
--(axis cs:77,0.0120271579151833)
--(axis cs:76,0.0120271579151833)
--(axis cs:75,0.0120271579151833)
--(axis cs:74,0.0120271579151833)
--(axis cs:73,0.0120271579151833)
--(axis cs:72,0.0120271579151833)
--(axis cs:71,0.0120271579151833)
--(axis cs:70,0.0120271579151833)
--(axis cs:69,0.0120271579151833)
--(axis cs:68,0.0120271579151833)
--(axis cs:67,0.0120271579151833)
--(axis cs:66,0.0120271579151833)
--(axis cs:65,0.0120271579151833)
--(axis cs:64,0.0120271579151833)
--(axis cs:63,0.0120271579151833)
--(axis cs:62,0.0120271579151833)
--(axis cs:61,0.0120271579151833)
--(axis cs:60,0.0120271579151833)
--(axis cs:59,0.0120271579151833)
--(axis cs:58,0.0120271579151833)
--(axis cs:57,0.0128078103421346)
--(axis cs:56,0.0128078103421346)
--(axis cs:55,0.0128078103421346)
--(axis cs:54,0.0128530216811463)
--(axis cs:53,0.0128530216811463)
--(axis cs:52,0.0128530216811463)
--(axis cs:51,0.0252450472797306)
--(axis cs:50,0.0252450472797306)
--(axis cs:49,0.0291072047791707)
--(axis cs:48,0.0291072047791707)
--(axis cs:47,0.0291072047791707)
--(axis cs:46,0.0348247185456746)
--(axis cs:45,0.0348247185456746)
--(axis cs:44,0.0348247185456746)
--(axis cs:43,0.0348247185456746)
--(axis cs:42,0.0348247185456746)
--(axis cs:41,0.0348412986644462)
--(axis cs:40,0.0564317761192478)
--(axis cs:39,0.0607950926297774)
--(axis cs:38,0.0677956682743171)
--(axis cs:37,0.0690059503413044)
--(axis cs:36,0.0690059503413044)
--(axis cs:35,0.0690059503413044)
--(axis cs:34,0.0690059503413044)
--(axis cs:33,0.0701309138826549)
--(axis cs:32,0.0708986408396472)
--(axis cs:31,0.0768323441977406)
--(axis cs:30,0.0768323441977406)
--(axis cs:29,0.0769886403420274)
--(axis cs:28,0.0769886403420274)
--(axis cs:27,0.0793384414524378)
--(axis cs:26,0.0801025954028124)
--(axis cs:25,0.0801025954028124)
--(axis cs:24,0.0801025954028124)
--(axis cs:23,0.0801025954028124)
--(axis cs:22,0.0907829867113053)
--(axis cs:21,0.0943848678899614)
--(axis cs:20,0.181178922442581)
--(axis cs:19,0.195379187002386)
--(axis cs:18,0.195845654176137)
--(axis cs:17,0.205608776256991)
--(axis cs:16,0.315165363391663)
--(axis cs:15,0.322319415830644)
--(axis cs:14,0.39455696078473)
--(axis cs:13,0.532336037943851)
--(axis cs:12,0.847829109855262)
--(axis cs:11,0.857049051507768)
--(axis cs:10,1.12330900415056)
--cycle;

\path [draw=blue, fill=blue, opacity=0.3]
(axis cs:10,1.12330901622772)
--(axis cs:10,0.681536316871643)
--(axis cs:11,0.480751037597656)
--(axis cs:12,0.45916610956192)
--(axis cs:13,0.447372615337372)
--(axis cs:14,0.374762773513794)
--(axis cs:15,0.286128133535385)
--(axis cs:16,0.272911339998245)
--(axis cs:17,0.255928695201874)
--(axis cs:18,0.195884168148041)
--(axis cs:19,0.18355156481266)
--(axis cs:20,0.154925405979156)
--(axis cs:21,0.145658627152443)
--(axis cs:22,0.145658627152443)
--(axis cs:23,0.116533555090427)
--(axis cs:24,0.116533555090427)
--(axis cs:25,0.116533555090427)
--(axis cs:26,0.0953144282102585)
--(axis cs:27,0.0953144282102585)
--(axis cs:28,0.0953144282102585)
--(axis cs:29,0.0919974595308304)
--(axis cs:30,0.0846218913793564)
--(axis cs:31,0.0738816112279892)
--(axis cs:32,0.0722583159804344)
--(axis cs:33,0.0644483417272568)
--(axis cs:34,0.0644483417272568)
--(axis cs:35,0.054282933473587)
--(axis cs:36,0.0498577952384949)
--(axis cs:37,0.0411083661019802)
--(axis cs:38,0.0359012223780155)
--(axis cs:39,0.0290218740701675)
--(axis cs:40,0.0290218740701675)
--(axis cs:41,0.0274904668331146)
--(axis cs:42,0.0272616222500801)
--(axis cs:43,0.023991797119379)
--(axis cs:44,0.023991797119379)
--(axis cs:45,0.023496687412262)
--(axis cs:46,0.023496687412262)
--(axis cs:47,0.023496687412262)
--(axis cs:48,0.023496687412262)
--(axis cs:49,0.019101407378912)
--(axis cs:50,0.019101407378912)
--(axis cs:51,0.019101407378912)
--(axis cs:52,0.019101407378912)
--(axis cs:53,0.0178216472268105)
--(axis cs:54,0.0157764218747616)
--(axis cs:55,0.012816809117794)
--(axis cs:56,0.012816809117794)
--(axis cs:57,0.012816809117794)
--(axis cs:58,0.012816809117794)
--(axis cs:59,0.012816809117794)
--(axis cs:60,0.012816809117794)
--(axis cs:61,0.0106518287211657)
--(axis cs:62,0.0106518287211657)
--(axis cs:63,0.00911596417427063)
--(axis cs:64,0.00911596417427063)
--(axis cs:65,0.00802942365407944)
--(axis cs:66,0.00802942365407944)
--(axis cs:67,0.00802942365407944)
--(axis cs:68,0.00578641425818205)
--(axis cs:69,0.00578641425818205)
--(axis cs:70,0.00578641425818205)
--(axis cs:71,0.00578641425818205)
--(axis cs:72,0.00535881798714399)
--(axis cs:73,0.00535881798714399)
--(axis cs:74,0.00535881798714399)
--(axis cs:75,0.00535881798714399)
--(axis cs:76,0.00535881798714399)
--(axis cs:77,0.00535881798714399)
--(axis cs:78,0.00535881798714399)
--(axis cs:79,0.00535881798714399)
--(axis cs:80,0.00535881798714399)
--(axis cs:81,0.00535881798714399)
--(axis cs:82,0.00535881798714399)
--(axis cs:83,0.00535881798714399)
--(axis cs:84,0.00535881798714399)
--(axis cs:85,0.00535881798714399)
--(axis cs:86,0.00535881798714399)
--(axis cs:87,0.00535881798714399)
--(axis cs:88,0.00535881798714399)
--(axis cs:89,0.00535881798714399)
--(axis cs:90,0.00535881798714399)
--(axis cs:91,0.00535881798714399)
--(axis cs:92,0.00535881798714399)
--(axis cs:93,0.00535881798714399)
--(axis cs:94,0.00535881798714399)
--(axis cs:95,0.00535881798714399)
--(axis cs:96,0.00535881798714399)
--(axis cs:97,0.00535881798714399)
--(axis cs:98,0.00535881798714399)
--(axis cs:99,0.00535881798714399)
--(axis cs:100,0.00535881798714399)
--(axis cs:101,0.00535881798714399)
--(axis cs:102,0.00535881798714399)
--(axis cs:103,0.00535881798714399)
--(axis cs:104,0.00535881798714399)
--(axis cs:105,0.00535881798714399)
--(axis cs:106,0.00535881798714399)
--(axis cs:107,0.00535881798714399)
--(axis cs:108,0.00535881798714399)
--(axis cs:109,0.00535881798714399)
--(axis cs:110,0.00535881798714399)
--(axis cs:111,0.00535881798714399)
--(axis cs:111,0.00994421076029539)
--(axis cs:111,0.00994421076029539)
--(axis cs:110,0.00994421076029539)
--(axis cs:109,0.00994421076029539)
--(axis cs:108,0.00994421076029539)
--(axis cs:107,0.00994421076029539)
--(axis cs:106,0.00994421076029539)
--(axis cs:105,0.00994421076029539)
--(axis cs:104,0.00994421076029539)
--(axis cs:103,0.00994421076029539)
--(axis cs:102,0.00994421076029539)
--(axis cs:101,0.00994421076029539)
--(axis cs:100,0.00994421076029539)
--(axis cs:99,0.00994421076029539)
--(axis cs:98,0.00994421076029539)
--(axis cs:97,0.00994421076029539)
--(axis cs:96,0.00994421076029539)
--(axis cs:95,0.00994421076029539)
--(axis cs:94,0.00994421076029539)
--(axis cs:93,0.00994421076029539)
--(axis cs:92,0.00994421076029539)
--(axis cs:91,0.00994421076029539)
--(axis cs:90,0.00994421076029539)
--(axis cs:89,0.00994421076029539)
--(axis cs:88,0.00994421076029539)
--(axis cs:87,0.00994421076029539)
--(axis cs:86,0.00994421076029539)
--(axis cs:85,0.00994421076029539)
--(axis cs:84,0.00994421076029539)
--(axis cs:83,0.00994421076029539)
--(axis cs:82,0.00994421076029539)
--(axis cs:81,0.00994421076029539)
--(axis cs:80,0.00994421076029539)
--(axis cs:79,0.00994421076029539)
--(axis cs:78,0.00994421076029539)
--(axis cs:77,0.00994421076029539)
--(axis cs:76,0.00994421076029539)
--(axis cs:75,0.00994421076029539)
--(axis cs:74,0.00994421076029539)
--(axis cs:73,0.00994421076029539)
--(axis cs:72,0.00994421076029539)
--(axis cs:71,0.0111009860411286)
--(axis cs:70,0.0111009860411286)
--(axis cs:69,0.0111009860411286)
--(axis cs:68,0.0111009860411286)
--(axis cs:67,0.0137663260102272)
--(axis cs:66,0.0137663260102272)
--(axis cs:65,0.0137663260102272)
--(axis cs:64,0.0176971182227135)
--(axis cs:63,0.0176971182227135)
--(axis cs:62,0.0192482452839613)
--(axis cs:61,0.0192482452839613)
--(axis cs:60,0.0220443829894066)
--(axis cs:59,0.0220443829894066)
--(axis cs:58,0.0220443829894066)
--(axis cs:57,0.0220443829894066)
--(axis cs:56,0.0220443829894066)
--(axis cs:55,0.0220443829894066)
--(axis cs:54,0.0236698277294636)
--(axis cs:53,0.0270531810820103)
--(axis cs:52,0.0304963327944279)
--(axis cs:51,0.0304963327944279)
--(axis cs:50,0.0304963327944279)
--(axis cs:49,0.0304963327944279)
--(axis cs:48,0.0354502946138382)
--(axis cs:47,0.0354502946138382)
--(axis cs:46,0.0354502946138382)
--(axis cs:45,0.0354502946138382)
--(axis cs:44,0.0369845479726791)
--(axis cs:43,0.0369845479726791)
--(axis cs:42,0.0502027496695518)
--(axis cs:41,0.0503165870904922)
--(axis cs:40,0.0518041998147964)
--(axis cs:39,0.0518041998147964)
--(axis cs:38,0.0654606372117996)
--(axis cs:37,0.0683722198009491)
--(axis cs:36,0.0837633162736893)
--(axis cs:35,0.086966410279274)
--(axis cs:34,0.109802797436714)
--(axis cs:33,0.109802797436714)
--(axis cs:32,0.114606373012066)
--(axis cs:31,0.115671813488007)
--(axis cs:30,0.128088280558586)
--(axis cs:29,0.141978219151497)
--(axis cs:28,0.143428608775139)
--(axis cs:27,0.143428608775139)
--(axis cs:26,0.143428608775139)
--(axis cs:25,0.172946780920029)
--(axis cs:24,0.172946780920029)
--(axis cs:23,0.172946780920029)
--(axis cs:22,0.241503432393074)
--(axis cs:21,0.241503432393074)
--(axis cs:20,0.255369454622269)
--(axis cs:19,0.277249723672867)
--(axis cs:18,0.311657905578613)
--(axis cs:17,0.419813871383667)
--(axis cs:16,0.445296853780746)
--(axis cs:15,0.452857941389084)
--(axis cs:14,0.777299046516418)
--(axis cs:13,0.835884988307953)
--(axis cs:12,0.844011127948761)
--(axis cs:11,0.858208298683167)
--(axis cs:10,1.12330901622772)
--cycle;

\path [draw=color1, fill=color1, opacity=0.3]
(axis cs:10,1.12330900415056)
--(axis cs:10,0.681536299601072)
--(axis cs:11,0.681536299601072)
--(axis cs:12,0.637797171807277)
--(axis cs:13,0.556970641149844)
--(axis cs:14,0.415992274880979)
--(axis cs:15,0.413689592166292)
--(axis cs:16,0.279987765496064)
--(axis cs:17,0.236046514828164)
--(axis cs:18,0.157005201764707)
--(axis cs:19,0.101166316580079)
--(axis cs:20,0.054663251054026)
--(axis cs:21,0.0512515687789123)
--(axis cs:22,0.04153603794081)
--(axis cs:23,0.0355575692075155)
--(axis cs:24,0.0207912350291185)
--(axis cs:25,0.0207912350291185)
--(axis cs:26,0.0189156516328703)
--(axis cs:27,0.0137856651940626)
--(axis cs:28,0.0123171916589982)
--(axis cs:29,0.0101066770096118)
--(axis cs:30,0.00915730450407119)
--(axis cs:31,0.00786575211580374)
--(axis cs:32,0.00711857421376805)
--(axis cs:33,0.00602291101220104)
--(axis cs:34,0.00569033040280043)
--(axis cs:35,0.00519913366175472)
--(axis cs:36,0.00440998614172064)
--(axis cs:37,0.00399395573869184)
--(axis cs:38,0.00243373748370406)
--(axis cs:39,0.00210899175564647)
--(axis cs:40,0.00210899175564647)
--(axis cs:41,0.0019872207635696)
--(axis cs:42,0.00165093171267166)
--(axis cs:43,0.00165093171267166)
--(axis cs:44,0.00165093171267166)
--(axis cs:45,0.00133008195685125)
--(axis cs:46,0.00133008195685125)
--(axis cs:47,0.00101232955685751)
--(axis cs:48,0.00101232955685751)
--(axis cs:49,0.00101232955685751)
--(axis cs:50,0.00101232955685751)
--(axis cs:51,0.000726150296716094)
--(axis cs:52,0.000726150296716094)
--(axis cs:53,0.000726150296716094)
--(axis cs:54,0.000726150296716094)
--(axis cs:55,0.000724292988953258)
--(axis cs:56,0.000724292988953258)
--(axis cs:57,0.000724292988953258)
--(axis cs:58,0.000724292988953258)
--(axis cs:59,0.000646235699219984)
--(axis cs:60,0.000569998905312407)
--(axis cs:61,0.000569998905312407)
--(axis cs:62,0.000569998905312407)
--(axis cs:63,0.000562835978286282)
--(axis cs:64,0.000562835978286282)
--(axis cs:65,0.000562835978286282)
--(axis cs:66,0.000562835978286282)
--(axis cs:67,0.000562835978286282)
--(axis cs:68,0.000562835978286282)
--(axis cs:69,0.000562835978286282)
--(axis cs:70,0.000562835978286282)
--(axis cs:71,0.000562835978286282)
--(axis cs:72,0.000562835978286282)
--(axis cs:73,0.000562835978286282)
--(axis cs:74,0.000562835978286282)
--(axis cs:75,0.000562835978286282)
--(axis cs:76,0.000562835978286282)
--(axis cs:77,0.000562835978286282)
--(axis cs:78,0.000562835978286282)
--(axis cs:79,0.00051501694866979)
--(axis cs:80,0.00051501694866979)
--(axis cs:81,0.00051501694866979)
--(axis cs:82,0.00051501694866979)
--(axis cs:83,0.00051501694866979)
--(axis cs:84,0.00051501694866979)
--(axis cs:85,0.00051501694866979)
--(axis cs:86,0.00051501694866979)
--(axis cs:87,0.00051501694866979)
--(axis cs:88,0.00051501694866979)
--(axis cs:89,0.00051501694866979)
--(axis cs:90,0.00051501694866979)
--(axis cs:91,0.00051501694866979)
--(axis cs:92,0.00051501694866979)
--(axis cs:93,0.00051501694866979)
--(axis cs:94,0.00051501694866979)
--(axis cs:95,0.00051501694866979)
--(axis cs:96,0.00051501694866979)
--(axis cs:97,0.00048021947548026)
--(axis cs:98,0.00048021947548026)
--(axis cs:99,0.00048021947548026)
--(axis cs:100,0.00048021947548026)
--(axis cs:101,0.00048021947548026)
--(axis cs:102,0.00048021947548026)
--(axis cs:103,0.00048021947548026)
--(axis cs:104,0.00048021947548026)
--(axis cs:105,0.00048021947548026)
--(axis cs:106,0.000479431444117781)
--(axis cs:107,0.000479431444117781)
--(axis cs:108,0.000479431444117781)
--(axis cs:109,0.000479431444117781)
--(axis cs:110,0.000461466394464433)
--(axis cs:111,0.000461466394464433)
--(axis cs:111,0.000747756398994154)
--(axis cs:111,0.000747756398994154)
--(axis cs:110,0.000747756398994154)
--(axis cs:109,0.000766844079627847)
--(axis cs:108,0.000766844079627847)
--(axis cs:107,0.000766844079627847)
--(axis cs:106,0.000766844079627847)
--(axis cs:105,0.000772076668937599)
--(axis cs:104,0.000772076668937599)
--(axis cs:103,0.000772076668937599)
--(axis cs:102,0.000772076668937599)
--(axis cs:101,0.000772076668937599)
--(axis cs:100,0.000772076668937599)
--(axis cs:99,0.000772076668937599)
--(axis cs:98,0.000772076668937599)
--(axis cs:97,0.000772076668937599)
--(axis cs:96,0.00078862983921396)
--(axis cs:95,0.00078862983921396)
--(axis cs:94,0.00078862983921396)
--(axis cs:93,0.00078862983921396)
--(axis cs:92,0.00078862983921396)
--(axis cs:91,0.00078862983921396)
--(axis cs:90,0.00078862983921396)
--(axis cs:89,0.00078862983921396)
--(axis cs:88,0.00078862983921396)
--(axis cs:87,0.00078862983921396)
--(axis cs:86,0.00078862983921396)
--(axis cs:85,0.00078862983921396)
--(axis cs:84,0.00078862983921396)
--(axis cs:83,0.00078862983921396)
--(axis cs:82,0.00078862983921396)
--(axis cs:81,0.00078862983921396)
--(axis cs:80,0.00078862983921396)
--(axis cs:79,0.00078862983921396)
--(axis cs:78,0.000946610829855544)
--(axis cs:77,0.000946610829855544)
--(axis cs:76,0.000946610829855544)
--(axis cs:75,0.000946610829855544)
--(axis cs:74,0.000946610829855544)
--(axis cs:73,0.000946610829855544)
--(axis cs:72,0.000946610829855544)
--(axis cs:71,0.000946610829855544)
--(axis cs:70,0.000946610829855544)
--(axis cs:69,0.000946610829855544)
--(axis cs:68,0.000946610829855544)
--(axis cs:67,0.000946610829855544)
--(axis cs:66,0.000946610829855544)
--(axis cs:65,0.000946610829855544)
--(axis cs:64,0.000946610829855544)
--(axis cs:63,0.000946610829855544)
--(axis cs:62,0.000949895943016707)
--(axis cs:61,0.000949895943016707)
--(axis cs:60,0.000949895943016707)
--(axis cs:59,0.00101429009996408)
--(axis cs:58,0.00114237194597233)
--(axis cs:57,0.00114237194597233)
--(axis cs:56,0.00114237194597233)
--(axis cs:55,0.00114237194597233)
--(axis cs:54,0.00114345478616333)
--(axis cs:53,0.00114345478616333)
--(axis cs:52,0.00114345478616333)
--(axis cs:51,0.00114345478616333)
--(axis cs:50,0.00172090186746289)
--(axis cs:49,0.00172090186746289)
--(axis cs:48,0.00172090186746289)
--(axis cs:47,0.00172090186746289)
--(axis cs:46,0.0022616453877287)
--(axis cs:45,0.0022616453877287)
--(axis cs:44,0.00257938946969272)
--(axis cs:43,0.00257938946969272)
--(axis cs:42,0.00257938946969272)
--(axis cs:41,0.00298377218019794)
--(axis cs:40,0.0031743118500152)
--(axis cs:39,0.0031743118500152)
--(axis cs:38,0.00468089463058971)
--(axis cs:37,0.00643091570138573)
--(axis cs:36,0.00686852332356768)
--(axis cs:35,0.00771586319240501)
--(axis cs:34,0.00816991128723181)
--(axis cs:33,0.0111201181456321)
--(axis cs:32,0.0137670711258765)
--(axis cs:31,0.014334227218588)
--(axis cs:30,0.0175137907900241)
--(axis cs:29,0.02133943173955)
--(axis cs:28,0.0324920695745396)
--(axis cs:27,0.0338363534738582)
--(axis cs:26,0.0394607810373302)
--(axis cs:25,0.0407156931025785)
--(axis cs:24,0.0407156931025785)
--(axis cs:23,0.0825961611055541)
--(axis cs:22,0.0953646836617726)
--(axis cs:21,0.107686197008657)
--(axis cs:20,0.140339626843101)
--(axis cs:19,0.212212780724229)
--(axis cs:18,0.257445890325946)
--(axis cs:17,0.399954240076709)
--(axis cs:16,0.443320762856379)
--(axis cs:15,0.583673467886278)
--(axis cs:14,0.593025917489945)
--(axis cs:13,0.90234232689452)
--(axis cs:12,0.972502327519405)
--(axis cs:11,1.12330900415056)
--(axis cs:10,1.12330900415056)
--cycle;

\path [draw=blue, fill=blue, opacity=0.3]
(axis cs:10,1.12330913543701)
--(axis cs:10,0.681536436080933)
--(axis cs:11,0.582499265670776)
--(axis cs:12,0.441318094730377)
--(axis cs:13,0.317734956741333)
--(axis cs:14,0.220179125666618)
--(axis cs:15,0.211940243840218)
--(axis cs:16,0.174286186695099)
--(axis cs:17,0.168571829795837)
--(axis cs:18,0.105511754751205)
--(axis cs:19,0.105511754751205)
--(axis cs:20,0.0853786319494247)
--(axis cs:21,0.080194003880024)
--(axis cs:22,0.0611120313405991)
--(axis cs:23,0.0495409891009331)
--(axis cs:24,0.0422049537301064)
--(axis cs:25,0.0422049537301064)
--(axis cs:26,0.0382485315203667)
--(axis cs:27,0.0382485315203667)
--(axis cs:28,0.0376374945044518)
--(axis cs:29,0.0300386250019073)
--(axis cs:30,0.0271887443959713)
--(axis cs:31,0.0271887443959713)
--(axis cs:32,0.0269179493188858)
--(axis cs:33,0.0269179493188858)
--(axis cs:34,0.0209533274173737)
--(axis cs:35,0.0209533274173737)
--(axis cs:36,0.0209533274173737)
--(axis cs:37,0.0202567595988512)
--(axis cs:38,0.0143252741545439)
--(axis cs:39,0.0115881226956844)
--(axis cs:40,0.0115881226956844)
--(axis cs:41,0.0115881226956844)
--(axis cs:42,0.0115881226956844)
--(axis cs:43,0.0115881226956844)
--(axis cs:44,0.0112699419260025)
--(axis cs:45,0.0112699419260025)
--(axis cs:46,0.0112699419260025)
--(axis cs:47,0.0112699419260025)
--(axis cs:48,0.0112699419260025)
--(axis cs:49,0.0112699419260025)
--(axis cs:50,0.0112699419260025)
--(axis cs:51,0.0112699419260025)
--(axis cs:52,0.00921052880585194)
--(axis cs:53,0.00721628358587623)
--(axis cs:54,0.00691604753956199)
--(axis cs:55,0.00691604753956199)
--(axis cs:56,0.00691604753956199)
--(axis cs:57,0.00691604753956199)
--(axis cs:58,0.00691604753956199)
--(axis cs:59,0.00691604753956199)
--(axis cs:60,0.00691604753956199)
--(axis cs:61,0.00691604753956199)
--(axis cs:62,0.00673503614962101)
--(axis cs:63,0.00673503614962101)
--(axis cs:64,0.00673503614962101)
--(axis cs:65,0.00673503614962101)
--(axis cs:66,0.00673503614962101)
--(axis cs:67,0.00673503614962101)
--(axis cs:68,0.00673503614962101)
--(axis cs:69,0.00634046504274011)
--(axis cs:70,0.00634046504274011)
--(axis cs:71,0.00634046504274011)
--(axis cs:72,0.00634046504274011)
--(axis cs:73,0.00634046504274011)
--(axis cs:74,0.00634046504274011)
--(axis cs:75,0.00634046504274011)
--(axis cs:76,0.00634046504274011)
--(axis cs:77,0.00634046504274011)
--(axis cs:78,0.00634046504274011)
--(axis cs:79,0.00634046504274011)
--(axis cs:80,0.0062227426096797)
--(axis cs:81,0.0062227426096797)
--(axis cs:82,0.0062227426096797)
--(axis cs:83,0.0062227426096797)
--(axis cs:84,0.0062227426096797)
--(axis cs:85,0.0062227426096797)
--(axis cs:86,0.0049559585750103)
--(axis cs:87,0.0049559585750103)
--(axis cs:88,0.0049559585750103)
--(axis cs:89,0.0049559585750103)
--(axis cs:90,0.0049559585750103)
--(axis cs:91,0.00469446275383234)
--(axis cs:92,0.00469446275383234)
--(axis cs:93,0.00469446275383234)
--(axis cs:94,0.00469446275383234)
--(axis cs:95,0.00469446275383234)
--(axis cs:96,0.00469446275383234)
--(axis cs:97,0.00469446275383234)
--(axis cs:98,0.00469446275383234)
--(axis cs:99,0.00469446275383234)
--(axis cs:100,0.00469446275383234)
--(axis cs:101,0.00469446275383234)
--(axis cs:102,0.00469446275383234)
--(axis cs:103,0.00469446275383234)
--(axis cs:104,0.00456151831895113)
--(axis cs:105,0.00456151831895113)
--(axis cs:106,0.00456151831895113)
--(axis cs:107,0.00456151831895113)
--(axis cs:108,0.00456151831895113)
--(axis cs:109,0.00456151831895113)
--(axis cs:110,0.00456151831895113)
--(axis cs:111,0.00456151831895113)
--(axis cs:111,0.00666132103651762)
--(axis cs:111,0.00666132103651762)
--(axis cs:110,0.00666132103651762)
--(axis cs:109,0.00666132103651762)
--(axis cs:108,0.00666132103651762)
--(axis cs:107,0.00666132103651762)
--(axis cs:106,0.00666132103651762)
--(axis cs:105,0.00666132103651762)
--(axis cs:104,0.00666132103651762)
--(axis cs:103,0.0067851273342967)
--(axis cs:102,0.0067851273342967)
--(axis cs:101,0.0067851273342967)
--(axis cs:100,0.0067851273342967)
--(axis cs:99,0.0067851273342967)
--(axis cs:98,0.0067851273342967)
--(axis cs:97,0.0067851273342967)
--(axis cs:96,0.0067851273342967)
--(axis cs:95,0.0067851273342967)
--(axis cs:94,0.0067851273342967)
--(axis cs:93,0.0067851273342967)
--(axis cs:92,0.0067851273342967)
--(axis cs:91,0.0067851273342967)
--(axis cs:90,0.00715554598718882)
--(axis cs:89,0.00715554598718882)
--(axis cs:88,0.00715554598718882)
--(axis cs:87,0.00715554598718882)
--(axis cs:86,0.00715554598718882)
--(axis cs:85,0.00854050647467375)
--(axis cs:84,0.00854050647467375)
--(axis cs:83,0.00854050647467375)
--(axis cs:82,0.00854050647467375)
--(axis cs:81,0.00854050647467375)
--(axis cs:80,0.00854050647467375)
--(axis cs:79,0.00874486099928617)
--(axis cs:78,0.00874486099928617)
--(axis cs:77,0.00874486099928617)
--(axis cs:76,0.00874486099928617)
--(axis cs:75,0.00874486099928617)
--(axis cs:74,0.00874486099928617)
--(axis cs:73,0.00874486099928617)
--(axis cs:72,0.00874486099928617)
--(axis cs:71,0.00874486099928617)
--(axis cs:70,0.00874486099928617)
--(axis cs:69,0.00874486099928617)
--(axis cs:68,0.00914703123271465)
--(axis cs:67,0.00914703123271465)
--(axis cs:66,0.00914703123271465)
--(axis cs:65,0.00914703123271465)
--(axis cs:64,0.00914703123271465)
--(axis cs:63,0.00914703123271465)
--(axis cs:62,0.00914703123271465)
--(axis cs:61,0.00987672712653875)
--(axis cs:60,0.00987672712653875)
--(axis cs:59,0.00987672712653875)
--(axis cs:58,0.00987672712653875)
--(axis cs:57,0.00987672712653875)
--(axis cs:56,0.00987672712653875)
--(axis cs:55,0.00987672712653875)
--(axis cs:54,0.00987672712653875)
--(axis cs:53,0.0151547435671091)
--(axis cs:52,0.0178768951445818)
--(axis cs:51,0.0202364921569824)
--(axis cs:50,0.0202364921569824)
--(axis cs:49,0.0202364921569824)
--(axis cs:48,0.0202364921569824)
--(axis cs:47,0.0202364921569824)
--(axis cs:46,0.0202364921569824)
--(axis cs:45,0.0202364921569824)
--(axis cs:44,0.0202364921569824)
--(axis cs:43,0.0204763673245907)
--(axis cs:42,0.0204763673245907)
--(axis cs:41,0.0204763673245907)
--(axis cs:40,0.0204763673245907)
--(axis cs:39,0.0204763673245907)
--(axis cs:38,0.0233271662145853)
--(axis cs:37,0.0333635285496712)
--(axis cs:36,0.0337523967027664)
--(axis cs:35,0.0337523967027664)
--(axis cs:34,0.0337523967027664)
--(axis cs:33,0.0433370620012283)
--(axis cs:32,0.0433370620012283)
--(axis cs:31,0.043716412037611)
--(axis cs:30,0.043716412037611)
--(axis cs:29,0.0468892455101013)
--(axis cs:28,0.0729753226041794)
--(axis cs:27,0.0740221440792084)
--(axis cs:26,0.0740221440792084)
--(axis cs:25,0.0784702375531197)
--(axis cs:24,0.0784702375531197)
--(axis cs:23,0.0920289978384972)
--(axis cs:22,0.111983940005302)
--(axis cs:21,0.129944384098053)
--(axis cs:20,0.152615398168564)
--(axis cs:19,0.203153520822525)
--(axis cs:18,0.203153520822525)
--(axis cs:17,0.320770651102066)
--(axis cs:16,0.331688225269318)
--(axis cs:15,0.47860711812973)
--(axis cs:14,0.487066566944122)
--(axis cs:13,0.57402515411377)
--(axis cs:12,0.688609540462494)
--(axis cs:11,1.03677201271057)
--(axis cs:10,1.12330913543701)
--cycle;

\addplot [semithick, color1, dash dot]
table {%
10 0.902422651875814
11 0.657812862547331
12 0.647850486118626
13 0.405862967464142
14 0.311600904393858
15 0.23897921088225
16 0.229282094951748
17 0.143314210590966
18 0.134457690892252
19 0.133906952891784
20 0.118462972027434
21 0.0701610599600657
22 0.0664840839860035
23 0.0584584602651948
24 0.0584584602651948
25 0.0584584602651948
26 0.0584584602651948
27 0.0574569690717025
28 0.0562116062608157
29 0.0562116062608157
30 0.0560463011800529
31 0.0560463011800529
32 0.0506643263049933
33 0.0498570597301582
34 0.0486161125302313
35 0.0486161125302313
36 0.0486161125302313
37 0.0486161125302313
38 0.0471087583385445
39 0.0425590118618202
40 0.0402790227089274
41 0.0264548944956125
42 0.0264346331989601
43 0.0264346331989601
44 0.0264346331989601
45 0.0264346331989601
46 0.0264346331989601
47 0.0210019250250508
48 0.0210019250250508
49 0.0210019250250508
50 0.0177948972706287
51 0.0177948972706287
52 0.0105936624905425
53 0.0105936624905425
54 0.0105936624905425
55 0.0105652040819644
56 0.0105652040819644
57 0.0105652040819644
58 0.00999291094103781
59 0.00999291094103781
60 0.00999291094103781
61 0.00999291094103781
62 0.00999291094103781
63 0.00999291094103781
64 0.00999291094103781
65 0.00999291094103781
66 0.00999291094103781
67 0.00999291094103781
68 0.00999291094103781
69 0.00999291094103781
70 0.00999291094103781
71 0.00999291094103781
72 0.00999291094103781
73 0.00999291094103781
74 0.00999291094103781
75 0.00999291094103781
76 0.00999291094103781
77 0.00999291094103781
78 0.00999291094103781
79 0.00999291094103781
80 0.00999291094103781
81 0.00999291094103781
82 0.00999291094103781
83 0.00999291094103781
84 0.00999291094103781
85 0.00999291094103781
86 0.00999291094103781
87 0.00999291094103781
88 0.00999291094103781
89 0.00999291094103781
90 0.00999291094103781
91 0.00999291094103781
92 0.00999291094103781
93 0.00999291094103781
94 0.00999291094103781
95 0.00999291094103781
96 0.00999291094103781
97 0.00999291094103781
98 0.00999291094103781
99 0.00999291094103781
100 0.00999291094103781
101 0.00999291094103781
102 0.00999291094103781
103 0.00999291094103781
104 0.00999291094103781
105 0.00999291094103781
106 0.00999291094103781
107 0.00999291094103781
108 0.00999291094103781
109 0.00999291094103781
110 0.00999291094103781
111 0.00999291094103781
};
\addplot [semithick, blue]
table {%
10 0.902422666549683
11 0.669479668140411
12 0.651588618755341
13 0.641628801822662
14 0.576030910015106
15 0.369493037462234
16 0.359104096889496
17 0.33787128329277
18 0.253771036863327
19 0.230400636792183
20 0.205147430300713
21 0.193581029772758
22 0.193581029772758
23 0.144740164279938
24 0.144740164279938
25 0.144740164279938
26 0.119371518492699
27 0.119371518492699
28 0.119371518492699
29 0.116987839341164
30 0.106355085968971
31 0.0947767123579979
32 0.0934323444962502
33 0.0871255695819855
34 0.0871255695819855
35 0.0706246718764305
36 0.0668105557560921
37 0.0547402948141098
38 0.0506809279322624
39 0.040413036942482
40 0.040413036942482
41 0.0389035269618034
42 0.038732185959816
43 0.0304881725460291
44 0.0304881725460291
45 0.0294734910130501
46 0.0294734910130501
47 0.0294734910130501
48 0.0294734910130501
49 0.0247988700866699
50 0.0247988700866699
51 0.0247988700866699
52 0.0247988700866699
53 0.0224374141544104
54 0.0197231248021126
55 0.0174305960536003
56 0.0174305960536003
57 0.0174305960536003
58 0.0174305960536003
59 0.0174305960536003
60 0.0174305960536003
61 0.0149500370025635
62 0.0149500370025635
63 0.0134065411984921
64 0.0134065411984921
65 0.0108978748321533
66 0.0108978748321533
67 0.0108978748321533
68 0.00844370014965534
69 0.00844370014965534
70 0.00844370014965534
71 0.00844370014965534
72 0.00765151437371969
73 0.00765151437371969
74 0.00765151437371969
75 0.00765151437371969
76 0.00765151437371969
77 0.00765151437371969
78 0.00765151437371969
79 0.00765151437371969
80 0.00765151437371969
81 0.00765151437371969
82 0.00765151437371969
83 0.00765151437371969
84 0.00765151437371969
85 0.00765151437371969
86 0.00765151437371969
87 0.00765151437371969
88 0.00765151437371969
89 0.00765151437371969
90 0.00765151437371969
91 0.00765151437371969
92 0.00765151437371969
93 0.00765151437371969
94 0.00765151437371969
95 0.00765151437371969
96 0.00765151437371969
97 0.00765151437371969
98 0.00765151437371969
99 0.00765151437371969
100 0.00765151437371969
101 0.00765151437371969
102 0.00765151437371969
103 0.00765151437371969
104 0.00765151437371969
105 0.00765151437371969
106 0.00765151437371969
107 0.00765151437371969
108 0.00765151437371969
109 0.00765151437371969
110 0.00765151437371969
111 0.00765151437371969
};
\addplot [semithick, color1]
table {%
10 0.902422651875814
11 0.902422651875814
12 0.805149749663341
13 0.729656484022182
14 0.504509096185462
15 0.498681530026285
16 0.361654264176221
17 0.318000377452436
18 0.207225546045327
19 0.156689548652154
20 0.0975014389485638
21 0.0794688828937845
22 0.0684503608012913
23 0.0590768651565348
24 0.0307534640658485
25 0.0307534640658485
26 0.0291882163351002
27 0.0238110093339604
28 0.0224046306167689
29 0.0157230543745809
30 0.0133355476470477
31 0.0110999896671959
32 0.0104428226698223
33 0.0085715145789166
34 0.00693012084501612
35 0.00645749842707987
36 0.00563925473264416
37 0.00521243572003878
38 0.00355731605714688
39 0.00264165180283084
40 0.00264165180283084
41 0.00248549647188377
42 0.00211516059118219
43 0.00211516059118219
44 0.00211516059118219
45 0.00179586367228997
46 0.00179586367228997
47 0.0013666157121602
48 0.0013666157121602
49 0.0013666157121602
50 0.0013666157121602
51 0.000934802541439714
52 0.000934802541439714
53 0.000934802541439714
54 0.000934802541439714
55 0.000933332467462793
56 0.000933332467462793
57 0.000933332467462793
58 0.000933332467462793
59 0.000830262899592032
60 0.000759947424164557
61 0.000759947424164557
62 0.000759947424164557
63 0.000754723404070913
64 0.000754723404070913
65 0.000754723404070913
66 0.000754723404070913
67 0.000754723404070913
68 0.000754723404070913
69 0.000754723404070913
70 0.000754723404070913
71 0.000754723404070913
72 0.000754723404070913
73 0.000754723404070913
74 0.000754723404070913
75 0.000754723404070913
76 0.000754723404070913
77 0.000754723404070913
78 0.000754723404070913
79 0.000651823393941875
80 0.000651823393941875
81 0.000651823393941875
82 0.000651823393941875
83 0.000651823393941875
84 0.000651823393941875
85 0.000651823393941875
86 0.000651823393941875
87 0.000651823393941875
88 0.000651823393941875
89 0.000651823393941875
90 0.000651823393941875
91 0.000651823393941875
92 0.000651823393941875
93 0.000651823393941875
94 0.000651823393941875
95 0.000651823393941875
96 0.000651823393941875
97 0.000626148072208929
98 0.000626148072208929
99 0.000626148072208929
100 0.000626148072208929
101 0.000626148072208929
102 0.000626148072208929
103 0.000626148072208929
104 0.000626148072208929
105 0.000626148072208929
106 0.000623137761872814
107 0.000623137761872814
108 0.000623137761872814
109 0.000623137761872814
110 0.000604611396729293
111 0.000604611396729293
};
\addplot [semithick, blue, dash dot]
table {%
10 0.902422785758972
11 0.809635639190674
12 0.564963817596436
13 0.445880055427551
14 0.353622853755951
15 0.345273673534393
16 0.252987205982208
17 0.244671240448952
18 0.154332637786865
19 0.154332637786865
20 0.118997015058994
21 0.105069190263748
22 0.0865479856729507
23 0.0707849934697151
24 0.060337595641613
25 0.060337595641613
26 0.0561353377997875
27 0.0561353377997875
28 0.0553064085543156
29 0.0384639352560043
30 0.0354525782167912
31 0.0354525782167912
32 0.0351275056600571
33 0.0351275056600571
34 0.02735286206007
35 0.02735286206007
36 0.02735286206007
37 0.0268101431429386
38 0.0188262201845646
39 0.0160322450101376
40 0.0160322450101376
41 0.0160322450101376
42 0.0160322450101376
43 0.0160322450101376
44 0.0157532170414925
45 0.0157532170414925
46 0.0157532170414925
47 0.0157532170414925
48 0.0157532170414925
49 0.0157532170414925
50 0.0157532170414925
51 0.0157532170414925
52 0.0135437119752169
53 0.0111855138093233
54 0.00839638710021973
55 0.00839638710021973
56 0.00839638710021973
57 0.00839638710021973
58 0.00839638710021973
59 0.00839638710021973
60 0.00839638710021973
61 0.00839638710021973
62 0.00794103369116783
63 0.00794103369116783
64 0.00794103369116783
65 0.00794103369116783
66 0.00794103369116783
67 0.00794103369116783
68 0.00794103369116783
69 0.00754266325384378
70 0.00754266325384378
71 0.00754266325384378
72 0.00754266325384378
73 0.00754266325384378
74 0.00754266325384378
75 0.00754266325384378
76 0.00754266325384378
77 0.00754266325384378
78 0.00754266325384378
79 0.00754266325384378
80 0.00738162454217672
81 0.00738162454217672
82 0.00738162454217672
83 0.00738162454217672
84 0.00738162454217672
85 0.00738162454217672
86 0.00605575228109956
87 0.00605575228109956
88 0.00605575228109956
89 0.00605575228109956
90 0.00605575228109956
91 0.00573979504406452
92 0.00573979504406452
93 0.00573979504406452
94 0.00573979504406452
95 0.00573979504406452
96 0.00573979504406452
97 0.00573979504406452
98 0.00573979504406452
99 0.00573979504406452
100 0.00573979504406452
101 0.00573979504406452
102 0.00573979504406452
103 0.00573979504406452
104 0.00561141967773438
105 0.00561141967773438
106 0.00561141967773438
107 0.00561141967773438
108 0.00561141967773438
109 0.00561141967773438
110 0.00561141967773438
111 0.00561141967773438
};
\end{axis}

\end{tikzpicture}

%% file: figures/ppd_dataset_1.tex
\begin{tikzpicture}

\definecolor{darkgray176}{RGB}{176,176,176}
\definecolor{lightgray204}{RGB}{204,204,204}
\definecolor{steelblue31119180}{RGB}{31,119,180}
\Large
\begin{groupplot}[group style={group size=3 by 1}]
\nextgroupplot[
tick align=outside,
tick pos=left,
title={ACE},
x grid style={darkgray176},
xlabel={Time},
xmin=-0.65, xmax=13.65,
xtick style={color=black},
xtick = {0, 5, 10},
y grid style={darkgray176},
ylabel={Count},
ymin=-10, ymax=420,
ytick style={color=black}
]
\addplot [semithick, steelblue31119180]
table {%
0 3.35500001907349
1 10.0514001846313
2 29.7569999694824
3 81.8942031860352
4 187.819198608398
5 300.008605957031
6 316.071594238281
7 246.874206542969
8 164.564407348633
9 101.247596740723
10 59.771598815918
11 35.0097999572754
12 20.0891990661621
13 11.5928001403809
};

\path [fill=steelblue31119180, fill opacity=0.25]
(axis cs:0,8)
--(axis cs:0,0)
--(axis cs:1,4)
--(axis cs:2,18)
--(axis cs:3,61)
--(axis cs:4,152)
--(axis cs:5,255)
--(axis cs:6,274)
--(axis cs:7,208)
--(axis cs:8,135)
--(axis cs:9,79)
--(axis cs:10,42)
--(axis cs:11,22)
--(axis cs:12,10)
--(axis cs:13,4)
--(axis cs:13,20)
--(axis cs:13,20)
--(axis cs:12,31.0249999999996)
--(axis cs:11,50)
--(axis cs:10,79)
--(axis cs:9,125)
--(axis cs:8,196)
--(axis cs:7,287)
--(axis cs:6,362)
--(axis cs:5,346)
--(axis cs:4,226)
--(axis cs:3,104)
--(axis cs:2,44)
--(axis cs:1,18)
--(axis cs:0,8)
--cycle;

\addplot [draw=black, fill=black, mark=*, only marks]
table{%
x  y
0 3
1 8
2 26
3 76
4 225
5 298
6 258
7 233
8 189
9 128
10 68
11 29
12 14
13 4
};

\draw (axis cs:0.05,410) node[
  anchor=north west,
  text=black,
  rotate=0.0
]{log-prob $\uparrow$ -64.4};

\nextgroupplot[
scaled y ticks=manual:{}{\pgfmathparse{#1}},
tick align=outside,
tick pos=left,
title={NPE},
x grid style={darkgray176},
xlabel={Time},
xmin=-0.65, xmax=13.65,
xtick style={color=black},
xtick = {0, 5, 10},
y grid style={darkgray176},
ymin=-10, ymax=420,
ytick style={color=black},
yticklabels={}
]
\addplot [semithick, steelblue31119180]
table {%
0 2.93700003623962
1 9.17000007629395
2 28.1522006988525
3 81.0818023681641
4 192.024398803711
5 308.960998535156
6 320.503814697266
7 246.366592407227
8 162.420593261719
9 99.0968017578125
10 58.3871994018555
11 34.0989990234375
12 19.6580009460449
13 11.2936000823975
};

\path [fill=steelblue31119180, fill opacity=0.25]
(axis cs:0,7)
--(axis cs:0,0)
--(axis cs:1,3)
--(axis cs:2,16)
--(axis cs:3,61)
--(axis cs:4,157)
--(axis cs:5,263.975)
--(axis cs:6,278)
--(axis cs:7,205)
--(axis cs:8,132)
--(axis cs:9,78)
--(axis cs:10,42)
--(axis cs:11,22)
--(axis cs:12,11)
--(axis cs:13,5)
--(axis cs:13,20)
--(axis cs:13,20)
--(axis cs:12,31)
--(axis cs:11,48)
--(axis cs:10,76)
--(axis cs:9,122)
--(axis cs:8,193)
--(axis cs:7,289)
--(axis cs:6,364)
--(axis cs:5,357)
--(axis cs:4,232)
--(axis cs:3,103)
--(axis cs:2,42)
--(axis cs:1,17)
--(axis cs:0,7)
--cycle;

\addplot [draw=black, fill=black, mark=*, only marks]
table{%
x  y
0 3
1 8
2 26
3 76
4 225
5 298
6 258
7 233
8 189
9 128
10 68
11 29
12 14
13 4
};

\draw (axis cs:0.05,410) node[
  anchor=north west,
  text=black,
  rotate=0.0
]{log-prob $\uparrow$ -64.6};

\nextgroupplot[
legend cell align={left},
legend style={
  fill opacity=0.8,
  draw opacity=1,
  text opacity=1,
  at={(1.65,0.5)},
  anchor=east,
  draw=lightgray204
},
scaled y ticks=manual:{}{\pgfmathparse{#1}},
tick align=outside,
tick pos=left,
title={MCMC},
x grid style={darkgray176},
xlabel={Time},
xmin=-0.65, xmax=13.65,
xtick style={color=black},
xtick = {0, 5, 10},
y grid style={darkgray176},
ymin=-10, ymax=420,
ytick style={color=black},
yticklabels={}
]
\addplot [semithick, steelblue31119180]
table {%
0 3.66820001602173
1 10.9191999435425
2 31.4109992980957
3 84.7503967285156
4 188.496795654297
5 295.789794921875
6 310.153198242188
7 243.214202880859
8 163.110397338867
9 100.798599243164
10 60.4127998352051
11 35.1990013122559
12 20.4108009338379
13 11.8531999588013
};
\addlegendentry{PPD mean}

\path [fill=steelblue31119180, fill opacity=0.25]
(axis cs:0,8)
--(axis cs:0,0)
--(axis cs:1,4)
--(axis cs:2,20)
--(axis cs:3,65)
--(axis cs:4,156)
--(axis cs:5,255)
--(axis cs:6,271)
--(axis cs:7,206.975)
--(axis cs:8,136)
--(axis cs:9,80)
--(axis cs:10,44)
--(axis cs:11,23)
--(axis cs:12,11)
--(axis cs:13,5)
--(axis cs:13,20)
--(axis cs:13,20)
--(axis cs:12,31)
--(axis cs:11,49)
--(axis cs:10,78)
--(axis cs:9,123)
--(axis cs:8,192)
--(axis cs:7,280)
--(axis cs:6,350)
--(axis cs:5,338)
--(axis cs:4,223)
--(axis cs:3,107)
--(axis cs:2,45)
--(axis cs:1,19)
--(axis cs:0,8)
--cycle;
\addlegendimage{area legend, fill=steelblue31119180, fill opacity=0.25}
\addlegendentry{PPD 95\% CI}

\addplot [draw=black, fill=black, mark=*, only marks]
table{%
x  y
0 3
1 8
2 26
3 76
4 225
5 298
6 258
7 233
8 189
9 128
10 68
11 29
12 14
13 4
};
\addlegendentry{observed}

\draw (axis cs:0.05,410) node[
  anchor=north west,
  text=black,
  rotate=0.0
]{log-prob $\uparrow$ -62.9};
\end{groupplot}

\end{tikzpicture}